# Recursive Distributed Collaborative Aided Inertial Navigation

## D I S S E R T A T I O N

submitted in fulfilment of the requirements for the degree of

## Doktor der Technischen Wissenschaften

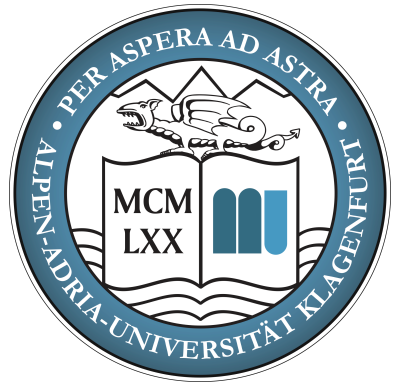

## University Klagenfurt

Faculty of Technical Sciences

by
Roland Jung, MSc.

Supervisors

Univ.-Prof. Dr. Stephan Weiss
University Klagenfurt
Department of Smart System Technologies

Univ.-Prof. Dipl.-Ing. Dr. Christian Bettstetter
University Klagenfurt
Dept. of Networked and Embedded Systems

Expert Reviewers

Assoc. Prof. James R. Forbes
McGill University
Department of Mechanical Engineering

Univ.-Prof. Dipl.-Ing.Dr. Wilfried Elmenreich
University Klagenfurt
Dept. of Networked and Embedded Systems

Klagenfurt, August 2023

# Affidavit

I hereby declare in lieu of an oath that

- the submitted academic paper is entirely my own work and that no auxiliary materials have been used other than those indicated,
- I have fully disclosed all assistance received from third parties during the process of writing the thesis, including any significant advice from supervisors,
- any contents taken from the works of third parties or my own works that have been included either literally or in spirit have been appropriately marked and the respective source of the information has been clearly identified with precise bibliographical references (e.g., in footnotes),
- have fully and truthfully declared the use of generative models (Artificial Intelligence, e.g., ChatGPT, Grammarly Go, Midjourney) including the product version,
- to date, I have not submitted this paper to an examining authority either in Austria or abroad and that
- when passing on copies of the academic thesis (e.g., in bound, printed or digital form), I will ensure that each copy is fully consistent with the submitted digital version.

I am aware that a declaration contrary to the facts will have legal consequences.

Roland Jung, MSc. *m.p.*          University Klagenfurt, August 24, 2023

# Contents

# Preface

I owe gratitude to all people who made it possible to fulfill this important milestone. Especially to my mentor and supervisor, Prof. Stephan Weiss, with his farsightedness, spirit for hard work, and gift to regard problems in science and every-day-life in an objective and disillusioning way. He pushed me beyond my limits and inspired me to dive into filter-based localization approaches. Besides, Prof. Weiss, I am grateful to my co-supervisor Prof. Christian Bettstetter and the other supervisors of the Karl Popper Kolleg on Networked Autonomous Aerial Vehicles (KPK-NAV): Prof. Hermann Hellwagner and Prof. Bernhard Rinner. Their valuable inputs, fruitful discussions, and the provided insights in the indivudals' domains was helpful to obtain a clear vision and to find solutions to our project goals. In total four doctoral students were supervises in the KPK-NAV. The other three (already graduated) PhD candidates involved were: Agata and Michał Barciś, and Petra Mazdin. Besides spending a lot of time in the office, we share great memories from our activities in our leisure time.

After the KPK-NAV, I got the opportunity to move to the Prof. Weiss's Control of Networked Systems (CNS) group, where I am currently sharing a office with Jan Michalczyk and Rohit Dhakate, and would like to thank the other fellow graduate students and Prof. Jan Steinbrener, with all whom I shared great time during discussions, paper writing, our retreats, and social activities.

At this point, I want to dedicate special thanks to Dr. Lukas Luft for our discussions regarding the consistency and properties of the IKF pardigm, to my Dr. Stefan Brennsteiner for editorial comments, and to my college Christoph Böhm for our discussion on the nonlinear observability analysis for multi-agent systems and the interpretation of the results, and all other co-others for their valuable inputs.

Special thanks goes to my former adviser Dr. Martin Humenberger, other colleagues and friends from the Austrian Institute of Technology (AIT), who encouraged me to start a PhD at the University Klagenfurt, to quit my job as scientist, and leave my settled life behind. This was certainly one of the most challenging decisions and I am often thinking back to inspiring, great and joyful moments.

This decision, quickly turned out to enrich my life on the personal side as well, starting with an unforgettable timid smile from the love of my life, Sandra, in the hallway of the University. She makes me a better person and is unconditionally supporting and encouraging me through all ups and downs. Our son, Florian, gives meaning to my life. The slightest smile from his cute face is able to enlighten my heart – a feeling that I could not have imagined before.

Last, but not least, I would like to thank my siblings and family. Sincerest gratitude is devoted to my parents for their unconditioned love, support and encouragement. I owe them more than I could ever give back and I will try me best follow their way of living.

Finally, I gratefully acknowledge financial support from the Karl Popper Kolleg on Network Autonomous Aerial Vehicles funded by the University of Klagenfurt and the Control of Networked Systems group, which is part of the Institute of Smart Systems Technologies at the University of Klagenfurt.

# Abstract

In this dissertation, we investigate the issue of robust localization in swarms of heterogeneous mobile agents with multiple and time-varying sensing modalities. Our focus is the development of filter-based and decoupled estimators under the assumption that agents possess communication and processing capabilities.

In the beginning, we study the filter-based Collaborative State Estimation (CSE) on a networked system model and propose a novel and approximation-based algorithm for Distribued CSE (DCSE). Then, we apply our insights from DCSE, to address the problem of modularity and scalability in multi-sensor fusion on a single agent. We demonstrate successfully that filter decoupling strategies originating from CSE can be applied to decouple the state of an individual agent, leading to a truly modular estimation framework. In total, three different strategies originating in the domain of DCSE were applied and compare them against a state-of-the-art approach. In evaluations, we show that our proposed approach, is scaling best with the number of sensors and outperforms the state-of-the-art approach in terms of accuracy. We underline that the application of our CSE-inspired method in such a context breaks the computational barrier. Otherwise, it would, for the sake of complexity-reduction, prohibit the use of all available information or would lead to significant estimator inconsistencies due to coarse approximations.

The modular fusion approaches allow us to revisit indoor localization using ranging devices and to incorporate fully-meshed, tightly-coupled range measurements. Being able to consider range measurements between the stationary ranging devices, allows us to improve the self-calibration of those devices and to improve the estimation accuracy.

Based on the findings from DCSE and modular sensor fusion, we propose a novel Kalman filter decoupling paradigm, which is termed Isolated Kalman Filtering (IKF). This paradigm is formally discussed and the treatment of delayed measurement is studied. The impact of approximation made was investigated on different observation graphs and the filter credibility was evaluated on a linear system in a Monte Carlo simulation.

Finally, we propose multi-agent modular sensor fusion approach based on the IKF paradigm, in order to cooperatively estimate the global state of a multi-agent system in a distributed way and fuse information provided by different on-board sensors in a computationally efficient way. As a consequence, this approach can be performed distributed among agents, while (i) communication between agents is only required at the moment of inter-agent joint observations, (ii) one agent acts as interim master to process state corrections isolated, (iii) agents can be added and removed from the swarm, (iv) each agent's full state can vary during mission (each local sensor suite can be truly modular), and (v) delayed and multi-rate sensor updates are supported.

Extensive evaluation on realistic simulated and real-world data sets show that the proposed Isolated Kalman Filtering (IKF) paradigm, is applicable for both, truly modular single agent estimation and distributed collaborative multi-agent estimation problems. By providing detailed pseudocode for each approach and a source code for the IKF paradigm, we hope to pave the way towards generic plug-and-play systems for challenging real-world and multi-agent applications.

# Kurzfassung

In dieser Dissertation untersuchen wir die robuste Lokalisierung in Schwärmen von heterogenen, beweglichen Agenten mit mehreren und sich einer zeitlich ändernden Anzahl an unterschiedlichen Sensoren. Der Fokus liegt hierbei auf filterbasierte Systemparameterschätzer (sogenannte Zustandsschätzer) unter der Annahme, dass individuelle Agenten die Fähigkeit zur Kommunikation und Datenverarbeitung besitzen.

Zunächst betrachten wir das Problem der kollaborativen (agentenübergreifende) Zustandsschätzung, basierend auf Kalman-Filter Formulierungen, anhand eines generisch modelierten und vernetzen Systems. Anschließend formulieren wir einen annäherungsbasierten Filter für verteilte kollaborative Zustandsschätzung.

Die gewonnenen Einsichten aus der kollaborativen Zustandsschätzung, werden anschließend im Bereich der modularen Multi-Sensor-Zustandsschätzung für individuelle Agenten angewendet. Dabei können wir zeigen, dass Ansätze aus der kollaborativen Zustandsschätzung im Bereich der modularen Zustandsschätzung erfolgreich angewendet werden können. Insgesamt werden drei Ansätze aus der kollaborativen Zustandsschätzung in den Bereich der modularen Zustandsschätzung für individuelle Agent überführt und mit einem weiteren modularen Ansatz verglichen. In unseren Evaluierungen zeigen wir, dass der präsentierte Ansatz am besten mit der Anzahl an verwendeten Sensor skaliert und dabei genauere Schätzungen als der native Ansatz liefert. Dadurch wird deutlich, dass die gewonnenen Inspirationen aus dem Bereich der kollaborativen Zustandsschätzung die Grenzen und Limitierungen in der Datenverarbeitung bändigen können. Alternativ müsste entweder auf verfügbaren Daten verzichtet oder Annäherungen getroffen werden, was wiederum einen signifikanten Einfluss auf die Genauigkeit und Konsistenz der Schätzung.

Unser modular Ansatz erlaubt es, ein bekanntes und teilweise ungelöstes Problem der Lokalisierung im Innenbereich, basierend auf Distanzmessungen, anzugehen. Dabei werden sämtliche Distanzmessungen eines voll vermaschten Netzes berücksichtigt. Das Einbeziehen von Messungen zwischen stationären Messgeräten erlaubt uns, neben einer Verbesserung der Lokalisierungsgenauigkeit, auch die geschätzte Position dieser stationären Geräte zu verbessern. Aufgrund der gewonnen Erkenntnisse aus den Bereichen der kollaborativen und der modularen Zustandsschätzung, können wir ein neues Paradigma zur Entkoppelung von Kalman-Filter Instanzen formulieren, welches wir als isolierten Kalman-Filter, kurz IKF, bezeichnen.

Schließlich beschreiben wir einen Ansatz mit Hilfe des IKF Paradigmas, welcher die kollaborative und modulare Zustandsschätzung vereint. Dies erlaubt es den globalen Zustand des Schwarms verteilt und die individuellen Zustände der Agenten modular und effizient zu schätzen. Dieser verteilte Ansatz zeigt folgende Eigenschaften auf: (i) Kommunikation zwischen Agenten wird nur während der Verarbeitung von gemeinsamen Messungen benötigt, (ii) ein Agent übernimmt vorübergehend die Verantwortung über die Berechnung der gemeinsamen Messung, (iii) Agenten können zur Laufzeit vom Schwarm entfernt oder hinzugefügt werden, (iv) die lokale Zustandsgröße der individuellen Agenten kann sich zur Laufzeit ändern, und (v) verzögerte oder asynchrone Sensordaten können ordnungsgemäß berücksichtigt und verarbeitet werden.

Wir führen extensive Evaluierungen mittels realistisch simulierten und echten Sensordaten durch und können dabei zeigen, dass das IKF Paradigma sowohl auf die modulare Zustandsschätzung auf individuellen Agenten als auch auf Agenten-übergreifenden kollaborativen Zustandsschätzung anwendbar und vorteilhaft ist. Zudem wird jeder Ansatz, neben der Formulierung, mit einem detaillierten Pseudocode beschrieben und für das IKF Paradigma der Quellcode bereitgestellt, in der Hoffnung, damit den Weg für generische Plug-and-Play Navigationssysteme im Bereich von herausfordernden Schwarmanwendungen bereitet zu haben.

# Chapter 1

# Introduction

Collaboration plays a crucial role in the success and survival of numerous species across diverse domains [171]. Performing actions collaboratively, enables, e.g., small groups of ants to astonishingly transport significantly larger insects into their habitat. Through this collaborative effort, ants leverage their individual strength, allowing them to overcome challenges that would otherwise be insurmountable.

In a different domain, the field of astronomical observations, collaborative sensing is necessary for optimal results. Notably, very-long-baseline interferometry (VLBI) are established by combining receivers all over the world to capture and process a single image of deep space. The collaboration within this sparsely distributed network of sensors allows the acquisition of information that would otherwise remain beyond reach [172].

In the domain of automation, Henry Ford's introduction of moving assembly lines revolutionized production processes. In the beginning, manufacturing took advantage from purely mechanical machines or actuators, but nowadays, modern robotics such as multi-axis lift arms are accelerating and improving production lines immensely [170].

Expanding the scope of automation, it is noteworthy that robotics is no longer confined to stationary systems. Instead, the inclusion of mobile entities, including agile devices like multi-copters, has become part of humans world and form a pillar of the Anthropocene – the geochronological epoch where the humankind has the most significant impact on the earths biological, geological and atmospheric processes [169].

Collectively, these examples underline that collaboration among members of a group, team, or species offers benefits that surpass those of isolated individuals. Therefore, collaboration between mobile robots is a natural next step, leading to significant benefits of individual robots and allows them to achieve tasks that an individual would not be able to. These advantages are typically realized through the distribution of workload, which, in turn, can lead to a reduction in overall mission duration, enhanced accuracy, and improved efficiency. However, it is important to mention that such benefits come along with the necessity for communication and task coordination, which entail additional overhead costs.

In context of this thesis, the problem of localization and navigation of groups of robots, and more generally agents is studied. More specifically, how agent's belief about their pose (position and orientation) can be estimated by considering the belief of a single agent as part of a group belief (constituting of many individual agents' beliefs) and how individual's perception of its environment can be used to improve the group's belief. Basically, the question "Where am I?" is extended to the question "Where are we?" and "What do we known about each other?".

## 1.1   Related Work

This chapter will illustrate that distributed state estimation in the area of networked robotics and autonomous navigation is still an open and cluttered research field. A general solution addressing all individual problems reaching from tracking of moving objects, relative measurement to stationary object of interest/landmarks, relative measurements between moving objects, over to measuring collaboratively a single phenomena, does not exist. Solutions and approaches to these individual problems are fairly well understood, but a generalized approach that can address and take advantage of the individual solutions is still missing. An evidence is (and in spite of) the large number of (recent) publication in individual domains.

Generally, the domain of distributed state estimation and Kalman filtering [77] is rather broad, as the bibliographic review by Mahmoud and Khalid unveils in [99]. In [142], Sun *et al.* provide a review on distributed sensor fusion in networked system, covering different centralized and distributed solutions and network specific phenomena, such as packet dropouts, random latency due to transmissions delays, information quantization, etc..

The related work discussed in this thesis is split into three – autonomous navigation specific – domains with subcategories: (i) Collaborative Localization (CL), Collaborative Target Tracking, and (iii) Modular Sensor Fusion (MSF) with focus on filter-based approaches.

Optimization-based approaches are becoming more and more important due to increased processing capabilities of today's hardware. Their batch-processing is more accurate, computationally intensive, and leads typically to a higher processing latency, which in the case of estimation has a negative impact on the control behavior in dynamic environments. In optimization-based approaches, the issue of high sensor rates (e.g., of an IMU) is well known. For instance, Forster *et al.* proposed in [41] a pre-integration technique reducing the computational burden. Nonetheless, by assuming a swarm consisting of multiple agents, where each individual is equipped with many fast sensors, is still challenging for these approaches. The same holds for a naive and centralized filter-based estimator, which renders CSE impractical.

Therefore, Kalman filter decoupling strategies need to be applied, enabling parallelism for load distribution and a distributed architecture to address a problem collaboratively with individual agents that posses communication and processing capabilities.

Our focus is in particular on collaborative localization by encountering relative measurements between robots using distributed filter-based approaches, but the other domains are covered to provide a wider view on the topic. We apologize in advance for any omission of publications and potentially disregarding relevant aspects.

We do not covered high level aspects such as collaborative exploration (e.g., [19]), self-aware navigation (e.g., [158]), or formation control (e.g., [30]).

### 1.1.1   Historical Review

This historical review will focus on filter-based approaches and more specifically on those based on the Kalman filter introduced by Rudolf E. Kalman in 1960 [77], to obtain an impression which topics have been addressed until the first distributed collaborative localization approach was published four decades later by Roumeliotis and Bekey in 2000 [123]. In 1966, Schmidt introduced nuisance parameters, where the mean is assumed to be constant and known, but the parameter's uncertainty and correlation to other estimates is considered [129].

Later in 1976, parallel Kalman filtering for a multi-sensor system consisting of, e.g., multiple radar or cameras are observing the same object of interest (e.g. an aircraft, pedestrian, re-entry body) from different stationary and a priori known positions was introduced by Willner *et al.* in [151].

This approach was further refined by Hashemipour *et al.* in 1988, with a particular interest on scalability with respect to the total number of sensors, as the processing of obtained measurements on a centralized entity becomes impractical [56]. Solutions are parallel filters, where measurement updates are computed in parallel, which is in contrast to sequential filters, where the filter is updated sequentially (with zero prediction time) in blocks of statistically independent data. In parallel filtering, one needs to distinguish between synchronously (all measurements can be related to a single timestamp) and randomly collected data. Hashemipour *et al.* mentioned in [56] different levels of parallelizing the Kalman filter equation: parallelism at the time-update level, at the measurement-update level, via segmentation and via decoupling the time-prediction (predictor) and the measurement update step (corrector).

In 1990, Durrant-Whyte *et al.* describe a fully decentralized architecture for data fusion problems for networks of sensors nodes [34]. In 1992, Grime *et al.* investigated on the communication in decentralized data-fusion systems [51]. In 1993, a multisensor surveillance system for, e.g., target tracking in a decentralized sensing architecture was presented by Roa *et al.* in [119]. They extended the decentralized Kalman filter formulation from a hierachically distributed system [56] (central processing and communication facilities) into a fully connected decentralized processing architecture.

In 1994, Kurazume *et al.* addressed the cooperative positioning with multiple robots as portable landmarks that need to follow certain motion strategies [84], followed by studies on optimal movement. These approaches have some significant limitations as only one robot of the team is allowed to move for a certain time (some scheduling is needed) and robots must maintain direct visual and communication contact over time.

In 2000, Roumeliotis and Bekey, provide a more generic and versatile solution, allowing robots to move simultaneously without the need for persistent communication and visual contact [123]. The proposed cross-covariance factorization allows a fully distributed state propagation step/cycle, which is a key ingredient of the proposed approaches.

## 1.1.2   Collaborative Localization

Regarding DCSE for autonomous robotic navigation, a large body of work and approaches were developed to tackle the cooperative localization problem. Typically, proprioceptive measurements that monitor the motion of the robot are combined with exteroceptive measurements, that provide information about the environment and its signals (pressure, noise, gas) or signatures (features landmarks) [123]. Collaborative localization is achieved when multiple agents are sharing their location and sensor information among others to improve both, their own (ego) estimates and, at the same time, the estimates of collaborating agents [39]. Additionally, Collaborative Localization (CL) enables to consider another sort of exterocetive measurements: exteroceptive measurements that describe relations between individual agents, e.g., relative angle/bearing, position, orientation, distance, etc. measurements, which we call joint or collective observations, as these allow to estimate multiple agents states jointly, see e.g., [101]. A disadvantage of joint observation is the introduced complexity in the filter formulation, due to statistical coupling between individual agents.

In the past decades, different approaches for collective multi-agent localization have been presented, while the challenges in filter-based approaches are (i) to *decouple* the individual agents (statistically) to relax the communication constraints, and at the same time (ii) to *maintain* and account for correlations between agents to achieve consistent estimates. Another objective is to reduce the substantial computational and bookkeeping efforts for the correlations in swarms of communicating agents. These distributed approaches can be roughly classified as (i) centralized-equivalent (e.g., [8, 67, 80, 97, 108, 121]), (ii) approximated correlations (e.g., [76, 95]), (iii) CI-based methods for unknown correlations (e.g., [7, 23, 91, 148]), (iv) optimizing correlations (e.g., , [162]), and (v) optimization-based methods (e.g., , Maximum A Posteriori (MAP)-based, e.g., [114], factor-graph based, e.g., [69, 81]), Maximum Likelihood Estimation (MLE)-based, e.g., [64]), (vi) sampling based approaches Monte Carlo localization (MCL), represents the *a-posteriori* belief as a set of weighted random samples or particle, name particle filter (PF)), which was, e.g., proposed by Dellaert *et al.* [31] or Fox *et al.* [44].

With respect to our approaches, the seminal work [121, 123] by Roumeliotis and Bekey investigated on decoupling the state propagation equation of a centralized multiagent EKF. Their proposed factorization/decomposition of cross-covariance terms allows distributed local (isolated) state propagation, and plays an important role in the proposed approaches. However, each private and joint measurements require mesh-based connectivity among all $N$ agents to exchange information among themselves, resulting in a communication link complexity of $\mathcal{O}(N^2)$ per update. Despite that, they explicitly addressed the problem of sensor data interdependencies that appear from joint observations. In an empirical study, they show an improved accuracy and reduced uncertainty due to information flow through interdependencies (correlations), which constitutes a form of *sensor sharing*.

Later, a theoretical analysis of the influence of the homogeneous swarms' size and sensors configuration with respect the localization accuracy and the upper uncertainty bound was covered, for instance in [101, 102, 108, 109, 111, 122, 126] and are summarized in a discussion on the properties of CSE in Section 3.2.6.

Madhavan *et al.* [97], applied heterogeneous cooperative localization in absence of absolute sensors in outdoor terrain.

In [80], Kia *et al.* present a DCSE algorithm that is exactly equivalent to a centralized EKF algorithm. In their algorithm, the propagation step is decoupled, but only one pairwise relative update can happen simultaneously in the network, since the intermediate correction terms, which each node uses to update its estimates, must be broadcasted. Further this imposes that the communication graph must be a direct graph, with a root at the interim master. To overcome communication range limits, each node needs to re-broadcast every received message, introducing latency, causes potential network congestion, and acquiring additional bandwidth. Further this approach assumes single pair-wise relative observations at the time to be the only source of state correction/updates, meaning individual nodes cannot perform private state observation (e.g. GPS measurements) meanwhile/simultaneously. Multiple, synchronized pairwise relative measurements can be performed, but a certain *sequential-updating-order* must be ensured.

In our approach, no restrictions regarding the network topology, nor regarding type of observation and sequence are given, since updates are performed *isolated* among *participating* agents, allowing a communication link complexity of $\mathcal{O}(1)$ among the participating nodes, while in [80], the worst case communication complexity per node per relative measurement is $\mathcal{O}(N^2)$. Similar to our approach, the size of a message is independent from the group size, thus the communication message size is of order $\mathcal{O}(1)$.

The use of CI can help to reduce extensive bookkeeping in a decentralized approach. CI, introduced by Julier and Uhlman in [71], is a convex optimization problem to account for unmodeled/unknown correlations between beliefs of the same stochastic process in a conservative but provable consistent fashion. Therefore, CI is used in CL to fuse the beliefs of multiple group states that are maintained per agent [7], or to fuse projected estimates from other agents relative observation accounting the local belief, e.g., in [23, 148].

The downsides of the later are twofold. First, it is rather limited to a specific type of relative observation, which has to be directly and statistically combined with local state variables to obtain an observation on a different agent. Second, only one estimator benefits from the relative measurement, while in our approach all estimators participating in a joint update obtain a correction. According to Wanashinghe *et al.* [148], this leads to overly pessimistic estimates, as the belief is assumed to be maximally correlated, on the other hand Carillo-Arce *et al.* state in [23] that, Split-Covariance Intersection Filter (SCIF) proposed by Li *et al.* in [91] may lead to overly optimistic and inconsistent estimates, as the beliefs are either directly or indirectly correlated to other agents' beliefs. To fix that, Wanashinghe *et al.* proposed in [148] to *reset* the independent part in the SCIF.

In [161, 162], Zhu and Kia proposed the Distributed Discorrelated Minimum Variance (DDMV) approach to discorrelate the agents' local beliefs by weighting and underestimating them. This reduces both communication and maintenance cost to $\mathcal{O}(1)$, yielding a highly scalable, but pessimistic approach. Unfortunately, this approach did not work as shown in Figure 3.10 due to inconsistencies.

Later, Zhu and Kia presented in [162] the Estimated Cross-covariance Minimum Variance (ECMV) approach to reduce the conservatism of DDMV by estimating the unknown cross-covariance between agents in a cascaded optimization. In their evaluations, ECMV outperforms DDMV at the cost of $150 \times$ longer processing times for joint updates, thus it is not real-time capable.

In [95], Luft *et al.* proposed an approximation to account for unavailable correlations between participating and non-participating agents in isolated joint observations. This approximation approach requires communication only when agents meet ($\mathcal{O}(1)$) and the maintenance effort for the interdependencies scales linearly with $\mathcal{O}(N)$ for $N$ correlated agents.

Jung and Weiss successfully applied this approximation in a more advanced estimation problem on a group of Micro Aerial Vehicles (MAVs) in [73], using an error-state EKF (ESEKF) with an IMU as state propagation sensor. We identified high sensor rates (in aided inertial systems the IMU's rate is typically between $100\,\mathrm{Hz}$ and $1\,\mathrm{kHz}$), causing tremendous maintenance effort and to be a limiting factor for the applicability in large swarms. Therefore, Jung and Weiss proposed in [76] a common fixed size sliding buffer for correction terms within each agent, allowing the maintenance cost to be invariant to the number of known/correlated agents. We advance these approximation-based approaches in Chapter 7, by (i) generalizing Luft *et al.'s* approximation [95] in Chapter 6, (ii) supporting growing and shrinking state vectors on agents, and (iii) supporting delayed (out-of-order) measurements.

Recently, Shalaby *et al.* investigated on relative position estimation using range measurements between multiple MAVs [132]. While the relative position, would be in case of single ranging devices per agent unobservable, they studied the sufficient conditions in case agents are equipped with a 9-Degrees of Freedom (DoF) IMU (including an accelerometer, gyroscope, and magnetometer) and two or one ranging devices. They prove, that at least two agents with two ranging devices are required to render the relative localization problem observable. Obviously, there are degenerate cases, when e.g., the baseline between the ranging devices is zero, if the baselines of the two agents are parallel, non-line-of-sight conditions, or if the sensing and communication range is insufficient. Still, these conditions can be mitigated and are relevant for formation control. Alternatively, Cossette *et al.* showed in [27], that the relative position between two agents, each equipped with a single ranging devices and a 9-DoF IMU, can be recovered by persistent and sufficient motion in a sliding window filter (a optimization-based batch estimation framework with a constant time-window) that exploits relative constraints at selected reference points by combining range measurements and acceleration estimated. Recently, Cossette *et al.* investigated in [28] on improving the accuracy of relative pose estimates by optimizing the formation/constellation of a swarm of MAVs that is obtaining inter-agent range measurements. Each MAV is equipped with an IMU and two ranging devices (one on front left and right arm of the quad-copter). Multiple tags per agents, can mitigate the requirement for persist excitation, i.e. persistent motion to render the state space observable, see e.g., [13, 49]. Alternatively, additional sensor that render the full state observable can be used, e.g., additionally stationary anchors or a Global Navigation Satellite System (GNSS) sensor. Still having multiple tags per agents, tuples of tags must be non-collinear with other agents' tags. The optimal formation is obtained by maximizing the Fisher information, which is similar to [16], where optimal positions for the moving ultra-wideband (UWB) tag to initialize the positions of stationary UWB anchors were obtained by the same method. In simulation and experiments, it was shown that the accuracy can be improved by a proper choice of geometry, which depends also on the observation/measurement graph. Avoiding collisions with obstacles and between agents is inherently supported by the proposed approach and is realized by adding terms in the cost function.

Another distributed, but centralized-equivalent approach with sporadic and asynchronous communication, was proposed by Allak *et al.* in [4]. At the moment of pair-wise measurements, pre-computed scattering factors from both agents are exchanged and used in the joint correction step to account for information obtained since their last encounter in a statistically correct way. By encountering this factors, a centralized-equivalent estimate can be obtained by very few operations at constant maintenance effort. The disadvantage remains in the limit for pairwise state estimation and, thus, poor scalability.

### 1.1.3   Collaborative VIO

Visual-Inertial Odometry (VIO) tracks the ego-motion of locally fast drifting IMU odometry by constraining/correcting the drift by visual observation of the environment, e.g., in [92]. In case agents are performing VIO, one can aim at using information obtained by overlapping camera views.

CL assumes that agents can directly observe one another, which is, depending on the sensor modalities not always possible, therefore indirect observations, e.g., via overlapping camera views might be exploit to improve/correct the estimates [106].

In [1], Achtelik *et al.* proposed a pair-wise method to recover the relative pose between two MAVs in absolute scale based on the individual's IMU and monocular camera. However, this centralized EKF approach has certain limitation with respect to the flexibility and scalability, which was the motivation of the distributed loosely-coupled VIO approach proposed by Jung *et al.* in [73]. As in [1], it was assumed that the metrically scaled relative pose based on overlapping (stereo) camera views, is provided by a black box, while the challenge is probabilistic re-alignment of the individual agents navigation frames to a common one, such that they converge to a common reference. The first pair-wise relative pose measurement was used to eliminate the navigation frame of one of the agents, e.g., based on the reference ID.

In [106], Melnyk *et al.* investigated on exploiting trails of a common feature in case of overlapping camera views between two agents in a centralize filter formulation, in order to improve the localization accuracy and studied the observability properties of the relative transformation under different measurement scenarios. They proofed that at least five common features must be determined at a single time step in order to recover 5-DoFs of the relative pose. All 6-DoFs of the relative pose can be recovered, if at least three common features are observed in two consecutive images, will this can be relaxed to two features if the direction of the gravity vector is known.

Recently, Zhu *et al.* proposed in [163], a distributed collaborative VIO, where each of the agents maintain locally it's states and tries to leverage information obtained by other robots, with global/centralized optimization or a centralized Fusion Center (FC), which builds upon [166]. A distributed collaborative thermal-inertial odometry approach was proposed by Polizzi *et al.* in [117] for a team of MAVs, that exchange information when possible to refine their states using covariance intersection. Xu *et al.* proposed in

[154] Omni-swarm, which allows each agent to incorporate stereo omnidirectional VIO and ultra-wideband in the front-end, while in the back-end a graph-based optimization is performed on own and broadcasted sensor data from agents to compute relative state constraints.

### 1.1.4 Collaborative Simultaneous Localization and Mapping

Autonomous systems that need to navigate in unknown and infrastructure-free (e.g., GNSS-denied) environments, must rely on their proprioceptive and exteroceptive sensors, which can be used to create an abstract representation of the environment by detecting and mapping landmarks, which referred to as Simultaneous Localization and Mapping (SLAM). Filter-based localization and mapping looks back on more than three decades of research and is the union of two domains, navigation and mapping, with many relevant existing work in each subdomain[39].

In 1991, Leonard and Durrant-Whyte investigated in [88] on simultaneous map building and localization for an autonomous mobile robot, and highlighted the problem of correlation between estimated robot pose and the estimated target/landmark of interest. A potential solution, proposed by Smith, Self, and Cheeseman [135] in 1990, is to maintain a stochastic map using an EKF, with limitations that the system state vector size increases linearly with the map size, but the computational complexity in an EKF increased cubically with state dimension. Further, the data association is uncertain, e.g., by spurious measurements or a wrong data association, which have negative impact on the estimation performance and stability. The fundamental assumption that the observed landmarks in the environment are static, might be violated by dynamic elements.

Similarly, Fox *et al.* identified in [43] as key problem of EKF-SLAM, that the uncertainty of the mobile robot's pose while exploring the environment, as it needs to be considered during generating the map. These interdependencies between the estimated pose and the estimated landmarks makes the problem computationally demanding. In [144], Thrun *et al.* introduced the sparse extended information filter (SEIF) for SLAM, and showed, by forcing sparsity of the information matrix, that update can be processed in constant time. A consistent EKF-based SLAM with linear complexity (with the number of landmarks) was proposed by Nerurkar and Roumeliotis in [113].

In 2002, Fenwick *et al.* generalized EKF-SLAM to support collaborative robots [39], denoted as Collaborative SLAM (CSLAM), in order to map an environment more quickly and robustly than a single robot using a centralized filter. They proposed a theorem to quantify the performance gain through collaboration and allows determining how many vehicles are required to achieve a certain task. They stressed that in case of relative localization between stochastic landmarks and the vehicle, the lower bound for the uncertainty of the agent's state is determined by the initial covariance of the agent at the time of the first landmark observation. Meaning that, by performing stochastic mapping of the environment, in the best case, the accumulated drift error can be bounded, if the initial landmarks are revisited, but no gain with respect to the absolute error is obtained. In case of multiple agents, the lower bound is determined by the sum of all collaborative agents initial information matrix (inverse of covariance) at the moment of first landmark or inter-agent observation [39]. Meaning that uncertainty of the entire swarm is, in the limit of infinite observations, lower than the most accurate vehicle, while all agents are contributing to the total information.

In [111], they investigated on an analytical upper bound of the position uncertainty in a steady state condition in collaborative EKF-SLAM.

In [160], Zhou and Roumeliotis studied the problem of merging multiple agents maps in case of a rendezvous (when agents are sensing each other and are within the communication range), which allows a swarm to jointly build a map (abstract representation of the environment) starting independently without any *a-priori* knowledge about the initial spatial distribution swarm – a key to achieve a modular, flexible, and scalable swarm. The idea is to measure a noisy relative pose between two agents to obtain a transformation for aligning the individual maps and to compute a new joint covariance for the joint system. After the (course) alignment, duplicated landmarks are identified using a search tree and a hypothesis check based on the Mahalanobis distance (see Section 2.5.3). Then, identified duplicated landmarks are used to impose constraints that improve the accuracy of the merged map and reduces the state vector size. In general, the identification of duplicates is sensitive to rotational errors, especially in large maps (i.e. due to the longer lever arm), therefore landmarks close to the rendezvous need to be matched first and sequentially.

Apart from filter-based CSLAM, centralized, smoothing-based, or factor graph-based, approaches have been proposed. In [32], Dellaert and Kaess introduced square root Smoothing and Mapping (SAM). Later, a collaborative SAM approach was proposed by Andersson and Nygards in  [6] and Kim *et al.* introduces a relative formulation between multiple pose graphs that allows to solve pose graphs jointly in order to provide a globally consistent map [81]. In [168], Zou and Tan present a collaborative visual SLAM, denoted as CoSLAM, which is gathering information from agents on a local server and is capable of clustering dynamic elements in the environment.

Due to the computational complexity and constraints on, for instance MAVs, it is reasonable or required to offload computational intensive processes to a central ground station or to exploit edge computing facilities [59]. Forster *et al.* proposed in [42], a monocular CSLAM using multiple MAVs that stream selected features and relative poses to a central FC, that creates and merges individual maps in case of detected overlaps. Similarly, Schmuck and Chli [130] propose a mono CSLAM employing MAVs with limited memory, that are transmitting collected information to a fusion center, which performs time-consuming operations, such as loop-closure detection and optimizations (bundle-adjustment) on large scale, and provides the agents with corrections. Karrer *et al.* proposed in [78] a centralized collaborative visual-inertial SLAM (CVI-SLAM), where individual agents are equipped with a visual-inertial sensor suite, while a fusion center (server) is again performing the optimization on a large scale and providing agents with corrections.

### 1.1.5  Collaborative Target Tracking

The problem of combining multiple estimates of a state into a single and more accurate estimate, is denoted as track-to-track fusion [164] and can be addressed by parallel or distributed Kalman filtering, e.g., in [34, 56, 119].

For target tracking we can distinguish between immobile/stationary systems (e.g., [119]) and mobile system (e.g., [164]), while mobile systems are advantageous as they may allow reducing the overall number of sensors for a specific scenario, and they can be reconfigured to such that the perception/information gain of the target is increased [72].

In [119], Rao *et al.* applied a decentralized Kalman filter for multi-sensor data fusion in a sensor network for (multi) target tracking, where each sensor node has processing and fully-meshed communication capabilities. Each sensor is estimating a global state of each target in a local KF. It fuses locally the obtained measurements by predicting the local estimate based on a known system model. The measurement is validated based on a hypothesis check and then used for updating the local belief. The error of the belief is transmitted to other nodes, while at the same time, state error beliefs are received from others. These errors are assimilated to obtain further variance and mean corrections. One disadvantage is that potential correlations between individual sensors' noise is not considered. Apart from the synchronous communication constraint, they investigated on asynchronous communication between nodes, which plays an important role for system modularity and nodal autonomy.

Generally, if the cross-correlation between sensors is known, the beliefs can be fused using BLUE [9], if cross-correlations are unknown, they can be fused consistently using CI [71], and if no correlation exist, they can be fused by exchanging and assimilating correction terms [119].

### 1.1.6 Collaborative Localization and Target Tracking

Mirzaei *et al.* studied in [108], the accuracy of Collaborative Localization and Target Tracking (CLATT) based on an EKF in a multi-robot system and derived a theoretical upper bound at the steady state condition for the position uncertainty and showed that jointly estimating the targets results in a better accuracy in the estimates of the individual agents. Later, Chang *et al.* [24], studied the feasibility of [108] in different scenarios and on different platforms.

In [3] Ahmad *et al.* proposed a optimization-based solution to cooperative robot localization and target tracking. Later, Ahmad *et al.* proposed in [2] an particle filter-based approach that outperform a joint KF formulation and achieves similar results as an offline batch optimization approach.

In [165], Zhu and Ren proposed a distributed filter-based multi-robot VIO approach with collaborative object tracking, where one agent performs visual inertial SLAM to bound the long-term drift of the individual agent's VIO. Recently, Zhu and Ren introduced in [165] and [164] a framework to distributively solve the CLATT problem without motion strategies for the robots and requiring only one-hop communication with neighbors, while preserving consistency.

### 1.1.7 Modular Sensor Fusion

Multi-sensor fusion can be regarded from different views. For instance, in [34], Durrant-Whyte proposed a Decentralized Kalman Filter (DKF) as an implementation of a multi-sensor EKF, divided into modules associated with each sensor, which can be applied for, e.g., target tracking using a sensor network that is observing a common random process. A different view might be the fusion of certain, maybe time-varying, sensor constellations/configurations in multi-sensor EKF formulation – in best case in a plug and play fashion – to achieve a certain navigation task. An example for a plug-and play Visual-Inertial Navigation System (VINS) that support online self-calibration of certain sensor related state is MIMC-VINS [36], proposed by Eckenhoff *et al.* . To avoid confusion we use for the first case the term *track-to-track fusion*, and for the second case, *modular sensor fusion*, while both are fusing observation from multiple sensors.

Existing multi-sensor fusion algorithms for navigation can be classified as optimization-based (e.g. [25, 29, 128]) or filter-based (e.g. [18, 46, 58, 96, 150]) approaches. Optimization-based methods result in more accurate estimates at the cost of computation time, while filter-based approaches are computationally more efficient through marginalization but struggle in correcting erroneous information in the past.

In [150], Weiss and Siegwart proposed a real-time capable loosely coupled visual aided inertial estimator (based on an EKF), supporting any arbitrary exteroceptive position or pose sensor.

Lynen *et al.* presented in [96] a filter framework for navigation as an extension of [150] to support any number of sensors, iterative re-linearization, and an efficient strategy for handling relative time measurements. Their architecture distinguishes between a core state that is propagated by the IMU and auxiliary states that hold spatial and other calibration information for the heterogeneous sensor suite. The simple approach tackling delayed measurements impedes scaling in practice for several sensors due to increased computational complexity. To address the computational burden to incorporate delayed measurements, Allak *et al.* proposed in [5] the use of star products between updates to perform approximated one step recalculations.

In [58], Hausman *et al.* investigated on the self-initialization and self-calibration of their multi-sensor fusion approach combining vision, Global Positioning System (GPS) and range measurements. They could seamlessly switch between different sensor modalities during an evaluation on a MAV. Further, an online failure detection was proposed, evaluating the Chi-Square test (a hypothesis test) on the innovation/residual to avoid incorporating erroneous out-of-order measurements.

Shen *et al.* proposed in [133], an Unscented Kalman Filter (UKF) approach that supports both heterogeneous absolute and relative time (multi-state constraint) updates. The UKF simplifies the filter formulation compared to an EKF as no explicit calculation of Jacobians or Hessian are necessary (deriving the Jacobians is known to be challenging) as these are evaluated precisely through its sigma point propagation [147]. However, the UKF introduces new challenges in propagating the augmented states.

In [37], Eckenhoff *et al.* proposed a tightly coupled EKF-based estimation framework, that fuses asynchronous measurements from multiple IMUs and a single camera. This framework allows for an online spatial and temporal calibration refinement between all sensors, which reduces the offline calibration efforts and allows to compensate calibration errors. Due to redundant IMU, it is resilient to a single IMU failure and provides uninterrupted (smooth) localization estimates in case of a detected failure.

Similarly, Geneva *et al.* recently published OpenVINS [46], which is a well documented open source modular Multi-State Contraint Kalman Filter (MSCKF) framework including a profound comparison against other state-of-the-art VINS algorithms. Our presented approach does not explicitly address the problem of fusing visual information, based, e.g., on *relative time updates* of MSCKF algorithms [110][1].

We are not explicitly addressing the problem of fusing visual information, but our proposed algorithm inherently supports, up to the time horizon of the history buffer, *relative time updates* that can be used to account for relative time constraints between states, which are the basis for MSCKF algorithms [110].

More recently, Eckenhoff *et al.* proposed in [36], a multi-IMU multi-camera VINS, that is capable of fusion an arbitrary number of IMUs and uncalibrated cameras. It supports a seamless fusion of asynchronous measurements, even in case of measurement depletion or sensor failures, in a MSCKF formulation. In contrast to our formulation, the propagation/prediction of the full state happens at the moment a new camera image is available. In our formulation, the IMU state is propagated at the moment IMU measurement is received. Lee *et al.* proposed recently in [87] a MSCKF-based MMSF framework incorporating a Laser Imaging Detection and Ranging (LIDAR) feature tracking algorithm that extracts plane patches which are used to from a motion constraint for the MSCKF update.

The previously mentioned multi-sensor fusion approaches operate on a full state, meaning no state decoupling strategies are applied such that (i) sensors cannot be added and removed during the mission, (ii) they are potentially ill-scaling with the total amount of sensors, sensor delays and updates rates. To overcome this issue, Brommer *et al.* published in [18] an open-source MMSF framework denoted as MaRS, that allows the addition and removal of sensors at run-time. A significant speed-up is achieved by decoupling the *core* state that is predicted (propagated) by a proprioceptive sensors (e.g., an IMU) from the generic exteroceptive *sensor* states. As correlations are just maintained between the core and sensor states (no inter-sensor correlations are considered), the propagation step is invariant to the total number of sensors. In case of ill conditioned joint covariances between the core and a sensor state, different Eigenvalue correction strategies have been proposed.

---

[1]Relative time update, require stochastic clone of states [124], that allows incorporate relative time constraints between beliefs in observations, while at the same time accounting for correlation between these past beliefs. The proposed approach maintains a sliding pastime horizon of previous beliefs and the corrections buffer allows tracking the change in correlation between two consecutive beliefs, meaning that stochastic clones and their interdependencies can not be reconstructed directly, as not all correlations can be recovered/maintained. Consequently, stochastic clones need to augment local full state, which can in our modular formulation be realized by adding a new estimator instance to the system. The proposed state decoupling approach allows reducing the maintenance effort for tracking correlations between stochastic clones.

More recently, Jung and Weiss bridged in [74] the gap between CSE and MMSF by showing that techniques used in CSE for distributed agents to decouple their estimators can be applied on local estimators to decouple states from different sensors. This new perspective allowed us to port existing CSE methods and to compare them, including our native MaRS approach. The Modular Multi-Sensor Fusion Decoupled Approximated History (MMSF-DAH) approach [74] uses approximated correction terms for sensor observation and considers inter-sensor correlations, which is advantageous in terms of accuracy. By using a common correction buffer for all correlations, the approach is invariant to the total number of sensors.

Based on the MMSF-DAH approach, we proposed a probabilistic filter based MMSF approach in [75] with the capability of using efficiently all information in a fully meshed UWB ranging network for infrastructure-based UWB AINS with continuous anchor self-calibration. This allows a simultaneous accurate mobile agent state estimation and calibration of the ranging network's spatial constellation. We advocate a new paradigm that includes elements from DCSE and allows us considering all stationary UWB anchors and the mobile agent as a distributed set of estimators/filters. Thus, our method can include all meshed (inter-)sensor observations tightly coupled in a modular estimator and show that decoupling paradigm can break the computational barrier in such a context.

A factor graph based modular estimation framework, WOLF, with a wrapper for Robot Operating System (ROS)[2], was introduced by Solà *et al.* in [138]. A key component is the WOLF tree, a centralized data structure containing abstract base classes to address robotics problems, for instance, sensors, features, etc.. These elements can be added to the tree in a modular way and allows to setup different applications with great flexibility. In the back-end a factor graph with it's states and factors is solved with a nonlinear graph solver. The author mentioned that the real-time performance need to improved, which is a common weak point of related estimation frameworks. This provides us with additional motivation to pursue with our recursive and modular estimation approach, that aims at reducing the computational effort.

## 1.2   Contribution and Outline

Although the topic of multi-agent navigation has being addressed and studied for several decades, the previously mentioned approaches have not been focusing on combining MMSF and CSE in generalized framework. Additionally, in DCSE, the handling of asynchronous and delayed measurements has not yet been explicitly covered and introduces additional challenges. In fact, sensor measurements undergo a certain delay from the moment information is perceived until it is actually processed. Due to the complexity of compensating delayed measurements, it is mostly neglected. Having multiple sensors, the sensor delay can vary, with the consequence that measurements are processed in a wrong order (out-of-sequence). Consequently, the sensor data needs to be processed asynchronously or delayed with respect to other sensor data or a common reference time.

The challenge for distributed processing architectures remains to allow for a loose coupling between agents. Agents shall be able to join and exit the group at any time. This way the need for a centralized processing unit can be mitigated, and the workload can be distributed among them. Further, the group or swarm of agents is most probably heterogeneous (in their type, shape, locomotion, or drive train) and capabilities (e.g., sensor configuration). For instance, in an exploration mission, different agents can be equipped with different exteroceptive sensors or posses different capabilities, as shown in Figure 1.1. During a mission, it might be needed to enable or disable sensors sporadically in order to save energy or sensors may suffer from dropouts, etc.. Meaning the configuration of the swarm should be able to scale in two dimensions. First with respect to the amount of agents and second with respect to the number of active sensors per agent.

---

[2] https://www.ros.org/

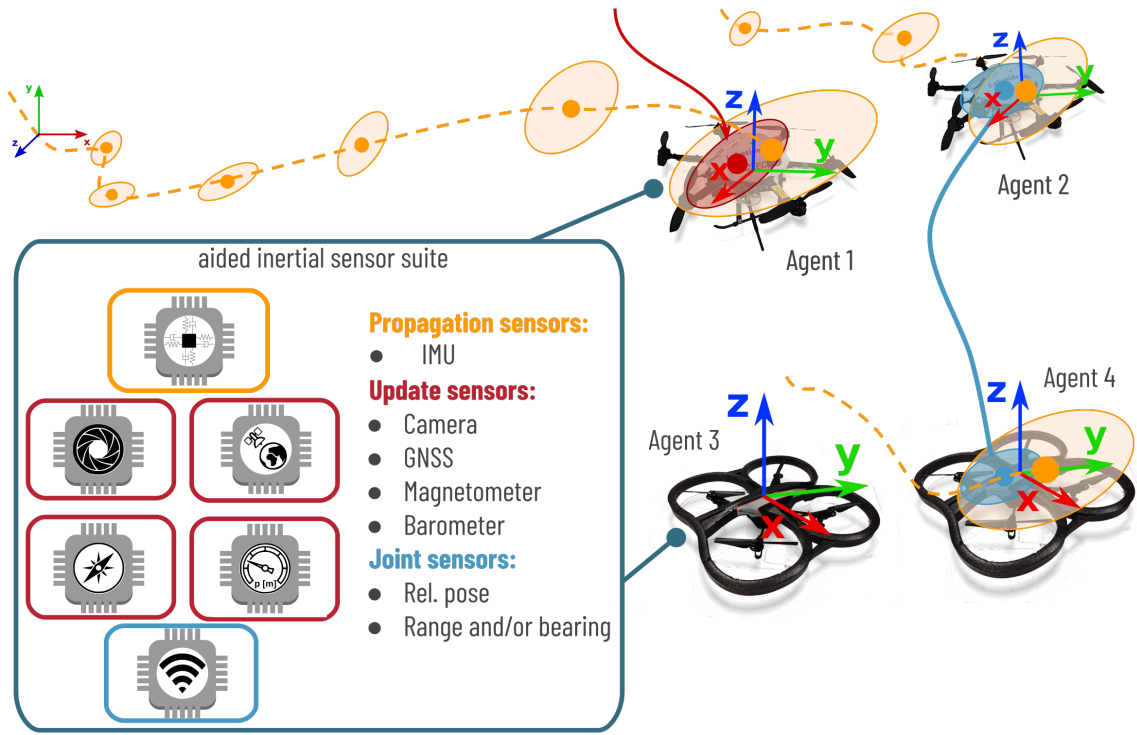

**Figure 1.1:** Use case for CSE: A heterogeneous swarm of four communicating agents, with potentially a different sensor configuration, is navigating in space to achieve tasks. Proprioceptive sensor data, held in orange, allows for a coarse and drifting dead-reckoning, while exteroceptive sensor data, held in red, can reduce that drift and bound the uncertainty of the estimated pose. Additionally, relative inter-agent measurements, held in blue, can further refine the estimates, but lead to correlations between the agents' estimates which need to be maintained and considered properly.

This thesis carefully addresses these issues and studies the impact of sensor delay on the processing time. The final algorithm is capable of handling (i) many sensors, (ii) a time-varying number of sensors, (iii) asynchronous measurement, (iv) fast propagation sensors, (v) and collaborative updates.

The key contribution of this thesis can be summarized as follows:

- We introduce a buffering scheme that allows for an efficient, but approximated, propagation and correction of dated correlations between distributed estimator in Section 3.3.5, which is beneficial in case of high state propagation rates.

- We bridge the gap between CSE and MMSF: three existing DCSE approaches are implemented as modularized and decoupled EKFs. We propose in Section 4.3.1 a novel scalable and general modular sensor fusion strategy for Kalman filters with constant maintenance complexity for propagation and private update steps (if buffer history suffices).

- We formulate and propose a Kalman filter decoupling paradigm referred to as Isolated Kalman Filtering (IKF) in Chapter 6. It builds upon approximations proposed by Luft *et al.* in [95] and supports out-of-sequence measurements. We provide the source code of a generic IKF framework, study the steady-state behavior in different observation graphs, and perform a filter credibility analysis.

- We propose a unified filter architecture for MMSF and CSE based on the IKF paradigm, termed DC-MMSF, in Chapter 7. It supports delayed measurements and the maintenance effort for each decoupled sensor nodes is invariant to the number of correlated sensors ($\mathcal{O}(1)$), if the buffer size is sufficient.

- We provide a detailed pseudocode for all proposed approaches.

The manuscript is structured as follows: In Chapter 2, we describe our notation, definitions, metrics, and filter formulation. In Chapter 3, we aim at generalizing filter-based DCSE and compare different filter architectures. In Chapter 4, we propose a novel modular sensor fusion algorithm for a single agent, which finds application in Chapter 5 for infrastructure-based UWB aided inertial navigation. In Chapter 6, we propose the Isolated Kalman Filtering (IKF) paradigm, a key concept to decouple filter formulations, which allows us to present a modular and multi-agent sensor fusion approach with support for delayed measurements in Chapter 7. We conclude the thesis in Chapter 8, by summarizing the contributions, discussing limitations, and providing an outlook on future research directions.

# Chapter 2

# Fundamentals

In this chapter, we introduce our notation and nomenclature to improve readability and to avoid ambiguity. In that spirit, fundamentals on spatial geometry, error-definitions and metrics for state estimation, and our filter formulation are covered.

## 2.1 Basic Notation

Vectors are lower case bold, matrices capitalized bold. The transpose, inverse, conjugate are $\{\bullet\}^\mathsf{T}$, $\{\bullet\}^{-1}$, and $\{\bullet\}^*$, respectively. The mean of a variable is $\{\bar{\bullet}\}$, the estimated quantity $\{\hat{\bullet}\}$, and the error $\{\tilde{\bullet}\}$. The time index, e.g., $k$, of a variable is in the right superscript $\{\bullet\}^k$ which related it a certain time $t(k) \equiv t^k$. $\{\bullet\}^{\#}$ specifies measured (perturbed) quantities, and $|\bullet|$ specifies the cardinality of a finite set.

For vectors and block matrices, semicolons and colons improve the readability such that $[\mathbf{A}; \mathbf{B}] = \begin{bmatrix} \mathbf{A} \\ \mathbf{B} \end{bmatrix}$ and $[\mathbf{A}, \mathbf{B}] = \begin{bmatrix} \mathbf{A} & \mathbf{B} \end{bmatrix}$.

Names of reference frames are capitalized and calligraphic, e.g., $\{\mathcal{I}\}$ for IMU (spatial geometry is covered in Section 2.3). The upwards pointing z-axis of the navigation reference frame $\{\mathcal{G}\}$ is gravity aligned.

The operators $\boxplus$ and $\boxminus$ should emphasize that state vector elements need to be composed appropriately and depends on the state space (product manifold[1]), the error type (see Section 2.4), and error representation (see Section 2.8.6). Rotational and translational components of a state vector have to be treated differently and are not necessary commutative, e.g., biases are additive, while positions and velocities are defined in the local frame, e.g., ${}^{\mathcal{G}}_{\mathcal{G}}\mathbf{p}_{\mathcal{I}} = {}^{\mathcal{G}}_{\mathcal{G}}\hat{\mathbf{p}}_{\mathcal{I}} \boxplus {}^{\mathcal{G}}_{\mathcal{G}}\tilde{\mathbf{p}}_{\mathcal{I}} = {}^{\mathcal{G}}_{\mathcal{G}}\hat{\mathbf{p}}_{\mathcal{I}} + {}^{\mathcal{G}}\hat{\mathbf{R}}_{\mathcal{I}\mathcal{I}}{}^{\mathcal{G}}_{\mathcal{I}}\tilde{\mathbf{p}}_{\mathcal{I}}$ and ${}^{\mathcal{G}}_{\mathcal{I}}\tilde{\mathbf{p}}_{\mathcal{I}} = \boxminus{}^{\mathcal{G}}\hat{\mathbf{p}}_{\mathcal{I}} \boxplus {}^{\mathcal{G}}_{\mathcal{G}}\mathbf{p}_{\mathcal{I}} = {}^{\mathcal{G}}\hat{\mathbf{R}}_{\mathcal{I}}^{\mathsf{T}}({}^{\mathcal{G}}_{\mathcal{G}}\mathbf{p}_{\mathcal{I}} - {}^{\mathcal{G}}_{\mathcal{G}}\hat{\mathbf{p}}_{\mathcal{I}})$.

### Multivariate random variable

A normally distributed multivariate variable is defined as $\mathbf{x}_i \sim \mathcal{N}(\hat{\mathbf{x}}_i, \boldsymbol{\Sigma}_{ii})$, with a mean $\hat{\mathbf{x}}_i$ and covariance (uncertainty) $\boldsymbol{\Sigma}_{ii}$, which is called the *belief* of $i$. Covariance matrices $\boldsymbol{\Sigma}$ of dimension $n \times n$ are positive semidefinite and are within a convex (half) cone $\mathbb{S}^n_+ = \{\mathbf{M} \in \mathbb{S}^n | \mathbf{M} \succeq 0\}$, with $\mathbb{S}^n := \{\mathbf{M} \in \mathbb{R}^{n \times n} | \mathbf{M} = \mathbf{M}^\mathsf{T}\}$ [21] . Positive semidefinite means that all leading determinants are greater equal zero. Due to the symmetry property, we abbreviate lower triangular elements by a $\{\bullet\}$.

A normally distributed multivariate variable is defined by a N-dimensional mean and covariance matrix

$$\mathbf{x} = \begin{bmatrix} \mathbf{x}_i \\ \mathbf{x}_j \end{bmatrix} \sim \mathcal{N}\left( \begin{bmatrix} \hat{\mathbf{x}}_i \\ \hat{\mathbf{x}}_j \end{bmatrix}, \begin{bmatrix} \boldsymbol{\Sigma}_{ii} & \boldsymbol{\Sigma}_{ij} \\ \boldsymbol{\Sigma}_{ji} & \boldsymbol{\Sigma}_{jj} \end{bmatrix} \right), \tag{2.1}$$

---

[1]Please note that elements might not be commutative, such as rotations or poses.

$\mathbf{\Sigma}_{ij}$ is denoted as the *cross-covariance* between the belief $i$ and belief $j$. The lower right subscript of block matrices $\{\bullet\}_{ij} = \{\bullet\}_{i,j}$ describe the corresponding indexing of the $i^{\text{th}}$ row(s) and $j^{\text{th}}$ column(s).

### Correlation matrices

Correlation matrices $\mathcal{K}$ are *elliptopes* $\mathcal{E} := \{\mathcal{K} \in \mathbb{S}|\mathcal{K} \succeq 0, \text{diag}(\mathcal{K}) = \mathbf{I}\}$, which are symmetric positive semidefinite matrices where all the diagonal elements are unity [70]. The correlation of multivariate random variable $\mathbf{x}_a \sim \mathcal{N}(\mathbf{x}_a, \mathbf{\Sigma}_{aa})$, with $\mathbf{\Sigma}_{aa} = \mathbf{E}[\mathbf{x}_a \mathbf{x}_a^{\mathsf{T}}] - \mathbf{E}[\mathbf{x}_a]\mathbf{E}[\mathbf{x}_a]^{\mathsf{T}}$ is defined as $\mathcal{K}_{aa} = \mathbf{E}[\mathbf{x}_a \mathbf{x}_a^{\mathsf{T}}]$.

The *correlation coefficient* $\rho_{ij}$ between the random variables $i$ and $j$ ($i \neq j$) of the correlation matrix $\mathcal{K}$ is defined as:

$$\rho_{ij} = \frac{\sigma_{ij}}{\sigma_i \sigma_j} = \frac{\sigma_{ij}}{\sqrt{\sigma_i^2 \sigma_j^2}} \in [-1, 1] \tag{2.2}$$

with the $\sigma_{i,j}$ as the standard deviation of $i$ and $j$ receptively and covariance between them $\sigma_{ij} = \text{cov}(i, j) = \sigma_i \rho_{ij} \sigma_j$. Thus, the structure of a correlation matrix $\mathcal{K}_{aa}$ is

$$\mathcal{K}_{aa} = \begin{bmatrix} 1 & \rho_{12} & \cdots & \rho_{1n} \\ \rho_{12} & 1 & \cdots & \rho_{2n} \\ \vdots & \vdots & \ddots & \vdots \\ \rho_{1n} & \cdots & \cdots & 1 \end{bmatrix} \tag{2.3}$$

### Skew symmetric matrices

A skew or cross product matrix of a vector $\mathbf{a} \in \mathbb{R}^3$ is defined as

$$[\mathbf{a}]_\times = \begin{bmatrix} 0 & -a_z & a_y \\ a_z & 0 & -a_x \\ -a_y & a_x & 0 \end{bmatrix} \tag{2.4a}$$

$$[\mathbf{a}]_\times^{\mathsf{T}} = [-\mathbf{a}]_\times = -[\mathbf{a}]_\times \tag{2.4b}$$

The skew matrix is used when computing the cross product of two vectors $\mathbf{a}$ and $\mathbf{b}$

$$\mathbf{a} \times \mathbf{b} = [\mathbf{a}]_\times \mathbf{b} = \begin{bmatrix} -a_z b_y + a_y b_z \\ a_z b_x - a_x b_z \\ -a_y b_x + a_x b_y \end{bmatrix}. \tag{2.4c}$$

Note that just as $\mathbf{a} \times \mathbf{b} = -\mathbf{b} \times \mathbf{a}$ that

$$[\mathbf{a}]_\times \mathbf{b} = [-\mathbf{b}]_\times \mathbf{a}. \tag{2.4d}$$

Additionally, the square of $[\mathbf{a}]_\times$ is

$$[\mathbf{a}]_\times^2 = \mathbf{a}\mathbf{a}^{\mathsf{T}} - \mathbf{a}^{\mathsf{T}}\mathbf{a}\mathbf{I} \tag{2.4e}$$

## 2.2 Nomenclature

### Centralized, Decentralized, Distributed

The conceptual differences between a centralized, decentralized, and distributed processing architecture is depicted in Figure 2.1. In a centralized architecture, a single node is processing data that is obtained by all nodes, meaning that a star-based communication graph rooted at the central node is required. In a decentralized architecture, multiple nodes can share the workload to process the information. A distributed architecture, distributes the workload among all nodes and potential data interdependencies might require a denser communication mesh, in the worst case, a fully-meshed communication graph. Distributed architectures tend to be more complex and require strategies to distribute and handle information exchange, while at the same time, they tend to scale well, are resilient to single point of failures, and allow distributing the payload between available nodes.

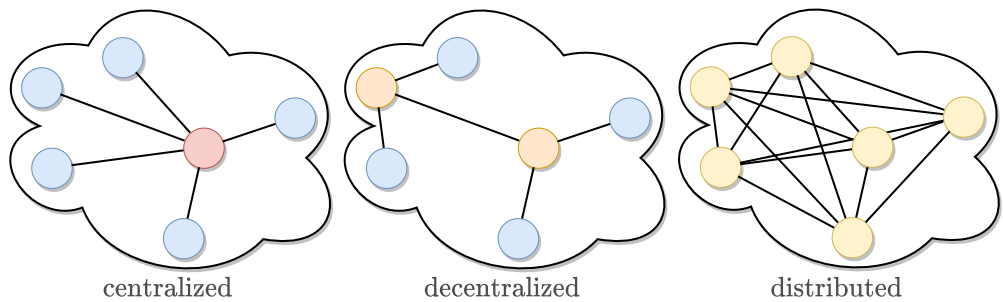

**Figure 2.1:** Shows the topological differences between a centralized, decentralized and distributed processing network, while nodes held in non-blue colors, are performing computation. The processing effort is indicated by red, orange, yellow, representing high to low, respectively.

### Agent

An agent can be something or someone with processing and communication capabilities that is able to perform mission specific tasks, e.g., a ground robot, Unmanned Aerial Vehicle (UAV), MAV, humanoid, Autonomous Underwater Vehicle (AUV), human equipped with a sensor suite, autonomous vehicle, etc..

### Swarm

We define swarm as a set $\mathsf{A} := \{\mathsf{A}_i | i = 1, \dots, N\}$ of agents $\mathsf{A}$ with communication and processing capabilities in a wireless network. Each agent is associated with a unique identifier (ID), $\mathsf{id}_\mathsf{A}$, allowing us the following notation $\mathsf{A}_i = \mathsf{A}\left(\mathsf{id}_{\mathsf{A}_i}\right)$. The swarm consists of $N = |\mathsf{A}|$ agents ($|\bullet|$ is the cardinality of a set).

A set of *met* agents $\mathsf{M}_i$ refers to the experience of an individual agent $\mathsf{A}_i$ and is defined as $\mathsf{M}_i := \{\mathsf{M}_j | j = 1, \dots, M_i\}$.

### Sliding history container

A sliding history container $\mathsf{Hist}$ is a smart cyclic (memory) buffer that allows to hold elements within a certain time span. Elements are inserted chronologically ascending sorted (like a FIFO that is considering the timestamp rather than the actual order). The time span is referred from most recent to the oldest element, once elements exceed the time-horizon they are removed automatically. Each entry is a tuple $< t^k, \bullet^k >$ of the associated timestamp and the actual element, e.g., $\mathcal{H} := \mathsf{Hist}\{\mathsf{x}^k\}$. Elements are inserted chronologically sorted into the buffer by $\mathsf{Hist}(t^k) = \bullet^k$ and accessed by $\bullet^k = \mathsf{Hist}(t^k)$.

### Dictionary container

A dictionary container – a so-called hashed-map– Dict, is (memory) buffer that is indexed via a unique identifier, meaning that it maintains a set of indices and a set of elements (a set of tuples $< \text{id}, \bullet >$), e.g., $\mathcal{M} := \text{Dict}\{\mathbf{v}\}$. Elements are inserted into the dictionary by $\text{Dict}(\text{id}) = \bullet^k$ and accessed by $\bullet^k = \text{Dict}(\text{id})$. A list $\mathbf{l}$ of unique identifiers is obtained by $\mathbf{l} = \text{Dict} \rightarrow \text{keys}()$

### Real-time capability

Regarding real-time capability, we can distinguish between hard and soft real-time. Depending on which type is needed and if the condition/deadline (the time instant at which the data should be process is called deadline) are met, the system is real-time capable. Soft real-time, means that no catastrophic failure modes are issued if the time constraint is not met, and that results are still valid/usable after the deadline. Hard real-time is required when the deadline is firm and critical state can be reached by missing the deadline. A hard real-time system must therefore provide guarantees on the temporal behaviors under all possible conditions [82].

## 2.3  Spatial Geometry

In Collaborative Localization (CL), the ability to communicate and the measure quantities of the spatial relations between robots/agent is exploited to improve the localization. Therefore, a neat notation for expressing spatial relations is required to avoid ambiguities.

### 2.3.1  Representing Poses

In general, a pose describe the position (or translation) between the origins of two frames of reference, and the orientation between the orthogonal axes of the coordinate reference frames. Poses are elements of the *Special Euclidean* group $SE(N)$ of dimension $N$,

In 3D space, a pose $\mathbf{T} \in SE(3)$ between the reference frames $\{\mathcal{A}\}$ and $\{\mathcal{B}\}$ (distance ${}^{\mathcal{A}}_{\mathcal{A}}\mathbf{p}_{\mathcal{B}}$ and orientation ${}^{\mathcal{A}}\mathbf{R}_{\mathcal{B}}$) is realized as a so-called homogeneous transformation matrix, forming a matrix Lie group, in the form of

$$ {}^{\mathcal{A}}\mathbf{T}_{\mathcal{B}} \in \text{SE}(3) := \left\{ \begin{bmatrix} {}^{\mathcal{A}}\mathbf{R}_{\mathcal{B}} & {}^{\mathcal{A}}_{\mathcal{A}}\mathbf{p}_{\mathcal{B}} \\ \mathbf{0}^{\mathsf{T}} & 1 \end{bmatrix} \middle| \ \mathbf{R} \in SO(3), \mathbf{p} \in \mathbb{R}^3 \right\}. \tag{2.5}$$

We use $(\mathbf{T})_{\{x,y,z,\mathbf{p},\mathbf{R}\}}$ and $(\mathbf{p})_{\{x,y,z\}}$ to access the corresponding elements, e.g., ${}^{\mathcal{A}}\mathbf{R}_{\mathcal{B}} = \left( {}^{\mathcal{A}}\mathbf{T}_{\mathcal{B}} \right)_{\mathbf{R}}$.

The composition of a homogeneous transformation matrices is ${}^{\mathcal{A}}\mathbf{T}_{\mathcal{C}} = {}^{\mathcal{A}}\mathbf{T}_{\mathcal{B}}{}^{\mathcal{B}}\mathbf{T}_{\mathcal{C}}$ and the transformation of a coordinate vector ${}^{\mathcal{C}}_{\mathcal{C}}\mathbf{p}_{P_1}$ pointing from the origin of the reference frame $\mathcal{C}$ to a point $P_1$, expressed in $\{\mathcal{C}\}$, can be transformed into the frame $\{\mathcal{A}\}$ by the homogeneous vector representation

$$ \begin{bmatrix} {}^{\mathcal{A}}_{\mathcal{A}}\mathbf{p}_{P_1} \\ 1 \end{bmatrix} = {}^{\mathcal{A}}\mathbf{T}_{\mathcal{C}} \begin{bmatrix} {}^{\mathcal{C}}_{\mathcal{C}}\mathbf{p}_{P_1} \\ 1 \end{bmatrix}. \tag{2.6}$$

The inverse of a homogeneous transformation matrix ${}^{\mathcal{A}}\mathbf{T}_{\mathcal{B}}^{-1}$ is

$$ {}^{\mathcal{A}}\mathbf{T}_{\mathcal{B}}^{-1} = \begin{bmatrix} {}^{\mathcal{A}}\mathbf{R}_{\mathcal{B}}^{\mathsf{T}} & -{}^{\mathcal{A}}\mathbf{R}_{\mathcal{B}}^{\mathsf{T}}{}^{\mathcal{A}}_{\mathcal{A}}\mathbf{p}_{\mathcal{B}} \\ \mathbf{0}^{\mathsf{T}} & 1 \end{bmatrix} \in \text{SE}(3) \tag{2.7}$$

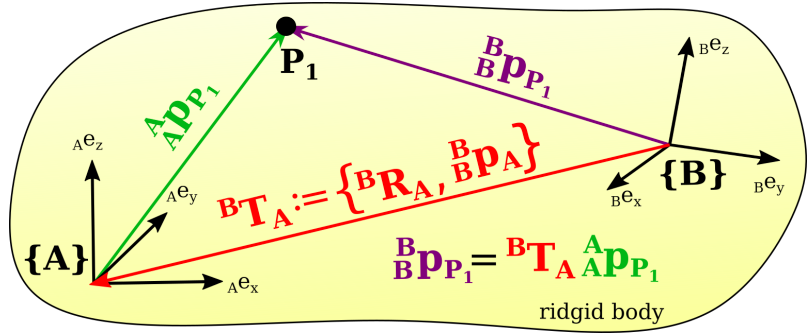

**Figure 2.2:** Referencing scheme: A point $P_1$ expressed in the reference frame $\{\mathcal{A}\}$ can be expressed in the reference frame $\{\mathcal{B}\}$ by applying the homogeneous transformation matrix ${}^{\mathcal{B}}\mathbf{T}_{\mathcal{A}}$.

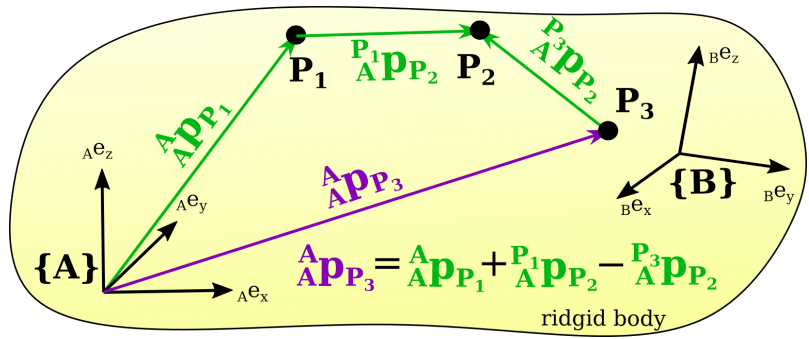

**Figure 2.3:** Referencing scheme: Three points $\{P_1, P_2, P_3\}$ on a rigid body expressed in the reference frame $\{\mathcal{A}\}$ are combined/concatenated to express a sum between the origin of $\{\mathcal{A}\}$ and the point $P_3$.

The reference scheme for vectors is read as

$$\substack{\text{from} \\ \text{in}}\{\bullet\}_{\text{to}}, \tag{2.8}$$

where *in* is the reference frame the element is expressed in, *from* is the tail of the vector, and *to* is the tip of the vector. An example is shown in Figure 2.2.

At this point, one can argue that the *in* specifier is superfluous, but that the tail of a vector is not necessary in the origin of the reference frame it is expressed in, e.g., the tail could be located in a different reference frame or another point of the rigid body configuration, as shown in Figure 2.3. The direction of a vector is changed by negating the sign

$$\substack{P_1 \\ \mathcal{A}}\mathbf{p}_{\mathcal{A}} = (-1)\substack{\mathcal{A} \\ \mathcal{A}}\mathbf{p}_{P_1}. \tag{2.9}$$

The referencing scheme for orientations and poses allows mitigating the *in* specifier as they always refer to the origin of their coordinate reference frame, as shown in Figure 2.2,

$$\substack{\text{from}}\{\bullet\}_{\text{to}}. \tag{2.10}$$

The time index (*at*) is located at the right superscript

$$\substack{\text{from} \\ \text{in}}\{\bullet\}\substack{\text{at} \\ \text{to}}. \tag{2.11}$$

Alternatively, the time index can be added to the right superscript of the reference frame, e.g., the point $\mathcal{P}_1$ expressed in the reference frame $\{\mathcal{A}\}$ at time $t(k)$ can be expressed as

$$\substack{\mathcal{A} \\ \mathcal{A}}\mathbf{p}_{P_1}^{k} = \substack{\mathcal{A} \\ \mathcal{A}}\mathbf{p}_{P_1^k}, \tag{2.12}$$

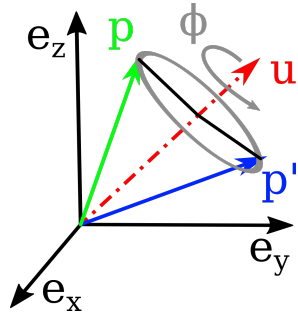

**Figure 2.4:** Rotation of a vector $\mathbf{p}$, by an angle $\phi$, around the axis $\mathbf{u}$.

and simplifies the notation to express spatial-temporal relations between reference frames.

### 2.3.2 Representing Rotations

The attitude of an Inertial Navigation System (INS) in 3D space can be represented in various ways. Prominent realization are rotations matrices – matrix Lie group – of the special orthogonal group $SO(3)$ or as unit quaternions as elements of the Hamiltonian space $\mathbb{H}$ with unit length. There are other representations such as *Axis-Angle*, and *Euler-angles*.

In this subsection, we briefly review these representations and their properties, in order to avoid ambiguities in definitions, derivations, and evaluations performed in the thesis. Generally, rotations are not commutative, meaning the order and sequence of consecutive rotations matters.

#### Axis–Angle Rotation

The *Axis–angle representation* describes a rotation in the 3D space about the axis of a unit vector $\mathbf{u} = \begin{bmatrix} u_x; u_y; u_z \end{bmatrix}$ with $\|\mathbf{u}\| = 1$, through the origin by an angle of $\phi$, as shown in Figure 2.4. This representation plays a major role as distance metric, as a meta parameterization in the conversion between different other representation, and is the Cartesian vector representation $\boldsymbol{\phi} = \phi \mathbf{u} \in \mathbb{R}^3$ of the Lie algebra $[\boldsymbol{\phi}]_\times \in \mathfrak{so}(3)$. For a constant angular velocity $\boldsymbol{\omega}$ it can be used to obtain the integrated rotation in the axis angle form $\boldsymbol{\phi} = \boldsymbol{\omega} t$ [137]. Note that $\phi = \frac{\phi}{\|\phi\|}$ and $\mathbf{u} = \frac{\phi}{\phi}$, satisfying $\|\mathbf{u}\| = 1$.

This parameterization suffers from a lacking composition operator, meaning that a sequence of rotation cannot be computed easily, as the *group axiom* is not satisfied (please refer to [137])

#### Euler-Angles

The *Euler's rotation theorem*, states that any orientation of a rigid body with respect to a fixed three-dimensional Cartesian coordinate system can be presented with three proper elemental rotations actions about different axis [26], e.g.,

$$\mathbf{R} = \mathbf{R}_{zyx}(\psi, \theta, \phi) = \mathbf{R}_z(\psi) \mathbf{R}_y(\theta) \mathbf{R}_x(\phi) \in SO(3), \tag{2.13}$$

about the x-, y-, and z-axis[2] with

$$\mathbf{R}_x(\phi) = \begin{pmatrix} 1 & 0 & 0 \\ 0 & \cos\phi & -\sin\phi \\ 0 & \sin\phi & \cos\phi \end{pmatrix}, \tag{2.14a}$$

---

[2]Note that a vector $\mathbf{p}$ is transformed by the right most matrix first, in this case $\mathbf{R}_x$, $\mathbf{p}' = \mathbf{R}_z(\psi) \mathbf{R}_y(\theta) \mathbf{R}_x(\phi) \mathbf{p}$

$$\mathbf{R}_y(\theta) = \begin{pmatrix} \cos\theta & 0 & \sin\theta \\ 0 & 1 & 0 \\ -\sin\theta & 0 & \cos\theta \end{pmatrix}, \tag{2.14b}$$

$$\mathbf{R}_z(\psi) = \begin{pmatrix} \cos\psi & -\sin\psi & 0 \\ \sin\psi & \cos\psi & 0 \\ 0 & 0 & 1 \end{pmatrix}. \tag{2.14c}$$

In total twelve possible sequences of rotation about the axes exist to a achieve a dedicated rotation, which are divided into two groups the *Eulerian* and the *Cardanian*[3] angles [26]:

- Eulerian angles {z-x-z, x-y-x, y-z-y, z-y-z, x-z-x, y-x-y} (2 distinct axes),
- Cardanian angles {x-y-z, y-z-x, z-x-y, x-z-y, y-x-z, z-y-x} (3 distinct axes).

This means, that each representation leads to different rotation matrices and special care needs to be taken if existing third-party source code is used. Representing a system state in terms Euler angles can lead, in case of a so-called Gimbal lock, to a loss of a degree of freedom in the orientation, meaning that changes in two parameters effect in a rotation about the same axis. For example, if the state of the pitch is $\beta = \frac{\pi}{2}$, then modifying either the roll or yaw angle results in a rotation about the z-axis.

Like the axis-angle representation, Euler-angles are lacking of a proper composition operator, meaning that the composition of angles about more than one axis does not relate to physical rotation of applying the individual rotations sequentially.

Finally, the inverse of a rotation matrix to Euler-angles does not provide a unique solution and the visualization of continues or error angles needs to mitigate the effect of discontinuities due to overflows ($\phi = \phi + 2N\pi$)

Due to their intuitive way of representing rotations, we use Euler angles for visualizing the estimation performance in 2D plots, but they should be avoided as error metric, see Section 2.6. Throughout the thesis, we use the *Cardanian* angles following the z-y-x sequence.

### Unit Quaternions

Unit quaternions are a commonly used representation for 3D rotations in robotics and, in our case, to estimate the attitude of the INS. A quaternions is described by four parameters, a real scalar $q_w$ and three scalars $\mathbf{q}_v = [q_x; q_y; q_z]$ for the imaginary parts [136]

$$\mathbf{q} = q_w + q_x i + q_y j + q_z k = q_w + \mathbf{q}_v = \begin{bmatrix} q_w \\ \mathbf{q}_v \end{bmatrix} = \begin{bmatrix} q_w \\ q_x \\ q_y \\ q_z \end{bmatrix} \in \mathbb{H}. \tag{2.15a}$$

The Hamiltonian product (with the $\otimes$ operator) of two quaternions $\{\mathbf{p}, \mathbf{q}\} \in \mathbb{H}$ is defined as

$$\mathbf{p} \otimes \mathbf{q} = \begin{bmatrix} p_w q_w - \mathbf{p}_v^\mathsf{T} \mathbf{q}_v \\ p_w \mathbf{q}_v + q_w \mathbf{p}_v + \mathbf{p}_v \left[\mathbf{q}_v\right]_\times \end{bmatrix}, \tag{2.15b}$$

which is not commutative

$$\mathbf{p} \otimes \mathbf{q} \neq \mathbf{q} \otimes \mathbf{p}, \tag{2.15c}$$

but associative

$$(\mathbf{p} \otimes \mathbf{q}) \otimes \mathbf{r} = \mathbf{p} \otimes (\mathbf{q} \otimes \mathbf{r}). \tag{2.15d}$$

---

[3]Which are also known as *Tait-Brayan angles*

The identity quaternion is

$$\mathbf{q_I} = \begin{bmatrix} 1 \\ \mathbf{0} \end{bmatrix}, \tag{2.15e}$$

the conjugate is defined by the symbol $\{\bullet\}^*$ as

$$\mathbf{q}^* = \begin{bmatrix} q_w \\ -\mathbf{q}_v \end{bmatrix}, \tag{2.15f}$$

and has – similar to the matrix transpose –, the following property

$$(\mathbf{p} \otimes \mathbf{q})^* = \mathbf{q}^* \otimes \mathbf{p}^*. \tag{2.15g}$$

Unit quaternions satisfy the unit length

$$\bar{\mathbf{q}} \in \{\bar{\mathbf{q}} \in \mathbb{H} \,|\, \|\bar{\mathbf{q}}\| = 1\}, \tag{2.15h}$$

with

$$\|\mathbf{q}\| = \sqrt{\mathbf{q} \otimes \mathbf{q}^*} = \sqrt{q_w^2 + q_x^2 + q_y^2 + q_z^2} \in \mathbb{R}. \tag{2.15i}$$

The inverse of a unit quaternion is its conjugate [136].

$$\bar{\mathbf{q}}^{-1} = \bar{\mathbf{q}}^*. \tag{2.15j}$$

Since unit quaternions constitutes a double cover of $SO(3)$,

$$\bar{\mathbf{q}} = \begin{bmatrix} q_w \\ -\mathbf{q}_v \end{bmatrix} = \begin{bmatrix} -q_w \\ -\mathbf{q}_v \end{bmatrix}, \tag{2.15k}$$

they are element of a hyper-sphere $\bar{\mathbf{q}} = \in S^3$. A unit-quaternion can be obtained from the axis-angle representation by

$$\bar{\mathbf{q}}(\mathbf{u}, \phi) = \begin{bmatrix} \cos(\phi/2) \\ \mathbf{u}\sin(\phi/2) \end{bmatrix}, \ \|\mathbf{u}\| = 1. \tag{2.15l}$$

The composition of quaternions is computationally more efficient compared to rotation matrices. Rotation matrices consist of nine parameters meaning that a matrix multiplication requires $9 \times 3$ multiplications and $9 \times 2$ additions. Quaternions consist of four parameters, while a multiplication requires $4 \times 4$ scalar multiplications and $4 \times 3$ scalar additions.

There are two different quaternion definitions, the Hamiltonian and JPL/Shuster quaternions [139], requiring to interpret the 4 parameters differently. Further, a different multiplication operator and mapping operator between quaternions and rotation matrices are needed. Since most software libraries and frameworks (Eigen[4], ROS[5], MATLAB[6]) are using Hamilton quaternions, special care needs to be taken if state values of JPL-frameworks are used directly, since a conversion needs to done.

### Rotation Matrix

A rotation matrix $\mathbf{R} \in SO(3) := \{\mathbf{R} \in \mathbb{R}_3 | \mathbf{R}^\mathsf{T}\mathbf{R} = \mathbf{I}, \det(\mathbf{R}) = 1\}$ belongs to the special orthogonal group of order three. Consequently, a rotation matrix fulfills the following two conditions:

---

[4] https://eigen.tuxfamily.org/

[5] https://www.ros.org/

[6] https://de.mathworks.com/

- **special:** $\det(\mathbf{R}) = +1$ (to avoid mirror images),
- **orthogonal:** $\mathbf{I} = \mathbf{R}\mathbf{R}^\top = \mathbf{R}^\top \mathbf{R}$.

The column vectors in rotation matrices have an intuitive underlying geometric interpretation. Given an Euclidean vector ${}^{\mathcal{B}}_{\mathcal{B}}\mathbf{p}_{P_1}$ in the reference frame $\mathcal{B}$ pointing from the origin to a point $P_1$, can be decomposed into vectors along the orthogonal axes ${}^{\mathcal{B}}_{\mathcal{B}}\mathbf{e}_{\{x,y,z\}}$ with unit length

$$
{}^{\mathcal{B}}_{\mathcal{B}}\mathbf{p}_{P_1} = {}^{\mathcal{B}}_{\mathcal{B}}\mathbf{e}_x x + {}^{\mathcal{B}}_{\mathcal{B}}\mathbf{e}_y y + {}^{\mathcal{B}}_{\mathcal{B}}\mathbf{e}_z z, \quad \left\| {}^{\mathcal{B}}_{\mathcal{B}}\mathbf{p}_{P_1} \right\| = \sqrt{x^2 + y^2 + z^2}. \tag{2.16}
$$

Given a rotation matrix ${}^{\mathcal{A}}\mathbf{R}_{\mathcal{B}}$, describing the orientation between a reference frame $\mathcal{A}$ and $\mathcal{B}$, then the columns are in the form

$$
{}^{\mathbf{A}}\mathbf{R}_{\mathbf{B}} = \begin{bmatrix} {}^{\mathcal{B}}_{\mathcal{A}}\mathbf{e}_x & {}^{\mathcal{B}}_{\mathcal{A}}\mathbf{e}_y & {}^{\mathcal{B}}_{\mathcal{A}}\mathbf{e}_z, \end{bmatrix} \tag{2.17}
$$

with the orthogonal axes ${}^{\mathcal{A}}_{\mathcal{B}}\mathbf{e}_{\{x,y,z\}}$ of the frame $\mathcal{B}$ expressed in the reference frame $\mathcal{A}$. This allows to express the point $P_1$ w.r.t. to the reference frame $\mathcal{A}$ by ${}^{\mathcal{B}}_{\mathcal{A}}\mathbf{p}_{P_1} = {}^{\mathcal{A}}\mathbf{R}_{\mathcal{B}}{}^{\mathcal{B}}_{\mathcal{B}}\mathbf{p}_{P_1}$, which is equivalent to rotating/transforming the bases

$$
{}^{\mathcal{B}}_{\mathcal{A}}\mathbf{p}_{P_1} = {}^{\mathcal{A}}\mathbf{R}_{\mathcal{B}}{}^{\mathcal{B}}\mathbf{e}_x x + {}^{\mathcal{A}}\mathbf{R}_{\mathcal{B}}{}^{\mathcal{B}}\mathbf{e}_x y + {}^{\mathcal{A}}\mathbf{R}_{\mathcal{B}}\mathbf{e}_x z, \tag{2.18}
$$

We must ensure that the transformed model/vector does not become distorted, which is achieved by ensuring that a rotation matrix satisfies the following criteria [85]:

- No stretching of axes: The norm of the columns must have the unit length.
- No shearing: The inner (dot) product is zero.
- No mirror image: The determinant must be positive.

Further properties of rotation matrices are:

- The transformation does not change the amplitude of the vector: $\|\mathbf{R}\mathbf{a}\| = \|\mathbf{a}\|$
- The inner product, and then the angle between two vectors, is invariant with respect to rotations: $\mathbf{a}^\top \mathbf{b} = (\mathbf{R}\mathbf{a})^\top (\mathbf{R}\mathbf{b})$
- Since $\mathbf{R}$ is an orthogonal matrix, the following property holds: $\mathbf{R}(\mathbf{a} \times \mathbf{b}) = (\mathbf{R}\mathbf{a}) \times (\mathbf{R}\mathbf{b})$
- The product of rotation matrices does not commute: $\mathbf{R}_a \mathbf{R}_b \neq \mathbf{R}_b \mathbf{R}_a$
- Negating the angle about the rotation axis is equal to the inverse of the corresponding rotation matrix: $\mathbf{R}(\mathbf{u}, -\phi) = \mathbf{R}(\mathbf{u}, \phi)^\top$
- Double representation $\mathbf{R}(\mathbf{u}, \phi) = \mathbf{R}(-\mathbf{u}, 2\pi - \phi)$
- Two consecutive rotations about the rotation axis is equal to a product of two rotations: $\mathbf{R}(\mathbf{u}, \phi_1 + \phi_2) = \mathbf{R}_{\phi_2} \mathbf{R}_{\phi_1}$

Manipulating rotation matrices in numerical implementations, will not preserve this structure, e.g., the determinant of the numerical solutions is not equal to one. A common solution is to project the ordinary matrix back onto the manifold. This projection is in case of rotation matrices more complex (e.g.,based on a singular value decomposition) than for unit quaternions – which can be achieved by the quaternion normalization) – and gives no guarantee that the projected result is physically meaningful.

## 2.4 Estimation Error definitions

### 2.4.1 Error types

In this thesis, we study CSE primarily on a widely used indirect or error-state formulation of a Kalman filter, meaning that the definition of the estimation error plays a crucial role in the state correction and the linearization of the underlying nonlinear functions. In this section, we will show that different choices can be made and that it is important to explicitly state which error definitions was chosen.

For Lie Groups, the following group axioms need to be fulfilled: closure under composition ($\boxplus$), identity, inverse, and associativity [137]. Lie Groups $\mathcal{G}$ are most likely not commutative (e.g., the rotation matrix realization of $SO(3)$), thus the order of compositing matters. Having two elements of a group, the error/perturbation $\sim$ and the estimate $\wedge$, and a composition operator ($\boxplus$), naturally two compositions arise to define the true value $\top$:

$$\mathcal{G} \text{ type-1 error}: \top = \wedge \boxplus \sim_1 \quad \text{(local pertubations)} \tag{2.19}$$

$$\mathcal{G} \text{ type-2 error}: \top = \sim_2 \boxplus \wedge \quad \text{(global pertubations)} \tag{2.20}$$

One can distinguish between local and global perturbations, depending on *where* the error is located, as shown in Figure 2.5. A global perturbations is in the origin or global frame, while the local error is located in the origin of the local/body frame. This interpretation stem from transforming vectors, e.g., $^{\mathcal{G}}_{\mathcal{G}}\mathbf{p}_{P_1} = {}^{\mathcal{G}}\mathbf{R}_{\mathcal{L}}{}^{\mathcal{L}}_{\mathcal{L}}\mathbf{p}_{P_1}$, assuming in this example, that the origins of $\{\mathcal{G}\}$ and $\{\mathcal{L}\}$ are in the same place.

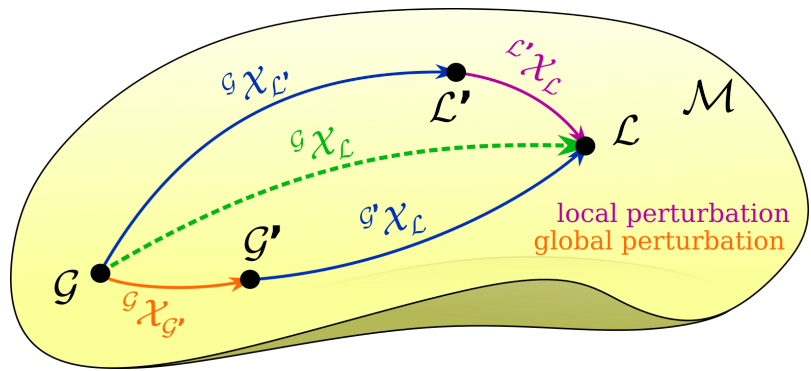

**Figure 2.5:** Two paths on a manifold $\mathcal{M}$ to join the origin $\{\mathcal{G}\}$ with the element $\{\mathcal{L}\}$. Both compose an element $\hat{\mathcal{X}} = {}^{\mathcal{G}}\mathcal{X}_{\mathcal{L}'} = {}^{\mathcal{G}'}\mathcal{X}_{\mathcal{L}}$ with increments ${}_{\mathcal{G}}\tilde{\mathcal{X}} = {}^{\mathcal{G}}\mathcal{X}_{\mathcal{G}'}$ and ${}_{\mathcal{L}'}\tilde{\mathcal{X}} = {}^{\mathcal{L}'}\mathcal{X}_{\mathcal{L}}$, expressed either in and the local frame $\{\mathcal{L}'\}$ or in the origin $\{\mathcal{G}\}$. Due to non-cummutativity the elements ${}_{\mathcal{G}}\tilde{\mathcal{X}}$ and ${}_{\mathcal{L}'}\tilde{\mathcal{X}}$ are not equal. Figure is inspired by Solà *et al.* [137].

From vector algebra, one is used to have a $+$ and $-$ operator, which can be interpreted as $\top = \wedge \boxminus \sim = \wedge \boxplus (\sim)^{-1}$, meaning that the minus operator $\boxminus$ is the composition with the inverse of an element. By considering all possible paths, eight different error definitions can be found, as shown in Figure 2.6, while four definitions – with the estimate in the opposite direction of the path – are rather counterintuitive. This figure reveals that four invariant error definitions (with the estimate in the direction of the forward) exist, despite in literature on invariant Kalman filtering, only one left- and right invariant error definition is mentioned, and the filter formulation is defined according to them, see e.g., [11, 55].

This also allows introducing eight error definitions (four parameters and non-commutative

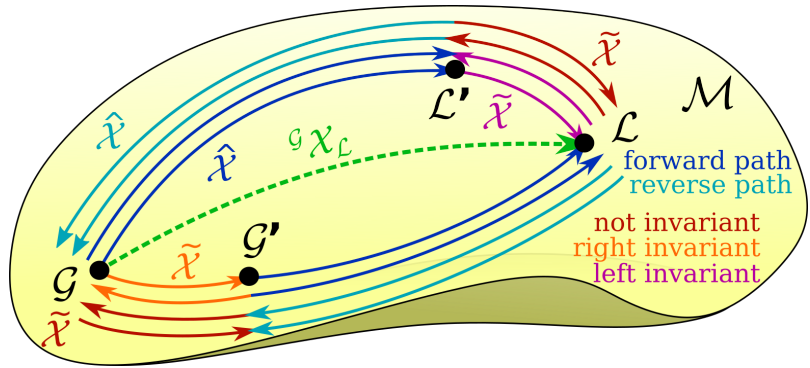

**Figure 2.6:** All possible paths on a manifold $\mathcal{M}$ to join the origin $\{\mathcal{G}\}$ with an element $\{\mathcal{L}\}$, by composing an element $\hat{\mathcal{X}}$ with an increment $\tilde{\mathcal{X}}$. Having estimates in the forward path ( $\hat{\mathcal{X}}$ held in blue), the increments or errors are either left- or right invariant, held in magenta and orange respectively. Defining estimates in the opposite direction ($\hat{\mathcal{X}}$ held in cyan) does not lead to invariant errors, thus, in total, four invariant error definitions are possible on Lie groups.

compositions)

$$\mathcal{G} \text{ type-3 error} : \top = \wedge \boxminus \sim_3 = \wedge \boxplus \sim_3^{-1} \tag{2.21}$$

$$\mathcal{G} \text{ type-4 error} : \top = \boxminus \sim_4 \boxplus \wedge = \sim_4^{-1} \boxplus \wedge \tag{2.22}$$

$$\mathcal{G} \text{ type-5 error} : \top = \boxminus \wedge \boxplus \sim_5 = \wedge^{-1} \boxplus \sim_5 \tag{2.23}$$

$$\mathcal{G} \text{ type-6 error} : \top = \sim_6 \boxminus \wedge = \sim_6 \boxplus \wedge^{-1} \tag{2.24}$$

$$\mathcal{G} \text{ type-7 error} : \top = \boxminus \wedge \boxminus \sim_7 = \wedge^{-1} \boxplus \sim_7^{-1} \tag{2.25}$$

$$\mathcal{G} \text{ type-8 error} : \top = \boxminus \sim_8 \boxminus \wedge = \sim_8^{-1} \boxplus \wedge^{-1} \tag{2.26}$$

Please note that the error types $\sim_5$ to $\sim_8$ based on the reverse path of the estimate are counterintuitive with respect to real estimation problems, e.g., estimating the pose of an object in space, and is thus not further considered.

As shown in Figure 2.6, and four invariant error types for Lie groups $\mathcal{G}$ exist

$$\mathcal{G} \text{ type-1 error} : \sim_1 = \wedge^{-1} \boxplus \top = (\mathbf{L}\wedge)^{-1} \boxplus (\mathbf{L}\top) \text{ (left invariant)} \tag{2.27}$$

$$\mathcal{G} \text{ type-2 error} : \sim_2 = \top \boxplus \wedge^{-1} = (\top\mathbf{R}) \boxplus (\wedge\mathbf{R})^{-1} \text{ (right invariant)} \tag{2.28}$$

$$\mathcal{G} \text{ type-3 error} : \sim_3 = \top^{-1} \boxplus \wedge = (\mathbf{L}\top)^{-1} \boxplus (\mathbf{L}\wedge) \text{ (left invariant)} \tag{2.29}$$

$$\mathcal{G} \text{ type-4 error} : \sim_4 = \wedge \boxplus \top^{-1} = (\wedge\mathbf{R}) \boxplus (\top\mathbf{R})^{-1} \text{ (right invariant)} \tag{2.30}$$

with $\mathbf{L}$ and $\mathbf{R}$ as elements of the group $\mathcal{G}$. An error classified as left or right invariant, if an arbitrary element ´of the group $\mathcal{G}$ can be multiplied left or right on both the estimate $\wedge$ and true $\top$, without changing the error $\sim$. For instance, assuming $\{\wedge, \top, \mathbf{L}\}$ being elements of a matrix Lie group

$$\sim_1 = (\mathbf{L}\wedge)^{-1}(\mathbf{L}\top) = (\wedge^{-1}\mathbf{L}^{-1})(\mathbf{L}\top) = \wedge^{-1}\top. \tag{2.31}$$

There are two left- and right invariant errors, whereas in literature regarding invariant Kalman filtering, e.g., Barrau and Bonnabel in [11], Hartley *et al.* in [55], and Yang *et al.* [157], only the left $\sim_2$ and right invariant error $\sim_4$ were defined. Later in [12], Barrau and Bonnabel defined the left invariant $\sim_1$ and right invariant $\sim_2$ error.

In the vector space $\mathbb{R}^N$, which is commutative $(-a + b = b - a)$, the composition operator is + and allows defining the following error types

$$\mathbb{R}^N \text{ type-1 error} : \sim_1 = - \wedge + \top = \top - \wedge \tag{2.32}$$

$$\mathbb{R}^N \text{ type-2 error} : \sim_2 = \top + (-\wedge) = \top - \wedge = \sim_1 \tag{2.33}$$

$$\mathbb{R}^N \text{ type-3 error} : \sim_3 = (-\top) + \wedge = \wedge - \top \tag{2.34}$$

$$\mathbb{R}^N \text{ type-4 error} : \sim_4 = \wedge + (-\top) = \wedge - \top = \sim_3 \tag{2.35}$$

Meaning that the four invariant error types from a general Lie Group $\mathcal{G}$, reduced to two error types in vector space $\mathbb{R}^N$.

### 2.4.2 Pose error definitions

In this section we define the invariant error definition of a 6-DoF pose $^{\mathcal{G}}\mathbf{T}_{\mathcal{L}} \in SE(3)$

$$SE(3) \text{ type-1 error} : \ ^{\mathcal{L}'}\tilde{\mathbf{T}}_{\mathcal{L}} = \ ^{\mathcal{G}}\hat{\mathbf{T}}_{\mathcal{L}'}^{-1} \boxplus \ ^{\mathcal{G}}\mathbf{T}_{\mathcal{L}} \text{ (left invariant)} \tag{2.36}$$

$$SE(3) \text{ type-2 error} : \ ^{\mathcal{G}}\tilde{\mathbf{T}}_{\mathcal{G}'} = \ ^{\mathcal{G}}\mathbf{T}_{\mathcal{L}} \boxplus \ ^{\mathcal{G}'}\hat{\mathbf{T}}_{\mathcal{L}}^{-1} = \text{ (right invariant)} \tag{2.37}$$

$$SE(3) \text{ type-3 error} : \ ^{\mathcal{L}}\tilde{\mathbf{T}}_{\mathcal{L}'} = \ ^{\mathcal{G}}\mathbf{T}_{\mathcal{L}}^{-1} \boxplus \ ^{\mathcal{G}}\hat{\mathbf{T}}_{\mathcal{L}'} = \text{ (left invariant)} \tag{2.38}$$

$$SE(3) \text{ type-4 error} : \ ^{\mathcal{G}'}\hat{\mathbf{T}}_{\mathcal{G}} = \ ^{\mathcal{G}'}\hat{\mathbf{T}}_{\mathcal{L}} \boxplus \ ^{\mathcal{G}}\mathbf{T}_{\mathcal{L}}^{-1} = \text{ (right invariant)} \tag{2.39}$$

## 2.5 Distance Metrics

In order to measure an error, we need to measure a distance between the true and estimated value. Note, as already mentioned in Section 2.4, there many ways to define an error, especially if values are not commutative.

### 2.5.1 Euclidean distance

In the Euclidean space, a bi-invariant metric is the $l^2$-norm of the difference between two vectors

$$d_E(\mathbf{p}_1, \mathbf{p}_2) = \|\mathbf{p}_1 - \mathbf{p}_2\|_2 \,, \tag{2.40}$$

or

$$d_E(\tilde{\mathbf{p}}) = \|\tilde{\mathbf{p}}\|_2 \tag{2.41}$$

### 2.5.2 Angle distance

Various metrics to determine a difference/distance between two 3D rotations exist [68]. In 2D on elements of $S(1)$, the unit circle, the difference between rotations is the absolute value between two angles in radians. Naively in 3D space, one might think the metric on $S(1)$ can be applied on the in individual axis using the Euler-Angle representation of two 3D rotations. First, the Euler angle representations do not satisfy the group axioms. Second, this would lead to three values that need to be weighted in some way.

The absolute difference between 3D rotations, $\{\mathbf{R}_1, \mathbf{R}_2\} \in SO(3)$ corresponds to the angle $\phi$ from the axis-angle representation as a bi-invariant metric [159] in the form

$$d_\mathbf{R} = d_\mathbf{R}(\mathbf{R}_1, \mathbf{R}_2) = \phi = \cos^{-1}\left(\frac{\text{tr}(\mathbf{R_2}^\top \mathbf{R}_1) - 1}{2}\right), \tag{2.42}$$

or

$$d_\mathbf{R}(\tilde{\mathbf{R}}) = \cos^{-1}\left(\frac{\text{tr}(\tilde{\mathbf{R}}) - 1}{2}\right), \tag{2.43}$$

where $\mathbf{R}^{-1} = \mathbf{R}^\top$ and describes the length of the geodesic on the unit sphere in radians.

### 2.5.3   Mahalanobis distance

Mahalanobis introduced in [98], a metric to measure a distance between a sample point $\mathbf{p}$ and a Gaussian normal distribution $\mathbf{x} \sim \mathcal{N}(\hat{\mathbf{x}}, \boldsymbol{\Sigma})$. The Mahalanobis distance allows determining how many standard deviations the given point $\mathbf{p}$ is away from the expected mean value of the distribution $\hat{\mathbf{x}}$ and is, thus, unit less and scale invariant

$$d_M = d_M(\mathbf{p}, \mathbf{x}) = d_M(\mathbf{p}, \hat{\mathbf{x}}, \boldsymbol{\Sigma}) = \sqrt{(\mathbf{p} - \hat{\mathbf{x}})^\mathsf{T} \boldsymbol{\Sigma}^{-1} (\mathbf{p} - \hat{\mathbf{x}})} = \|\mathbf{p} - \hat{\mathbf{x}}\|_{\boldsymbol{\Sigma}}. \tag{2.44}$$

Since $\boldsymbol{\Sigma}$ is positive semidefinite, the inverse is always defined.

## 2.6   Estimation Error Metrics

To determine the (estimation) error, we use according to the parameter space an appropriate measure for distance Section 2.5 between an estimated and the actual (true) value. For elements in Euclidean space $\mathbb{R}^3$ we use the Euclidean distance and for elements describing rotations we use the angle or Euler-angles.

### 2.6.1   Root Mean Square Error (RMSE)

The RMSE at every time step $k$ of given a set of $M$ estimated trajectories and the true trajectory can be computed given an error-definition $e$ or distance metric $d$ as follows

$$e_{\mathrm{RMSE}}^k = \sqrt{\frac{1}{M} \sum_{i=1}^{M} \left\| e(\mathbf{x}_i^k, \hat{\mathbf{x}}_i^k) \right\|_2^2} = \sqrt{\frac{1}{M} \sum_{i=1}^{M} d(\mathbf{x}_i^k, \hat{\mathbf{x}}_i^k)^2}. \tag{2.45}$$

The average RMSE, the ARMSE, of a single run over the entire duration of a trajectory with $K$ steps is defined as

$$e_{\mathrm{ARMSE}} = \frac{1}{K} \sum_{k=1}^{K} e_{\mathrm{RMSE}}^k. \tag{2.46}$$

### 2.6.2   Normalized Innovation Squared (NIS)

The NIS hypothesis check (also known as gating test) is used to detect measurement outliers based on the analysis of the innovation/residual $\mathbf{r}$ and the innovation covariance in Equation (2.65b) [50]

$$\mathbf{S} = \mathbf{H} \boldsymbol{\Sigma} \mathbf{H}^\mathsf{T} + \mathbf{R}.$$

A Kalman filter, assumes that the residual has a (multivariate) zero-mean Gaussian distribution of dimension $n = \dim(\mathbf{r})$, i.e., the innovations are independent of each other and thus the sequence is zero-mean and white [10]. Therefore, the actual residual must fall into the estimated uncertainty $\mathbf{S}$, if the system is modeled correctly and the estimates are consistent. Similar to the NEES, we compute the squared Mahalanobis distance $s$ using Equation (2.44), between the residual $\mathbf{r}$ and the innovation belief with zero mean $\mathcal{S} \sim \mathcal{N}(\mathbf{0}, \mathbf{S})$, in the form

$$s = \mathbf{r}^\mathsf{T} \mathbf{S}^{-1} \mathbf{r} = \|\mathbf{r}\|_{\mathbf{S}}^2 \in [0, \infty], \tag{2.47}$$

as a measure of how many squared standard deviations the residual is away from the expect mean of $\mathbf{0}$.

The aim is to evaluate, if the innovation falls within a certain population of the distribution, e.g., $p = 0.997$ or $99.7\%$ (meaning that the point lies within an interval of $99.7\%$ of values closest to the mean of the normal distribution). Consequently, the lower the $p$ value, the stricter is the criteria to accept a measurement. The summed squares of $n$ independent univariate Gaussian variables is known to be $\chi_n^2$-distributed with $n$ degrees of freedom and is related to the innovation belief $\mathcal{S}$ by [26]

$$s = \chi_n^2(p), \ p \in [0, 1].\tag{2.48}$$

From the inverse of the chi-square cumulative distribution of the residual's dimension and the desired $p$ value, we obtain the reference value for the hypothesis/significance check

$$\text{outlier} = s \geq \chi_n^2(p).\tag{2.49}$$

Compared to the aforementioned estimation error metrics, this metric requires no ground truth information and can always be calculated from the existing parameters in the Kalman filter update step (see Section 2.8.1), but relies on consistent estimates.

### 2.6.3 Normalized Estimation Error Squared (NEES)

The NEES is the squared Mahalanobis distance (see Section 2.5.3) of the estimated and true value at every timestep $k$ of the trajectory and is, thus, a unit less scalar. It measures how many squared standard deviations the true values is away from the expected mean value of the state's distribution $\mathbf{X} \sim \mathcal{N}\left(\hat{\mathbf{x}}_i^k, \boldsymbol{\Sigma}_i^k\right)$

$$e_{\text{NEES}}^k = d_M\left(\mathbf{x}_i^k, \hat{\mathbf{x}}_i^k, \boldsymbol{\Sigma}_i^k\right)^2 = e\left(\mathbf{x}_i^k, \hat{\mathbf{x}}_i^k\right)^\mathsf{T} \boldsymbol{\Sigma}_i^{k,-1} e\left(\mathbf{x}_i^k, \hat{\mathbf{x}}_i^k\right) = \left\| e\left(\mathbf{x}_i^k, \hat{\mathbf{x}}_i^k\right) \right\|_{\boldsymbol{\Sigma}_i^k}^2,\tag{2.50}$$

with, for instance, an error definition $e\left(\mathbf{x}_i^k, \hat{\mathbf{x}}_i^k\right) = \mathbf{x}_i^k - \hat{\mathbf{x}}_i^k$, which depends on the estimator's error definition, e.g., *where* the error is defined (globally or locally) and *which* error representation was used, e.g., the small angle approximation for estimated orientations. The NEES is assumed to be $\chi_n^2$ distributed with $n$ DoF and a mean value of $n$, where $n$ is the dimension of the error, and can be used for hypothesis checks. NEES values below and above $n$ indicate under- and over-confidence, respectively.

Our filter is based on an indirect formulation (see Section 2.8.6), meaning that the uncertainty of the error-state is estimated. Both, position and orientation errors are defined to be in the local frame with ${}^{\mathcal{A}}_{\mathcal{A}}\mathbf{p}_{\mathcal{B}} = {}^{\mathcal{A}}_{\mathcal{A}}\hat{\mathbf{p}}_{\mathcal{B}} + {}^{\mathcal{A}}\hat{\mathbf{R}}_{\mathcal{B}}{}^{\mathcal{A}}_{\mathcal{A}}\tilde{\mathbf{p}}_{\mathcal{B}}$ and ${}^{\mathcal{A}}\mathbf{R}_{\mathcal{B}} = {}^{\mathcal{A}}\hat{\mathbf{R}}_{\mathcal{B}}{}^{\mathcal{A}}\tilde{\mathbf{R}}_{\mathcal{B}}$ with ${}^{\mathcal{A}}\tilde{\mathbf{R}}_{\mathcal{B}} \approx \mathbf{I} + \left[{}^{\mathcal{A}}\tilde{\boldsymbol{\theta}}_{\mathcal{B}}\right]_\times$. Consequently, the uncertainties/covariances of positions and orientation estimates are expressed in these spaces and the error definition $e(\mathbf{x}, \hat{\mathbf{x}})$ needs to be considered in the NEES calculation for each particular state.

The NEES for the local position error-state is obtained by

$$e_{\text{NEES},\tilde{\mathbf{p}}}^k = \left({}^{\mathcal{A}}\hat{\mathbf{R}}_{\mathcal{B}}^\mathsf{T}({}^{\mathcal{A}}_{\mathcal{A}}\mathbf{p}_{\mathcal{B}} - {}^{\mathcal{A}}_{\mathcal{A}}\hat{\mathbf{p}}_{\mathcal{B}})\right)^\mathsf{T} \boldsymbol{\Sigma}_{i,\tilde{\mathbf{p}}}^{k,-1} \left({}^{\mathcal{A}}\hat{\mathbf{R}}_{\mathcal{B}}^\mathsf{T}({}^{\mathcal{A}}_{\mathcal{A}}\mathbf{p}_{\mathcal{B}} - {}^{\mathcal{A}}_{\mathcal{A}}\hat{\mathbf{p}}_{\mathcal{B}})\right).\tag{2.51}$$

The NEES for the local orientation error-state is obtained by

$$e_{\text{NEES},\tilde{\boldsymbol{\theta}}}^k = \left(\left(\mathbf{I} - {}^{\mathcal{A}}\hat{\mathbf{R}}_{\mathcal{B}}^\mathsf{T}{}^{\mathcal{A}}\mathbf{R}_{\mathcal{B}}\right)^\vee\right)^\mathsf{T} \boldsymbol{\Sigma}_{i,\tilde{\boldsymbol{\theta}}}^{k,-1} \left(\mathbf{I} - {}^{\mathcal{A}}\hat{\mathbf{R}}_{\mathcal{B}}^\mathsf{T}{}^{\mathcal{A}}\mathbf{R}_{\mathcal{B}}\right)^\vee.\tag{2.52}$$

The average (mean) NEES over a single trajectory[7], is denoted as $\overline{\text{NEES}}$

$$e_{\overline{\text{NEES}}} = \frac{1}{K} \sum_{k=0}^{K} e_{\text{NEES}}^k.\tag{2.53}$$

---

[7]In OpenVINS [46], the NEES computed over a single trajectory is denoted as *single run NEES*.

Despite providing useful insights, a single run NEES does not adequately measure the filter consistency and a thorough examination is needed by evaluating the single run NEES over (multiple) $M$ Monte Carlo simulation runs, leading to an average NEES, the ANEES at the time instance $t^k$

$$e_{\text{ANEES}}^k = \frac{1}{M} \sum_{i=1}^{M} e_{\text{NEES},i}^k. \tag{2.54}$$

Assuming a zero mean estimation error $\tilde{\mathbf{x}}^k \sim \mathcal{N}(\mathbf{0}, \boldsymbol{\Sigma})$ with $\tilde{\mathbf{x}}^k = \mathbf{x}^k \boxminus \hat{\mathbf{x}}^k$, the $e_{\text{ANEES}}$ should have a chi-squared distribution $\chi_n^2$ of dimension $n = \dim(\tilde{\mathbf{x}})$. Therefore, the NEES should be on average $n$. This allows to assess the credibility by defining a lower and upper boundary, $r1$, $r2$, for the observed $e_{\text{ANEES}}$. In general, lower values indicate pessimism (under-confidence), while higher ones optimism. The boundaries of a commonly used two-sided 95 % confidence regions ($\alpha = 0.05$) is computed using the inverse of the chi-square cumulative distribution $\chi_{nM}^2$, defined as

$$[r_1, r_2] = \left[ \frac{\chi_{nM}^2(0.5\alpha)}{M}, \frac{\chi_{nM}^2(1-0.5\alpha)}{M} \right], \tag{2.55}$$

with the state dimension $n$, $M$ Monte-Carlo runs, and the chi-squared distribution $\chi_{nM}^2$ [10]. For $M = 10$ and $n = 3$, e.g., the IMU position, the lower and upper credibility bounds for the ANEES in the 95 % region are $[1.68, 4.7]$.

**Remark 1** *In contrast to a single run NEES, an ANEES evaluation can only be performed in simulation, based on multiple Monte-Carlo runs on a single reference/true trajectory, with different initial conditions for the initial state $\mathbf{x}^0$ drawn as sample from the initial distribution of the state $\mathbf{x}^0 \sim \mathcal{N}\left(\hat{\mathbf{x}}^0, \boldsymbol{\Sigma}^0\right)$, and the same set of measurements, but with randomly generated measurement noise for each individual run. Typically, these requirements cannot be met in robotics by repeating an experiment $M$ times, e.g., by following predefined set of waypoints multiple times.*

## 2.7 Estimated Trajectory Evaluation

The evaluation of error metrics is typically an offline process, that requires an estimated- and the actual (true) trajectory. A trajectory is not necessarily a 3D path in space time, rather a set/sequence of timestamped states. Apart from simulations, it is typically not possible or hard to obtain the true values of a dynamical system. In robotics, motion capture systems became standard to obtain highly accurate pose (position and orientation) measurement of a tracked device at high rates and allows evaluating the quality of the pose estimates of real robots. Several tools to quantify the accuracy of the estimated trajectory (pose) of SLAM or VIO systems by computing the ATE or Relative Pose Error (RPE) have been published, e.g., by Sturm *et al.* [141] Zhang and Scaramuzza [159], and Geneva *et al.* [46].

Apart from the absolute and relative pose trajectory errors, the consistency of pose estimates based on the estimated uncertainty is an indispensable metric for a reliable and credible localization. For instance, if different sources of information are fused or statistical outlier rejection methods are applied (see Section 2.6.2). This was one of the primary motivations of our open-sourced *statistical evaluation toolbox*[8] in our VINSEval framework[9]. The framework was proposed by Fornasier *et al.* in [40] and introduces a credibility measures based on the NEES (see Section 2.6.3) for VINS algorithms building upon the open-source trajectory evaluation framework[10] published in [159].

---
[8] https://github.com/aau-cns/cnspy_trajectory_evaluation
[9] https://github.com/aau-cns/vins_eval

Before error metrics can be computed, an association between the estimated and reference data points need to be established, as these sequences typically have different sample rates, might suffer from missing data, and different number of samples/length [141]. Missing true and estimated values can be approximated by interpolation, such that both sequences have the same amount of samples. Alternatively, associations between the timestamps of the estimated and reference trajectory can be found with a certain tolerance, which might reduce the total number sample points after the association.

### 2.7.1 Absolute Trajectory Error (ATE)

The ATE is a measure for the global consistency of the estimated trajectory and is obtained by averaging the distances (Euclidean and angular distance) between the true and estimated (pose) trajectory [141].

The pose trajectory evaluation also requires some preliminary information about the system and configuration. For instance, if the state space would generally be observable or not, or if the system was initialized at the true value or randomly. As discussed later in details, in a IMU/camera configuration the absolute position and rotation about a known gravity vector cannot be recovered. Meaning that this sensor constellation only allows to infer relative information, which might result, when directly compared to the true trajectory, in huge errors. Therefore, trajectory alignment strategies should be applied before computing the ATE.

After the data association and timestamp alignment was performed, the ATE is computed as the RMSE of the position- and orientation error over an estimated trajectory $m$ of $M$ (Monte-Carlo simulation) runs with respect to one true trajectory. The ATE for the position is computed using Equation (2.40) as

$$e_{\text{ATE},\mathbf{p}}^m = \sqrt{\frac{1}{K} \sum_{k=1}^{K} d_E \left( {}_{\mathcal{G}}^{\mathcal{G}}\mathbf{p}_{\mathcal{B}}^k, {}_{\mathcal{G}}^{\mathcal{G}}\hat{\mathbf{p}}_{\mathcal{B}}^{k,m} \right)}, \tag{2.56}$$

and for the orientation using Equation (2.42) as

$$e_{\text{ATE},\mathbf{R}}^m = \sqrt{\frac{1}{K} \sum_{k=1}^{K} d_R \left( {}^{\mathcal{G}}\mathbf{R}_{\mathcal{B}}^k, {}^{\mathcal{G}}\hat{\mathbf{R}}_{\mathcal{B}}^{k,m} \right)}. \tag{2.57}$$

The average over $M$ runs is obtained by

$$\bar{e}_{\text{ATE},\mathbf{p}} = \frac{1}{M} \sum_{m=1}^{M} e_{\text{ATE},\mathbf{p}}^m \tag{2.58}$$

and

$$\bar{e}_{\text{ATE},\mathbf{R}} = \frac{1}{M} \sum_{m=1}^{M} e_{\text{ATE},\mathbf{R}}^m \tag{2.59}$$

### 2.7.2 Credibility Evaluation

The estimator's credibility can be evaluated by defining a threshold, e.g., by a probability interval of $p = 0.99$, for the double-sided confidence boundary using Equation (2.55). If the threshold is reasonably high and any ANEES value exceeds these boundaries, the estimator can be classified as inconsistent, as shown in Figure 2.7.

---

[10]https://github.com/uzh-rpg/rpg_trajectory_evaluation

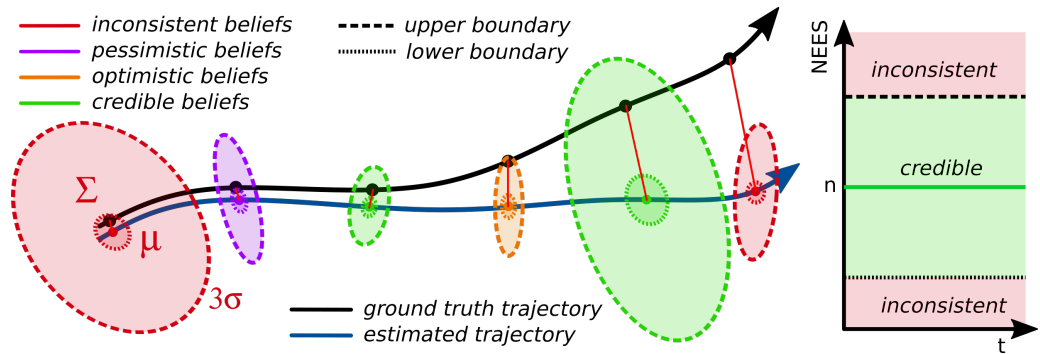

**Figure 2.7:** Shows the classification into inconsistent, pessimistic, optimistic and credible beliefs of an estimated trajectory with respect to the true trajectory given upper and lower credibility boundaries.

The uncertainty of pose estimates was previously used to obtain the NEES, e.g., in Eustice *et al.* [38], Walter *et al.* [146], and Geneva [46], but, as correctly mention in [83], a generalized NEES evaluation suffers from an appropriate error measure and error definition of the estimates. Typically, for filter-based VINS, an indirect filter formulation (see Section 2.8.6) is applied, which implies that the covariance of the error is estimated and many different choices of the error definition exist, as already discussed in Section 2.4.1. Therefore, the uncertainty of the estimated nominal-state is expressed in the local coordinates of the error-state space, rather than in the nominal-state space.

For the NEES computation in Equation (2.50), we need the *estimation error* which depends on the error-state definition. Since a spatial trajectory alignment might be required for the estimated trajectory, as discussed in Section 2.7.1, the estimated covariances might be transformed as well (which depends again on the error-state definition). Alternatively, the true trajectory can be aligned with the estimated trajectory, which mitigates the need for transforming the covariances, but leads to counterintuitive/wrong results in case the trajectories are visualized in plots. For instance, if multiple estimated and a single true trajectory are visualized in a single 3D plot.

Summarized, it is not trivial and not standardized how to compute the ANEES over different estimators' trajectories, given a set of estimates for the position, orientation, and the associated covariance matrix/matrices[11].

## 2.8 Estimator Types

In order to introduce a common notation and wording, we briefly revisit various types of estimators.

### 2.8.1 Kalman Filter (KF)

The Kalman filter [77], estimates the state of a linear dynamic system (or physical plant), that is perturbed by a zero-mean white Gaussian noise, using measurements that are linear functions of the systems state and perturbed by additive zero-mean white Gaussian noise. The linear dynamic system is described by the differential or difference equations of the state-space model to represent the deterministic and random phenomena/process [50]. The state variables of the system are, thus, uni- or multivariate random variables, which are changing over time.

---

[11]The covariance of the position and orientation might be provided separately by two mutually independent $3 \times 3$ covariance matrices, or by single $6 \times 6$ covariance matrix for the 6-DoF pose including in the off-diagonal block matrices – the cross-covariance between the position and orientation estimates. Consequently, the $6 \times 6$ covariance matrix would allow considering the correlation between the position and orientation in the computation of the NEES of the estimated pose.

Providing a finite-dimensional representation of the problem at hand by its variables, parameters, and differential equations are the basis of the state-space approach. Dependent variables of the differential equations are so-called state variables of the dynamic system [50].

We are interested in discrete-time over a continuous-time dynamic system – a system that varies continuously of a real interval with the time variable $t$. In particular, we are interested in state variables at a certain point in time, e.g., when a measurement occurred. Therefore, we use the following shorthand notation of [50], to relate a state variable to a certain point in time $\mathbf{x}^k = \mathbf{x}(t^k)$, with $t^k = k\Delta t$, with a uniform sampling period $\Delta t$ and consequently $t^k = t^{k-1} + \Delta t$.

A discrete-time system model is described by the discrete-time state equations in the form of

$$\mathbf{x}^k = \mathbf{\Phi}^{k|k-1}\mathbf{x}^{k-1} + \mathbf{\Gamma}^{k-1}\mathbf{u}^{k-1}, \tag{2.60}$$

$$\mathbf{z}^k = \mathbf{H}^k\mathbf{x}^k, \tag{2.61}$$

with the state variable $\mathbf{x}$, the state transition matrix from time $t^{k-1}$ to time $t^k$ is $\mathbf{\Phi}^{k|k-1} = \mathbf{\Phi}(t^k, t^{k-1})$, the discrete-time control input or input coupling matrix $\mathbf{\Gamma}^{k-1}$, the control input $\mathbf{u}^{k-1}$, the measurement (sensitivity) matrix $\mathbf{H}^k$, and the measurement or output vector $\mathbf{z}^k$ [50].

Treating the underlying dynamics of the system as a random process, by $\mathbf{x} \sim \mathcal{N}(\hat{\mathbf{x}}, \mathbf{\Sigma})$, we can describe the discrete-time linear stochastic system with control input in the general form

$$\mathbf{x}^k = \mathbf{\Phi}^{k|k-1}\mathbf{x}^{k-1} + \mathbf{\Gamma}^{k-1}\mathbf{u}^{k-1} + \mathbf{G}^{k-1}\mathbf{w}^{k-1}, \tag{2.62}$$

$$\mathbf{z}^k = \mathbf{H}^k\mathbf{x}^k + \mathbf{v}^k, \tag{2.63}$$

with the state $\mathbf{x}$ vector, $\mathbf{G}$ the process noise coupling matrix, $\mathbf{w} \sim \mathcal{N}(\mathbf{0}, \mathbf{Q})$ the zero-mean white and uncorrelated Gaussian system/plant noise, and $\mathbf{v} \sim \mathcal{N}(\mathbf{0}, \mathbf{R})$ the independent measurement noise [50].

The standard Kalman filter formulation [77] for the state of previously described linear stochastic system can be divided in a prediction step Equation (2.64) and update step Equation (2.65) as follows: The prediction step of the estimated state $\hat{\mathbf{x}}$ and the corresponding covariance $\mathbf{\Sigma}$ has, e.g., the form

$$\hat{\mathbf{x}}^k = \mathbf{\Phi}^{k|k-1}\hat{\mathbf{x}}^{k-1}, \tag{2.64a}$$

$$\mathbf{\Sigma}^k = \mathbf{\Phi}^{k|k-1}\mathbf{\Sigma}^{k-1}\mathbf{\Phi}^{k|k-1^\mathsf{T}} + \mathbf{G}^{k-1}\mathbf{Q}^{k-1}\mathbf{G}^{k-1^\mathsf{T}}. \tag{2.64b}$$

The correction step of the *a-priori* belief $\mathbf{x}^{k(-)}$ is defined by the following equations:

$$\mathbf{r}^k = \mathbf{z}^k - \mathbf{H}^k\hat{\mathbf{x}}^{k(-)}, \tag{2.65a}$$

with the residual $\mathbf{r}^k$, describing the difference between the measured and estimated measurement $\hat{\mathbf{z}}^k = \mathbf{H}^k\hat{\mathbf{x}}^{k(-)}$. The corresponding residual covariance $\mathbf{S}$ is defined as

$$\mathbf{S}^k = \mathbf{H}^k\mathbf{\Sigma}_{ii}^{k(-)}(\mathbf{H}^k)^\mathsf{T} + \mathbf{R}^k. \tag{2.65b}$$

The Kalman gain $\mathbf{K}^k$ is defined as[12]:

$$\mathbf{K}^k = \mathbf{\Sigma}_{ii}^{k(-)}(\mathbf{H}^k)^\mathsf{T}(\mathbf{S}^k)^{-1}. \tag{2.65c}$$

Alternatively, the Kalman gain can be computed equivalently, according to Xu and Zhang in [156], in the form of

$$\mathbf{K}^k = \left(\mathbf{H}^\mathsf{T}\mathbf{\Sigma}_{ii}^{k(-)}\mathbf{H} + \left(\mathbf{\Sigma}_{ii}^{k(-)}\right)^{-1}\right)^{-1}\mathbf{H}^\mathsf{T}\mathbf{R}^{-1}, \tag{2.65d}$$

to mitigate computational limits in case the measurement vector is significantly larger than the state vector.

The *a priori* state $\mathbf{x}^{k(-)}$ is finally updated/corrected by

$$\hat{\mathbf{x}}^{k(+)} = \hat{\mathbf{x}}^{k(-)} + \mathbf{K}^k\mathbf{r}^k \tag{2.65e}$$

and

$$\mathbf{\Sigma}^{k(+)} = (\mathbf{I} - \mathbf{K}^k\mathbf{H}^k)\mathbf{\Sigma}^{k(-)}, \tag{2.65f}$$

or the more numerical robust Josephs' form [50]

$$\mathbf{\Sigma}^{k(+)} = (\mathbf{I} - \mathbf{K}^k\mathbf{H}^k)\mathbf{\Sigma}^{k(-)}(\mathbf{I} - \mathbf{K}^k\mathbf{H}^k)^\mathsf{T} + \mathbf{K}^k\mathbf{R}^k(\mathbf{K}^k)^\mathsf{T}, \tag{2.65g}$$

leading to the *a posteriori* belief $\mathbf{x}^{k(+)} \sim \mathcal{N}\left(\hat{\mathbf{x}}^{k(+)}, \mathbf{\Sigma}^{k(+)}\right)$.

The Kalman filter is often referred to as a recursive filter, as the process of state prediction steps in Equation (2.64), followed by a state correction steps described in Equation (2.65) are repeated with the rate the measurements are obtained, without the need to store previous information.

**Remark 2** *The KF, is an* optimal *linear estimator, if the system can be described by a linear model, in which the system and measurement noise are zero-mean white Gaussian. The filter is implemented as a recursive data processing algorithm, that incorporates all information that can be processed (regardless of the precision) to estimate the current variable (state) of interest, making use of the system dynamic and measurement model, the* a priori *knowledge about the system and measurement noise characteristics, and the information about the initial condition of the state variables [103].*

### 2.8.2 Extended Kalman Filter (EKF)

The majority of dynamic systems are nonlinear and are modeled by a discrete-time nonlinear state transformation in, e.g., the form

$$\mathbf{x}^k = \phi(\mathbf{x}^{k-1}, \mathbf{u}^{k-1}) + \mathbf{w}^{k-1}, \tag{2.66a}$$

$$\mathbf{z}^k = h(\mathbf{x}^k) + \mathbf{v}^k. \tag{2.66b}$$

Since a Kalman filter operates in terms of means and covariances of the assumed Gaussian probability distribution, general nonlinear transformation are likely to violate/change the values in unpredictable ways. Therefore, nonlinear filtering cannot be implemented exactly [50] and tend to be suboptimal.

Still, in practical applications, linearization techniques are typically applied to use a Kalman filter in nonlinear estimation problems. The EKF is linearizing the nonlinear plant and measurement functions, by computing the Jacobian with respect to the state, about the estimated trajectory $\hat{\mathbf{x}}^{(-)}$ in every time/filter step [50]

$$\boldsymbol{\Phi} \approx \left.\frac{\partial \phi(\mathbf{x}, \mathbf{u})}{\partial \mathbf{x}}\right|_{\mathbf{x}=\hat{\mathbf{x}}, \mathbf{u}}, \ \mathbf{H} \approx \left.\frac{\partial h(\mathbf{x})}{\partial \mathbf{x}}\right|_{\mathbf{x}=\hat{\mathbf{x}}^{(-)}}. \tag{2.67}$$

---

[12]In a univariate estimation problem (1 DoF), the Kalman gain $K$ is a value between 0 and 1 optimally weighting the measurement and, thus, the amount of correction applied on the belief.

**Remark 3** *In contrast to the KF, the EKF has no guaranties and does not lead to optimal estimates due to the nonlinear nature of the problem, which is linearized at the best (current) estimate.*

### 2.8.3 Schmidt-Kalman Filter (SKF)

In order to reduce the computational requirements of Kalman filtering, Stanley F. Schmidt introduced the concept of *nuisance variables* in [129], which are state variables that are not corrected/updated, but considered as part of the state vector. It allows the partitioning of the dynamics model's state $\mathbf{x} = \begin{bmatrix} \mathbf{x}_e \\ \mathbf{x}_n \end{bmatrix}$ into essential variables $\mathbf{x}_e$ and nuisance variables $\mathbf{x}_n$, which are generally used for modeling correlated measurement noise. This approach is denoted as SKF and leads to suboptimal estimation performance, by applying the so-called *Schmidt-Kalman Gain* instead of the Kalman gain. A primary motivation is, that the dynamics of these nuisance parameters are typically not linked/coupled with the dynamics of the system of interest, meaning that the off-diagonal blocks of the state transition matrix and process noise matrix are zero and the prediction step can be partitioned reducing the order [50].

In contrast to Equation (2.62), the partitioned discrete-time linear stochastic system for the process $\mathbf{x} \sim \mathcal{N}\left(\begin{bmatrix} \hat{\mathbf{x}}_e \\ \hat{\mathbf{x}}_n \end{bmatrix}, \begin{bmatrix} \boldsymbol{\Sigma}_{ee} & \boldsymbol{\Sigma}_{en} \\ \boldsymbol{\Sigma}_{ne} & \boldsymbol{\Sigma}_{nn} \end{bmatrix}\right)$ obtains the form

$$\begin{bmatrix} \mathbf{x}_e \\ \mathbf{x}_n \end{bmatrix}^k = \begin{bmatrix} \boldsymbol{\Phi}_{ee} & \mathbf{0} \\ \mathbf{0} & \boldsymbol{\Phi}_{nn} \end{bmatrix}^{k|k-1} \begin{bmatrix} \mathbf{x}_e \\ \mathbf{x}_n \end{bmatrix}^{k-1} + \begin{bmatrix} \boldsymbol{\Gamma}_e \\ \boldsymbol{\Gamma}_n \end{bmatrix}^{k-1} \begin{bmatrix} \mathbf{u}_e \\ \mathbf{u}_n \end{bmatrix}^{k-1} + \begin{bmatrix} \mathbf{w}_e \\ \mathbf{w}_n \end{bmatrix}^{k-1}, \tag{2.68}$$

with $\boldsymbol{\Gamma}_n^{k-1} = \mathbf{0}$, $\boldsymbol{\Phi}_{nn}^{k|k-1} = \mathbf{I}$. As process noise vectors $\mathbf{w}_e$ and $\mathbf{w}_n$ of the essential system and the nuisance parameter (describing e.g., the correlated measurement noise) are uncorrelated, the process noise matrix obtains the following form

$$\mathbf{Q} = \begin{bmatrix} \mathbf{Q}_{ee} & \mathbf{0} \\ \mathbf{0} & \mathbf{Q}_{nn} \end{bmatrix} \tag{2.69}$$

and allows computing the covariance prediction in four individual blocks by exploiting the sparsity of state transition and process noise matrix

$$\begin{bmatrix} \boldsymbol{\Sigma}_{ee} & \boldsymbol{\Sigma}_{en} \\ \boldsymbol{\Sigma}_{ne} & \boldsymbol{\Sigma}_{nn} \end{bmatrix}^k = \begin{bmatrix} \boldsymbol{\Phi}_{ee} & \mathbf{0} \\ \mathbf{0} & \boldsymbol{\Phi}_{nn} \end{bmatrix}^{k|k-1} \begin{bmatrix} \boldsymbol{\Sigma}_{ee} & \boldsymbol{\Sigma}_{en} \\ \boldsymbol{\Sigma}_{ne} & \boldsymbol{\Sigma}_{nn} \end{bmatrix}^{k-1} \begin{bmatrix} \boldsymbol{\Phi}_{ee} & \mathbf{0} \\ \mathbf{0} & \boldsymbol{\Phi}_{nn} \end{bmatrix}^{k|k-1\mathsf{T}} + \begin{bmatrix} \mathbf{Q}_{ee} & \mathbf{0} \\ \mathbf{0} & \mathbf{Q}_{nn} \end{bmatrix}^{k-1}. \tag{2.70}$$

The partitioned measurement is, like Equation (2.63), in the form

$$\mathbf{z}^k = \begin{bmatrix} \mathbf{H}_{ee} & \mathbf{H}_{nn} \end{bmatrix}^k \begin{bmatrix} \mathbf{x}_e \\ \mathbf{x}_n \end{bmatrix}^k + \mathbf{v}^k = \mathbf{H}_{ee}\mathbf{x}_e + \mathbf{H}_{nn}\mathbf{x}_n + \mathbf{v}, \tag{2.71}$$

with the essential state dependence $\mathbf{H}_{ee}\mathbf{x}_e$, the correlated noise $\mathbf{H}_{nn}\mathbf{x}_n$, and uncorrelated noise $\mathbf{v} \sim \mathcal{N}(\mathbf{0}, \mathbf{R})$.

The suboptimal Schmidt-Kalman filter update is derived based on Kalman filter update steps described in Equation (2.65). The residual $\mathbf{r}^k$ is computed by

$$\mathbf{r}^k = \mathbf{z}^k - \begin{bmatrix} \mathbf{H}_{ee} & \mathbf{H}_{nn} \end{bmatrix}^k \begin{bmatrix} \mathbf{x}_e \\ \mathbf{x}_n \end{bmatrix}^{k(-)}. \tag{2.72a}$$

The corresponding partitioned residual covariance $\mathbf{S}$ is defined as

$$\mathbf{S}^k = \begin{bmatrix} \mathbf{H}_{ee} & \mathbf{H}_{nn} \end{bmatrix}^k \begin{bmatrix} \boldsymbol{\Sigma}_{ee} & \boldsymbol{\Sigma}_{en} \\ \boldsymbol{\Sigma}_{ne} & \boldsymbol{\Sigma}_{nn} \end{bmatrix}^{k(-)} \left( \begin{bmatrix} \mathbf{H}_{ee} & \mathbf{H}_{nn} \end{bmatrix}^k \right)^{\mathsf{T}} + \mathbf{R}^k. \tag{2.72b}$$

The partitioned Kalman gain $\mathbf{K}^k = \begin{bmatrix} \mathbf{K}_{ee} \\ \mathbf{K}_{nn} \end{bmatrix}^k$ is obtained by

$$\mathbf{K}^k = \begin{bmatrix} \mathbf{K}_{ee} \\ \mathbf{K}_{nn} \end{bmatrix}^k = \begin{bmatrix} \boldsymbol{\Sigma}_{ee} & \boldsymbol{\Sigma}_{en} \\ \boldsymbol{\Sigma}_{ne} & \boldsymbol{\Sigma}_{nn} \end{bmatrix}^{k(-)} \left( \begin{bmatrix} \mathbf{H}_{ee} & \mathbf{H}_{nn} \end{bmatrix}^k \right)^{\mathsf{T}} \left( \mathbf{S}^k \right)^{-1}, \tag{2.72c}$$

leading to $\mathbf{K}_{ee}^k = \left( \boldsymbol{\Sigma}_{ee}^{k(-)} (\mathbf{H}_{ee}^k)^{\mathsf{T}} + \boldsymbol{\Sigma}_{en}^{k(-)} (\mathbf{H}_{nn}^k)^{\mathsf{T}} \right) \left( \mathbf{S}^k \right)^{-1}$. The suboptimal Schmidt-Kalman gain $\breve{\mathbf{K}}^k$ sets the Kalman gain for nuisance parameters $\mathbf{K}_{nn}$ to zero, which leads to

$$\breve{\mathbf{K}}^k = \begin{bmatrix} \mathbf{K}_{ee} \\ \mathbf{0} \end{bmatrix}^k. \tag{2.72d}$$

The suboptimal *a priori* state $\mathbf{x}^{k(-)}$ is finally updated/correct by

$$\hat{\mathbf{x}}^{k(+)} = \hat{\mathbf{x}}^{k(-)} + \breve{\mathbf{K}}^k \mathbf{r}^k = \hat{\mathbf{x}}^{k(-)} + \begin{bmatrix} \mathbf{K}_{ee} \\ \mathbf{0} \end{bmatrix}^k \mathbf{r}^k \tag{2.72e}$$

meaning that only the essential parameters obtain a correction $\hat{\mathbf{x}}_e^{k(+)} = \hat{\mathbf{x}}_e^{k(-)} + \mathbf{K}_{ee}\mathbf{r}^k$ and $\hat{\mathbf{x}}_n^{k(+)} = \hat{\mathbf{x}}_n^{k(-)}$. The partitioned covariance is also updated using the Schmidt-Kalman gain

$$\begin{aligned} \boldsymbol{\Sigma}^{k(+)} &= (\mathbf{I} - \breve{\mathbf{K}}^k \mathbf{H}^k) \boldsymbol{\Sigma}^{k(-)} \\ &= \begin{bmatrix} (\mathbf{I} - \mathbf{K}_{ee}\mathbf{H}_{ee}) \boldsymbol{\Sigma}_{ee}^{(-)} - \mathbf{K}_{ee}\mathbf{H}_{nn}\boldsymbol{\Sigma}_{ne}^{(-)} & \mathbf{I} - \mathbf{K}_{ee}\mathbf{H}_{ee}) \boldsymbol{\Sigma}_{en}^{(-)} - \mathbf{K}_{ee}\mathbf{H}_{nn}\boldsymbol{\Sigma}_{nn}^{(-)} \\ \bullet & \boldsymbol{\Sigma}_{nn}^{(-)} \end{bmatrix}^k, \end{aligned} \tag{2.72f}$$

meaning that $\boldsymbol{\Sigma}_{ne}^{k(+)} = \boldsymbol{\Sigma}_{en}^{k(+)^{\mathsf{T}}}$ and $\boldsymbol{\Sigma}_{nn}^{k(+)} = \boldsymbol{\Sigma}_{nn}^{k(-)}$ [50].

**Remark 4** *The cross-covariance between essential and nuisance parameters is tracked/maintained and is, together with the* a-priori *covariance of the nuisance parameters, considered in the update step. The partitioning of the state vector, the constraints in the dynamic systems, and the Schmidt-Kalman gain, allow reducing the computational effort significantly. Some of these concepts are heavily used in the proposed algorithms.*

### 2.8.4 Covariance Intersection (CI)

Despite being not directly related to Kalman filtering, CI is important concept of combining beliefs about a process, when the cross-correlation between the beliefs are unknown. If they are known the optimal estimate can be obtained using BLUE (see Section 2.8.5). Generally, CI solves the problem of fusing two beliefs to obtain a new belief by a convex combination that is provable consistent and was proposed by Julier and Uhlmann in [71]. The CI algorithm is in the form

$$\boldsymbol{\Sigma}_{cc}^{-1} = \omega \boldsymbol{\Sigma}_{aa}^{-1} + (1 - \omega) \boldsymbol{\Sigma}_{bb}^{-1} \tag{2.73a}$$

$$\hat{\mathbf{x}}_c = \boldsymbol{\Sigma}_{cc} \left( \omega \boldsymbol{\Sigma}_{aa}^{-1} \hat{\mathbf{x}}_a + (1 - \omega) \boldsymbol{\Sigma}_{bb}^{-1} \hat{\mathbf{x}}_b \right) \tag{2.73b}$$

where $\omega \in [0,1]$. The CI algorithm results in a consistent belief $\mathbf{x}_c$ for any correlations between the beliefs $\mathbf{x}_{\{a,b\}}$ and any choice on $\omega$.

One common way to obtain the weight $\omega$ is e.g., to find the minimum of the (logarithm of the) determinant or trace of the final uncertainty in the interval $\omega \in [0,1]$

$$
\begin{aligned}
\mathbf{\Sigma}_{cc}^* &= \underset{0 \leqslant \omega \leqslant 1}{\arg\min} \det \left( \left( \omega \mathbf{\Sigma}_{aa}^{-1} + (1-\omega) \mathbf{\Sigma}_{bb}^{-1} \right)^{-1} \right) \\
&= \underset{0 \leqslant \omega \leqslant 1}{\arg\min} \frac{1}{\det \left( \omega \mathbf{\Sigma}_{aa}^{-1} + (1-\omega) \mathbf{\Sigma}_{bb}^{-1} \right)}
\end{aligned}
\tag{2.73c}
$$

given

$$
\det(\mathbf{M}) = \frac{1}{\det \left( \mathbf{M}^{-1} \right)},
\tag{2.73d}
$$

for an invertible squared matrix $\mathbf{M}$ and allows avoiding the covariance inversion.

In general, the determinant is a measure for spanned volume of the covariance matrix and indicates the overall uncertainty. In CI, one aims for the most certain result, therefore, the (logarithm of the) determinant should be minimized by $\omega$. In [120], Reinhardt *et al.*, pointed out that, CI may lead to weights $\omega$ of only 0 and 1, if one of the uncertainties is within the other, i.e., $\mathbf{\Sigma}_{aa} - \mathbf{\Sigma}_{bb} \succeq \mathbf{0}$. This means, fusing the two beliefs would always yield in either $\mathbf{x}_c = \mathbf{x}_a$ or $\mathbf{x}_c = \mathbf{x}_b$, depending on which one is more certain, and potential information provided by the second belief is ignored.

### 2.8.5 Best Linear Unbiased Estimator (BLUE)

The optimal solution to fuse two beliefs $\mathbf{x}_{\{a,b\}}$ of a common process linearly, in existence of cross-correlations $\mathbf{\Sigma}_{ab} = \mathbf{\Sigma}_{ba}^{\mathsf{T}} \neq \mathbf{0}$, is given by the Bar-Shalom-Campo formula [9]

The optimal mean $\hat{\mathbf{x}}_c$ is obtained by

$$
\begin{aligned}
\hat{\mathbf{x}}_c =&(\mathbf{\Sigma}_{bb} - \mathbf{\Sigma}_{ba})(\mathbf{\Sigma}_{aa} + \mathbf{\Sigma}_{bb} - \mathbf{\Sigma}_{ab} - \mathbf{\Sigma}_{ba})^{-1}\hat{\mathbf{x}}_a + \\
&(\mathbf{\Sigma}_{aa} - \mathbf{\Sigma}_{ab})(\mathbf{\Sigma}_{aa} + \mathbf{\Sigma}_{bb} - \mathbf{\Sigma}_{ab} - \mathbf{\Sigma}_{ba})^{-1}\hat{\mathbf{x}}_b
\end{aligned}
\tag{2.74a}
$$

and the optimal covariance in the form

$$
\mathbf{\Sigma}_{cc} = \mathbf{\Sigma}_{aa} - (\mathbf{\Sigma}_{aa} - \mathbf{\Sigma}_{ab})^{\mathsf{T}}(\mathbf{\Sigma}_{aa} + \mathbf{\Sigma}_{bb} - \mathbf{\Sigma}_{ab} - \mathbf{\Sigma}_{ba})^{-1}(\mathbf{\Sigma}_{aa} - \mathbf{\Sigma}_{ab})
\tag{2.74b}
$$

### 2.8.6 Error-State EKF (ESEKF)

For inertial navigation systems, it became standard to implement the Kalman filter rather on the *error-state space* than on the *total state space* [103].

In the so-called *indirect* or *error-state* formulation, the Kalman filter estimates the uncertainty of error-state $\tilde{\mathbf{x}}$, while at the same time the nominal-state $\bar{\mathbf{x}}$ is maintained/integrated and corrected by injection the estimated error-state. The true state is expressed as combination of the nominal- and the error-state, e.g., $\mathbf{x} = \bar{\mathbf{x}} \boxplus \tilde{\mathbf{x}}$ [136] (please refer to the error definitions in Section 2.4). The nominal-state $\bar{\mathbf{x}}$ is considered as large and high frequent signal which is integrable in non-linear fashion, while the error-state $\tilde{\mathbf{x}}$ is a small, low frequent zero-mean signal, thus linearly integrable and suitable for Kalman filtering [136]. Meaning that the model of the nominal-states dynamics do not have to be modeled explicitly in the filter, instead the dynamics of the filter are based on the system's error propagation equations, which are assumed to be low frequent and adequately represented as linear [125].

In the indirect feedback formulation, measurements are used to correct the nominal-state as difference between the measured and integrated data, which results in the optimal estimate of the error-state. The corrected error-state is added/applied to the nominal-state to establish an estimate of the total state. At every update step, the error-state correction is included/injected in the nominal (full) state. Thus, the error-state is assumed to be zero after applying the correction to the nominal-state[13]. In the indirect formulation, it is common to use both high frequent propagation and low frequent update sensors, leading to a saw tooth evolution of the error-state's uncertainty. Another property is that the estimator is lagging on indirectly observed states (that are not measured directly), as it measure only its effects on other states [125].

In contrast to the EKF, the linearized nonlinear discrete-time plant and measurement function at a certain time-step $t^k$, is obtained by computing the Jacobian with respect to the error-state $\tilde{\mathbf{x}}$, about the nominal-state's trajectory $\bar{\mathbf{x}} = \hat{\mathbf{x}}^{(-)}$ in the form of

$$\mathbf{\Phi} \approx \left.\frac{\partial \phi(\mathbf{x}, \mathbf{u})}{\partial \tilde{\mathbf{x}}}\right|_{\mathbf{x}=\bar{\mathbf{x}}, \mathbf{u}} = \left.\frac{\partial \phi(\mathbf{x}, \mathbf{u})}{\partial \mathbf{x}}\right|_{\mathbf{x}=\bar{\mathbf{x}}, \mathbf{u}} \left.\frac{\partial \mathbf{x}}{\partial \tilde{\mathbf{x}}}\right|_{\mathbf{x}=\bar{\mathbf{x}}} \tag{2.75}$$

and the measurement matrix is obtained by

$$\mathbf{H} \approx \left.\frac{\partial h(\mathbf{x})}{\partial \tilde{\mathbf{x}}}\right|_{\mathbf{x}=\bar{\mathbf{x}}^{(-)}} = \left.\frac{\partial h(\mathbf{x})}{\partial \mathbf{x}}\right|_{\mathbf{x}=\bar{\mathbf{x}}^{(-)}} \left.\frac{\partial \mathbf{x}}{\partial \tilde{\mathbf{x}}}\right|_{\mathbf{x}=\bar{\mathbf{x}}} . \tag{2.76}$$

### Error propagation step

Summarizing [136], the ESEKF maintains two states, the estimated mean of nominal-state $\bar{\mathbf{x}}$ and the uncertainty of the error-state $\tilde{\mathbf{x}} \sim \mathcal{N}(\mathbf{0}, \mathbf{\Sigma})$. Assuming the following nonlinear stochastic system for the process $\mathbf{x} \sim \mathcal{N}(\hat{\mathbf{x}}, \mathbf{\Sigma})$ in the form

$$\dot{\mathbf{x}} = f(\mathbf{x}, \mathbf{u}, \mathbf{w}), \tag{2.77a}$$

with the noisy control input $\mathbf{u} = \mathbf{u}^{\#} - \tilde{\mathbf{u}}$, $\tilde{\mathbf{u}} \sim \mathcal{N}(\mathbf{0}, \mathbf{U}_c)$, and a Gaussian perturbation $\mathbf{w} \sim \mathcal{N}(\mathbf{0}, \mathbf{W}_c)$. The continuous-time error-state dynamics of, e.g., $\dot{\tilde{\mathbf{x}}} = \boxminus \dot{\mathbf{x}} \boxplus \dot{\mathbf{x}}$, can be linearized to

$$\dot{\tilde{\mathbf{x}}} = \mathbf{F}\tilde{\mathbf{x}} + \mathbf{B}\tilde{\mathbf{u}} + \mathbf{C}\mathbf{w}, \tag{2.77b}$$

with

$$\mathbf{F} \approx \left.\frac{\partial f}{\partial \tilde{\mathbf{x}}}\right|_{\mathbf{x}, \mathbf{u}^{\#}}, \; \mathbf{B} \approx \left.\frac{\partial f}{\partial \tilde{\mathbf{u}}}\right|_{\mathbf{x}, \mathbf{u}^{\#}}, \; \mathbf{C} \approx \left.\frac{\partial f}{\partial \mathbf{w}}\right|_{\mathbf{x}, \mathbf{u}^{\#}} \tag{2.77c}$$

The discrete-time error-state dynamics, integrated over $\Delta t$, with $\mathbf{\Phi}^{k|k-1} = \exp(\mathbf{F}^{k-1}\Delta t)$, is

$$\tilde{\mathbf{x}}^k = \mathbf{\Phi}^{k|k-1}\tilde{\mathbf{x}}^{k-1} + \Delta t \mathbf{B}\tilde{\mathbf{u}}^{k-1} + \mathbf{C}\mathbf{w}^{k-1}, \tag{2.77d}$$

with $\mathbf{w}^{k-1} \sim \mathcal{N}(\mathbf{0}, \Delta t \mathbf{W}_c)$ and $\tilde{\mathbf{u}}^{k-1} \sim \mathcal{N}(\mathbf{0}, \mathbf{U}_c)$. The covariance of the error-state is propagated by

$$\mathbf{\Sigma}^k = \mathbf{\Phi}^{k|k+1}\mathbf{\Sigma}^{k-1}(\mathbf{\Phi}^{k|k+1})^{\mathsf{T}} + \mathbf{Q}^{k|k-1}, \tag{2.77e}$$

with the integrated process noise $\mathbf{Q}^{k|k-1} = \Delta t^2 \mathbf{B}\mathbf{U}_c(\mathbf{B})^{\mathsf{T}} + \Delta t \mathbf{C}\mathbf{W}_c(\mathbf{C})^{\mathsf{T}}$, further details can be found in [136].

### Error correction step

Assuming a sensor measurement provides information depending on the total state $\mathbf{x}$ in the form

$$\mathbf{z} = h(\mathbf{x}^k) + \mathbf{v}^k, \tag{2.78a}$$

---

[13]The error-state is not explicitly maintained in filter formulation, only it's covariance and the mean of the nominal-state

with the nonlinear measurement function $h()$ and the uncorrelated noise $\mathbf{v} \sim \mathcal{N}(\mathbf{0}, \mathbf{R})$. The measurement presented to the indirect filter is the difference between the estimated measurement at nominal-state and the actual measurement in the form

$$\bar{\mathbf{z}} = \boxminus \hat{\mathbf{z}} \boxplus \mathbf{z} = \boxminus h(\bar{\mathbf{x}}) \boxplus h(\mathbf{x}) + \mathbf{v}, \tag{2.78b}$$

which can be approximated by

$$\bar{\mathbf{z}} \approx \mathbf{H}\tilde{\mathbf{x}} + \mathbf{v}. \tag{2.78c}$$

The linearized measurement matrix is obtained by linearizing Equation (2.78b) with respect to the error-state about the nominal-state (best linearization point available) by

$$\mathbf{H}^k \approx \left.\frac{\partial \bar{\mathbf{z}}}{\partial \tilde{\mathbf{x}}}\right|_{\mathbf{x}=\hat{\mathbf{x}}^{(-)}} = \left.\frac{\partial \bar{\mathbf{z}}}{\partial \mathbf{x}}\right|_{\mathbf{x}=\hat{\mathbf{x}}^{k(-)}} \left.\frac{\partial \mathbf{x}}{\partial \tilde{\mathbf{x}}}\right|. \tag{2.78d}$$

The residual $\mathbf{r}^k$ is the difference between observation and estimated observation $\hat{\mathbf{z}}^k = h(\hat{\mathbf{x}}^k)$

$$\mathbf{r}^k = \boxminus h(\hat{\mathbf{x}}^k) \boxplus \mathbf{z}^k = \boxminus \hat{\mathbf{z}}^k \boxplus \mathbf{z}^k \tag{2.78e}$$

The Kalman gain $\mathbf{K}^k$ and *a-posteriori* covariance $\mathbf{\Sigma}^{k(+)}$ is computed as in the Kalman filter (see Equation (2.65)) and the nominal-state is corrected by injecting the best estimate of the error-state $\hat{\tilde{\mathbf{x}}}^{(+)} = \mathbf{0} = \hat{\tilde{\mathbf{x}}}^{(-)} \boxplus \mathbf{K}^k \mathbf{r}^k$ (which is zero after the update step)

$$\hat{\mathbf{x}}^{k(+)} = \hat{\mathbf{x}}^{k(+)} \boxplus \hat{\tilde{\mathbf{x}}}^{(+)} = \hat{\mathbf{x}}^{k(+)} \boxplus \mathbf{K}^k \mathbf{r}^k, \tag{2.78f}$$

leading to a reset of the error-state's mean $\hat{\tilde{\mathbf{x}}}^{(+)} = \mathbf{0}$.

## 2.9 Filter-based Aided Inertial Navigation System (AINS)

For aided inertial navigation, we use a triaxial accelerometer and gyroscope of a Micro Electro Mechanical Systems (MEMS) IMU (6-DoF) which is strapped down on a vehicle/agent. Low-cost MEMS IMUs are well suited for autonomous navigation systems, where weight, power consumption and costs are constrained, such as MAVs. The INS provides attitude and translation information based on proprioceptive sensor data, which are in generally good high frequency information. An INS based on low-cost and consumer MEMS IMUs tend to drift unbounded at faster rates due to biases, other systematic, and random effects, such as random walk or other colored noise, temperature or mechanical stress of the package. Compared to white noise on gyroscope and accelerometer biases, the angle and velocity random walk is changing the biases on average with time. Consequently, these biases need to be estimated in order to reduce systematically errors in the sensor readings. Other systematic errors, such as input axis misalignment, scale factors or temperature drift, are not considered[14].

Additionally, it requires aiding by other navigation/exteroceptive sensors which is in the best case a navigation sensor that has complementary characteristics, e.g., position data from GNSS or altitude information from a barometric altimeter. Given the INS and other exteroceptive sources of data, the aim of a Kalman filter approach is to use the statistical characteristics of their errors and to determine the optimal combination of the information available, such that the errors of the estimates are minimized statistically [103].

By integrating gyroscope readings, the indirect filter becomes reliable for high frequent motion, thus should be corrected by a sensor with a complementary noise profile, which is small at lower frequencies/motion [125].

---

[14]Gimbal-based INS tend to drift at slow rates, but still unbounded, meaning that in long term, the content of the data is more accurate. These systems are generally bigger, heavier, more expensive, and are found in spacecrafts, aircrafts and submarines.

Some successful sensor constellations for AINS are, e.g., loosely-coupled barometric fusion [15], tightly-coupled UWB-aided inertial odometry [61], tightly-coupled GNSS-aided inertial odometry [22], VIO [110], VINS [46], radar-aided inertial odometry [107], LIDAR-aided inertial odometry [155], thermal camera-aided inertial odometry [33], loosely-coupled visual inertial SLAM [150], and other multi-sensor based approaches, e.g., [18, 36, 58, 87].

In the following sections, we briefly describe the system- and sensor models used in the filter-based aided INS for our evaluations. As the Jacobians in indirect filter formulations depend on the error-state definition, it can be a laborious process to derive Jacobians. Therefore, we provide a generic set of Jacobians for our error definition (the type-1 error as discussed in Section 2.4.1) and filter formulations.

### 2.9.1 Inertial Measurement Unit (IMU)

As state propagation sensor, an IMU is used to predict the position, velocity and orientation based on noisy and biased accelerometer and gyroscope measurements.

The measured linear acceleration and angular velocity of the IMU's body, are assumed to be noisy and biased

$$_{\mathcal{I}}\mathbf{a}^{\#} = {}_{\mathcal{I}}\mathbf{a} + ({}^{\mathcal{G}}\mathbf{R}_{\mathcal{I}})^{\mathsf{T}}{}_{\mathcal{G}}\mathbf{g} + {}_{\mathcal{I}}\mathbf{b}_{\mathbf{a}} + \mathbf{n}_{\mathbf{a}} = {}_{\mathcal{I}}\mathbf{a} + {}_{\mathcal{I}}\mathbf{g} + {}_{\mathcal{I}}\mathbf{b}_{\mathbf{a}} + \mathbf{n}_{\mathbf{a}}, \tag{2.79a}$$

$$_{\mathcal{I}}\boldsymbol{\omega}^{\#} = {}_{\mathcal{I}}\boldsymbol{\omega} + {}_{\mathcal{I}}\mathbf{b}_{\boldsymbol{\omega}} + \mathbf{n}_{\boldsymbol{\omega}}, \tag{2.79b}$$

with $\mathbf{n}_{\mathbf{a}} \sim \mathcal{N}\left(\mathbf{0}, \boldsymbol{\Sigma}_{\mathbf{a}}\right)$ and $\mathbf{n}_{\boldsymbol{\omega}} \sim \mathcal{N}\left(\mathbf{0}, \boldsymbol{\Sigma}_{\boldsymbol{\omega}}\right)$. $_{\mathcal{I}}\mathbf{b}_{\mathbf{a}}$ and $_{\mathcal{I}}\mathbf{b}_{\boldsymbol{\omega}}$ are the accelerometer and gyroscope biases, respectively. The known gravitational acceleration is aligned with the vertically upward z-axis of the global reference frame $\{\mathcal{G}\}$ $_{\mathcal{G}}\mathbf{g} = \begin{bmatrix} 0; 0; g \end{bmatrix}$ which is not changing $_{\mathcal{G}}\dot{\mathbf{g}} = \mathbf{0}$ and with a magnitude of $g \approx 9.81 \,\mathrm{m/s^2}$. We are ignoring the Coriolis effect, earth rotation, axis miss-alignments. We assume that the origin of both sensor, the accelerometer and gyroscope, coincide.

We assume that the additive white Gaussian noise on the IMU measurements is equal on all three axes, meaning that the noise and perturbations are isotropic. The same assumption is made for the random-walk on the biases. The following standard deviations are typically found in IMU datasets [136]

$$\sigma_{\mathbf{a}} \, [m/s^2], \ \sigma_{\boldsymbol{\omega}} \, [rad/s], \ \sigma_{\mathbf{b}_{\mathbf{a}}} \, [m/s^2\sqrt{s}], \ \sigma_{\mathbf{b}_{\boldsymbol{\omega}}} \, [rad/s\sqrt{s}], \tag{2.80}$$

leading to the covariance matrices with diagonal elements

$$\boldsymbol{\Sigma}_{\mathbf{a}} = \left[\mathbf{I}\sigma_{\mathbf{a}}^2\right], \boldsymbol{\Sigma}_{\boldsymbol{\omega}} = \left[\mathbf{I}\sigma_{\boldsymbol{\omega}}^2\right], \boldsymbol{\Sigma}_{\mathbf{b}_{\mathbf{a}}} = \left[\mathbf{I}\sigma_{\mathbf{b}_{\mathbf{a}}}^2\right], \boldsymbol{\Sigma}_{\mathbf{b}_{\boldsymbol{\omega}}} = \left[\mathbf{I}\sigma_{\mathbf{b}_{\boldsymbol{\omega}}}^2\right]. \tag{2.81}$$

Note, if an ideal accelerometer is stationary $_{\mathcal{I}}\mathbf{a} = \mathbf{0}$ and aligned with the global reference frame $^{\mathcal{G}}\mathbf{R}_{\mathcal{I}} = \mathbf{I}$, it measures an upward positive acceleration in the z-axis due to the earth's gravity field of $_{\mathcal{I}}\mathbf{a}^{\#} = \mathbf{0} + {}_{\mathcal{G}}\mathbf{g}$ and is sometimes referred to as *specific acceleration* as it has included the Newtonian gravity force acting on the internal reference/proof mass of the IMU [26].

#### System kinematics in continuous time

The dynamics of the IMU true/total state $\mathbf{x}_I$ are

$$_{\mathcal{G}}^{\mathcal{G}}\dot{\mathbf{p}}_{\mathcal{I}} = {}_{\mathcal{G}}^{\mathcal{G}}\mathbf{v}_{\mathcal{I}}, \tag{2.82a}$$

$$_{\mathcal{G}}^{\mathcal{G}}\dot{\mathbf{v}}_{\mathcal{I}} = {}_{\mathcal{G}}^{\mathcal{G}}\mathbf{a}_{\mathcal{I}} = {}^{\mathcal{G}}\mathbf{R}_{\mathcal{I}\mathcal{I}}\mathbf{a} = {}^{\mathcal{G}}\mathbf{R}_{\mathcal{I}}({}_{\mathcal{I}}\mathbf{a}^{\#} - {}_{\mathcal{I}}\mathbf{b}_{\mathbf{a}} - \mathbf{n}_{\mathbf{a}}) - {}_{\mathcal{G}}\mathbf{g} \tag{2.82b}$$

$$^{\mathcal{G}}\dot{\mathbf{R}}_{\mathcal{I}} = {}^{\mathcal{G}}\mathbf{R}_{\mathcal{I}} \left[_{\mathcal{I}}\boldsymbol{\omega}\right]_{\times} = {}^{\mathcal{G}}\mathbf{R}_{\mathcal{I}} \left[_{\mathcal{I}}\boldsymbol{\omega}^{\#} - {}_{\mathcal{I}}\mathbf{b}_{\boldsymbol{\omega}} - \mathbf{n}_{\boldsymbol{\omega}}\right]_{\times}, \tag{2.82c}$$

$$_{\mathcal{I}}\dot{\mathbf{b}}_{\mathbf{a}} = \mathbf{n}_{\mathbf{b}_{\mathbf{a}}}, \tag{2.82d}$$

$$_{\mathcal{I}}\dot{\mathbf{b}}_{\boldsymbol{\omega}} = \mathbf{n}_{\mathbf{b}_{\boldsymbol{\omega}}}, \tag{2.82e}$$

with $^{\mathcal{G}}_{\mathcal{G}}\{\mathbf{p}, \mathbf{v}, \mathbf{R}\}_{\mathcal{I}}$ as the position, velocity and orientation of the IMU $\{\mathcal{I}\}$ expressed in the global navigation frame $\{\mathcal{G}\}$. $_{\mathcal{I}}\mathbf{a}$ is the acceleration of the IMU's body, and $_{\mathcal{I}}\mathbf{a}^{\#}$ is the measured specific acceleration. Note that we need to subtract the gravitation vector from the measured acceleration to obtain the acceleration of the IMU frame $\{\mathcal{I}\}$. $\mathbf{n}_{\mathbf{b}_{\mathbf{a}}} \sim \mathcal{N}\left(\mathbf{0}, \boldsymbol{\Sigma}_{\mathbf{b}_{\mathbf{a}}}\right)$ and $\mathbf{n}_{\mathbf{b}_{\boldsymbol{\omega}}} \sim \mathcal{N}\left(\mathbf{0}, \boldsymbol{\Sigma}_{\mathbf{b}_{\boldsymbol{\omega}}}\right)$ is the random walk noise of the biases.

We can summarize the kinematics in the form $\dot{\mathbf{x}}_I = f_I(\mathbf{x}_I, \mathbf{u}, \mathbf{w})$ with $\mathbf{x}_I = \left[^{\mathcal{G}}_{\mathcal{G}}\mathbf{p}_{\mathcal{I}}; ^{\mathcal{G}}_{\mathcal{G}}\mathbf{v}_{\mathcal{I}}; ^{\mathcal{G}}\mathbf{R}_{\mathcal{I}}; _{\mathcal{I}}\mathbf{b}_{\mathbf{a}}; _{\mathcal{I}}\mathbf{b}_{\boldsymbol{\omega}}\right]$, the control input $\mathbf{u} = \begin{bmatrix} _{\mathcal{I}}\mathbf{a}^{\#} \\ _{\mathcal{I}}\boldsymbol{\omega}^{\#} \end{bmatrix}$ and the Gaussian white noise $\mathbf{w} = \left[\mathbf{n}_{\mathbf{a}}; \mathbf{n}_{\boldsymbol{\omega}}; \mathbf{n}_{\mathbf{b}_{\mathbf{a}}}; \mathbf{n}_{\mathbf{b}_{\boldsymbol{\omega}}}\right] \sim \mathcal{N}\left(\mathbf{0}, \mathbf{Q}_I\right)$.

For the ESEKF formation the nominal-state $\bar{\mathbf{x}}_I$ is defined without perturbations and noise

$$^{\mathcal{G}}_{\mathcal{G}}\dot{\bar{\mathbf{p}}}_{\mathcal{I}} = {}^{\mathcal{G}}_{\mathcal{G}}\bar{\mathbf{v}}_{\mathcal{I}}, \tag{2.83a}$$

$$^{\mathcal{G}}_{\mathcal{G}}\dot{\bar{\mathbf{v}}}_{\mathcal{I}} = {}^{\mathcal{G}}\bar{\mathbf{R}}_{\mathcal{I}}(_{\mathcal{I}}\mathbf{a}^{\#} - {}_{\mathcal{I}}\bar{\mathbf{b}}_{\mathbf{a}}) - {}_{\mathcal{G}}\mathbf{g}, \tag{2.83b}$$

$$^{\mathcal{G}}\dot{\bar{\mathbf{R}}}_{\mathcal{I}} = {}^{\mathcal{G}}\bar{\mathbf{R}}_{\mathcal{I}} \left[_{\mathcal{I}}\boldsymbol{\omega}^{\#} - {}_{\mathcal{I}}\bar{\mathbf{b}}_{\boldsymbol{\omega}}\right]_{\times}, \tag{2.83c}$$

$$_{\mathcal{I}}\dot{\bar{\mathbf{b}}}_{\mathbf{a}} = \mathbf{0}, \tag{2.83d}$$

$$_{\mathcal{I}}\dot{\bar{\mathbf{b}}}_{\boldsymbol{\omega}} = \mathbf{0}, \tag{2.83e}$$

Therefore, the nominal-state of the IMU lies on a product manifold consisting of $\bar{\mathbf{x}}_I \in \{\mathbb{R}^6 \times SO(3) \times \mathbb{R}^6\}$ with 15 DoFs [46]. The kinematics of the nominal-state is in from $\dot{\bar{\mathbf{x}}}_I = f_I(\bar{\mathbf{x}}_I, \mathbf{u})$ with $\bar{\mathbf{x}}_I = \left[^{\mathcal{G}}_{\mathcal{G}}\bar{\mathbf{p}}_{\mathcal{I}}; ^{\mathcal{G}}_{\mathcal{G}}\bar{\mathbf{v}}_{\mathcal{I}}; ^{\mathcal{G}}\bar{\mathbf{R}}_{\mathcal{I}}; _{\mathcal{I}}\bar{\mathbf{b}}_{\mathbf{a}}; _{\mathcal{I}}\bar{\mathbf{b}}_{\boldsymbol{\omega}}\right]$ and the control input $\mathbf{u} = \begin{bmatrix} _{\mathcal{I}}\mathbf{a}^{\#} \\ _{\mathcal{I}}\boldsymbol{\omega}^{\#} \end{bmatrix}$.

The kinematics of the type-1 errors-state is the difference between the total- and nominal-state's kinematic $\dot{\tilde{\mathbf{x}}}_I = \boxminus \dot{\bar{\mathbf{x}}}_I \boxplus \dot{\mathbf{x}}_I$[15]. For rotations, we use the small angle approximation describing a local rotation error

$$^{\mathcal{G}}\mathbf{R}_{\mathcal{I}} \approx {}^{\mathcal{G}}\bar{\mathbf{R}}_{\bar{\mathcal{I}}} \left(\mathbf{I} + \left[^{\bar{\mathcal{I}}}_{\mathcal{I}}\tilde{\boldsymbol{\theta}}_{\mathcal{I}}\right]_{\times}\right) \tag{2.84a}$$

and we assume local position and velocity errors, such that

$$^{\mathcal{G}}_{\mathcal{G}}\{\mathbf{p}, \mathbf{v}\}_{\mathcal{I}} = {}^{\mathcal{G}}_{\mathcal{G}}\{\bar{\mathbf{p}}, \bar{\mathbf{v}}\}_{\bar{\mathcal{I}}} + {}^{\mathcal{G}}\bar{\mathbf{R}}_{\bar{\mathcal{I}}} \left(^{\bar{\mathcal{I}}}_{\bar{\mathcal{I}}}\{\tilde{\mathbf{p}}, \tilde{\mathbf{v}}\}_{\mathcal{I}}\right) \tag{2.84b}$$

the biases are additive

$$_{\mathcal{I}}\mathbf{b}_{\{\mathbf{a}, \boldsymbol{\omega}\}} = {}_{\mathcal{I}}\bar{\mathbf{b}}_{\{\mathbf{a}, \boldsymbol{\omega}\}} + {}_{\mathcal{I}}\tilde{\mathbf{b}}_{\{\mathbf{a}, \boldsymbol{\omega}\}} \tag{2.84c}$$

---

[15]The order matters, since not all elements of the state vector are commutative, e.g., the rotation $^{\mathcal{G}}\mathbf{R}_{\mathcal{I}} \in SO(3)$.

The error-state of the IMU lies on a product manifold consisting of $\bar{\mathbf{x}}_I \in \{\mathbb{R}^3 \times \mathbb{R}^3 \times \mathbb{R}^3 \times \mathbb{R}^3 \times \mathbb{R}^3\}$ with 15-DoFs. The kinematics of the error-state can be described after neglecting second-order terms by

$$\dot{\bar{\bar{\mathbf{p}}}}_{\mathcal{I}}^{\bar{\mathcal{I}}} = \dot{\bar{\bar{\mathbf{v}}}}_{\mathcal{I}}^{\bar{\mathcal{I}}}, \tag{2.84d}$$

$$\dot{\bar{\bar{\mathbf{v}}}}_{\mathcal{I}}^{\bar{\mathcal{I}}} \approx - \left[ {}_{\mathcal{I}}\mathbf{a}^{\#} - \bar{\mathbf{b}}_{\mathbf{a}} \right]_{\times} \tilde{\bar{\boldsymbol{\theta}}}_{\mathcal{I}}^{\bar{\mathcal{I}}} - \left[ {}_{\mathcal{I}}\boldsymbol{\omega}^{\#} - {}_{\mathcal{I}}\bar{\mathbf{b}}_{\boldsymbol{\omega}} \right]_{\times} \tilde{\bar{\mathbf{v}}}_{\mathcal{I}}^{\bar{\mathcal{I}}} - {}_{\mathcal{I}}\tilde{\mathbf{b}}_{\mathbf{a}} - \mathbf{n_a} \tag{2.84e}$$

$$\dot{\bar{\mathbf{R}}}_{\mathcal{I}}^{\mathcal{G}} \approx - \left[ {}_{\mathcal{I}}\boldsymbol{\omega}^{\#} - {}_{\mathcal{I}}\bar{\mathbf{b}}_{\boldsymbol{\omega}} \right]_{\times} \tilde{\bar{\mathbf{R}}}_{\mathcal{I}} - {}_{\mathcal{I}}\tilde{\mathbf{b}}_{\boldsymbol{\omega}} - \mathbf{n_{\boldsymbol{\omega}}} \tag{2.84f}$$

$$\dot{{}_{\mathcal{I}}\tilde{\mathbf{b}}}_{\mathbf{a}} = \mathbf{n_{b_a}}, \tag{2.84g}$$

$$\dot{{}_{\mathcal{I}}\tilde{\mathbf{b}}}_{\boldsymbol{\omega}} = \mathbf{n_{b_{\boldsymbol{\omega}}}}, \tag{2.84h}$$

The continuous-time error dynamics can be rearranged in the form

$$\dot{\bar{\mathbf{x}}} = f_I(\bar{\mathbf{x}}, \tilde{\mathbf{x}}, \mathbf{u}, \mathbf{w}) = \mathbf{F}\bar{\mathbf{x}} + \mathbf{G}\mathbf{w} \tag{2.85}$$

with the control input $\mathbf{u} = \begin{bmatrix} {}_{\mathcal{I}}\mathbf{a}^{\#} - {}_{\mathcal{I}}\bar{\mathbf{b}}_{\mathbf{a}} \\ {}_{\mathcal{I}}\boldsymbol{\omega}^{\#} - {}_{\mathcal{I}}\bar{\mathbf{b}}_{\boldsymbol{\omega}} \end{bmatrix}$ the noise vector $\mathbf{w} = \begin{bmatrix} \mathbf{n_a}; \mathbf{n_{\boldsymbol{\omega}}}; \mathbf{n_{b_a}}; \mathbf{n_{b_{\boldsymbol{\omega}}}} \end{bmatrix} \sim \mathcal{N}(\mathbf{0}, \mathbf{Q}_I)$, and

$$\mathbf{F} = \begin{bmatrix} \mathbf{0} & \mathbf{I} & \mathbf{0} & \mathbf{0} & \mathbf{0} \\ \mathbf{0} & -\left[ {}_{\mathcal{I}}\boldsymbol{\omega}^{\#} - {}_{\mathcal{I}}\bar{\mathbf{b}}_{\boldsymbol{\omega}} \right]_{\times} & -\left[ {}_{\mathcal{I}}\mathbf{a}^{\#} - \bar{\mathbf{b}}_{\mathbf{a}} \right]_{\times} & \mathbf{0} & -\mathbf{I} \\ \mathbf{0} & \mathbf{0} & -\left[ {}_{\mathcal{I}}\boldsymbol{\omega}^{\#} - {}_{\mathcal{I}}\bar{\mathbf{b}}_{\boldsymbol{\omega}} \right]_{\times} & -\mathbf{I} & \mathbf{0} \\ \mathbf{0} & \mathbf{0} & \mathbf{0} & \mathbf{0} & \mathbf{0} \\ \mathbf{0} & \mathbf{0} & \mathbf{0} & \mathbf{0} & \mathbf{0} \end{bmatrix}, \mathbf{G} = \begin{bmatrix} \mathbf{0} & \mathbf{0} & \mathbf{0} & \mathbf{0} \\ \mathbf{0} & -\mathbf{I} & \mathbf{0} & \mathbf{0} \\ -\mathbf{I} & \mathbf{0} & \mathbf{0} & \mathbf{0} \\ \mathbf{0} & \mathbf{0} & \mathbf{I} & \mathbf{0} \\ \mathbf{0} & \mathbf{0} & \mathbf{0} & \mathbf{I} \end{bmatrix}. \tag{2.86}$$

**Linear velocity error kinematics** Inspired by [136], the linear velocity error in Equation (2.84e) is obtained by writing the true velocity differential equation (see Equation (2.82b)) in two different forms, to the left using the nominal- and error-state, and to the right by the true kinematics

$$\frac{d}{dt}\left[ {}_{\mathcal{G}}^{\mathcal{G}}\bar{\mathbf{v}}_{\bar{\mathcal{I}}} + {}^{\mathcal{G}}\bar{\mathbf{R}}_{\bar{\mathcal{I}}\mathcal{I}}^{\bar{\mathcal{I}}}\tilde{\mathbf{v}}_{\mathcal{I}} \right] = {}_{\mathcal{G}}^{\mathcal{G}}\dot{\mathbf{v}}_{\mathcal{I}} = {}^{\mathcal{G}}\mathbf{R}_{\mathcal{I}}\left( {}_{\mathcal{I}}\mathbf{a}^{\#} - {}_{\mathcal{I}}\mathbf{b}_{\mathbf{a}} - \mathbf{n_a} \right) - {}_{\mathcal{G}}\mathbf{g} \tag{2.87}$$

After applying the chain rule $f(t)\dot{g}(t) = \dot{f}(t)g(t) + f(t)\dot{g}(t)$ the left side expands to

$$ {}^{\mathcal{G}}\dot{\bar{\mathbf{v}}}_{\bar{\mathcal{I}}} + {}^{\mathcal{G}}\dot{\bar{\mathbf{R}}}_{\mathcal{I}\bar{\mathcal{I}}}^{\bar{\mathcal{I}}}\tilde{\mathbf{v}}_{\mathcal{I}} + {}^{\mathcal{G}}\bar{\mathbf{R}}_{\mathcal{I}\bar{\mathcal{I}}}^{\bar{\mathcal{I}}}\dot{\tilde{\mathbf{v}}}_{\mathcal{I}} = {}_{\mathcal{G}}^{\mathcal{G}}\dot{\mathbf{v}}_{\mathcal{I}} \tag{2.88}$$

By inserting Equation (2.83b) and Equation (2.83c) is expands to

$$ {}^{\mathcal{G}}\bar{\mathbf{R}}_{\mathcal{I}}({}_{\mathcal{I}}\mathbf{a}^{\#} - {}_{\mathcal{I}}\bar{\mathbf{b}}_{\mathbf{a}}) - {}_{\mathcal{G}}\mathbf{g} + {}^{\mathcal{G}}\bar{\mathbf{R}}_{\mathcal{I}}\left[ {}_{\mathcal{I}}\boldsymbol{\omega}^{\#} - {}_{\mathcal{I}}\bar{\mathbf{b}}_{\boldsymbol{\omega}} \right]_{\times} \tilde{\bar{\mathbf{v}}}_{\mathcal{I}}^{\bar{\mathcal{I}}} + {}^{\mathcal{G}}\bar{\mathbf{R}}_{\mathcal{I}\bar{\mathcal{I}}}^{\bar{\mathcal{I}}}\dot{\tilde{\mathbf{v}}}_{\mathcal{I}} = {}_{\mathcal{G}}^{\mathcal{G}}\dot{\mathbf{v}}_{\mathcal{I}} \tag{2.89}$$

The right side can be expanded by inserting the error definition in Equation (2.84a) and in Equation (2.84c)

$$ {}_{\mathcal{G}}^{\mathcal{G}}\dot{\mathbf{v}}_{\mathcal{I}} = {}^{\mathcal{G}}\bar{\mathbf{R}}_{\bar{\mathcal{I}}}\left( \mathbf{I} + \left[ \tilde{\bar{\boldsymbol{\theta}}}_{\mathcal{I}}^{\bar{\mathcal{I}}} \right]_{\times} \right)\left( {}_{\mathcal{I}}\mathbf{a}^{\#} - ({}_{\mathcal{I}}\bar{\mathbf{b}}_{\mathbf{a}} + {}_{\mathcal{I}}\tilde{\mathbf{b}}_{\mathbf{a}}) - \mathbf{n_a} \right) - {}_{\mathcal{G}}\mathbf{g}. \tag{2.90}$$

After eliminating $_{\mathcal{G}}\mathbf{g}$, a right multiplication by $^{\mathcal{G}}\bar{\mathbf{R}}_{\bar{\mathcal{I}}}{}^{\mathsf{T}}$, rearranging skew symmetric matrices by $[\mathbf{a}]_{\times}\mathbf{b} = [-\mathbf{b}]_{\times}\mathbf{a}$, and further rearrangements to isolate $_{\bar{\mathcal{I}}}^{\bar{\mathcal{I}}}\dot{\tilde{\mathbf{v}}}_{\mathcal{I}}$, we arrive at the differential equation for the velocity error

$$_{\bar{\mathcal{I}}}^{\bar{\mathcal{I}}}\dot{\tilde{\mathbf{v}}}_{\mathcal{I}} = -\left[_{\mathcal{I}}\boldsymbol{\omega}^{\#} - _{\mathcal{I}}\bar{\mathbf{b}}_{\boldsymbol{\omega}}\right]_{\times}{}_{\bar{\mathcal{I}}}^{\bar{\mathcal{I}}}\tilde{\mathbf{v}}_{\mathcal{I}} - \left[_{\mathcal{I}}\mathbf{a}^{\#} - \bar{\mathbf{b}}_{\mathbf{a}}\right]_{\times}{}^{\bar{\mathcal{I}}}\tilde{\boldsymbol{\theta}}_{\mathcal{I}} + \left[\tilde{\mathbf{b}}_{\mathbf{a}}\right]_{\times}{}^{\bar{\mathcal{I}}}\tilde{\boldsymbol{\theta}}_{\mathcal{I}} - \tilde{\mathbf{b}}_{\mathbf{a}} - \left(\mathbf{I} + \left[^{\bar{\mathcal{I}}}\tilde{\boldsymbol{\theta}}_{\mathcal{I}}\right]_{\times}\right)\mathbf{n}_{\mathbf{a}}$$
(2.91)

By neglecting higher-order terms, we arrive finally at Equation (2.84e).

**Local orientation error kinematics**   The local orientation error kinematics are equal to the one presented in [136], for details on the please follow the reference.

**System kinematics in discrete time**

In the ESEKF formulation, we are interested in the integrated nominal-state and the predicted error-state's covariance (the mean of the error-state is and remains always zero). To propagate the covariance of the error-state from $t^{k-1}$ to $t^k$, we need to compute the discrete-time process noise matrix $\mathbf{Q}_d^{k-1} = \mathbf{G}^k\mathbf{Q}_i^{k-1}(\mathbf{G}^k)^{\mathsf{T}}$ and the discrete-time state transition matrix $\boldsymbol{\Phi}^{k|k-1}$. The discrete-time error dynamics are in the form

$$\tilde{\mathbf{x}}^k = \boldsymbol{\Phi}^{k|k-1}\tilde{\mathbf{x}}^{k-1} + \mathbf{G}^k\mathbf{i}^k,$$
(2.92)

where $\mathbf{i} \sim \mathcal{N}\left(\mathbf{0}, \mathbf{Q}_i\right) = \left[\mathbf{i}_{\mathbf{a}}; \mathbf{i}_{\boldsymbol{\omega}}; \mathbf{i}_{\mathbf{b}_{\mathbf{a}}}; \mathbf{i}_{\mathbf{b}_{\boldsymbol{\omega}}}\right]$ is a vector of random Gaussian impulses. This leads to the corresponding covariance propagation in the form

$$\boldsymbol{\Sigma}^k = \boldsymbol{\Phi}^{k|k-1}\boldsymbol{\Sigma}^{k-1}(\boldsymbol{\Phi}^{k|k-1})^{\mathsf{T}} + \mathbf{G}^k\mathbf{Q}_i^{k-1}(\mathbf{G}^k)^{\mathsf{T}},$$
(2.93)

In order to obtain the system kinematics of the error-state in discrete time, the differential equations need to be integrated into differences equations to account for the sampling intervals $\Delta t$ of the IMU [136]. In our case, we perform a numerical integration of the state transition matrix by a truncated Tyler-expansion $\boldsymbol{\Phi}^{k|k-1} = \exp(\mathbf{F}\Delta t) \approx \sum_{k=0}^{3}\frac{1}{k!}\mathbf{F}^k\Delta t^k$.

Given the covariances of the noise and perturbations in continuous-time $\mathbf{w}$ (see Equation (2.81)), we need to integrate covariance of these random variables for the uncertainty propagation of the error-state. Since the measurement noise and the perturbations, caused by the random-walk behavior, are different stochastic processes, their integration over $\Delta t$ differs. The later results in discrete white Gaussian impulse [136]

$$\mathbf{i}_{\mathbf{b}_{\{\mathbf{a},\boldsymbol{\omega}\}}} \sim \mathcal{N}\left(\mathbf{0}, \Delta t\boldsymbol{\Sigma}_{\mathbf{b}_{\{\mathbf{a},\boldsymbol{\omega}\}}}\right)$$
(2.94)

and the former leads to a quadratic term

$$\mathbf{i}_{\{\mathbf{a},\boldsymbol{\omega}\}} \sim \mathcal{N}\left(\mathbf{0}, \Delta t^2\boldsymbol{\Sigma}_{\{\mathbf{a},\boldsymbol{\omega}\}}\right),$$
(2.95)

leading to the impulse variance $\mathbf{Q}_i$

$$\mathbf{Q}_i = \text{Diag}\left(\left[\Delta t^2\boldsymbol{\Sigma}_{\mathbf{a}}, \Delta t^2\boldsymbol{\Sigma}_{\boldsymbol{\omega}}, \Delta t\boldsymbol{\Sigma}_{\mathbf{b}_{\mathbf{a}}}, \Delta t\boldsymbol{\Sigma}_{\mathbf{b}_{\boldsymbol{\omega}}}\right]\right).$$
(2.96)

### 2.9.2   Zero Velocity Update

If we knew that the INS is standing still, e.g., be detecting a contact with a surface we assume to remain static/fixed in the global navigation frame $\{\mathcal{G}\}$, the body acceleration, body velocity as shown in Figure 2.8, and angular velocity have to be zero, which is an ideal case to observe the biases in the IMU measurements. The concept of zero-velocity updates, was applied in, e.g., contact aided inertial navigation for legged robots by Hartley *et al.* in [55] or in VINSs, e.g., in [46].

**Figure 2.8:** Contact detection between a rigid body configuration and surface that is fixed in the world reference.

Once a stand-still is detected, we induce noisy *pseudo* observation in the form

$$\mathbf{z_0} = \begin{bmatrix} {}^{\mathcal{G}}_{\mathcal{G}}\mathbf{v}_{\mathcal{I}} + \mathbf{v_v} = \mathbf{0} \\ {}_{\mathcal{I}}\mathbf{a} + \mathbf{v_a} = \mathbf{0} \\ {}_{\mathcal{I}}\boldsymbol{\omega} + \mathbf{v_{\omega}} = \mathbf{0} \end{bmatrix} = h_{\mathbf{0}}(\mathbf{x}) + \mathbf{v}, \tag{2.97}$$

with a measurement noise to account for potential vibrations and slippage $\mathbf{v} = \begin{bmatrix} \mathbf{v_v} \\ \mathbf{v_a} \\ \mathbf{v_{\omega}} \end{bmatrix} \sim$ $\mathcal{N}\left(\mathbf{0}, \text{Diag}\left(\mathbf{R_v}, \mathbf{R_a}, \mathbf{R_{\omega}}\right)\right)$.

According to the documentation of OpenVINS [46], we can relate the zero velocity measurement to the latest IMU readings. Recalling the IMU measurement defined in Equation (2.79), we obtain

$${}_{\mathcal{I}}\mathbf{a} = \mathbf{0} = {}_{\mathcal{I}}\mathbf{a}^{\#} - ({}^{\mathcal{G}}\mathbf{R}_{\mathcal{I}})^{\mathsf{T}}{}_{\mathcal{G}}\mathbf{g} - {}_{\mathcal{I}}\mathbf{b_a} - \mathbf{n_a},$$

$${}_{\mathcal{I}}\boldsymbol{\omega} = \mathbf{0} = {}_{\mathcal{I}}\boldsymbol{\omega}^{\#} - {}_{\mathcal{I}}\mathbf{b_{\omega}} - \mathbf{n_{\omega}}.$$

In the indirect filter formulation, the measurement presented to the filter is the difference between the estimated and actual measurement, following our error definition $\tilde{\mathbf{z}} = \hat{\mathbf{z}}^{-1}\mathbf{z}^{\#}$, and leads to

$$\tilde{\mathbf{z}}_{\mathbf{a}} = {}_{\mathcal{I}}\mathbf{a} - {}_{\mathcal{I}}\mathbf{a}^{\#} - ({}^{\mathcal{G}}\mathbf{R}_{\mathcal{I}})^{\mathsf{T}}{}_{\mathcal{G}}\mathbf{g} - {}_{\mathcal{I}}\mathbf{b_a} - \mathbf{n_a}, \tag{2.99a}$$

$$\tilde{\mathbf{z}}_{\boldsymbol{\omega}} = {}_{\mathcal{I}}\boldsymbol{\omega} - {}_{\mathcal{I}}\boldsymbol{\omega}^{\#} - {}_{\mathcal{I}}\mathbf{b_{\omega}} - \mathbf{n_{\omega}} \tag{2.99b}$$

$$\tilde{\mathbf{z}}_{\mathbf{v}} = {}^{\mathcal{G}}_{\mathcal{G}}\mathbf{v}_{\mathcal{I}} - {}^{\mathcal{G}}_{\mathcal{G}}\hat{\mathbf{v}}_{\mathcal{I}}, \tag{2.99c}$$

We can linearize this measurement model around the current estimated state $\hat{\mathbf{x}}$ to obtain the measurement Jacobian $\mathbf{H} = \begin{bmatrix} \mathbf{H_v} \\ \mathbf{H_a} \\ \mathbf{H_{\omega}} \end{bmatrix} = \frac{\partial \tilde{\mathbf{z}}}{\partial \mathbf{x}}\big|_{\hat{\mathbf{x}}} \frac{\partial \mathbf{x}}{\partial \tilde{\mathbf{x}}}\big|$ for our filter formulation.

Therefore, we express the measured position error by replacing the true value in Equation (2.101b) with the definition of the total state and insert it in Equation (2.102a) to

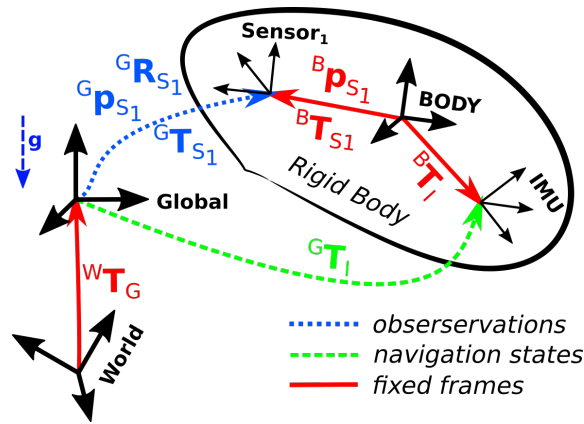

**Figure 2.9:** Spatial frame constellation of an absolute position, orientation, or pose sensor with respect to the navigation frame $\{\mathcal{G}\}$.

derive the partial derivatives with respect to the error-states

$$\left. \frac{\partial \tilde{\mathbf{z}}_{\mathbf{v}}}{\partial {}^{\mathcal{G}}_{\mathcal{G}} \tilde{\mathbf{v}}_{\mathcal{I}}} \right|_{\hat{\mathbf{x}}} = {}^{\mathcal{G}} \hat{\mathbf{R}}_{\mathcal{I}} \tag{2.100a}$$

$$\left. \frac{\partial \tilde{\mathbf{z}}_{\mathbf{a}}}{\partial {}^{\mathcal{G}}_{\mathcal{G}} \tilde{\boldsymbol{\omega}}_{\mathcal{I}}} \right|_{\hat{\mathbf{x}}} = - \left[ {}^{\mathcal{G}} \hat{\mathbf{R}}_{\mathcal{I}}^{\mathsf{T}} {}_{\mathcal{G}} \mathbf{g} \right]_{\times} \tag{2.100b}$$

$$\left. \frac{\partial \tilde{\mathbf{z}}_{\mathbf{a}}}{\partial {}_{\mathcal{I}} \tilde{\mathbf{b}}_{\mathbf{a}}} \right|_{\hat{\mathbf{x}}} = -\mathbf{I} \tag{2.100c}$$

$$\left. \frac{\partial \tilde{\mathbf{z}}_{\boldsymbol{\omega}}}{\partial {}_{\mathcal{I}} \tilde{\mathbf{b}}_{\boldsymbol{\omega}}} \right|_{\hat{\mathbf{x}}} = -\mathbf{I} \tag{2.100d}$$

### 2.9.3   Absolute Position, Orientation, and Pose

Absolute information with respect to the navigation frame is needed to correct the drifting estimates provided by the INS. A rudimentary and generic way to model noisy absolute position, pose, or orientations is shown in Figure 2.9, by defining displacements between the IMU, the body reference, and the sensor that is obtaining the information. Outdoors, GNSS receivers, provide typically earth-fixed ellipsoidal latitude, longitude and altitude above mean-sea-level estimates that stem from a multilateration of range measurements to at least four satellites in line of sight (LOS). To determine the distances, the propagation time of the electromagnetic signal between the satellites and the receiver is measured [131]. Given an earth surface model, these polar angles can be converted into earth-fixed and earth-centered position information in Cartesian coordinates. Since, we assume that the navigation frame is near the earth's crust and that the z-axis is aligned with the gravity vector, the pose between the earth-fixed and earth-centered reference frame and the navigation frame $\{\mathcal{G}\}$ needs to be considered.

Indoors, absolute information can be provided by additional infrastructure, such as motion capture systems, that track a reflecting rigid marker configuration on rigid body using (extrinsically and intrinsically calibrated) infrared cameras. These systems allow tracking the pose at high rates (greater 100 Hz) and a high accuracy (sub-centimeters and sub-degrees).

A generic absolute pose measurement, that considers the displacements between the sensors attached on the rigid body, is in the form

$$\mathbf{z}^{\#} = \begin{bmatrix} \mathbf{z_p} \\ \mathbf{z_R} \end{bmatrix} = \begin{bmatrix} {}^{\mathcal{G}}_{\mathcal{G}}\mathbf{p}_{\mathcal{S}} \\ {}^{\mathcal{G}}_{\mathcal{G}}\mathbf{R}_{\mathcal{S}} \end{bmatrix} + \mathbf{n} \tag{2.101a}$$

where $\mathbf{n} \sim \mathcal{N}\left(\mathbf{0}, \mathbf{R}\right)$ is a white Gaussian noise vector with the covariance matrix $\mathbf{R}$ and with

$$\begin{bmatrix} {}^{\mathcal{G}}_{\mathcal{G}}\mathbf{p}_{\mathcal{S}} \\ {}^{\mathcal{G}}_{\mathcal{G}}\mathbf{R}_{\mathcal{S}} \end{bmatrix} = \begin{bmatrix} \left( {}^{\mathcal{G}}\mathbf{T}_{\mathcal{S}} \right)_{\mathbf{p}} \\ \left( {}^{\mathcal{G}}\mathbf{T}_{\mathcal{S}} \right)_{\mathbf{R}} \end{bmatrix} = \begin{bmatrix} {}^{\mathcal{G}}_{\mathcal{G}}\mathbf{p}_{\mathcal{I}} + {}^{\mathcal{G}}\mathbf{R}_{\mathcal{I}} \left( {}^{\mathcal{B}}\mathbf{R}_{\mathcal{I}}^{\mathsf{T}} \left( -{}^{\mathcal{B}}_{\mathcal{B}}\mathbf{p}_{\mathcal{I}} + {}^{\mathcal{B}}_{\mathcal{B}}\mathbf{p}_{\mathcal{S}} \right) \right) \\ {}^{\mathcal{G}}\mathbf{R}_{\mathcal{S}} = {}^{\mathcal{G}}\mathbf{R}_{\mathcal{I}} {}^{\mathcal{B}}\mathbf{R}_{\mathcal{I}}^{\mathsf{T}} {}^{\mathcal{B}}\mathbf{R}_{\mathcal{S}} \end{bmatrix} \tag{2.101b}$$

and

$$\mathbf{^{\mathcal{G}}T}_{\mathcal{S}} = {}^{\mathcal{G}}\mathbf{T}_{\mathcal{I}} {}^{\mathcal{B}}\mathbf{T}_{\mathcal{I}}^{-1\mathcal{B}}\mathbf{T}_{\mathcal{S}}. \tag{2.101c}$$

In the indirect filter formulation, the measurement presented to the filter is the difference between the estimated and actual measurement, following our error definition $\tilde{\mathbf{z}} = \hat{\mathbf{z}}^{-1}\mathbf{z}^{\#}$, and leads to

$$\tilde{\mathbf{z}} = \begin{bmatrix} \tilde{\mathbf{z}}_{\mathbf{p}} \\ \tilde{\mathbf{z}}_{\mathbf{R}} \end{bmatrix} = \begin{bmatrix} \left( {}^{\mathcal{G}}\tilde{\mathbf{T}}_{\mathcal{S}} \right)_{\mathbf{p}} \\ \left( {}^{\mathcal{G}}\tilde{\mathbf{T}}_{\mathcal{S}} \right)_{\mathbf{R}} \end{bmatrix} + \mathbf{n} = \begin{bmatrix} {}^{\mathcal{G}}\hat{\mathbf{R}}_{\mathcal{S}}^{\mathsf{T}} \left( -{}^{\mathcal{G}}_{\mathcal{G}}\hat{\mathbf{p}}_{\mathcal{S}} + {}^{\mathcal{G}}_{\mathcal{G}}\mathbf{p}_{\mathcal{S}}^{\#} \right) \\ {}^{\mathcal{G}}\hat{\mathbf{R}}_{\mathcal{S}}^{\mathsf{T}\mathcal{G}}\mathbf{R}_{\mathcal{S}}^{\#} \end{bmatrix} + \mathbf{n} \tag{2.102a}$$

with

$$\mathbf{^{\mathcal{G}}\tilde{T}}_{\mathcal{S}} = {}^{\mathcal{G}}\hat{\mathbf{T}}_{\mathcal{S}}^{-1\mathcal{G}}\mathbf{T}_{\mathcal{S}}^{\#} = \begin{bmatrix} {}^{\mathcal{G}}\hat{\mathbf{R}}_{\mathcal{S}}^{\mathsf{T}} & -{}^{\mathcal{G}}\hat{\mathbf{R}}_{\mathcal{S}}^{\mathsf{T}\mathcal{G}}\hat{\mathbf{p}}_{\mathcal{S}} \\ \mathbf{0} & 1 \end{bmatrix} \begin{bmatrix} {}^{\mathcal{G}}\mathbf{R}_{\mathcal{S}}^{\#} & {}^{\mathcal{G}}_{\mathcal{G}}\mathbf{p}_{\mathcal{S}}^{\#} \\ \mathbf{0} & 0 \end{bmatrix}. \tag{2.102b}$$

The estimated measurement is obtained using the measurement model in Equation (2.101b).

We can linearize this measurement model around the current estimated state $\hat{\mathbf{x}}$ to obtain the measurement Jacobian $\mathbf{H} = \begin{bmatrix} \mathbf{H_p} \\ \mathbf{H_R} \end{bmatrix} = \left.\frac{\partial \tilde{\mathbf{z}}}{\partial \mathbf{x}}\right|_{\hat{\mathbf{x}}} \left.\frac{\partial \mathbf{x}}{\partial \mathbf{x}}\right|_{\hat{\mathbf{x}}}$ for our filter formulation.

Therefore, we express the measured position error by replacing the true value in Equation (2.101b) with the definition of the total state and insert it in Equation (2.102a) to derive the partial derivatives with respect to the error-states

$$\left.\frac{\partial \tilde{\mathbf{z}}_{\mathbf{p}}}{\partial {}^{\mathcal{G}}_{\mathcal{G}}\tilde{\mathbf{p}}_{\mathcal{I}}}\right|_{\hat{\mathbf{x}}} = \mathbf{I}^{\mathcal{G}}\hat{\mathbf{R}}_{\mathcal{I}} \tag{2.103a}$$

$$\left.\frac{\partial \tilde{\mathbf{z}}_{\mathbf{p}}}{\partial {}^{\mathcal{B}}_{\mathcal{B}}\tilde{\mathbf{p}}_{\mathcal{I}}}\right|_{\hat{\mathbf{x}}} = -{}^{\mathcal{G}}\hat{\mathbf{R}}_{\mathcal{I}} \tag{2.103b}$$

$$\left.\frac{\partial \tilde{\mathbf{z}}_{\mathbf{p}}}{\partial {}^{\mathcal{B}}_{\mathcal{B}}\tilde{\mathbf{p}}_{\mathcal{S}}}\right|_{\hat{\mathbf{x}}} = {}^{\mathcal{G}}\hat{\mathbf{R}}_{\mathcal{I}} {}^{\mathcal{B}}\hat{\mathbf{R}}_{\mathcal{I}}^{\mathsf{T}\mathcal{B}}\hat{\mathbf{R}}_{\mathcal{S}} \tag{2.103c}$$

$$\left.\frac{\partial \tilde{\mathbf{z}}_{\mathbf{p}}}{\partial {}^{\mathcal{G}}\tilde{\boldsymbol{\theta}}_{\mathcal{I}}}\right|_{\hat{\mathbf{x}}} = -{}^{\mathcal{G}}\hat{\mathbf{R}}_{\mathcal{I}} \left[ {}^{\mathcal{B}}\hat{\mathbf{R}}_{\mathcal{I}}^{\mathsf{T}} \left( -{}^{\mathcal{B}}_{\mathcal{B}}\hat{\mathbf{p}}_{\mathcal{I}} + {}^{\mathcal{B}}_{\mathcal{B}}\hat{\mathbf{p}}_{\mathcal{S}} \right) \right]_{\times} \tag{2.103d}$$

$$\left.\frac{\partial \tilde{\mathbf{z}}_{\mathbf{p}}}{\partial {}^{\mathcal{B}}\tilde{\boldsymbol{\theta}}_{\mathcal{I}}}\right|_{\hat{\mathbf{x}}} = {}^{\mathcal{G}}\hat{\mathbf{R}}_{\mathcal{I}} \left[ {}^{\mathcal{B}}\hat{\mathbf{R}}_{\mathcal{I}}^{\mathsf{T}} \left( -{}^{\mathcal{B}}_{\mathcal{B}}\hat{\mathbf{p}}_{\mathcal{I}} + {}^{\mathcal{B}}_{\mathcal{B}}\hat{\mathbf{p}}_{\mathcal{S}} \right) \right]_{\times}. \tag{2.103e}$$

Expressing the measured rotation error by replacing the true value in Equation (2.101b) with the definition of the total state, the equation in Equation (2.102a) expands to

$$\tilde{\mathbf{z}}_{\mathbf{R}} = {}^{\mathcal{B}}\hat{\mathbf{R}}_{\mathcal{S}}^{\mathsf{T}\mathcal{B}}\hat{\mathbf{R}}_{\mathcal{I}} {}^{\mathcal{G}}\hat{\mathbf{R}}_{\mathcal{I}}^{\mathsf{T}\mathcal{G}}\mathbf{R}_{\mathcal{I}} {}^{\mathcal{B}}\mathbf{R}_{\mathcal{I}}^{\mathsf{T}\mathcal{B}}\mathbf{R}_{\mathcal{S}} + \mathbf{n_R} \tag{2.104}$$

$$\tilde{\mathbf{z}}_{\mathbf{R}} = {}^{\mathcal{B}}\hat{\mathbf{R}}_{\mathcal{S}}^{\mathsf{T}\mathcal{B}}\hat{\mathbf{R}}_{\mathcal{I}} {}^{\mathcal{G}}\hat{\mathbf{R}}_{\mathcal{I}}^{\mathsf{T}\mathcal{G}}\hat{\mathbf{R}}_{\mathcal{I}} \mathbf{R}({}^{\mathcal{G}}\tilde{\boldsymbol{\theta}}_{\mathcal{I}}) \mathbf{R}({}^{\mathcal{B}}\tilde{\boldsymbol{\theta}}_{\mathcal{I}})^{\mathsf{T}\mathcal{B}}\hat{\mathbf{R}}_{\mathcal{I}}^{\mathsf{T}\mathcal{B}}\hat{\mathbf{R}}_{\mathcal{S}} \mathbf{R}({}^{\mathcal{B}}\tilde{\boldsymbol{\theta}}_{\mathcal{S}}) + \mathbf{n_R} \tag{2.105}$$

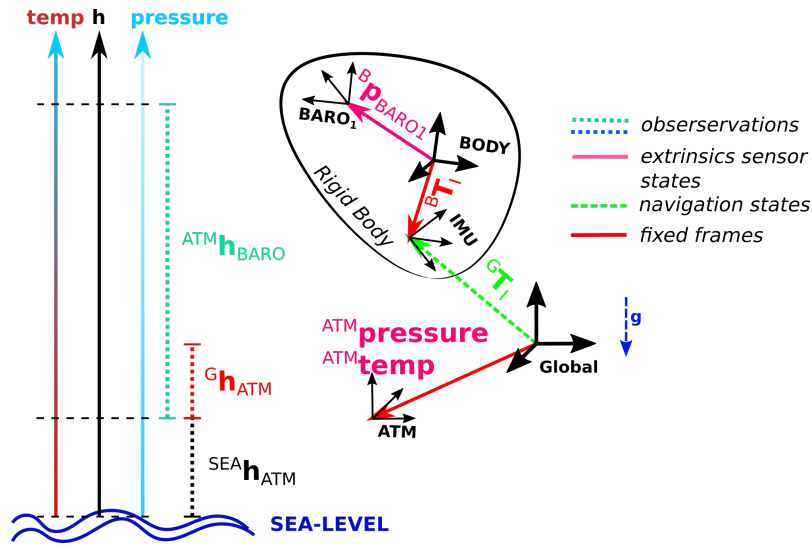

**Figure 2.10:** Shows a differential barometric sensor configuration, with a static reference sensor ATM, which observes the local pressure and temperature at a known height with respect to the navigation frame. The pressure sensor, BARO, rigidly attached to the body observes a pressure locally, which can be used to determine the relative height with respect to the reference sensor ATM based on the isotherm atmosphere model.

This allows to derive the following partial derivatives with respect to the error-states

$$\frac{\partial \tilde{\mathbf{z}}_{\mathbf{R}}}{\partial {}^{\mathcal{G}}\tilde{\boldsymbol{\theta}}_{\mathcal{I}}}\bigg|_{\hat{\mathbf{x}}} = \frac{\partial}{\partial}\left\{\left({}^{\mathcal{B}}\hat{\mathbf{R}}_{\mathcal{S}}^{\mathsf{T}\mathcal{B}}\hat{\mathbf{R}}_{\mathcal{I}}\left[{}^{\mathcal{G}}\tilde{\boldsymbol{\theta}}_{\mathcal{I}}\right]_{\times}{}^{\mathcal{B}}\hat{\mathbf{R}}_{\mathcal{I}}^{\mathsf{T}\mathcal{B}}\hat{\mathbf{R}}_{\mathcal{S}}\right)^{\vee}\right\} = {}^{\mathcal{B}}\hat{\mathbf{R}}_{\mathcal{S}}^{\mathsf{T}\mathcal{B}}\hat{\mathbf{R}}_{\mathcal{I}} \tag{2.106a}$$

$$\frac{\partial \tilde{\mathbf{z}}_{\mathbf{R}}}{\partial {}^{\mathcal{B}}\tilde{\boldsymbol{\theta}}_{\mathcal{I}}}\bigg|_{\hat{\mathbf{x}}} = \frac{\partial}{\partial}\left\{\left({}^{\mathcal{B}}\hat{\mathbf{R}}_{\mathcal{S}}^{\mathsf{T}\mathcal{B}}\hat{\mathbf{R}}_{\mathcal{I}}\left[-{}^{\mathcal{B}}\tilde{\boldsymbol{\theta}}_{\mathcal{I}}\right]_{\times}{}^{\mathcal{B}}\hat{\mathbf{R}}_{\mathcal{I}}^{\mathsf{T}\mathcal{B}}\hat{\mathbf{R}}_{\mathcal{S}}\right)^{\vee}\right\} = -{}^{\mathcal{B}}\hat{\mathbf{R}}_{\mathcal{S}}^{\mathsf{T}\mathcal{B}}\hat{\mathbf{R}}_{\mathcal{I}} \tag{2.106b}$$

$$\frac{\partial \tilde{\mathbf{z}}_{\mathbf{R}}}{\partial {}^{\mathcal{B}}\tilde{\boldsymbol{\theta}}_{\mathcal{S}}}\bigg|_{\hat{\mathbf{x}}} = \frac{\partial}{\partial}\left\{\left(\mathbf{I}\left[{}^{\mathcal{B}}\tilde{\boldsymbol{\theta}}_{\mathcal{S}}\right]_{\times}\right)^{\vee}\right\} = \mathbf{I} \tag{2.106c}$$

by using the following identities $\mathbf{R}(\boldsymbol{\theta}) \approx \mathbf{I} + [\boldsymbol{\theta}]_{\times}$ and $\left(\mathbf{R}\left[\boldsymbol{\theta}\right]_{\times}\mathbf{R}^{\mathsf{T}}\right)^{\vee} = \left([\mathbf{R}\boldsymbol{\theta}]_{\times}\right)^{\vee} = \mathbf{R}\boldsymbol{\theta}$ [136].

### 2.9.4  Barometric Altimeter

In this section, we derive a loosely coupled barometric altimeter, based on a standard atmosphere model as shown in Figure 2.10.

The barometer measures the local absolute pressure at its position. The sensor's extrinsic with respect to the body reference frame is denoted as

$$\mathbf{x}_{\mathcal{P}} = \left[{}^{\mathcal{B}}_{\mathcal{B}}\mathbf{p}_{\mathcal{P}}\right], \tag{2.107}$$

which is assumed to be rigid ${}^{\mathcal{B}}_{\mathcal{B}}\dot{\mathbf{p}}_{\mathcal{P}} = \mathbf{0}$.

The relative height can be estimated based on the barometric formula of the isotherm atmosphere model [15]. Therefore, a reference temperature ${}_{\mathcal{G}}T$ and pressure ${}_{\mathcal{G}}P$ at the global reference frame $\mathcal{G}$ needs to be specified (e.g., by a sensor on the ground or by initial samples at the ground before moving). The barometric pressure ${}_{\mathcal{P}}P$ measured by the sensor onboard is modeled as

$$_{\mathcal{P}}P = {}_{\mathcal{G}}P\left(1 - \frac{c_1 {}_{\mathcal{G}}^{\mathcal{G}}h_{\mathcal{P}}}{{}_{\mathcal{G}}T}\right)^{c_2}, \tag{2.108}$$

with $_\mathcal{G}T$ and $_\mathcal{G}P$ being the temperature in Kelvin and pressure in hectopascal. The standard atmosphere at sea-level on the equator specifies a constant (average) temperature lapse rate of $c_1 = 0.0065\,\text{K/m}$ valid up to 11 km and the constant $c_2 = 5.257$ for dry air[16]. By rearranging Equation (2.108), the absolute height of the barometer is obtained in the form

$$_\mathcal{G}^\mathcal{G}h_\mathcal{P}(_\mathcal{P}P) = \frac{_\mathcal{G}T}{c_1}\left(\left(\frac{_\mathcal{P}P}{_\mathcal{G}P}\right)^{\frac{1}{c_2}} - 1\right) \qquad (2.109)$$

The estimated height of the barometer expressed by the states is

$$_\mathcal{G}^\mathcal{G}h_\mathcal{P} = \left(_\mathcal{G}^\mathcal{G}\mathbf{T}_\mathcal{P}\right)_z = \left(_\mathcal{I}^\mathcal{G}\mathbf{T}_\mathcal{I}{}^\mathcal{B}\mathbf{T}_\mathcal{I}^{-1}{}^\mathcal{B}\mathbf{T}_\mathcal{P}\right)_z \qquad (2.110)$$

with the mapping $(\mathbf{p})_z = [0\ 0\ 1]\,\mathbf{p}$

The barometer provides noisy pressure readings

$$z_\mathcal{P} = {}_\mathcal{P}P + n_\mathcal{P}, \qquad (2.111)$$

with Gaussian pressure noise $n_\mathcal{P} \sim \mathcal{N}\left(0, \sigma_\mathcal{P}^2\right)$.

In a loosely coupled barometric altimeter fusion, the pressure measurement is converted into a pseudo height measurement $z'_\mathcal{P}$ using Equation (2.109). Also the noise of the pressure sensor needs to be converted into a distance variation at the reference pressure.

$$z'_\mathcal{P} = {}_\mathcal{G}^\mathcal{G}h'_\mathcal{P} + n'_\mathcal{P} = {}_\mathcal{G}^\mathcal{G}h_\mathcal{P}(z_\mathcal{P}) + n'_\mathcal{P}, \qquad (2.112)$$

with $n'_\mathcal{P} = {}_\mathcal{G}^\mathcal{G}h_\mathcal{P}(_\mathcal{G}P + n_\mathcal{P})$.

In the indirect filter formulation, the measurement presented to the filter is the difference between the estimated and actual measurement, following our error definition $\tilde{z} = -\hat{z} + z^\#$, and leads to

$$\tilde{z} = {}_\mathcal{G}^\mathcal{G}h'^\#_\mathcal{P} - {}_\mathcal{G}^\mathcal{G}\hat{h}_\mathcal{P} = {}_\mathcal{G}^\mathcal{G}h'^\#_\mathcal{P} - \left(_\mathcal{G}^\mathcal{G}\hat{\mathbf{T}}_\mathcal{P}\right)_z \qquad (2.113)$$

The partial derivatives with respect to the error-states are similar to Equation (2.103) with a constant factor

$$\left.\frac{\partial \tilde{\mathbf{z}}_\mathbf{P}}{\partial _\mathcal{G}^\mathcal{G}\mathbf{p}_\mathcal{I}}\right|_{\hat{\mathbf{x}}} = [0\ 0\ 1]\,{}^\mathcal{G}\hat{\mathbf{R}}_\mathcal{I} \qquad (2.114a)$$

$$\left.\frac{\partial \tilde{\mathbf{z}}_\mathbf{P}}{\partial _\mathcal{B}^\mathcal{B}\mathbf{p}_\mathcal{I}}\right|_{\hat{\mathbf{x}}} = -[0\ 0\ 1]\,{}^\mathcal{G}\hat{\mathbf{R}}_\mathcal{I} \qquad (2.114b)$$

$$\left.\frac{\partial \tilde{\mathbf{z}}_\mathbf{P}}{\partial _\mathcal{B}^\mathcal{B}\mathbf{p}_\mathcal{P}}\right|_{\hat{\mathbf{x}}} = [0\ 0\ 1]\,{}^\mathcal{G}\hat{\mathbf{R}}_\mathcal{I}{}^\mathcal{B}\hat{\mathbf{R}}_\mathcal{I}^\mathsf{T} \qquad (2.114c)$$

$$\left.\frac{\partial \tilde{\mathbf{z}}_\mathbf{P}}{\partial ^\mathcal{G}\boldsymbol{\theta}_\mathcal{I}}\right|_{\hat{\mathbf{x}}} = -[0\ 0\ 1]\,{}^\mathcal{G}\hat{\mathbf{R}}_\mathcal{I}\left[{}^\mathcal{B}\hat{\mathbf{R}}_\mathcal{I}^\mathsf{T}\left(-{}_\mathcal{B}^\mathcal{B}\hat{\mathbf{p}}_\mathcal{I} + {}_\mathcal{B}^\mathcal{B}\hat{\mathbf{p}}_\mathcal{P}\right)\right]_\times \qquad (2.114d)$$

$$\left.\frac{\partial \tilde{\mathbf{z}}_\mathbf{P}}{\partial ^\mathcal{B}\boldsymbol{\theta}_\mathcal{I}}\right|_{\hat{\mathbf{x}}} = [0\ 0\ 1]\,{}^\mathcal{G}\hat{\mathbf{R}}_\mathcal{I}\left[{}^\mathcal{B}\hat{\mathbf{R}}_\mathcal{I}^\mathsf{T}\left(-{}_\mathcal{B}^\mathcal{B}\hat{\mathbf{p}}_\mathcal{I} + {}_\mathcal{B}^\mathcal{B}\hat{\mathbf{p}}_\mathcal{P}\right)\right]_\times. \qquad (2.114e)$$

---

[16]The temperature lapse rate is highly influenced by the current weather conditions, e.g., warm weather leads to a lapse rate of $0.003\,\text{K/m}$ to $0.005\,\text{K/m}$, while cold weather $0.006\,\text{K/m}$ to $0.008\,\text{K/m}$.

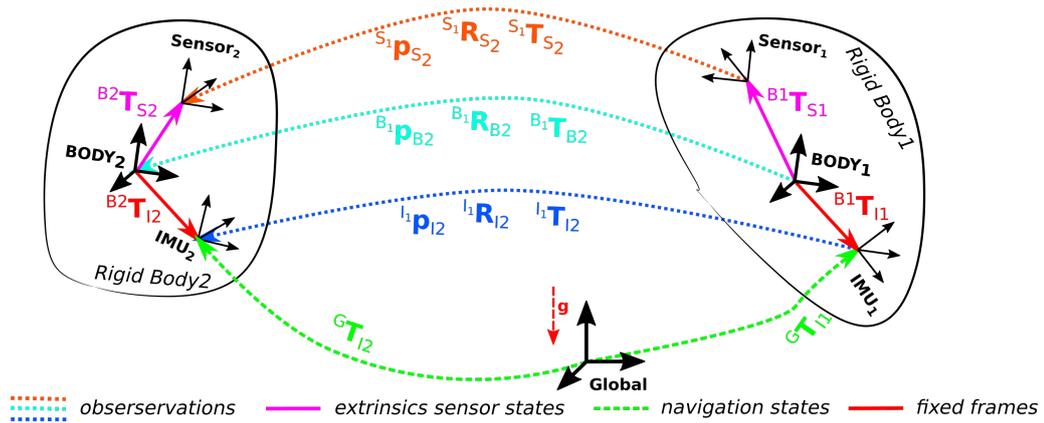

**Figure 2.11:** Spatial constellation for local relative observations between two agents.

### 2.9.5   Relative Position, Orientation, and Pose

In this section we describe *local* relative position, orientation, and pose observations in 3D between two rigid bodies IMU $\{\mathcal{I}\}$, body $\{\mathcal{B}\}$, and sensor $\{\mathcal{S}\}$ frames, to cover a wide variety of possible constellations that are used in our evaluations for CSE. Please note, that we can distinguish between local and global relative observations. Local means in this context, from the perspective of an agent, while global means from the perspective of a global meta-observer. For example, a local relative position information between the IMUs $\{\mathcal{I}_1\}$ and $\{\mathcal{I}_2\}$ is defined as $^{\mathcal{I}_1}_{\mathcal{I}_1}\mathbf{p}_{\mathcal{I}_2}$, while this observation can be expressed as a global observation by $^{\mathcal{I}_1}_{\mathcal{G}}\mathbf{p}_{\mathcal{I}_2}$. This subtle distinction is just relevant for relative position observation, as translations can be expressed in different reference frames, in this example either in $\{\mathcal{I}_1\}$ or in $\{\mathcal{G}\}$.

In Figure 2.11, relative observations between to sensors in orange is shown, which is a generic representation for different realizations. To obtain the relative pose between two agents, a visual camera at one agent and a visual tag placed on the other agent [63] or a known LED marker configuration [143] on the other agent can be used. A similar setup is e.g., used for precision landing for UAVs with visual markers placed on the ground [58]. Alternatively, in [153], the authors proposed to detect other agents using i.e. an Convolutional Neural Network (CNN) and measure the mean depth at the center using a depth camera. In [48], Gietler *et al.* proposed a wireless electromagnetic field-based sensor system that allows to obtain the relative pose between an exciter and a sensor.

A relative position between agents can be obtained by sensors that allow to measure both the bearing angles and the distance to an object with respect to the sensor, such as radars, sonars, or LIDARs.

#### Local Relative Sensor Observations

A generic local relative pose measurement, that considers the displacements between senors attached on two rigid bodies, is in the form

$$\mathbf{z}^{\#} = \begin{bmatrix} \mathbf{z}_{\mathbf{p}} \\ \mathbf{z}_{\mathbf{R}} \end{bmatrix} = \begin{bmatrix} ^{\mathcal{S}_1}_{\mathcal{S}_1}\mathbf{p}_{\mathcal{S}_2} \\ ^{\mathcal{S}_1}_{\mathcal{S}_1}\mathbf{R}_{\mathcal{S}_2} \end{bmatrix} + \mathbf{n} = \begin{bmatrix} \left(^{\mathcal{S}_1}\mathbf{T}_{\mathcal{S}_2}\right)_{\mathbf{p}} \\ \left(^{\mathcal{S}_1}\mathbf{T}_{\mathcal{S}_2}\right)_{\mathbf{R}} \end{bmatrix} + \mathbf{n} \tag{2.115a}$$

where $\mathbf{n} \sim \mathcal{N}\left(\mathbf{0}, \mathbf{R}\right)$ is a white Gaussian noise vector with a covariance matrix $\mathbf{R}$ and with

$$\begin{bmatrix} \left({}^{\mathcal{S}_1}\mathbf{T}_{\mathcal{S}_2}\right)_{\mathbf{p}} \\ \left({}^{\mathcal{S}_1}\mathbf{T}_{\mathcal{S}_2}\right)_{\mathbf{R}} \end{bmatrix} = \begin{bmatrix} {}^{\mathcal{B}_1}\mathbf{R}_{\mathcal{S}_1}^{\mathsf{T}}\left(-{}^{\mathcal{B}_1}_{\mathcal{B}_1}\mathbf{p}_{\mathcal{S}_1} + {}^{\mathcal{B}_1}_{\mathcal{B}_1}\mathbf{p}_{\mathcal{I}_1} + {}^{\mathcal{B}_1}\mathbf{R}_{\mathcal{I}_1}{}^{\mathcal{G}}\mathbf{R}_{\mathcal{I}_1}^{\mathsf{T}}\left(-{}^{\mathcal{G}}_{\mathcal{G}}\mathbf{p}_{\mathcal{I}_1} + \mathbf{M}_2\right)\right) \\ {}^{\mathcal{B}_1}\mathbf{R}_{\mathcal{S}_1}^{\mathsf{T}}{}^{\mathcal{B}_1}\mathbf{R}_{\mathcal{I}_1}{}^{\mathcal{G}}\mathbf{R}_{\mathcal{I}_1}^{\mathsf{T}}{}^{\mathcal{G}}\mathbf{R}_{\mathcal{I}_2}{}^{\mathcal{B}_2}\mathbf{R}_{\mathcal{I}_2}^{\mathsf{T}}{}^{\mathcal{B}_2}\mathbf{R}_{\mathcal{S}_2} \end{bmatrix} \quad (2.115\text{b})$$

with $\mathbf{M}_2 = {}^{\mathcal{G}}_{\mathcal{G}}\mathbf{p}_{\mathcal{I}_2} + {}^{\mathcal{G}}\mathbf{R}_{\mathcal{I}_2}\left({}^{\mathcal{B}_2}\mathbf{R}_{\mathcal{I}_2}^{\mathsf{T}}\left(-{}^{\mathcal{B}_2}_{\mathcal{B}_2}\mathbf{p}_{\mathcal{I}_2} + {}^{\mathcal{B}_2}_{\mathcal{B}_2}\mathbf{p}_{\mathcal{S}_2}\right)\right)$ and

$$ {}^{\mathcal{S}_1}\mathbf{T}_{\mathcal{S}_2} = {}^{\mathcal{B}_1}\mathbf{T}_{\mathcal{S}_1}^{-1}{}^{\mathcal{B}_1}\mathbf{T}_{\mathcal{I}_1}{}^{\mathcal{G}}\mathbf{T}_{\mathcal{I}_1}^{-1}{}^{\mathcal{G}}\mathbf{T}_{\mathcal{I}_2}{}^{\mathcal{B}_2}\mathbf{T}_{\mathcal{I}_2}^{-1}{}^{\mathcal{B}_2}\mathbf{T}_{\mathcal{S}_2}. \quad (2.115\text{c})$$

In the indirect filter formulation, the measurement presented to the filter is the difference between the estimated and actual measurement, following our error definition $\tilde{\mathbf{z}} = \hat{\mathbf{z}}^{-1}\mathbf{z}^{\#}$, and leads to

$$\tilde{\mathbf{z}} = \begin{bmatrix} \tilde{\mathbf{z}}_{\mathbf{p}} \\ \tilde{\mathbf{z}}_{\mathbf{R}} \end{bmatrix} = \begin{bmatrix} \left({}^{\mathcal{S}_1}\tilde{\mathbf{T}}_{\mathcal{S}_2}\right)_{\mathbf{p}} \\ \left({}^{\mathcal{S}_1}\tilde{\mathbf{T}}_{\mathcal{S}_2}\right)_{\mathbf{R}} \end{bmatrix} + \mathbf{n} = \begin{bmatrix} {}^{\mathcal{S}_1}\hat{\mathbf{R}}_{\mathcal{S}_2}^{\mathsf{T}}\left(-{}^{\mathcal{S}_1}_{\mathcal{S}_1}\hat{\mathbf{p}}_{\mathcal{S}_2} + {}^{\mathcal{S}_1}_{\mathcal{S}_1}\mathbf{p}_{\mathcal{S}_2}^{\#}\right) \\ {}^{\mathcal{S}_1}\hat{\mathbf{R}}_{\mathcal{S}_2}^{\mathsf{T}}{}^{\mathcal{S}_1}\mathbf{R}_{\mathcal{S}_2}^{\#} \end{bmatrix} + \mathbf{n} \quad (2.116\text{a})$$

with

$$ {}^{\mathcal{S}_1}\tilde{\mathbf{T}}_{\mathcal{S}_2} = {}^{\mathcal{S}_1}\hat{\mathbf{T}}_{\mathcal{S}_2}^{-1}{}^{\mathcal{S}_1}\mathbf{T}_{\mathcal{S}_2}^{\#} = \begin{bmatrix} {}^{\mathcal{S}_1}\hat{\mathbf{R}}_{\mathcal{S}_2}^{\mathsf{T}} & -{}^{\mathcal{S}_1}\hat{\mathbf{R}}_{\mathcal{S}_2}^{\mathsf{T}}{}^{\mathcal{S}_1}_{\mathcal{S}_1}\hat{\mathbf{p}}_{\mathcal{S}_2} \\ \mathbf{0} & 1 \end{bmatrix}\begin{bmatrix} {}^{\mathcal{S}_1}\mathbf{R}_{\mathcal{S}_2}^{\#} & {}^{\mathcal{S}_1}_{\mathcal{S}_2}\mathbf{p}_{\mathcal{S}_2}^{\#} \\ \mathbf{0} & 0 \end{bmatrix}. \quad (2.116\text{b})$$

The estimated measurement is obtained using the measurement model in Equation (2.115b).

We can linearize this measurement model around the current estimated state $\hat{\mathbf{x}}$ to obtain the measurement Jacobian $\mathbf{H} = \begin{bmatrix} \mathbf{H}_{\mathbf{p}} \\ \mathbf{H}_{\mathbf{R}} \end{bmatrix} = \frac{\partial \tilde{\mathbf{z}}}{\partial \mathbf{x}}\big|_{\hat{\mathbf{x}}}\frac{\partial \mathbf{x}}{\partial \mathbf{x}}\big|_{\hat{\mathbf{x}}}$ for our filter formulation. Please note that the measurement matrix of this collaborative/joint observations has a particular structure $\mathbf{H} = \begin{bmatrix} \mathbf{H}_{\mathbf{p}} \\ \mathbf{H}_{\mathbf{R}} \end{bmatrix} = \begin{bmatrix} \mathbf{H}_{1,\mathbf{p}} & \mathbf{H}_{2,\mathbf{p}} \\ \mathbf{H}_{1,\mathbf{R}} & \mathbf{H}_{2,\mathbf{R}} \end{bmatrix}$ as the estimated beliefs of the individual agents are stacked $\hat{\mathbf{x}} = \begin{bmatrix} \hat{\mathbf{x}}_1 \\ \hat{\mathbf{x}}_2 \end{bmatrix}$ (for details please refer to Section 3.2).

Therefore, we express the measured position error by replacing the true value in Equation (2.101b) with the definition of the total state and insert it in Equation (2.102a) to derive the partial derivatives with respect to the error-states by considering the chain rule

$$\frac{\partial \bar{\mathbf{z}}_\mathbf{p}}{\partial^{\mathcal{B}_1}\tilde{\boldsymbol{\theta}}_{\mathcal{S}_1}}\bigg|_{\hat{\mathbf{x}}} = {}^{\mathcal{S}_1}\hat{\mathbf{R}}_{\mathcal{S}_2}^\mathsf{T}\left[{}^{\mathcal{B}_1}\hat{\mathbf{R}}_{\mathcal{S}_1}^\mathsf{T}\left(-{}^{\mathcal{B}_1}_{\mathcal{B}_1}\hat{\mathbf{p}}_{\mathcal{S}_1} + {}^{\mathcal{B}_1}_{\mathcal{B}_1}\hat{\mathbf{p}}_{\mathcal{I}_1} + {}^{\mathcal{B}_1}\hat{\mathbf{R}}_{\mathcal{I}_1}{}^{\mathcal{G}}\hat{\mathbf{R}}_{\mathcal{I}_1}^\mathsf{T}\left(-{}^{\mathcal{G}}_{\mathcal{G}}\hat{\mathbf{p}}_{\mathcal{I}_1} + \hat{\mathbf{M}}_2\right)\right)\right]_\times \quad (2.117\text{a})$$

$$\frac{\partial \bar{\mathbf{z}}_\mathbf{p}}{\partial^{\mathcal{B}_1}_{\mathcal{B}_1}\tilde{\mathbf{p}}_{\mathcal{S}_1}}\bigg|_{\hat{\mathbf{x}}} = -{}^{\mathcal{S}_1}\hat{\mathbf{R}}_{\mathcal{S}_2}^\mathsf{T}{}^{\mathcal{B}_1}\hat{\mathbf{R}}_{\mathcal{S}_1}^\mathsf{T}{}^{\mathcal{B}_1}\hat{\mathbf{R}}_{\mathcal{S}_1} = -{}^{\mathcal{S}_1}\hat{\mathbf{R}}_{\mathcal{S}_2}^\mathsf{T} \quad (2.117\text{b})$$

$$\frac{\partial \bar{\mathbf{z}}_\mathbf{p}}{\partial^{\mathcal{B}_1}_{\mathcal{B}_1}\tilde{\mathbf{p}}_{\mathcal{I}_1}}\bigg|_{\hat{\mathbf{x}}} = {}^{\mathcal{S}_1}\hat{\mathbf{R}}_{\mathcal{S}_2}^\mathsf{T}{}^{\mathcal{B}_1}\hat{\mathbf{R}}_{\mathcal{S}_1}^\mathsf{T}{}^{\mathcal{B}_1}\hat{\mathbf{R}}_{\mathcal{I}_1} = {}^{\mathcal{B}_2}\hat{\mathbf{R}}_{\mathcal{S}_2}^\mathsf{T}{}^{\mathcal{B}_2}\hat{\mathbf{R}}_{\mathcal{I}_2}{}^{\mathcal{G}}\hat{\mathbf{R}}_{\mathcal{I}_2}^\mathsf{T}\hat{\mathbf{R}}_{\mathcal{I}_1} \quad (2.117\text{c})$$

$$\frac{\partial \bar{\mathbf{z}}_\mathbf{p}}{\partial^{\mathcal{B}_1}\tilde{\boldsymbol{\theta}}_{\mathcal{I}_1}}\bigg|_{\hat{\mathbf{x}}} = -{}^{\mathcal{B}_2}\hat{\mathbf{R}}_{\mathcal{S}_2}^\mathsf{T}{}^{\mathcal{B}_2}\hat{\mathbf{R}}_{\mathcal{I}_2}{}^{\mathcal{G}}\hat{\mathbf{R}}_{\mathcal{I}_2}^\mathsf{T}{}^{\mathcal{G}}\hat{\mathbf{R}}_{\mathcal{I}_1}\left[{}^{\mathcal{G}}\hat{\mathbf{R}}_{\mathcal{I}_1}^\mathsf{T}\left(-{}^{\mathcal{G}}_{\mathcal{G}}\hat{\mathbf{p}}_{\mathcal{I}_1} + \hat{\mathbf{M}}_2\right)\right]_\times \quad (2.117\text{d})$$

$$\frac{\partial \bar{\mathbf{z}}_\mathbf{p}}{\partial^{\mathcal{G}}\tilde{\boldsymbol{\theta}}_{\mathcal{I}_1}}\bigg|_{\hat{\mathbf{x}}} = {}^{\mathcal{B}_2}\hat{\mathbf{R}}_{\mathcal{S}_2}^\mathsf{T}{}^{\mathcal{B}_2}\hat{\mathbf{R}}_{\mathcal{I}_2}{}^{\mathcal{G}}\hat{\mathbf{R}}_{\mathcal{I}_2}^\mathsf{T}{}^{\mathcal{G}}\hat{\mathbf{R}}_{\mathcal{I}_1}\left[{}^{\mathcal{G}}\hat{\mathbf{R}}_{\mathcal{I}_1}^\mathsf{T}\left(-{}^{\mathcal{G}}_{\mathcal{G}}\hat{\mathbf{p}}_{\mathcal{I}_1} + \hat{\mathbf{M}}_2\right)\right]_\times \quad (2.117\text{e})$$

$$\frac{\partial \bar{\mathbf{z}}_\mathbf{p}}{\partial^{\mathcal{G}}_{\mathcal{G}}\tilde{\mathbf{p}}_{\mathcal{I}_1}}\bigg|_{\hat{\mathbf{x}}} = -{}^{\mathcal{S}_1}\hat{\mathbf{R}}_{\mathcal{S}_2}^\mathsf{T}{}^{\mathcal{B}_1}\hat{\mathbf{R}}_{\mathcal{S}_1}^\mathsf{T}{}^{\mathcal{B}_1}\hat{\mathbf{R}}_{\mathcal{I}_1} = -{}^{\mathcal{B}_2}\hat{\mathbf{R}}_{\mathcal{S}_2}^\mathsf{T}{}^{\mathcal{B}_2}\hat{\mathbf{R}}_{\mathcal{I}_2}{}^{\mathcal{G}}\hat{\mathbf{R}}_{\mathcal{I}_2}^\mathsf{T}\hat{\mathbf{R}}_{\mathcal{I}_1} \quad (2.117\text{f})$$

$$\frac{\partial \bar{\mathbf{z}}_\mathbf{p}}{\partial^{\mathcal{G}}_{\mathcal{G}}\tilde{\mathbf{p}}_{\mathcal{I}_2}}\bigg|_{\hat{\mathbf{x}}} = {}^{\mathcal{S}_1}\hat{\mathbf{R}}_{\mathcal{S}_2}^\mathsf{T}{}^{\mathcal{B}_1}\hat{\mathbf{R}}_{\mathcal{S}_1}^\mathsf{T}{}^{\mathcal{B}_1}\hat{\mathbf{R}}_{\mathcal{I}_1}{}^{\mathcal{G}}\hat{\mathbf{R}}_{\mathcal{I}_1}^\mathsf{T}{}^{\mathcal{G}}\hat{\mathbf{R}}_{\mathcal{I}_2} = {}^{\mathcal{B}_2}\hat{\mathbf{R}}_{\mathcal{S}_2}^\mathsf{T}{}^{\mathcal{B}_2}\hat{\mathbf{R}}_{\mathcal{I}_2} \quad (2.117\text{g})$$

$$\frac{\partial \bar{\mathbf{z}}_\mathbf{p}}{\partial^{\mathcal{G}}\tilde{\boldsymbol{\theta}}_{\mathcal{I}_2}}\bigg|_{\hat{\mathbf{x}}} = -{}^{\mathcal{B}_2}\hat{\mathbf{R}}_{\mathcal{S}_2}^\mathsf{T}{}^{\mathcal{B}_2}\hat{\mathbf{R}}_{\mathcal{I}_2}\left[{}^{\mathcal{B}_2}\hat{\mathbf{R}}_{\mathcal{I}_2}^\mathsf{T}\left(-{}^{\mathcal{B}_2}_{\mathcal{B}_2}\hat{\mathbf{p}}_{\mathcal{I}_2} + {}^{\mathcal{B}_2}_{\mathcal{B}_2}\hat{\mathbf{p}}_{\mathcal{S}_2}\right)\right]_\times \quad (2.117\text{h})$$

$$\frac{\partial \bar{\mathbf{z}}_\mathbf{p}}{\partial^{\mathcal{B}_2}\tilde{\boldsymbol{\theta}}_{\mathcal{I}_2}}\bigg|_{\hat{\mathbf{x}}} = {}^{\mathcal{B}_2}\hat{\mathbf{R}}_{\mathcal{S}_2}^\mathsf{T}{}^{\mathcal{B}_2}\hat{\mathbf{R}}_{\mathcal{I}_2}\left[{}^{\mathcal{B}_2}\hat{\mathbf{R}}_{\mathcal{I}_2}^\mathsf{T}\left(-{}^{\mathcal{B}_2}_{\mathcal{B}_2}\hat{\mathbf{p}}_{\mathcal{I}_2} + {}^{\mathcal{B}_2}_{\mathcal{B}_2}\hat{\mathbf{p}}_{\mathcal{S}_2}\right)\right]_\times \quad (2.117\text{i})$$

$$\frac{\partial \bar{\mathbf{z}}_\mathbf{p}}{\partial^{\mathcal{B}_2}_{\mathcal{B}_2}\tilde{\mathbf{p}}_{\mathcal{I}_2}}\bigg|_{\hat{\mathbf{x}}} = -{}^{\mathcal{S}_1}\hat{\mathbf{R}}_{\mathcal{S}_2}^\mathsf{T}{}^{\mathcal{B}_1}\hat{\mathbf{R}}_{\mathcal{S}_1}^\mathsf{T}{}^{\mathcal{B}_1}\hat{\mathbf{R}}_{\mathcal{I}_1}{}^{\mathcal{G}}\hat{\mathbf{R}}_{\mathcal{I}_1}^\mathsf{T}{}^{\mathcal{G}}\hat{\mathbf{R}}_{\mathcal{I}_2} = -{}^{\mathcal{B}_2}\hat{\mathbf{R}}_{\mathcal{S}_2}^\mathsf{T}{}^{\mathcal{B}_2}\hat{\mathbf{R}}_{\mathcal{I}_2} \quad (2.117\text{j})$$

$$\frac{\partial \bar{\mathbf{z}}_\mathbf{p}}{\partial^{\mathcal{B}_2}_{\mathcal{B}_2}\tilde{\mathbf{p}}_{\mathcal{S}_2}}\bigg|_{\hat{\mathbf{x}}} = {}^{\mathcal{S}_1}\hat{\mathbf{R}}_{\mathcal{S}_2}^\mathsf{T}{}^{\mathcal{B}_1}\hat{\mathbf{R}}_{\mathcal{S}_1}^\mathsf{T}{}^{\mathcal{B}_1}\hat{\mathbf{R}}_{\mathcal{I}_1}{}^{\mathcal{G}}\hat{\mathbf{R}}_{\mathcal{I}_1}^\mathsf{T}{}^{\mathcal{G}}\hat{\mathbf{R}}_{\mathcal{I}_2}{}^{\mathcal{B}_2}\hat{\mathbf{R}}_{\mathcal{I}_2}^\mathsf{T}{}^{\mathcal{B}_2}\hat{\mathbf{R}}_{\mathcal{S}_2} = \mathbf{I} \quad (2.117\text{k})$$

$$(2.117\text{l})$$

Please note, that the chain rule for translational components, due to the local position error definition was applied. For instance, for partial derivatives w.r.t. the IMU position error ${}^{\mathcal{G}}_{\mathcal{G}}\tilde{\mathbf{p}}_{\mathcal{I}_1}$ we obtain

$$\frac{\partial^{\mathcal{G}}_{\mathcal{G}}\mathbf{p}_{\mathcal{I}_1}}{\partial^{\mathcal{G}}_{\mathcal{G}}\tilde{\mathbf{p}}_{\mathcal{I}_1}}\bigg|_{\hat{\mathbf{x}}} = \frac{\partial}{\partial}\left\{{}^{\mathcal{G}}\hat{\mathbf{R}}_{\mathcal{I}_1}\left({}^{\mathcal{G}}_{\mathcal{G}}\hat{\mathbf{p}}_{\mathcal{I}_1} + {}^{\mathcal{G}}_{\mathcal{G}}\tilde{\mathbf{p}}_{\mathcal{I}_1}\right)\right\} = {}^{\mathcal{G}}\hat{\mathbf{R}}_{\mathcal{I}_1}. \quad (2.117\text{m})$$

Expressing the measured rotation error by replacing the true value in Equation (2.115b) with the definition of the total state, the equation in Equation (2.116a) expands to

$$\begin{aligned}\tilde{\mathbf{z}}_\mathbf{R} =&{}^{\mathcal{B}_2}\hat{\mathbf{R}}_{\mathcal{S}_2}^\mathsf{T}{}^{\mathcal{B}_2}\hat{\mathbf{R}}_{\mathcal{I}_2}{}^{\mathcal{G}}\hat{\mathbf{R}}_{\mathcal{I}_2}^\mathsf{T}{}^{\mathcal{G}}\hat{\mathbf{R}}_{\mathcal{I}_1}{}^{\mathcal{B}_1}\hat{\mathbf{R}}_{\mathcal{I}_1}^\mathsf{T}{}^{\mathcal{B}_1}\hat{\mathbf{R}}_{\mathcal{S}_1}\\ &\mathbf{R}({}^{\mathcal{B}_1}\tilde{\boldsymbol{\theta}}_{\mathcal{S}_1})^\mathsf{T}{}^{\mathcal{B}_1}\hat{\mathbf{R}}_{\mathcal{S}_1}^\mathsf{T}{}^{\mathcal{B}_1}\hat{\mathbf{R}}_{\mathcal{I}_1}\mathbf{R}({}^{\mathcal{B}_1}\tilde{\boldsymbol{\theta}}_{\mathcal{I}_1})\mathbf{R}({}^{\mathcal{G}}\tilde{\boldsymbol{\theta}}_{\mathcal{I}_1})^\mathsf{T}{}^{\mathcal{G}}\hat{\mathbf{R}}_{\mathcal{I}_1}^\mathsf{T}\\ &{}^{\mathcal{G}}\hat{\mathbf{R}}_{\mathcal{I}_2}\mathbf{R}({}^{\mathcal{G}}\tilde{\boldsymbol{\theta}}_{\mathcal{I}_2})\mathbf{R}({}^{\mathcal{B}_2}\tilde{\boldsymbol{\theta}}_{\mathcal{I}_2})^\mathsf{T}{}^{\mathcal{B}_2}\hat{\mathbf{R}}_{\mathcal{I}_2}^\mathsf{T}{}^{\mathcal{B}_2}\hat{\mathbf{R}}_{\mathcal{S}_2}\mathbf{R}({}^{\mathcal{B}_2}\tilde{\boldsymbol{\theta}}_{\mathcal{S}_2}) + \mathbf{n}_\mathbf{R}\end{aligned} \quad (2.118)$$

This allows to derive the following partial derivatives with respect to the error-states,

with $\mathbf{R}(\boldsymbol{\theta}) \approx \mathbf{I} + [\boldsymbol{\theta}]_\times$,

$$\left.\frac{\partial \tilde{\mathbf{z}}_\mathbf{R}}{\partial^{\mathcal{B}_1} \tilde{\boldsymbol{\theta}}_{\mathcal{S}_1}}\right|_{\hat{\mathbf{x}}} = -{}^{\mathcal{B}_2}\hat{\mathbf{R}}_{\mathcal{S}_2}^\mathsf{T}\, {}^{\mathcal{B}_2}\hat{\mathbf{R}}_{\mathcal{I}_2}\, {}^{\mathcal{G}}\hat{\mathbf{R}}_{\mathcal{I}_2}^\mathsf{T}\, {}^{\mathcal{G}}\hat{\mathbf{R}}_{\mathcal{I}_1}\, {}^{\mathcal{B}_1}\hat{\mathbf{R}}_{\mathcal{I}_1}^\mathsf{T}\, {}^{\mathcal{B}_1}\hat{\mathbf{R}}_{\mathcal{S}_1} \tag{2.119a}$$

$$\left.\frac{\partial \tilde{\mathbf{z}}_\mathbf{R}}{\partial^{\mathcal{B}} \tilde{\boldsymbol{\theta}}_{\mathcal{I}}}\right|_{\hat{\mathbf{x}}} = {}^{\mathcal{B}_2}\hat{\mathbf{R}}_{\mathcal{S}_2}^\mathsf{T}\, {}^{\mathcal{B}_2}\hat{\mathbf{R}}_{\mathcal{I}_2}\, {}^{\mathcal{G}}\hat{\mathbf{R}}_{\mathcal{I}_2}^\mathsf{T}\, {}^{\mathcal{G}}\hat{\mathbf{R}}_{\mathcal{I}_1} \tag{2.119b}$$

$$\left.\frac{\partial \tilde{\mathbf{z}}_\mathbf{R}}{\partial^{\mathcal{G}} \tilde{\boldsymbol{\theta}}_{\mathcal{I}_1}}\right|_{\hat{\mathbf{x}}} = -{}^{\mathcal{B}_2}\hat{\mathbf{R}}_{\mathcal{S}_2}^\mathsf{T}\, {}^{\mathcal{B}_2}\hat{\mathbf{R}}_{\mathcal{I}_2}\, {}^{\mathcal{G}}\hat{\mathbf{R}}_{\mathcal{I}_2}^\mathsf{T}\, {}^{\mathcal{G}}\hat{\mathbf{R}}_{\mathcal{I}_1} \tag{2.119c}$$

$$\left.\frac{\partial \tilde{\mathbf{z}}_\mathbf{R}}{\partial^{\mathcal{G}} \tilde{\boldsymbol{\theta}}_{\mathcal{I}_2}}\right|_{\hat{\mathbf{x}}} = {}^{\mathcal{B}_2}\hat{\mathbf{R}}_{\mathcal{S}_2}^\mathsf{T}\, {}^{\mathcal{B}_2}\hat{\mathbf{R}}_{\mathcal{I}_2} \tag{2.119d}$$

$$\left.\frac{\partial \tilde{\mathbf{z}}_\mathbf{R}}{\partial^{\mathcal{B}_2} \tilde{\boldsymbol{\theta}}_{\mathcal{I}_2}}\right|_{\hat{\mathbf{x}}} = -{}^{\mathcal{B}_2}\hat{\mathbf{R}}_{\mathcal{S}_2}^\mathsf{T}\, {}^{\mathcal{B}_2}\hat{\mathbf{R}}_{\mathcal{I}_2} \tag{2.119e}$$

$$\left.\frac{\partial \tilde{\mathbf{z}}_\mathbf{R}}{\partial^{\mathcal{B}_2} \tilde{\boldsymbol{\theta}}_{\mathcal{S}_2}}\right|_{\hat{\mathbf{x}}} = \mathbf{I}. \tag{2.119f}$$

Note, by neglecting the rotational part of the observation $\mathbf{z}_\mathbf{R} = \mathbf{0}$ or the transnational part $\mathbf{z}_\mathbf{p} = \mathbf{0}$ entirely, it can be regarded as relative position or relative orientation measurement between two sensors, respectively.

### Local Relative Body Observations

The local relative measurement model between sensor in Equation (2.115), can be simply modified to model local relative measurements between two bodies, by setting ${}^{\mathcal{B}_{\{1,2\}}}\hat{\mathbf{R}}_{\mathcal{S}_{\{1,2\}}}$ and ${}^{\mathcal{B}_{\{1,2\}}}\hat{\mathbf{p}}_{\mathcal{S}_{\{1,2\}}}$ to the neutral elements $\mathbf{p}_\mathbf{I} = \mathbf{0}$ and $\mathbf{R}_\mathbf{I} = \mathbf{I}$, respectively.

A generic local relative pose measurement between bodies, that considers the displacements to the IMUs, is in the form

$$\mathbf{z}^\# = \begin{bmatrix} \mathbf{z}_\mathbf{p} \\ \mathbf{z}_\mathbf{R} \end{bmatrix} = \begin{bmatrix} {}^{\mathcal{B}_1}_{\mathcal{B}_1}\mathbf{p}_{\mathcal{B}_2} \\ {}^{\mathcal{B}_1}_{\mathcal{B}_1}\mathbf{R}_{\mathcal{B}_2} \end{bmatrix} + \mathbf{n} = \begin{bmatrix} \left({}^{\mathcal{B}_1}\mathbf{T}_{\mathcal{B}_2}\right)_\mathbf{p} \\ \left({}^{\mathcal{B}_1}\mathbf{T}_{\mathcal{B}_2}\right)_\mathbf{R} \end{bmatrix} + \mathbf{n} \tag{2.120a}$$

where $\mathbf{n} \sim \mathcal{N}\left(\mathbf{0}, \mathbf{R}\right)$ is a white Gaussian noise vector with the covariance matrix $\mathbf{R}$ and with

$$\begin{bmatrix} \left({}^{\mathcal{B}_1}\mathbf{T}_{\mathcal{B}_2}\right)_\mathbf{p} \\ \left({}^{\mathcal{B}_1}\mathbf{T}_{\mathcal{B}_2}\right)_\mathbf{R} \end{bmatrix} = \begin{bmatrix} {}^{\mathcal{B}_1}_{\mathcal{B}_1}\mathbf{p}_{\mathcal{I}_1} + {}^{\mathcal{B}_1}\mathbf{R}_{\mathcal{I}_1}\, {}^{\mathcal{G}}\mathbf{R}_{\mathcal{I}_1}^\mathsf{T}\left(-{}^{\mathcal{G}}_{\mathcal{G}}\mathbf{p}_{\mathcal{I}_1} + {}^{\mathcal{G}}_{\mathcal{G}}\mathbf{p}_{\mathcal{I}_2} + {}^{\mathcal{G}}\mathbf{R}_{\mathcal{I}_2}\left(-{}^{\mathcal{B}_2}\mathbf{R}_{\mathcal{I}_2}^\mathsf{T}\, {}^{\mathcal{B}_2}\mathbf{p}_{\mathcal{I}_2}\right)\right) \\ {}^{\mathcal{B}_1}\mathbf{R}_{\mathcal{I}_1}\, {}^{\mathcal{G}}\mathbf{R}_{\mathcal{I}_1}^\mathsf{T}\, {}^{\mathcal{G}}\mathbf{R}_{\mathcal{I}_2}\, {}^{\mathcal{B}_2}\mathbf{R}_{\mathcal{I}_2}^\mathsf{T} \end{bmatrix} \tag{2.120b}$$

$${}^{\mathcal{B}_1}\mathbf{T}_{\mathcal{B}_2} = {}^{\mathcal{B}_1}\mathbf{T}_{\mathcal{I}_1}\, {}^{\mathcal{G}}\mathbf{T}_{\mathcal{I}_1}^{-1}\, {}^{\mathcal{G}}\mathbf{T}_{\mathcal{I}_2}\, {}^{\mathcal{B}_2}\mathbf{T}_{\mathcal{I}_2}^{-1}. \tag{2.120c}$$

Consequently, there are no partial derivatives with respect to these error-states, such that

$$\left.\frac{\partial \tilde{\mathbf{z}}_\mathbf{p}}{\partial^{\mathcal{B}_{\{1,2\}}} \tilde{\boldsymbol{\theta}}_{\mathcal{S}_{\{1,2\}}}}\right|_{\hat{\mathbf{x}}} = \mathbf{0}, \quad \left.\frac{\partial \tilde{\mathbf{z}}_\mathbf{p}}{\partial^{\mathcal{B}_{\{1,2\}}}_{\mathcal{B}_{\{1,2\}}} \tilde{\mathbf{p}}_{\mathcal{S}_{\{1,2\}}}}\right|_{\hat{\mathbf{x}}} = \mathbf{0}, \quad \left.\frac{\partial \tilde{\mathbf{z}}_\mathbf{R}}{\partial^{\mathcal{B}_{\{1,2\}}1} \tilde{\boldsymbol{\theta}}_{\mathcal{S}_{\{1,2\}}}}\right|_{\hat{\mathbf{x}}} = \mathbf{0}. \tag{2.121}$$

Note, by neglecting the rotational part of the observation $\mathbf{z}_\mathbf{R} = \mathbf{0}$ or the transnational part $\mathbf{z}_\mathbf{p} = \mathbf{0}$ entirely, it can be regarded as relative position or relative orientation measurement between two bodies, respectively.

### Local Relative IMU Observations

The local relative measurement model between bodies in Equation (2.120), can be simply modified to model local relative measurements between two IMUs, by setting $^{\mathcal{B}_{\{1,2\}}}\hat{\mathbf{R}}_{\mathcal{I}_{\{1,2\}}}$ and $^{\mathcal{B}_{\{1,2\}}}\hat{\boldsymbol{p}}_{\mathcal{I}_{\{1,2\}}}$ to the neutral elements, respectively.

A generic local relative pose measurement between IMUs is in the form

$$\mathbf{z}^{\#} = \begin{bmatrix} \mathbf{z}_{\mathbf{p}} \\ \mathbf{z}_{\mathbf{R}} \end{bmatrix} = \begin{bmatrix} ^{\mathcal{I}_1}\mathbf{p}_{\mathcal{I}_2} \\ ^{\mathcal{I}_1}\mathbf{R}_{\mathcal{I}_2} \end{bmatrix} + \mathbf{n} = \begin{bmatrix} \left(^{\mathcal{I}_1}\mathbf{T}_{\mathcal{I}_2}\right)_{\mathbf{p}} \\ \left(^{\mathcal{I}_1}\mathbf{T}_{\mathcal{I}_2}\right)_{\mathbf{R}} \end{bmatrix} + \mathbf{n} \tag{2.122a}$$

where $\mathbf{n} \sim \mathcal{N}\left(\mathbf{0}, \mathbf{R}\right)$ is a white Gaussian noise vector with the covariance matrix $\mathbf{R}$ and with

$$\begin{bmatrix} \left(^{\mathcal{I}_1}\mathbf{T}_{\mathcal{I}_2}\right)_{\mathbf{p}} \\ \left(^{\mathcal{I}_1}\mathbf{T}_{\mathcal{I}_2}\right)_{\mathbf{R}} \end{bmatrix} = \begin{bmatrix} ^{\mathcal{G}}\mathbf{R}_{\mathcal{I}_1}^{\mathsf{T}}\left(-^{\mathcal{G}}_{\mathcal{G}}\mathbf{p}_{\mathcal{I}_1} + ^{\mathcal{G}}_{\mathcal{G}}\mathbf{p}_{\mathcal{I}_2}\right) \\ ^{\mathcal{G}}\mathbf{R}_{\mathcal{I}_1}^{\mathsf{T}}\,^{\mathcal{G}}\mathbf{R}_{\mathcal{I}_2} \end{bmatrix} \tag{2.122b}$$

$$^{\mathcal{I}_1}\mathbf{T}_{\mathcal{I}_2} = ^{\mathcal{G}}\mathbf{T}_{\mathcal{I}_1}^{-1}\,^{\mathcal{G}}\mathbf{T}_{\mathcal{I}_2}. \tag{2.122c}$$

Consequently, there are no partial derivatives with respect to these error-states, such that

$$\frac{\partial \tilde{\mathbf{z}}_{\mathbf{p}}}{\partial^{\mathcal{B}_{\{1,2\}}}\tilde{\boldsymbol{\theta}}_{\mathcal{I}_{\{1,2\}}}}\bigg|_{\hat{\mathbf{x}}} = \mathbf{0}, \quad \frac{\partial \tilde{\mathbf{z}}_{\mathbf{p}}}{\partial^{\mathcal{B}_{\{1,2\}}}\tilde{\mathbf{p}}_{\mathcal{I}_{\{1,2\}}}}\bigg|_{\hat{\mathbf{x}}} = \mathbf{0}, \quad \frac{\partial \tilde{\mathbf{z}}_{\mathbf{R}}}{\partial^{\mathcal{B}_{\{1,2\}}}\tilde{\boldsymbol{\theta}}_{\mathcal{I}_{\{1,2\}}}}\bigg|_{\hat{\mathbf{x}}} = \mathbf{0}. \tag{2.123}$$

Note, by neglecting the rotational part of the observation $\mathbf{z}_{\mathbf{R}} = \mathbf{0}$ or the transnational part $\mathbf{z}_{\mathbf{p}} = \mathbf{0}$ entirely, it can be regarded as relative position or relative orientation measurement between two IMUs, respectively.

### 2.9.6 Range Measurements

In this section, we consider four different ranging constellations as depict in Figure 2.12. First, range measurement between a rigidly attached moving ranging device, a so-called tag, and a stationary ranging device, a so-called *anchor*. Second, range measurements between a stationary anchor and a moving tag. Third, range measurements between two stationary anchors, and finally, range measurements between two moving tags.

Despite being of lower dimensional (1D), in contrast to, for instance, position measurements, relative range or distance information is easier to obtain by sensors, e.g., based on time-of-arrival or received signal strength. Besides, these sensors are typically cheaper. In general, range-based localization systems use time difference of arrival (TDOA), time of arrival (TOA), return time of flight (RTOF) [115] or received signal strength (RSS) metrics/lateration techniques to estimate the distance between the antennas [127]. A promising technology for both data transmission and localization is based on UWB radio frequency (RF) signals [127]. Beneficial characteristics for estimating distances between two transceivers, are the large bandwidths, that allow UWB receivers to accurately estimate the arrival time of the first signal path [127], and it's capabilitiy to penetrate obstacles [152].

### Tag to Anchor (T-A) Range Measurement

The range sensor measures the distance between two devices, e.g., a ranging device based on UWB measures the distance between two antennas, as shown in Figure 2.12.

For moving ranging tags and stationary ranging anchors, a spatial displacement to a reference frame is estimated

$$\mathbf{x}_{\{\mathcal{T},\mathcal{A}\}} = \begin{bmatrix} \{\mathcal{B},\mathcal{G}\} \\ \{\mathcal{B},\mathcal{G}\}\mathbf{P}_{\{\mathcal{T},\mathcal{A}\}} \end{bmatrix} \tag{2.124a}$$

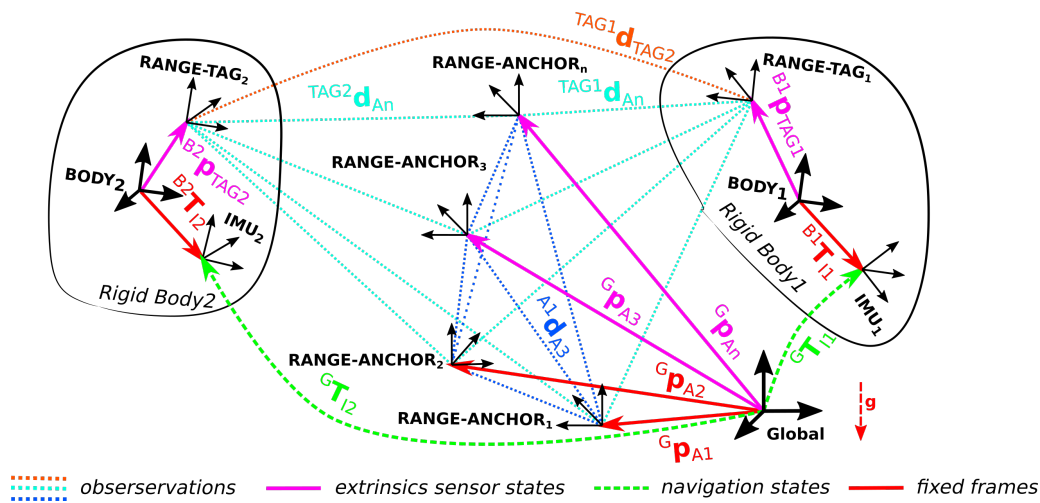

**Figure 2.12:** Spatial constellation of ranging (modified image that was published by Jung and Weiss in [75]).

which is not changing in time $_{\{\mathcal{B},\mathcal{G}\}}^{\{\mathcal{B},\mathcal{G}\}}\dot{\mathbf{P}}_{\{\mathcal{T},\mathcal{A}\}} = \mathbf{0}$.

The range measurement between a tag and an anchor (T-A) is modeled as

$$^{\mathcal{T}}z_{\mathcal{A}} = {}^{\mathcal{T}}d_{\mathcal{A}} + n_{\mathrm{r}}, \tag{2.124b}$$

with Gaussian ranging noise $n_{\mathrm{r}} \sim \mathcal{N}(0, \sigma_r^2)$ and with

$$^{\mathcal{T}}d_{\mathcal{A}} = \left\|{}_{\mathcal{T}}^{\mathcal{T}}\hat{\mathbf{P}}_{\mathcal{A}}\right\|_2 \tag{2.124c}$$

and

$$\begin{aligned}
{}_{\mathcal{T}}^{\mathcal{T}}\mathbf{P}_{\mathcal{A}} &= \left({}^{\mathcal{T}}\mathbf{T}_{\mathcal{A}}\right)_{\mathbf{p}} = \left({}^{\mathcal{B}}\mathbf{T}_{\mathcal{T}}^{-1}{}^{\mathcal{B}}\mathbf{T}_{\mathcal{I}}{}^{\mathcal{G}}\mathbf{T}_{\mathcal{I}}^{-1}{}^{\mathcal{G}}\mathbf{T}_{\mathcal{A}}\right)_{\mathbf{p}} \\
&= {}^{\mathcal{B}}\mathbf{R}_{\mathcal{T}}^{\mathsf{T}}\left(-{}_{\mathcal{B}}^{\mathcal{B}}\mathbf{p}_{\mathcal{T}} + {}_{\mathcal{B}}^{\mathcal{B}}\mathbf{p}_{\mathcal{I}} + {}^{\mathcal{B}}\mathbf{R}_{\mathcal{I}}{}^{\mathcal{G}}\mathbf{R}_{\mathcal{I}}^{-1}\left(-{}_{\mathcal{G}}^{\mathcal{G}}\mathbf{p}_{\mathcal{I}} + {}_{\mathcal{G}}^{\mathcal{G}}\mathbf{p}_{\mathcal{A}}\right)\right).
\end{aligned} \tag{2.124d}$$

As the distance measurement is invariant to the orientation ${}^{\mathcal{B}}\mathbf{R}_{\mathcal{T}} = \mathbf{I}$ of the tag $\{\mathcal{T}\}$ with respect to the body frame, it can be set to the neutral element. If the displacement of, e.g., the UWB antenna and the tag needs to be modeled, this rotation is required.

In the indirect filter formulation, the measurement presented to the filter is the difference between the estimated and actual measurement, following our error definition $\tilde{\mathbf{z}} = \hat{\mathbf{z}}^{-1}\mathbf{z}^{\#}$, and leads to

$$\tilde{\mathbf{z}} = -\left\|\left({}^{\mathcal{T}}\tilde{\mathbf{T}}_{\mathcal{A}}\right)_{\mathbf{p}}\right\| + {}^{\mathcal{T}}d_{\mathcal{A}}^{\#} + \mathbf{n} = \left\|{}_{\mathcal{T}}^{\mathcal{T}}\mathbf{p}_{\mathcal{A}}\right\| - \left\|{}_{\mathcal{T}}^{\mathcal{T}}\hat{\mathbf{p}}_{\mathcal{A}}\right\| + \mathbf{n}. \tag{2.125a}$$

We can linearize this measurement model, by replacing the true value with the total state definition, at the current estimated state $\hat{\mathbf{x}}$ to obtain the measurement Jacobian $\mathbf{H} = \frac{\partial \tilde{\mathbf{z}}}{\partial \mathbf{x}}\big|_{\hat{\mathbf{x}}} \frac{\partial \mathbf{x}}{\partial \bar{\mathbf{x}}}\big|_{\hat{\mathbf{x}}} = \frac{\partial \left\|{}_{\mathcal{T}}^{\mathcal{T}}\mathbf{p}_{\mathcal{A}}\right\|}{\partial \mathbf{x}}\Big|_{\hat{\mathbf{x}}} \frac{\partial \mathbf{x}}{\partial \bar{\mathbf{x}}}\big|_{\hat{\mathbf{x}}}$ with respect to the error-state for our filter formulation.

The Jacobian of the vector norm ($l_2$-norm), which can be split up into two functions/operations

$$f(\mathbf{x}) = \|\mathbf{x}\| = s(p(\mathbf{x})) = \sqrt{\|\mathbf{x}\|^2} \tag{2.126a}$$

with $s(\mathbf{t}) = \sqrt{\mathbf{t}}$ and $p(\mathbf{v}) = \|\mathbf{v}\|^2 = \mathbf{v}^{\mathsf{T}}\mathbf{v}$, is obtained by applying the chain rule

$$\frac{\partial \|\mathbf{x}\|}{\partial \mathbf{x}} = \frac{\partial f}{\partial \mathbf{x}} = \frac{\partial s(\cdot)}{\partial \mathbf{t}}\Big|_{\mathbf{x}} \frac{\partial p(\cdot)}{\partial \mathbf{v}}\Big|_{\mathbf{x}} = \frac{1}{2\sqrt{\mathbf{x}}}2\mathbf{x}^{\mathsf{T}} = \frac{\mathbf{x}^{\mathsf{T}}}{\sqrt{\mathbf{x}}} \tag{2.126b}$$

with

$$\frac{\partial s}{\partial \mathbf{t}} = \frac{1}{2\sqrt{\mathbf{t}}}, \ \frac{\partial p}{\partial \mathbf{v}} = 2\mathbf{v}^{\mathsf{T}}. \tag{2.126c}$$

Consequently, the Jacobian with respect to the error-state is in the form

$$\frac{\partial \tilde{\mathbf{z}}}{\partial \tilde{\mathbf{x}}} = \frac{\partial \left\| {}_{\mathcal{T}}\mathbf{p}_{\mathcal{A}} \right\|}{\partial {}_{\mathcal{T}}\mathbf{p}_{\mathcal{A}}} \bigg|_{\hat{\mathbf{x}}} \ \frac{\partial {}_{\mathcal{T}}\mathbf{p}_{\mathcal{A}}}{\partial \mathbf{x}} \bigg|_{\hat{\mathbf{x}}} \ \frac{\partial \mathbf{x}}{\partial \tilde{\mathbf{x}}} \bigg|_{\hat{\mathbf{x}}}. \tag{2.127a}$$

According to Equation (2.126), we obtain for the norm a $1 \times 3$ matrix

$$\frac{\partial \left\| {}_{\mathcal{T}}\mathbf{p}_{\mathcal{A}} \right\|}{\partial {}_{\mathcal{T}}\mathbf{p}_{\mathcal{A}}} \bigg|_{\hat{\mathbf{x}}} = \frac{{}_{\mathcal{T}}\hat{\mathbf{p}}_{\mathcal{A}}^{\mathsf{T}}}{\left\| {}_{\mathcal{T}}\hat{\mathbf{p}}_{\mathcal{A}} \right\|}, \tag{2.127b}$$

and the following partial derivatives of the translation ${}_{\mathcal{T}}\mathbf{p}_{\mathcal{A}}$ with respect to the error-states

$$\frac{\partial {}_{\mathcal{T}}\mathbf{p}_{\mathcal{A}}}{\partial {}_{\mathcal{B}}\tilde{\mathbf{p}}_{\mathcal{T}}} \bigg|_{\hat{\mathbf{x}}} = \frac{\partial {}_{\mathcal{T}}\mathbf{p}_{\mathcal{A}}}{\partial {}_{\mathcal{B}}\mathbf{p}_{\mathcal{T}}} \bigg|_{\hat{\mathbf{x}}} \frac{\partial {}_{\mathcal{B}}\mathbf{p}_{\mathcal{T}}}{\partial {}_{\mathcal{B}}\tilde{\mathbf{p}}_{\mathcal{T}}} \bigg|_{\hat{\mathbf{x}}} = -\mathbf{I}^{\mathcal{B}}\hat{\mathbf{R}}_{\mathcal{T}} = -\mathbf{I} \tag{2.127c}$$

$$\frac{\partial {}_{\mathcal{T}}\mathbf{p}_{\mathcal{A}}}{\partial {}_{\mathcal{B}}\tilde{\mathbf{p}}_{\mathcal{I}}} \bigg|_{\hat{\mathbf{x}}} = \mathbf{I}^{\mathcal{B}}\hat{\mathbf{R}}_{\mathcal{I}} \tag{2.127d}$$

$$\frac{\partial {}_{\mathcal{T}}\mathbf{p}_{\mathcal{A}}}{\partial {}^{\mathcal{B}}\tilde{\boldsymbol{\theta}}_{\mathcal{I}}} \bigg|_{\hat{\mathbf{x}}} = -{}^{\mathcal{B}}\hat{\mathbf{R}}_{\mathcal{I}} \left[ {}^{\mathcal{G}}\hat{\mathbf{R}}_{\mathcal{I}}^{\mathsf{T}}(-{}^{\mathcal{G}}_{\mathcal{G}}\hat{\mathbf{p}}_{\mathcal{I}} + {}^{\mathcal{G}}_{\mathcal{G}}\hat{\mathbf{p}}_{\mathcal{A}}) \right]_{\times} \tag{2.127e}$$

$$\frac{\partial {}_{\mathcal{T}}\mathbf{p}_{\mathcal{A}}}{\partial {}^{\mathcal{G}}\tilde{\boldsymbol{\theta}}_{\mathcal{I}}} \bigg|_{\hat{\mathbf{x}}} = {}^{\mathcal{B}}\hat{\mathbf{R}}_{\mathcal{I}} \left[ {}^{\mathcal{G}}\hat{\mathbf{R}}_{\mathcal{I}}^{\mathsf{T}}(-{}^{\mathcal{G}}_{\mathcal{G}}\hat{\mathbf{p}}_{\mathcal{I}} + {}^{\mathcal{G}}_{\mathcal{G}}\hat{\mathbf{p}}_{\mathcal{A}}) \right]_{\times} \tag{2.127f}$$

$$\frac{\partial {}_{\mathcal{T}}\mathbf{p}_{\mathcal{A}}}{\partial {}^{\mathcal{G}}_{\mathcal{G}}\tilde{\mathbf{p}}_{\mathcal{I}}} \bigg|_{\hat{\mathbf{x}}} = -{}^{\mathcal{B}}\hat{\mathbf{R}}_{\mathcal{I}} {}^{\mathcal{G}}\hat{\mathbf{R}}_{\mathcal{I}}^{\mathsf{T}} {}^{\mathcal{G}}\hat{\mathbf{R}}_{\mathcal{I}} = -{}^{\mathcal{B}}\hat{\mathbf{R}}_{\mathcal{I}} \tag{2.127g}$$

$$\frac{\partial {}_{\mathcal{T}}\mathbf{p}_{\mathcal{A}}}{\partial {}^{\mathcal{G}}_{\mathcal{G}}\tilde{\mathbf{p}}_{\mathcal{A}}} \bigg|_{\hat{\mathbf{x}}} = {}^{\mathcal{B}}\hat{\mathbf{R}}_{\mathcal{I}} {}^{\mathcal{G}}\hat{\mathbf{R}}_{\mathcal{I}}^{\mathsf{T}} {}^{\mathcal{G}}\hat{\mathbf{R}}_{\mathcal{A}} = {}^{\mathcal{B}}\hat{\mathbf{R}}_{\mathcal{I}} {}^{\mathcal{G}}\hat{\mathbf{R}}_{\mathcal{I}}^{\mathsf{T}}. \tag{2.127h}$$

Note that the following Jacobians of the estimated translations with respect to the error-states were applied

$$\frac{\partial {}^{\mathcal{B}}_{\mathcal{B}}\mathbf{p}_{\mathcal{T}}}{\partial {}^{\mathcal{B}}_{\mathcal{B}}\tilde{\mathbf{p}}_{\mathcal{T}}} \bigg|_{\hat{\mathbf{x}}} = {}^{\mathcal{B}}_{\mathcal{B}}\hat{\mathbf{p}}_{\mathcal{T}} + {}^{\mathcal{B}}\hat{\mathbf{R}}_{\mathcal{T}} {}^{\mathcal{B}}_{\mathcal{B}}\tilde{\mathbf{p}}_{\mathcal{T}} = {}^{\mathcal{B}}\hat{\mathbf{R}}_{\mathcal{T}} = \mathbf{I} \tag{2.127i}$$

$$\frac{\partial {}^{\mathcal{B}}_{\mathcal{B}}\mathbf{p}_{\mathcal{I}}}{\partial {}^{\mathcal{B}}_{\mathcal{B}}\tilde{\mathbf{p}}_{\mathcal{I}}} \bigg|_{\hat{\mathbf{x}}} = {}^{\mathcal{B}}\hat{\mathbf{R}}_{\mathcal{I}} \tag{2.127j}$$

$$\frac{\partial {}^{\mathcal{G}}_{\mathcal{G}}\mathbf{p}_{\mathcal{I}}}{\partial {}^{\mathcal{G}}_{\mathcal{G}}\tilde{\mathbf{p}}_{\mathcal{I}}} \bigg|_{\hat{\mathbf{x}}} = {}^{\mathcal{G}}\hat{\mathbf{R}}_{\mathcal{I}} \tag{2.127k}$$

$$\frac{\partial {}^{\mathcal{G}}_{\mathcal{G}}\mathbf{p}_{\mathcal{A}}}{\partial {}^{\mathcal{G}}_{\mathcal{G}}\tilde{\mathbf{p}}_{\mathcal{A}}} \bigg|_{\hat{\mathbf{x}}} = {}^{\mathcal{G}}\hat{\mathbf{R}}_{\mathcal{A}} = \mathbf{I} \tag{2.127l}$$

### Anchor to Tag (A-T) Range Measurement

Range measurements between a stationary anchor and a moving tag results in equal distances as from a moving tag and a stationary anchor, as described in Section 2.9.6, but will result in a different linearized measurement matrix.

For the moving ranging tags and stationary ranging anchors, a spatial displacement to a reference frame is estimated, see Equation (2.124a), which is not changing in time ${}^{\{\mathcal{B},\mathcal{G}\}}_{\{\mathcal{B},\mathcal{G}\}}\dot{\mathbf{p}}_{\{\mathcal{T},\mathcal{A}\}} = \mathbf{0}$. The range measurement between an anchor and an tag (A-T) is modeled as

$$\mathcal{A}_{z_{\mathcal{T}}} = {}^{\mathcal{A}}d_{\mathcal{T}} + n_{\mathrm{r}}, \tag{2.128a}$$

with Gaussian ranging noise $n_{\mathrm{r}} \sim \mathcal{N}(0, \sigma_r^2)$ and with

$$\mathcal{A}d_{\mathcal{T}} = \left\| {}^{\mathcal{A}}_{\mathcal{A}}\hat{\mathbf{p}}_{\mathcal{T}} \right\|_2 \tag{2.128b}$$

and

$$\begin{aligned}
{}^{\mathcal{A}}_{\mathcal{A}}\mathbf{p}_{\mathcal{T}} &= \left( {}^{\mathcal{A}}\mathbf{T}_{\mathcal{T}} \right)_{\mathbf{p}} = \left( {}^{\mathcal{G}}\mathbf{T}_{\mathcal{A}}^{-1}{}^{\mathcal{G}}\mathbf{T}_{\mathcal{I}}{}^{\mathcal{B}}\mathbf{T}_{\mathcal{I}}^{-1}{}^{\mathcal{B}}\mathbf{T}_{\mathcal{A}} \right)_{\mathbf{p}} \\
&= {}^{\mathcal{G}}\mathbf{R}_{\mathcal{A}}^{\mathsf{T}} \left( -{}^{\mathcal{G}}_{\mathcal{G}}\mathbf{p}_{\mathcal{A}} + {}^{\mathcal{G}}_{\mathcal{G}}\mathbf{p}_{\mathcal{I}} + {}^{\mathcal{G}}\mathbf{R}_{\mathcal{I}}{}^{\mathcal{B}}\mathbf{R}_{\mathcal{I}}^{-1} \left( -{}^{\mathcal{B}}_{\mathcal{B}}\mathbf{p}_{\mathcal{I}} + {}^{\mathcal{B}}_{\mathcal{A}}\mathbf{p}_{\mathcal{T}} \right) \right).
\end{aligned} \tag{2.128c}$$

As the distance measurement is invariant to the orientation of the sensor's reference frame, it can be set to the neutral element by ${}^{\mathcal{B}}\mathbf{R}_{\mathcal{T}} = \mathbf{I}$ ${}^{\mathcal{G}}\mathbf{R}_{\mathcal{A}} = \mathbf{I}$ . If the displacement of, e.g., the UWB antenna of the tag needs to be modeled, these rotations are required.

In the indirect filter formulation, the measurement presented to the filter is the difference between the estimated and actual measurement, following our error definition $\tilde{\mathbf{z}} = \hat{\mathbf{z}}^{-1}\mathbf{z}^{\#}$, and leads to

$$\tilde{\mathbf{z}} = -\left\| \left( {}^{\mathcal{A}}\tilde{\mathbf{T}}_{\mathcal{T}} \right)_{\mathbf{p}} \right\| + {}^{\mathcal{A}}d_{\mathcal{T}}^{\#} + \mathbf{n} = \left\| {}^{\mathcal{A}}_{\mathcal{A}}\mathbf{p}_{\mathcal{T}} \right\| - \left\| {}^{\mathcal{A}}_{\mathcal{A}}\hat{\mathbf{p}}_{\mathcal{T}} \right\| + \mathbf{n}. \tag{2.129a}$$

We can linearize this measurement model, by replacing the true value with the total state definition, at the current estimated state $\hat{\mathbf{x}}$ to obtain the measurement Jacobian $\mathbf{H} = \frac{\partial \tilde{\mathbf{z}}}{\partial \mathbf{x}}\big|_{\hat{\mathbf{x}}} \frac{\partial \mathbf{x}}{\partial \tilde{\mathbf{x}}}\big|_{\hat{\mathbf{x}}} = \frac{\partial \left\| {}^{\mathcal{T}}_{\mathcal{T}}\mathbf{p}_{\mathcal{A}} \right\|}{\partial \mathbf{x}}\bigg|_{\hat{\mathbf{x}}} \frac{\partial \mathbf{x}}{\partial \tilde{\mathbf{x}}}\big|_{\hat{\mathbf{x}}}$ with respect to the error-state for our filter formulation.

The measurement matrix Jacobian with respect to the error-state is in the form

$$\frac{\partial \tilde{\mathbf{z}}}{\partial \tilde{\mathbf{x}}} = \frac{\partial \left\| {}^{\mathcal{A}}_{\mathcal{A}}\mathbf{p}_{\mathcal{T}} \right\|}{\partial {}^{\mathcal{A}}_{\mathcal{A}}\mathbf{p}_{\mathcal{T}}}\bigg|_{\hat{\mathbf{x}}} \frac{\partial {}^{\mathcal{A}}_{\mathcal{A}}\mathbf{p}_{\mathcal{T}}}{\partial \mathbf{x}}\bigg|_{\hat{\mathbf{x}}} \frac{\partial \mathbf{x}}{\partial \tilde{\mathbf{x}}}\bigg|_{\hat{\mathbf{x}}}. \tag{2.130a}$$

According to Equation (2.126), we obtain for the norm a $1 \times 3$ matrix

$$\frac{\partial \left\| {}^{\mathcal{A}}_{\mathcal{A}}\mathbf{p}_{\mathcal{T}} \right\|}{\partial {}^{\mathcal{A}}_{\mathcal{A}}\mathbf{p}_{\mathcal{T}}}\bigg|_{\hat{\mathbf{x}}} = \frac{{}^{\mathcal{A}}_{\mathcal{A}}\hat{\mathbf{p}}_{\mathcal{T}}^{\mathsf{T}}}{\left\| {}^{\mathcal{A}}_{\mathcal{A}}\hat{\mathbf{p}}_{\mathcal{T}} \right\|}, \tag{2.130b}$$

and the following partial derivatives of the translation ${}^{\mathcal{A}}_{\mathcal{A}}\mathbf{p}_{\mathcal{T}}$ with respect to the error-states

$$\frac{\partial {}^{\mathcal{A}}_{\mathcal{A}}\mathbf{p}_{\mathcal{T}}}{\partial {}^{\mathcal{B}}_{\mathcal{B}}\tilde{\mathbf{p}}_{\mathcal{T}}}\bigg|_{\hat{\mathbf{x}}} = \frac{\partial {}^{\mathcal{A}}_{\mathcal{A}}\mathbf{p}_{\mathcal{T}}}{\partial {}^{\mathcal{B}}_{\mathcal{B}}\mathbf{p}_{\mathcal{T}}}\bigg|_{\hat{\mathbf{x}}} \frac{\partial {}^{\mathcal{B}}_{\mathcal{B}}\mathbf{p}_{\mathcal{T}}}{\partial {}^{\mathcal{B}}_{\mathcal{B}}\tilde{\mathbf{p}}_{\mathcal{T}}}\bigg|_{\hat{\mathbf{x}}} = {}^{\mathcal{G}}\hat{\mathbf{R}}_{\mathcal{A}}^{\mathsf{T}}{}^{\mathcal{G}}\hat{\mathbf{R}}_{\mathcal{I}}{}^{\mathcal{B}}\hat{\mathbf{R}}_{\mathcal{I}}^{\mathsf{T}}\mathbf{I} \tag{2.130c}$$

$$\frac{\partial {}^{\mathcal{A}}_{\mathcal{A}}\mathbf{p}_{\mathcal{T}}}{\partial {}^{\mathcal{B}}_{\mathcal{B}}\tilde{\mathbf{p}}_{\mathcal{I}}}\bigg|_{\hat{\mathbf{x}}} = -{}^{\mathcal{G}}\hat{\mathbf{R}}_{\mathcal{A}}^{\mathsf{T}}{}^{\mathcal{G}}\hat{\mathbf{R}}_{\mathcal{I}}{}^{\mathcal{B}}\hat{\mathbf{R}}_{\mathcal{I}}^{\mathsf{T}}{}^{\mathcal{B}}\hat{\mathbf{R}}_{\mathcal{I}} = -{}^{\mathcal{G}}\hat{\mathbf{R}}_{\mathcal{A}}^{\mathsf{T}}{}^{\mathcal{G}}\hat{\mathbf{R}}_{\mathcal{I}} \tag{2.130d}$$

$$\frac{\partial {}^{\mathcal{A}}_{\mathcal{A}}\mathbf{p}_{\mathcal{T}}}{\partial {}^{\mathcal{B}}\tilde{\boldsymbol{\theta}}_{\mathcal{I}}}\bigg|_{\hat{\mathbf{x}}} = +{}^{\mathcal{G}}\hat{\mathbf{R}}_{\mathcal{A}}^{\mathsf{T}}{}^{\mathcal{G}}\hat{\mathbf{R}}_{\mathcal{I}} \left[ {}^{\mathcal{B}}\hat{\mathbf{R}}_{\mathcal{I}}^{\mathsf{T}}(-{}^{\mathcal{B}}_{\mathcal{B}}\hat{\mathbf{p}}_{\mathcal{I}} + {}^{\mathcal{B}}_{\mathcal{B}}\hat{\mathbf{p}}_{\mathcal{T}}) \right]_{\times} \tag{2.130e}$$

$$\frac{\partial {}^{\mathcal{A}}_{\mathcal{A}}\mathbf{p}_{\mathcal{T}}}{\partial {}^{\mathcal{G}}\tilde{\boldsymbol{\theta}}_{\mathcal{I}}}\bigg|_{\hat{\mathbf{x}}} = -{}^{\mathcal{G}}\hat{\mathbf{R}}_{\mathcal{A}}^{\mathsf{T}}{}^{\mathcal{G}}\hat{\mathbf{R}}_{\mathcal{I}} \left[ {}^{\mathcal{B}}\hat{\mathbf{R}}_{\mathcal{I}}^{\mathsf{T}}(-{}^{\mathcal{B}}_{\mathcal{B}}\hat{\mathbf{p}}_{\mathcal{I}} + {}^{\mathcal{B}}_{\mathcal{B}}\hat{\mathbf{p}}_{\mathcal{T}}) \right]_{\times} \tag{2.130f}$$

$$\frac{\partial {}^{\mathcal{A}}_{\mathcal{A}}\mathbf{p}_{\mathcal{T}}}{\partial {}^{\mathcal{G}}_{\mathcal{G}}\tilde{\mathbf{p}}_{\mathcal{I}}}\bigg|_{\hat{\mathbf{x}}} = {}^{\mathcal{G}}\hat{\mathbf{R}}_{\mathcal{A}}^{\mathsf{T}}{}^{\mathcal{G}}\hat{\mathbf{R}}_{\mathcal{I}} \tag{2.130g}$$

$$\frac{\partial {}^{\mathcal{A}}_{\mathcal{A}}\mathbf{p}_{\mathcal{T}}}{\partial {}^{\mathcal{G}}_{\mathcal{G}}\tilde{\mathbf{p}}_{\mathcal{A}}}\bigg|_{\hat{\mathbf{x}}} = -{}^{\mathcal{G}}\hat{\mathbf{R}}_{\mathcal{A}}^{\mathsf{T}}{}^{\mathcal{G}}\hat{\mathbf{R}}_{\mathcal{A}} = -\mathbf{I}, \tag{2.130h}$$

with $^{\mathcal{G}}\hat{\mathbf{R}}_{\mathcal{A}} = \mathbf{I}$.

**Anchor to Anchor (A-A) Range Measurement**

In this section, we cover the range measurements (A-A) between two stationary anchors. For the stationary ranging anchors, a spatial displacement to the global reference frame $\{\mathcal{G}\}$ is estimated, see Equation (2.124a), which is not changing in time $^{\mathcal{G}}_{\mathcal{G}}\dot{\mathbf{p}}_{\mathcal{A}} = \mathbf{0}$. The range measurement between an anchor and an tag (A-A) is modeled as

$$^{\mathcal{A}_1}z_{\mathcal{A}_2} = {}^{\mathcal{A}_1}d_{\mathcal{A}_2} + n_{\mathrm{r}}, \tag{2.131a}$$

with Gaussian ranging noise $n_{\mathrm{r}} \sim \mathcal{N}(0, \sigma_r^2)$ and with

$$^{\mathcal{A}_1}d_{\mathcal{A}_2} = \left\| {}^{\mathcal{A}_1}_{\mathcal{A}_1}\mathbf{p}_{\mathcal{A}_2} \right\|_2 \tag{2.131b}$$

and

$$\begin{aligned}
{}^{\mathcal{A}_1}_{\mathcal{A}_1}\mathbf{p}_{\mathcal{A}_2} &= \left( {}^{\mathcal{A}_1}\mathbf{T}_{\mathcal{A}_2} \right)_{\mathbf{p}} = \left( {}^{\mathcal{G}}\mathbf{T}^{-1}_{\mathcal{A}_1}\,{}^{\mathcal{G}}\mathbf{T}_{\mathcal{A}_2} \right)_{\mathbf{p}} \\
&= {}^{\mathcal{G}}\mathbf{R}^{\mathsf{T}}_{\mathcal{A}_1} \left( -{}^{\mathcal{G}}_{\mathcal{G}}\mathbf{p}_{\mathcal{A}_1} + {}^{\mathcal{G}}_{\mathcal{G}}\mathbf{p}_{\mathcal{A}_2} \right).
\end{aligned} \tag{2.131c}$$

As the distance measurement is invariant to the orientation of the sensor's reference frame, it can be set to the neutral element by $^{\mathcal{G}}\mathbf{R}_{\mathcal{A}} = \mathbf{I}$. If the displacement of, e.g., the UWB antenna of the tag needs to be modeled, these rotations is required. In the indirect filter formulation, the measurement presented to the filter is the difference between the estimated and actual measurement, following our error definition $\tilde{\mathbf{z}} = \hat{\mathbf{z}}^{-1}\mathbf{z}^{\#}$, and leads to

$$\tilde{\mathbf{z}} = -\left\| \left( {}^{\mathcal{A}_1}\tilde{\mathbf{T}}_{\mathcal{A}_2} \right)_{\mathbf{p}} \right\| + {}^{\mathcal{A}_1}d^{\#}_{\mathcal{A}_2} + \mathbf{n} = \left\| {}^{\mathcal{A}_1}_{\mathcal{A}_1}\mathbf{p}_{\mathcal{A}_2} \right\| - \left\| {}^{\mathcal{A}_1}_{\mathcal{A}_1}\hat{\mathbf{p}}_{\mathcal{A}_2} \right\| + \mathbf{n}. \tag{2.132a}$$

We can linearize this measurement model, by replacing the true value with the total state definition, at the current estimated state $\hat{\mathbf{x}}$ to obtain the measurement Jacobian $\mathbf{H} = \frac{\partial \tilde{\mathbf{z}}}{\partial \mathbf{x}}\big|_{\hat{\mathbf{x}}}$ with respect to the error-state for our filter formulation. The measurement Jacobian with respect to the error-state is in the form

$$\frac{\partial \tilde{\mathbf{z}}}{\partial \tilde{\mathbf{x}}} = \frac{\partial \left\| {}^{\mathcal{A}_1}_{\mathcal{A}_1}\mathbf{p}_{\mathcal{A}_2} \right\|}{\partial {}^{\mathcal{A}_1}_{\mathcal{A}_1}\mathbf{p}_{\mathcal{A}_2}}\bigg|_{\hat{\mathbf{x}}} \frac{\partial {}^{\mathcal{A}_1}_{\mathcal{A}_1}\mathbf{p}_{\mathcal{A}_2}}{\partial \mathbf{x}}\bigg|_{\hat{\mathbf{x}}} \frac{\partial \mathbf{x}}{\partial \tilde{\mathbf{x}}}\bigg|_{\hat{\mathbf{x}}}. \tag{2.132b}$$

According to Equation (2.126), we obtain for the norm a $1 \times 3$ matrix

$$\frac{\partial \left\| {}^{\mathcal{A}_1}_{\mathcal{A}_1}\mathbf{p}_{\mathcal{A}_2} \right\|}{\partial {}^{\mathcal{A}_1}_{\mathcal{A}_1}\mathbf{p}_{\mathcal{A}_2}}\bigg|_{\hat{\mathbf{x}}} = \frac{{}^{\mathcal{A}_1}_{\mathcal{A}_1}\hat{\mathbf{p}}^{\mathsf{T}}_{\mathcal{A}_2}}{\left\| {}^{\mathcal{A}_1}_{\mathcal{A}_1}\hat{\mathbf{p}}_{\mathcal{A}_2} \right\|}, \tag{2.132c}$$

and the following partial derivatives of the translation $^{\mathcal{A}_1}_{\mathcal{A}_1}\mathbf{p}_{\mathcal{A}_2}$ with respect to the error-states

$$\frac{\partial {}^{\mathcal{A}_1}_{\mathcal{A}_1}\mathbf{p}_{\mathcal{A}_2}}{\partial {}^{\mathcal{G}}_{\mathcal{G}}\tilde{\mathbf{p}}_{\mathcal{A}_1}}\bigg|_{\hat{\mathbf{x}}} = -{}^{\mathcal{G}}\hat{\mathbf{R}}^{\mathsf{T}}_{\mathcal{A}_1}\,{}^{\mathcal{G}}\hat{\mathbf{R}}_{\mathcal{A}_1} = -\mathbf{I}, \tag{2.132d}$$

$$\frac{\partial {}^{\mathcal{A}_1}_{\mathcal{A}_1}\mathbf{p}_{\mathcal{A}_2}}{\partial {}^{\mathcal{G}}_{\mathcal{G}}\tilde{\mathbf{p}}_{\mathcal{A}_2}}\bigg|_{\hat{\mathbf{x}}} = {}^{\mathcal{G}}\hat{\mathbf{R}}^{\mathsf{T}}_{\mathcal{A}_1}\,{}^{\mathcal{G}}\hat{\mathbf{R}}_{\mathcal{A}_2} = \mathbf{I}, \tag{2.132e}$$

with $^{\mathcal{G}}\hat{\mathbf{R}}_{\mathcal{A}_{\{1,2\}}} = \mathbf{I}$.

### Tag to Tag (T-T) Range Measurement

Finally, we model the range measurement between two moving tags, e.g., a ranging device based on UWB measures the distance between two antennas, as shown in Figure 2.12.

For the moving ranging tags, a spatial displacement to a reference frame $\{\mathcal{T}\}$ is estimated

$$\mathbf{x}_{\mathcal{T}} = \begin{bmatrix} \mathcal{B}_{\{1,2\}} \mathbf{p}_{\mathcal{T}_{\{1,2\}}} \end{bmatrix} \tag{2.133a}$$

which is not changing in time $\mathcal{B}_{\{1,2\}}_{\mathcal{B}_{\{1,2\}}} \dot{\mathbf{p}}_{\mathcal{T}_{\{1,2\}}} = \mathbf{0}$.

The range measurement between two tags (T-T) is modeled as

$$\mathcal{T}_1 z_{\mathcal{T}_2} = \mathcal{T}_1 d_{\mathcal{T}_2} + n_{\mathrm{r}}, \tag{2.133b}$$

with Gaussian ranging noise $n_{\mathrm{r}} \sim \mathcal{N}(0, \sigma_r^2)$ and with

$$\mathcal{T}_1 d_{\mathcal{T}_2} = \left\| \mathcal{T}_1_{\mathcal{T}_1} \hat{\mathbf{p}}_{\mathcal{T}_2} \right\|_2 \tag{2.133c}$$

and

$$\begin{aligned}
\mathcal{T}_1_{\mathcal{T}_1} \mathbf{p}_{\mathcal{T}_2} &= \left( \mathcal{T}_1 \mathbf{T}_{\mathcal{T}_2} \right)_{\mathbf{p}} = \left( \mathcal{B}_1 \mathbf{T}_{\mathcal{T}_1}^{-1} \mathcal{B}_1 \mathbf{T}_{\mathcal{I}_1} \, \mathcal{G} \mathbf{T}_{\mathcal{I}_1}^{-1} \mathcal{G} \mathbf{T}_{\mathcal{I}_2} \, \mathcal{B}_2 \mathbf{T}_{\mathcal{I}_2}^{\mathsf{T}} \, \mathcal{B}_2 \mathbf{T}_{\mathcal{T}_2} \right)_{\mathbf{p}} \\
&= \mathcal{B}_1 \mathbf{R}_{\mathcal{T}_1}^{\mathsf{T}} \left( -\mathcal{B}_1_{\mathcal{B}_1} \mathbf{p}_{\mathcal{T}_1} + \mathcal{B}_1_{\mathcal{B}_1} \mathbf{p}_{\mathcal{I}_1} + \mathcal{B}_1 \mathbf{R}_{\mathcal{I}_1} \, \mathcal{G} \mathbf{R}_{\mathcal{I}_1}^{\mathsf{T}} \left( -\mathcal{G}_{\mathcal{G}} \mathbf{p}_{\mathcal{I}_1} \right. \right. \\
&\qquad \left. \left. + \mathcal{G}_{\mathcal{G}} \mathbf{p}_{\mathcal{I}_2} + \mathcal{G} \mathbf{R}_{\mathcal{I}_2} \left( \mathcal{B}_2 \mathbf{R}_{\mathcal{I}_2}^{\mathsf{T}} \left( -\mathcal{B}_2_{\mathcal{B}_2} \mathbf{p}_{\mathcal{I}_2} + \mathcal{B}_2_{\mathcal{B}_2} \mathbf{p}_{\mathcal{T}_2} \right) \right) \right) \right).
\end{aligned} \tag{2.133d}$$

As the distance measurement is invariant to the orientation of the sensor's reference frame, it can be set to the neutral element by $\mathcal{B} \mathbf{R}_{\mathcal{T}_1} = \mathbf{I}$ and $\mathcal{B} \mathbf{R}_{\mathcal{T}_2} = \mathbf{I}$. If the displacement of, e.g., the UWB antenna of the tag needs to be modeled, these rotations are required. In the indirect filter formulation, the measurement presented to the filter is the difference between the estimated and actual measurement, following our error definition $\tilde{\mathbf{z}} = \hat{\mathbf{z}}^{-1} \mathbf{z}^{\#}$, and leads to

$$\tilde{\mathbf{z}} = - \left\| \left( \mathcal{T}_1 \tilde{\mathbf{T}}_{\mathcal{T}_2} \right)_{\mathbf{p}} \right\| + \mathcal{T}_1 d_{\mathcal{T}_2}^{\#} + \mathbf{n} = \left\| \mathcal{T}_1_{\mathcal{T}_1} \mathbf{p}_{\mathcal{T}_2} \right\| - \left\| \mathcal{T}_1_{\mathcal{T}_1} \hat{\mathbf{p}}_{\mathcal{T}_2} \right\| + \mathbf{n}. \tag{2.134a}$$

We can linearize this measurement model, by replacing the true value with the total state definition, at the current estimated state $\hat{\mathbf{x}}$ to obtain the measurement Jacobian $\mathbf{H} = \frac{\partial \tilde{\mathbf{z}}}{\partial \mathbf{x}} \big|_{\hat{\mathbf{x}}}$ with respect to the error-state for our filter formulation. The measurement matrix with respect to the error-state is in the form

$$\frac{\partial \tilde{\mathbf{z}}}{\partial \tilde{\mathbf{x}}} = \frac{\partial \left\| \mathcal{T}_1_{\mathcal{T}_1} \mathbf{p}_{\mathcal{T}_2} \right\|}{\partial \mathcal{T}_1_{\mathcal{T}_1} \mathbf{p}_{\mathcal{T}_2}} \bigg|_{\hat{\mathbf{x}}} \frac{\partial \mathcal{T}_1_{\mathcal{T}_1} \mathbf{p}_{\mathcal{T}_2}}{\partial \mathbf{x}} \bigg|_{\hat{\mathbf{x}}} \frac{\partial \mathbf{x}}{\partial \tilde{\mathbf{x}}} \bigg|_{\hat{\mathbf{x}}}. \tag{2.134b}$$

According to Equation (2.126), we obtain for the norm a $1 \times 3$ matrix

$$\frac{\partial \left\| \mathcal{T}_1_{\mathcal{T}_1} \mathbf{p}_{\mathcal{T}_2} \right\|}{\partial \mathcal{T}_1_{\mathcal{T}_1} \mathbf{p}_{\mathcal{T}_2}} \bigg|_{\hat{\mathbf{x}}} = \frac{\mathcal{T}_1_{\mathcal{T}_1} \hat{\mathbf{p}}_{\mathcal{T}_2}^{\mathsf{T}}}{\left\| \mathcal{T}_1_{\mathcal{T}_1} \hat{\mathbf{p}}_{\mathcal{T}_2} \right\|}, \tag{2.134c}$$

and the following partial derivatives of the translation $\mathcal{T}_1_{\mathcal{T}_1} \mathbf{p}_{\mathcal{T}_2}$ with respect to the error-states, which is similar to the Jacobians for the local relative position measurement between two sensors in Equation (2.117),

$$\frac{\partial^{\mathcal{T}_1}_{\mathcal{T}_1}\mathbf{p}_{\mathcal{T}_2}}{\partial^{\mathcal{B}_1}\tilde{\boldsymbol{\theta}}_{\mathcal{T}_1}}\bigg|_{\hat{\mathbf{x}}} = \left[{}^{\mathcal{B}_1}\hat{\mathbf{R}}^{\mathsf{T}}_{\mathcal{T}_1}\left(-{}^{\mathcal{B}_1}_{\mathcal{B}_1}\hat{\mathbf{p}}_{\mathcal{T}_1} + {}^{\mathcal{B}_1}_{\mathcal{B}_1}\hat{\mathbf{p}}_{\mathcal{I}_1} + {}^{\mathcal{B}_1}\hat{\mathbf{R}}_{\mathcal{I}_1}{}^{\mathcal{G}}\hat{\mathbf{R}}^{\mathsf{T}}_{\mathcal{I}_1}\left(-{}^{\mathcal{G}}_{\mathcal{G}}\hat{\mathbf{p}}_{\mathcal{I}_1} + \hat{\mathbf{M}}_2\right)\right)\right]_{\times} \tag{2.134d}$$

$$\hat{\mathbf{M}}_2 = \tag{2.134e}$$

$$\frac{\partial^{\mathcal{T}_1}_{\mathcal{T}_1}\mathbf{p}_{\mathcal{T}_2}}{\partial^{\mathcal{B}_1}_{\mathcal{B}_1}\tilde{\mathbf{p}}_{\mathcal{T}_1}}\bigg|_{\hat{\mathbf{x}}} = -{}^{\mathcal{B}_1}\hat{\mathbf{R}}^{\mathsf{T}}_{\mathcal{T}_1}{}^{\mathcal{B}_1}\hat{\mathbf{R}}_{\mathcal{T}_1} = -\mathbf{I} \tag{2.134f}$$

$$\frac{\partial^{\mathcal{T}_1}_{\mathcal{T}_1}\mathbf{p}_{\mathcal{T}_2}}{\partial^{\mathcal{B}_1}_{\mathcal{B}_1}\tilde{\mathbf{p}}_{\mathcal{I}_1}}\bigg|_{\hat{\mathbf{x}}} = {}^{\mathcal{B}_1}\hat{\mathbf{R}}^{\mathsf{T}}_{\mathcal{T}_1}{}^{\mathcal{B}_1}\hat{\mathbf{R}}_{\mathcal{I}_1} \tag{2.134g}$$

$$\frac{\partial^{\mathcal{T}_1}_{\mathcal{T}_1}\mathbf{p}_{\mathcal{T}_2}}{\partial^{\mathcal{B}_1}\tilde{\boldsymbol{\theta}}_{\mathcal{I}_1}}\bigg|_{\hat{\mathbf{x}}} = -{}^{\mathcal{B}_1}\hat{\mathbf{R}}^{\mathsf{T}}_{\mathcal{T}_1}{}^{\mathcal{B}_1}\hat{\mathbf{R}}_{\mathcal{I}_1}\left[{}^{\mathcal{G}}\hat{\mathbf{R}}^{\mathsf{T}}_{\mathcal{I}_1}\left(-{}^{\mathcal{G}}_{\mathcal{G}}\hat{\mathbf{p}}_{\mathcal{I}_1} + \hat{\mathbf{M}}_2\right)\right]_{\times} \tag{2.134h}$$

$$\frac{\partial^{\mathcal{T}_1}_{\mathcal{T}_1}\mathbf{p}_{\mathcal{T}_2}}{\partial^{\mathcal{G}}\tilde{\boldsymbol{\theta}}_{\mathcal{I}_1}}\bigg|_{\hat{\mathbf{x}}} = {}^{\mathcal{B}_1}\hat{\mathbf{R}}^{\mathsf{T}}_{\mathcal{T}_1}{}^{\mathcal{B}_1}\hat{\mathbf{R}}_{\mathcal{I}_1}\left[{}^{\mathcal{G}}\hat{\mathbf{R}}^{\mathsf{T}}_{\mathcal{I}_1}\left(-{}^{\mathcal{G}}_{\mathcal{G}}\hat{\mathbf{p}}_{\mathcal{I}_1} + \hat{\mathbf{M}}_2\right)\right]_{\times} \tag{2.134i}$$

$$\frac{\partial^{\mathcal{T}_1}_{\mathcal{T}_1}\mathbf{p}_{\mathcal{T}_2}}{\partial^{\mathcal{G}}_{\mathcal{G}}\tilde{\mathbf{p}}_{\mathcal{I}_1}}\bigg|_{\hat{\mathbf{x}}} = -{}^{\mathcal{B}_1}\hat{\mathbf{R}}^{\mathsf{T}}_{\mathcal{T}_1}{}^{\mathcal{B}_1}\hat{\mathbf{R}}_{\mathcal{I}_1} \tag{2.134j}$$

$$\frac{\partial^{\mathcal{T}_1}_{\mathcal{T}_1}\mathbf{p}_{\mathcal{T}_2}}{\partial^{\mathcal{G}}_{\mathcal{G}}\tilde{\mathbf{p}}_{\mathcal{I}_2}}\bigg|_{\hat{\mathbf{x}}} = {}^{\mathcal{B}_1}\hat{\mathbf{R}}^{\mathsf{T}}_{\mathcal{T}_1}{}^{\mathcal{B}_1}\hat{\mathbf{R}}_{\mathcal{I}_1}{}^{\mathcal{G}}\hat{\mathbf{R}}^{\mathsf{T}}_{\mathcal{I}_1}{}^{\mathcal{G}}\hat{\mathbf{R}}_{\mathcal{I}_2} \tag{2.134k}$$

$$\frac{\partial^{\mathcal{T}_1}_{\mathcal{T}_1}\mathbf{p}_{\mathcal{T}_2}}{\partial^{\mathcal{G}}\tilde{\boldsymbol{\theta}}_{\mathcal{I}_2}}\bigg|_{\hat{\mathbf{x}}} = -{}^{\mathcal{B}_1}\hat{\mathbf{R}}^{\mathsf{T}}_{\mathcal{T}_1}{}^{\mathcal{B}_1}\hat{\mathbf{R}}_{\mathcal{I}_1}{}^{\mathcal{G}}\hat{\mathbf{R}}^{\mathsf{T}}_{\mathcal{I}_1}{}^{\mathcal{G}}\hat{\mathbf{R}}_{\mathcal{I}_2}\left[{}^{\mathcal{B}_2}\hat{\mathbf{R}}^{\mathsf{T}}_{\mathcal{I}_2}\left(-{}^{\mathcal{B}_2}_{\mathcal{B}_2}\hat{\mathbf{p}}_{\mathcal{I}_2} + {}^{\mathcal{B}_2}_{\mathcal{B}_2}\hat{\mathbf{p}}_{\mathcal{T}_2}\right)\right]_{\times} \tag{2.134l}$$

$$\frac{\partial^{\mathcal{T}_1}_{\mathcal{T}_1}\mathbf{p}_{\mathcal{T}_2}}{\partial^{\mathcal{B}_2}\tilde{\boldsymbol{\theta}}_{\mathcal{I}_2}}\bigg|_{\hat{\mathbf{x}}} = {}^{\mathcal{B}_1}\hat{\mathbf{R}}^{\mathsf{T}}_{\mathcal{T}_1}{}^{\mathcal{B}_1}\hat{\mathbf{R}}_{\mathcal{I}_1}{}^{\mathcal{G}}\hat{\mathbf{R}}^{\mathsf{T}}_{\mathcal{I}_1}{}^{\mathcal{G}}\hat{\mathbf{R}}_{\mathcal{I}_2}\left[{}^{\mathcal{B}_2}\hat{\mathbf{R}}^{\mathsf{T}}_{\mathcal{I}_2}\left(-{}^{\mathcal{B}_2}_{\mathcal{B}_2}\hat{\mathbf{p}}_{\mathcal{I}_2} + {}^{\mathcal{B}_2}_{\mathcal{B}_2}\hat{\mathbf{p}}_{\mathcal{T}_2}\right)\right]_{\times} \tag{2.134m}$$

$$\frac{\partial^{\mathcal{T}_1}_{\mathcal{T}_1}\mathbf{p}_{\mathcal{T}_2}}{\partial^{\mathcal{B}_2}_{\mathcal{B}_2}\tilde{\mathbf{p}}_{\mathcal{I}_2}}\bigg|_{\hat{\mathbf{x}}} = -{}^{\mathcal{B}_1}\hat{\mathbf{R}}^{\mathsf{T}}_{\mathcal{T}_1}{}^{\mathcal{B}_1}\hat{\mathbf{R}}_{\mathcal{I}_1}{}^{\mathcal{G}}\hat{\mathbf{R}}^{\mathsf{T}}_{\mathcal{I}_1}{}^{\mathcal{G}}\hat{\mathbf{R}}_{\mathcal{I}_2} \tag{2.134n}$$

$$\frac{\partial^{\mathcal{T}_1}_{\mathcal{T}_1}\mathbf{p}_{\mathcal{T}_2}}{\partial^{\mathcal{B}_2}_{\mathcal{B}_2}\tilde{\mathbf{p}}_{\mathcal{T}_2}}\bigg|_{\hat{\mathbf{x}}} = {}^{\mathcal{B}_1}\hat{\mathbf{R}}^{\mathsf{T}}_{\mathcal{T}_1}{}^{\mathcal{B}_1}\hat{\mathbf{R}}_{\mathcal{I}_1}{}^{\mathcal{G}}\hat{\mathbf{R}}^{\mathsf{T}}_{\mathcal{I}_1}{}^{\mathcal{G}}\hat{\mathbf{R}}_{\mathcal{I}_2}{}^{\mathcal{B}_2}\hat{\mathbf{R}}^{\mathsf{T}}_{\mathcal{I}_2}{}^{\mathcal{B}_2}\hat{\mathbf{R}}_{\mathcal{T}_2} \tag{2.134o}$$

$$\tag{2.134p}$$

with ${}^{\mathcal{B}_{\{1,2\}}}\hat{\mathbf{R}}_{\mathcal{T}_{\{1,2\}}} = \mathbf{I}$.

### 2.9.7 Observability of states

A desirable feature of an estimator framework is an automatic self-calibration of system's extrinsic (position of sensor reference frame with respect to the body) and intrinsic, e.g., the focal length of a camera) states. Although the calibration could be performed *a-priori* during the manufacturing process, it involves a lot of engineering effort and might be a time-consuming process, that might result in unsatisfying accuracy and external factors, such as aging, temperature, humidity, external mechanical forces, and might gravely bias the factory calibration [105].

If the extrinsic (spatial relation) or intrinsic (internal parameters) of a sensor are assumed to be perfectly known or the initial guess are not exact, a degradation in both, the consistency and the accuracy of the filter estimates may be caused. Therefore, online self-calibration or re-calibration during mission/operation of the system is desired.

Despite spatial calibration, online temporal calibration, e.g., the offset between sensors timestamps, is an indispensable or integral part of high-precision pose estimation algorithms [36, 37, 45, 93] Despite being able to compensate calibration errors or biases in the initial beliefs, additional self-calibration states lead to an increased state space and computational complexity, while the number of observations/measurements remain the same [57]. Therefore, it is recommended to aim at a minimal set of calibration parameters and a minimal state representation.

A nonlinear system is classified as observable, if it is possible to compute a single initial state $\mathbf{x}^0$ given a sequence of control inputs $\mathbf{u}(t)$ and measurements $\mathbf{z}(t) = h(\mathbf{x}(t))$. It is globally observable if there exist no initial points $\mathbf{x}_0^0$ and $\mathbf{x}_1^0$ in state space with the same input-output-maps for any control inputs. A system is defined as weekly observable the there is no point $\mathbf{x}_1^0$ in the local neighborhood of $\mathbf{x}_0^0$ [57, 60].

The state variable $\mathbf{x}$ may or may not be directly observable by the measurement and might be used to represent the memory of the system [60]. If the state variable is not directly observable, the variable might be inferred from a local set of measurements, meaning that it requires measurements obtained by the entire system, which again might strongly depend on the operation performed. Further, normal operation might lead to unobservable directions in the parameter space [105]. Assuming a system constituting of an 6-DoF IMU (accelerometer and gyroscope) and GNSS sensor that provides only absolute position and given a known gravity vector, horizontal motion is required to recover the absolute orientation about the gravity vector, e.g., yaw [57].

Consequently, not all components can be estimated as they lack of observability, e.g, the pose between the body $\{\mathcal{B}\}$ and IMU reference frame $\{\mathcal{I}\}$ is generally not observable. Estimating this transformation would lead to an over parameterization of the state space that introduces unobservable states, which in turn may harm the convergence of the entire state. Generally, this pose needs to be know *a-priori* and is constrained as constant/rigid. This transformation is used to simplify the modeling of a modular estimation problem. The sensor extrinsic between the body $\{\mathcal{B}\}$ and the sensor $\{\mathcal{S}\}$ reference frame might suffer from poor observability, which might depend on the IMU characteristics, motion performed, the system's sensor configuration, and the characteristics of the sensor observations. In [57], Hausman *et al.* investigated on how to move in order to generate motions that render the entire state space of the system observable. In this context, the aim of classical trajectory/motion planning algorithms is, e.g, to minimize the total energy, which might lead to unobservable subspaces of the system state.

Another important aspect is that discretization and linearization of continuous nonlinear system, induce additional error and lead to wrong observability results [57], see for instance the well-known studies on EKF-SLAM [66] and EKF-based VIO [92]. In [66], Huang *et al.* identified as fundamental root cause for inconsistencies a mismatch between the dimension of observable subspaces between the nonlinear system and the linearized system used in the EKF and introduced the first estimate Jacobian (FEJ)-EKF and the observability-constrained (OC)-EKF, which were derived based on the nonlinear system's observability properties [67]. Meaning, that if a certain quantity is unobservable in the actual nonlinear system, its should be unobservable in the linearized model as well [92].

In [92], Li and Mourikis performed the observability analysis on an EKF-based VIO framework and proved that the same problem as in [66] appear. Theoretically, a visual-inertial navigation system, constituting of an IMU and camera, that is navigation in space with a known gravity vector but unknown landmark positions, has four unobservable degrees of freedom. Three relating to the global (absolute) position and one corresponding the rotation about the gravitation vector, i.e. global yaw, if z-axis is aligned with the gravity vector. Erroneously, the rotation about the gravity vector is rendered observable in the linearized system [92] and decreases the accuracy and degrades the consistency of the filter formulation.

Similarly, Huang *et al.* studied in [65] the consistency of CL in a multi-robot system with relative pose measurements between robots from the perspective of observability and proved that the linearized global multi-robot system has an observable subspace higher than the underlying nonlinear global system. Again, the estimated covariances obtain corrections where theoretically no information is available and leads to inconsistencies in the estimates and it was shown that the linearization points for the Jacobians affects the observability properties of the linearized system. Please note, that in EKF-based aided INS where absolute information, for instance absolute position measurement, and sufficient motion, i.e. excitation in at least 2 directions, is provided, the nonlinear and linearized system is able to recover the absolute position, velocity, and orientation of the IMU. Summarized, the observability properties of the system model play a major role in assessing the estimator's accuracy and consistency. It is essential to ensure, that the linearized system has an observable subspace of the same dimension as the underlying nonlinear system [65].

In general, it is highly recommended as initial step to perform *a-priori* an observability analysis of the underlying nonlinear system to proper understand the behavior of the method. In a second step, if the observability properties of the nonlinear system satisfies the requirements, an observability analysis on the linearized system might be inevitable, as discussed previously. Unfortunately, the analysis cannot be performed on integrated and/or approximated state transition matrices. Due to their numerically nature, they are not amenable to theoretical analysis, and analytical expression need to be found, otherwise the observability of the linearized system cannot be guaranteed [93].

### Nonlinear observability analysis

The observability of a linear or nonlinear system can be determined by performing the so-called *rank test* on the observability matrix $\mathbf{O}$, which should be equal to the dimension of the state space. For nonlinear system, the observability matrix $\mathbf{O}$ is constructed using the Lie derivatives of the sensor model(s) $h(\mathbf{x})$, which are defined recursively up to a certain order without considering measurement noise $\mathbf{v} = \mathbf{0}$ [57]. The 0-th Lie derivative is the measurement model

$$\mathbf{L}_0^h = h(\mathbf{x}(t)) \tag{2.135}$$

while the next $i$-th Lie derivatives are constructed recursively by

$$\mathbf{L}_{i+1}^h = \frac{\partial \mathbf{L}_i^h}{\partial t} = \frac{\partial \mathbf{L}_i^h}{\partial \mathbf{x}} \frac{\partial \mathbf{x}}{\partial \mathbf{t}} = \frac{\partial \mathbf{L}_i^h}{\partial \mathbf{x}} f(\mathbf{x}, \mathbf{u}), \tag{2.136}$$

and allows constructing an observability matrix $\mathbf{O}(\mathbf{x}, \mathbf{u})$ up to a certain order by computing the Jacobian of the previous Lie derivatives with respect to the state $\nabla \mathbf{L}_0^h = \frac{\partial \mathbf{L}_0^h}{\partial \mathbf{x}}$

$$\mathbf{O}_i(\mathbf{x}, \mathbf{u}) = \left[ \nabla \mathbf{L}_0^h; \nabla \mathbf{L}_1^h; \dots; \nabla \mathbf{L}_i^h \right]. \tag{2.137}$$

The computation of the i-th order observability matrix allows making a binary assessment, if the nonlinear system is weakly observable [60]

$$\text{observable} = \text{rank}(\mathbf{O}_i(\mathbf{x}, \mathbf{u})) \geq \dim(\mathbf{x}) \tag{2.138}$$

Note that this analysis neither quantifies how *well* the system is observable, nor does it consider noise/perturbations of the system [57].

Despite these limitations, it is a systematic approach that can be processed symbolically in tools like as MATLAB.

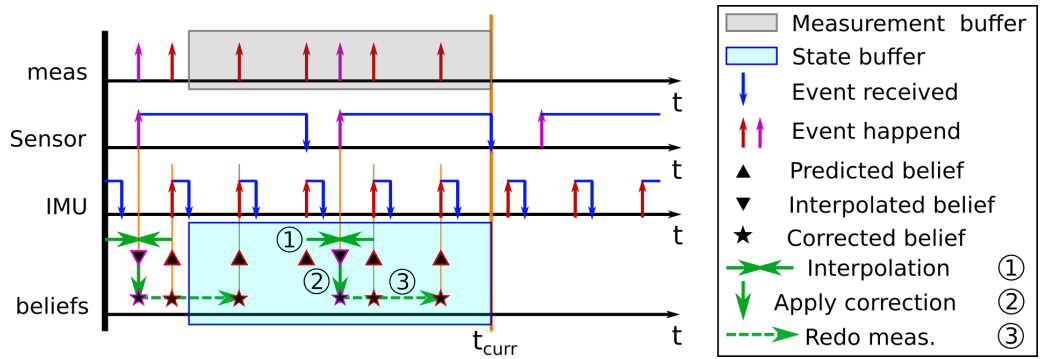

**Figure 2.13:** Sensor activity: Given two sensors, an IMU for state propagation and an exteroceptive sensor S for state correction. The sensors provide delayed measurement at different rates, and it is assumed that the sources of the timestamps are synchronized. At $t_{curr}$, the filter receives a sensor measurement, which relates to an event that falls between two IMU measurements. In order to fuse that measurement, first, an interpolated (pseudo) belief needs to be created. Second, the pseudo belief is corrected using the sensor measurement. Third, all measurements in the measurement buffer after that event, are reprocessed in order to propagate the correction forward in time.

### 2.9.8 Delayed Measurements

Each sensor measurement has a certain delay from the actual moment/event the information was perceived by the sensor, until it is actually processed in the filter. Different sensors can have individual reference times, meaning that all measurements need to be related to a common reference time.

In a Kalman filter, a measurement needs to be associated to an *a-priori* belief of the filter. In our ESEKF formulation, the state evolution happens periodically at the rate of the IMU measurements, meaning that each processed IMU measurement will result in a new (predicted) belief, as shown in Figure 2.13. We assume that the IMU measurements have the least latency, such that there is always a more recent predicted state. If this would not be the case, these update sensor measurements need to be buffered until a more recent IMU measurement arrives. Since it is more likely that a timestamp of update measurements falls between two IMU measurements, then exactly coincides with a single IMU measurement, a synthetic state needs to be created by interpolating between two IMU measurements and by performing a prediction step using that pseudo measurement. After the correction was applied on the synthetic/interpolated state of the past, we need to *redo updates*, in order to apply all changes that happened in the future of that measurement event. In case of a nonlinear system, all Jacobians, e.g., in the propagation and update steps, need to be re-linearized about a new linearization pint – the new best estimate.

To support delayed measurements, certain measurements and beliefs (mean and covariance) need to be buffered, e.g., in a ring buffer [149] with a fixed size or in a fixed time horizon buffer [74], see Section 2.2.

Please note, that propagating corrected beliefs forward in time scales linearly with the number of measurements to be re-applied, while the covariance propagation is computationally more expensive than the mean propagation [5, 149]. To mitigate the costly covariance propagation based on IMU measurements, Allak *et al.* proposed in [5] an efficient method for covariance propagation during recalculation of delayed measurements, by applying pre-computed scattering factors using the star product. This renders the computational complexity of IMU propagation invariant to the number of propagation steps between filter updates, at the cost of approximations made as the scattering factors are not re-linearized.

# Chapter 3

# Filter-based Distributed Collaborative State Estimation

From Section 1.1, we observe that distributed state estimation is a key component of various research fields (target tracking, localization, etc.) with various solutions for different problems. In this chapter, we study the problem of Collaborative State Estimation (CSE) on a networked dynamic model with decoupled inputs and sparsely coupled outputs. We investigate some algorithms and architectures that are used in the field of Collaborative Localization (CL) and compare them in terms of their capabilities and properties. We propose a novel filter-based algorithm for Distribued CSE (DCSE) which was published in [76] and is particularly designed to address the problem of collaborative aided inertial navigation. Furthermore, we evaluate the scalability and the filter credibility in Monte Carlo simulations and on a real world dataset for MAVs. Parts of this chapter have already been published in [73, 76].

## 3.1  Introduction

**Definition 1** *Collaborative State Estimation*[1] *facilitates nodes of a heterogeneous network, which are modeled as stochastic processes with decoupled dynamics, communication, and computation capabilities, to work/sense jointly/collaboratively by coupling their outputs.*

The aim of the collaboration is to improve the accuracy of individual's states, and thus, the global state, by exchanging information via communication, in order to achieve a common goal or task. Wireless communication is needed when the nodes are spatially distributed or moving in space in order to exchange, e.g., beliefs (mean and covariances), correlations, unique identifiers, measurements, time synchronization messages, etc..

If nodes are non-stationary, meaning that the spatial constellation of them is changing, the network can be seen as swarm organism with multiple non-rigid limbs/nodes that is spreading and observing the environment at different places, while information is shared wirelessly across the network [123]. By sharing information, the networks has – to a certain extent – swarm intelligence, enabling the network to be more than the sum of its individual nodes.

A self-aware swarm organism, acts based on the observed environment, leading to information based swarm control, see e.g., [158], which is not covered in this thesis. We investigate only open loop control performance, and thus, focusing only on the estimation problem. Without loss of generality, all approaches that try to improve the observability of the state space or to maximize information obtained from the environment, e.g., by steering the swarm into information rich areas or by forming swarm constellations that reduces e.g., the positional dilution of precision (PDOP), can be applied.

---

[1]Wiktionary: Latin *collaboro*, from *con-* ("with") + *laboro* ("to work"): to work together with others.

Collaboratively estimating states or performing observations jointly in such a wireless sensor/robot network can significantly improve an individual's estimate. This abrogates the statistical independence of the involved nodes as it leads to (statistical) correlations (coupling) between their estimates. Again, these dependencies need to be considered and properly treated in order to obtain consistent estimates. Naively processed in a centralized fusion entity, tremendous challenges regarding the communication and computation complexity are imposed, as the state space of the dynamic system is increasing linearly with the number of agents.

Depending on the problem at hand, different collaborative estimation architectures can be applied, while the major challenge remains: *to preserve consistent estimates by properly treating correlations between nodes.* Further, a balance between the estimation accuracy and the effort needed for communication, memory, computation, and book-keeping/maintenance of correlations has to be found, as it directly impacts the scalability of the system with respect to the number of nodes.

Joint observation among nodes differ from individual observations by relating an observation to multiple nodes' states compared to just an individual node's estimate. Joint observations are generally exteroceptive ones, e.g., a relative position measurement between two moving robots. This measurement would require the estimated poses of both robots from the moment the measurement was performed in the update step of the estimator.

Joint observations are not limited to direct relative observation between nodes. A joint observation could also be an observation of a commonly known object of interest, e.g., multiple robots are tracking jointly the position of a moving target[2], see e.g., Mirzaei *et al.* [108] or multiple robots are tracking jointly landmarks to obtain a globally consistent representation of the environment [160].

As already mention in Chapter 1, the aim of this thesis is to perform filter-based CSE fully distributed and with minimal overhead in size-varying groups/swarms of mobile agents in order to improve the localization performance. Thus, DCSE can be regarded as generalization of the Distributed Collaborative Localization (DCL) problem, that might be extended to support mapping and target tracking. The proposed approaches for DCSE could even be applied beyond robotics applications requiring estimation processes.

Due to the generalization made from the collaborative localization problem in mobile robotics application to wireless sensor network, the terms *agent* or *robot* are interchangeable with the term *node*, and the terms *swarm* or *group* are interchangeable with *network*.

Before presenting different filter-based DCSE architectures in Section 3.4, we briefly define the CSE problem in Section 3.2. After that, our proposed DAH Collaborative State Estimation (DAH-CSE) approach is presented in Section 3.3 and is evaluated on different use cases in Section 3.5. Conclusion are made in Section 3.6.

Please note that the presented DCSE approaches within this chapter assume, that the state space of individual nodes is not changing arbitrarily during mission and measurements are received and processed without delay, and thus, they are received and processed in correct order. Eliminating these limitations are the primary objectives and the motivation of the final chapter (see Chapter 7).

---

[2]If the target would have communication and processing capabilities, it would be able to treat the target as a node of the collaborative estimation network and it would reduce the observation to two direct relative observation between robots and the target. Note that this would simplify the distributed collaborative estimation approaches as no consensus about multiple hypothesis of the tracked object's belief needs to be found, whereas in a centralized estimation architecture, it would make no difference as there is globally only one belief relating to the target.

## 3.2 The filter-based Collaborative State Estimation problem

In this section, we study the estimation problem in a setup of three agents performing CSE (like in [121]), to provide details and challenges in filter-based CSE, as the cross-covariance need to be handed properly for optimal estimation (given a linear problem, known dynamic models, and signal noise characteristics) in the prediction step (Section 3.2.2), the private (Section 3.2.3), and joint update steps (Section 3.2.4).

To visualize the influence of cross-covariance terms due to the output coupling of the dynamic systems, a private observation on $A_1$ (Section 3.2.3), a relative joint observation between $A_1$ and $A_2$ (Section 3.2.4) and the state propagation (Section 3.2.2) will be analyzed based on a (global) full state estimator.

### 3.2.1 Network dynamics model

Consider a network of three interconnected agents, $A_i$, with $i = 1, \ldots, 3$, each of them has a computational unit, while each system is dynamically decoupled and described by an individual discrete-time linear stochastic process (see Equation (2.62)). In contrast to the dynamics, the outputs of an arbitrary set of models are/can be coupled via joint observation.

Assume that the physical process of an agent $A_i$ is represented as linear stochastic system in the form

$$\mathbf{x}_i^k = \mathbf{\Phi}_{ii}^{k|k-1} \mathbf{x}_i^{k-1} + \mathbf{G}_{ii}^{k-1} \mathbf{w}_i^{k-1}, \tag{3.1a}$$

$$\mathbf{z}_i^k = \mathbf{H}_{ii}^k \mathbf{x}_i^k + \mathbf{v}_{ii}^k, \tag{3.1b}$$

$$\mathbf{z}_{i,j}^k = \mathbf{H}_{i,j}^k \begin{bmatrix} \mathbf{x}_i^k \\ \mathbf{x}_j^k \end{bmatrix} + \mathbf{v}_{i,j}^k, \tag{3.1c}$$

where $\mathbf{x}_i$ is the state vector, and $\mathbf{z}_i$ and $\mathbf{z}_{i,j}$ are output vectors. $\mathbf{\Phi}_{ii}$ and $\mathbf{G}_{ii}$ are the state transition and process noise coupling matrices, respectively. $\mathbf{H}_{ii}$ and $\mathbf{H}_{i,j}$ are the measurement matrices for the private and joint observations, respectively. Please note that the output coupling is not limited to pairwise observation. $\mathbf{w}_i \sim \mathcal{N}(\mathbf{0}, \mathbf{W}_{ii})$, and $\mathbf{v}_{\{ii,i,j\}} \sim \mathcal{N}(\mathbf{0}, \mathbf{R}_{\{ii,i,j\}})$ are the known process and observation noise, that are assumed to be zero-mean white Gaussian with an associated covariance $\mathbf{W}_{ii}$ and $\mathbf{R}_{\{ii,i,j\}}$, respectively.

An agent's belief (state) $\mathbf{x}_i$ needs to be regarded as part of the *global full state* $\mathbf{x}$

$$\mathbf{x} = \begin{bmatrix} \mathbf{x}_1 \\ \mathbf{x}_2 \\ \mathbf{x}_3 \end{bmatrix} \sim \mathcal{N} \left( \begin{bmatrix} \hat{\mathbf{x}}_1 \\ \hat{\mathbf{x}}_2 \\ \hat{\mathbf{x}}_3 \end{bmatrix}, \begin{bmatrix} \mathbf{\Sigma}_{11} & \mathbf{\Sigma}_{12} & \mathbf{\Sigma}_{13} \\ \bullet & \mathbf{\Sigma}_{22} & \mathbf{\Sigma}_{23} \\ \bullet & \bullet & \mathbf{\Sigma}_{33} \end{bmatrix} \right). \tag{3.2}$$

Therefore, the global dynamics of the swarm of $N$ agents can be represented in the form

$$\mathbf{x}^k = \mathbf{\Phi}^{k|k-1} \mathbf{x}^{k-1} + \mathbf{G}^{k-1} \mathbf{w}^{k-1}, \tag{3.3a}$$

$$\mathbf{z}^k = \mathbf{H}^k \mathbf{x}^k + \mathbf{v}^k, \tag{3.3b}$$

where $\mathbf{z}^k = \begin{bmatrix} \mathbf{z}_1^k; \ldots; \mathbf{z}_N^k \end{bmatrix}$ is the concatenation of measurement available at each agent at $t^k$, $\mathbf{x}^k = \begin{bmatrix} \mathbf{x}_1^k; \ldots; \mathbf{x}_N^k \end{bmatrix}$ is the stacked state, $\mathbf{w}^k = \begin{bmatrix} \mathbf{w}_1^k; \ldots; \mathbf{w}_N^k \end{bmatrix}$ and $\mathbf{v}^k = \begin{bmatrix} \mathbf{v}_1^k; \ldots; \mathbf{v}_N^k \end{bmatrix}$ are the concatenated zero-mean white Gaussian process and observation noise, respectively. Similar to the assumptions made in the Schmidt-Kalman filter (see 2.8.3), the dynamics of the individual agents are typically not linked/coupled with each other, meaning that the off-diagonal blocks of the state transition matrix and process noise matrix are zero. Such that the matrices $\mathbf{\Phi} = \text{Diag}\left([\mathbf{\Phi}_{11}; \ldots; \mathbf{\Phi}_{NN}]\right)$ and $\mathbf{W} = \text{Diag}\left([\mathbf{W}_{11}; \ldots; \mathbf{W}_{NN}]\right)$ are obtained from the individual agents' dynamics [145]. Similarly, the global measurement matrix $\mathbf{H} = [\mathbf{H}_{11}, \ldots, \mathbf{H}_{NN}]$ is built from the individual agents' measurement sensitivity matrices.

### 3.2.2 Prediction/Propagation step

We aim at a Kalman filter formulation for the network of three agents, which is described by the global dynamics in Equation (3.3). The prediction step of the global full state $\mathbf{x}$ is defined as follows.

Assuming the discrete-time linear stochastic process is evolving with a sampling period $\Delta t = t^k - t^{k-1}$, the estimated state and the state covariance evolves according to

$$\mathbf{x}^k = \mathbf{\Phi}^{k|k-1}\mathbf{x}^{k-1} + \mathbf{G}^{k-1}\mathbf{w}^{k-1}, \tag{3.4a}$$

$$\hat{\mathbf{x}}^k = \mathbf{\Phi}^{k|k-1}\mathbf{x}^{k-1}, \tag{3.4b}$$

$$\mathbf{\Sigma}^k = \mathbf{\Phi}^{k|k-1}\mathbf{\Sigma}^{k-1}\mathbf{\Phi}^{k|k-1\mathsf{T}} + \mathbf{Q}^{k-1}, \tag{3.4c}$$

with the process noise $\mathbf{Q}^{k-1} = \mathbf{G}^{k-1}\mathbf{W}^{k-1}\mathbf{G}^{k-1\mathsf{T}}$.

Since we assume no kinematic and noise interference between agents[3], their dynamics and process error evolve independently. Therefore, the joint transition and process matrices are typically found to be sparse

$$\mathbf{\Phi}^{k|k-1} = \begin{bmatrix} \mathbf{\Phi}_{11}^{k|k-1} & \mathbf{0} & \mathbf{0} \\ \mathbf{0} & \mathbf{\Phi}_{22}^{k|k-1} & \mathbf{0} \\ \mathbf{0} & \mathbf{0} & \mathbf{\Phi}_{33}^{k|k-1} \end{bmatrix}, \tag{3.5a}$$

$$\mathbf{Q}^{k-1} = \begin{bmatrix} \mathbf{Q}_{11}^{k-1} & \mathbf{0} & \mathbf{0} \\ \mathbf{0} & \mathbf{Q}_{22}^{k-1} & \mathbf{0} \\ \mathbf{0} & \mathbf{0} & \mathbf{Q}_{33}^{k-1} \end{bmatrix}. \tag{3.5b}$$

Assuming no cross-covariances between the agents, the propagated joint covariance is

$$\mathbf{\Sigma}^{k|k-1} = \begin{bmatrix} \mathbf{\Phi}_{11}^{k|k-1}\mathbf{\Sigma}_{11}^{k-1}\mathbf{\Phi}_{11}^{k|k-1\mathsf{T}} + \mathbf{Q}_{11}^{k-1} & \mathbf{0} & \mathbf{0} \\ \mathbf{0} & \ddots & \mathbf{0} \\ \mathbf{0} & \mathbf{0} & \mathbf{\Phi}_{33}^{k|k-1}\mathbf{\Sigma}_{33}^{k-1}\mathbf{\Phi}_{33}^{k|k-1\mathsf{T}} + \mathbf{Q}_{33}^{k-1} \end{bmatrix}. \tag{3.6a}$$

---

[3]A reasonable assumption in robotics applications, since agents are typically not physically connected/strapped together. In case of multi-copter, we assume that the downwash turbulence created by individual's propulsion system is not influencing other agents.

Once the agents are correlated, the cross-covariance are propagated as well

$$
\boldsymbol{\Sigma}^k = \begin{bmatrix} \boldsymbol{\Phi}_{11} & \mathbf{0} & \mathbf{0} \\ \mathbf{0} & \boldsymbol{\Phi}_{22} & \mathbf{0} \\ \mathbf{0} & \mathbf{0} & \boldsymbol{\Phi}_{33} \end{bmatrix}^{k|k-1} \begin{bmatrix} \boldsymbol{\Sigma}_{11} & \boldsymbol{\Sigma}_{12} & \boldsymbol{\Sigma}_{13} \\ \boldsymbol{\Sigma}_{21} & \boldsymbol{\Sigma}_{22} & \boldsymbol{\Sigma}_{23} \\ \boldsymbol{\Sigma}_{31} & \boldsymbol{\Sigma}_{32} & \boldsymbol{\Sigma}_{33} \end{bmatrix}^{k-1} \begin{bmatrix} \boldsymbol{\Phi}_{11}{}^{\mathsf{T}} & \mathbf{0} & \mathbf{0} \\ \mathbf{0} & \boldsymbol{\Phi}_{22}{}^{\mathsf{T}} & \mathbf{0}+ \\ \mathbf{0} & \mathbf{0} & \boldsymbol{\Phi}_{33}{}^{\mathsf{T}} \end{bmatrix}^{k|k-1}
$$

$$
+ \begin{bmatrix} \mathbf{Q}_{11} & \mathbf{0} & \mathbf{0} \\ \mathbf{0} & \mathbf{Q}_{22} & \mathbf{0} \\ \mathbf{0} & \mathbf{0} & \mathbf{Q}_{33} \end{bmatrix}^{k-1} =
$$

$$
\begin{bmatrix} \boldsymbol{\Phi}_{11}^{k|k-1}\boldsymbol{\Sigma}_{11}^k\boldsymbol{\Phi}_{11}^{k|k-1\mathsf{T}} + \mathbf{Q}_{11}^{k-1} & \boldsymbol{\Phi}_{11}^{k|k-1}\boldsymbol{\Sigma}_{12}^k\boldsymbol{\Phi}_{22}^{k|k-1\mathsf{T}} & \boldsymbol{\Phi}_{11}^{k|k-1}\boldsymbol{\Sigma}_{13}^k\boldsymbol{\Phi}_{33}^{k|k-1\mathsf{T}} \\ \boldsymbol{\Phi}_{22}^{k|k-1}\boldsymbol{\Sigma}_{21}^k\boldsymbol{\Phi}_{11}^{k|k-1\mathsf{T}} & \boldsymbol{\Phi}_{22}^{k|k-1}\boldsymbol{\Sigma}_{22}^k\boldsymbol{\Phi}_{22}^{k|k-1\mathsf{T}} + \mathbf{Q}_{22}^{k-1} & \boldsymbol{\Phi}_{22}^{k|k-1}\boldsymbol{\Sigma}_{23}^k\boldsymbol{\Phi}_{33}^{k|k-1\mathsf{T}} \\ \boldsymbol{\Phi}_{33}^{k|k-1}\boldsymbol{\Sigma}_{31}^k\boldsymbol{\Phi}_{11}^{k|k-1\mathsf{T}} & \boldsymbol{\Phi}_{33}^{k|k-1}\boldsymbol{\Sigma}_{32}^k\boldsymbol{\Phi}_{22}^{k|k-1\mathsf{T}} & \boldsymbol{\Phi}_{33}^{k|k-1}\boldsymbol{\Sigma}_{33}^k\boldsymbol{\Phi}_{33}^{k|k-1\mathsf{T}} + \mathbf{Q}_{33}^{k-1} \end{bmatrix}.
$$

$$\tag{3.6b}$$

**Remark 5** *This filter formulation assumes that all agents $\mathsf{A}_i, i = 1, \ldots, 3$ perform the state propagation step at the same point in time. In real world application, the propagation is sometimes driven by a control input $\mathbf{u}$ or a proprioceptive sensor input (wheel encoder, IMU). In a distributed system, we cannot assume that the sensor read-out is synchronized and that, in the case of a heterogeneous system, the sampling rate is the same across agents. If these assumptions are not fulfilled, interpolation techniques might be applied. Another problem is the rate of propagation has to be performed, e.g, if it is based on the IMU readings, it might be processed at a rate of, e.g., $10\,\mathrm{Hz} - 1\,\mathrm{kHz}$. This would cause a distributed propagation every few milliseconds. Depending on the size of the individual state vector and the number of agents, a full connection with low latency and high throughput is required.*

### 3.2.3 Private update step

In general, a measurement/observation $\mathbf{z}$ is linearly related to the state $\mathbf{x}$ by the measurement sensitivity-/output coupling matrix $\mathbf{H}$ in the form

$$
\mathbf{z}^k = \mathbf{H}^k\mathbf{x}^k + \mathbf{v}^k, \tag{3.7}
$$

with an uncorrelated and additive measurement noise $\mathbf{v}^k \sim \mathcal{N}\left(\mathbf{0}, \mathbf{R}\right)$ (see Section 2.8.1).

An individual or private update step is performed in cases an exteroceptive measurement is referring only to a single agent's (ego-) state, meaning that the output of the individual dynamic system is not interconnected to other dynamical systems of the swarm, e.g., if a mobile robot obtains an absolute position measurement.

**Definition 2** *In a private observation, the measurement is linearly related (only) to the states of an individual agent's system, meaning that the output is just coupled with one dynamic (sub)system.*

Furthermore, we analyze the influence of cross-covariances $\boldsymbol{\Sigma}_{ij}$ between each robot, based on the standard Kalman filter update (see Equation (2.65)) and show that observations on an individual agent have an impact on the other members of the swarm.

We assume that agent $\mathsf{A}_1$ receives a private measurements, meaning that $\mathbf{H}_1^\ell \neq \mathbf{0}$. The global and consequently spare measurement matrix is

$$
\mathbf{H}^k = \begin{bmatrix} \mathbf{H}_1^k & \mathbf{0} & \mathbf{0} \end{bmatrix}. \tag{3.8a}
$$

The global Kalman gain (2.65c)

$$\mathbf{K}^k = \begin{bmatrix} \boldsymbol{\Sigma}_{11} & \boldsymbol{\Sigma}_{12} & \boldsymbol{\Sigma}_{13} \\ \boldsymbol{\Sigma}_{21} & \boldsymbol{\Sigma}_{22} & \boldsymbol{\Sigma}_{23} \\ \boldsymbol{\Sigma}_{31} & \boldsymbol{\Sigma}_{32} & \boldsymbol{\Sigma}_{33} \end{bmatrix}^{k(-)} \begin{bmatrix} \mathbf{H}_1^{k\mathsf{T}} \\ \mathbf{0} \\ \mathbf{0} \end{bmatrix} \left( \begin{bmatrix} \mathbf{H}_1^k & \mathbf{0} & \mathbf{0} \end{bmatrix} \begin{bmatrix} \boldsymbol{\Sigma}_{11} & \boldsymbol{\Sigma}_{12} & \boldsymbol{\Sigma}_{13} \\ \boldsymbol{\Sigma}_{21} & \boldsymbol{\Sigma}_{22} & \boldsymbol{\Sigma}_{23} \\ \boldsymbol{\Sigma}_{31} & \boldsymbol{\Sigma}_{32} & \boldsymbol{\Sigma}_{33} \end{bmatrix}^{k(-)} \begin{bmatrix} \mathbf{H}_1^{k\mathsf{T}} \\ \mathbf{0} \\ \mathbf{0} \end{bmatrix} + \mathbf{R}^k \right)^{-1}$$

$$= \begin{bmatrix} \boldsymbol{\Sigma}_{11}^{k(-)}\mathbf{H}_1^{k\mathsf{T}} \\ \boldsymbol{\Sigma}_{21}^{k(-)}\mathbf{H}_1^{k\mathsf{T}} \\ \boldsymbol{\Sigma}_{31}^{k(-)}\mathbf{H}_1^{k\mathsf{T}} \end{bmatrix} \left( \mathbf{H}_1^k \boldsymbol{\Sigma}_{11}^{k(-)}\mathbf{H}_1^{k\mathsf{T}} + \mathbf{R}^k \right)^{-1} = \begin{bmatrix} \mathbf{K}_1^k \\ \mathbf{K}_2^k \\ \mathbf{K}_3^k \end{bmatrix}$$

$$(3.8b)$$

The corrected covariance matrix $\boldsymbol{\Sigma}^{k(+)}$ is then (2.65f)

$$\boldsymbol{\Sigma}^{k(+)} = (\mathbf{I} - \mathbf{K}^k\mathbf{H}^k)\boldsymbol{\Sigma}^{k(-)} = \begin{bmatrix} \mathbf{I} - \mathbf{K}_1^k\mathbf{H}_1^k & \mathbf{0} & \mathbf{0} \\ -\mathbf{K}_2^k\mathbf{H}_1^k & \mathbf{I} & \mathbf{0} \\ -\mathbf{K}_3^k\mathbf{H}_1^k & \mathbf{0} & \mathbf{I} \end{bmatrix} \boldsymbol{\Sigma}^{k(-)} \tag{3.8c}$$

which results in the following updated covariance and cross-covariance blocks $\{\boldsymbol{\Sigma}_{ij} : i, j = 1, \ldots, 3\}$

$$\boldsymbol{\Sigma}_{11}^{k(+)} = (\mathbf{I} - \mathbf{K}_1^k\mathbf{H}_1^k)\boldsymbol{\Sigma}_{11}^{k(-)}$$
$$\boldsymbol{\Sigma}_{12}^{k(+)} = (\mathbf{I} - \mathbf{K}_1^k\mathbf{H}_1^k)\boldsymbol{\Sigma}_{12}^{k(-)}$$
$$\boldsymbol{\Sigma}_{13}^{k(+)} = (\mathbf{I} - \mathbf{K}_1^k\mathbf{H}_1^k)\boldsymbol{\Sigma}_{13}^{k(-)}$$
$$\boldsymbol{\Sigma}_{21}^{k(+)} = -\mathbf{K}_2^k\mathbf{H}_1^k\boldsymbol{\Sigma}_{11}^{k(-)} + \boldsymbol{\Sigma}_{21}^{k(-)}$$
$$\boldsymbol{\Sigma}_{22}^{k(+)} = -\mathbf{K}_2^k\mathbf{H}_1^k\boldsymbol{\Sigma}_{12}^{k(-)} + \boldsymbol{\Sigma}_{22}^{k(-)} \tag{3.8d}$$
$$\boldsymbol{\Sigma}_{23}^{k(+)} = -\mathbf{K}_2^k\mathbf{H}_1^k\boldsymbol{\Sigma}_{13}^{k(-)} + \boldsymbol{\Sigma}_{23}^{k(-)}$$
$$\boldsymbol{\Sigma}_{31}^{k(+)} = -\mathbf{K}_3^k\mathbf{H}_1^k\boldsymbol{\Sigma}_{11}^{k(-)} + \boldsymbol{\Sigma}_{31}^{k(-)}$$
$$\boldsymbol{\Sigma}_{32}^{k(+)} = -\mathbf{K}_3^k\mathbf{H}_1^k\boldsymbol{\Sigma}_{12}^{k(-)} + \boldsymbol{\Sigma}_{32}^{k(-)}$$
$$\boldsymbol{\Sigma}_{33}^{k(+)} = -\mathbf{K}_3^k\mathbf{H}_1^k\boldsymbol{\Sigma}_{13}^{k(-)} + \boldsymbol{\Sigma}_{33}^{k(-)}$$

The residual is

$$\mathbf{r}^k = \mathbf{z}^k \boxminus \mathbf{H}^k\hat{\mathbf{x}}^{k(-)} \tag{3.8e}$$

The global full state is corrected by

$$\hat{\mathbf{x}}^{k(+)} = \begin{bmatrix} \hat{\mathbf{x}}_1^{k(+)} \\ \hat{\mathbf{x}}_2^{k(+)} \\ \hat{\mathbf{x}}_3^{k(+)} \end{bmatrix} = \begin{bmatrix} \hat{\mathbf{x}}_1^{k(-)} \\ \hat{\mathbf{x}}_2^{k(-)} \\ \hat{\mathbf{x}}_3^{k(-)} \end{bmatrix} + \begin{bmatrix} \mathbf{K}_1^k \\ \mathbf{K}_2^k \\ \mathbf{K}_3^k \end{bmatrix} \mathbf{r}^k \tag{3.8f}$$

If the individual systems are independent (has no cross-covariance terms between agents) the above updates reduces, as expected, to an update on the individual agent:

$$\boldsymbol{\Sigma}_{11}^{k(+)} = (\mathbf{I} - \mathbf{K}_1^k\mathbf{H}_1^k)\boldsymbol{\Sigma}_{11}^{k(-)} \tag{3.9a}$$

$$\hat{\mathbf{x}}_1^{k(+)} = \hat{\mathbf{x}}_1^{k(-)} \boxplus \mathbf{K}_1^k\mathbf{r}^k \tag{3.9b}$$

If two agents $\mathsf{A}_1$ and $\mathsf{A}_2$ are correlated ($\boldsymbol{\Sigma}_{12} = \boldsymbol{\Sigma}_{21}^{\mathsf{T}} \neq \mathbf{0}$) then the global belief is updated by

$$
\begin{aligned}
\boldsymbol{\Sigma}_{11}^{k(+)} &= (\mathbf{I} - \mathbf{K}_1^k \mathbf{H}_1^k) \boldsymbol{\Sigma}_{11}^{k(-)} \\
\boldsymbol{\Sigma}_{12}^{k(+)} &= (\mathbf{I} - \mathbf{K}_1^k \mathbf{H}_1^k) \boldsymbol{\Sigma}_{12}^{k(-)} \\
\boldsymbol{\Sigma}_{21}^{k(+)} &= -\mathbf{K}_2^k \mathbf{H}_1^k \boldsymbol{\Sigma}_{11}^{k(-)} + \boldsymbol{\Sigma}_{21}^{k(-)} \\
\boldsymbol{\Sigma}_{22}^{k(+)} &= -\mathbf{K}_2^k \mathbf{H}_1^k \boldsymbol{\Sigma}_{12}^{k(-)} + \boldsymbol{\Sigma}_{22}^{k(-)}
\end{aligned}
\tag{3.10a}
$$

$$
\begin{aligned}
\hat{\mathbf{x}}_1^{k(+)} &= \hat{\mathbf{x}}_1^{k(-)} \boxplus \mathbf{K}_1^k \mathbf{r}^k \\
\hat{\mathbf{x}}_2^{k(+)} &= \hat{\mathbf{x}}_2^{k(-)} \boxplus \mathbf{K}_2^k \mathbf{r}^k
\end{aligned}
\tag{3.10b}
$$

which means, if individual agents are correlated, an observation on any of them inherently corrects/influences the estimates of the other correlated ones.

**Remark 6** *Aiming at a consistent DCSE, the impact of inter-agent correlations need to be considered and requires information exchange between agents for each individual observation.*

### 3.2.4 Joint update step

A joint/collective update step is performed, in cases an exteroceptive observation is referring to multiple agents (ego-) beliefs, meaning that the output of multiple dynamic systems is interconnected. Those agents are denoted as so-called *participants*, while the other agents of the swarm are denoted as *non-participants*. A joint observation is for instance, a relative pose observation from a mobile robot $\mathsf{A}_1$ to another robot $\mathsf{A}_2$.

**Definition 3** *In a joint observation, the measurement is linearly related to states of multiple agents' systems, meaning that the output of an arbitrary sets of dynamic models is coupled.*

A set of so-called participants $\mathbf{P}$ is defined as

$$
\mathbf{P} := \{\mathsf{P}_i \in \mathsf{A} | 1, \dots, P\},
\tag{3.11}
$$

while the set of non-participants is defined as

$$
\bar{\mathbf{P}} := \mathbf{A} \setminus \mathbf{P}.
\tag{3.12}
$$

Assuming a joint measurement between agent $\mathsf{A}_1$ and $\mathsf{A}_2$, meaning that $\mathbf{H}_1^k \neq \mathbf{0}$ and $\mathbf{H}_2^k \neq \mathbf{0}$, the resulting global and sparse measurement matrix is

$$
\mathbf{H}^k = \begin{bmatrix} \mathbf{H}_1^k & \mathbf{H}_2^k & \mathbf{0} \end{bmatrix}.
\tag{3.13a}
$$

The global Kalman gain (2.65c) is

$$
\begin{aligned}
\mathbf{K}^k &= \begin{bmatrix} \boldsymbol{\Sigma}_{11} & \boldsymbol{\Sigma}_{12} & \boldsymbol{\Sigma}_{13} \\ \boldsymbol{\Sigma}_{21} & \boldsymbol{\Sigma}_{22} & \boldsymbol{\Sigma}_{23} \\ \boldsymbol{\Sigma}_{31} & \boldsymbol{\Sigma}_{32} & \boldsymbol{\Sigma}_{33} \end{bmatrix}^{k(-)} \begin{bmatrix} \mathbf{H}_1^{k\mathsf{T}} \\ \mathbf{H}_2^{k\mathsf{T}} \\ \mathbf{0} \end{bmatrix} \left( \begin{bmatrix} \mathbf{H}_1^k & \mathbf{H}_2^k & \mathbf{0} \end{bmatrix} \begin{bmatrix} \boldsymbol{\Sigma}_{11} & \boldsymbol{\Sigma}_{12} & \boldsymbol{\Sigma}_{13} \\ \boldsymbol{\Sigma}_{21} & \boldsymbol{\Sigma}_{22} & \boldsymbol{\Sigma}_{23} \\ \boldsymbol{\Sigma}_{31} & \boldsymbol{\Sigma}_{32} & \boldsymbol{\Sigma}_{33} \end{bmatrix}^{k(-)} \begin{bmatrix} \mathbf{H}_1^{k\mathsf{T}} \\ \mathbf{H}_2^{k\mathsf{T}} \\ \mathbf{0} \end{bmatrix} + \mathbf{R}^k \right)^{-1} \\
&= \begin{bmatrix} \boldsymbol{\Sigma}_{11} \mathbf{H}_1^{k\mathsf{T}} + \boldsymbol{\Sigma}_{12}^{k(-)} \mathbf{H}_2^{k\mathsf{T}} \\ \boldsymbol{\Sigma}_{21} \mathbf{H}_1^{k\mathsf{T}} + \boldsymbol{\Sigma}_{22}^{k(-)} \mathbf{H}_2^{k\mathsf{T}} \\ \boldsymbol{\Sigma}_{31} \mathbf{H}_1^{k\mathsf{T}} + \boldsymbol{\Sigma}_{32}^{k(-)} \mathbf{H}_2^{k\mathsf{T}} \end{bmatrix} \left( \mathbf{S}^k \right)^{-1} = \begin{bmatrix} \mathbf{K}_1^k \\ \mathbf{K}_2^k \\ \mathbf{K}_3^k \end{bmatrix}
\end{aligned}
\tag{3.13b}
$$

with

$$\mathbf{S}^k = \left(\mathbf{H}_1^k \boldsymbol{\Sigma}_{11}^{k(-)} + \mathbf{H}_2 \boldsymbol{\Sigma}_{21}^{k(-)}\right) \mathbf{H}_1^{k\mathsf{T}} + \left(\mathbf{H}_1^k \boldsymbol{\Sigma}_{12}^{k(-)} + \mathbf{H}_2^k \boldsymbol{\Sigma}_{22}^{k(-)}\right) \mathbf{H}_2^{k\mathsf{T}} + \mathbf{R}^k \qquad (3.13c)$$

The corrected *a-posteriori* covariance matrix $\boldsymbol{\Sigma}^{k(+)}$ is then (2.65f)

$$\boldsymbol{\Sigma}^{k(+)} = \left(\mathbf{I} - \mathbf{K}^k \mathbf{H}^k\right) \boldsymbol{\Sigma}^{k(-)} = \begin{bmatrix} \mathbf{I} - \mathbf{K}_1^k \mathbf{H}_1^k & -\mathbf{K}_1^k \mathbf{H}_2^k & \mathbf{0} \\ -\mathbf{K}_2^k \mathbf{H}_1^k & \mathbf{I} - \mathbf{K}_2^k \mathbf{H}_2^k & \mathbf{0} \\ -\mathbf{K}_3^k \mathbf{H}_1^k & -\mathbf{K}_3^k \mathbf{H}_2^k & \mathbf{I} \end{bmatrix} \boldsymbol{\Sigma}^{k(-)} \qquad (3.13d)$$

which results in the following updated covariance and cross-covariance blocks $\{\boldsymbol{\Sigma}_{ij}^{k(+)} : i,j = 1,\ldots,3\}$

$$\begin{aligned}
\boldsymbol{\Sigma}_{11}^{k(+)} &= (\mathbf{I} - \mathbf{K}_1^k \mathbf{H}_1^k) \boldsymbol{\Sigma}_{11}^{k(-)} - \mathbf{K}_1^k \mathbf{H}_2^k \boldsymbol{\Sigma}_{21}^{k(-)} \\
\boldsymbol{\Sigma}_{12}^{k(+)} &= (\mathbf{I} - \mathbf{K}_1^k \mathbf{H}_1^k) \boldsymbol{\Sigma}_{12}^{k(-)} - \mathbf{K}_1^k \mathbf{H}_2^k \boldsymbol{\Sigma}_{22}^{k(-)} \\
\boldsymbol{\Sigma}_{13}^{k(+)} &= (\mathbf{I} - \mathbf{K}_1^k \mathbf{H}_1^k) \boldsymbol{\Sigma}_{13}^{k(-)} - \mathbf{K}_1^k \mathbf{H}_2^k \boldsymbol{\Sigma}_{23}^{k(-)} \\
\boldsymbol{\Sigma}_{21}^{k(+)} &= -\mathbf{K}_2^k \mathbf{H}_1^k \boldsymbol{\Sigma}_{11}^{k(-)} - (\mathbf{I} - \mathbf{K}_2^k \mathbf{H}_2^k) \boldsymbol{\Sigma}_{21}^{k(-)} \\
\boldsymbol{\Sigma}_{22}^{k(+)} &= -\mathbf{K}_2^k \mathbf{H}_1^k \boldsymbol{\Sigma}_{12}^{k(-)} - (\mathbf{I} - \mathbf{K}_2^k \mathbf{H}_2^k) \boldsymbol{\Sigma}_{22}^{k} \\
\boldsymbol{\Sigma}_{23}^{k(+)} &= -\mathbf{K}_2^k \mathbf{H}_1^k \boldsymbol{\Sigma}_{13}^{k(-)} - (\mathbf{I} - \mathbf{K}_2^k \mathbf{H}_2^k) \boldsymbol{\Sigma}_{23}^{k} \\
\boldsymbol{\Sigma}_{31}^{k(+)} &= -\mathbf{K}_3^k \mathbf{H}_1^k \boldsymbol{\Sigma}_{11}^{k(-)} - \mathbf{K}_3^k \mathbf{H}_2^k \boldsymbol{\Sigma}_{21}^{k(-)} + \boldsymbol{\Sigma}_{31}^{k(-)} \\
\boldsymbol{\Sigma}_{32}^{k(+)} &= -\mathbf{K}_3^k \mathbf{H}_1^k \boldsymbol{\Sigma}_{12}^{k(-)} - \mathbf{K}_3^k \mathbf{H}_2^k \boldsymbol{\Sigma}_{22}^{k(-)} + \boldsymbol{\Sigma}_{32}^{k(-)} \\
\boldsymbol{\Sigma}_{33}^{k(+)} &= -\mathbf{K}_3^k \mathbf{H}_1^k \boldsymbol{\Sigma}_{13}^{k(-)} - \mathbf{K}_3^k \mathbf{H}_2^k \boldsymbol{\Sigma}_{23}^{k(-)} + \boldsymbol{\Sigma}_{33}^{k(-)}
\end{aligned} \qquad (3.13e)$$

The residual is

$$\mathbf{r}^k = \mathbf{z}^k \boxminus \begin{bmatrix} \mathbf{H}_1^k & \mathbf{H}_2^k & \mathbf{0} \end{bmatrix} \hat{\mathbf{x}}^{k(-)} \qquad (3.13f)$$

The corrected *a-posteriori* global state is

$$\hat{\mathbf{x}}^{k(+)} = \begin{bmatrix} \hat{\mathbf{x}}_1^{k(+)} \\ \hat{\mathbf{x}}_2^{k(+)} \\ \hat{\mathbf{x}}_3^{k(+)} \end{bmatrix} = \begin{bmatrix} \hat{\mathbf{x}}_1^{k(-)} \\ \hat{\mathbf{x}}_2^{k(-)} \\ \hat{\mathbf{x}}_3^{k(-)} \end{bmatrix} \boxplus \begin{bmatrix} \mathbf{K}_1^k \\ \mathbf{K}_2^k \\ \mathbf{K}_3^k \end{bmatrix} \mathbf{r}^k \qquad (3.13g)$$

If the system has only cross-covariances between agent $\mathsf{A}_1$ and $\mathsf{A}_2$ then *a-posteriori* belief is corrected by

$$\begin{aligned}
\boldsymbol{\Sigma}_{11}^{k(+)} &= (\mathbf{I} - \mathbf{K}_1^k \mathbf{H}_1^k) \boldsymbol{\Sigma}_{11}^{k} - \mathbf{K}_1^k \mathbf{H}_2^k \boldsymbol{\Sigma}_{21}^{k} \\
\boldsymbol{\Sigma}_{12}^{k(+)} &= (\mathbf{I} - \mathbf{K}_1^k \mathbf{H}_1^k) \boldsymbol{\Sigma}_{12}^{k} - \mathbf{K}_1^k \mathbf{H}_2^k \boldsymbol{\Sigma}_{22}^{k} \\
\boldsymbol{\Sigma}_{21}^{k(+)} &= -\mathbf{K}_2^k \mathbf{H}_1^k \boldsymbol{\Sigma}_{11}^{k} - (\mathbf{I} - \mathbf{K}_2^k \mathbf{H}_2^k) \boldsymbol{\Sigma}_{21}^{k} \\
\boldsymbol{\Sigma}_{22}^{k(+)} &= -\mathbf{K}_2^k \mathbf{H}_1^k \boldsymbol{\Sigma}_{12}^{k} - (\mathbf{I} - \mathbf{K}_2^k \mathbf{H}_2^k) \boldsymbol{\Sigma}_{22}^{k}
\end{aligned} \qquad (3.14a)$$

and

$$\begin{aligned}
\hat{\mathbf{x}}_1^{k(+)} &= \hat{\mathbf{x}}_1^k \boxplus \mathbf{K}_1^k \mathbf{r} \\
\hat{\mathbf{x}}_2^{k(+)} &= \hat{\mathbf{x}}_2^k \boxplus \mathbf{K}_2^k \mathbf{r}
\end{aligned} \qquad (3.14b)$$

If correlations to agent $\mathsf{A}_3$ exist, a correction on the full belief needs to be performed.

**Lemma 1** *Corrections are obtained on statistically independent agents' states in the course of filter update steps, if any of them are indirectly correlated.*

**Proof:** Assuming three agents $\mathsf{A}_{\{1,2,3\}}$ and that two states, $\mathbf{x}_2$ and $\mathbf{x}_3$, are correlated, while $\mathbf{x}_1$ is independent from the others. By performing a joint observation (see Equation (3.13)) incorporating the $\mathsf{A}_1$ and $\mathsf{A}_2$, assuming per definition $\mathbf{H}_{\{1,2\}} \neq \mathbf{0}$, then the Kalman gain $\mathbf{K}_3^k \neq \mathbf{0}$ and *a-posteriori* belief of the non-participating agent $\mathsf{A}_3$ obtains a correction $\mathbf{x}_3^{k(+)} \neq \mathbf{x}_3^{k(-)}$. ∎

**Lemma 2** *Correlations between independent agents' states are obtained in the course of filter update steps, if any of them are indirectly correlated.*

**Proof:** Assuming three agents $\mathsf{A}_{\{1,2,3\}}$ and that two states, $\mathbf{x}_2$ and $\mathbf{x}_3$, are correlated ($\mathbf{\Sigma}_{23} \neq \mathbf{0}$), while $\mathbf{x}_1$ is independent from the others ($\mathbf{\Sigma}_{\{12,13\}} = \mathbf{0}$). By performing a joint observation (see Equation (3.13)) incorporating the $\mathsf{A}_1$ and $\mathsf{A}_2$, assuming per definition $\mathbf{H}_{\{1,2\}} \neq \mathbf{0}$, then *a-posteriori* belief $\mathbf{x}_1^{(+)}$ of $\mathsf{A}_1$ is correlated with to the two states ($\mathbf{\Sigma}_{\{12,13\}}^{(+)} \neq \mathbf{0}$). ∎

**Remark 7** *Aiming at a consistent DCSE, all beliefs and cross-covariances between non-participating but correlated agents to the participating ones are needed. In the worst case, it would require all-to-all communication during processing the joint observation.*

### 3.2.5 Generalized decoupled update step

Private and joint observations are technically the same, while later relates, in addition to the local state estimate, to estimates from one or multiple other agents. Therefore, we can distinguish between *participants* (with the subscript $p$) and *non-participants* (with the subscript $o$ for "others").

The *generalized decoupled update step* according to the Kalman filter (see Equation (2.65)) can be computed for a set of participants $\mathbf{P}$ and non-participants $\bar{\mathbf{P}}$ as follows. For legibility we will neglect the discrete time index $\{\}^k$.

The stacked/partitioned random variable is in the form

$$\mathbf{x} = \begin{bmatrix} \mathbf{x}_p \\ \mathbf{x}_o \end{bmatrix}, \tag{3.15a}$$

while $\mathbf{x}_p$ is a joint belief of participants, e.g., constituting of $\mathsf{A}_i$'s and $\mathsf{A}_j$'s belief $\mathbf{x}_p^\mathsf{T} = \begin{bmatrix} \mathbf{x}_i^\mathsf{T} & \mathbf{x}_j^\mathsf{T} \end{bmatrix}$ and $\mathbf{x}_o$ a joint belief of others.

An the corresponding partitioned joint covariance is given by

$$\mathbf{\Sigma}^k = \begin{bmatrix} \mathbf{\Sigma}_{pp} & \mathbf{\Sigma}_{po} \\ \mathbf{\Sigma}_{po}^\mathsf{T} & \mathbf{\Sigma}_{oo} \end{bmatrix}^k, \tag{3.15b}$$

while the participants might be correlated by $\mathbf{\Sigma}_{po} \neq \mathbf{0}$ with any of the non-participating agents $\mathsf{A}_o \in \bar{\mathbf{P}}$.

Per definition the states of non-participants is not related to the observation, meaning that $\mathbf{H}_o = \mathbf{0}$. Consequently, the partitioned measurement sensitivity matrix in the form

$$\mathbf{H} = \begin{bmatrix} \mathbf{H}_p & \mathbf{0} \end{bmatrix}. \tag{3.15c}$$

The measurement residual is

$$\mathbf{r} = \mathbf{z} \boxminus \begin{bmatrix} \mathbf{H}_p & \mathbf{0} \end{bmatrix} \begin{bmatrix} \hat{\mathbf{x}}_p^{(-)} \\ \hat{\mathbf{x}}_o^{(-)} \end{bmatrix} = \mathbf{z} \boxminus \mathbf{H}_p \hat{\mathbf{x}}_p^{(-)} \tag{3.15d}$$

The partitioned Kalman gain is

$$\mathbf{K} = \begin{bmatrix} \mathbf{K}_p \\ \mathbf{K}_o \end{bmatrix} = \begin{bmatrix} \boldsymbol{\Sigma}_{pp}^{(-)}\mathbf{H}_p^\mathsf{T} \\ \boldsymbol{\Sigma}_{po}^{(-)}\mathbf{H}_p^\mathsf{T} \end{bmatrix} \left(\mathbf{S}^k\right)^{-1}, \tag{3.15e}$$

with the innovation covariance $\mathbf{S}^k$

$$\mathbf{S}^k = \mathbf{H}_p\boldsymbol{\Sigma}_{pp}^{(-)}\mathbf{H}_p^\mathsf{T} + \mathbf{R}, \tag{3.15f}$$

and $\mathbf{R}$ being the measurement noise covariance.

The partitioned *a posteriori* covariance is

$$\boldsymbol{\Sigma}^{(+)} = \begin{bmatrix} (\mathbf{I} - \mathbf{K}_p\mathbf{H}_p)\boldsymbol{\Sigma}_{pp}^{(-)} - \mathbf{K}_p & (\mathbf{I} - \mathbf{K}_p\mathbf{H}_p)\boldsymbol{\Sigma}_{po}^{(-)} \\ \bullet & -\mathbf{K}_o\mathbf{H}_p\boldsymbol{\Sigma}_{po}^{(-)} + \boldsymbol{\Sigma}_{oo}^{(-)} \end{bmatrix} \tag{3.15g}$$

The *a posteriori* mean $\hat{\mathbf{x}}^{(+)}$ is

$$\hat{\mathbf{x}}^{(+)} = \begin{bmatrix} \hat{\mathbf{x}}_p^{(+)} \\ \hat{\mathbf{x}}_o^{(+)} \end{bmatrix} = \begin{bmatrix} \hat{\mathbf{x}}_p^{(-)} \boxplus \mathbf{K}_p\mathbf{r} \\ \hat{\mathbf{x}}_o^{(-)} \boxplus \mathbf{K}_o\mathbf{r} \end{bmatrix}, \tag{3.15h}$$

### 3.2.6  Analysis of joint observation and properties

In past decades, the effect of different relative observations on the localization accuracy in a swarm of agents was studied in-depth.

In [122, 126], Roumeliotis and Rekleitis, studied the improvement in localization accuracy per additional agent as the size of swarm increases and provided an analytical expression for the theoretical upper bound on the position uncertainty increase rate in case of a missing absolute information. Meaning that the entire swarm is drifting globally, since only odometry and only relative position measurements are obtained. Two interesting findings were made: First, the growth-rate of the uncertainty is inversely proportional to the number of agents. Therefore, the contribution of each additional agent follows a law of diminishing return. Second, the rate of growth depends only on the number of agents and odometric uncertainty, and not on the accuracy of the relative position measurements.

In [101] and [102], Martinelli *et al.* investigated in the effects when a swarm of agents is capable of simultaneously observing one another in a EKF formulation based on the work of Roumeliotis and Bekey [121]. Different relative observation where studied: relative bearing, relative distance, and relative orientation. They showed and proved that relative bearing provides more information than distance and orientation observations.

Later in [109], Mourikis and Roumeliotis studied the localization performance and uncertainty growth given relative position measurement graphs (RPMG) of different topologies. In [111], they investigated on the upper bound analysis for collaborative SLAM. In [108], Mirzaei *et al.* studied the problem of determining the upper bounds for the covariance of the position estimates in CLATT. This analysis also unveiled that the agents' position estimates are always better when, in addition to inter-agent measurements, the agents fuse agent-to-target measurements.

These studies give important insights and propositions, that are applicable for distributed CSE. Important insights from those studies are that (i) even if one estimator observes absolute position information, e.g., provided by a GNSS, the position information of all agents in the swarm remain bounded and converge, (ii) the agent equipped with the most precise sensors has the greatest impact on the overall accuracy of the swarm, and (iii) upper bounds for the covariance can be determined, therefore, allowing to predict the worst-case performance of a swarm with a particular sensor suite.

### 3.2.7 Conclusion

The ability to assume that individual agent's dynamic system are not coupled, see Equation (3.5), is a fundamental pillar on which DCSE is built on and allows us to propose various decoupling strategies as described later in Section 3.4.

The challenge in Distributed Collaborative Localization (DCL) is a decoupled, decentralized architecture that performs equivalently to a joint centralized solution (denoted as *exact* solution).

**Definition 4** *A filter formulation that provides (temporary) equal estimates as a centralized filter formulation, which is operating on a global full state in every filter step, is denoted as* exact.

## 3.3 DCSE based on a Distributed Approximated History (DCSE-DAH)

In this section, we describe our DCSE-DAH approach [76] to address the DCSE problem in an *approximated/inexact* way, which is an extension to the DCSE-DACC approach, that was proposed by Luft *et al.* in [94, 95], and allows us to reduce the maintenance effort for factorized cross-covariances therein.

### 3.3.1 Introduction

Recent work by Luft *et al.* in [94, 95] and ourselves [73] have shown that communication is just required for joint observations and that the communication complexity in pairwise joint observations can be reduced to $\mathcal{O}(1)$. Yet, the local *maintenance* effort for interdependencies (i.e. inter-agent cross-covariances) still increases linearly with the number of met agents. This renders the DCSE based on Distributed Approximated Cross-Covariances (DCSE-DACC) approach [95] ill-suited for large swarms and long-duration missions with many encounters.

Therefore, the primary motivation for our DCSE-DAH approach is the aim for a constant complexity $\mathcal{O}(1)$ in both, maintenance and communication, with respect to number of agents in the swarm.

In aided-inertial navigation systems, the state propagation is typically performed at a rate of the IMU which is mostly above 100 Hz. Not only the state, but also all interdependencies due to cross-correlations with met agents have to be propagated at this rate. The more agents are correlated, the more *maintenance* is required, even after not seeing each other for a long time and hinders the scalability with respect to the number of met agents.

To overcome that issue, we propose to keep locally just (i) the most recent factorized cross-covariance between agents, (ii) a timestamp of the event, and (iii) a sliding window buffer $\mathcal{B}$ keeping track of the last corrections. The factorized cross-covariance and timestamp can, e.g., be stored in a dictionary Dict accessed via the other agent's ID. Meaning that the DCSE-DAH approach is restoring the dated cross-correlation between agents at the moment they meet again as an extension to DCSE-DACC. Our experiments in Section 3.5 manifest the improvement while accomplishing the same accuracy.

As DCSE-DACC and DCSE-DAH operate only on a subset of swarm agents' beliefs, we denote the filter steps as *isolated*, emphasizing that they are performed isolated from the rest of the swarm.

### 3.3.2 Isolated Propagation step

This approach assumes that the beliefs/states of the individual nodes/agents can be predicted independently. Meaning that the prediction model (e.g., a motion model) allows an independent state transition and that the distributed agents do not require information about the latest cross-covariance of the swarms nor the current global full-state, while they are predicting.

One of the most trivial solutions to avoid communications between agents while they perform their state propagation steps, e.g., based on a control input or proprioceptive sensor readings, can be obtained by a delayed cross-covariance prediction.

In [121, 123], Roumeliotis and Bekey proposed in their approach, which we term DCSE-DP, to factorize cross-covariances between agents and to exploit the symmetry of the covariance matrix, in order to perform the state propagation step fully distributed, while it is possible to restore cross-covariances between agents exactly.

As covariance matrices are symmetric $\boldsymbol{\Sigma}_{ij} = \boldsymbol{\Sigma}_{ji}^{\mathsf{T}}$, the information about the cross-covariances between agents is kept redundantly. The lower- or upper-triangular components can be restored exploiting the symmetry of full state's covariance

$$\boldsymbol{\Sigma}_{ji} = \left(\boldsymbol{\Sigma}_{ij}\right)^{\mathsf{T}}, \ \{i, j\} := 1, \dots, N, \ i \neq j. \tag{3.16}$$

Therefore, the upper-right triangular matrix hold basically the same information as the lower-left triangular matrix, e.g, in case of three agents

$$\boldsymbol{\Sigma} = \begin{bmatrix} \boldsymbol{\Sigma}_{11} & \boldsymbol{\Sigma}_{12} & \boldsymbol{\Sigma}_{13} \\ \boldsymbol{\Sigma}_{12}^{\mathsf{T}} & \boldsymbol{\Sigma}_{22} & \boldsymbol{\Sigma}_{23} \\ \boldsymbol{\Sigma}_{13}^{\mathsf{T}} & \boldsymbol{\Sigma}_{23}^{\mathsf{T}} & \boldsymbol{\Sigma}_{33} \end{bmatrix} = \begin{bmatrix} \boldsymbol{\Sigma}_{11} & \boldsymbol{\Sigma}_{12} & \boldsymbol{\Sigma}_{13} \\ \bullet & \boldsymbol{\Sigma}_{22} & \boldsymbol{\Sigma}_{23} \\ \bullet & \bullet & \boldsymbol{\Sigma}_{33} \end{bmatrix}. \tag{3.17}$$

In [121], it was proposed to *factorize* blocks of the upper-triangular cross-covariance by performing a singular value decomposition (SVD)

$$\boldsymbol{\Sigma}_{ij} = \mathbf{U}_{ij} \mathbf{S}_{ij} \mathbf{V}_{ij}^{\mathsf{T}} = \left(\mathbf{U}_{ij} \mathbf{S}_i\right) \left(\mathbf{V}_{ij} \mathbf{S}_j\right)^{\mathsf{T}} = \mathcal{S}_{ij} \mathcal{S}_{ji}^{\mathsf{T}}, \tag{3.18}$$

with diagonal matrix of eigenvalues $\mathbf{S}_{ij}$, and the diagonal matrices $\mathbf{S}_i$ and $\mathbf{S}_j$ such that $\mathbf{S}_{ij} = \mathbf{S}_i \mathbf{S}_j^{\mathsf{T}}$, e.g., by $\mathbf{S}_j = \mathbf{I}$ and $\mathbf{S}_i = \mathbf{S}_{ij}$. $\mathcal{S}$ is a so-called *cross-covariance factor*.

As proposed by Luft *et al.* in [94], one can avoid the SVD – at the cost of more memory needed to store the factors, if the dimensions of individual agents' states are different in size – by decomposing the cross-covariance $\boldsymbol{\Sigma}_{ij} = \mathcal{S}_{ij} \left(\mathcal{S}_{ji}\right)^{\mathsf{T}}$ into $\mathcal{S}_{ij} = \boldsymbol{\Sigma}_{ij}$ and $\mathcal{S}_{ji} = \mathbf{I}_{n \times n}$ with $n = \dim(\mathbf{x}_j)$.

By inspecting Equation (3.6b), cross-covariances can be propagated from a timestamp $t^m$ to a timestamp $t^k$ by

$$\boldsymbol{\Sigma}_{ij}^k = \boldsymbol{\Phi}_{ii}^{k|m} \boldsymbol{\Sigma}_{ij}^m \left(\boldsymbol{\Phi}_{jj}^{k|m}\right)^{\mathsf{T}} \tag{3.19}$$

In case of a factorized covariance matrix of the global full state of, e.g., three agents, is propagated as

$$\boldsymbol{\Sigma}^k = \begin{bmatrix} \boldsymbol{\Phi}_{11}^{k|m} \boldsymbol{\Sigma}_{11}^m \boldsymbol{\Phi}_{11}^{k|m\mathsf{T}} + \mathbf{Q}_{11}^m & \boldsymbol{\Phi}_{11}^{k|m} \mathcal{S}_{12}^m \left(\boldsymbol{\Phi}_{22}^{k|m} \mathcal{S}_{21}^m\right)^{\mathsf{T}} & \boldsymbol{\Phi}_{11}^{k|m} \mathcal{S}_{13}^m \left(\boldsymbol{\Phi}_{33}^{k|m} \mathcal{S}_{31}^m\right)^{\mathsf{T}} \\ \bullet & \boldsymbol{\Phi}_{22}^{k|m} \boldsymbol{\Sigma}_{22}^m \boldsymbol{\Phi}_{22}^{k|m\mathsf{T}} + \mathbf{Q}_{22}^m & \boldsymbol{\Phi}_{22}^{k|m} \mathcal{S}_{23}^m \left(\boldsymbol{\Phi}_{33}^{k|m} \mathcal{S}_{32}^m\right)^{\mathsf{T}} \\ \bullet & \bullet & \boldsymbol{\Phi}_{33}^{k|m} \boldsymbol{\Sigma}_{33}^m \boldsymbol{\Phi}_{33}^{k|m\mathsf{T}} + \mathbf{Q}_{33}^m \end{bmatrix}. \tag{3.20}$$

From above, factorized cross-covariances can be propagated from a timestamp $t^m$ to a timestamp $t^k$ by

$$\boldsymbol{\Sigma}_{ij}^k = \boldsymbol{\Phi}_{ii}^{k|m} \mathcal{S}_{ij}^m \left(\boldsymbol{\Phi}_{jj}^{k|m} \mathcal{S}_{ji}^m\right)^{\mathsf{T}}. \tag{3.21}$$

In order to obtain a fully distributed propagation step, each agent $\mathsf{A}_i$ needs to maintain a set of cross-covariance factors $\mathcal{S}_{ij}, j = 1, \dots, N, j \neq i$ which can be (uniquely) associated to another agent $A_j$, e.g., in a dictionary (see Section 2.2) $\mathcal{C}_i := \text{Dict}\{\mathcal{S}_{ij}\}$ that is accessed via the other agent's unique ID $\text{id}_{A_j}$ by

$$\mathcal{C}_i \left( \text{id}_{A_j} \right) = \mathcal{S}_{ij}. \tag{3.22}$$

In DCSE-DP [121], this set of cross-covariance factors needs to be propagated in each (ego-) propagation step by

$$\mathcal{C}_i \left( \text{id}_{A_j} \right)^k = \boldsymbol{\Phi}_{ii}^{k|m} \mathcal{C}_i \left( \text{id}_{A_j} \right)^m, \; j = 1, \ldots, |A|, \; j \neq i. \tag{3.23}$$

In DCSE-DACC [95], this set is limited per agent $A_i$ to a set of *met* agents $M_i$ – those who participated in joint observations – only. According to the remark 7, this leads to an *inexact* (see Definition 4) global full state. As only subset of cross-covariance factors is considered, we denote this distributed, but inexact propagation step as *isolated propagation step*.

$$\mathcal{C}_i \left( \text{id}_{A_j} \right)^k = \boldsymbol{\Phi}_{ii}^{k|m} \mathcal{C}_i \left( \text{id}_{A_j} \right)^m, \; j = 1, \ldots, |A|, \; j \neq i, \; A_j \in M_i \tag{3.24}$$

Despite being *inexact*, the *isolated propagation step* is based on two simplifications made.

**Lemma 3** *Correlations between independent agents' states can only be obtained in the curse of filter update steps, in case of decoupled system inputs (no state, no control input coupling and no process noise coupling).*

    *Proof:* If two agents $A_i$ and $A_j$ beliefs $\mathbf{x}_i$ and $\mathbf{x}_j$ are uncorrelated prior the propagation step at $t^k$, meaning that their cross-covariance has a block zero entries $\boldsymbol{\Sigma}_{ij}^k = \mathbf{0}$ ($\boldsymbol{\Sigma}_{ij} = \boldsymbol{\Sigma}_{ij}^{\mathsf{T}}$) in the global full state's covariance, and the network dynamics model's state transition, input coupling, and process noise matrices only have block diagonal entries (corresponding only the individual agent's belief) by $\{\boldsymbol{\Phi}, \boldsymbol{\Gamma}, \mathbf{G}\}_{ij} = \mathbf{0}, \{i, j\} = 1, \ldots, |A|, i \neq j$, then the beliefs remain independent after the prediction step $\boldsymbol{\Sigma}_{ij}^{k+1} = \mathbf{0}$, following Equation (3.4). ∎

In large swarms, the number of met agents is typically smaller than the total number of agents $|M_i| << |A|$, meaning that the maintenance effort in the propagation step for factorized cross-covariances in DCSE-DACC is lower than in CCSE, while in the limit, the maintenance effort of both strategies has a complexity of $\mathcal{O}(|A|)$.

Our inspiration to achieve a constant maintenance complexity stem from the observation that, in case no intermittent corrections happened, the decomposed covariances can be propagated *exactly* over a wider time span by applying two transition matrix series $\bar{\boldsymbol{\Phi}}_{\{ii,jj\}}^{k|m}$ from $t^m$ to $t^k$. A state transition matrix series $\bar{\boldsymbol{\Phi}}^{k|m}$ from $t^m$ to $t^k$ is obtained by accumulating consecutive state transition matrices in the form

$$\bar{\boldsymbol{\Phi}}_{ii}^{k|m} = \prod_{i=m}^{k} \boldsymbol{\Phi}_{ii}^{i|i-1}. \tag{3.25}$$

A propagated cross-covariance at $t^k$ is obtained by applying $\bar{\boldsymbol{\Phi}}_{\{ii,jj\}}^{k|m}$ in the form

$$\boldsymbol{\Sigma}_{ij}^k = \bar{\boldsymbol{\Phi}}_{ii}^{k|m} \mathcal{S}_{ij}^m \left( \bar{\boldsymbol{\Phi}}_{jj}^{k|m} \mathcal{S}_{ji}^m \right)^{\mathsf{T}}. \tag{3.26}$$

To achieve a constant maintenance complexity for factorized cross-covariances in the propagation step, we propose to store a transition matrix $\boldsymbol{\Phi}_{ii}$ – a so-called correction term – in a sliding time-horizon buffer (see Section 2.2) Hist for corrections $\boldsymbol{\mathcal{B}}_i := \mathsf{Hist}\{\{\boldsymbol{\Phi}_{ii}, \boldsymbol{\Upsilon}_{ii}, \boldsymbol{\Lambda}_{ii}\}\}$ on each agent $\mathsf{A}_i$ (the terms $\boldsymbol{\Upsilon}_{ii}$ and $\boldsymbol{\Lambda}_{ii}$ are defined in Section 3.3.3)

$$\boldsymbol{\mathcal{B}}_i\left(t^k\right) = \boldsymbol{\Phi}_{ii}^k. \tag{3.27}$$

Once a cross-covariance factor is needed again, e.g., during an isolated joint update at $t^k$ with another agent $\mathsf{A}_j$, the dated cross-covariance factors stemming from $t^m$ can be restored exactly by applying an accumulated state correction matrix $\mathbf{M}^4$ using the correction terms of the correction buffer $\boldsymbol{\mathcal{B}}_i$ by

$$\mathcal{S}_{ij}^m = \mathcal{C}_i(\mathsf{id}_{\mathsf{A}_j}), \tag{3.28a}$$

$$\mathbf{M}_{ii}^{k|m} = \prod_{l=m}^{k} \boldsymbol{\mathcal{B}}_i\left(t^l\right), \tag{3.28b}$$

and

$$\mathcal{S}_{ij}^k = \mathbf{M}_{ii}^{k|m} \mathcal{S}_{ij}^m. \tag{3.28c}$$

The steps of Equation (3.28) (see Algorithm 3.7) need to be performed on $\mathsf{A}_j$ to obtain $\mathcal{S}_{ji}^k = \mathbf{M}_{jj}^{k|m} \mathcal{S}_{ji}^m$. This allows to restore $\boldsymbol{\Sigma}_{ij}^k$ by applying Equation (3.18) and to obtain $\boldsymbol{\Sigma}_{ji}^k$ given the symmetry in Equation (3.16).

### 3.3.3  Isolated Decoupled Observations

Starting from the generalized decoupled update step (see Section 3.2.5), we will describe approximations made to achieve isolated decoupled observations in the DCSE-DACC and DCSE-DAH approaches.

**Definition 5** *Isolated decoupled observations require communication only among participants and the communication complexity scales linearly with the number of participating agent $\mathcal{O}(P)$ (two links per participant are needed), while in general $|\mathbf{P}| << |\mathbf{A}|$.*

We assume that the computation of an observation is performed on an interim master agent $\mathsf{A}_i \in \mathbf{P}$, e.g., based on the lowest integer ID among participants[5].

In order to achieve this isolated communication constraint, we assume that the interim master obtains the *a-priori* beliefs and cross-covariance factors from all participants, which allows us to recover the joint belief of participants at $t^k$. Once the interim master $\mathsf{A}_i$ obtains the noisy coupled measurement $\mathbf{z}^k$, it requests from the participants the *a-priori* ego-belief and cross-covariance factors relating to the participants $\mathbf{P}$. While in DCSE-DACC the *a-priori* cross-covariance factors relating to $t^k$ would be already available in the dictionary $\mathcal{C}_p, p \in \mathbf{P}$, in DCSE-DAH the dated factors need to be first forward propagated using the accumulated state correction matrix series obtained in the individuals correction buffers $\boldsymbol{\mathcal{B}}_p, p \in \mathbf{P}$ (see Equation (3.28) and Algorithm 3.7). This step is performed distributed (in parallel) on the participants.

After the interim master agent $\mathsf{A}_i$ obtained the set of *a-priori* beliefs $\{\hat{\mathbf{x}}_p^{k(-)} \boldsymbol{\Sigma}_p^{k(-)}, p \in \mathbf{P}\}$ and factorized cross-covariances $\{\mathcal{S}_{mn}^{k(-)}, \{m, n\} \in \mathbf{P}, m \neq n\}$. The upper triangular cross-covariance block entries can be restored by

$$\boldsymbol{\Sigma}_{mn}^{k(-)} = \mathcal{S}_{mn}^{k(-)} \left(\mathcal{S}_{nm}^{k(-)}\right)^{\mathsf{T}}, \{m, n\} \in \mathbf{P}, m \neq n. \tag{3.29a}$$

---

[4] We named it on purpose *accumulated state correction matrix* instead of *state transition matrix series*, since we will later introduce other correction terms for private and joint update steps in Section 3.3.3.

[5] The primary motivations are (i) to reduce the complexity and improve the readability of the algorithm and (ii) to avoid additional communication and process synchronization in the update step.

The lower triangular cross-covariance blocks are obtained by symmetry

$$\boldsymbol{\Sigma}_{nm}^{k(-)} = \left(\boldsymbol{\Sigma}_{mn}^{k(-)}\right)^{\mathsf{T}}, \{m, n\} \in \mathbf{P}, m \neq n. \tag{3.29b}$$

Finally, all blocks need to be stacked to obtain the joint belief of participants $\mathbf{x}_p \sim \mathcal{N}\left(\hat{\mathbf{x}}_p, \boldsymbol{\Sigma}_{pp}\right)$ (see Equation (3.2)) in the form

$$\mathbf{x} = \begin{bmatrix} \mathbf{x}_1 \\ \vdots \\ \mathbf{x}_P \end{bmatrix}^{k(-)} \sim \mathcal{N}\left(\begin{bmatrix} \hat{\mathbf{x}}_1 \\ \vdots \\ \hat{\mathbf{x}}_P \end{bmatrix}, \begin{bmatrix} \boldsymbol{\Sigma}_{11} & \dots & \boldsymbol{\Sigma}_{1P} \\ \bullet & \ddots & \dots \\ \bullet & \bullet & \boldsymbol{\Sigma}_{PP} \end{bmatrix}\right)^{k(-)}, P = |\mathbf{P}|. \tag{3.29c}$$

Note that the order of participants' ego-beliefs needs to be aligned with the columns of the measurement sensitivity matrix $\mathbf{H}$ for the particular decoupled observation, meaning that the columns of $\mathbf{H}^k = \begin{bmatrix} \mathbf{H}_p^k & \mathbf{H}_o^k \end{bmatrix}$, with $\mathbf{H}_o^k = \mathbf{0}$ needs to be aligned with the $\hat{\mathbf{x}}_p$.

$$\mathbf{H}^k = \begin{bmatrix} \begin{bmatrix} \mathbf{H}_1 & \dots & \mathbf{H}_P \end{bmatrix} & \mathbf{0} \end{bmatrix}^k, P = |\mathbf{P}|. \tag{3.29d}$$

Up to the previous step, no approximations were made, and the beliefs are exact. In the following we describe, why approximations are needed in isolated update steps. As defined in Equation (3.15e), the partitioned Kalman gain is

$$\mathbf{K} = \begin{bmatrix} \mathbf{K}_p \\ \mathbf{K}_o \end{bmatrix} = \begin{bmatrix} \boldsymbol{\Sigma}_{pp}^{(-)} \mathbf{H}_p^{\mathsf{T}} \\ \boldsymbol{\Sigma}_{po}^{(-)} \mathbf{H}_p^{\mathsf{T}} \end{bmatrix} \left(\mathbf{S}^k\right)^{-1}, \tag{3.30}$$

with $\mathbf{S}^k = \mathbf{H}_p \boldsymbol{\Sigma}_{pp}^{(-)} \mathbf{H}_p^{\mathsf{T}} + \mathbf{R}$. We are not able to restore the cross-covariance to non-participants $\boldsymbol{\Sigma}_{po}$ in an isolated update, despite some participants might have factorized cross-covariances $\mathcal{S}_{po} \neq \mathbf{0}$, but the corresponding counterpart $\mathcal{S}_{op}$ is temporarily unavailable. Consequently, the Kalman gain for other $\mathbf{K}_o$ cannot be calculated.

Inspired by the suboptimal SKF, the Kalman gain of non-participants is forced to be zero, resulting in the Schmidt-Kalman gain $\breve{\mathbf{K}}$ according to Equation (2.72d)

$$\breve{\mathbf{K}} = \begin{bmatrix} \mathbf{K}_p \\ \mathbf{0} \end{bmatrix}. \tag{3.31}$$

Consequently, non-participants do not directly benefit from participants' observations.

**Lemma 4** *In isolated decoupled observations, correlations can only be obtained among participant's belief, despite some participants might be correlated to some non-participants.*

   *Proof:* Assuming a correlation between a participant $\mathsf{A}_i$ and non-participants is $\boldsymbol{\Sigma}_{io} \neq \mathbf{0}$, but no correlation between the participant $\mathsf{A}_j$ and non-participants $\boldsymbol{\Sigma}_{jo} = \mathbf{0}$. In an isolated joint update between $\mathsf{A}_i$ and $\mathsf{A}_j$, the Kalman gain for non-participants is forcibly set to $\mathbf{K}_p = \mathbf{0}$. Therefore, no state correction will be applied $\{\hat{\mathbf{x}}_o, \boldsymbol{\Sigma}_{oo}\}^{(+)} = \{\hat{\mathbf{x}}_o, \boldsymbol{\Sigma}_{oo}\}^{(-)}$ and no correlation will be obtained $\boldsymbol{\Sigma}_{jo}^{(+)} = \boldsymbol{\Sigma}_{jo}^{(-)} = \mathbf{0}$.    ∎

The approximated partitioned *a posteriori* covariance $\breve{\boldsymbol{\Sigma}}^{(+)}$ according to Equation (3.15g) is consequently

$$\breve{\boldsymbol{\Sigma}}^{(+)} = \begin{bmatrix} (\mathbf{I} - \mathbf{K}_p \mathbf{H}_p) \boldsymbol{\Sigma}_{pp}^{(-)} - \mathbf{K}_p & (\mathbf{I} - \mathbf{K}_p \mathbf{H}_p) \boldsymbol{\Sigma}_{po}^{(-)} \\ \bullet & \boldsymbol{\Sigma}_{oo}^{(-)} \end{bmatrix}. \tag{3.32}$$

The *a posteriori* mean $\breve{\mathbf{x}}^{(+)}$ is

$$\breve{\mathbf{x}}^{(+)} = \begin{bmatrix} \hat{\mathbf{x}}_p^{(+)} \\ \breve{\mathbf{x}}_o^{(+)} \end{bmatrix} = \begin{bmatrix} \hat{\mathbf{x}}_p^{(-)} \boxplus \mathbf{K}_p \mathbf{r} \\ \hat{\mathbf{x}}_o^{(-)} \end{bmatrix}. \tag{3.33}$$

Despite the approximations made for non-participants' beliefs, the cross-covariance of non-participants can be corrected $\boldsymbol{\Sigma}_{po}^{(+)} = \boldsymbol{\Upsilon}_p \boldsymbol{\Sigma}_{po}^{(-)}$ (see Equation (3.32)) using

$$\boldsymbol{\Upsilon}_p = (\mathbf{I} - \mathbf{K}_p \mathbf{H}_p). \tag{3.34}$$

For private observations (case of a single participant) the participant, e.g., $\mathsf{A}_i$, applies the correction factor $\boldsymbol{\Upsilon}$ (Equation (3.34)) on the corresponding element of buffer.

$$\boldsymbol{\mathcal{B}}_i \left( t^k \right) = \boldsymbol{\Upsilon}_i^k \boldsymbol{\mathcal{B}}_i \left( t^k \right), \; i \in \mathbf{P}. \tag{3.35}$$

In DCSE-DACC, this correction factor is directly applied on all factorized cross-covariances held in $\boldsymbol{\mathcal{C}}_i$, meaning this operation scales linearly with the number of met agents.

For joint observations with multiple participants, $|\mathbf{P}| > 1$ , Luft *et al.* proposed in [94, 95], that each participant applies a correction in relation to gained information

$$\boldsymbol{\Lambda}_i^k = \boldsymbol{\Sigma}_{ii}^{k(+)} \left( \boldsymbol{\Sigma}_{ii}^{k(-)} \right)^{-1}, i \in \mathbf{P} \tag{3.36}$$

on the individual participants cross-covariance factors. In DCSE-DAH these individual correction terms are applied on the corresponding element of the buffer by

$$\boldsymbol{\mathcal{B}}_i \left( t^k \right) = \boldsymbol{\Lambda}_i^k \boldsymbol{\mathcal{B}}_i \left( t^k \right), \; i \in \mathbf{P} \tag{3.37}$$

and allows reducing the maintenance effort for cross-covariance factors held in $\boldsymbol{\mathcal{C}}$.

The correction factor $\boldsymbol{\Lambda}_i$, is reasonable approximation if participants $\mathbf{P}$ are strongly directly or indirectly correlated with non-participants $\bar{\mathbf{P}}$ before the joint observation.

### 3.3.4 Buffer maintenance and propagation strategy

In this subsection, we describe how the cross-covariances are propagated and updated, once needed for joint observations, and how we prevent cross-covariances to fall out of the past horizon. Figure 3.1 shows how correction factors from different events are used to propagate a previous cross-covariance forward, which is also described in Algorithm 3.7.

To keep cross-covariances in the buffer's time horizon, we suggest performing a sanity check, e.g., at the end of each propagation step. The aim is to find factorized cross-covariances by their timestamps, that are exactly at the border of the time horizon. In that case, we perform immediately a forward propagation using the entire history, which is described in Algorithm 3.3.

Consequently, a smaller buffer increases the chance that a forward propagation is performed in propagation steps (assuming sporadic joint observations). In the best case, the buffer size matches the ratio between the rate of the propagation sensor and the rate of joint observations rendering the approach constant in maintenance complexity. Note that setting the buffer size of $\boldsymbol{\mathcal{B}}$ to 1, DCSE-DAH emulates DCSE-DACC.

### 3.3.5 Algorithm

In this section, the DCSE-DAH algorithm [76] based on the EKF formulation is described and should be executed on each agent of a swarm (see  Algorithm 3.1), as depict in the block diagram in Figure 3.2. Each agent's estimator consist of a unique identifier $\mathsf{id}_p$, the latest belief $\{\hat{\mathbf{x}}_p^k, \boldsymbol{\Sigma}_{pp}^k\}$, a correction buffer $\boldsymbol{\mathcal{B}}_p$, and a dictionary with the factorized cross-covariance $\mathsf{Dict}_p$. The isolated state propagation step is described in Algorithm 3.2 for an agent $\mathsf{A}_p$ with a non-linear state transition function $\phi_p(\mathbf{x}_p, \mathbf{u}_p)$. A private isolated update step based on a non-linear measurement function $h_{\mathrm{private}}(\mathbf{x})$ is described in Algorithm 3.4. A joint isolated update step with two participants, $\mathsf{A}_{\{i,j\}}$, based on a non-linear measurement function $h_{\mathrm{joint}}(\mathbf{x}_i, \mathbf{x}_j)$, $\{i, j\} \in \mathbf{P}$ is described in Algorithm 3.8, while this can be extended to an arbitrary number of participants. Please note, that joint observations need to identify other agents. Communication between agents is just required during processing joint observations, while the other filter steps are performed independently. In order to fully exploit the benefits of the correction buffer $\mathcal{B}_p$, factorized cross-covariance need to be forward propagated with a common *accumulated state correction matrix* $\mathbf{M}$ (see Equation (3.28)), which is satisfied by the $set\_fcc()$ and $get\_fcc()$ algorithms described in Algorithm 3.6 and Algorithm 3.5, respectively.

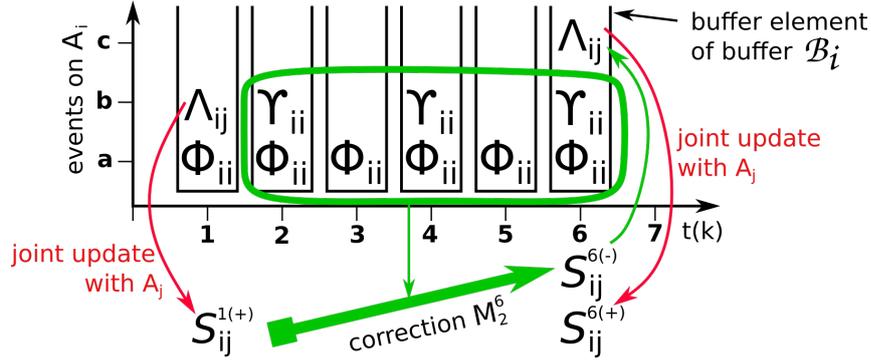

**Figure 3.1:** Decomposed cross-covariance forward propagation scheme using elements with accumulated correction terms from the buffer $\mathcal{B}_i$ on $\mathsf{A}_i$. At $t(k) = 1$, $\mathsf{A}_i$ performs a joint observation with the uncorrelated $\mathsf{A}_j$ resulting in a correction $\mathbf{\Lambda}_{ij}^1$ (event b at $t^1$) and a decomposed cross-covariance $\mathcal{S}_{ij}^{1(+)}$. Propagations and private updates result in $\mathbf{\Phi}$ and $\mathbf{\Upsilon}$, respectively. The events are left multiplied in order on the buffer elements. At $t(k) = 6$, the correlated agents perform again a joint observation. Therefore, each agent accumulates its buffer elements to forward propagate $\mathcal{S}_{ij}^{1(+)} \rightarrow \mathcal{S}_{ij}^{6(-)}$ by $\mathbf{M}_2^6$ (Equation (3.28) and Algorithm 3.7). After that, a new correction factor $\mathbf{\Lambda}_{ij}^6$ and cross-covariance factor $\mathcal{S}_{ij}^{6(+)}$ is obtained. Image reused from [76].

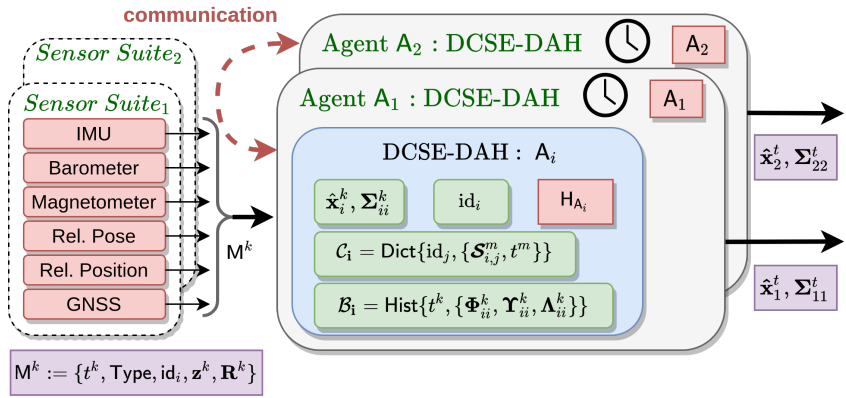

**Figure 3.2:** Shows the block diagram of the DCSE-DAH approach for individual agents. The sensor suite provides measurements to the estimator. For inter-agent joint observation, information exchange between estimators is needed. Details are provided in Section 3.3.

---

**Algorithm 3.1:** DCSE-DAH: Measurement handling on $\mathsf{A}_p$

**Input** : $\mathrm{id}_p, \hat{\mathbf{x}}_p^k, \mathbf{\Sigma}_{ii}^k, \boldsymbol{\mathcal{B}}_p, \mathsf{Dict}_p, \mathbf{z}^k, \mathbf{R}^k, t^k$

1 **if** $\mathbf{z}^k$ *is a proprioceptive measurement or control input* **then**
2     isolated_propagation($\hat{\mathbf{x}}_p^{k(-)}, \mathbf{\Sigma}_{pp}^{k(-)}, \boldsymbol{\mathcal{B}}_p, \mathbf{z}^k, \mathbf{R}^k, \mathsf{Dict}_p, t^k$) (Alg. 3.2)
3 **end**
4 **if** $\mathbf{z}^k$ *is a private observation on* $\mathsf{A}_i$ **then**
5     isolated_private($\hat{\mathbf{x}}_p^k, \mathbf{\Sigma}_{pp}^k, \boldsymbol{\mathcal{B}}_p, \mathbf{z}^k, \mathbf{R}^k, t^k$) (Alg. 3.4)
6 **end**
7 **if** $\mathbf{z}^k$ *is a joint observation between* $\mathsf{A}_i$ *and* $\mathsf{A}_j$ **then**
8     isolated_joint($\hat{\mathbf{x}}_p^{k(-)}, \mathbf{\Sigma}_{pp}^{k(-)} \mathbf{z}^k, \mathbf{R}^k, \mathrm{id}_{\{i,j\}}, \mathsf{Dict}_p, \boldsymbol{\mathcal{B}}_p, t^k$) (Alg. 3.8)
9 **end**

---

**Algorithm 3.2:** DCSE-DAH: Isolated Propagation on $\mathsf{A}_p$

**Input** : $\hat{\mathbf{x}}_p^{k(-)}, \mathbf{\Sigma}_{pp}^{k(-)}, \boldsymbol{\mathcal{B}}_p, \mathbf{u}^k, \mathbf{N}^k, \mathsf{Dict}_p, t^k$

1 $\mathbf{\Phi}_p^k = \left[ \frac{\partial \phi_p(\mathbf{x}_p, \mathbf{u}_p)}{\partial \mathbf{x}_p}(\hat{\mathbf{x}}_p, \mathbf{u}_p) \right]^{k-1}$
2 $\mathbf{G}_p^k = \left[ \frac{\partial \phi_p(\mathbf{x}_p, \mathbf{u}_p)}{\partial \mathbf{u}}(\hat{\mathbf{x}}_p, \mathbf{u}_p) \right]^{k-1}$
3 $\mathbf{Q}^k = \mathbf{G}_p^k \mathbf{N}^k (\mathbf{G}_p^k)^\mathsf{T}$
4 $\hat{\mathbf{x}}_p^k = \phi_p(\hat{\mathbf{x}}_p^k, \mathbf{u}^k)$
5 $\mathbf{\Sigma}_{pp}^k = \mathbf{\Phi}_p^{k|k-1} \mathbf{\Sigma}_{pp}^{k-1} (\mathbf{\Phi}_p^{k|k-1})^\mathsf{T} + \mathbf{Q}^{k-1}$
6 $\boldsymbol{\mathcal{B}}_p\left(t^k\right) = \mathbf{\Phi}^{k|k-1}$ (push back)
7 check_horizon($\boldsymbol{\mathcal{B}}_p, \mathsf{Dict}_p, t^k$) (Alg. 3.3)

---

**Algorithm 3.3:** DCSE-DAH: check_horizon on $\mathsf{A}_p$

**Input** : $\boldsymbol{\mathcal{B}}_p, \mathsf{Dict}, t^k$

1 $t^{\mathrm{oldest}} = min(\boldsymbol{\mathcal{B}}_p)$
2 **for** $\{t^m, \mathcal{S}^{m(-)}, \mathrm{id}\}$ in $\mathsf{Dict}$ **do**
3     $\mathbf{M}_m^k = $ compute_corr($\boldsymbol{\mathcal{B}}_p, t^m, t^k$) (Alg. 3.7)
4     **if** $t^m \equiv t^{oldest}$ **then**
5        $\mathsf{Dict}(\mathrm{id}) = \{\mathbf{M}_m^k \mathcal{S}^{m(-)}, t^k\}$ (forward prop.)
6     **end**
7 **end**

---

**Algorithm 3.4:** DCSE-DAH: Isolated Private Observation on $\mathsf{A}_p$

**Input** : $\hat{\mathbf{x}}_p^{k(-)}, \mathbf{\Sigma}_{pp}^{k(-)}, \boldsymbol{\mathcal{B}}_p, \mathbf{z}^k, \mathbf{R}^k, t^k$

1 $\mathbf{H}_p = \left[ \frac{\partial h_{\mathrm{private}}(\mathbf{x}_p)}{\partial \mathbf{x}_p}(\hat{\mathbf{x}}_p) \right]^{k(-)}$
2 $\mathbf{K}_p = \mathbf{\Sigma}_{pp}^{k(-)} \mathbf{H}_p^\mathsf{T} (\mathbf{H}_p \mathbf{\Sigma}_{pp}^{k(-)} \mathbf{H}_p^\mathsf{T} + \mathbf{R}^k)^{-1}$
3 $\hat{\mathbf{x}}_p^{k(+)} = \hat{\mathbf{x}}_p^{k(-)} \boxplus \mathbf{K}_p \left( \mathbf{z}^k \boxminus h(\hat{\mathbf{x}}_p) \right)$
4 $\mathbf{\Sigma}_{pp}^{k(+)} = (\mathbf{I} - \mathbf{K}_p \mathbf{H}_p) \mathbf{\Sigma}_{pp}^{k(-)}$
5 $\mathbf{\Upsilon}_p^k = (\mathbf{I} - \mathbf{K}_p \mathbf{H}_p)$
6 $\boldsymbol{\mathcal{B}}_p\left(t^k\right) = \mathbf{\Upsilon}_p^k \boldsymbol{\mathcal{B}}_p\left(t^k\right)$

---

**Algorithm 3.5:** DCSE-DAH: get_fcc on $A_i$

---

 **Input** : $\mathcal{B}_i, \mathrm{id}_j, \mathrm{Dict}_i, t^m$
1 $\{\mathcal{S}_{ij}^k, t^k\} = \mathrm{Dict}_i(\mathrm{id}_j)$
2 **if** $t^k \neq t^m$ **then**
3     /* Note: make sure that all $\mathcal{S}$ are forward propagated */
4     $\mathbf{M}^{k|m} = \mathrm{compute\_correction}(\mathcal{B}_i, t^m, t^k)$ (Alg. 3.7)
5     **for** $\{t^n, \mathcal{S}^{(-)}, \mathrm{id}\}$ **in** $\mathrm{Dict}_i$ **do**
6         **if** $t^n \equiv t^m$ **then**
7             $\mathrm{Dict}_i(\mathrm{id}) = \{\mathbf{M}^{k|m}\mathcal{S}^{n(-)}, t^k\}$ (forward prop.)
8         **end**
9     **end**
10 **end**
11 $\{\mathcal{S}_{ij}^k, t^k\} = \mathrm{Dict}_i(\mathrm{id}_j)$
 **Output:** $\mathcal{S}_{ij}^k$

---

**Algorithm 3.6:** DCSE-DAH: set_fcc on $A_i$

---

 **Input** : $\mathcal{B}_i, \mathrm{id}_j, \mathrm{Dict}_i, t^m, \mathcal{S}_{ij}^m$
1 $\mathrm{Dict}_i(\mathrm{id}_j) = \{\mathcal{S}_{ij}^{m(+)}, t^m\}$
2 /* Note: make sure that all $\mathcal{S}$ are forward propagated to the latest one */
3 $t^k = min(\mathcal{B}_i)$
4 /* Note: this branch will only be issued, if a new agent was met */
5 **if** $t^m \neq t^k$ **then**
6     $\mathbf{M}^{m|k} = \mathrm{compute\_corr}(\mathcal{B}_i, t^k, t^m)$ (Alg. 3.7)
7     **for** $\{t^n, \mathcal{S}^n, \mathrm{id}\}$ **in** $\mathrm{Dict}$ **do**
8         **if** $t^n \equiv t^k$ **then**
9             $\mathrm{Dict}(\mathrm{id}) = \{\mathbf{M}^{m|k}\mathcal{S}^n, t^m\}$ (forward prop.)
10         **end**
11     **end**
12 **end**

---

**Algorithm 3.7:** DCSE-DAH: compute_correction on $A_p$

---

 **Input** : $\mathcal{B}_p, t^{m-1}, t^m, \Delta t$
1 $\mathbf{M}_{m-1}^m = \mathbf{I}$
2 **for** $i \leftarrow t^{m-1} + \Delta t$ **to** $t^m$ **by** $\Delta t$ **do**
3     $\mathbf{M}_{m-1}^m = \mathcal{B}_p(i)\mathbf{M}_{m-1}^m$
4 **end**
 **Output:** $\mathbf{M}_{m-1}^m$

**Algorithm 3.8:** DCSE-DAH: Isolated Joint Observation on $\mathsf{A}_{\{i,j\}}$

**Input** : $\hat{\mathbf{x}}_{\{i,j\}}^{k(-)}, \boldsymbol{\Sigma}_{\{ii,jj\}}^{k(-)} {}^{i}\mathbf{z}_j^{\ k}, \mathbf{R}^k, \mathsf{id}_{\{i,j\}}, \mathsf{Dict}_{\{i,j\}}, \boldsymbol{\mathcal{B}}_{\{i,j\}}$

1  **if** $\mathsf{id}_i < \mathsf{id}_j$ /* one possibility */ **then**

2    |  /* Interim Master */  $\mathsf{A}_i$ receives $\{\hat{\mathbf{x}}_j^{k(-)}, \boldsymbol{\Sigma}_{jj}^{k(-)}, \mathsf{id}_j, \mathcal{S}_{ji}^{k(-)}\}$ from $\mathsf{A}_j$

3    |  $\mathcal{S}_{ij}^{k(-)} = \mathsf{get\_fcc}(\boldsymbol{\mathcal{B}}_i, \mathsf{id}_j, \mathsf{Dict}_i, t^k)$ (Alg. 3.5)

4    |  $\boldsymbol{\Sigma}_{ij}^{k(-)} = \mathcal{S}_{ij}^{k(-)}(\mathcal{S}_{ji}^{k(-)})^{\mathsf{T}}$

5    |  $\boldsymbol{\Sigma}_{pp}^{k(-)} = \begin{bmatrix} \boldsymbol{\Sigma}_{ii} & \boldsymbol{\Sigma}_{ij} \\ \boldsymbol{\Sigma}_{ij}^{\mathsf{T}} & \boldsymbol{\Sigma}_{jj} \end{bmatrix}^{k(-)}$

6    |  $\mathbf{H}_p = \left[ \frac{\partial h_{\text{joint}}(\mathbf{x}_i, \mathbf{x}_j)}{\partial \mathbf{x}_i}(\hat{\mathbf{x}}_i, \hat{\mathbf{x}}_j) \quad \frac{\partial h_{\text{joint}}(\mathbf{x}_i, \mathbf{x}_j)}{\partial \mathbf{x}_j}(\hat{\mathbf{x}}_i, \hat{\mathbf{x}}_j) \right]^{k(-)}$

7    |  $\mathbf{K}_p = \boldsymbol{\Sigma}_{pp}^{k(-)} \mathbf{H}_p^{\mathsf{T}} (\mathbf{H}_p \boldsymbol{\Sigma}_{pp}^{k(-)} \mathbf{H}_p^{\mathsf{T}} + \mathbf{R}^k)^{-1}$

8    |  $\hat{\mathbf{x}}_p^{k(-)} = \left[ (\hat{\mathbf{x}}_i^{k(-)})^{\mathsf{T}} \quad (\hat{\mathbf{x}}_j^{k(-)})^{\mathsf{T}} \right]^{\mathsf{T}}$

9    |  $\hat{\mathbf{x}}_p^{k(+)} = \hat{\mathbf{x}}_p^{k(-)} \boxplus \mathbf{K}_p \left( {}^{i}\mathbf{z}_j^{\ k} \boxminus h(\hat{\mathbf{x}}_i, \hat{\mathbf{x}}_j) \right)$

10   |  $\boldsymbol{\Sigma}_{pp}^{k(+)} = (\mathbf{I} - \mathbf{K}_p \mathbf{H}_p) \boldsymbol{\Sigma}_{pp}^{k(-)}$

11   |  /* Note: split $\boldsymbol{\Sigma}_{pp}^{k(+)}$ and $\hat{\mathbf{x}}_p^{k(+)}$ again */

12   |  $\mathcal{S}_{ij}^{k(+)} = \boldsymbol{\Sigma}_{ij}^{k(+)}$

13   |  $\mathsf{set\_fcc}(\boldsymbol{\mathcal{B}}_i, \mathsf{id}_j, \mathsf{Dict}_i, t^k, \mathcal{S}_{ij}^{k(-)})$ (Alg. 3.6)

14   |  $\mathcal{S}_{ji}^{k(+)} = \mathbf{I}$

15   |  $\mathsf{A}_i$ sends $\{\hat{\mathbf{x}}_i^{k(+)}, \boldsymbol{\Sigma}_{jj}^{k(+)}, \mathsf{id}_i, \mathcal{S}_{ji}^{k(+)}\}$ to $\mathsf{A}_j$

16   |  $\boldsymbol{\Lambda}_i^k = \boldsymbol{\Sigma}_{ii}^{k(+)} (\boldsymbol{\Sigma}_{ii}^{k(-)})^{-1}$

17   |  $\boldsymbol{\mathcal{B}}_i\left(t^k\right) = \boldsymbol{\Lambda}_i^k \boldsymbol{\mathcal{B}}_i\left(t^k\right)$

18   |  $\mathsf{Dict}_i(\mathsf{id}_j) = \{\mathcal{S}_{ij}^{k(+)}, t^k\}$

19  **else**

20   |  $\mathcal{S}_{ji}^{k(-)} = \mathsf{get\_fcc}(\boldsymbol{\mathcal{B}}_j, \mathsf{id}_i, \mathsf{Dict}_j, t^k)$ (Alg. 3.5)

21   |  $\mathsf{A}_j$ sends $\{\hat{\mathbf{x}}_j^{k(-)}, \boldsymbol{\Sigma}_{jj}^{k(-)}, \mathsf{id}_j, \mathcal{S}_{ji}^{k(-)}\}$ to $\mathsf{A}_i$

22   |  $\mathsf{A}_j$ receives $\{\hat{\mathbf{x}}_j^{k(+)}, \boldsymbol{\Sigma}_{jj}^{k(+)}, \mathsf{id}_i, \mathcal{S}_{ji}^{k(+)}\}$ from $\mathsf{A}_i$

23   |  $\boldsymbol{\Lambda}_j^k = \boldsymbol{\Sigma}_{jj}^{k(+)} (\boldsymbol{\Sigma}_{jj}^{k(-)})^{-1}$

24   |  $\boldsymbol{\mathcal{B}}_j\left(t^k\right) = \boldsymbol{\Lambda}_j^k \boldsymbol{\mathcal{B}}_j\left(t^k\right)$

25   |  $\mathsf{set\_fcc}(\boldsymbol{\mathcal{B}}_j, \mathsf{id}_i, \mathsf{Dict}_j, t^k, \mathcal{S}_{ji}^{k(+)})$ (Alg. 3.6)

26  **end**

## 3.4 Architecture Overview

In this section, we present diagrammatically some system architectures for filter-based CSE on a group of three heterogeneous agents (with different sensor modalities $S_1, \ldots, S_3$) to emphasize the difference between our proposed DCSE-DAH approach and closely related approaches. These diagrams should give an answer to the questions: "Where is the data processed?" and "What data is maintained?".

All obtain proprioceptive measurements (orange arrow), agent $A_i$ and $A_j$ private ones (red arrow), and a relative joint observation is obtained between agent $A_k$ and $A_i$ (blue arrow). $A_i$ and $A_j$ are jointly observing a commonly known object of interest (held in blue), e.g., the synchronous relative bearing angles to a moving object in order to triangulate its position. In this context, an *object of interest* might be a remarkable landmark or observed/tracked target. Its state is part of the estimation problem/global full state. These arrows indicate, what sensor measurements are obtained and shows in which filter they are processed, while thickness of these arrows should indicate the amount/frequency.

Each agent is associated with a unique identifier and has a globally synchronized reference time (held in purple). The time synchronization needs to be at least twice as accurate as the measurement update rate of the fastest sensor in the swarm, allowing accurately associating estimated states with obtained measurements [6].

The measurements are assumed to be processed in the order they are obtained (no out-of-sequence updates).

The communication link is assumed to be perfect, meaning a full and permanent connectivity, insignificant latency, sufficient bandwidth, and no packet are drops.

For our analysis of different filter strategies, we assume a group of $N$ homogeneous agents with a global full state of $M$ elements[7], and $m = \frac{M}{N}$ being the state vector length of the individual agent's state. We assume that all agents provide synchronously and alternating proprioceptive and exteroceptive sensor readings. We assume homogeneous senors with a single sensor measurement containing $r$ elements and in total $R = Nr$ elements per filter step.

### 3.4.1 Centralized CSE (CCSE)

In a centralized architecture, as shown in Figure 3.3, available information is processed in a central entity, denoted as Fusion Center (FC), with a global full state estimator as described in Section 3.2. Agents require a persistent communication with the FC, to receive the latest beliefs and to provide extero- and proprioceptive sensor data to the FC. Given known models and noise characteristics of the individual agent's systems (in a heterogeneous swarm, each node might be modeled individually), the FC provides exact estimates (see Definition 4).

A disadvantage of a naive implementation is that, individual agents' proprioceptive sensor readings need to be processed synchronously to perform a simultaneous state prediction step on the global full state. Additionally, we assume that all exteroceptive sensor readings are processed synchronously as well.

#### Properties

Regarding memory consumption, the global full belief (mean and covariance) needs to be stored. Assuming global full state with $M$ elements, in total $M(M + 1)$ elements for the

---

global belief (mean and covariance) need to be stored in the FC.

Assuming a synchronous information exchange, each synchronous filter step requires $2N$ communication links: $N$ for transmitting the measurement from the agents to the FC and $N$ messages to reset/correct the beliefs of $N$ agents by the FC. Depending on the problem at hand, it might not be required to immediately update all agents with the latest beliefs. But when it comes to fast motions and aggressive maneuvers in closed-loop control, immediate estimates with low latency are needed to satisfy a set trajectory. Meaning that the round-trip time between a measurement is received, it was processed on the FC, and the corrected belief is received again at the agent might be a limitation for real-world applications. Please note that, once agents are correlated, any measurement among them will result in updated beliefs (see Lemma 1).

Regarding message size and latency, two naive possibilities to transmit the corrected belief exist. Either a global full-state is broadcasted by the fusion center, which can, based on the network type and topology, minimize the latency, or individual direct messages are sent sequentially to the individual agents, requiring a routing protocol. Assuming a homogeneous swarm with a global full state of $M$ elements and $N$ agents, either $M + Nm^2$ elements in case of belief broadcasting, or $N$ times $(m + 1)m$ elements in case of sequential messages have to be transmitted. Before the corrected belief can be processed, the agents need to transmit their latest sensor reading with $R$ elements to the FC, leading to a total message size per filter step of $R + N(m + 1)m$ elements.

Regarding connectivity, messages sent by the FC must be received by the agents, if no direct link can be established, the existence of a spanning tree rooted at the FC, a routing protocol/scheme, and a persistent star-graph-based communication is required.

Despite communication constraints, offloading the computation effort from a mobile and computationally constrained agent to a centralized fusion center, might pave the way for new possibilities/applications. It might allow reducing the energy consumption on the agents and, thus, to extend the mission time. It might allow processing resource intensive algorithms, which might be well-scaling on the platform the FC is executed by exploiting different hardware accelerators (multiple CPU cores or GPUs acceleration, etc..). In [59], we investigated on offloading a computation intense VIO algorithm at different resolution, from small MAVs partially or fully on a cloud or edge facility, assuming a high bandwidth and low latency of a 5G mobile network in order to improve the localization accuracy.

### 3.4.2 CCSE with Distributed Propagation (CCSE-DP)

One of the most trivial solutions to avoid communication between agents while they perform their state propagation steps, e.g., based on a control input or proprioceptive sensor readings, can be obtained by a delayed cross-covariance prediction.

This approach assumes that the beliefs/states of the individual nodes can be predicted independently, meaning that the prediction model (e.g., a motion model) allows an independent state transition, and that the distributed agents do not require information about the latest cross-covariance of the swarms nor the current global full-state, while they are predicting.

In Equation (3.6b), the cross-covariance between agent $\mathsf{A}_1$ and $\mathsf{A}_2$ is propagated by their state transition matrices $\mathbf{\Phi}_{11}^{k|k-1}$ and $\mathbf{\Phi}_{22}^{k|k-1}$, by $\mathbf{\Sigma}_{12}^k = \mathbf{\Phi}_{11}^{k|k-1} \mathbf{\Sigma}_{12}^{k-1} \mathbf{\Phi}_{22}^{k|k-1^\mathsf{T}}$.

As shown in Figure 3.4, proprioceptive sensor readings are processed locally on each agent to evolve an agent's local belief. In order to propagate the cross-covariance relating to other agents, a history/sequence of the state transition matrix $\mathbf{\Phi}$ is maintained (cross-covariances are only required in the update step and can be recovered at any update incidence), a so-called state transition history $\mathcal{F} = \mathsf{Hist}\{\mathbf{\Phi}\}$ or an accumulated state transition matrix $\bar{\bar{\mathbf{\Phi}}}_{ii}^{k|m} = \prod_{l=m}^k \mathbf{\Phi}_{ii}^{k|l}$, which gets reset after the latest update step on the FC by $\bar{\boldsymbol{\Phi}}_{ii}^{k|m} = \mathbf{I}$. Once any of them obtains a private or joint observation at $t^k$, all agents need to send their current beliefs $\{\hat{\mathbf{x}}_i^k, \boldsymbol{\Sigma}_{ii}^k\}$ and an accumulated history of the state transition matrices $\bar{\boldsymbol{\Phi}}_{ii}^{k|m}$ to the FC. This allows to restore the cross-covariance between agents stemming from the previous update at $t^m$ exactly in the FC by

$$\boldsymbol{\Sigma}_{ij}^k = \bar{\boldsymbol{\Phi}}_{ii}^{k|m} \boldsymbol{\Sigma}_{ij}^m \left(\bar{\boldsymbol{\Phi}}_{jj}^{k|m}\right)^\mathsf{T}, \ \{i,j\} = 1,\ldots,N, \ i \neq j. \tag{3.38}$$

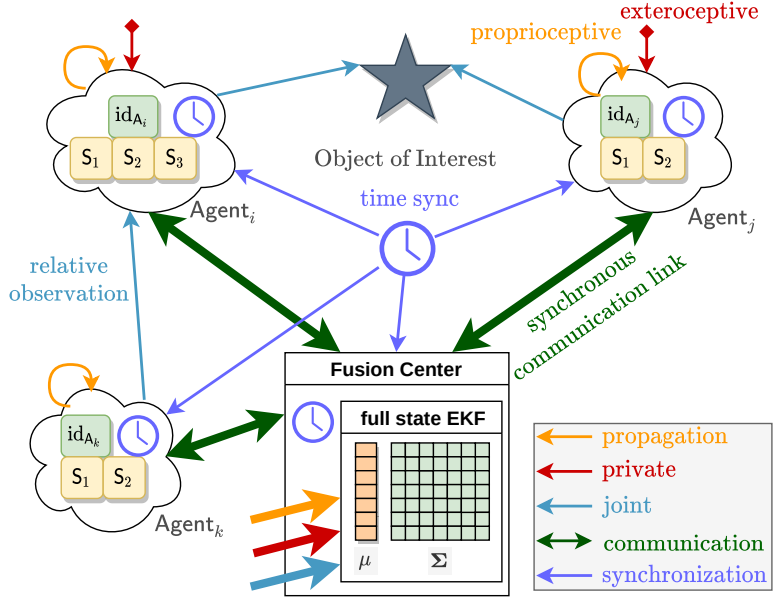

**Figure 3.3:** CCSE: Time-synchronized agents transmit all locally obtained sensor measurements to a central sensor fusion entity, the Fusion Center (FC). The FC maintains a filter that estimates the entire swarms' full state and sends the corrected states to the corresponding agents after each filter step.

Then, the FC is able to process the observation given the global full state according to Equation (2.65), sends the corrected/updated beliefs back to all agents, and if agents are using a state transition matrix series, they need to reset it by $\bar{\boldsymbol{\Phi}}_{ii}^{k|m} = \mathbf{I}$ or in case of the history buffer, they can clear it.

**Properties**

Regarding memory consumption, each agent maintains its local belief and additionally needs to store the accumulate state transition matrix. Assuming a homogeneous swarm with $N$ agents and global full state with $M$ elements, in total $\frac{M}{N}(1 + \frac{M}{N} + \frac{M}{N})$ (mean, covariance, and accumulated state transition matrix) elements need to be stored per agent. Additionally, the FC needs to store the global full belief with $M(M+1)$ elements. In total, $M(M+1) + Nm(1+2m)$ elements.

Assuming a synchronous information exchange, each synchronous filter update step requires $2N$ communication links: $N$ for transmitting the measurement, the latest beliefs, and accumulated transition matrices from the agents to the FC and $N$ links from the FC to send the corrected beliefs back to all agents[8].

---

[8]For simplicity, we assume that all nodes can be reached directly from the FC, otherwise a routing mechanism is needed, which inherently increases the number of links per message. Further, we assume that the transmitted data fits in a single message.

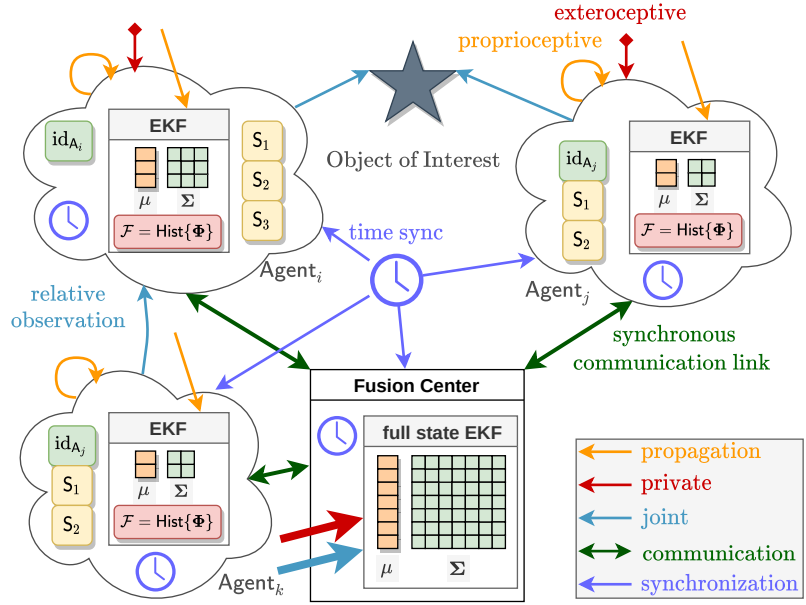

**Figure 3.4:** CCSE-DP: Each time-synchronized agent maintains a filter locally, that allows processing prediction steps (evolving the estimate in time) based on obtained proprioceptive sensor measurements locally. Each agent maintains a history of state transition matrices $\boldsymbol{\Phi}$, which are accumulated and sent with the most recent beliefs to a FC once any agent obtains a exteroceptive measurement to forward propagate the dated cross-covariances of the full state. After the update was processed on the full state, the FC sends the corrected beliefs back to the agents.

Regarding message size, each agent has to transmit and receive the local belief (mean and covariance), an accumulated transition matrix with equal size as its covariance and the measurement. Assuming a homogeneous swarm with $N$ agents, each agent needs to send $r + m(1 + 2m)$ elements and receive $m(1 + m)$ elements per exteroceptive measurement. Thus, in total $R + Nm(2 + 3m)$ elements need to be communicated per update step.

Regarding connectivity, the same constraints as in Section 3.4.1 apply.

### 3.4.3  DCSE with Distributed Propagation (DCSE-DP)

Like CCSE-DP, this approach assumes that the beliefs/states of the individual nodes can be predicted independently. Meaning that the prediction model (e.g., a motion model) allows an independent state transition. In [121, 123], Roumeliotis and Bekey proposed to factorize cross-covariances between agents to exploit the symmetry of the covariance matrix, in order to perform the state propagation step fully distributed. At the same time, it is possible to restore cross-covariances between agents exactly, and thus, the estimation results are centralized-equivalent.

Figure 3.5 shows the fully distributed approach by Roumeliotis and Bekey, which distributes the update step across the agents, meaning that no FC is needed, but it requires a persistent all-to-all communication at the moment of a private or joint update. Each agent maintains factorized off-diagonal block-elements of their corresponding row of the global full state in, e.g., a dictionary container (see Section 2.2).

Proprioceptive sensor readings are processed locally on each agent to evolve the agent's local belief and factorized cross-covariance relating to other agents in time. For instance, if agents are using a high-rate IMU as state propagation sensor, these measurements can be processed independently, asynchronously, and locally on each agent, requiring no communication. Therefore, many other approaches directly build upon this concept, e.g.,

[80][79][94][95][73][76].

Once any of them obtains a private or joint observation, agents need to exchange factorized cross-covariances and their beliefs to correlated ones. Then $\frac{N(N-1)}{2}$ measurement innovations and updated cross-covariances need to be processed and exchanged. After that, individual agents can compute their Kalman gain, update their beliefs, and need to factorize their cross-covariances again (locally). This allows to restore the cross-covariance between agents exactly and the calculations in update steps are distributed among all agents. A downside is that many communication links are required to obtain the needed information on each agent and that a generalization for joint updates relating to any number of agents is rather complex.

### Properties

Regarding memory consumption, each agent maintains its local belief and additionally needs to keep track of the factorized cross-covariances referring to other agents, e.g., in a dictionary that is access via the other agent's ID. Assuming a homogeneous swarm with $N$ agents and global full state with $M$ elements, in total $m + Nm$ (mean, covariance, and factorized cross-covariances) elements need to be stored per agent. In total, $M(M + 1)$ elements.

Regarding communication, $N + 2\frac{N(N-1)}{2}$ links are required once an exteroceptive measurement is obtained: $N$ links to request an information exchange, $\frac{N(N-1)}{2}$ links to exchange the latest beliefs and $N - 1$ cross-covariance factors across the agents (in the limit), and $\frac{N(N-1)}{2}$ messages to exchange updated cross-covariance factors and measurement innovations again.

Assuming a homogeneous swarm with the global full state of $M$ elements and $m = \frac{M}{N}$ individual states, $\frac{N(N-1)}{2}(m + m^2 + (N-1)m^2 + m^2 + r^2)$ elements need to be sent per update, with $r$ being the dimension of the measurement/observation.

As a consequence, a persistent all-to-all communication required, while processing exteroceptive measurements and the amount of exteroceptive ones increases typically with the group size.

### 3.4.4   Decentralized CSE with Distributed Propagation (DCSE-DP*)

DCSE-DP* is a *decentralized* flavor of DCSE-DP, as described in  Section 3.4.3, where an agent acts as interim master at the moment it receives a private or joint measurement. This agent requests all other agents' beliefs and cross-covariance factors. It behaves equal to the FC in CCSE-DP, as described in  Section 3.4.2, and performs the update exactly. After the update step was computed, it sends the corrected beliefs and factorized cross-covariance back to the other agents. Compared to DCSE-DP, this results in less communication links and less data to exchange, but increases the computational complexity. Please note, that measurements are processed sequentially, not parallel, meaning that other agents' processes are idle, while one acts as interim master, meaning that a task parallelization on measurement level can only be achieved for proprioceptive measurements. Consequently, processing a private or joint updates on an individual agent remains the critical path/bottleneck.

### Properties

Regarding memory consumption, each agent maintains its local belief and factorized cross-covariances referring to other agents, e.g., in a dictionary that is access via the other agents identifiers. Assuming a homogeneous swarm with $N$ agents, in total $m + Nm$ (mean, covariance, and factorized cross-covariances) elements need to be stored per agent. In total, $M(M + 1)$ elements.

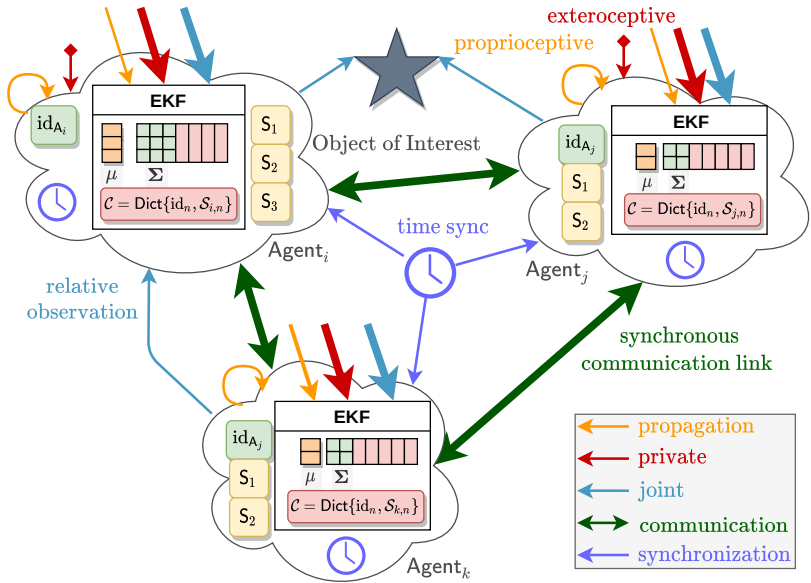

**Figure 3.5:** DCSE-DP: Each time-synchronized agent maintains a filter locally, that allows processing prediction steps (evolving the estimate in time) based on obtained proprioceptive sensor measurements locally. Each agent maintains a set of factorized cross-covariances $\mathcal{S}$, which are exchanged with each other, once any agent obtains an exteroceptive measurement, as the update steps are processed distributed.

Communication is just required to process update steps. Assuming synchronous exteroceptive measurements and a known interim master, $N - 1$ links to provide the measurement, the local belief, and cross-covariance terms to the interim master, which responds with the corrected beliefs and factorized cross-covariances. Thus, in total $2(N - 1)$ links are required.

Regarding message size, each agent has to transmit and receive the local belief (mean and covariance), the elements in the dictionary, and the measurement. Assuming a homogeneous swarm with $N$ agents, $N - 1$ agent needs to send $r + m + Nm$ elements to the interim master agent and receives $m + Nm$ corrected elements per exteroceptive measurement. Thus, in total $R + M + M^2$ elements need to be communicated per update step.

Regarding connectivity, any agents can act as interim master. During processing update steps, a star-graph based and persistent communication with the root at the interim master is needed.

### 3.4.5 DCSE based on Distributed Approximated Cross-Covariances (DCSE-DACC)

In Figure 3.7, an architecture derived from Section 3.4.3 supporting distributed propagation and distributed private exteroceptive sensor updates is shown. In addition to the assumption of decoupled input dynamics of systems, as in the SKF (see Section 2.8.3), similar assumption regarding decoupled output dynamics are made. This means that non-participating agents are assumed to be *nuisance* variables with respect to the participant's belief – which is considered as *essential* variables – and the full state is partitioned, as in the SKF, during the Kalman filter update step. The state prediction is equal to DCSE-DP. Note, that the proposed update step is not equal to the SKF update step (see Equation (2.72)), as the uncertainty of the nuisance parameters is not or just partially available.

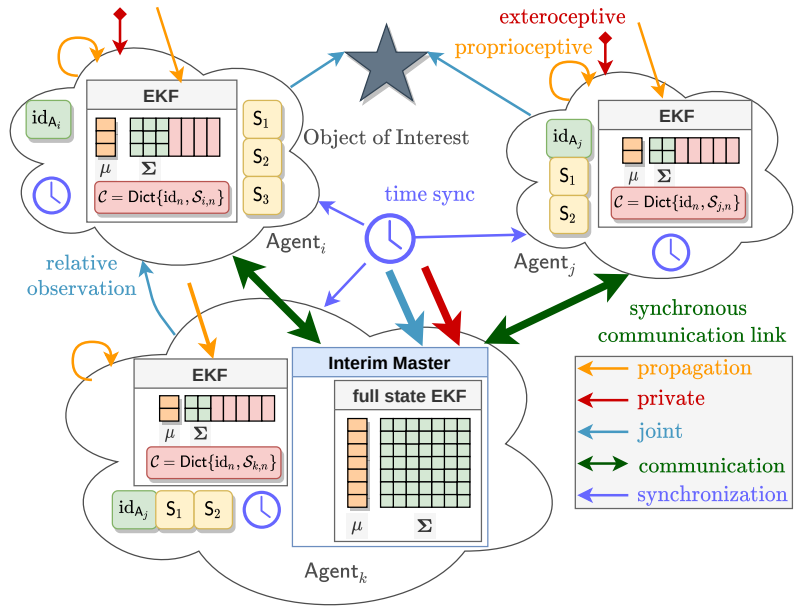

**Figure 3.6:** DCSE-DP*: Each time-synchronized agent maintains a filter locally, that allows processing prediction steps (evolving the estimate in time) based on obtained proprioceptive sensor measurements locally. Each agent maintains a set of factorized cross-covariances $\mathcal{S}$. Once a filter update steps needs to be performed, one agent acts as interim master. The other agents send their most recent beliefs and factorized cross-covariances to the interim master. After the update was processed on the full state, the interim master agent sends the corrected beliefs and factorized cross-covariances back to the other agents.

Therefore, the nuisance parameters' uncertainty and the interdependencies to the essential variables cannot be considered and corrected exactly, and approximations need to be applied. Due to approximations made regarding non-participating agents' cross-covariance factors, joint updates requires communication only among participating agents.

This approach was proposed by Luft *et al.* in [94, 95] and allows processing propagation and private update steps independently, while joint update steps require just *participating* agent. Participating agents are defined as those, who's beliefs are directly coupled to the systems' dynamics in the update step[9]. In private updates, only the ego-belief (or parts if it) are directly coupled with the ego-system's output. In contrast to fully distributed DCSE-DP approach, one agent of $P$ participants acts in DCSE-DACC as *interim master* to process the joint update step and sends the corrections back to the other participants. This means that the approach has among participant a temporary centralized processing topology.

### Properties

The memory consumption is equal to DCSE-DP as described in Section 3.4.3 and in total/globally seen $M(M+1)$ elements, which are distributed among agents.

Regarding communication, $3(P-1)$ links are required for each joint observation: $P-1$ links to request the data by the interim master from other participants, $P-1$ links to obtain the information, and $P-1$ links to send the corrected information back.

Assuming a homogeneous swarm with a global full state of $M$ elements, $m = \frac{M}{N}$

---

[9]More precisely, whose belief is needed in the linearization of the measurement Jacobian, meaning that the sparsity of the measurement Jacobian is exploited and assumes in the manner of the SKF [129], that non-participants are nuisance parameters that obtain no correction.

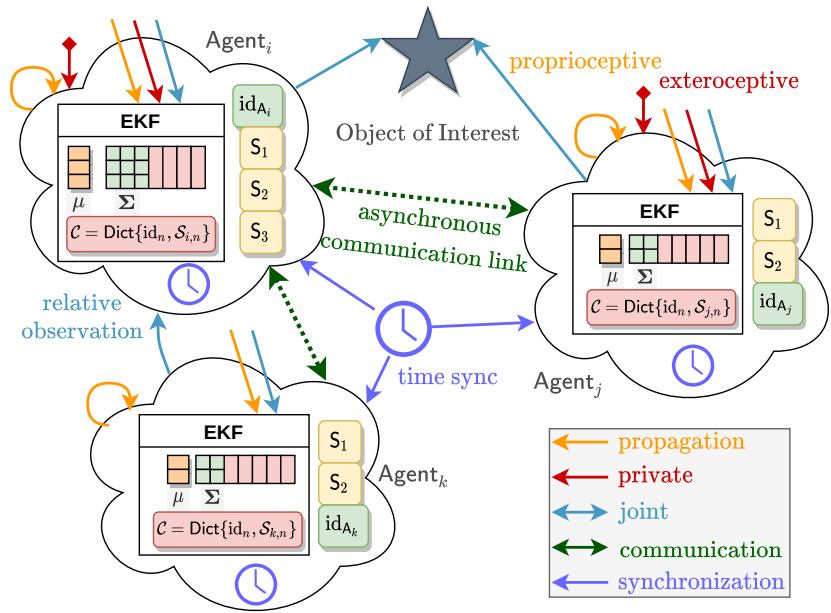

**Figure 3.7:** DCSE-DACC: Each time-synchronized agent maintains a filter locally that holds approximated factorized cross-covariance $\mathcal{S}$ between agents beliefs. This allows agents to process propagation and private filter steps isolated as the cross-covariance factors obtain correction in each filter step. Once agents perform joint observations, one agent acts as interim master. It obtains the other agents' beliefs and cross-covariance factors, processes the joint update step locally, and sends corrected states back to the participating agents. Correlated but non-participating agents' cross-covariances obtain an approximated correction, thus this approach is not *exact*.

individual state elements, and $P$ participants. In total $(P-1)+2(P-1)(m+m^2+(P-1)m^2)$ elements need to be sent per joint update step. Per participant an element for the request, the *a-priori* belief and cross-covariance factors relating to the other participants, and the *a-posteriori* belief and factors. Please note, that no measurement data need to be exchanged as the measurement data is received and processed by the interim master. Private observations are processed locally and independently.

Consequently, communication is required among participants, while processing joint measurements.

### 3.4.6  DCSE based on a Distributed Approximated History (DCSE-DAH)

In Figure 3.8, an architecture of the DCSE-DAH approach is shown and was covered in Section 3.3.5. Jung and Weiss identified in [76], high propagation sensor update rates a limitation of DCSE-DACC (see Section 3.4.5) regarding scalability with respect to the number of known/met agents.

That effect was mitigated by introducing a buffering scheme for correction terms ($\boldsymbol{\Phi}$, $\boldsymbol{\Upsilon}$, $\boldsymbol{\Lambda}$) of different filter steps (propagation, private, and joint filter steps). Consequently, the maintenance effort for factorized cross-covariances can be shifted to the moment they are actually needed – at the moment of joint observation.

In DCSE-DACC, the factorized cross-covariances obtain corrections in each filter step. The number of factorized cross-covariances relates to the (directly) correlated/*known* agents[10]. In DCSE-DAH, the factorized cross-covariances obtain corrections, once they are needed again, by applying an accumulated history of correction terms onto dated factors, which forward-propagates and corrects them. Assuming $C$ correlated agents and a history of $L$ elements, in the best case, $L - 1 + C$ matrix multiplications are needed to forward propagate all factors, while in DCSE-DACC, $L \times C$ matrix multiplications are needed. In case of fast propagation sensors with $L >> C$, a common correction buffer and applying an accumulated correction term on all factorized cross-covariances is beneficial.

---

[10]In contrast to *exact* filter formulations, correlations between agents can only be obtained via joint observations, while in *exact* formulations, correlations can be established indirectly via other agents cor-

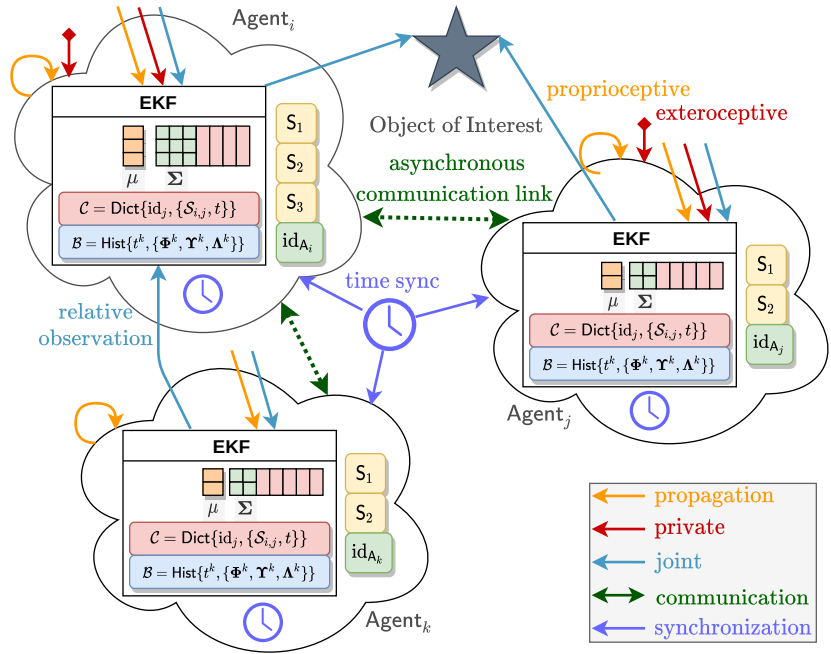

**Figure 3.8:** DCSE-DAH: Each time-synchronized agent maintains a filter locally that holds approximated factorized cross-covariance $\mathcal{S}$ between agents beliefs and a history of recent correction terms obtained in the different filter steps. This allows agents to process propagation and private filter steps *isolated*. In contrast to Figure 3.7, the cross-covariance factors are, in the best case, corrected only at the moment of joint observation – when they are needed again – and reduces the maintenance effort. Once agents perform joint observations, one agent acts as interim master, obtains the other agents' beliefs and corrected cross-covariance factors, processes the joint update step locally, and sends corrected states back to the participating agents. Correlated but non-participating agents' cross-covariances obtain an approximated correction, thus it is not *exact*.

### Properties

The memory consumption differs slightly from DCSE-DP and DCSE-DACC by requiring a buffer to keep track of recent correction terms, e.g., a fixed size sliding window buffer. The length $L$ of this buffer has implications on *when* and *where* the correction is applied and is further discussed in Section 3.3.5. Consequently, each agent maintains a buffer with $Lm^2$ elements, meaning that $m(1 + Nm + Lm)$ elements are maintained per agent. In total $Nm(1 + Nm + Lm)$ elements.

Regarding connectivity, message size, and communication links, the same constraints as in Section 3.4.5 apply.

---

relations, e.g., if agent $\mathsf{A}_1$ and $\mathsf{A}_2$ are correlated, a joint update between $\mathsf{A}_1$ and $\mathsf{A}_3$ would introduce correlations between $\mathsf{A}_3$ and $\mathsf{A}_2$.

### 3.4.7 Summary

In Table 3.1, we summarized the memory consumption, the communication, and the computational complexity for five fusion strategies, CCSE, CCSE-DP, DCSE-DP*, DCSE-DACC, and DCSE-DAH. Generally, the listing of architectures in this section is by far not complete, e.g., the interim master decentralized cooperative localization (IMDCL) algorithm proposed by Kia *et al.* in [79, 80], a flavor of DCSE-DP, is missing[11].

Table 3.1 does not unveil the advantage of DCSE-DAH over DCSE-DACC with respect to the maintenance effort as our scalability evaluation in manifest Section 3.5.2.

---

[11]This approach does not support private updates as it assumes cross-covariances are only changing at the event of joint updates.

| topology | steps | CCSE | CCSE-DP | DCSE-DP* | DCSE-DACC | DCSE-DAH |
|---|---|---|---|---|---|---|
| | | star | star | star | star | star |
| connectivity | prop | permanent | – | permanent | – | – |
| | priv | permanent | permanent | permanent | sporadic | sporadic |
| | joint | permanent | permanent | permanent | sporadic | sporadic |
| msg size | prop | $\{r, m(1+m)\}$ | – | – | – | – |
| | priv | $\{r, m(1+m)\}$ | $\{r+m(1+2m), m(1+m)\}$ | $\{r, m+Nm\}$ | – | – |
| | joint | $\{r, m(1+m)\}$ | $\{r+m(1+2m), m(1+m)\}$ | $\{r, m+Nm\}$ | $\{1, m(1+m)+(P-1)m^2\}$ | $\{1, m(1+m)+(P-1)m^2\}$ |
| msg total | prop | $R+M+Nm^2$ | 0 | 0 | 0 | 0 |
| | priv | $R+M+Nm^2$ | $R+Nm(2+3m)$ | $R+M+M^2$ | 0 | 0 |
| | joint | $R+M+Nm^2$ | $R+Nm(2+3m)$ | $R+M+M^2$ | $2(P-1)(m(1+m)+(P-1)m^2)$ | $2(P-1)(m(1+m)+(P-1)m^2)$ |
| links/step | prop | $2N$ | 0 | 0 | 0 | 0 |
| | priv | $2N$ | $2N$ | $2(N-1)$ | 0 | 0 |
| | joint | $2N$ | $2N$ | $2(N-1)$ | $3(P-1)$ | $3(P-1)$ |
| complexity | prop | $\mathcal{O}(n^3)$ | $\mathcal{O}(n^2)$ | $\mathcal{O}(n^2)$ | $\mathcal{O}(n^2)$ | $\mathcal{O}(n^2)$ |
| | priv | $\mathcal{O}(n^3)$ | $\mathcal{O}(n^3)$ | $\mathcal{O}(n^3)$ | $\mathcal{O}(n^3)$ | $\mathcal{O}(n^2)$ |
| | joint | $\mathcal{O}(n^3)$ | $\mathcal{O}(n^3)$ | $\mathcal{O}(n^3)$ | $\mathcal{O}(n^3)$ | $\mathcal{O}(n^3)$ |
| memory | per agent | – | $m(1+2m)$ | $m+Nm$ | $m+Nm$ | $m(1+M+Lm)$ |
| | FC | $M(1+M)$ | $M(1+M)$ | $M(1+M)$ | – | – |
| | total | $M(1+M)$ | $M(M+1)+Nm(1+2m)$ | $M(1+M)$ | $M(1+M)$ | $M\left(1+M+\frac{LM}{N}\right)$ |

**Table 3.1:** Communication, complexity, and memory characteristics of different filter strategies in a group of $N$ homogeneous agents with a global full state of $M$ elements, and $m = \frac{M}{N}$ being the state vector length of the individual agent's state. $r$ is the length of a single agent's measurement vector and $R = Nr$ is length of the global measurement vector. $P$ is the number of participants in joint updates. $L$ is the length of the correction buffer. The O-complexity of the different fusion strategies, is in the limit of the global full state vector length, number of agents, and number of participating agents. The different fusion strategies are discussed in Section 3.4.

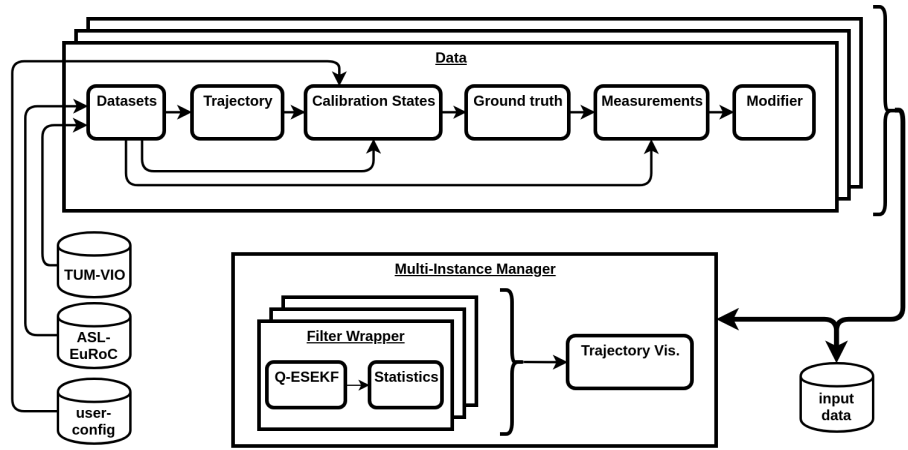

**Figure 3.9:** Shows the block diagram of our custom MATLAB simulation framework used to evaluate different DCSE algorithms.

## 3.5 Evaluation

A series of experiments in simulation were conducted, in order to validate the applicability of CSE using an ESEKF formulation that is based on an IMU as state propagation sensor. At this point, it is worth emphasizing that most evaluation are conducted using rather trivial dynamic systems – most commonly wheeled robots, e.g., in [79, 89, 95, 121]. In [95], Luft *et al.* compared and validated different approaches on the UTIAS dataset [90] for wheeled robots by Leung *et al.* .

The aim of this thesis is to go one step further and to apply and compare collaborative fusion strategies on a more advanced and versatile filter formulation based on an indirect EKF or ESEKF (see Section 2.8.6) using a strapped-down IMU as state propagation sensor, which became in recent years a standard approach to address various localization problems in robotics.

In the first experiment, we want to evaluate if the global information can be propagated through relative position measurements using different DCSE fusion strategies. In a scenario with two agents, only one is obtaining absolute position information, while the other agent is just participating in relative position measurements.

In the second experiment, we investigate on the scalability of the proposed DCSE-DAH approach on larger swarms of agents and compared it against other fusion strategies.

The experiments were conducted in a custom MATLAB simulation framework, that allows to load existing datasets or to generate trajectories, as shown in Figure 3.9. The exteroceptive measurements (private or joint observations) are generated based on the ground truth trajectory, the sensor calibration states, and noise parameters. The real-world IMU samples provided by the datasets are used directly (without modification) as measurement. Finally, all measurements are sorted chronologically and are locally processed in a multi-instance manager. It is maintaining multiple filter instances, while communication between filter instances is handled locally.

In the evaluation, we used DCSE-DP* instead of DCSE-DP, as this algorithm is easier to implement and easier to generalize for an arbitrary number of participants in joint observations. In our simulation framework, each agent has the same processing capabilities as a FC, as the evaluation is per design processed within one common process. Meaning that the measurement of the execution time of a fully distributed approach (without interim master) is per design in our simulation framework not possible[12].

---

[12] By the time of writing, launching of multiple threads in one process was not supported natively in

### 3.5.1 DCSE applied on Aided Inertial Navigation using relative position measurements

In this section, we investigate on a fundamental lemma of Mourikis and Roumeliotis [109], to show that only a single agent needs absolute information such that others are also navigating absolutely by receiving relative position measurements, which constitutes a form of *sensor sharing* or *sensor relaying*.

In CL with agents using an INS and joint relative observations among themselves, neither the absolute position nor the orientation about the gravity vector is observable. If a single agent has access to global position information, e.g., provided by a GNSS receiver, and by observing sufficient linear acceleration in two axes, all 6-DoF can be recovered [149]. Furthermore, if one recovers its absolute pose, relative position measurements and linear acceleration in two axes are sufficient for other robots to observe their absolute pose as well [121].

We evaluate the estimation performance using the Machine Hall (MH) sequences (MH_04 for $A_1$ and MH_05 for $A_2$) from the EuRoC dataset [20].

This evaluation was adapted from [76], and details on the filter formulation can be found in Section 2.9.

#### Problem Formulation

A team of $N$ distributed and communicating agents are equipped with an IMU and running an ESEKF (see Section 2.8.6). It provides them with drifting pose estimates, while exteroceptive observations are fused to correct them. Joint observations between agents are provided by a black box. For instance, the relative pose between agents can be obtained by recognizing a known visual tag on another agents.

In order to fuse joint observations in a distributed filter formulation, different DCSE algorithms are evaluated: DCSE-DP (see Section 3.4.2), DCSE-DACC (see Section 3.4.5), DCSE-DAH (see Section 3.4.6), and Distributed Discorrelated Minimum Variance (DDMV) proposed by Zhu and Kia in [161, 162].

The following simplifications are made:

- synchronized system clocks, e.g., by network based synchronization protocols,
- each agent has a unique identifier,
- the period of exteroceptive sensors is an integer multiple of the IMU period,
- extrinsic calibration between the IMU and other sensors is known and constant,
- communication range is larger than the sensing distance,
- and exchanged information between agents and sensor measurements arrive without delay, while proprioceptive ones are processed before exteroceptive ones (first prediction, then correction).

The state space of individual agents' filters comprises the IMU states (defined in Section 2.9) yielding a 15-DoF state vector

$$\mathbf{x}_{I_i} = \left[ {}^{\mathcal{G}}_{\mathcal{G}}\mathbf{p}_{\mathcal{I}}; {}^{\mathcal{G}}_{\mathcal{G}}\mathbf{v}_{\mathcal{I}}; {}^{\mathcal{G}}\mathbf{R}_{\mathcal{I}}; {}_{\mathcal{I}}\mathbf{b_a}; {}_{\mathcal{I}}\mathbf{b}_{\boldsymbol{\omega}} \right]_i, i \in 1, \dots, N, \tag{3.39}$$

with ${}^{\mathcal{G}}\mathbf{p}_{\mathcal{I}}, {}^{\mathcal{G}}_{\mathcal{G}}\mathbf{v}_{\mathcal{I}}$, and ${}^{\mathcal{G}}\mathbf{q}_{\mathcal{I}}$ as the position, velocity and orientation of the IMU $\mathcal{I}$ w.r.t. the global frame $\mathcal{G}$ (or navigation frame). ${}_{\mathcal{I}}\mathbf{b}_{\boldsymbol{\omega}}$ and ${}_{\mathcal{I}}\mathbf{b}_a$ are the estimated gyroscope and accelerometer biases to correct the related IMU readings. The body $\{\mathcal{B}\}$ and IMU $\{\mathcal{I}\}$ frame are aligned $\left( {}^{\mathcal{B}}\mathbf{T}_{\mathcal{I}} = \mathbf{I} \right)$. The error-state kinematic used for the IMU propagation and corrections through exteroceptive sensors is well studied and can be found in Section 2.9. The global full state is $\mathbf{x} = \begin{bmatrix} \mathbf{x}_{I_1} \\ \mathbf{x}_{I_2} \end{bmatrix}$. The corresponding error-state comprises 30-elements.

---

MATLAB.

The absolute position measurement, that considers the displacements between the sensors attached on the rigid body, is, as described in Equation (2.101), in the form

$$\mathbf{z}_{\mathrm{abs}}^{\#} = {}_{\mathcal{G}}^{\mathcal{G}}\mathbf{p}_{\mathcal{S}} + \mathbf{n}_{\mathrm{abs}} \tag{3.40}$$

where $\mathbf{n}_{\mathrm{abs}} \sim \mathcal{N}\left(\mathbf{0}, \mathbf{R}_{\mathrm{abs}}\right)$ is a white Gaussian noise vector with the covariance matrix $\mathbf{R}_{\mathrm{abs}}$ and with

$$
{}_{\mathcal{G}}^{\mathcal{G}}\mathbf{p}_{\mathcal{S}} = {}_{\mathcal{G}}^{\mathcal{G}}\mathbf{p}_{\mathcal{I}} + {}^{\mathcal{G}}\mathbf{R}_{\mathcal{I}} \left( {}^{\mathcal{B}}\mathbf{R}_{\mathcal{I}}^{\mathsf{T}} \left( -{}_{\mathcal{B}}^{\mathcal{B}}\mathbf{p}_{\mathcal{I}} + {}_{\mathcal{B}}^{\mathcal{B}}\mathbf{p}_{\mathcal{S}} \right) \right). \tag{3.41}
$$

We assume no displacement between the body $\{\mathcal{B}\}$ and position sensor $\{\mathcal{S}\}$ frame, such that observation can be simplified to ${}_{\mathcal{G}}^{\mathcal{G}}\mathbf{p}_{\mathcal{S}} = {}_{\mathcal{G}}^{\mathcal{G}}\mathbf{p}_{\mathcal{I}}$.

The local relative position measurement, that considers the displacements between senors attached on two rigid bodies, as described in Equation (2.115), is in the form

$$\mathbf{z}_{\mathrm{rel}}^{\#} = {}_{\mathcal{S}_1}^{\mathcal{S}_1}\mathbf{p}_{\mathcal{S}_2} + \mathbf{n} = \left( {}^{\mathcal{S}_1}\mathbf{T}_{\mathcal{S}_2} \right)_{\mathbf{p}} + \mathbf{n}_{\mathrm{rel}} \tag{3.42}$$

where $\mathbf{n}_{\mathrm{rel}} \sim \mathcal{N}\left(\mathbf{0}, \mathbf{R}_{\mathrm{rel}}\right)$ is a white Gaussian noise vector with the covariance matrix $\mathbf{R}_{\mathrm{rel}}$. We assume no displacement between the body $\{\mathcal{B}\}$ and sensor $\{\mathcal{S}_{\{1,2\}}\}$ frames, such that observation simplifies to

$$
\left( {}^{\mathcal{S}_1}\mathbf{T}_{\mathcal{S}_2} \right)_{\mathbf{p}} = {}^{\mathcal{G}}\mathbf{R}_{\mathcal{I}_1}^{\mathsf{T}} \left( -{}_{\mathcal{G}}^{\mathcal{G}}\mathbf{p}_{\mathcal{I}_1} + {}_{\mathcal{G}}^{\mathcal{G}}\mathbf{p}_{\mathcal{I}_2} \right). \tag{3.43}
$$

### Simulation

We evaluate this configuration on two Machine Hall (MH) sequences (MH_04 for $\mathsf{A}_1$ and MH_05 for $\mathsf{A}_2$) of the EuRoC dataset [20]. The trajectory of $\mathsf{A}_1$ starts at the origin of the global frame $\{\mathcal{G}\}$, while the trajectory of $\mathsf{A}_2$ obtained an offset of $\Delta\mathbf{p} = [5; 1; 0]$ from the origin.

The initial filter state of the individual agents is obtained by drawing a random sample centered at the ground truth value and given the initial uncertainty, according to Table 3.2. This demonstrates the self-calibration capabilities and allows analyzing the state convergence. Both agents obtain noisy IMU data at a rate of $200\,\mathrm{Hz}$, which is provided by the dataset. The noise characteristics are summarized in Table 3.3.

$\mathsf{A}_1$ is provided by synthetic absolute IMU position measurements at a rate of $10\,\mathrm{Hz}$ with an isotropic standard deviation of $\sigma_{\mathrm{abs}} = 0.1\,\mathrm{m}$, leading to a measurement covariance $\mathbf{R}_{\mathrm{abs}} = \mathbf{I}\sigma_{\mathrm{abs}}$. Synthetic relative IMU position measurements between $\mathsf{A}_1$ and $\mathsf{A}_2$ are performed from $t = 5.05\,\mathrm{s}$ on with an isotropic standard deviation of $\sigma_{\mathrm{pos}} = 0.1\,\mathrm{m}$, a measurement covariance of $\mathbf{R}_{\mathrm{rel}} = \mathbf{I}\sigma_{\mathrm{rel}}$, and at a rate of $10\,\mathrm{Hz}$. The measurement activity using DCSE-DAH of both agents is shown in Figure 3.11.

Figure 3.10 shows that agent $\mathsf{A}_2$ is drifting heavily, due to randomly initialized gyroscope and accelerometer biases ($_{\mathcal{I}}\mathbf{b}_\omega$ and $_{\mathcal{I}}\mathbf{b}_a$). Using DDMV agent $\mathsf{A}_2$ is diverging and $\mathsf{A}_1$ receives significantly wrong corrections from joint updates, that can fortunately be compensated by private ones. Thus, our implementation of DDMV is *not applicable* for our ESEKF formulation and not included in the following evaluation.

Table 3.4 lists the ARMSE and the mean of the NEES ($\overline{\mathrm{NEES}}$) over the entire trajectory (including the drift phases) of the estimated states for different CSE approaches. Therefore, the ARMSE values for $\mathsf{A}_2$ are significantly higher than for $\mathsf{A}_2$, although they achieve similar estimation results towards the end of the trajectory, see Figure 3.12. No remarkable differences between either fusion techniques are noticeable, while DCSE-DAH is the best scalable. DCSE-DP* has to be considered as ground-truth as it fuses the observations in an exact (centralized-equivalent) way. The NEES for all states should be on average 3; lower than that indicates conservatism, but all states are far from being considered inconsistent.

| | $^{\mathcal{G}}_{\mathcal{G}}\mathbf{p}_{\mathcal{I}}$ | $^{\mathcal{G}}_{\mathcal{G}}\mathbf{v}_{\mathcal{I}}$ | $^{\mathcal{G}}\mathbf{q}_{\mathcal{I}}$ | $_{\mathcal{I}}\mathbf{b}_{\boldsymbol{\omega}}$ | $_{\mathcal{I}}\mathbf{b}_{\mathbf{a}}$ |
|---|---|---|---|---|---|
| $\boldsymbol{\sigma}^0$ | 1 m | 1 m/s | 5 deg | 0.1 rad/s | 0.05 m/s$^2$ |

**Table 3.2:** Isotopic initial uncertainty for the IMU states. Problem is formulated in Section 3.5.1.

| $\sigma_{\mathbf{a}}$ | $\sigma_{\mathbf{b}_{\mathbf{a}}}$ | $\sigma_{\boldsymbol{\omega}}$ | $\sigma_{\mathbf{b}_{\boldsymbol{\omega}}}$ | $\sigma_{\text{abs}}$ | $\sigma_{\text{rel}}$ |
|---|---|---|---|---|---|
| 0.002 m/s$^2\sqrt{\text{Hz}}$ | 0.003 m/s$^3\sqrt{\text{Hz}}$ | $1.69e^{-4}$ rad/s$\sqrt{\text{Hz}}$ | $1.939e^{-5}$ rad/s$^2\sqrt{\text{Hz}}$ | 0.1 m | 0.1 m |

**Table 3.3:** Isotopic sensor noise characteristics of the accelerometer, gyroscope, absolute-, and relative position sensor. Problem is formulated in Section 3.5.1.

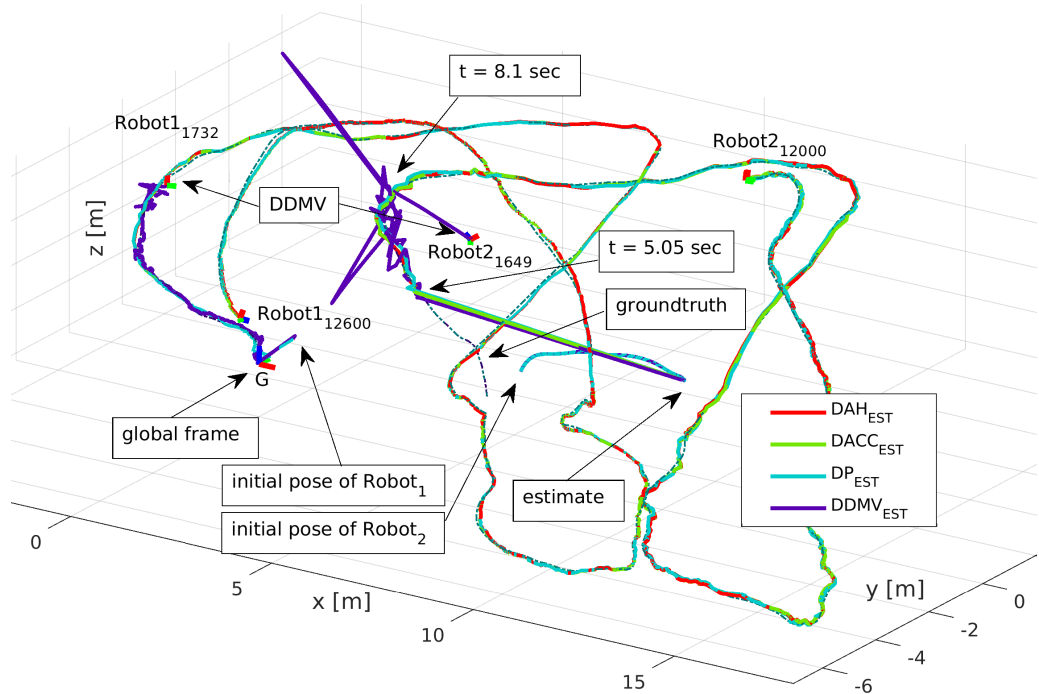

**Figure 3.10:** Estimated trajectory comparison using four different CSE fusion techniques: DCSE-DAH (red, proposed), DCSE-DACC (green), DCSE-DP* (cyan), and DDMV (purple). $\mathsf{A}_1$ receives global pose measurements, while $\mathsf{A}_2$ receives correction from joint relative position measurements with $\mathsf{A}_1$, starting from $t = 5.05$ s on. Due to wrongly initialized states, $\mathsf{A}_2$ is drifting heavily, but in most cases it can quickly converge towards ground truth, despite no direct global information being available. The agents fusing joint observation using DDMV receive strong state correction and $\mathsf{A}_2$ starts diverging. The problem is described in Section 3.5.1. Reused from [76].

The lower and upper 99.97 % bound are 0.05 and 13.93, respectively. For details regarding estimator credibility and the NEES please refer to Section 2.6.3.

In Figure 3.12, the estimated pose and errors of both agents using DCSE-DAH are shown. In the first 5.05 s, the estimated pose of $\mathsf{A}_2$ is heavily drifting. From the first relative position measurement on, the position error of $\mathsf{A}_2$ remains bounded and the uncertainty converged fast. The uncertainties of the orientation converge slower, but the error remains bounded and is not exceeding 7 °.

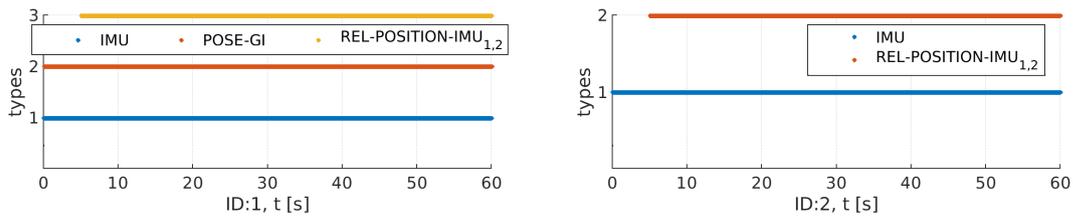

**Figure 3.11:** Shows the processed measurements on agent $A_1$ (top) and $A_2$ (bottom) using DCSE-DAH. At $t = 5.05$, joint relative position updates are performed between $A_1$ and $A_2$. Problem is formulated in Section 3.5.1.

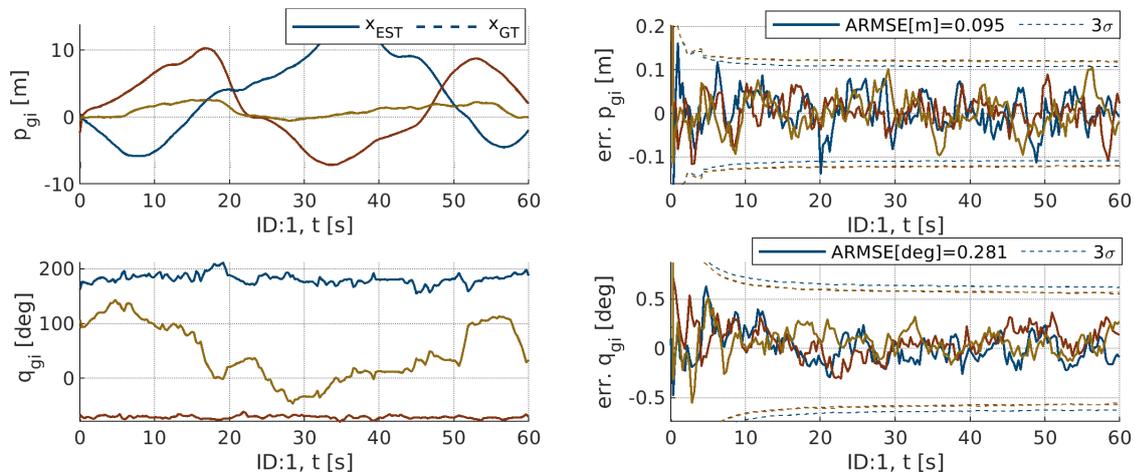

**(a)** Estimated and true pose, and estimation error of agent $A_1$.

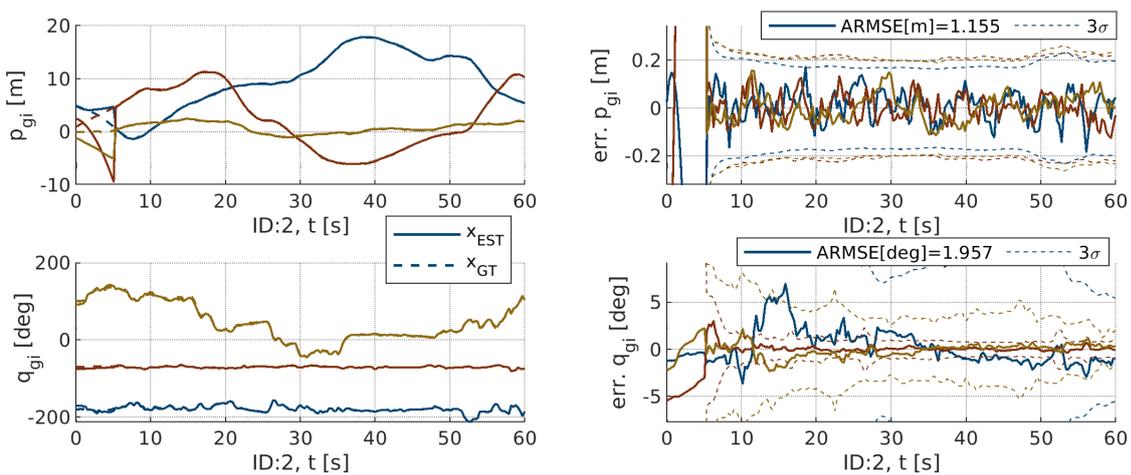

**(b)** Estimated and true pose, and estimation error of agent $A_2$.

**Figure 3.12:** The error of the estimated position (top row) and attitude (bottom row) in yellow, blue, red for position x, y, z respectively roll, pitch, yaw of agent $A_1$ (left) and $A_2$ (right)) using DCSE-DAH. At $t = 5.05\,\mathrm{s}$ the relative position measurements between the robots began and $A_2$ can restore all 6-DoF w.r.t. the absolute position. The estimation errors of both remain between the $3\sigma$ boundaries. Problem is formulated in Section 3.5.1. Adapted from [76].

| | | $^{\mathcal{G}}_{\mathcal{G}}\mathbf{p}_{\mathcal{I}}$ [m] | | $^{\mathcal{G}}_{\mathcal{G}}\mathbf{v}_{\mathcal{I}}$ [m/s] | | $^{\mathcal{G}}\mathbf{R}_{\mathcal{I}}$ [deg] | | $_{\mathcal{I}}\mathbf{b}_{\mathbf{a}}$ [m/s$^2$] | | $_{\mathcal{I}}\mathbf{b}_{\omega}$ [ rad/s] | |
|----|-----------|-------|-------|-------|-------|-------|------|--------|------|--------|--------|
| ID | CSE | ARMSE | $\overline{\text{NEES}}$ | ARMSE | $\overline{\text{NEES}}$ | ARMSE | $\overline{\text{NEES}}$ | ARMSE | $\overline{\text{NEES}}$ | ARMSE | $\overline{\text{NEES}}$ |
| 1 | DCSE-DP* | 0.065 | 3.1 | 0.086 | 2.86 | 0.264 | 1.56 | 0.0127 | 2.8 | 0.0007 | 0.33 |
| 1 | DCSE-DACC | 0.065 | 3.013 | 0.086 | 2.7 | 0.263 | 1.54 | 0.0134 | 3.38 | 0.0007 | 0.313 |
| 1 | DCSE-DAH | 0.065 | 3.017 | 0.086 | 2.7 | 0.263 | 1.54 | 0.0134 | 3.37 | 0.0007 | 0.314 |
| 2 | DCSE-DP* | 1.09 | 2.53 | 0.48 | 3.74 | 1.65 | 0.8 | 0.015 | 0.96 | 0.0024 | 1.25 |
| 2 | DCSE-DACC | 1.09 | 2.35 | 0.49 | 3.13 | 1.93 | 0.73 | 0.012 | 1.06 | 0.0025 | 0.92 |
| 2 | DCSE-DAH | 1.09 | 2.35 | 0.49 | 3.13 | 1.93 | 0.73 | 0.012 | 1.06 | 0.0025 | 0.92 |

**Table 3.4:** ARMSE and $\overline{\text{NEES}}$ values of the states estimated in the proposed ESEKF. Real-world IMU measurements from UAVs are used and described in Section 3.5.1, using different fusion approaches: DCSE-DP*, DCSE-DACC, and the proposed DCSE-DAH. Note: DCSE-DACC and DCSE-DAH should perform in this setup equivalently. Due to random noise on measurements a slight deviation is given. Adapted from [76].

### 3.5.2   Scalability of DCSE algorithms for Aided Inertial Navigation

The main motivation of the proposed DCSE-DAH approach is to reduce the maintenance effort for estimators with high propagation and update rates, as extension to DCSE-DACC. The following evaluation is adapted from our evaluation conducted in [76]. In simulation with 20 agents arranged in a circle, we compare different DCSE fusion techniques, as shown in Figure 3.13. Each performs a unique take-off (altitude of 20 m, circle (diameter of 10 m, with a height variation of $\pm 2.5$ m), and landing procedure. The duration of a trajectory is 60 s.

The problem formulation and assumptions made, are identical to those of the previous evaluation, see Section 3.5.1.

To stress the maintenance effort, each agent observes relative position measurements with three other agents (counter clock-wise on the formed circle), as shown in the measurement activity on agent $\mathsf{A}_1$ in  Figure 3.15. Thus, each agent know/met in total six other agents and all agents are directly correlated or indirectly correlated through others. A quarter of the agents are provided with noisy absolute IMU position measurements, meaning that 14 agents rely on IMU and relative IMU position measurements only. The absolute position update is received at a rate of 10 Hz, with a standard deviation of $\boldsymbol{\sigma}_{\text{abs}} = 0.3$ m, from $t = 0.1$ s on, and with a random message drop rate of 20 %. Local relative position observations are obtained at a rate of 10 Hz, with $\boldsymbol{\sigma}_{\text{rel}} = 0.1$ m, and a random message drop rate of 60 %. All receive unbiased and very noisy IMU measurements at 200 Hz with a standard deviation of $\boldsymbol{\sigma}_{\text{acc}} = 0.01$ m/s$^2$ and $\boldsymbol{\sigma}_{\text{gyr}} = 0.01$ rad/s for the accelerometer and gyroscope, respectively. All measurements are generated from the ground truth trajectory. All agent's states are initialized correctly with a reasonable uncertainty. The size of the correction buffer $\boldsymbol{\mathcal{B}}$ was set to 100 elements, which corresponds to a time horizon of 0.5 s.

Figure 3.16 and  Table 3.5 show that DCSE-DAH outperforms DCSE-DACC in terms of total execution time. Using DCSE-DAH, the execution time of the estimator was in total 7.87 s, while using DCSE-DACC, it was 11.26 s which is  43 % slower. It can also be seen, that the processing time of joint observations are in DCSE-DAH approximately 33 % higher as in DCSE-DACC. Concluding, DCSE-DAH keeps constant maintenance complexity for propagation and private filter steps, if the buffer size satisfies the period of joint observations. The estimation performance of DCSE-DAH seems to be a bit worse than DCSE-DACC, despite there should be no difference . Figure 3.14 shows a snapshot at $t = 8.5$ s of the simulation performing DCSE-DAH.

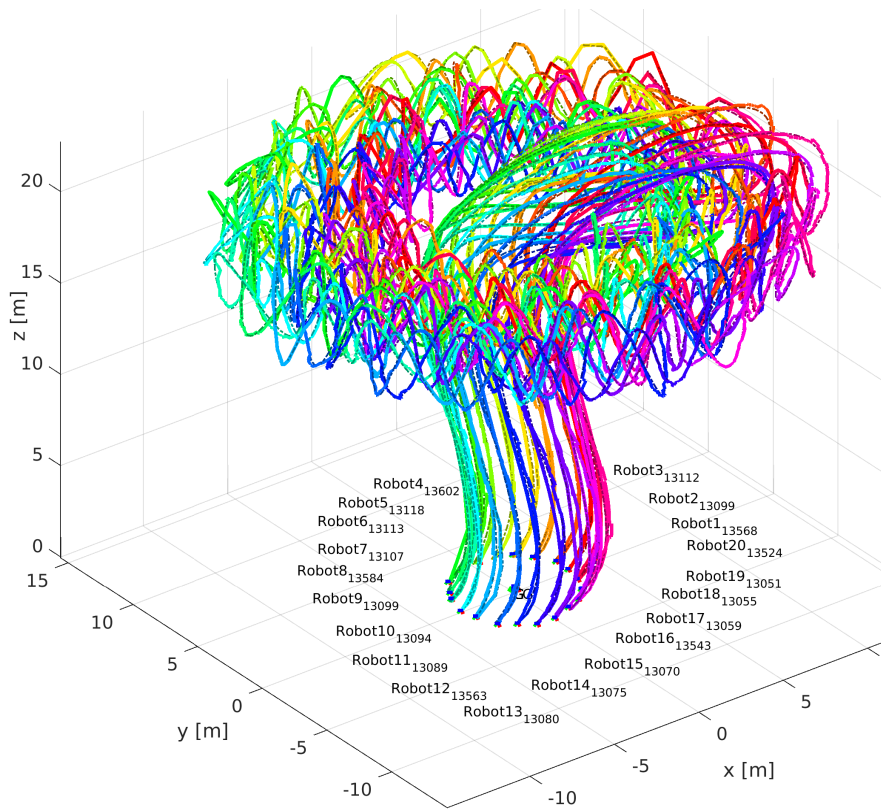

**Figure 3.13:** Shows the true and estimated trajectories using DCSE-DAH of 20 agents. Problem is formulated in Section 3.5.2.

| | $^{\mathcal{G}}_{\mathcal{G}}\mathbf{p}_{\mathcal{I}}$ [m] | | $^{\mathcal{G}}\mathbf{R}_{\mathcal{I}}$ [deg] | | [ms] | [ms] | [ms] |
|---|---|---|---|---|---|---|---|
| algorithm | AR | AN | AR | AN | $\bar{t}_{\text{prop}}$ | $\bar{t}_{\text{priv}}$ | $\bar{t}_{\text{joint}}$ |
| DCSE-DP* | **0.076** | **2.41** | **1.54** | 0.69 | **0.4** | 35 | 58 |
| DCSE-DACC | 0.105 | 5.19 | 1.91 | 1.04 | 0.83 | 0.28 | **1.4** |
| DCSE-DAH | 0.106 | 5.21 | 1.94 | **1.09** | 0.53 | **0.12** | 1.6 |

**Table 3.5:** Shows the average over 20 agents of the (i) ARMSE (AR), (ii) $\overline{\text{NEES}}$ (AN), (iii) average propagation time $\bar{t}_{\text{prop}}$, (iv) average private update time $\bar{t}_{\text{priv}}$, and (v) average joint update time $\bar{t}_{\text{joint}}$. The agents used an ESEKF formulation with different DCSE algorithms: DCSE-DP*, DCSE-DACC, and the proposed DCSE-DAH. Problem is formulated in Section 3.5.2. Best values in bold. Adapted from [76].

## 3.6 Conclusion

Driven by the need to reduce local maintenance effort using fast propagation sensors, we proposed the DCSE-DAH approach in Section 3.3.5, that performs the forward propagation of interdependencies at the moment they are needed.

Our experiments manifest that the proposed DCSE-DAH approach outperforms DCSE-DACC, while achieving the same filter performance in terms of accuracy and credibility. It can be seen as direct extension to DCSE-DACC that requires additionally a small amount of statically allocated storage per estimator. DCSE-DAH allows rendering CSE distributed with support for generic measurement and propagation models, communication is required only in case of joint observation. By choosing an appropriate buffer size, the maintenance effort in propagation and private update steps can be completely shifted to the moment of joint observations, making DCSE-DAH a fully scalable CSE approach with constant complexity both in communication and maintenance.

There are numerous interesting open topics, such as analyzing the bandwidth utilization and message delays of different fusion strategies in a real-world application using different communication technologies and protocols.

As future work, an extension for target-tracking (track-to-track fusion) while performing localization in a group of agents, denoted as CLATT, would be interesting. With respect to DCSE, the main challenge, apart from the unique data association, is to maintain multiple hypothesis of targets distributed across the agents with limited communication range, while considering correlation between the agents and the targets properly – which would be rather trivial in a centralized formulation/realization. Once in range, agents are able exchange their hypothesis and to find a consensus via, e.g., CI (see Section 2.8.4), to replace their local beliefs, and to account for the correlations to other (local) states.

Another important aspect is the missing support for delayed/out-of-order updates due to sensor, communication, and processing delay in the presented DCSE approaches, which is address in the final chapter Chapter 7.

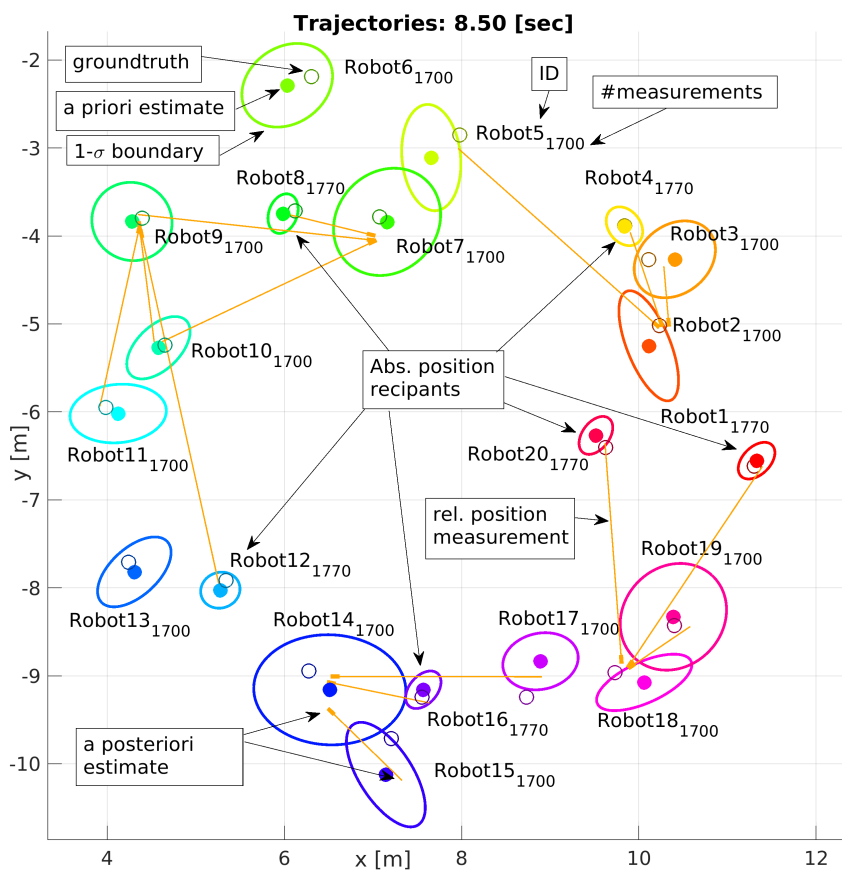

**Figure 3.14:** Top view on 20 agents' *a-priori* estimates, their $1\sigma$ uncertainty bound, as well as the ground truth position at $t = 8.5$ s. Some are performing joint observations (indicated by orange lines connecting the corrected poses) that compensate drifting states due to noisy IMU propagation. As only 6 agents obtain absolute position measurements, these joint observations between agents guarantee that their uncertainty and error remains bounded. The problem is described in Section 3.5.2. Image reused from [76].

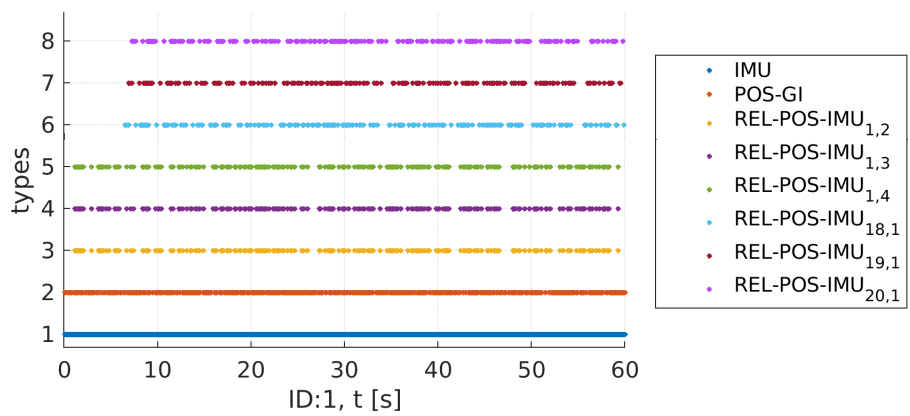

**Figure 3.15:** Measurement activity on $A_1$ using DCSE-DAH. The problem is described in Section 3.5.2.

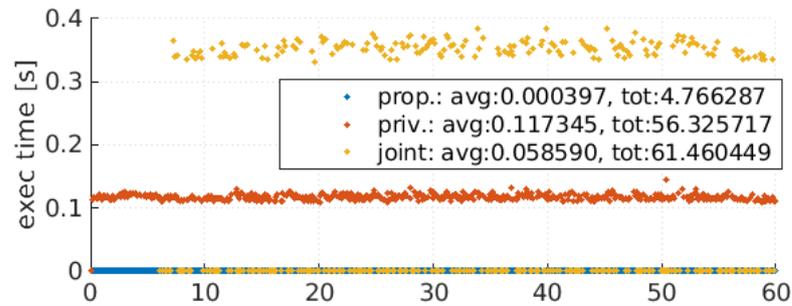

**(a)** Execution time on $A_{20}$ using DCSE-DP*.

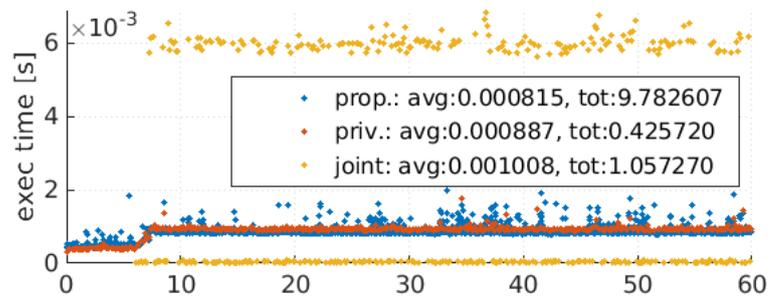

**(b)** Execution time on $A_{20}$ using DCSE-DACC.

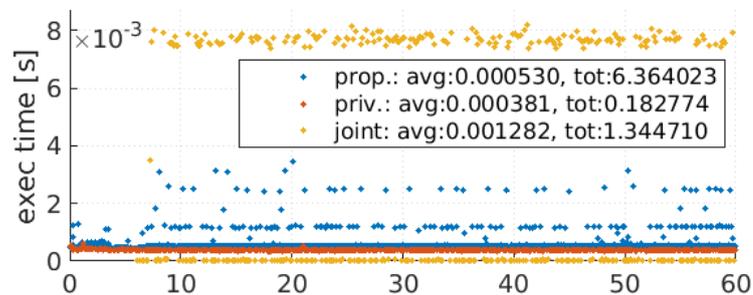

**(c)** Execution time on $A_{20}$ using DCSE-DAH.

**Figure 3.16:** Execution time plots of $A_{20}$, showing the execution time for each filter step: propagation (blue), private observation (red), and joint observation (yellow). The legends inform about the average and the accumulated time for each type. Note: As joint observations are performed on an interim master, the execution is rather short for the other participants and lowers the average of joint observations (e.g in Figure 3.16a). The increasing maintenance effort can be clearly seen for DACC in Figure 3.16b, which increases at $t \approx 10\,$s as joint observations began. Average timings are listed in Table 3.5. Further details are given in Section 3.5.2. Image reused from [76].

# Chapter 4

# Filter-based Modular Multi-Sensor Fusion

In this chapter, we are generalizing filter-based multi-sensor fusion from a DCSE perspective, which we published in [74]. The plug-and-play-like sensor handling of the proposed estimation framework builds upon the concept of performing local filter steps isolated. By bridging the gap between multi-agent collaborative state estimation and modular sensor fusion on a single agent, we propose a novel and generic algorithm for modular sensor fusion with constant maintenance complexity for propagation and private update steps. The proposed approach is evaluated on both, synthetic and real-world data and is compared against a native modular sensor fusion approach and two other approaches that have been ported from the DCSE domain.

## 4.1 Introduction

Modular, versatile, and robust state estimation at high-rates is a crucial component of autonomous robotic systems to control and navigate accurately in known, unknown, and dynamic environments.

Combining sensor information is denoted as sensor fusion, for which typically Bayes filters or optimization techniques are applied. Optimization-based methods are known to result in more accurate estimates at the cost of computation time, while filter-based approaches, due to their recursive nature, provide theoretically optimal estimates at each correction step (up to linearization and modeling errors).

Each additional sensor in a modular fusion framework may complement others or may increase the system's redundancy. However, it also introduces more (computational) complexity to the overall system, hindering many current approaches to scale well in a modular fashion for larger numbers of (different) sensor modalities. It is known that the computation time of a naive filter formulation increases cubically with the state vector size, and it scales typically linearly with amount of measurements, the update rate, and sensor delay.

From the perspective of combining a multitude of redundant or complementary elements in a scalable and modular fashion during mission, one can argue that this is exactly what distributed collaborative state estimation algorithms do in swarms of robots – with the difference that in the Modular Multi-Sensor Fusion (MMSF) aspect *swarm agents* are now considered *sensors of a single platform*. On the other hand, scalability and modularity, comes at the cost of computational complexity and is contradictory to the requirement of low latency for smooth and accurate control.

In view of the above, MMSF can be seen as problem of multi-agent CSE, aiming for decoupled and distributed estimation that reduces computation, communication, and maintenance cost [73].

As mention in the literature review (see Section 1.1.7), closely related modular sensor fusion approaches are, e.g., [18, 58, 96, 150].

In this chapter, we aim at generalizing multi-sensor fusion for Kalman filter formulations originating from a CSE perspective. Our approach presents a modular, yet scalable MMSF strategy, capable of processing not only measurements from exteroceptive sensors tightly-coupled for system-dynamics correction, but also of including isolated private and joint sensor observations for and between different exteroceptive sensors (see Section 4.2).

Nonetheless, it leads to suboptimal estimation results as it is based on the DCSE-DAH approach described in Section 3.4.6. It allows us to decouple estimation problem into individual nodes which are combined in our MMSF-DAH approach to a modular estimator with partially coupled outputs, but decoupled inputs. Depending on the application, MMSF-DAH can be configured according to the user's need and increases the flexibility and re-usability, as e.g., [18, 138]. The framework inherently supports plug and play, i.e. hot swapping, of sensors during mission, in case an initial belief and other sensor relevant information, such as sensor type, noise characteristics, unique identified, etc., are provided.

In our experiments on real data using an ESEKF for an AINS, we demonstrate that filter decoupling strategies used in CSE reduce the compute complexity also in MMSF formulations. The here proposed MMSF-DAH approach outperforms other fusion strategies we ported from DCSE to MMSF formulations regarding scalability and timing, while keeping up with the accuracy and credibility of centralized-equivalent and thus, statistically optimal architectures (Figure 4.3).

Our contributions are:

- Bridging the gap between CSE and MMSF: three existing DCSE approaches were implemented as modularized and decoupled MMSF frameworks.
- MMSF-DAH: A novel scalable and general multi-sensor fusion strategy with constant maintenance complexity for propagation and private update steps (if buffer history suffices).
- Comparison of the four different CSE fusion strategies and a classical MMSF approach using real-world data.

In the following, we briefly discuss related work, followed by a problem formulation in Section 4.2, and a description of our proposed MMSF-DAH approach in Section 4.3. In Section 4.4, we (i) evaluate the accuracy and consistency on real-world data, and (ii) the scalability with an increasing number of sensors among different fusion techniques. Conclusions are made in Section 4.5.

## 4.2  Problem Formulation

The problem formulation coincides with the one for the Isolated Kalman Filtering (IKF) paradigm in Section 6.2. In this chapter, we particularly address the problem handling and fusing multiple exteroceptive sensors in a modular (plug and play) fashion for an Aided Inertial Navigation System (AINS). In order to fuse the obtained sensor data properly, the spatial position or pose (extrinsic) with respect to the IMU or an common body reference frame $\{\mathcal{B}\}$ needs to be known. If the sensors are not synchronized via a hardware trigger or a software-based synchronization protocol, e.g., NTP [100], the time-offset between internal reference times or the sensor delays needs to be provided.

Please note that this approach is not limited to sensor related states only, e.g., the position of landmarks, visual feature, tracked objects can be modeled in individual filter instances, which can be added and removed online in a modular and lightweight fashion to the Aided Inertial Navigation System (AINS).

The support for plug-and-play lowers the technology barrier and if the system's sensor configuration allows a self-calibration of sensor intrinsic, extrinsic parameters or time offsets, the manufacturing process can be simplified, as discussed more thoroughly in MIMC-VINS by Eckenhoff *et al.* [36].

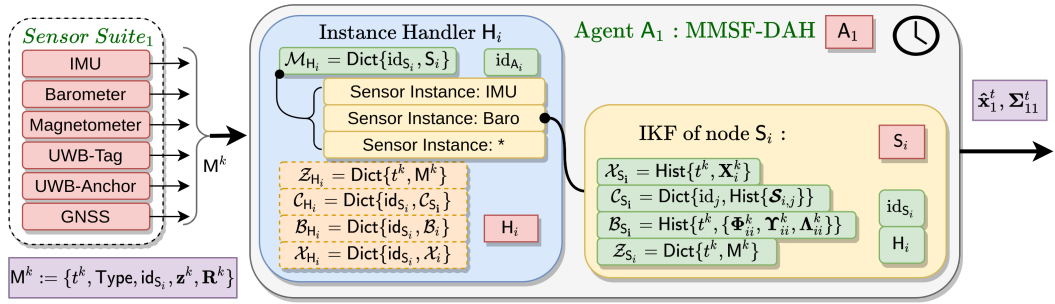

**Figure 4.1:** Block diagram of the proposed MMSF-DAH algorithm, consisting of an instance handler H maintaining various IKF instances (nodes) of a specific type. The sensor suite provides measurements to the H, which again delegates them to the appropriate node S, which performs the isolated filter steps. For joint observation, information exchange between IKF instances is needed, which is handled locally through the H. Details are provided in Section 4.2.

Either the sensor related parameters are assumed to be known *a-priori* , thus provided as constant values to the estimator, or if the parameters are observable, with an initial belief. Meaning, that each sensor has a certain parameter space, sensor specific characteristic (model), and a configuration (spatial relation, operation modes, etc.) that need to be accounted in the modular estimation framework.

Each sensor can be modeled as stochastic process (see Equation (6.1c)), in the simplest case with zero dynamics and no process noise. If the spatial configuration is not perfectly rigid or the initial calibration changes with temperature or humidity, one can model some process noise. A fundamental assumption of IKF is that the process noise of the individual sensors are not correlated, which is typically the case and needed for isolated sensor state propagation. For instance, the rate random walk of an IMU is independent of the temperature drift of a barometer, even if they are attached rigidly with respect to a common reference frame. This stochastic process can be estimated using the IKF paradigm in a sensor specific estimator deriving from the abstract IKF formulation, which is added as node to the modular Aided Inertial Navigation System (AINS).

Besides the modularity and abstraction provided by the IKF paradigm, it allows reducing the computational complexity over a modular centralized formulation. A potential downside regard convergence and the observability might arise, as discussed in Section 6.5.

## 4.3 Decoupled Approximated History based MMSF

In this section, the architecture of a local fusion entity, the *Instance Handler* H, to unify $N$ *locally* held IKF instances (nodes) is proposed, as shown in Figure 4.1. A set of $N$ heterogeneous sensors provide information, e.g., sensor measurements, to the handler H. Each sensor needs a unique identifier and belongs to a known class of sensor types, which defines the estimated parameter/state space, constant parameters, and sensors dynamic model, and a method to process sensor-specific measurement. Further, each sensor is associated to an IKF instance (node), maintaining a history of data as described in Section 6.3.1, a unique identified $\text{id}_S$ and a handle $H_A$ in order to access other IKF instances through the handler, in case of isolated joint observations.

The fusion entity H maintains handles to these IKFs instances, which can be added and removed at any point in time. Additionally, a sorted history of measurements $\mathcal{Z}_\mathsf{H}$ is maintained in the fusion entity to correct out-of-order measurements due to, e.g., latency. Technically, each IKF holds a history of measurements $\mathcal{Z}_\mathsf{S}$ itself, but as all instances are running in the same process and are managed by the fusion entity, it simplifies the handling of delayed measurements.

The measurement data $\mathsf{M}$ is provided to the handler $\mathsf{H}$ and needs to contain some meta information as defined in Equation (6.6). Once a measurement is received by the handler $\mathsf{H}$ (Algorithm 4.1), one can associate the corresponding estimator based on the identifier $\mathsf{id}_\mathsf{S}$ in the measurement data $\mathsf{M}$ and delegate it to the sensor-specific IKF node (Algorithm 4.3). Typically, measurements are received after the event of perception. Thus, a timestamp $t^k$ referring to that event must be provided in $\mathsf{M}$, the sensor delay needs to be known *a-priori* , or be estimated online.

The sensor node $\mathsf{S}$ is able to distinguish from different measurement types according to the meta information $\mathsf{Type}$ in $\mathsf{M}$. According to the field $\mathsf{Type}$, the sensor node $\mathsf{S}$ is performing either an isolated propagation (see Section 6.3.3), private (see Section 6.3.3), or joint update (see Section 6.3.3) step.

A *private observation* means that the observation/measurement relates just to the sensor's state. Meaning that it is *agnostic* (transparent) to other sensor states. For instance, a barometer sample is correcting a local pressure estimate. On the other hand, a *joint observation* means that estimates of various other sensors are rendered directly or indirectly observable by the measured quantity. For instance, the estimated temperature and pressure of reference station is used in a joint observation with a local pressure estimate to estimate the height based on the standard atmospheric model (see Section 2.9.4). Since isolated joint observations are related to at least one other IKF instance, access to these instances is realized by a handle to $\mathsf{H}_\mathsf{A}$ which allows access to all other IKF instances.

After a delayed measurement is processed and if it was not rejected from the NIS-based hypothesis check (Algorithm 4.7), all elements (beliefs, cross-covariance factors, correction terms) after the measurement event are deleted from all buffers of the IKF instances. Then, all interim measurements, held in $\mathcal{Z}_\mathsf{H}$, are re-applied in order (see Section 6.3.4). Finally, the new measurement $\mathsf{M}$ is inserted in the correct order into the measurement buffer $\mathcal{Z}_\mathsf{H}$ (Algorithm 4.1).

### 4.3.1 Algorithm

---

**Algorithm 4.1:** MMSF-DAH: handle sensor observation

**Input** : $\{\boldsymbol{\mathcal{X}}, \boldsymbol{\mathcal{C}}, \boldsymbol{\mathcal{B}}, \boldsymbol{\mathcal{M}}, \boldsymbol{\mathcal{Z}}\}_\mathsf{H}, \mathsf{M}$
1  $\{t^k, \mathsf{id}_{\mathsf{S}_i}, \mathbf{z}_{\mathsf{S}_i}^k, \mathbf{R}_{\mathsf{S}_i}^k, \mathsf{Type}\} = \mathsf{M}$
2  $\mathsf{S}_i = \boldsymbol{\mathcal{M}}_\mathsf{H}(\mathsf{id}_{\mathsf{S}_i})$
3  /* process measurement as sensor specific observation * /
4  /* e.g. propagation, private, joint (bi, tri, quad, ...) */
5  rejected = $\mathsf{S}_i \rightarrow$ process_sensor_measurement($\mathsf{M}$) (Alg. 4.3)
6  **if** *!rejected* **then**
7      redo_updates_after_t($\{\boldsymbol{\mathcal{X}}, \boldsymbol{\mathcal{C}}, \boldsymbol{\mathcal{B}}, \boldsymbol{\mathcal{M}}, \boldsymbol{\mathcal{Z}}\}_\mathsf{H}, t^k$) (Alg. 4.2)
8  **end**
9  $\boldsymbol{\mathcal{Z}}(t^k) = \mathsf{M}$   // store sensor measurement

---

---

**Algorithm 4.2:** MMSF-DAH: redo_updates_after_t

**Input** : $\{\boldsymbol{\mathcal{X}}, \boldsymbol{\mathcal{C}}, \boldsymbol{\mathcal{B}}, \boldsymbol{\mathcal{M}}, \boldsymbol{\mathcal{Z}}\}_\mathsf{H}, t^k$

1  /* delete newer corrections, cross-cov and states! */
2  delete($\{\boldsymbol{\mathcal{X}}_{\mathsf{H}i}, \boldsymbol{\mathcal{C}}_{\mathsf{H}i}, \boldsymbol{\mathcal{B}}_{\mathsf{H}i}\} > t^k$)
3  /* redo existing observation after $t^k$! */
4  **for** $\mathsf{M}_i$ **in** $\{sort(\boldsymbol{\mathcal{Z}}_\mathsf{H}) > t^k\}$ **do**
5  $\quad$ $\{t^i, \mathsf{id}_{\mathsf{S}_i}, \mathbf{z}^i_{\mathsf{S}_i}, \mathbf{R}^i_{\mathsf{S}_i}, \mathsf{Type}\} = \mathsf{M}_i$
6  $\quad$ $\mathsf{S}_l = \boldsymbol{\mathcal{M}}_\mathsf{H}(\mathsf{id}_{\mathsf{S}_i})$
7  $\quad$ rejected$_i = \mathsf{S}_l \to$ process_sensor_measurement($\mathsf{M}_i$) (Alg. 4.3)
8  **end**

---

**Algorithm 4.3:** MMSF-DAH: process sensor measurement

**Input** : $\boldsymbol{\mathcal{X}}, \boldsymbol{\mathcal{C}}, \boldsymbol{\mathcal{B}}, \mathsf{id}, \boldsymbol{\mathcal{Z}}, t^k, \mathbf{z}^k_\mathsf{S}, \mathbf{R}^k_\mathsf{S}$

1  rejected = True
2  **if** *propagation* **then**
3  $\quad$ propagate($\dots$) (Alg. 4.8)
4  **else if** *private observation* **then**
5  $\quad$ rejected = private_observation($\dots$) (Alg. 4.9)
6  **else**
7  $\quad$ rejected = joint_observation($\dots$) (Alg. 4.10)
8  **end**

---

**Algorithm 4.4:** MMSF-DAH: get_belief

**Input** : $\boldsymbol{\mathcal{X}}_i, \boldsymbol{\mathcal{B}}_i, \mathsf{id}_j, t^k$

1  **if** *not exist*($\boldsymbol{\mathcal{X}}_i(t^k)$) **then**
2  $\quad$ $\{\hat{\mathbf{x}}^a_i, \boldsymbol{\Sigma}^a_{ii}, t^a\} = \max(\mathrm{find}(\boldsymbol{\mathcal{X}}_i < t^k))$
3  $\quad$ /* predict from previous state to current timestamp */
4  $\quad$ $\hat{\mathbf{x}}^k_i, \boldsymbol{\Sigma}^k_{ii}, \boldsymbol{\Phi}^{k|a} = \mathrm{propagate}_i(\hat{\mathbf{x}}^a_i, \boldsymbol{\Sigma}^a_{ii}, \dots, t^a, t^k)$ (Alg. 4.8)
5  $\quad$ $\boldsymbol{\mathcal{B}}_i(t^k) = \boldsymbol{\Phi}^{k|a}$
6  **else**
7  $\quad$ $\{\hat{\mathbf{x}}^k_i, \boldsymbol{\Sigma}^k_{ii}\} = \boldsymbol{\mathcal{X}}_i(t^k)$
8  **end**

---

**Algorithm 4.5:** MMSF-DAH:check_horizon

**Input** : $\boldsymbol{\mathcal{B}}, \boldsymbol{\mathcal{C}}, t^k$

1  $t^o = \min(\mathrm{find}(\boldsymbol{\mathcal{B}} < t^k))$ (oldest correction term) (Equation (6.10))
2  $t^m = t^o + (t^k - t^o)/2$ (half of the time horizon Equation (6.11))
3  $\mathbf{M}^{m|o} = $ compute_corr($\boldsymbol{\mathcal{B}}, t^m, t^o$) (Alg. 4.6)
4  **for** $\{\mathsf{id}\}$ **in** $\boldsymbol{\mathcal{C}}$ **do**
5  $\quad$ **if** *find*($\boldsymbol{\mathcal{C}}(\mathsf{id})$) $\equiv t^o$ *(Equation (6.12))* **then**
6  $\quad\quad$ $\boldsymbol{\mathcal{C}}(\mathsf{id}) = \{\mathbf{M}^{m|o}\boldsymbol{\mathcal{S}}^{o(-)}, t^k\}$ (forward prop. Equation (6.13))
7  $\quad$ **end**
8  **end**

---

**Algorithm 4.6:** MMSF-DAH: compute_correction

**Input** : $\boldsymbol{\mathcal{B}}_i, t^a, t^k$

1  $\mathbf{M}^{k|a}_i = \mathbf{I}$
2  $t^b = \min(\boldsymbol{\mathcal{B}}_i > t^a)($
3  **for** $l \leftarrow t^b$ **to** $t^k$ **do**
4  $\quad$ $\mathbf{M}^{k|a}_i = \boldsymbol{\mathcal{B}}_i(l)\mathbf{M}^{k|a}_i$ (Equation (6.8))
5  **end**

---

**Algorithm 4.7:** MMSF-DAH: check_NIS

---

**Input** : $\mathbf{r}, \mathbf{S}, p = 0.997$

**1** $s = \mathbf{r}^\mathsf{T} \mathbf{S}^{-1} \mathbf{r}$ // Mahalanobis distance squared

**2** DoF = length($\mathbf{r}$) (Degrees of freedom)

**3** /* Inverse of the chi-square cumulative distribution */

**4** outlier = ($s > \text{chi2inv}(p, \text{DoF})$) (see Section 2.6.2)

---

**Algorithm 4.8:** MMSF-DAH: Isolated Propagation on $\mathsf{S}_p$

---

**Input** : $\hat{\mathbf{x}}_p^{k-1}, \mathbf{\Sigma}_{pp}^{k-1}, \mathcal{B}_p, \mathbf{u}^k, \mathbf{N}^k, \mathcal{C}_p, t^{k-1}, t^k$

**1** $\mathbf{\Phi}_p^{k|k-1} = \left[ \frac{\partial \phi_p(\mathbf{x}_p, \mathbf{u})}{\partial \mathbf{x}_p}(\hat{\mathbf{x}}_p, \mathbf{u}) \right]^{k|k-1}$

**2** $\mathbf{G}_p^{k|k-1} = \left[ \frac{\partial \phi_p(\mathbf{x}_p, \mathbf{u})}{\partial \mathbf{u}}(\hat{\mathbf{x}}_p, \mathbf{u}) \right]^{k|k-1}$

**3** $\mathbf{Q}^{k|k-1} = \mathbf{G}_p^{k|k-1} \mathbf{N}^k (\mathbf{G}_p^{k|k-1})^\mathsf{T}$

**4** $\hat{\mathbf{x}}_p^k = \phi_p(\hat{\mathbf{x}}_p^{k-1}, \mathbf{u}^k)$

**5** $\mathbf{\Sigma}_{pp}^k = \mathbf{\Phi}_p^{k|k-1} \mathbf{\Sigma}_{pp}^{k-1} (\mathbf{\Phi}_p^{k|k-1})^\mathsf{T} + \mathbf{Q}^{k|k-1}$

**6** $\mathcal{B}_p(t^k) = \mathbf{\Phi}^{k|k-1}$ // insert into sorted buffer

**7** check_horizon($\mathcal{B}_p, \mathcal{C}_p, t^k$) (Alg. 4.5)

---

**Algorithm 4.9:** MMSF-DAH: Isolated Private Observation on $\mathsf{S}_p$

---

**Input** : $\hat{\mathbf{x}}_p^{k(-)}, \mathbf{\Sigma}_{pp}^{k(-)}, \mathcal{B}_p, \mathcal{Z}_p, \mathbf{z}_p^k, \mathbf{R}_p^k$

**1** $\mathbf{H}_p = \left[ \frac{\partial h(\mathbf{x}_p)}{\partial \mathbf{x}_p}(\hat{\mathbf{x}}_p) \right]^{k(-)}$

**2** $\mathbf{S}_p = \mathbf{H}_p \mathbf{\Sigma}_{pp}^{k(-)} \mathbf{H}_p^\mathsf{T} + \mathbf{R}^k$

**3** $\mathbf{K}_p = \mathbf{\Sigma}_{pp}^{k(-)} \mathbf{H}_p^\mathsf{T} (\mathbf{S}_p)^{-1}$

**4** $\mathbf{r} = \mathbf{z}_p^k \boxminus h_p(\hat{\mathbf{x}}_p^{k(-)})$

**5** rejected = check_NIS($\mathbf{r}, \mathbf{S}_p$) (Alg. 4.7)

**6** **if** *not rejected* **then**

**7** $\quad$ $\hat{\mathbf{x}}_p^{k(+)} = \hat{\mathbf{x}}_p^{k(-)} \boxplus \mathbf{K}_p \mathbf{r}$

**8** $\quad$ $\mathbf{\Sigma}_{pp}^{k(+)} = (\mathbf{I} - \mathbf{K}_p \mathbf{H}_p) \mathbf{\Sigma}_{pp}^{k(-)}$

**9** $\quad$ $\mathbf{\Upsilon}_p^k = (\mathbf{I} - \mathbf{K}_p \mathbf{H}_p)$

**10** $\quad$ $\mathcal{B}_p(t^k) = \mathbf{\Upsilon}_p^k \mathcal{B}_p(t^k)$

**11** $\quad$ $\mathcal{X}_p(t^k) = \{\hat{\mathbf{x}}_p^{k(+)}, \mathbf{\Sigma}_p^{k(+)}\}$

**12** **end**

---

**Algorithm 4.10:** MMSF-DAH: Isolate Joint Observation of sensors $\mathsf{S}$ and $\mathsf{C}$

---

**Input** : $\{\boldsymbol{\mathcal{X}}, \boldsymbol{\mathcal{C}}, \boldsymbol{\mathcal{B}}, \mathsf{id}\}_{\{\mathsf{C},\mathsf{S}\}}, t^k, \mathbf{z}_\mathsf{S}^k, \mathbf{R}_\mathsf{S}^k$

1  /* get previous cross-covariance factors */
2  $\{\mathcal{S}_{\mathsf{SC}}^m, t^m\} = \max(\mathrm{find}(\boldsymbol{\mathcal{C}}_\mathsf{S}(\mathsf{id}_\mathsf{C}) < t^k))$
3  $\{\mathcal{S}_{\mathsf{CS}}^m, t^m\} = \max(\mathrm{find}(\boldsymbol{\mathcal{C}}_\mathsf{C}(\mathsf{id}_\mathsf{S}) < t^k))$
4  /* get existing beliefs or predict beliefs */
5  $\mathbf{x}_{\{\mathsf{C},\mathsf{S}\}}^m = \mathrm{get\_belief}(\boldsymbol{\mathcal{X}}_{\{\mathsf{C},\mathsf{S}\}}, \boldsymbol{\mathcal{B}}_{\{\mathsf{C},\mathsf{S}\}}, t^m)$ (Alg. 4.4)
6  $\mathbf{M}_{\{\mathsf{C},\mathsf{S}\}}^{k|m} = \mathrm{compute\_correction}(\boldsymbol{\mathcal{B}}_{\{\mathsf{C},\mathsf{S}\}}, t^m, t^k)$ (Alg. 4.6)
7  /* propagate previous cross-covariances */
8  $\boldsymbol{\Sigma}_{\mathsf{CS}}^k = (\mathbf{M}_\mathsf{C}^{k|m} \mathcal{S}_{\mathsf{CS}}^m)(\mathbf{M}_\mathsf{S}^{k|m} \mathcal{S}_{\mathsf{SC}}^m)^\mathsf{T}$
9  /* stack beliefs*/
10  $\boldsymbol{\Sigma}_{pp}^{k(-)} = \begin{bmatrix} \boldsymbol{\Sigma}_\mathsf{C} & \boldsymbol{\Sigma}_\mathsf{CS} \\ \boldsymbol{\Sigma}_\mathsf{CS}^\mathsf{T} & \boldsymbol{\Sigma}_\mathsf{S} \end{bmatrix}^k$
11  $\hat{\mathbf{x}}_p^{k(-)} = \begin{bmatrix} \hat{\mathbf{x}}_\mathsf{C} \\ \hat{\mathbf{x}}_\mathsf{S} \end{bmatrix}^k$
12  $\mathbf{H}_p = \begin{bmatrix} \frac{\partial h_\mathsf{S}(\mathbf{x}_\mathsf{C}, \mathbf{x}_\mathsf{S})}{\partial \mathbf{x}_\mathsf{C}}(\hat{\mathbf{x}}_\mathsf{C}, \hat{\mathbf{x}}_\mathsf{S}) & \frac{\partial h_\mathsf{S}(\mathbf{x}_\mathsf{C}, \mathbf{x}_\mathsf{S})}{\partial \mathbf{x}_\mathsf{S}}(\hat{\mathbf{x}}_\mathsf{C}, \hat{\mathbf{x}}_\mathsf{S}) \end{bmatrix}^{k(-)}$
13  $\mathbf{S}_p = \mathbf{H}_p \boldsymbol{\Sigma}_{pp}^{k(-)} \mathbf{H}_p^\mathsf{T} + \mathbf{R}^k$
14  $\mathbf{K}_p = \boldsymbol{\Sigma}_{pp}^{k(-)} \mathbf{H}_p^\mathsf{T} (\mathbf{S}_p)^{-1}$
15  $\mathbf{r} = \mathbf{z}_\mathsf{S}^k \boxminus h_\mathsf{S}(\hat{\mathbf{x}}_p^{k(-)})$
16  rejected = check_NIS$(\mathbf{r}, \mathbf{S}_p)$ (Alg. 4.7)
17  **if** *not rejected* **then**
18  $\quad \hat{\mathbf{x}}_p^{k(+)} = \hat{\mathbf{x}}_p^{k(-)} \boxplus \mathbf{K}_p \mathbf{r}$
19  $\quad \boldsymbol{\Sigma}_{pp}^{k(+)} = (\mathbf{I} - \mathbf{K}_p \mathbf{H}_p) \boldsymbol{\Sigma}_{pp}^{k(-)}$
20  $\quad$ /* Note: split $\boldsymbol{\Sigma}_{pp}^{k(+)}$ and $\hat{\mathbf{x}}_p^{k(+)}$ again */
21  $\quad \mathcal{S}_{\mathsf{CS}}^{k(+)} = \boldsymbol{\Sigma}_{\mathsf{CS}}^{k(+)}$
22  $\quad \mathcal{S}_{\mathsf{SC}}^{k(+)} = \mathbf{I}$
23  $\quad \boldsymbol{\Lambda}_{\{\mathsf{C},\mathsf{S}\}}^k = \boldsymbol{\Sigma}_{\{\mathsf{C},\mathsf{S}\}}^{k(+)} (\boldsymbol{\Sigma}_{\{\mathsf{C},\mathsf{S}\}}^{k(-)})^{-1}$
24  $\quad \boldsymbol{\mathcal{B}}_{\{\mathsf{C},\mathsf{S}\}}\left(t^k\right) = \boldsymbol{\Lambda}_{\{\mathsf{C},\mathsf{S}\}}^k \boldsymbol{\mathcal{B}}_{\{\mathsf{C},\mathsf{S}\}}\left(t^k\right)$
25  $\quad \boldsymbol{\mathcal{C}}_{\{\mathsf{C},\mathsf{S}\}}(\mathsf{id}_{\{\mathsf{S},\mathsf{C}\}}) = \{\mathcal{S}_{\{\mathsf{CS},\mathsf{SC}\}}^{k(+)}, t^k\}$
26  $\quad \boldsymbol{\mathcal{X}}_{\{\mathsf{C},\mathsf{S}\}}(t^k) = \{\hat{\mathbf{x}}_{\{\mathsf{C},\mathsf{S}\}}^{k(+)}, \boldsymbol{\Sigma}_{\{\mathsf{C},\mathsf{S}\}}^{k(+)}\}$
27  **end**

## 4.4   Evaluation

The experiments are done in our custom MATLAB framework, that allows to load real data from the EuRoC dataset [20] by Burri *et al.* , as shown in Figure 4.2. Exteroceptive measurements (private or joint observations) are generated based on the ground truth trajectory modified by the sensors' calibration states and noise parameters. The noisy and biased real-world IMU samples provided by the datasets are used without modifications. Finally, all measurements are processed in the instance handler H. It is maintaining multiple IKF instances/nodes S, while communication between instances is handled locally. The filter instances and handler support different fusion strategies as listed in Section 4.4.1.

For the scalability and accuracy evaluation, we add different exteroceptive sensors to a modular ESEKF using an IMU as propagation sensor. The sensor suite's state vector might consist of a varying number of (i) absolute position sensors, e.g., a GNSS sensor, (ii) a barometers, (iii) an atmospheric sensors (ATM):

$$\mathbf{x}_{\mathcal{I}} = \left[ {}^{\mathcal{G}}_{\mathcal{G}}\mathbf{p}_{\mathcal{I}}, {}^{\mathcal{G}}_{\mathcal{G}}\mathbf{v}_{\mathcal{I}}, {}^{\mathcal{G}}\mathbf{q}_{\mathcal{I}}, {}_{\mathcal{I}}\mathbf{b}_{\omega}, {}_{\mathcal{I}}\mathbf{b}_a \right] \tag{4.1}$$

$$\mathbf{x}_{\{\mathcal{ABS},\mathcal{BARO}\}} = \begin{bmatrix} {}^{\mathcal{B}}_{\mathcal{B}}\mathbf{p}_{\mathcal{S}} \end{bmatrix} \tag{4.2}$$

$$\mathbf{x}_{\mathcal{ATM}} = [P_{atm}, T_{atm}] \tag{4.3}$$

with ${}^{\mathcal{G}}_{\mathcal{G}}\mathbf{p}_{\mathcal{I}}, {}^{\mathcal{G}}_{\mathcal{G}}\mathbf{v}_{\mathcal{I}}$, and ${}^{\mathcal{G}}\mathbf{q}_{\mathcal{I}}$ as the position, velocity and orientation of the IMU $\mathcal{I}$ referring to the global frame $\mathcal{G}$. ${}_{\mathcal{I}}\mathbf{b}_{\omega}$ and ${}_{\mathcal{I}}\mathbf{b}_a$ are the estimated gyroscope and accelerometer biases to correct the related IMU readings. ${}^{\mathcal{B}}_{\mathcal{B}}\mathbf{p}_{\mathcal{I}}$ and ${}^{\mathcal{B}}\mathbf{q}_{\mathcal{I}}$ as constant position and orientation of the IMU $\mathcal{I}$ referring to the body frame $\mathcal{B}$. Sensors rigidly attached with respect to the body reference frame, require a spatial calibration state ${}^{\mathcal{B}}_{\mathcal{B}}\mathbf{p}_{\mathcal{S}}$, specifying the translation to the sensor frame $\mathcal{S}$. The static atmospheric reference sensor estimates the local temperature $T_{atm}$ in [°C] and pressure $P_{atm}$ in [Pa] as private observations. Barometer readings in [Pa] are processed as joint observations incorporating the IMU and ATM ($\mathbf{x}_{\{\mathcal{I},\mathcal{ATM},\mathcal{BARO}\}}$). The error-state kinematics for the IMU propagation and corrections through exteroceptive sensors is covered in Section 2.9. The evaluation is performed single-threaded in MATLAB on an AMD Ryzen 7 3700X CPU with a 32 GB DDR4 RAM.

### 4.4.1   Strategy Overview

We ported different CSE strategies to a MMSF formulation with support for delayed measurements: Centralized equivalent (MMSF-C), decoupled propagation (MMSF-DP), decoupled approximated cross-covariance (MMSF-DACC) and tested them as well as a state-of-the-art native modular approach (MMSF-MaRS) against our here proposed decoupled approximated history (MMSF-DAH) approach.

1. MMSF-C: Modularized, but centralized equivalent implementation and corresponds to CCSE (see Section 3.4.2).

2. MMSF-DP: Using the cross-covariance factorization proposed by Roumeliotis and Bekey [121], allowing decoupled propagation. Observations are performed on the full state thus it corresponds to DCSE-DP* (see Section 3.4.4).

3. MMSF-MaRS: Decoupled on-demand propagation of cross-covariance factors using a buffer for $\boldsymbol{\Phi}$. Originally, it is considering just cross-covariance between a *core* sensor and other sensors. Missing interdependencies between sensors and missing correction terms may lead to invalid joint covariances, which are then corrected by different Eigenvalue correction strategies as described by Brommer et al. in [18].

4. MMSF-DACC: Like MMSF-DP with approximations proposed by Luft et al. [95] directly applied on cross-covariance factors, to perform observation isolated and corresponds to DCSE-DACC (see Section 3.4.5).

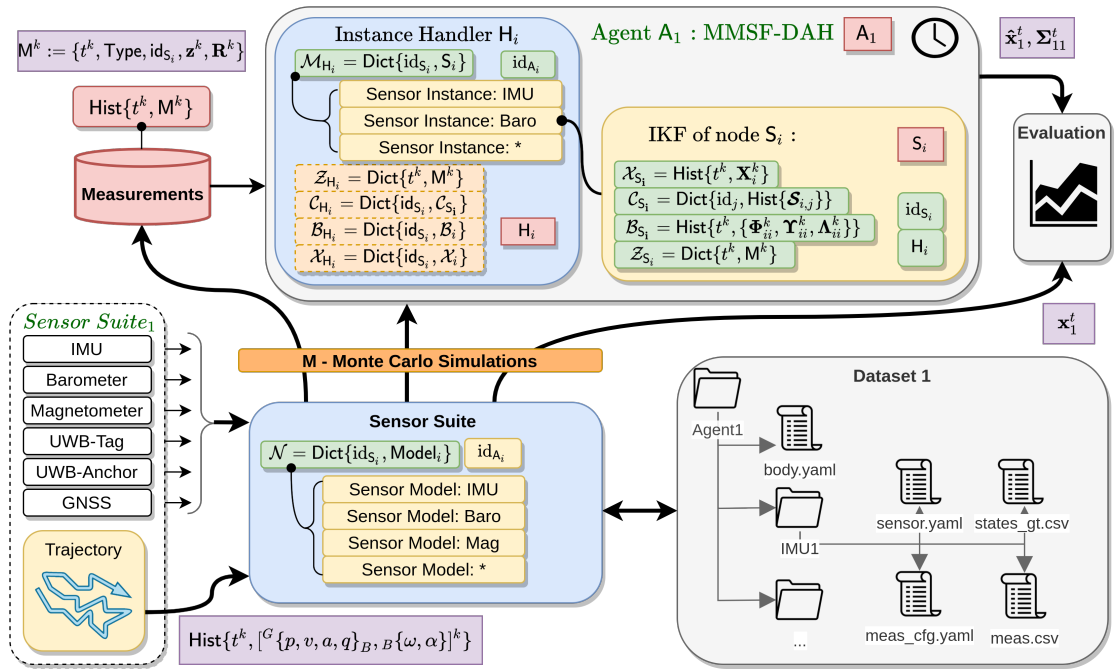

**Figure 4.2:** Shows the block diagram of the single agent's simulation framework used in Section 4.4. The agent's estimation framework is defined by a sensor suite consisting of multiple sensor models. It can be loaded and stored into a human-readable dataset or be generated from a set of emulated sensors. The sensor suite allows generating a set of measurements, initial beliefs, and to obtain the true state, which is a prerequisite for the ANEES evaluation. Based on the sensor suite's sensor models, each agent's instance handler H instantiates an appropriate sensor instance. All measurements from the sensor suites are processed sequentially and chronologically sorted, while individual sensor delays are configurable.

5. MMSF-DAH: Extending MMSF-DACC such that correction terms are buffered and applied on demand onto the cross-covariance factors. This allows constant complexity in propagation and private observations if the buffer history suffices as described in the CSE formulation of [76] and corresponds to DCSE-DAH (see Section 3.4.6).

### 4.4.2 Accuracy

We evaluate the filter performance using the *Machine Hall* sequences (MH_$\{01\ldots05\}$) of the EuRoC dataset [20] and different fusion strategies. The emulated sensor suite consists of (i) an IMU providing noisy and biased measurements at a rate of $200\,\mathrm{Hz}$, (ii) three absolute position sensors providing measurements at $5\,\mathrm{Hz}$ with a message drop rate of $20\,\%$, a standard deviation of $\boldsymbol{\sigma}_{abs} = 0.25\,\mathrm{m}$, a delay of $0.05\,\mathrm{s}$, and overlapping sensor switching phases (Figure 4.4), and (iii) a barometer providing pressure readings at a rate of $10\,\mathrm{Hz}$, a delay of $0.1\,\mathrm{s}$, a message drop rate of $20\,\%$, and a standard deviation of $\sigma_{baro} = 0.5\,\mathrm{Pa}$ (inspired from, e.g., the Bosch BME280 sensor). All states are initialized with a random offset and a reasonable uncertainty.

Table 4.1 lists the ARMSE (short AR), the mean of the NEES ($\overline{\mathrm{NEES}}$, short AN), and the average execution times for propagation and joint measurements ($\bar{t}_{\mathrm{prop}}$ and $\bar{t}_{\mathrm{joint}}$ respectively) over five sequences (including the convergence phases) of the estimated states for different MMSF approaches. Note, $\Delta$AR and $\Delta$AN are with respect to the centralized equivalent (MMSF-C) method.

As expected, MMSF-DP behaves similar to MMSF-C at the cost of computation time

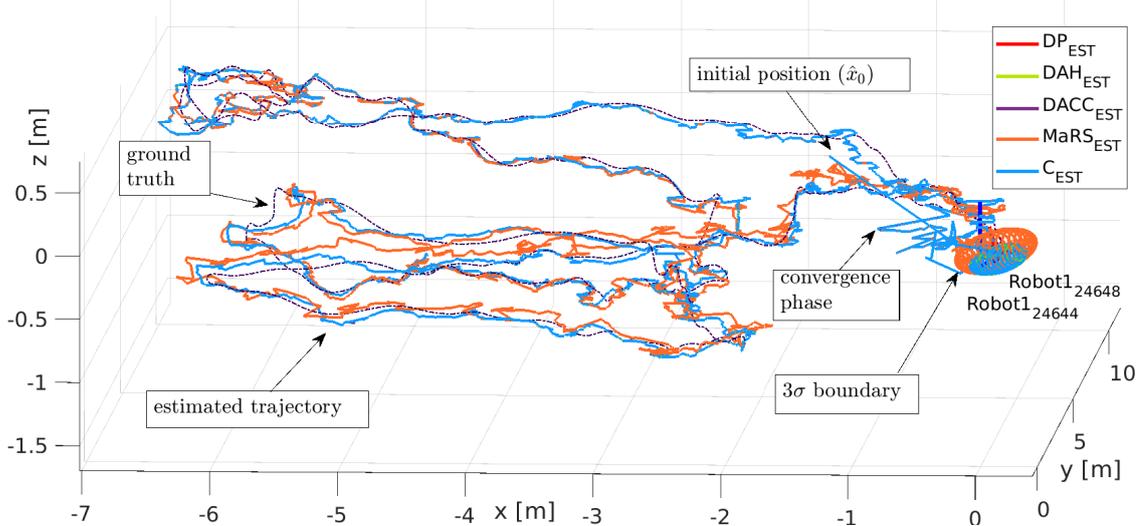

**Figure 4.3:** Estimated trajectory comparison between different CSE strategies we ported to MMSF formulations and our MMSF-MaRS approach: DAH (green, ours), DACC (purple), DP (red), MaRS (orange), and C (blue). Due to wrongly initialized states, some irregularities during the state convergence phase are noticeable. DP, DACC, and DAH perform almost identical as the statistically optimal strategy (C). DAH and MMSF-MaRS are fastest. The problem is described in Section 4.4.2. Image is reused from [74].

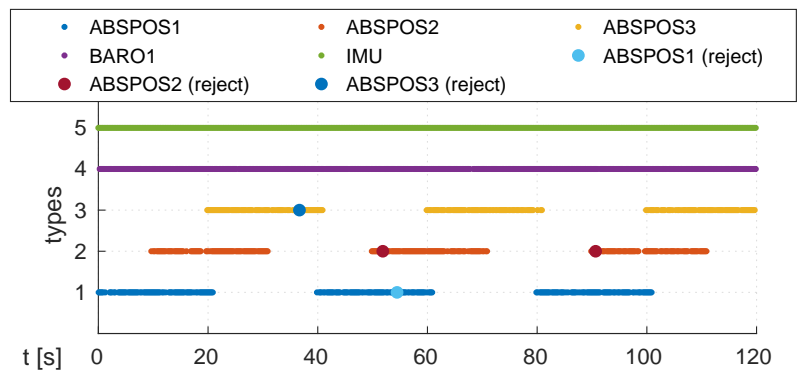

**Figure 4.4:** Measurements processed using MMSF-DAH on the MH_01 dataset. The absolute position measurements are interleaved and some are rejected. There were at most five sensor active at a time. The problem is described in Section 4.4.2. Image from [74].

(see Table 4.2) while MMSF-DACC and MMSF-DAH perform slightly worse (in absolute values) due to their approximations. Another decreasing step is shown for MMSF-MaRS due to further approximations and neglecting correction terms in update steps (see Figure 4.3 and Table 4.1). The $\overline{\text{NEES}}$ for all states should be in average 3. All states are far from being considered inconsistent in this experiment. Regarding the average execution time MMSF-MaRS and MMSF-DAH are the fastest approaches.

### 4.4.3 Scalability and latency

The scalability is demonstrated by adding sensors that provide private and joint observations to a sensor suite. It consists again of (i) an IMU providing noisy and biased measurements at a rate of 200 Hz, (ii) $K$ absolute position sensors at a rate of 5 Hz, $\boldsymbol{\sigma}_{abs} = 1$ m and a sensor delay $t_{delay}$, (iii) $K$ atmospheric sensors providing temperature and pressure information at 1 Hz and 4 Hz, $\sigma_T = 1\,°\text{C}$ and $\sigma_P = 1\,\text{Pa}$, respectively, and a sensor delay

| EuRoC | avg($\text{MH}_{\{1,2,3,4,5\}}$) | | | | | |
|---|---|---|---|---|---|---|
| | $^{\mathcal{G}}_{\mathcal{G}}\mathbf{p}_{\mathcal{I}}$ [m] | | $^{\mathcal{G}}\mathbf{R}_{\mathcal{I}}$ [deg] | | [ms] | [ms] |
| MMSF | $\Delta$AR | $\Delta$AN | $\Delta$AR | $\Delta$AN | $\bar{t}_{prop}$ | $\bar{t}_{joint}$ |
| C | - | - | - | - | 22.8 | 350 |
| DACC | 4e-8 | -7.9e-5 | -1.6e-5 | -2.5e-5 | 3.7 | 51.8 |
| DAH | -1.2e-7 | -9.7e-5 | -4e-6 | -2.1e-5 | **3.2** | **44.6** |
| DP | **-7e-16** | **-3.4e-14** | **-7e-14** | **-4e-13** | 4 | 84.9 |
| MaRS | 0.0152 | 0.438 | 0.085 | 0.24 | **3.2** | 44.7 |

| | AR | AN | AR | AN |
|---|---|---|---|---|
| C | 0.104 | 2.82 | 2.12 | 1.9 |

**Table 4.1:** Average absolute trajectory error over five sequences and different fusion strategies compared to the exact solution (C) (i) ARMSE (AR), (ii) $\overline{\text{NEES}}$ (AN), (iii) average propagation time $\bar{t}_{\text{prop}}$, (iv) average joint update time $\bar{t}_{\text{joint}}$. Different fusion approaches were used: MMSF-C, MMSF-DACC, MMSF-DAH, MMSF-DP, and MMSF-MaRS. The problem is formulated in Section 4.4.2. Best values in bold. Adapted from [74].

$t_{delay}$. In our evaluation, we set $K = \{1, 3, 5\}$ and $t_{delay} = \{0, 0.1, 0.2\}$ s.

As shown in Figure 4.6 and in Table 4.2, MMSF-C performs the worst (note that it should not be confused with a single full-state estimator, as the modularized version imposes a management overhead). Instead, MMSF-DP performs due to the high IMU propagation rate far better, while providing statistically same results.

MMSF-DACC and MMSF-DP perform the same propagation steps, while MMSF-DACC performs observations decoupled. As just the participants are required, the update execution time can be drastically reduced. MMSF-MaRS and MMSF-DAH perform a slightly different propagation strategy, than the previous ones by adding the correction terms in a buffer.

At zero latency, MMSF-DACC is the fastest strategy, but with increasing sensor delay, using a buffer for correction terms as done in MMSF-MaRS and MMSF-DAH is faster. Compared to MMSF-DACC and MMSF-DAH, MMSF-MaRS is not considering correction terms for private and joint updates which degrades the accuracy as discussed in Section 4.4.2. Figure 4.5 shows that MMSF-DACC scales in maintenance linearly with $\mathcal{O}(C)$, where $C$ denotes the number of cross-covariance factors to be corrected, while MMSF-DAH scales with $\mathcal{O}(1)$. Therefore, the buffer management overhead in MMSF-DAH, e.g., to check the time horizon (Algorithm 4.5) is with increasing number of sensors and delay smaller than MMSF-DACC's cross-covariance maintenance. This makes MMSF-DAH the favorite choice for complex real-world applications (see Table 4.2 last few columns).

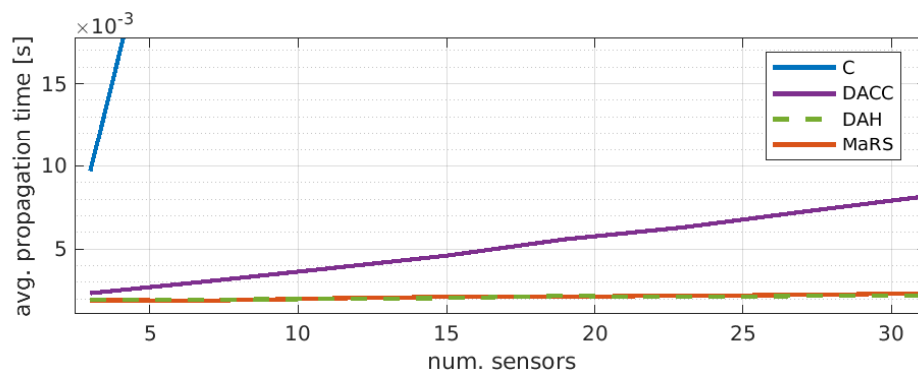

**Figure 4.5:** Average IMU propagation time with increasing number of non-delayed extero-ceptive sensors as described in Section 4.4.3. It can be seen that the propagation time using MMSF-DACC (purple) increases with the total number of sensors, while MMSF-MaRS (red) and MMSF-DAH (green, ours) remain unaffected. Image reused from [74].

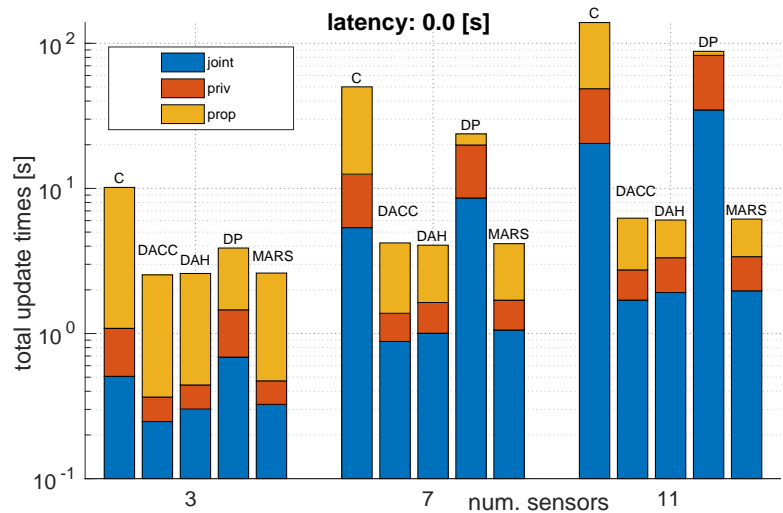

**(a)** no sensor delay.

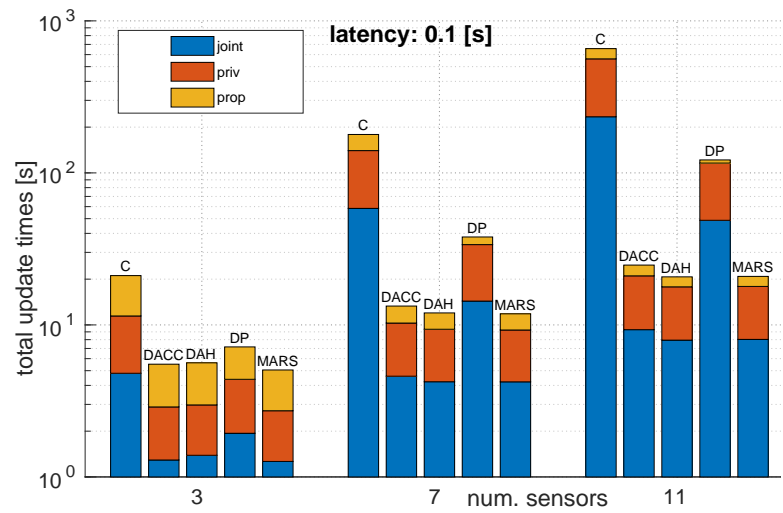

**(b)** sensor delay of 100 ms.

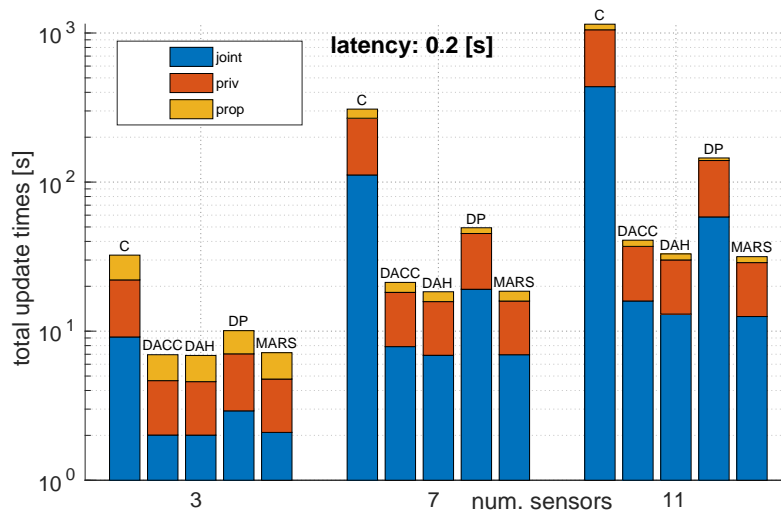

**(c)** sensor delay of 200 ms.

**Figure 4.6:** Scalability: Total execution over an increasing number of sensors and increasing sensor delay on a logarithmic scale (y-axis) for different fusion strategies and different filter steps. The problem is described in Section 4.4.3. Image reused from [74].

| | num. sensors = 3 | | | num. sensors = 7 | | | num. sensors = 11 | | | | | | | | |
| | $t_{delay} = 0\,\mathrm{ms}$ | | | $t_{delay} = 0\,\mathrm{ms}$ | | | $t_{delay} = 0\,\mathrm{ms}$ | | | $t_{delay} = 100\,\mathrm{ms}$ | | | $t_{delay} = 200\,\mathrm{ms}$ | | |
| MMSF | $\bar{t}_{pro}$ | $\bar{t}_{priv}$ | $\bar{t}_{joint}$ | $\bar{t}_{pro}$ | $\bar{t}_{priv}$ | $\bar{t}_{joint}$ | $\bar{t}_{pro}$ | $\bar{t}_{priv}$ | $\bar{t}_{joint}$ | $\bar{t}_{pro}$ | $\bar{t}_{priv}$ | $\bar{t}_{joint}$ | $\bar{t}_{pro}$ | $\bar{t}_{priv}$ | $\bar{t}_{joint}$ |
|---|---|---|---|---|---|---|---|---|---|---|---|---|---|---|---|
| C | 9.1 | 8.6 | 10.6 | 37.6 | 36.2 | 38.1 | 90.4 | 87.5 | 88.8 | 95.1 | 1040 | 1040 | 96 | 1982 | 1985 |
| DACC | 2.2 | **1.8** | **5.2** | 2.80 | **2.5** | **6.3** | 3.5 | **3.2** | **7.4** | 3.8 | 37.1 | 41.4 | 3.8 | 68.1 | 72.4 |
| DAH | 2.2 | 2.1 | 6.3 | **2.4** | 3.2 | 7.1 | **2.7** | 4.4 | 8.3 | **2.9** | **31.2** | **35.5** | 3 | 54.9 | 59.2 |
| DP | 2.4 | 11.5 | 14.3 | 3.8 | 57.2 | 60.9 | 5.4 | 148.3 | 151.1 | 5.8 | 212.8 | 216.7 | 5.7 | 261.7 | 265.4 |
| MaRS | **2.1** | 2.2 | 6.8 | 2.5 | 3.2 | 7.5 | 2.8 | 4.4 | 8.6 | 3 | 31.3 | 35.7 | **2.8** | **52.6** | **57** |

**Table 4.22:** Scalability: The table shows the average execution time in milliseconds for the three steps: propagation, private, and joint observation. With increasing sensor delay $t_{delay}$, the execution time on all methods increases. MMSF-DACC performs best with non-delayed measurements, while MMSF-DAH and MMSF-MaRS perform best with sensor delays. The reason is, that delayed measurements result in redoing updates, which are mostly propagation steps. The problem is described in Section 4.4.3. Best values in bold. Adapted from [74].

## 4.5   Conclusion

We have shown that distributed fusion techniques from Collaborative State Estimation (CSE) can be adapted and applied to decouple sensor estimates allowing Modular Multi-Sensor Fusion (MMSF) on a single agent. We hope that this new perspective on modular estimators from a CSE point of view paves the way towards efficient, consistent, and scalable generalized modular estimators.

In total, we have adapted three strategies to the MMSF domain and included the native MMSF approach MMSF-MaRS in comparisons against our proposed MMSF-DAH approach. Our evaluations on MMSF-DAH have shown (i) that using approximated correction terms for sensor observation and considering inter-sensor correlations is advantageous in terms of accuracy, and (ii) that using a common correction buffer for all factorized cross-covariances further reduce the computational effort over MMSF-DACC and other CSE-based methods.

This renders MMSF-DAH (i) ideal for high sensor rates, as it is typically the case for ESEKF based on IMU propagation, (ii) capable of performing any-sensor to any-sensor observations and private observations, (iii) re-applying fast updates after delayed (out-of-order) updates, and (iv) with minimal overhead for maintaining temporally disabled sensors.

Still, one remarkable aspect is, that the observability of hidden states and thus the self-calibration capabilities might be affected, as discussed in Section 6.5. In these cases, a two-step-approach might be implemented. If the conditions for self-calibration are given (sufficient motion in case of AINS), the instance handler could perform the estimation centralized equivalent like in MMSF-DP. Once the sensor parameters are calibrated, isolated observations can be performed again to reduce the computational complexity.

Furthermore, a non-modular estimator, which is tailored for a problem at hand and operating on the full-state can achieve very good performance, but modifying or scaling them up is difficult and limits the re-usability of the tailored implementation for other applications [138].

In future work, we consider extending our approach to add support for target tracking, where each target is modeled as individual IKF node, that is identified (maybe classified to decided for a proper model/filter specialization), added, tracked, and removed from the filter once it is not appearing for a certain duration. Landmark estimation and self-calibration was successfully addressed in the next chapter Chapter 5.

Another interesting addition would be differential fusion of sensor information, where a stationary reference sensor is used to compensate local environmental changes, e.g., a local stationary reference barometer for a barometric altimeter as illustrated in Figure 2.10, or a local stationary reference magnetometer/sun-sensor which is used as reference heading for an onboard magnetometer/sun-sensor to render a relative orientation about the gravity vector observable.

Finally, we plan to support multi-state constraints (relative time constraints between estimates) based on stochastic clones of certain states, which is an integral part of VINSs [46] or Radar-Inertial Odometry (RIO) [107]. This requires special care, since observability constraints need to be considered for our AINS formulation. Regarding robustness and resilience to single point of failures, support for multiple proprioceptive sensors is missing, like e.g., in [36] by introducing hard constraints between propagation sensors.

# Chapter 5

# Filter-based Modular Ultra-Wideband Aided Inertial Navigation

In this chapter, we demonstrate our filter-based MMSF-DAH approach's capability of processing information in a fully meshed UWB ranging network efficiently. We verify extensive Monte Carlo simulations on synthetic and real data for MAVs that the application of our CSE-inspired method in such a context breaks the computational barrier. Otherwise, it would, for the sake of complexity-reduction, prohibit the use of all available information or would lead to significant estimator inconsistencies due to coarse approximations. The content of this chapter was published in [75].

## 5.1  Introduction and Related-Work

Accurate localization is a crucial component of autonomous robotic systems, e.g., service robots, warehouse pallet robots, etc., to control and navigate accurately in unknown and GNSS-denied environments. The demand for a cost friendly, scalable and accurate indoor positioning infrastructure is growing. Typically, range-based localization systems use TDOA, TOA, RTOF [115] or RSS metrics/lateration methods to determine the distance between the antennas [127]. A promising technology for both data transmission and localization is based on UWB radio frequency (RF) signals [127]. It has desirable features for estimating distances between two transceivers, as its large bandwidths allow the UWB receiver to accurately estimate the arrival time of the first signal path [127].

In order to estimate a unique 3-DoF position of a non-stationary UWB device/node (tag), classical mulitlateration approaches requires simultaneous range measurements from at least four known stationary and non-coplanar UWB nodes (anchors). The precision of the mulitlateration is strongly dependent on (i) the relative pose between the ranging modules' antennas [86], (ii) the placement of stationary modules (anchors) as it influences the PDOP of the mobile modules, and (iii) multipath effects and non-line-of-sight (NLOS) condition may lead to wrong distance estimates [61]. Consequently, it is reasonable to combine range measurements of mobile modules with complementary sensors, e.g., an IMU and a barometer, as depicted in Figure 5.1, to increase the robustness against dropouts, precision, and accuracy [14, 49, 62, 116, 134, 140].

Hol *et al.* show in [61], that tightly fusing range measurements with inertial measurements obtained by an IMU in a probabilistic fusion algorithm allows (i) to recover a 6-DoF pose, (ii) to bridge periods with limited UWB range measurements, and (iii) to successfully detect and reject outliers. By assuming known and static anchor locations, unmodeled errors will degrade the estimation performance, in case the anchor positions were not accurately measured initially. To address this issue, calibration routines to es-timate the position of deployed anchors were proposed in, e.g., [14, 16, 54, 58, 62, 112, 134].

**Figure 5.1:** Spatial frame constellation of the proposed UWB-AINS framework. Image reused from [75].

In [58], Hausman *et al.* proposed a multi-sensor fusion approach combining inertial sensor data with loosely coupled vision-based pose measurements, GPS measurements, and tightly-coupled UWB range measurements for precision landing. They proposed a Linear Least Squares (LLS) initialization scheme for anchors, based on estimated tag positions, that provides the estimator with an initial belief for three estimated anchor positions.

In recent years, relative state estimation approaches to eliminate the need for stationary anchors in GPS-denied environment have been presented in [52, 53, 132, 153, 167]. For instance, Guo *et al.* present in [52, 53] an infrastructure-free cooperative approach to estimate the positions of neighboring MAVs. Similarly, Xu *et al.* presented in [153] an optimization-based, fully decentralized visual-inertial-UWB fusion framework for relative state estimation in a swarm of MAVs. Nguyen *et al.* extended in [116] a state-of-the-art optimization-based VIO algorithm to fuse camera, IMU and UWB range measurements from a single anchor to reduce the drift. Having only a single anchor in this configuration allows rendering the relative position with respect to the navigation frame observable, but the orientation about the gravity vector is still unobservable. Therefore, at least two anchors need to be known to achieve a fully observable system by assuming the gravity vector is parallel to one of the navigation frames axes, e.g., in our case, the z-axis.

Song *et al.* fused in [140], LIDAR, UWB, and inertial measurements in an EKF-SLAM algorithm, while the UWB ranging measurement reduced accumulated errors in the proposed LIDAR-based SLAM algorithm.

Recently, Goudar and Schoelling investigated in [49] on the spatial-temporal self-calibration of the UWB tag in a tightly-coupled UWB-aided INS. They proof the local identifiability of the time delay between the IMU and the UWB-tag by investigating on the input-output representation of the system. The temporal offset is identifiable, if the tag is not co-located with an anchor and at least on axis of the accelerometer or all axes of the gyroscope are exited and the tag is not co-located with the IMU. The observability of error-state was studied by assuming three range observations to anchors, we will show in Section 7.5.3, that two anchor positions are sufficient, if assumption on the gravity vector are made, i.e. we assume that the gravity vector is aligned with the z-axis of the navigation frame. Therefore, they proof that the system is locally weakly observable if three non-collinear anchors are available, the tag is non-coplanar with the anchors, and all axes of the accelerometer and gyroscope are exited.

Shi *et al.* investigated in [134] on the anchor self-calibration in a tightly-coupled UWB-ranging and IMU fusion algorithm in simulations with five anchors and varying ranging noise. In the first step of this algorithm, a coarse anchor position initialization, minimizing the residual between predicted tag positions and unknown anchor position in a least-squares problem. In the second step, these initial guesses are used for estimating the full state in a regular ESEKF [103] heavily limiting the scalability of the approach to more anchors. Also, the previously mentioned approaches suffer either from poor scalability or coarse approximations, affecting estimator consistency [10].

Therefore, we revisit the infrastructure-based UWB inertial localization and continuous/online anchor self-calibration for large UWB networks using our recently proposed MMSF-DAH approach, see Section 4.3. It is based on the IKF paradigm and has promising attributes that can be extended to render real-time tightly-coupled and scalable UWB aided inertial navigation. From that work, we borrow the idea to treat any inter-sensor observation in an isolated fashion, requiring only the participating sensor estimates. For our present approach, by associating each UWB device to a single sensor instance in the MMSF framework, measurements between any UWB device can be processed efficiently. This step allows combining the modular inclusion of states of external sensors with a regular MMSF with onboard sensors. Our main contributions can be summarized as:

- We propose a modular and scalable UWB-inertial based ESEKF, merging aspects from CSE and MMSF in order to estimate an agent's 6-DoF motion and sensor calibration states, and simultaneously estimate in a SLAM-like fashion geometry-states of very large UWB sensor networks in real-time.
- We perform extensive Monte Carlo simulations on synthetic and on real data from a MAVs dataset to verify both consistency and accuracy of the proposed approach.
- We evaluate the self-calibration of simulated anchor positions using different sensor configurations and against different state-of-the-art fusion strategies, showing improved performance by incorporating meshed range measurements between stationary UWB anchors.

## 5.2 Problem Formulation

In the proposed sensor constellation, an IMU is used as a proprioceptive state propagation sensor, a barometer is used for a tightly-coupled height estimation (both sensors are typically available on MAV) and a single UWB tag that performs ranging measurements with the UWB anchors in communication range. Also, UWB anchors perform measurements among them if in range.

We assume commercial UWB modules to work with the double-sided two-way-ranging protocol (High Precision Ranging (HiPR)) proposed by Neuhold *et al.* [115] allowing fast ranging acquisitions at a rate of 40 Hz by performing a Round-Robin scheduling to avoid network congestion, and measurement broadcasting to close-by sensor nodes. The UWB tag acquires/sniffs range measurement to and between nearby stationary UWB anchor upon fly-by in communication range and fuses them locally in a probabilistic modular filter framework, while, in contrast to other work, consistently accounting for correlations between individual sensor estimates. For simplicity, other than Gaussian noise, we do not assume any biases on the signals, nor extrinsic calibration between tag and IMU. We refer to [16] on how to include these elements into an UWB-inertial estimator and how to perform *a-priori* a coarse anchor position initialization.

Each sensor has a unique identifier and belongs to a known class of sensor types. Measurements are processed in a ESEKF-based MMSF framework [74], which associates each measurement to a sensor instance that performs the information fusion with potentially other instances of the sensor suite. Each sensor instance maintains sensor specific states, such as calibration parameters needed for self-calibration. Further, each instance performs a statistical NIS [10] hypothesis check to detect and reject outliers, and is capable of processing delayed (out-of-order) measurements (see [74]).

Applying joint observations, incorporating different estimates, results in cross-covariance terms between them and at some point, all sensor instances might be correlated.

Classical centralized filters such as [58] are typically not truly modular, as they operate on the predefined full state as one entity (including the full joint covariance matrix) making individual sensor filter step very costly in large networks.

We integrate our approach in three different MMSF strategies (cf. [74] for details on the general concepts of these approaches) and compare the resulting performance: MMSF-C is a centralized-equivalent EKF filter implementation performing all filter steps on the entire full state vector. MMSF-DP is a centralized-equivalent EKF filter implementation, performing the state propagation of individual sensor instances independently, while update steps are performed on the full state vector. MMSF-DAH performs all filter steps isolated, requiring only sensor instances that are directly involved (so-called participants) in the filter update steps, while correlations to non-participants are conservatively approximated. We show in Section 5.3 that the last strategy is best suited for our real-time UWB-inertial aided navigation and meshed anchor self-calibration.

### 5.2.1  Sensor Suite

The sensor suite of the proposed AINS consists of a varying number of stationary UWB anchors, an UWB tag, a barometer, and an IMU. Due to indirect error estimation [104], observations have to be expressed by their error $\bar{\mathbf{z}} = \mathbf{z} \ominus \hat{\mathbf{z}}$. This measurement error needs to be linearized with respect to the error state at the current estimate $\bar{\mathbf{z}} = \mathbf{H}\bar{\mathbf{x}}$ with the measurement Jacobian $\mathbf{H} = \frac{\partial h}{\partial \mathbf{x}} \frac{\mathbf{x}}{\partial \bar{\mathbf{x}}}\big|_{\hat{\mathbf{x}}}$ for the measurement function $\mathbf{z} = h(\mathbf{x})$. Details on the error-state definition, the measurement models, and Jacobians can be found in Section 2.9. The IMU state $\mathbf{x}_{\mathcal{I}}$ is defined as

$$\mathbf{x}_{\mathcal{I}} = \begin{bmatrix} {}^{\mathcal{G}}_{\mathcal{G}}\mathbf{p}_{\mathcal{I}}; {}^{\mathcal{G}}_{\mathcal{G}}\mathbf{v}_{\mathcal{I}}; {}^{\mathcal{G}}\mathbf{q}_{\mathcal{I}}; {}_{\mathcal{I}}\mathbf{b}_{\omega}; {}_{\mathcal{I}}\mathbf{b}_{a} \end{bmatrix} \tag{5.1a}$$

with ${}^{\mathcal{G}}_{\mathcal{G}}\mathbf{p}_{\mathcal{I}}, {}^{\mathcal{G}}_{\mathcal{G}}\mathbf{v}_{\mathcal{I}}$, and ${}^{\mathcal{G}}\mathbf{q}_{\mathcal{I}}$ as the position, velocity, and orientation of the IMU $\mathcal{I}$ w.r.t. to the global navigation frame $\mathcal{G}$. ${}_{\mathcal{I}}\mathbf{b}_{\omega}$ and ${}_{\mathcal{I}}\mathbf{b}_{a}$ are the estimated gyroscope and accelerometer biases to correct the related IMU readings (see Equation (2.79)). The unobservable pose between the body frame $\{\mathcal{B}\}$ and IMU $\{\mathcal{I}\}$, ${}^{\mathcal{B}}_{\mathcal{B}}\mathbf{p}_{\mathcal{I}}$ and ${}^{\mathcal{B}}\mathbf{q}_{\mathcal{I}}$, are constants that need to be known *a-priori* . The barometer state $\mathbf{x}_{\mathcal{P}}$ is defined as

$$\mathbf{x}_{\mathcal{P}} = \begin{bmatrix} {}^{\mathcal{B}}_{\mathcal{B}}\mathbf{p}_{\mathcal{P}} \end{bmatrix} \tag{5.1b}$$

with ${}^{\mathcal{B}}_{\mathcal{B}}\mathbf{p}_{\mathcal{P}}$ being the position between the body reference frame and the sensor, which is assumed to be fixed and known *a-priori* . To model the range measurements between two UWB nodes, e.g., between a tag $\{\mathcal{T}\}$ and anchor $\{\mathcal{A}\}$ or between anchors, $\{\mathcal{A}_i\}$ and $\{\mathcal{A}_j\}$, we include the 3D position of each anchor ${}^{\mathcal{G}}_{\mathcal{G}}\mathbf{p}_{\mathcal{A}}$ with respect to the navigation frame $\{\mathcal{G}\}$ in the state estimation process. Therefore, the anchor state $\mathbf{x}_{\mathcal{A}}$ is

$$\mathbf{x}_{\mathcal{A}} = \begin{bmatrix} {}^{\mathcal{G}}_{\mathcal{G}}\mathbf{p}_{\mathcal{A}} \end{bmatrix} \tag{5.1c}$$

and the tag state $\mathbf{x}_{\mathcal{T}}$ is

$$\mathbf{x}_{\mathcal{T}} = \begin{bmatrix} {}^{\mathcal{B}}_{\mathcal{B}}\mathbf{p}_{\mathcal{T}} \end{bmatrix} \tag{5.1d}$$

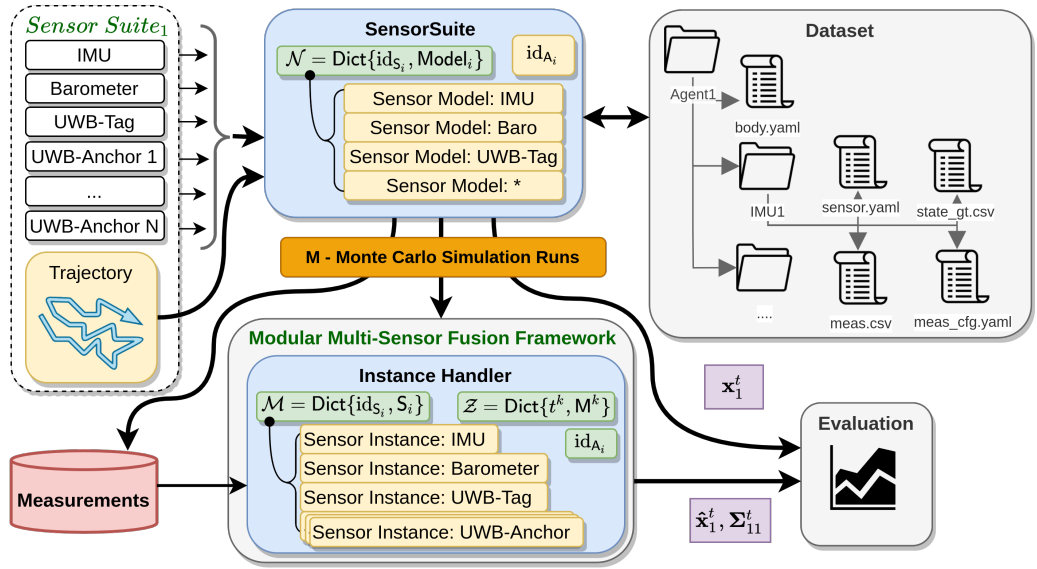

**Figure 5.2:** Shows the block diagram of the simulation framework. Image adapted from [75].

with ${}^{\mathcal{B}}_{\mathcal{B}}\mathbf{p}_{\mathcal{T}}$ being the transformation between the body reference frame $\mathcal{B}$ and the UWB tag, which is assumed to be constant and not included in the estimation.

Summarized and as depict in Figure 5.1, our modular ESEKF formulation is based on the IMU navigation states and their dynamics in Equation (2.84). The filter obtains corrections from the barometer (Equation (2.111) and Equation (2.110)), the measurements between UWB tag and anchors in range (Section 2.9.6) as well as the inter-anchor measurements observed from anchors in range (Section 2.9.6). The inclusion of the inter-anchor measurements as an extension to our CSE inspired MMSF framework ([74]) is key to enable consistent and scalable inclusion of the UWB mesh geometry in the estimation process in real-time. This renders our approach a real-time capable consistent UWB-inertial SLAM-like estimator.

## 5.3 Evaluation

The experiments are done in our MATLAB framework, that allows to load real data from the EuRoC [20] dataset or to generate smooth trajectories and noisy, biased IMU samples. Exteroceptive measurements are generated based on the ground truth trajectory and are modified by the sensors' calibration states and noise parameters. Furthermore, a delay and dropout rate can be applied to these measurements. The noisy and biased real-world IMU samples provided by the datasets are used without modifications. Finally, in multiple Monte Carlo simulation runs, all measurements are processed in an instance handler as depict in Figure 5.2. It maintains multiple sensor instances and communication between them is handled locally. The estimates and ground truth values from the dataset are used for deterministic and reproducible evaluation of the estimator credibility, which is described in the following section.

The simulated UWB range measurements are modeled based on the HiPR protocol [115], with a ranging standard deviation of $0.1\,\mathrm{m}$. Further, we assume to have coarse initial beliefs about each anchor's locations, e.g., by performing a calibration procedure described by Blueml *et al.* [16] or manual measurements.

The evaluations are performed single-threaded in MATLAB on an AMD Ryzen 7 3700X CPU with a $32\,\mathrm{GB}$ DDR4 RAM.

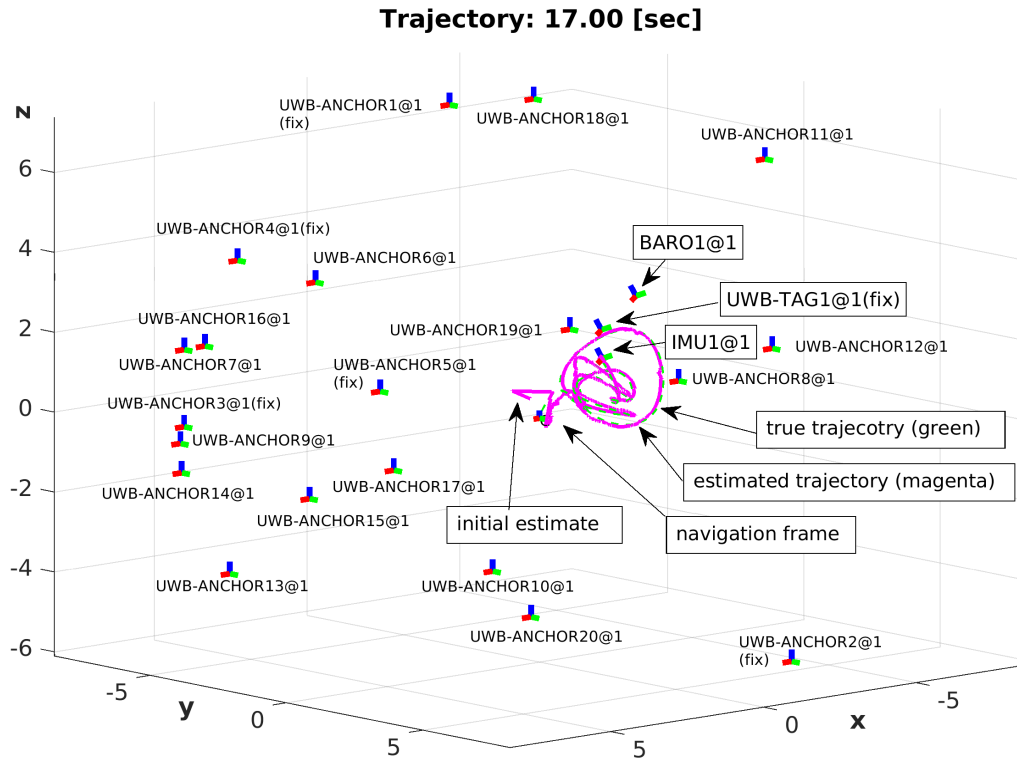

**Figure 5.3:** Scenario $S_1$: Shows the estimated trajectory at $t = 17$ s, the 15 estimated and 5 fixed anchor positions, randomly placed on a sphere with a radius of 7 m, and the estimate IMU pose with the exteroceptive sensor positions of the barometer and the fixed UWB tag. The problem is described in Section 5.3.1. Image reused from [75].

### 5.3.1 Scenario $S_1$

In the first scenario, $S_1$, we study the tightly-coupled anchor self-calibration in our modular aided inertial estimation framework. Therefore, 20 anchors are randomly distributed on a sphere with a radius of 7 m. Five anchor positions are assumed to be known and set to be fixed in the estimation framework, meaning that the corresponding sensor instances are excluded from sensor fusion. According to a nonlinear observability analysis, only two anchors need be known, in order to render the nonlinear estimation problem fully observable. Since the system suffers from approximated models, linearization errors, multi-rate measurements, and a low signal-to-noise ratios, we found five fixed anchors to render a good compromise in terms of convergence behavior.

The agent moves along a smooth and randomly generated trajectory within the sphere for a duration of $D = 150$ s as shown in Figure 5.3. All states are initialized with a randomly assigned $\pm 1\sigma$ offset from the true value, as described in Table 5.1. The IMU sample rate is 100 Hz and all UWB ranging devices are in communication range (leading to maximum complexity), while 10 % of measurements are randomly dropped. In total, three experiments regarding the anchor self-calibration are conducted. The first studies the anchor self-calibration by just obtaining tag to anchor (T-A) range measurements. In the second experiment, additionally range measurements between anchors (A-A) are used. In the third experiment, tag to anchor and readings from the barometer are used. Three different modular multi-sensor fusion strategies, MMSF-C, MMSF-DP, and MMSF-DAH are applied as already described in Section 5.2.

| | $^{\mathcal{G}}_{\mathcal{G}}\mathbf{P}_{\mathcal{A}}$ | $^{\mathcal{B}}_{\mathcal{B}}\mathbf{P}_{\{\mathcal{T},\mathcal{P}\}}$ | $^{\mathcal{G}}_{\mathcal{G}}\{\mathbf{p},\mathbf{v}\}_{\mathcal{I}}$ | $^{\mathcal{G}}\mathbf{q}_{\mathcal{I}}$ | $_{\mathcal{I}}\mathbf{b}_a$ | $_{\mathcal{I}}\mathbf{b}_\omega$ |
|---|---|---|---|---|---|---|
| $\boldsymbol{\sigma}^0$ | 30 cm | 10 cm | 1 {m, m/s} | 5 deg | 0.05 m/s$^2$ | 0.05 rad/s |
| $\Delta\mathbf{x}^0$ | ±30 cm | ±10 cm | ±1 {m, m/s} | ±5 deg | ±0.05 m/s$^2$ | ±0.05 rad/s |

**Table 5.1:** Scenario S$_1$: Initial uncertainty and initial state offsets $\Delta\mathbf{x}^0$ for the UWB anchor positions, IMU states, and exteroceptive sensor states. ± emphasis that a positive or negative values is assigned randomly per element. Table is reused from [75].

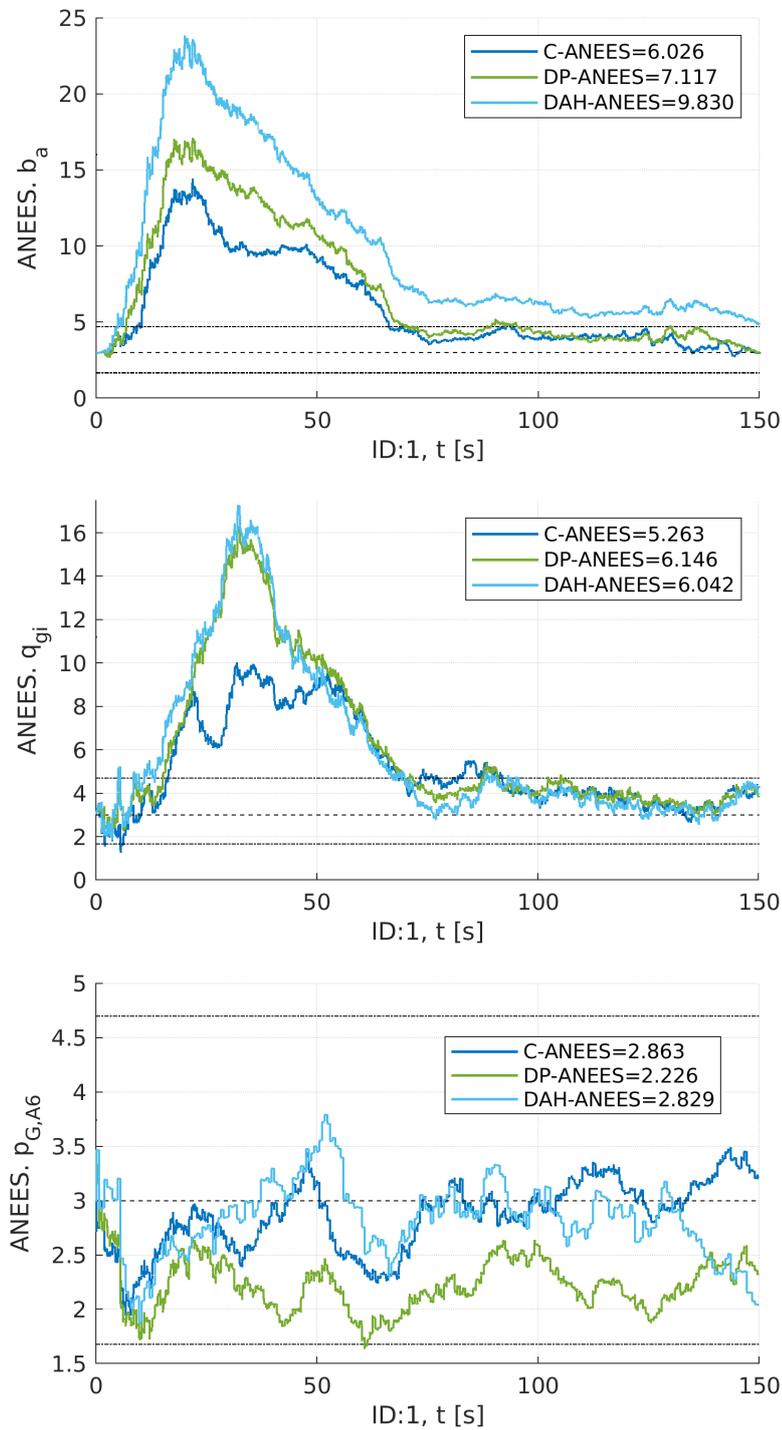

**Figure 5.4:** Scenario $S_1$: T-A ranging. Shows the ANEES, the double-sided 95 % confidence region (dotted lines), and expected ANEES value (dashed line) over 10 Monte Carlo simulation runs of the accelerometer bias $\mathbf{b}_a$, the IMU orientation, and the 6th UWB anchor position using MMSF-C (blue), MMSF-DP(green), and MMSF-DAH (cyan). The experiment is described in Section 5.3.1. Image is reused from [75].

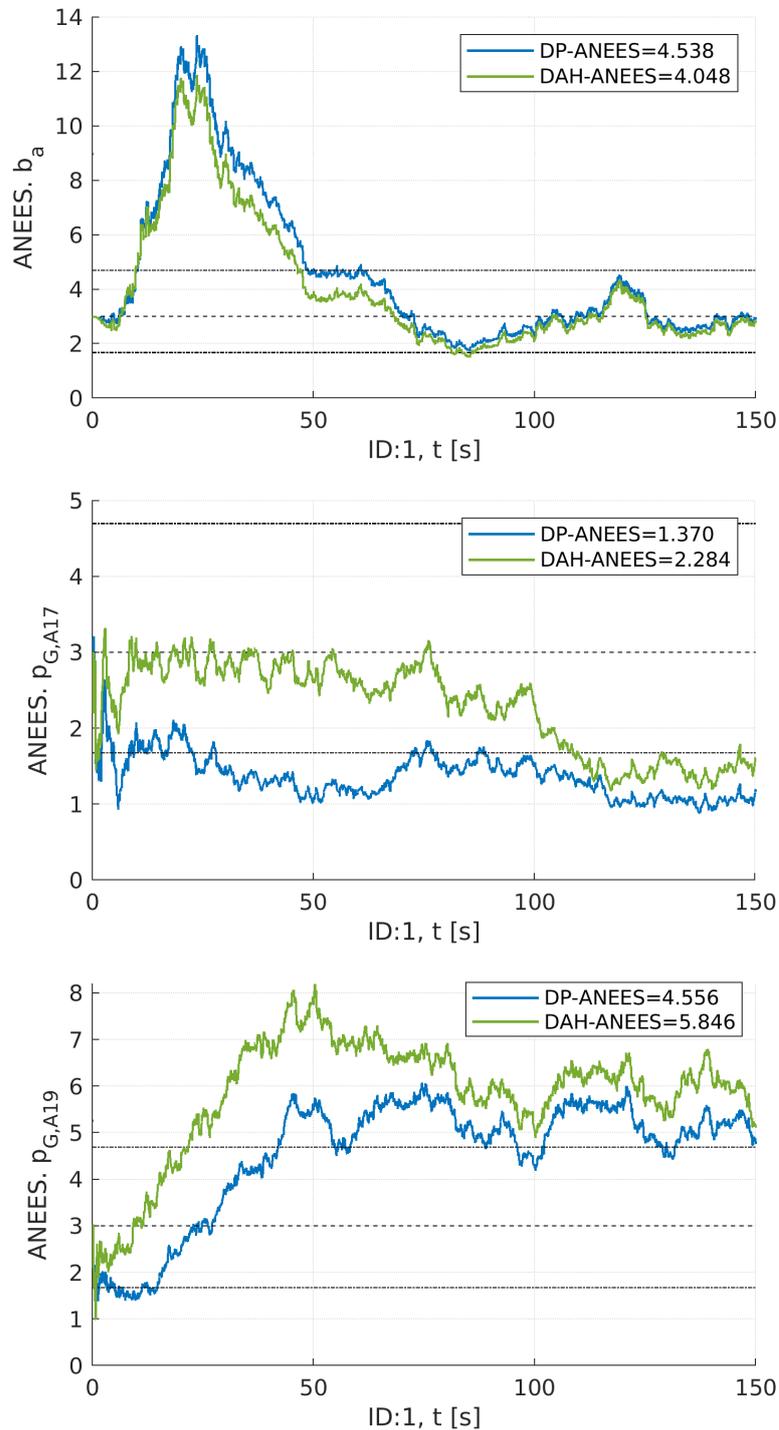

**Figure 5.5:** Scenario $S_1$: T-A and A-A ranging. Shows the ANEES, the double-sided 95 % confidence region (dotted lines), and expected ANEES value (dashed line) over 10 Monte Carlo simulation runs of the accelerometer bias $\mathbf{b}_a$, the 17th and 19th UWB anchor position using MMSF-DP (blue) and MMSF-DAH (green). The experiment is described in Section 5.3.1. Image reused from [75].

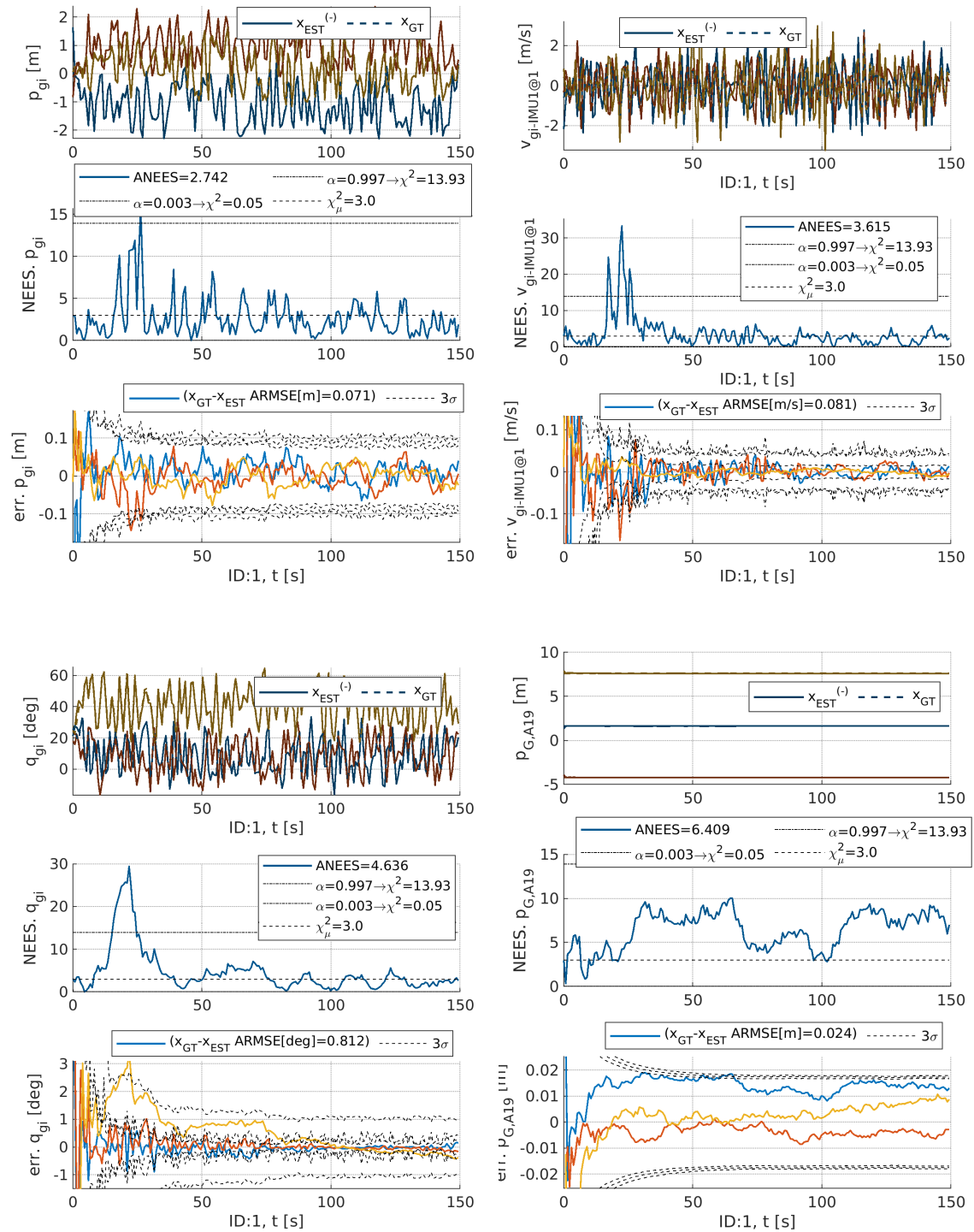

**Figure 5.6:** Scenario $S_1$: T-A and A-A ranging estimation results of the first Monte Carlo run using MMSF-DAH. The top row shows the true (dashed) and estimated values (solid), the second row the single run NEES with the double-sided 99.7 % confidence region (dotted lines), and the third row show the estimation error and the $3\sigma$ boundaries. The four quadrants are the estimated IMU position, velocity, orientation, and the estimated position of the 19th UWB anchor. In yellow, blue, red are for the x, y, z position, or the roll, pitch, yaw angle, respectively. The experiment is described in Section 5.3.1. Image adapted from [75].

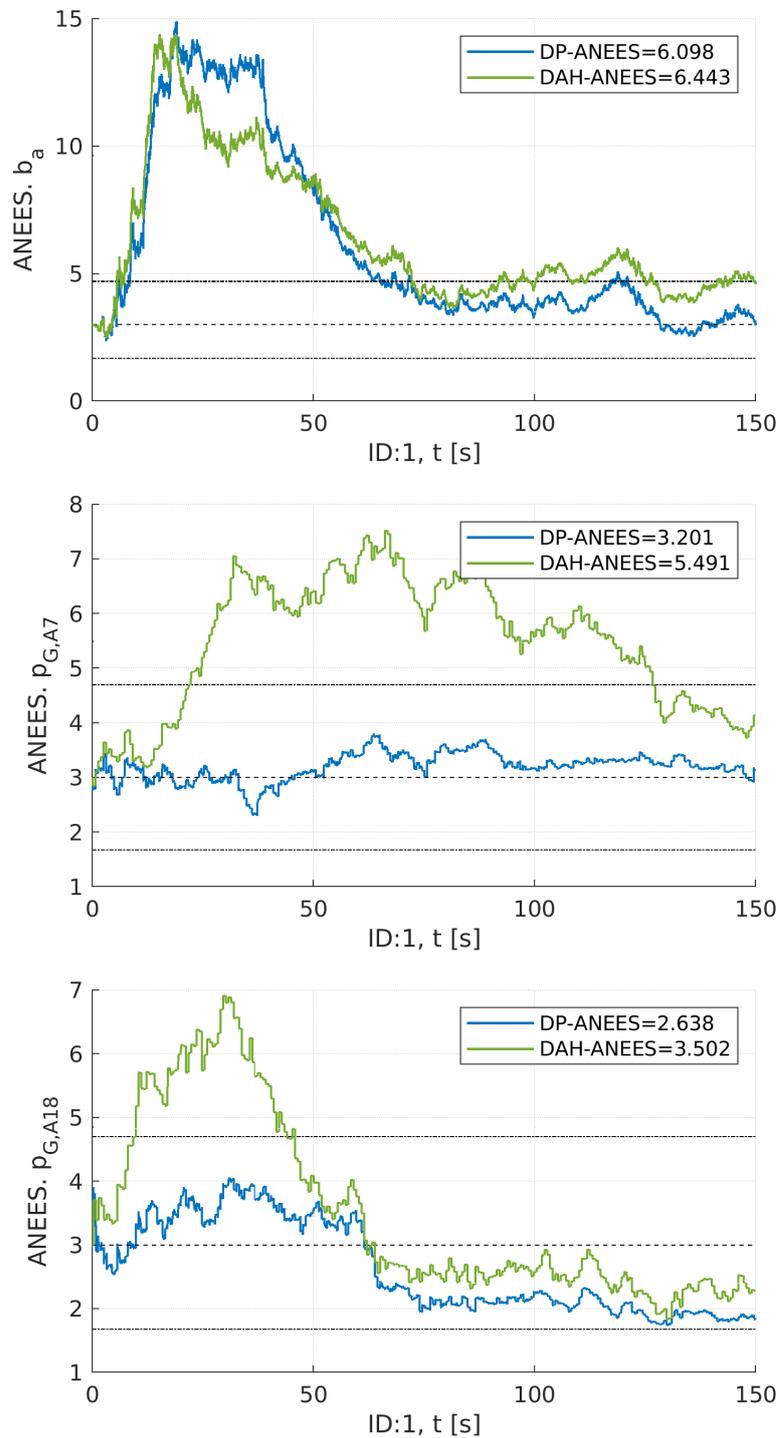

**Figure 5.7:** Scenario $S_1$: T-A ranging and pressure readings. Shows the ANEES, the double-sided 95 % confidence region (dotted lines), and expected ANEES value (dashed line) over 10 Monte Carlo simulation runs of the accelerometer bias $\mathbf{b}_a$, the 7th and 18th UWB anchor position using MMSF-DP (blue) and MMSF-DAH (green). The experiment is described in Section 5.3.1. Image reused from [75].

| 10 Monte Carlo runs | | | [s] | [s] | [s] | [cm] | | [cm/s] | | [deg] | | [m/s²] | | [rad/s] | | [cm] | |
|---|---|---|---|---|---|---|---|---|---|---|---|---|---|---|---|---|---|
| MMSF | A-A | Baro | $\bar{t}_{prop}$ | $\bar{t}_{joint}$ | $t_{tot}$ | $^{G}_{G}\mathbf{p}_I$ | | $^{G}_{G}\bar{\mathbf{v}}_G$ | | $^{G}\mathbf{q}_I$ | | $_I\bar{\mathbf{b}}_a$ | | $_I\bar{\mathbf{b}}_\omega$ | | $^{G}_{G}\mathbf{p}_A$ | |
| | | | | | | AR | AN | AR | AN | AR | AN | AR | AN | AR | AN | AR | AN |
| C | 0 | 0 | 0.089 | 0.096 | 1774 | **4.51** | 2.81 | **7.4** | 3.59 | **0.92** | 5.26 | **0.015** | 6.02 | **0.0066** | 3.016 | **11.1** | 2.88 |
| DP | 0 | 0 | 0.0063 | 0.253 | 1237 | 4.69 | 2.70 | 7.8 | 3.73 | 1.01 | 6.14 | 0.016 | 7.12 | 0.0068 | 2.89 | 11.3 | 2.46 |
| DAH | 0 | 0 | **0.0039** | **0.0045** | **79.4** | 5.3 | 4.49 | 7.9 | 3.8 | 0.98 | 6.04 | 0.018 | 9.82 | 0.0067 | 3.56 | 11.6 | 3.45 |
| DP | 1 | 0 | 0.0073 | 0.113 | 9741 | **4.3** | 2.53 | **6.9** | 3.14 | **0.91** | 3.43 | **0.015** | 4.54 | **0.0079** | 4.6 | **1.5** | 3.39 |
| DAH | 1 | 0 | **0.0052** | **0.0044** | **446.7** | 4.4 | 2.52 | 7.1 | 2.98 | **0.91** | 3.18 | **0.015** | 4.04 | **0.0079** | 3.95 | 1.6 | 4.82 |
| DP | 0 | 1 | 0.0066 | 0.274 | 2075 | **3.85** | 3.76 | **6.1** | 4.36 | **0.81** | 6.17 | **0.012** | 6.1 | 0.0062 | 3.56 | **12.1** | 2.96 |
| DAH | 0 | 1 | **0.0041** | **0.0047** | **95.2** | 4.52 | 6.6 | 6.2 | 4.7 | 0.83 | 6.94 | **0.012** | 6.44 | 0.0066 | 4.22 | 13 | 5.04 |

**Table 5.2:** Scenario $S_1$: Shows the average ARMSE (AR) and ANEES (AN) of the estimated states as well as the average over the estimates UWB anchor positions for different fusion strategies and different sensor configurations averaged over 10 Monte Carlo simulation runs. Please note the total execution time $t_{tot}$ is the average over a single run, while $\bar{t}_{prop}$ and $\bar{t}_{joint}$ are the average values for a single filter propagation or update step. Since the ARMSE calculation includes the entire trajectory, including the initialization and the state convergence phase, it is on average higher than the error after convergence. Best values in bold and problem is described in Section 5.3.1. Table is reused from [75].

### T-A ranging

In this experiment, IMU and T-A measurements are fused in the modular estimator framework using three different fusion strategies. The effective T-A measurement rate for each anchor is 2 Hz due to the Round Robin scheduling performed. As shown in Table 5.2, MMSF-DAH is the fastest approach that allows to significantly reduce the total filter execution time by a factor of 22.3 and a factor of 15.6 over MMSF-C and MMSF-DP, respectively. MMSF-DAH seems to be slightly more confident than the other two approaches and the ANEES converges slower to the desired mean value. Nonetheless, the ANEES plots of the weakly observable accelerometer bias $\mathbf{b}_a$, the IMU orientation ${}^{\mathcal{G}}\mathbf{q}_{\mathcal{I}}$, and the position of the sixth (as a representative and also random choice for space reasons) anchor ${}^{\mathcal{G}}\mathbf{p}_{\mathcal{A}_6}$ for all strategies shown in Figure 5.4, behave similarly. Summarized, the ARMSE of anchor positions reduced from an initial RMSE of 30 cm (see Table 5.1) below to approximately 12 cm on average over the entire trajectory using either fusion strategy in Table 5.2.

### T-A and A-A ranging

In this experiment, IMU, T-A and A-A range measurements were fused using MMSF-DP and MMSF-DAH. MMSF-C was excluded due to the huge single run computation time. The effective T-A and A-A measurement rate for each device is 1 Hz due to the time scheduling performed to avoid network congestion. As shown in Table 5.2, MMSF-DAH is 21.8 times faster then MMSF-DP. It can be clearly seen in the same table, that A-A range measurements significantly reduce the estimation error of the anchor positions down to 1.6 cm. The inclusion of these measurements comes at the cost of higher complexity, leading to 5.6 times higher computation time in the case of MMSF-DAH. This could motivate to perform a two-stage approach: first, performing an accurate anchor localization, then once the system is calibrated, only highly efficient T-A measurements are performed.

In Figure 5.5, the ANEES of the accelerometer bias, the 17th anchor position $\mathcal{A}_{17}$ and the 19th anchor position $\mathcal{A}_{19}$ is shown. Interestingly, the A-A measurements have a positive impact on the ANEES of the accelerometer bias, and it also seems to converge faster, compared to the T-A-only experiment. The ANEES of the 17th and 19th anchor are chosen as they show both under-confidence and over-confidence, which is independent of the selected fusion strategy. Apart from a position/constellation related dependency issue due to PDOP, the major impact stems form lacking observability in areas with no direct link to fixed anchors. Nonetheless, the state plots in Figure 5.6 of the first Monte Carlo simulation run using MMSF-DAH, show an exemplary and satisfying estimation behavior of the navigation states, while the estimated anchor position $\mathcal{A}_{19}$ is not converging, as it is not globally observable as discussed later in Section 7.5.4.

### T-A and pressure

In this experiment, we evaluate the impact of fusing pressure readings of a rigidly attached barometer loosely coupled in the modular estimation framework using MMSF-DP and MMSF-DAH. The calibration state between the barometer and the body reference frame are assumed to be known *a-priori* and are fixed. Pressure readings are processed at a rate of 20 Hz with measurement standard deviation of $\sigma_{\mathcal{P}} = 1$ Pa, translating to approximately 8.4 cm standard deviation at sea-level and is slightly beyond commercially integrated pressure sensors. Again, 10 % of pressure readings are dropped randomly. As shown in Table 5.2, MMSF-DAH is again 21.8 times faster as MMSF-DP. Fusing pressure reading tightly leads to more accurate IMU estimate in case of MMSF-DP than fusing A-A measurements. Interestingly, the pressure readings are not improving the anchor position estimates across approaches and is subject to future investigations. In Figure 5.5, the ANEES of the accelerometer bias, the 7th and 18th UWB anchor are depicted, indicating again that MMSF-DAH is more optimistic than MMSF-DP and that the ANEES converges slower, while all estimates tend to converge towards the defined confidence region.

| 10 Monte Carlo runs | | [cm] | | [cm/s] | | [deg] | | [m/s$^2$] | | [rad/s] | | [cm] | |
|---|---|---|---|---|---|---|---|---|---|---|---|---|---|
| MMSF | EuRoC | $^{\mathcal{G}}_{\mathcal{G}}\bar{p}_{\mathcal{I}}$ | | $^{\mathcal{G}}_{\mathcal{G}}\bar{v}_{\mathcal{G}}$ | | $^{\mathcal{G}}\bar{q}_{\mathcal{I}}$ | | $_{\mathcal{I}}\bar{b}_a$ | | $_{\mathcal{I}}\bar{b}_\omega$ | | $^{\mathcal{G}}_{\mathcal{G}}\bar{p}_{\mathcal{A}}$ | |
| | | AR | AN | AR | AN | AR | AN | AR | AN | AR | AN | AR | AN |
| DP | MH_04 | **9.82** | 3.28 | **9.6** | 3.97 | **1.52** | 2.96 | **0.03** | 2.91 | **0.003** | 2.1 | **5.9** | 4.77 |
| DAH | MH_04 | 10.7 | 3.81 | 9.8 | 3.92 | 1.58 | 2.87 | **0.03** | 2.72 | **0.003** | 2 | 6.6 | 4.7 |
| DP | MH_05 | **10.8** | 3.86 | **9.3** | 4.63 | **1.72** | 3.38 | **0.029** | 2.2 | **0.0028** | 2.62 | **5.65** | 5.78 |
| DAH | MH_05 | 12.56 | 4.3 | 9.77 | 4.87 | 1.73 | 3.77 | 0.03 | 3.37 | 0.0029 | 4.43 | 6.55 | 5.66 |

**Table 5.3:** Scenario S$_2$: Shows the average ARMSE (AR) and ANEES (AN) of the estimated states as well as the average over the estimates UWB anchor positions for different fusion strategies and different sensor configurations averaged over 10 Monte Carlo simulation runs. Best values in bold and problem is described in Section 5.3.2. Table is reused from [75].

### 5.3.2   Scenario S$_2$

For evaluating the estimator's credibility and the computation time of individual sensor instances in a realistic scenario, we use two *Machine Hall* sequences (MH_04 and MH_05) of the EuRoC dataset [20]. The simulation runs for 78 s and 92 s, which is the flight time of the *Machine Hall* sequences between take-off and landing. Each MAV is equipped with an IMU, barometer, and an UWB transceiver (tag), for both communication and pairwise ranging between other UWB modules in communication range. Figure 5.1 depicts the spatial frame constellation. Twenty-five stationary UWB transceiver (anchors) are assumed to be deployed to cover the area of interest as shown in Figure 5.8 with a communication range of 4 m. The communication range is on purpose short, to justify the deployment of 25 anchors and to challenge the estimation problem: anchors are revisited again, meaning that they are correlated and need to be considered properly, and some anchors have no direct link to (fixed) reference anchors. Five anchors close to the take-off position of the MAV are fixed (constant) to define the global coordinate reference frame.

The same initial values and parameters as in Section 5.3.1 are used, with the difference that the initial uncertainty of the anchor position was lowered to $\boldsymbol{\sigma}_{^{\mathcal{G}}\mathbf{p}_{\mathcal{A}}} = 0.1\,\mathrm{m}$ and the effective UWB ranging rate between devices in range is 8 Hz (due to the sparse configuration, fewer devices are in communication range, which allows a higher net rate for individuals) and the IMU rate in the dataset is 200 Hz. All UWB nodes are inserted artificially into the real-world dataset.

Figure 5.8 shows the estimation performance using MMSF-DP and MMSF-DAH. As confirmed in Table 5.3, hardly any difference in the estimated IMU states are noticeable, while operating on the full state vector (with 75 elements) in case of MMSF-DP in the filter update steps causes tremendous computation efforts over treating them isolated in case of MMSF-DAH. This processing speedup of almost 26.5 comes at the cost of a slightly degraded and slower converging anchor self-calibration, as discussed in Section 5.3.1.

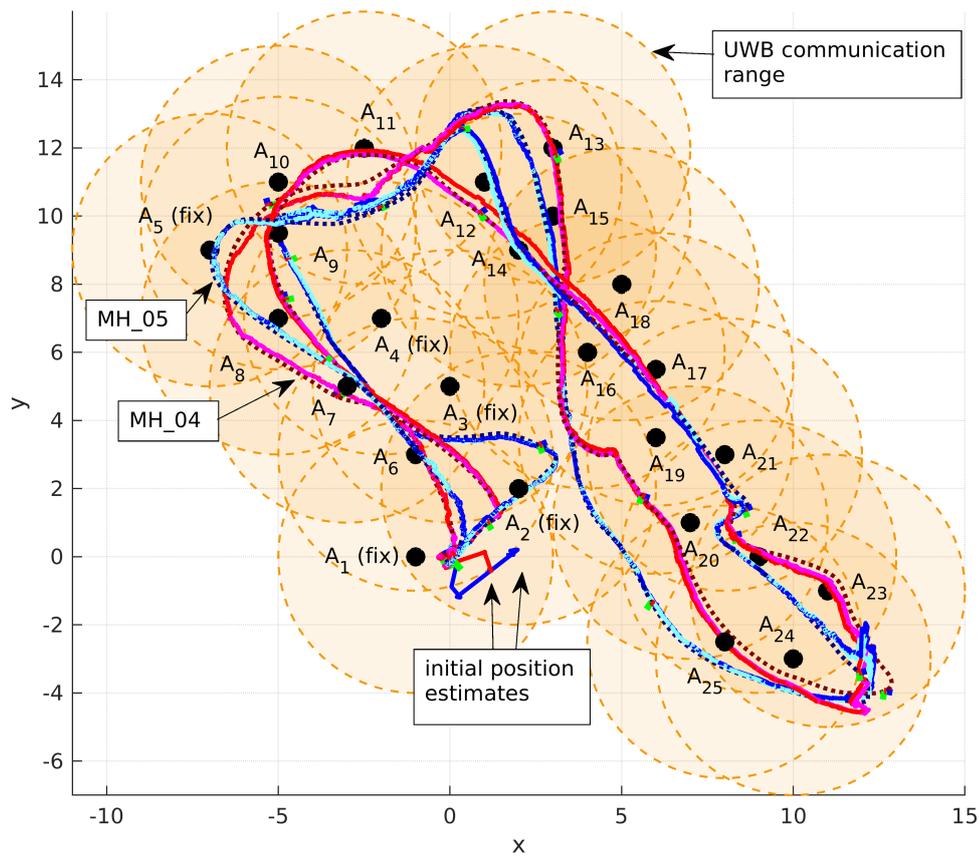

**Figure 5.8:** Scenario $S_2$: Top view on the anchor placement (black bullets) and the Machine Hall 4 and 5 trajectories of the *Machine Hall* sequences of the EuRoC dataset [20] in dashed dark red and dashed dark blue, respectively. In red and blue are the estimated position using MMSF-DAH, while in mageneta and cyan using MMSF-DP. The filled orange circles show the communication range of 4 m per UWB anchor. Fixed anchors at the starting position are assumed to be constant and known, to define the navigation reference frame. Image is reused from [75]

## 5.4   Conclusion

Considering inter-sensor observations not involving the core navigation states in a MMSF framework is a new paradigm originating from CSE. In our evaluations, we have shown that considering these as tightly-coupled range observations between UWB anchors can significantly improve the estimates of both, the navigation states and estimated anchor positions. In such formulations, the computational effort increases significantly using centralized-equivalent estimators. With the presented sensor extensions to our recently proposed MMSF-DAH approach (see Chapter 4), which builds upon the IKF paradigm (see Chapter 6), we show the feasibility of processing inter-sensor observation as isolated joint updates between sensor instances without requiring estimates of other sensors and, thus, resulting in a significant speedup. This breaks the computational barrier such that meshed (and generally ill scaling) inter-sensor observations can be fused to aided inertial navigation estimators, while maintaining consistency in the navigation states.

UWB-aided inertial motion estimation with simultaneous online UWB anchor position initialization (and extension by fly-by), and continuous self-calibration remains a challeng-ing task. An interesting addition would be an ad hoc UWB anchor position initialization during a fly-by with continuous self-calibration, once an UWB tag detects and obtains measurements from a new unknown anchor. In future work, we will also investigate in increasing the robustness in case of NLOS conditions (that lead to unmodeled offsets in range measurements), which requires consistent estimates for a statistical outlier rejection.

| 10 Monte Carlo runs | | [s] | [s] | [s] |
|---|---|---|---|---|
| MMSF | EuRoC | $\bar{t}_{prop}$ | $\bar{t}_{joint}$ | $t_{tot}$ |
| DP | MH_04 | 0.0071 | 0.209 | 2337.8 |
| DAH | MH_04 | **0.0036** | **0.0029** | **88.7** |
| DP | MH_05 | 0.0072 | 0.204 | 3087.9 |
| DAH | MH_05 | **0.0038** | **0.0032** | **116.6** |

**Table 5.4:** Scenario $S_2$: Shows the average execution times for different fusion strategies and different sensor configurations averaged over 10 Monte Carlo simulation runs. Please note the total execution time $t_{tot}$ is the average over a single run, while $\bar{t}_{prop}$ and $\bar{t}_{joint}$ are the average values for a single filter propagation or update step. Best values in bold and problem is described in Section 5.3.2. Table is reused from [75].

# Chapter 6

# The Isolated Kalman Filtering Paradigm

Our recent publications on DCSE [76] and on Modular Multi-Sensor Fusion (MMSF) [74], covered in Chapter 3 and Chapter 4, respectively, allow us to generalize these filter formulations and to propose a new Kalman filter decoupling paradigm, denoted as Isolated Kalman Filtering (IKF). This paradigm builds upon approximations introduced by Luft *et al.* in [94, 95].We formally discuss the paradigm on a generic estimation problem, we introduce a novel buffering scheme that allows processing delayed measurements, and prove that the approximation made, in order to achieve isolated joint update steps, is based on an implicit maximum determinant (and thus maximum entropy) completion of the incomplete full state covariance matrix. We provide a source code of a generic estimation framework called ikf_lib, that implements the IKF paradigm and supports out-of-sequence measurements. The steady-state behavior in eleven different observation graphs was studied and compared against an optimal and naive approach, followed by a filter credibility analysis. This paradigm and the gained insights are of particular interest for the robotics community, since DCSE-DACC [95] became almost a standard for distributed multi-robot filter-based cooperative localization. For instance, De Carli *et al.* used it in [30] as estimation approach for their perception aware path planing algorithm for multi-robot systems.

## 6.1  Introduction

This paradigm allows to decouple various estimated states of a single filter into multiple sub-states that are estimated in decoupled sub-filters in a suboptimal, but credible fashion. The decoupling is achieved in a sense that each sub-filter can perform certain steps independently – i.e. isolated – from other filters, while at the same time correlations between the sub-filters are encountered/considered and corrected approximately. Thus, it builds upon the fundamental assumption that dynamics of individual nodes are decoupled and exploits a spare output coupling between nodes, which was already covered thoroughly with respect to CSE in Section 3.2. Observations relating to multiple filters only require the presence or temporal availability of those during the process of performing the joint update, thus again, in an isolated fashion. The computation time of individual isolated filter steps is theoretically invariant to the total number of decoupled sub-filters in the system. Additionally, sub-filters can perform private observations isolated, in case the internal state is directly or partially observable by a measurement. These isolated updates might lead to suboptimal estimates with respect to the exact filter formulation, similar to the Schmidt-Kalman Filter (SKF), while at the same time, it reduces the computational complexity, as only sub-filters, that are coupled through the observation, are needed.

With this paradigm, originating from DCSE-DAH, we want to rethink the classical monolithic filter formulation based on a global full state space and treat the estimation problem rather as a set of filter nodes in an abstract/virtual network with communication and processing capabilities, where each node is estimating its subspace of the global full state[1]. The granularity, in which the global full state is divided, depends on problem at hand, the required modularity, e.g., if it desired to add or remove single nodes frequently, and if the dynamics of the states are decoupled[2].

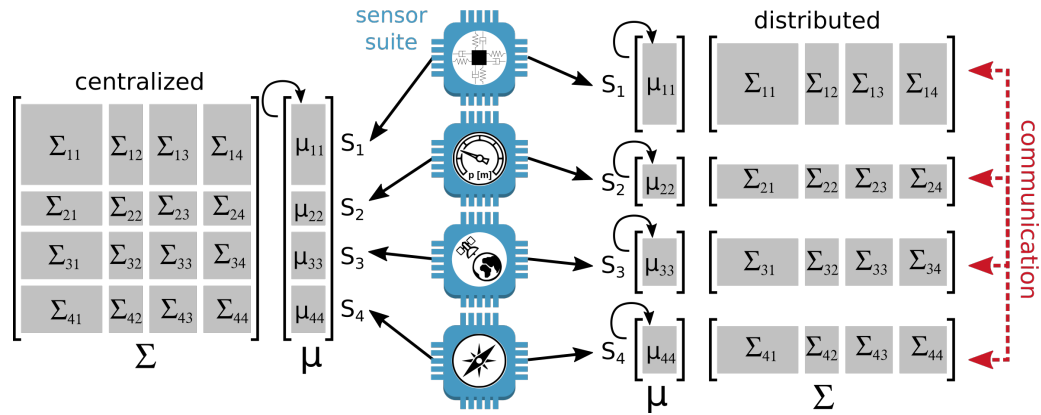

**Figure 6.1:** A centralized and distributed realization of fusing information provided by four sensors. In a centralized realization, a single filter instance is operation on a full state, while in a distributed realization, the full state is split up into reasonable subspaces, which are maintained in individual filter instances. In order to process measurements exactly, communication and information exchange between the instances is required. In fact and as shown in [74], various DCSE strategies can be applied to decouple these individual filter instances. The proposed IKF is built upon approximations, that reduces both, the computation and communication effort and allows to decouple states, in the sense that observation can be performed isolated among participants.

Figure 6.1 illustrates the difference of a centralized/monolithic and decoupled filter formulation. Another example can be made by revisiting the sensor constellation proposed by Hausmann *et al.* in [58], that consists of a single IMU as state propagation sensor, a camera, a GPS sensor, and a range sensor measuring distances to three stationary anchors. Splitting the estimation problem up into individual *nodes* would result in one node for the IMU, the camera, the GPS sensor, and one for each estimated anchor position. Isolated joint updates would be performed between the IMU and each individual exteroceptive sensor. This abstraction made, embodies inherently modularity by design and allowed us to propose a MMSF algorithm in Chapter 4 and a unified distributed EKF algorithm for both MMSF and CSE in Chapter 7, which typically has been addressed in literature separately.

While processing delayed measurement in a centralized formulation is rather trivial, as discussed in Section 2.9.8, it imposed challenges in a decoupled or distributed system as operations need to be reapplied among correlated estimators.

More precisely, our contributions can be summarized as:

- Formulating the Isolated Kalman Filtering paradigm as a general method to decouple Kalman filters by approximating and distributing relevant cross-covariance terms even of currently non-involved filters,
- introducing a buffering scheme that allows to process delayed measurements, and
- performing an extensive Monte Carlo simulation on a linear estimation problem, including a credibility analysis.

---

[1] Going even a step further, one could process tasks in parallel (e.g., multi-threaded), if, e.g., individual nodes would be spawn as threads of a single process. A considerable downside is the overhead and complexity for scheduling and task synchronization.

[2] This is often the case for calibration states which are modeled to have zero dynamics.

The rest of the chapter is structured as follows: in Section 6.2 we describe the estimation problem as a set communicating estimators with processing capabilities, while Section 6.3 presents the IKF paradigm on a generic node. In Section 6.4, we evaluate the paradigm on a linear estimation problem in a Monte Carlo simulation and our findings are concluded in Section 6.5.

## 6.2   Problem Formulation

The problem formulation for IKF is the counterpart to the one for filter-based CSE in Section 3.2, where we described the estimation problem on a swarm of communicating agents, with the difference that individual agents become now local nodes of an abstract network. While we stressed in Section 3.2, that it is important to regard/treat the individual estimates of agents in a swarm as part of a global full state, we want to stress with the IKF paradigm, that parts of global full state can be treated as individual nodes of an abstract (locally communicating) network, in order to exploit the sparsity of the input and output coupling, as shown in Figure 6.2.

Therefore, we start with the dynamics of a stochastic process, that consists of three groups of states that are dynamically not linked/coupled with each other, meaning that the off-diagonal blocks of the state transition and process noise matrix are zero. Each subgroup is associated to a *node* of the abstract *network*. Therefore, the matrices $\boldsymbol{\Phi} = \text{Diag}\left([\boldsymbol{\Phi}_{11}; \ldots; \boldsymbol{\Phi}_{NN}]\right)$, $\mathbf{Q} = \text{Diag}\left([\mathbf{Q}_{11}; \ldots; \mathbf{Q}_{NN}]\right)$, and $\mathbf{B} = [\mathbf{B}_{11}; \ldots; \mathbf{B}_{NN}]$ are obtained from the individual nodes' dynamics. The measurement sensitivity matrix $\mathbf{H} = \text{Diag}\left([\mathbf{H}_{11}, \ldots, \mathbf{H}_{NN}]\right)$ is built from the individual nodes' matrices. Typically, it has a sparse structure, as it is not always possible to relate the measured quantity to all internal states, meaning that parts of the full state space remains hidden or unobserved (see Section 2.9.7).

The dynamics of a system with $N$ nodes, $\mathsf{S} := \{\mathsf{S}_i | i = 1, \ldots, N\}$, can be represented, e.g., in the form

$$\mathbf{x}^k = \boldsymbol{\Phi}^{k|k-1} \mathbf{x}^{k-1} + \boldsymbol{\Gamma}^{k-1} \mathbf{u}^{k-1} + \mathbf{G}^{k-1} \mathbf{w}^{k-1}, \tag{6.1a}$$

$$\mathbf{z}^k = \mathbf{H}^k \mathbf{x}^k + \mathbf{v}^k, \tag{6.1b}$$

where $\mathbf{z}^k = \left[\mathbf{z}_1^k; \ldots; \mathbf{z}_N^k\right]$ is the concatenation of measurements available at each agent at $t^k$, $\mathbf{x}^k = \left[\mathbf{x}_1^k; \ldots; \mathbf{x}_N^k\right]$ is the stacked state, $\mathbf{u}^k = \left[\mathbf{u}_1^k; \ldots; \mathbf{u}_N^k\right]$ the stacked control input, $\mathbf{w}^k = \left[\mathbf{w}_1^k; \ldots; \mathbf{w}_N^k\right]$ and $\mathbf{v}^k = \left[\mathbf{v}_1^k; \ldots; \mathbf{v}_N^k\right]$ are the concatenated zero-mean white Gaussian process and observation noise, respectively.

While the physical process of an individual node $\mathsf{S}_i$ is represented as linear stochastic system in the form

$$\mathbf{x}_i^k = \boldsymbol{\Phi}_{ii}^{k|k-1} \mathbf{x}_i^{k-1} + \boldsymbol{\Gamma}_{ii}^{k-1} \mathbf{u}_i^{k-1} + \mathbf{G}_{ii}^{k-1} \mathbf{w}_i^{k-1}, \tag{6.1c}$$

$$\mathbf{z}_i^k = \mathbf{H}_{ii}^k \mathbf{x}_i^k + \mathbf{v}_{ii}^k, \tag{6.1d}$$

$$\mathbf{z}_{i,j}^k = \mathbf{H}_{i,j}^k \begin{bmatrix} \mathbf{x}_i^k \\ \mathbf{x}_j^k \end{bmatrix} + \mathbf{v}_{i,j}^k, \tag{6.1e}$$

where $\mathbf{x}_i$ is the state vector, $\mathbf{u}$ the (control) input, and $\mathbf{z}_i$ and $\mathbf{z}_{i,j}$ are output vectors. $\boldsymbol{\Phi}_{ii}$, $\boldsymbol{\Gamma}_{ii}$, and $\mathbf{G}_{ii}$ are the state transition, input coupling, and process noise coupling matrices, respectively. $\mathbf{H}_{ii}$ and $\mathbf{H}_{i,j}$ are the measurement sensitivity matrices for the private and joint observations, respectively. Note that joint observations are not limited to a pairwise coupling. $\mathbf{w}_i \sim \mathcal{N}\left(\mathbf{0}, \mathbf{Q}_{ii}\right)$ and $\mathbf{v}_{\{ii,i,j\}} \sim \mathcal{N}\left(\mathbf{0}, \mathbf{R}_{\{ii,i,j\}}\right)$ are the known process and observation noise, that are assumed to be zero-mean white Gaussian with an associated covariance $\mathbf{Q}_{ii}$ and $\mathbf{R}_{\{ii,i,j\}}$, respectively.

A set of nodes $\mathsf{S}$ that is (temporarily) coupled in their outputs are so-called participants $\mathsf{P}$ and defined as

$$\mathsf{P} := \{\mathsf{P}_i \in \mathsf{S} | 1, \ldots, P\}, \tag{6.1f}$$

while the set of non-participants is defined as

$$\bar{\mathsf{P}} := \mathsf{S} \setminus \mathsf{P}. \tag{6.1g}$$

Like in Section 3.2.1, this abstract network can be modeled as a graph $\mathcal{G}(\mathsf{S}, \mathcal{E})$, where $\mathsf{S} = \{1, \ldots, N\}$ is the vertex set of nodes and the edge set $\mathcal{E} \subseteq \mathcal{V} \times \mathcal{V}$, where the presence of an edge $e_o^k = (i, j)$ represents the possibility for an output coupling between nodes $i$ and $j$ at the time step $t(k)$. A bundle/set of edges represent an output coupling (measurement) that relates to multiple nodes. For the IKF, the communication graph is a bi-directed star graph rooted at the interim master node, containing all vertices of the observations. Both, the vertex and edge set are time-varying, e.g., by adding or removing a new IKF or in case of sensor depletion.

### 6.2.1 Summary

The IKF paradigm could be seen as alternative to Schmidt-states/nuisance parameters (see Section 2.8.3), whose beliefs remain unchanged, but the uncertainty of the nuisance parameters and the correlation with the essential parameters are considered in the update step. In the same sense, IKF states can be used to represent nuisance parameter, but they obtain correction via joint observations, meaning the mean and covariance of *isolated* parameters are changing.

## 6.3 Isolated Kalman Filtering for a generic (sensor) node

IKF is based on three concepts, (i) the *isolated state propagation* introduced by Roumeliotis and Bekey in [121] and (ii) the *isolated state correction* based on seminal work of Schmidt [129] and Luft *et al.* [95], and (iii) the correction buffering scheme that was introduced in our previous DCSE-DAH approach [76] (see Section 3.3) and refined in our MMSF-DAH approach [74] (see Section 4.3) to support delayed (out of sequence) measurements.

The support of delayed measurements requires a more sophisticated buffering scheme, as shown in Figure 6.3 and is described in following section. In the sections Section 6.3.2 and Section 6.3.3, we elaborate on the proposed correction terms inspired by [76].In Section 6.3.4, we handle asynchronously received measurements, which is in particular relevant for practical applications, since the assumption made in Section 3.3.5 can degrade the estimation performance when, e.g., agents perform aggressive motions. Although parts were already described in Section 3.3, we decided to repeat those instead of referencing them, to support readability and to provide a self-contained chapter.

### 6.3.1 Buffering Scheme

Figure 6.3 shows the block diagram of the IKF for a node $\mathsf{S}_i$, consisting of four buffers; $\boldsymbol{\mathcal{B}}_{\mathsf{S}_i} := \mathsf{Hist}\{t^k, \{\boldsymbol{\Phi}, \boldsymbol{\Upsilon}, \boldsymbol{\Lambda}\}^k\}$ for the correction terms, $\boldsymbol{\mathcal{X}}_{\mathsf{S}_i} := \mathsf{Hist}\{t^k, \mathbf{X}_i^k\}$ for the beliefs, $\boldsymbol{\mathcal{C}}_{\mathsf{S}_i} := \mathsf{Dict}\{\mathsf{id}_j, \mathsf{Hist}\{\mathcal{S}_{i,j}\}\}$ for the factorized cross-covariances, and $\boldsymbol{\mathcal{Z}}_{\mathsf{S}_i} := \mathsf{Hist}\{t^k, \mathsf{M}^k\}$

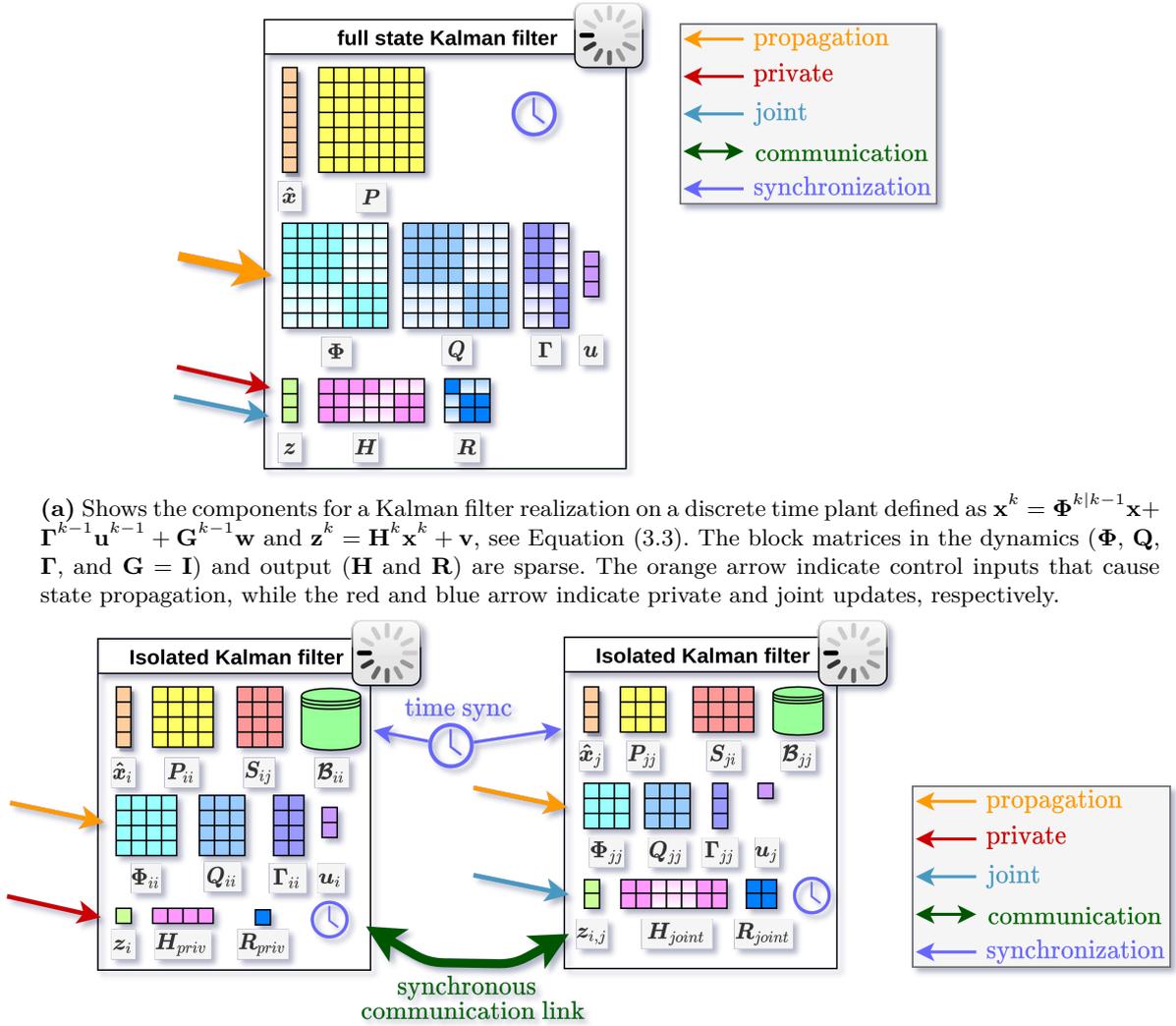

**(a)** Shows the components for a Kalman filter realization on a discrete time plant defined as $\mathbf{x}^k = \mathbf{\Phi}^{k|k-1}\mathbf{x} + \mathbf{\Gamma}^{k-1}\mathbf{u}^{k-1} + \mathbf{G}^{k-1}\mathbf{w}$ and $\mathbf{z}^k = \mathbf{H}^k\mathbf{x}^k + \mathbf{v}$, see Equation (3.3). The block matrices in the dynamics ($\mathbf{\Phi}$, $\mathbf{Q}$, $\mathbf{\Gamma}$, and $\mathbf{G} = \mathbf{I}$) and output ($\mathbf{H}$ and $\mathbf{R}$) are sparse. The orange arrow indicate control inputs that cause state propagation, while the red and blue arrow indicate private and joint updates, respectively.

**(b)** Shows the components of an isolated Kalman filter realization of Figure 6.2a. By exploiting the sparsity in the input dynamics of the plant, two IKF instances can be used to address the estimation problem. Each IKF instance needs to maintain, in addition the factorized cross-covariance $\mathcal{S}$, a correction buffer $\mathcal{B}$. The matrices to describe the isolated parts of the plant are smaller. Each instance can process its control input and private measurements independently (isolated from the rest). A coupling between the IKF instances happens at the moment a joint measurement $\mathbf{z}_{i,j}$ is obtained, and requires information from both instances. Please note that the estimates $\hat{\mathbf{x}}_i$ and $\hat{\mathbf{x}}_j$ are suboptimal, thus not exact.

**Figure 6.2:** Comparison between a full state Kalman filter formulation and corresponding, decoupled formulation based on two isolated Kalman filter instances. Please note that delayed (out-of-sequence) measurements are not supported in these realizations.

for the processed measurements. The definitions for a dictionary $\mathsf{Dict}$ and sliding time horizon buffer $\mathsf{Hist}$ can be found in Section 2.2.

$\mathcal{C}_\mathsf{S}$ stores a history of factorized cross-covariances in dictionary. According to [121], the cross-covariance between the estimators/nodes $\mathsf{S}_i$ and $\mathsf{S}_j$ at time $k$ can be factorized

$$\mathbf{\Sigma}_{ij}^k = \mathcal{S}_{ij}^k \left(\mathcal{S}_{ji}^k\right)^\mathsf{T},   \tag{6.2}$$

where the choice of decomposition is arbitrary, e.g., $\mathcal{S}_{ij}^k = \mathbf{\Sigma}_{ij}^k$ and $\mathcal{S}_{ji}^k = \mathbf{I}$.

The cross-covariance factors $\mathcal{S}_{ij}^k$ are stored in a dictionary $\mathcal{C}_{\mathsf{S}_i} := \{\mathsf{Hist}_j | j, \ldots, N\}$ of sliding time horizon buffers $\mathsf{Hist}$ denoted as $\mathcal{C}_{\mathsf{S}_i}$ which are accessed via another node's

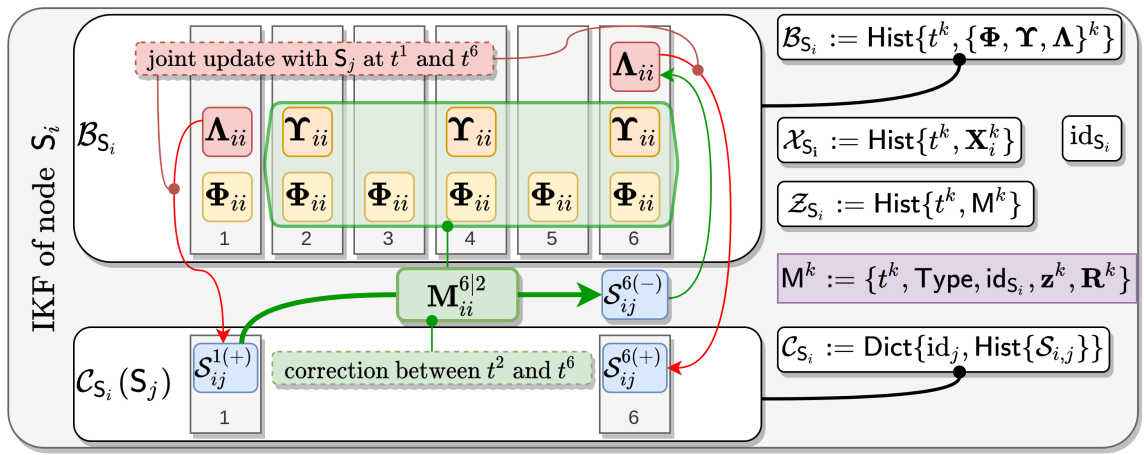

**Figure 6.3:** Block diagram of the IKF, which shows the accumulated elements (e.g., $\mathbf{M}_{ii}^2 = \mathbf{\Upsilon}_{ii}^2 \mathbf{\Phi}_{ii}^{2|1} = \mathcal{B}_{\mathsf{S}_i}(t^2)$) in the correction buffer $\mathcal{B}_{\mathsf{S}_i}$ of node $\mathsf{S}_i$. $\mathbf{\Phi}$ for propagation steps, $\mathbf{\Upsilon}$ for private, and $\mathbf{\Lambda}$ for joint observations (see Sections 6.3.2 and 6.3.3). At $t = 6$, the correlated estimators $\mathsf{S}_{\{i,j\}}$ perform again a joint observation. To account for meanwhile propagation and update steps, each estimator has to apply these corrections on the factors $\mathcal{S}_{\{ij,ji\}}^{1(+)}$ stemming from their last encounter before using them in the current joint observations. By accumulating corrections $\mathbf{M}_{\{ii,jj\}}^{6|2} = \prod_{l=2}^6 \mathcal{B}_{\mathsf{S}_{\{i,j\}}}(t^l)$ (held in green), factors can be forward propagated (green arrow and in Equation (6.8)) $\mathcal{S}_{\{ij,ji\}}^{6(-)} = \mathbf{M}_{\{ii,jj\}}^{6|2} \mathcal{S}_{\{ij,ji\}}^{1(+)}$ (see Equation (6.9)). After the joint update, a new factor is inserted into the correlation dictionary of the participating nodes, e.g., $\mathsf{P} := \{\mathsf{S}_i, \mathsf{S}_j\}$ by $\mathcal{C}_{\{\mathsf{S}_i, \mathsf{S}_j\}, 6}(\{\mathsf{id}_{\mathsf{S}_j}, \mathsf{id}_{\mathsf{S}_i}\}) = \mathcal{S}_{\{ij,ji\}}^{6(+)}$ (red arrow). A history of beliefs is maintained in $\mathcal{X}_{\mathsf{S}_i}$ and a history of measurements in $\mathcal{Z}_{\mathsf{S}_i}$.

unique ID $\mathsf{id}_{\mathsf{S}_j}$ and a timestamp $t^k$ in the form

$$\mathcal{C}_{\mathsf{S}_i}(\mathsf{id}_{\mathsf{S}_j}, t^k) = \{\mathcal{S}_{ij}^k\}. \tag{6.3}$$

We propose that each node $\mathsf{S}_i$ keeps a sliding time horizon buffer $\mathsf{Hist}$ for the latest beliefs denoted as $\mathcal{X}_{\mathsf{S}_i}$ which is accessed via a timestamp $t^k$ in the form

$$\mathcal{X}_{\mathsf{S}_i}(t^k) = \{\mathbf{x}_i^k\}. \tag{6.4}$$

Similarly, a sliding time horizon buffer $\mathcal{Z}_{\mathsf{S}_i}$ for the latest measurements to support redoing of measurement in their proper sequence (as discussion in Section 2.9.8), which is accessed via a timestamp $t^k$ in the form

$$\mathcal{Z}_{\mathsf{S}_i}(t^k) = \{\mathsf{M}^k\}. \tag{6.5}$$

Please note that, in case of joint observations, only one participant, e.g., the interim master, adds the measurement to the buffer. The measurement data

$$\mathsf{M} := \{t^k, \mathsf{Type}, \mathsf{id}_{\mathsf{S}_i}, \mathbf{z}^k, \mathbf{R}^k\} \tag{6.6}$$

contains beside the measurement vector and measurement noise, a timestamp of the event, the sensor id, and a meta information about the measurement type in $\mathsf{Type}$. $\mathcal{B}_{\mathsf{S}}$ is buffering a history of correction terms which is accessed via a timestamp $t^k$ in the form

$$\mathcal{B}_{\mathsf{S}_i}(t^k) = \{\mathbf{x}_i^k\}. \tag{6.7}$$

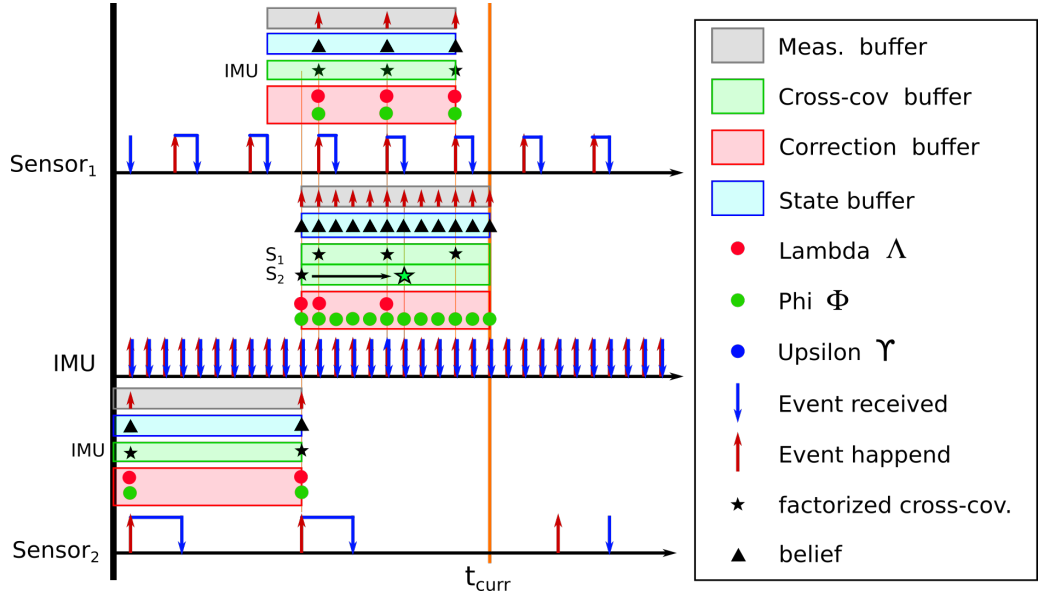

**Figure 6.4:** Sensor activity and buffering: Given three sensors nodes, Sensor$_1$ (S$_1$), Sensor$_2$ (S$_2$) and an IMU (I) for state propagation. S$_1$ and S$_2$ are due to previous joint updates correlated with I. S$_2$ has a low and time-varying rate. At $t_{curr}$ the IMU's time horizon is still reaching to the last joint update with S$_2$. In order not to discard the last known cross-covariance factor $\mathcal{S}_{\mathsf{IS_2}}$ relating to S$_2$, the IMU has to propagate that factor forward in time using elements from correction buffer $\mathcal{B}$ (depicted by a black arrow). Image adapted from [74].

According to [76], we are able to restore an approximated *a priori* cross-covariance $\mathbf{\Sigma}_{ij}^m$ between sensors at the moment they perform a joint observation at $t^k$ by applying their correction buffer's $\mathcal{B}_{\{\mathsf{S}_i, \mathsf{S}_j\}}$ accumulated histories $\mathbf{M}_{\{ii,jj\}}^{k|m}$

$$\mathbf{M}_{\{ii,jj\}}^{k|m} = \prod_{l=m}^{k} \mathcal{B}_{\{\mathsf{S}_i, \mathsf{S}_j\}}(t^l), \tag{6.8}$$

onto their individual factors $\mathcal{S}_{\{ij,ji\}}^m$ from their latest encounter at $t^m$

$$\mathbf{\Sigma}_{ij}^{k(-)} = \left(\mathbf{M}_{ii}^{k|m} \mathcal{S}_{ij}^m\right) \left(\mathbf{M}_{jj}^{k|m} \mathcal{S}_{ji}^m\right)^{\mathsf{T}}. \tag{6.9}$$

**Time horizon**

Note that the time horizon of the nodes' buffers ($\mathcal{C}_{\mathsf{S}_i}$, $\mathcal{X}_{\mathsf{S}_i}$, $\mathcal{Z}_{\mathsf{S}_i}$, and $\mathcal{B}_{\mathsf{S}_i}$) have an impact (i) on the memory footprint, (ii) the maximum allowed sensor delay, and (iii) on the frequency the cross-covariance factors have to be forward propagated.

To reduce the memory footprint, it is desirable to have a fixed time horizon for the buffers. As shown in Figure 6.4, this leads to the challenge that required correction terms within the correction buffer $\mathcal{B}$ can fall out of the past time horizon, before the cross-covariance factors have being used. Consequently those cannot be restored, when they would be needed the next time.

Therefore, the time horizons for each cross-covariance buffer Hist within $\mathcal{C}_{\mathsf{S}}$ have to be checked periodically, e.g., in the propagation steps, in order avoid information loss in the correction buffer by forward-propagating cross-covariance factors (Algorithm 4.5).

This is achieved by finding cross-covariance factors $\mathcal{S}$ that are associated to the last element in the correction buffer $\mathcal{B}$. The oldest element in the $\mathcal{B}$ by

$$t^o = \min(\mathcal{B} < t^k) \tag{6.10}$$

then we compute a time span for forward propagation, e.g., the half of the correction buffers current horizon,

$$t^m = t^o + (\max(\mathcal{B}) - t^o)/2. \tag{6.11}$$

We need to compute a correction factor $\mathbf{M}^{m|o}$ based on the correction buffers $\mathcal{B}$ elements using Equation (6.8).

Now we need to iterate through all $\mathsf{Hist}$ of the factorized cross-covariance dictionary $\mathcal{C}$ and find factorized cross-covariances $\mathcal{S}$ that need to be forward propagated by

$$\{\mathcal{S}_{i,j}^{o(-)}\} = \text{find}(\mathcal{C}(\mathsf{id}_{\mathsf{S}_j}) \equiv t^o), \ \mathsf{id}_{\mathsf{S}_j} \in \mathcal{C} \tag{6.12}$$

and apply the correction factor on all of them, leading to new elements

$$\mathcal{C}(\mathsf{id}_{\mathsf{S}_j}, t^m) = \mathbf{M}^{m|o}\{\mathcal{S}_{i,j}^{o(-)}\}, \ \mathsf{id}_{\mathsf{S}_j} \in \{\mathcal{S}_{i,j}^{o(-)}\}. \tag{6.13}$$

This ensures that the time horizon of the correction buffer $\mathcal{B}$ is always sufficient. While the correction buffer reduces the maintenance effort during the propagation and private updates steps – if many correlations exist – compared to [95, 121].

**Remark 8** *The length of that horizon has an impact on* when *and* where *that propagation happens and how much sensor measurements can be delayed. In the limit of infinitely long time horizons, it always happens in joint updates (when it is actually required). The maximum supported sensor delay is half of the history buffers time horizon, which results from the worst case, when an element was forcibly forward propagated by the half horizon (see Equation (6.11)), in Equation (6.13). If the size of the correction buffer $|\mathcal{B}_\mathsf{S}| \leq 1$, then the corrections are always directly applied on all factorized cross-covariances held in $\mathcal{C}_\mathsf{S}$*

### 6.3.2   Isolated State Propagation

The isolated state propagation of a node between the time instance $t^m$ to $t^k$, is assumed to be corrupted by Gaussian noise. Assuming decoupled inputs, each node independently propagates its belief. The state covariance matrix on $\mathsf{S}_i$ is propagated using

$$\boldsymbol{\Sigma}_{ii}^k = \boldsymbol{\Phi}_{ii}^{k|m}\boldsymbol{\Sigma}_{ii}^m\boldsymbol{\Phi}_{ii}^{k|m\mathsf{T}} + \mathbf{Q}_{ii}^{k|m}, \tag{6.14}$$

with $\boldsymbol{\Phi}^{k|m}$ as the discretized state transition matrix and $\mathbf{Q}^{k|m}$ the discretized process noise covariance.

As the node (participant) $\mathsf{S}_p$ is most likely correlated to other (non-participating) nodes $\mathsf{S}_o \subseteq \mathsf{S}$, these cross-covariances $\boldsymbol{\Sigma}_{po}^m$ need to be propagated as well. To illustrate the cross-covariance between participants ($p$) and non-participants ($o$), we assume a stacked belief

$$\mathbf{x}^m = \begin{bmatrix} \mathbf{x}_p \\ \mathbf{x}_o \end{bmatrix}^m = \mathcal{N}\left( \begin{bmatrix} \hat{\mathbf{x}}_p \\ \hat{\mathbf{x}}_o \end{bmatrix}^m, \begin{bmatrix} \boldsymbol{\Sigma}_{pp} & \boldsymbol{\Sigma}_{po} \\ \boldsymbol{\Sigma}_{po}^\mathsf{T} & \boldsymbol{\Sigma}_{oo} \end{bmatrix}^m \right) \tag{6.15}$$

The propagated stacked belief's uncertainty $\boldsymbol{\Sigma}^m$ from $t^m$ to $t^k$ is

$$\boldsymbol{\Sigma}^k = \begin{bmatrix} \boldsymbol{\Phi}_{pp}^{k|m}\boldsymbol{\Sigma}_{pp}^m\boldsymbol{\Phi}_{pp}^{k|m\mathsf{T}} + \mathbf{Q}_{pp}^{k|m} & \boldsymbol{\Phi}_{pp}^{k|m}\boldsymbol{\Sigma}_{po}^m(\boldsymbol{\Phi}_{oo}^{k|m})^\mathsf{T} + \mathbf{Q}_{po}^{k|m} \\ \bullet & \boldsymbol{\Phi}_{oo}^{k|m}\boldsymbol{\Sigma}_{oo}^m(\boldsymbol{\Phi}_{oo}^{k|m})^\mathsf{T} + \mathbf{Q}_{oo}^{k|m} \end{bmatrix}. \tag{6.16}$$

Assuming no inter-node process noise ($\mathbf{Q}_{po}^{k|m} = \mathbf{0}$) the cross-covariance between a participant and others can be factorized $\boldsymbol{\Sigma}_{po}^m = \mathcal{S}_{po}^m\left(\mathcal{S}_{op}^m\right)^\mathsf{T}$ (see Equation (6.2)).

By applying the factorization of Equation (6.2) on Equation (6.16), the propagation of the stacked uncertainty is

$$\boldsymbol{\Sigma}^k = \begin{bmatrix} \boldsymbol{\Phi}_{pp}^{k|m} \boldsymbol{\Sigma}_{pp}^m \boldsymbol{\Phi}_{pp}^{k|m^\top} + \mathbf{Q}_{pp}^{k|m} & \boldsymbol{\Phi}_{pp}^{k|m} \mathcal{S}_{po}^m \left(\mathcal{S}_{op}^m\right)^\top \left(\boldsymbol{\Phi}_{oo}^{k|m}\right)^\top \\ \bullet & \boldsymbol{\Phi}_{oo}^{k|m} \boldsymbol{\Sigma}_{oo}^m (\boldsymbol{\Phi}_{oo}^{k|m})^\top + \mathbf{Q}_{oo}^{k|m} \end{bmatrix}. \tag{6.17}$$

The factorization allows propagating the cross-covariance between participants and non-participants

$$\boldsymbol{\Sigma}_{po}^k = \mathcal{S}_{po}^k \left(\mathcal{S}_{op}^k\right)^\top = \boldsymbol{\Phi}_{pp}^{k|m} \mathcal{S}_{po}^m \left(\boldsymbol{\Phi}_{oo}^{k|m} \mathcal{S}_{op}^m\right)^\top, \tag{6.18}$$

and can be performed between two time instances $t^m$ and $t^k$ exactly, fully distributed, and isolated on each node by multiplying the state transition matrices $\boldsymbol{\Phi}_{\{pp,oo\}}^{k|m}$ on the corresponding factorized cross-covariances $\mathcal{S}_{\{op,po\}}^m$ [121].

As a state transition matrix $\boldsymbol{\Phi}^{k|m}$ can be the accumulated product of individual transition matrices, $\boldsymbol{\Phi}^{k|m} = \prod_{i=m}^{k} \boldsymbol{\Phi}^i$, the state propagation on individual nodes, can be performed at different rates without loss of generality.

In a node's isolated propagation step, only a subset of the full state's uncertainty is available (missing elements indicated by $-$ and $\widehat{=}$ means *corresponds to*)

$$\breve{\boldsymbol{\Sigma}}^m \widehat{=} \begin{bmatrix} \boldsymbol{\Sigma}_{pp} & \mathcal{S}_{po} \\ - & - \end{bmatrix}^m, \tag{6.19}$$

which can be propagated *exactly* by each node by applying the state transition matrix $\boldsymbol{\Phi}_{pp}^{k|m}$ on their partition of the factorized cross-covariance $\mathcal{S}_{po}^m$

$$\breve{\boldsymbol{\Sigma}}^k \widehat{=} \begin{bmatrix} \boldsymbol{\Phi}_{pp}^{k|m} \boldsymbol{\Sigma}_{pp}^m \boldsymbol{\Phi}_{pp}^{k|m^\top} + \mathbf{Q}_{pp}^{k|m} & \boldsymbol{\Phi}_{pp}^{k|m} \mathcal{S}_{po}^m \\ - & - \end{bmatrix}. \tag{6.20}$$

In order to be modular and scalable, the cross-covariances between $\mathsf{S}_p$ and individual non-participants $\mathsf{S}_{o_i} \in \mathsf{S}_o$ are held in $\mathcal{C}$ and not in a joint block $\boldsymbol{\Sigma}_{po}$. Therefore, applying the state transmission matrix directly on all factors (e.g. [95]) of $\mathcal{C}_p$ (of the node $\mathsf{S}_p$) scales linearly with the number of known and correlated nodes

$$\mathcal{S}_{po}^k = \boldsymbol{\Phi}_{pp}^{k|m} \mathcal{S}_{po}^m, \ o \in \{1 \ldots N | o \not\in p\}. \tag{6.21}$$

To restore cross-covariance factors per node at the moment they are needed (rather than on all factors at each filter step), we chronologically insert the correction term $\boldsymbol{\Phi}_{pp}$ for factorized cross-covariances in the propagation step into the correction buffer $\mathcal{B}$ at $t^k$

$$\mathcal{B}_{\mathsf{S}_p}(t^k) = \boldsymbol{\Phi}_{pp}^{k|m}. \tag{6.22}$$

Once nodes are able to exchange information, the exactly propagated cross-covariance can be restored according to Equation (6.9).

If the propagation step is triggered/driven either by a control input or a proprioceptive sensor measurement, these inputs are stored in measurement buffer $\mathcal{Z}_{\mathsf{S}_p}$.

### 6.3.3 Isolated Observation

An observation is modeled through a measurement function $h$ of a state $\mathbf{x}$ and $\mathbf{v} \sim \mathcal{N}(\mathbf{0}, \mathbf{R})$ defining an independent Gaussian noise

$$\mathbf{z}^m = h(\mathbf{x}^m) + \mathbf{v}^m. \tag{6.23}$$

For legibility, we will neglect the measurement time index $\{\}^m$ and assume that a *a-priori* belief exist at $t^m$. Conceptually, private and joint observations are the same, while the latter is coupling multiples nodes' outputs, i.e., it requires estimates from multiple nodes.

Similar to Equation (6.15), we assume a stacked belief $\mathbf{x} = \begin{bmatrix} \mathbf{x}_p; \ \mathbf{x}_o \end{bmatrix}$. In this case, the measurement is described by a nonlinear function. The joint measurement sensitivity matrix $\mathbf{H} = \begin{bmatrix} \mathbf{H}_p, \ \mathbf{H}_o \end{bmatrix}$, with measurement Jacobian of participants $\mathbf{H}_p$ linearized at the joint *a priori* estimate $\hat{\mathbf{x}}^{(-)}$

$$\mathbf{H}_p = \left[ \frac{\partial h(\mathbf{x})}{\partial \mathbf{x}_p} \big|_{\hat{\mathbf{x}}} \right]^{(-)}. \tag{6.24}$$

Following the definition of non-participating estimators, their beliefs are not involved in the current observation and thus their measurement Jacobian $\mathbf{H}_o$ is

$$\mathbf{H}_o = \left[ \frac{\partial h(\mathbf{x})}{\partial \mathbf{x}_o} \big|_{\hat{\mathbf{x}}} \right]^{(-)} = \mathbf{0}, \tag{6.25}$$

leading finally to a measurement Jacobian for isolated observations

$$\mathbf{H} = \left[ \left[ \frac{\partial h(\mathbf{x}_p)}{\partial \mathbf{x}_p} \big|_{\hat{\mathbf{x}}_p^{(-)}} \right] \quad \mathbf{0} \right]. \tag{6.26}$$

Therefore, the corrected stacked *a posteriori* covariance for participating and non-participating estimators is

$$\mathbf{\Sigma}^{(+)} = \begin{bmatrix} (\mathbf{I} - \mathbf{K}_p \mathbf{H}_p) \mathbf{\Sigma}_{pp}^{(-)} & (\mathbf{I} - \mathbf{K}_p \mathbf{H}_p) \mathbf{\Sigma}_{po}^{(-)} \\ (\mathbf{\Sigma}_{po}^{(+)})^{\mathsf{T}} & -\mathbf{K}_o \mathbf{H}_p \mathbf{\Sigma}_{po}^{(-)} + \mathbf{\Sigma}_{oo}^{(-)} \end{bmatrix} \tag{6.27}$$

with Kalman gain

$$\begin{bmatrix} \mathbf{K}_p \\ \mathbf{K}_o \end{bmatrix} = \begin{bmatrix} \mathbf{\Sigma}_{pp}^{(-)} \mathbf{H}_p^{\mathsf{T}} \\ \mathbf{\Sigma}_{po}^{(-)} \mathbf{H}_p^{\mathsf{T}} \end{bmatrix} \left( \mathbf{H}_p \mathbf{\Sigma}_{pp}^{(-)} \mathbf{H}_p^{\mathsf{T}} + \mathbf{R} \right)^{-1}, \tag{6.28}$$

and $\mathbf{R}$ being the measurement noise covariance. The *a posteriori* mean is

$$\begin{bmatrix} \hat{\mathbf{x}}_p^{(+)} \\ \hat{\mathbf{x}}_o^{(+)} \end{bmatrix} = \begin{bmatrix} \hat{\mathbf{x}}_p^{(-)} \boxplus \mathbf{K}_p \mathbf{r} \\ \hat{\mathbf{x}}_o^{(-)} \boxplus \mathbf{K}_o \mathbf{r} \end{bmatrix}, \tag{6.29}$$

with measurement residual $\mathbf{r} = \boxminus \hat{\mathbf{z}} \boxplus \mathbf{z}^3$.

Note that the impact of a cross-covariance between participants and non-participants $\mathbf{\Sigma}_{po}^{(-)}$ on the Kalman gain $\mathbf{K}_o$ can be seen in Equation (6.28).

In order to achieve an isolated observation among the participants and to avoid information exchange with non-participants, Luft *et al.* proposed in [95] to neglect non-participants' beliefs, which is equal to assume no Kalman gain for non-participants, $\mathbf{K}_o = \mathbf{0}$, and concurs with the idea of the Schmidt-Kalman filter to account for correlated but unknown constant states [129]. This assumption leads to the approximated *a posteriori* belief for non-participants

$$\check{\mathbf{x}}_o^{(+)} \sim \mathcal{N} \left( \check{\mathbf{x}}_o^{(+)}, \check{\mathbf{\Sigma}}_{oo}^{(+)} \right) = \mathbf{x}_o^{(-)} \sim \mathcal{N} \left( \check{\mathbf{x}}_o^{(-)}, \check{\mathbf{\Sigma}}_{oo}^{(-)} \right). \tag{6.30}$$

In consequence, non-participants do not benefit from the participants' observations as they obtain no corrections.

However, participants need to correct the *a priori* cross-covariance referring to non-participants, $\mathbf{\Sigma}_{po}^{(-)}$, which brings us to the differentiation between private and joint observations in the following subsections.

The measurement is stored in one participant's buffer $\mathcal{Z}_{\mathsf{S}_i}$, e.g., the interim master $\mathsf{S}_i$.

---

[3] According to the error-definition in our ESEKF formulation $\tilde{\mathbf{x}} = \boxminus \hat{\mathbf{x}} \boxplus \mathbf{x}$.

**Isolated private observations**

In an isolated private observation, which relates the observation to the estimate of a single node, only a subset of the full states' covariance is available

$$\breve{\boldsymbol{\Sigma}}^{(-)} \cong \begin{bmatrix} \boldsymbol{\Sigma}_{pp} & \mathcal{S}_{po} \\ - & - \end{bmatrix}^{(-)} \tag{6.31}$$

The participant can correct the cross-covariance factor with respect to non-participants $\mathcal{S}_{po}^{(+)}$ *exactly* by observing $\boldsymbol{\Sigma}_{po}^{(-)}$ in Equation (6.27) with

$$\mathcal{S}_{po}^{(+)} = \left(\mathbf{I} - \mathbf{K}_p \mathbf{H}_p\right) \mathcal{S}_{po}^{(-)}. \tag{6.32}$$

This leads to the correction factor for (isolated) private observations

$$\boldsymbol{\Upsilon}_{pp} = \left(\mathbf{I} - \mathbf{K}_p \mathbf{H}_p\right) = \boldsymbol{\Sigma}_{pp}^{(+)} \left(\boldsymbol{\Sigma}_{pp}^{(-)}\right)^{-1}, \tag{6.33}$$

from $\boldsymbol{\Sigma}_{pp}^{(+)} = \left(\mathbf{I} - \mathbf{K}_p \mathbf{H}_p\right) \boldsymbol{\Sigma}_{pp}^{(-)} \equiv \boldsymbol{\Upsilon}_{pp} \boldsymbol{\Sigma}_{pp}^{(-)}$.

Note, in a recursive filter, an estimate is always first predicted then corrected. Therefore, a buffer element for the observations at $t^m$ already exists as shown in Figure 6.3 and means that $\mathcal{B}_{\mathsf{S}_p}(t^m) \neq \mathbf{0}$. Thus, the correction factor $\boldsymbol{\Upsilon}$ has to be left-multiplied on that element in order to inject information from the current observation into existing buffer elements.

$$\mathcal{B}_{\mathsf{S}_p}(t^m) = \boldsymbol{\Upsilon}_{pp}^m \mathcal{B}_{\mathsf{S}_p}(t^m). \tag{6.34}$$

By considering this correction term for cross-covariances relating to non-participants, the approximated full covariance matrix neglecting updates on non-participating beliefs $\breve{\boldsymbol{\Sigma}}^{(+)} \in \mathbb{S}_+$ is conservative compared to the exact/optimal joint *a posteriori* covariance $\boldsymbol{\Sigma}^{(+)}$ (6.27), but remains credible as the resulting covariance error is positive semi-definite [35]

$$\tilde{\boldsymbol{\Sigma}}^{(+)} = \breve{\boldsymbol{\Sigma}}^{(+)} - \boldsymbol{\Sigma}^{(+)} = \begin{bmatrix} \mathbf{0} & \mathbf{0} \\ \mathbf{0} & \mathbf{K}_p \mathbf{H}_p \boldsymbol{\Sigma}_{po}^{(-)} \end{bmatrix} \in \mathbb{S}_+ \succeq \mathbf{0}. \tag{6.35}$$

**Isolated joint observations**

To derive the corrections for cross-covariances of non-participants $\boldsymbol{\Sigma}_{po}^{(+)}$ for each participant, we need to consider the individual block matrix terms. Assuming two participants $\mathsf{S}_i$ and $\mathsf{S}_j$, and others $\mathbf{S}_o \subseteq \mathbf{S}$ leads to a stacked belief $\mathbf{x}^\mathsf{T} = \begin{bmatrix} \mathbf{x}_i^\mathsf{T} & \mathbf{x}_j^\mathsf{T} & \mathbf{x}_o^\mathsf{T} \end{bmatrix}$ with a joint covariance

$$\boldsymbol{\Sigma} = \begin{bmatrix} \boldsymbol{\Sigma}_{ii} & \boldsymbol{\Sigma}_{ij} & \boldsymbol{\Sigma}_{io} \\ \bullet & \boldsymbol{\Sigma}_{jj} & \boldsymbol{\Sigma}_{jo} \\ \bullet & \bullet & \boldsymbol{\Sigma}_{oo} \end{bmatrix} \tag{6.36}$$

and the joint measurement matrix $\mathbf{H} = \begin{bmatrix} \mathbf{H}_i & \mathbf{H}_j & \mathbf{H}_o \end{bmatrix}$, with $\mathbf{H}_o = \mathbf{0}$ as the corresponding beliefs are not involved in the current observations. The Kalman gain for the full state is

$$\mathbf{K} = \boldsymbol{\Sigma}^{(-)} \left(\mathbf{H}\right)^\mathsf{T} \left(\mathbf{S}\right)^{-1} \tag{6.37}$$

with the innovation uncertainty $\mathbf{S} = \mathbf{H} \boldsymbol{\Sigma}^{(-)} \left(\mathbf{H}\right)^\mathsf{T} + \mathbf{R}$. The corresponding *a posteriori* joint covariance is obtained by

$$\boldsymbol{\Sigma}^{(+)} = \left(\mathbf{I} - \mathbf{K} \mathbf{H}\right) \boldsymbol{\Sigma}^{(-)}. \tag{6.38}$$

As $\mathbf{H}_o = \mathbf{0}$ (refer to Equation (6.25)), one can simplify the individual matrices and the innovation uncertainty becomes

$$
\begin{aligned}
\mathbf{S} &= \begin{bmatrix} \mathbf{H}_i & \mathbf{H}_j & \mathbf{0} \end{bmatrix} \mathbf{\Sigma}^{(-)} \begin{bmatrix} \mathbf{H}_i; & \mathbf{H}_j; & \mathbf{0} \end{bmatrix} + \mathbf{R} \\
&= \left( \mathbf{H}_i \mathbf{\Sigma}_{ii}^{(-)} + \mathbf{H}_j \mathbf{\Sigma}_{ji}^{(-)} \right) \mathbf{H}_i^\mathsf{T} + \left( \mathbf{H}_i \mathbf{\Sigma}_{ij}^{(-)} + \mathbf{H}_j \mathbf{\Sigma}_{jj}^{(-)} \right) \mathbf{H}_j^\mathsf{T} + \mathbf{R},
\end{aligned}
\tag{6.39}
$$

thus, the Kalman gain

$$
\mathbf{K} = \begin{bmatrix} \mathbf{K}_i \\ \mathbf{K}_j \\ \mathbf{K}_o \end{bmatrix} = \begin{bmatrix} \mathbf{\Sigma}_{ii} \mathbf{H}_i^\mathsf{T} + \mathbf{\Sigma}_{ij} \mathbf{H}_j^\mathsf{T} \\ \mathbf{\Sigma}_{ji} \mathbf{H}_i^\mathsf{T} + \mathbf{\Sigma}_{jj} \mathbf{H}_j^\mathsf{T} \\ \mathbf{\Sigma}_{oi} \mathbf{H}_i^\mathsf{T} + \mathbf{\Sigma}_{oj} \mathbf{H}_j^\mathsf{T} \end{bmatrix} (\mathbf{S})^{-1}.
\tag{6.40}
$$

As nodes are maintaining only factorized cross-covariances and communication to others $\mathbf{S}_o$ should be avoided, only a subset of the joint uncertainty in Equation (6.36) is available at the moment of the isolated joint observation

$$
\begin{bmatrix} \mathbf{\Sigma}_{ii} & \mathcal{S}_{ij} & \mathcal{S}_{io} \\ \mathcal{S}_{ji} & \mathbf{\Sigma}_{jj} & \mathcal{S}_{jo} \\ - & - & - \end{bmatrix}^{(-)} \ \widehat{=}\ \begin{bmatrix} \mathbf{\Sigma}_{ii} & \mathbf{\Sigma}_{ij} & \mathcal{S}_{io} \\ \bullet & \mathbf{\Sigma}_{jj} & \mathcal{S}_{jo} \\ - & - & - \end{bmatrix}^{(-)} \ \widehat{=}\ \breve{\mathbf{\Sigma}}^{(-)}.
\tag{6.41}
$$

The *a posteriori* covariance of participants $\mathbf{\Sigma}_{pp}^{(+)} = \begin{bmatrix} \mathbf{\Sigma}_{ii} & \mathbf{\Sigma}_{ij} \\ \bullet & \mathbf{\Sigma}_{jj} \end{bmatrix}^{(+)}$ can be calculated exactly

$$
\mathbf{\Sigma}_{ii}^{(+)} = \left[ (\mathbf{I} - \mathbf{K}_i \mathbf{H}_i) \mathbf{\Sigma}_{ii}^{(-)} - \mathbf{K}_i \mathbf{H}_j \mathbf{\Sigma}_{ji}^{(-)} \right],
\tag{6.42}
$$

$$
\mathbf{\Sigma}_{jj}^{(+)} = \left[ (\mathbf{I} - \mathbf{K}_j \mathbf{H}_j) \mathbf{\Sigma}_{jj}^{(-)} - \mathbf{K}_j \mathbf{H}_i \mathbf{\Sigma}_{ij}^{(-)} \right],
\tag{6.43}
$$

$$
\mathbf{\Sigma}_{ij}^{(+)} = \left[ (\mathbf{I} - \mathbf{K}_i \mathbf{H}_i) \mathbf{\Sigma}_{ij}^{(-)} - \mathbf{K}_i \mathbf{H}_j \mathbf{\Sigma}_{jj}^{(-)} \right].
\tag{6.44}
$$

and $\mathbf{\Sigma}_{ij}^k = \mathcal{S}_{ij}^k \left( \mathcal{S}_{ji}^k \right)^\mathsf{T}$, (see Equation (6.2)).

The exact *a posteriori* cross-covariance between the participant and non-participants $\mathbf{\Sigma}_{\{i,j\}o}^{(+)}$ would be

$$
\mathbf{\Sigma}_{io}^{(+)} = (\mathbf{I} - \mathbf{K}_i \mathbf{H}_i) \mathbf{\Sigma}_{io}^{(-)} + \mathbf{K}_i \mathbf{H}_j \mathbf{\Sigma}_{jo}^{(-)},
\tag{6.45}
$$

$$
\mathbf{\Sigma}_{jo}^{(+)} = (\mathbf{I} - \mathbf{K}_j \mathbf{H}_j) \mathbf{\Sigma}_{jo}^{(-)} + \mathbf{K}_j \mathbf{H}_i \mathbf{\Sigma}_{io}^{(-)}.
\tag{6.46}
$$

In an isolated/decoupled update, the *a priori* cross-covariances $\mathbf{\Sigma}_{io}^{(-)}$ and $\mathbf{\Sigma}_{jo}^{(-)}$ are not available, since we have only the factorized cross-covariance $\mathcal{S}_{\{i,j\}o}^{(-)}$ of the participants relating to the non-participants, as shown in Equation (6.41).

Therefore, Luft *et al.* proposed in [95] to express the relation between $\mathbf{\Sigma}_{jo}^{(-)}$ and $\mathbf{\Sigma}_{io}^{(-)}$ as

$$
\mathbf{\Sigma}_{jo}^{(-)} \approx \breve{\mathbf{\Sigma}}_{jo}^{(-)} = \mathbf{\Sigma}_{ji}^{(-)} (\mathbf{\Sigma}_{ii}^{(-)})^{-1} \mathbf{\Sigma}_{io}^{(-)},
\tag{6.47}
$$

$$
\mathbf{\Sigma}_{io}^{(-)} \approx \breve{\mathbf{\Sigma}}_{io}^{(-)} = \mathbf{\Sigma}_{ij}^{(-)} (\mathbf{\Sigma}_{jj}^{(-)})^{-1} \mathbf{\Sigma}_{jo}^{(-)}.
\tag{6.48}
$$

Similarly, Georgescu *et al.* derived in [47] an explicit solution to the problem of completing a partially specified symmetric positive-definite block matrix that maximizes its determinant, in order to complete a partially defined correlation matrix.

The maximal determinant completion is equivalent to the positive semidefinite matrix selection problem and requires solving a convex optimization problem on $\mathbb{S}_+$.

The explicit solution for a symmetric block matrix $\mathbf{S} \in \mathbb{S}$ with an unknown block element $\mathbf{U}_{ik}$

$$\mathbf{S}(\mathbf{U}_{ik}) = \begin{bmatrix} \mathbf{S}_{ii} & \mathbf{S}_{ij} & \mathbf{U}_{ik} \\ \mathbf{S}_{ij}^\mathsf{T} & \mathbf{S}_{jj} & \mathbf{S}_{jk} \\ \mathbf{U}_{ik}^\mathsf{T} & \mathbf{S}_{jk}^\mathsf{T} & \mathbf{S}_{kk} \end{bmatrix} \tag{6.49}$$

that maximizes its determinant (note that minimizing the negative of a determinant is equal to maximizing it)

$$\hat{\mathbf{U}}_{ik} = \underset{\mathbf{U}_{ik}}{\arg\min} -\det(\mathbf{S}(\mathbf{U}_{ik})) \ \ \text{subject to} \ \ \mathbf{S}(\mathbf{U}_{ik}) \in \mathbb{S}_+ \tag{6.50}$$

is given by

$$\hat{\mathbf{U}}_{ik} = \mathbf{S}_{ij}\mathbf{S}_{jj}^{-1}\mathbf{S}_{jk}. \tag{6.51}$$

The maximum determinant completion of a symmetric positive-definite block matrix results in relevant properties [47], such as (i) if a positive semi-definite completion exists, then there exists only one solution, meaning that a unique symmetric positive-definite matrix is obtained, and (ii) the maximum determinant completion equals the maximum entropy completion on a covariance matrix for a multivariate normal model, meaning that it results in the maximally uncertain/inflated/pessimistic covariance and that the block element of its information matrix is $(\mathbf{S}(\mathbf{U}_{ik}))_{ik}^{-1} = \mathbf{0}$.

**Lemma 5** *The approximation proposed by Luft et al. in [95] for the cross-covariance between individual participants and non-participants equals to the maximum determinant completion of a positive semidefinite block matrix, such that the a priori stacked joint covariance is assumed to be maximally uncertain with respect to the cross-covariances to non-participants at the moment of a joint update between participants.*

    ***Proof:*** It can be easily proven that Luft's approximation in Equation (6.47) equals to Equation (6.51) by swapping the participants $i$ and $j$ in the stacked belief leading to $\mathbf{x}' = \begin{bmatrix} \mathbf{x}_j; & \mathbf{x}_i; & \mathbf{x}_o \end{bmatrix}$. ∎

Inserting the maximum determinant completion $\breve{\boldsymbol{\Sigma}}_{jo}^{(-)}$ from Equation (6.47) into Equation (6.45) leads to an approximated *a posteriori* cross-covariance $\breve{\boldsymbol{\Sigma}}_{io}^{(+)}$

$$\begin{aligned} \boldsymbol{\Sigma}_{io}^{(+)} &\approx \breve{\boldsymbol{\Sigma}}_{io}^{(+)} \\ &= (\mathbf{I} - \mathbf{K}_i\mathbf{H}_i)\boldsymbol{\Sigma}_{io}^{(-)} + \mathbf{K}_i\mathbf{H}_j\breve{\boldsymbol{\Sigma}}_{jo}^{(-)} \\ &= (\mathbf{I} - \mathbf{K}_i\mathbf{H}_i)\boldsymbol{\Sigma}_{io}^{(-)} + \mathbf{K}_i\mathbf{H}_j\boldsymbol{\Sigma}_{ji}^{(-)}\left(\boldsymbol{\Sigma}_{ii}^{(-)}\right)^{-1}\boldsymbol{\Sigma}_{io}^{(-)}. \end{aligned} \tag{6.52}$$

Solving for $\boldsymbol{\Sigma}_{ji}^{(-)}$ in Equation (6.42) and inserting into Equation (6.52) leads to the *a posteriori* cross-covariance between the participant $i$ and others $o$

$$\boldsymbol{\Sigma}_{io}^{(+)} \approx \breve{\boldsymbol{\Sigma}}_{io}^{(+)} = \boldsymbol{\Sigma}_{ii}^{(+)}\left(\boldsymbol{\Sigma}_{ii}^{(-)}\right)^{-1}\boldsymbol{\Sigma}_{io}^{(-)}, \tag{6.53}$$

which can be generalized to a *generic* approximation for correcting cross-covariances between participants and non-participants in isolated joint updates

$$\boldsymbol{\Sigma}_{uv}^{(+)} \approx \boldsymbol{\breve{\Sigma}}_{uv}^{(+)} = \boldsymbol{\Sigma}_{uu}^{(+)} \left(\boldsymbol{\Sigma}_{uu}^{(-)}\right)^{-1} \boldsymbol{\Sigma}_{uv}^{(-)}, u \in \mathbf{P}, v \in \bar{\mathbf{P}}, \tag{6.54}$$

and allows us to define the joint correction term $\boldsymbol{\Lambda}$ for each participant $\mathsf{S}_u$

$$\boldsymbol{\Lambda}_{uu} = \boldsymbol{\Sigma}_{uu}^{(+)} \left(\boldsymbol{\Sigma}_{uu}^{(-)}\right)^{-1}, \ u \in \mathbf{P}. \tag{6.55}$$

Again, the correction factor of each participant $u \in \mathbf{P}$ for joint updates $\boldsymbol{\Lambda}_{uu}$ has to be left-multiplied on the participants' correction buffers

$$\boldsymbol{\mathcal{B}}_{\mathsf{S}_u}\left(t^k\right) = \boldsymbol{\Lambda}_{uu}^k \boldsymbol{\mathcal{B}}_{\mathsf{S}_u}\left(t^k\right), \ u \in \mathbf{P}. \tag{6.56}$$

At this point it is worth mentioning that, first, the correction term **implicitly** assumes the maximum determinant completion, meaning (a) it *is not explicitly computed*, (b) the stacked joint *a-priori* covariance $\boldsymbol{\breve{\Sigma}}_p^{(-)} \in \mathbb{S}_+$ of participants is positive semidefinite as the correction terms applied to the participants cross-covariance factors $\mathcal{S}$ stem from a maximum determinant completion, and (c) the stacked joint *a-posteriori* covariance $\boldsymbol{\breve{\Sigma}}_p^{(+)} \in \mathbb{S}_+$ of participants remains positive semidefinite. Second, each participant needs to compute only one common correction term $\boldsymbol{\Lambda}$ which is valid for **all** its cross-covariance factors. Third, the *a-posteriori* covariance of participants can be computed using the numerically more stable Joseph's form[4] providing better numerical stability to the computation of the elements in the correction buffer. Fourth, the correction term is equal to the correction term defined for private observations in Equation (6.33).

Similarly to private isolated observation and according to the Schmidt-Kalman filter, we assume the Kalman gain for non-participants to be $\mathbf{K}_o = \mathbf{0}$ leading finally to the approximated *a-posteriori* joint covariance $\boldsymbol{\breve{\Sigma}}^{(+)}$ of the full state

$$\boldsymbol{\Sigma}^{(+)} \approx \boldsymbol{\breve{\Sigma}}^{(+)} = \begin{bmatrix} \boldsymbol{\Sigma}_{ii}^{(+)} & \boldsymbol{\Sigma}_{ij}^{(+)} & \boldsymbol{\breve{\Sigma}}_{io}^{(+)} \\ \bullet & \boldsymbol{\Sigma}_{jj}^{(+)} & \boldsymbol{\breve{\Sigma}}_{jo}^{(+)} \\ \bullet & \bullet & \boldsymbol{\Sigma}_{oo}^{(-)} \end{bmatrix} \notin \mathbb{S}_+. \tag{6.57}$$

This leads in general to a non-positive semidefinite approximated *a-posteriori* global covariance $\boldsymbol{\breve{\Sigma}}^{(+)}$ and to two downsides of the proposed IKF scheme. First, non-participants do not become (indirectly) correlated to a participant through other participants' cross-correlations [95]. For instance, by observing Equation (6.45), if $\boldsymbol{\Sigma}_{io}^{(-)} = \mathbf{0}$ and $\boldsymbol{\Sigma}_{jo}^{(-)} \neq \mathbf{0}$, then, in an exact update an (indirect) correlation would be obtained by $\boldsymbol{\Sigma}_{io}^{(+)} \neq \mathbf{0}$. Since, Equation (6.55) is applied just on existing cross-covariance factors, these indirect correlations through other participants are neglected. In other words, correlations can **only** be obtained among participants of joint observations. Second, correlated non-participants obtain no correction[5]. These approximations are a cost to be paid to obtain much better scalability while still preserving credible estimates in various scenarios (see Section 6.4).

---

[4]The Joseph's form of the *a posteriori* covariance $\boldsymbol{\Sigma}^{(+)} = (\mathbf{I} - \mathbf{KH})\boldsymbol{\Sigma}^{(-)}(\mathbf{I} - \mathbf{KH})^{\mathsf{T}} + \mathbf{KRK}^{\mathsf{T}}$ is known to be numerically more stable than the simplified form $\boldsymbol{\Sigma}^{(+)} = (\mathbf{I} - \mathbf{KH})\boldsymbol{\Sigma}^{(-)}$.

[5]Worst case are totally correlated non-participants, e.g., if the belief of a participant $\mathsf{S}_i$ is a stochastic clone of a non-participant $\mathsf{S}_j$, with $\boldsymbol{\Sigma}_{ii} \equiv \boldsymbol{\Sigma}_{jj}$ and $\boldsymbol{\Sigma}_{ij} = \boldsymbol{\Sigma}_{ii}$, as any correction on $\mathsf{S}_i$ or $\mathsf{S}_j$ is influencing the belief equivalently, which can be proven easily by observing the individuals' Kalman gains in Equation (6.40) which are in this particular case identically $\mathbf{K}_i = \mathbf{K}_j$ assuming $\mathbf{H}_{ii} = \mathbf{I}$ and $\mathbf{H}_{jj} = \mathbf{0}$ or vice versa.

### 6.3.4  Delayed Observations

In Section 6.3.3, we made the assumption, that an *a-priori* belief $\mathbf{x}_p^{m(-)}$ of the participants from the stacked belief $\mathbf{x}^m = \begin{bmatrix} \mathbf{x}_p; \ \mathbf{x}_o \end{bmatrix}^m$ exist for the measurement taken at $t(m)$ in the form

$$\mathbf{z}^m = h(\mathbf{x}^m) + \boldsymbol{\nu}_z^m.$$

As already discussed in Section 2.9.8, sensor data in AINS are typically delayed or processed out-of-sequence. In such cases, we assume that the measurement at $t(m)$ is bounded by two previous beliefs $\mathbf{x}_p^a$ and $\mathbf{x}_p^b$ that are maintained in $\boldsymbol{\mathcal{X}}_i, i \in \mathbf{P}$ of the participants. Consequently, $t^a \leq t^m \leq t^b$ and each participant $\mathsf{S}_p \in \mathbf{P}$ has to create a pseudo/virtual belief at $t^m$.

Figure 6.5 shows the procedure in case of delayed isolated private observations.

For each participant, we need to obtain the closest belief before the delayed event happened

$$\{\hat{\mathbf{x}}_i^a, \boldsymbol{\Sigma}_{ii}^a, t^a\} = \max(\text{find}(\boldsymbol{\mathcal{X}}_i < t^k)), \ i \in \mathbf{P} \tag{6.58}$$

Then we have to predict starting from that previous belief until the event

$$\hat{\mathbf{x}}_i^m, \boldsymbol{\Sigma}_{ii}^m, \boldsymbol{\Phi}^{m|a} = \text{propagate}_i(\hat{\mathbf{x}}_i^a, \boldsymbol{\Sigma}_{ii}^a, \ldots, t^a, t^m), \ i \in \mathbf{P} \tag{6.59}$$

with

$$\{\hat{\mathbf{x}}_i^b, \boldsymbol{\Sigma}_{ii}^b, t^b\} = \min(\text{find}(\boldsymbol{\mathcal{X}}_i > t^k)), \ i \in \mathbf{P} \tag{6.60}$$

and insert the new correction factor into the participants buffers

$$\boldsymbol{\mathcal{B}}_i(t^m) = \boldsymbol{\Phi}_{ii}^{m|a}, \ i \in \mathbf{P}. \tag{6.61}$$

After performing an intermediate prediction step from $t^a$ until $t^m$, the appropriate state transition matrix is inserted into the correction buffer. Note that we will have to continue the propagation later (after the delayed measurement was applied) from the virtual/interpolated state.

Now the exact procedure as for non-delayed isolated observation can be performed, see Section 6.3.3.

After the pseudo belief was corrected (step two in Figure 6.5), we need to delete all entries after the measurement event from the belief history $\boldsymbol{\mathcal{X}}$, factorized cross-covariance dictionary $\boldsymbol{\mathcal{C}}$, and the correction buffer $\boldsymbol{\mathcal{B}}$

$$\text{delete}(\{\boldsymbol{\mathcal{X}}_{\mathsf{S}_i}, \boldsymbol{\mathcal{C}}_{\mathsf{S}_i}, \boldsymbol{\mathcal{B}}_{\mathsf{S}_i}\} > t^m), \ i \in \mathbf{C}, \tag{6.62}$$

with $\mathbf{C}$ being a set of *post-correlated* instances which led to factorized cross-covariances rooted at each participant after the delayed observations

$$\mathbf{C} = \mathbf{C}' \cap \text{unique}(\{\boldsymbol{\mathcal{C}}_{\mathsf{S}_i} \to \text{keys}() > t^m\}, \ i \in \mathbf{C}'), \text{until } \mathbf{C} \equiv \mathbf{C}', \mathbf{C}_0' = \mathbf{P}. \tag{6.63}$$

Note, that the set of post-correlated instances $\mathbf{C}$ can be computed either iteratively starting from the participants until convergence (see above) or recursively.

Then all measurements are reprocessed, beginning from the pseudo state onward using all measurement in all post-correlated instances

$$\mathbf{M}_i := \{\mathsf{M}_i^l\} := \boldsymbol{\mathcal{Z}}_{\mathsf{S}_i} > t^m, \ i \in \mathbf{C}. \tag{6.64}$$

by sorting them chronologically

$$\mathbf{M} = \text{sort}(\{\mathbf{M}_1, \ldots, \mathbf{M}_N\}), \text{with prop} < \text{private} < \text{joint}. \tag{6.65}$$

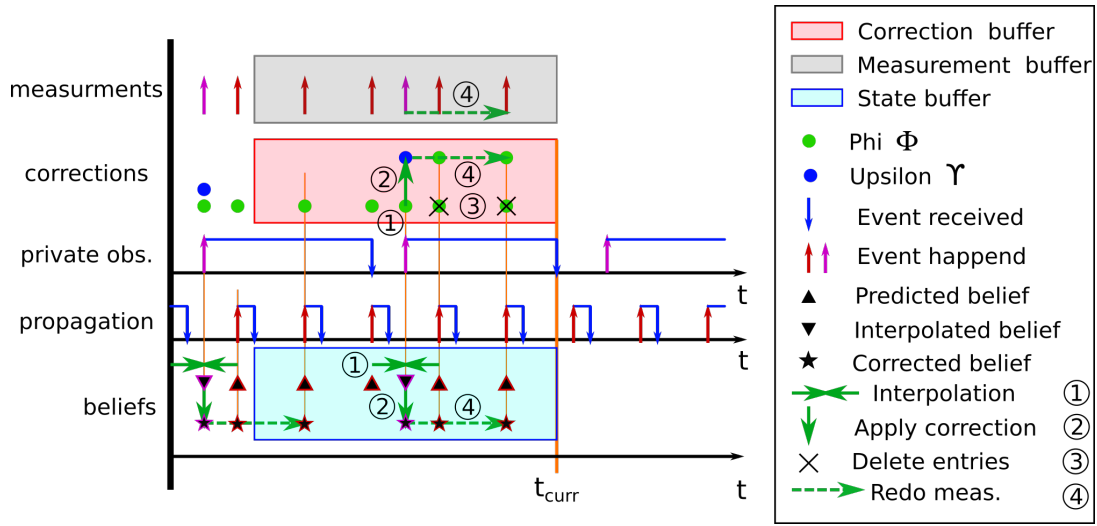

**Figure 6.5:** Delayed private measurement handling in an IKF. Given a state propagation and an exteroceptive sensor for state correction. The exteroceptive sensor provides delayed private measurements at different rates, and it is assumed that the sources for the timestamps are synchronized. At $t_{curr}$, the IKF receives a measurement, which relates to an event that falls between two propagation measurements. In order to fuse that measurement, first, an interpolated (pseudo/virtual) belief needs to be created. Second, the pseudo belief is corrected using the delayed measurement. Third, all elements in the correction and factorized cross-covariance buffers after that event need to be deleted. Lastly, all measurements in the measurement buffer after that event, are reprocessed in order to propagate the obtained correction forward in time. In this example, $\mathcal{C}$ is omitted, but the third and fourth step would affect the history of cross-covariances too.

Since measurements are processed isolated, concurrent measurements need to be sorted by prioritizing propagation over private observations and private over joint observations[6]. In a centralized equivalent fusion architecture, a prioritization of private over joint observation is not needed. This set of measurements $\mathbf{M}$ needs to be reapplied sequentially and will restore an updated and re-linearized version of the previously deleted elements within the buffers.

Processing measurements sequentially in a distributed system can be achieved, e.g., by triggering and synchronizing individual nodes via an interim master node. For instance, the interim master that processed the delayed measurements would need to gather all measurements from all post-correlated nodes (in case of limited communication range, they need to be relayed). After sorting them, measurements are directed to individual nodes (proprioceptive measurements can be processed in parallel) and waits until they complete their task by sending acknowledgments back to the interim master. Consequently, during the course of delayed measurements, in the limit a persistent all-to-all communication is needed. In other words: the communication link complexity of $\mathcal{O}(|\mathsf{P}|)$ for non-delayed isolated joint observations and $\mathcal{O}(1)$ for non-delayed private observations is $\mathcal{O}(|\mathsf{S}|)$ for delayed isolated observations.

The scenario depicted in Figure 6.6, illustrates the measurement history of four filter instances, while at $t = 5$ a delayed joint measurement between $\mathsf{S}_1$ and $\mathsf{S}_2$ is obtained. After processing the delayed measurement, all measurements within the gray box need to be processed synchronously and in an appropriate order (see Equation (6.65)). Note that performing only measurements among participants (highlighted by the green box), would violate correlations introduced with $\mathbf{H}_{3,2}^3$, between $\mathsf{S}_3$ and $\mathsf{S}_2$. $\mathsf{S}_2$ would delete the cross-covariance factor $\mathcal{S}_{23}^3$, while $\mathsf{S}_3$ still maintains the counterpart $\mathcal{S}_{32}^3$. As the observation $\mathbf{H}_{3,2}^3$, is not reprocessed, $\mathsf{S}_3$ would be correlated to a deleted belief of $\mathsf{S}_2$ and the stacked belief $[\mathbf{x}_2; \mathbf{x}_3]$ of $\mathsf{S}_2$ and $\mathsf{S}_3$ cannot be restored in a future joint observation.

---

[6]As already mentioned, some dynamical systems are modeled in a way, that a control input or a sensor measurement is advancing the state in time.

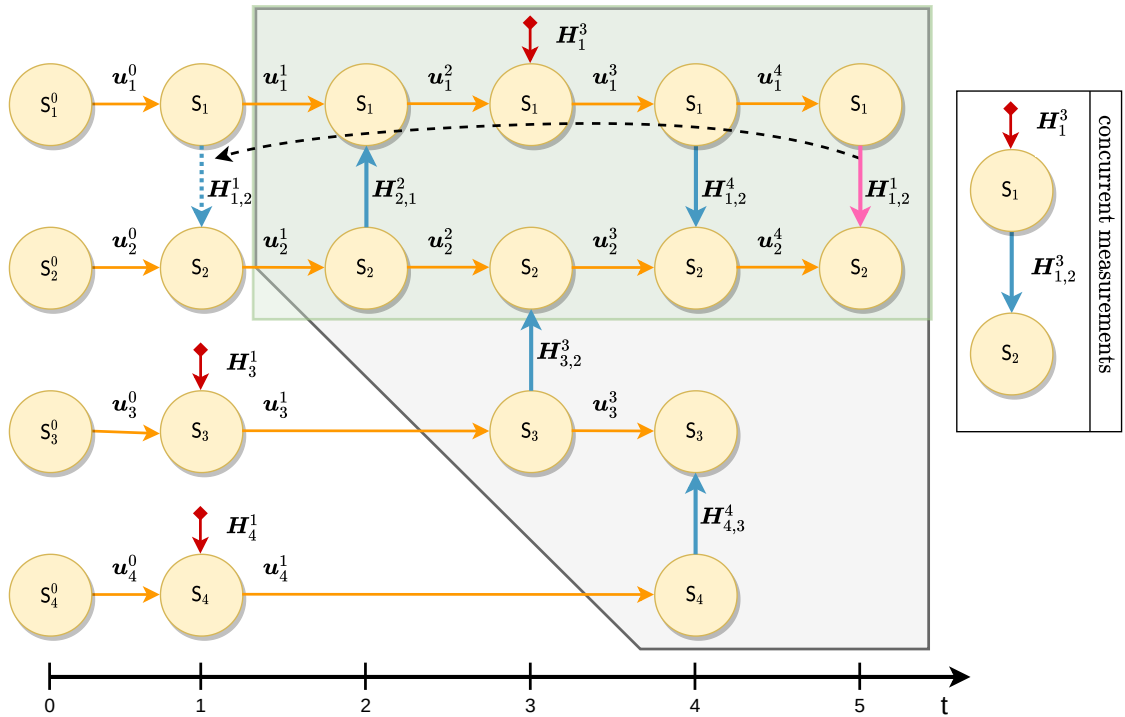

**Figure 6.6:** Shows the measurements obtained on a network of four instances $\{\mathsf{S}_1, \mathsf{S}_2, \mathsf{S}_3, \mathsf{S}_4\}$. Arrows in orange are control inputs $\mathbf{u}$, private observations are held in red, and joint observations in blue. The joint observation at $t = 5$ held in magenta indicates a delayed measurement $\mathbf{H}_{1,2}^1$ originating from $t = 1$. In order to restore the beliefs at $t = 5$, all intermediate measurements after that event (within the gray box) need to be reprocessed in an appropriate order.

### 6.3.5 The IKF library

We made the source of a Isolated Kalman Filtering (IKF) framework (ikf_lib) with support for out-of-sequence measurements available to the public[7]. This library implements the IKF paradigm in abstract classes, where actual estimation problems can derive from. It aims at demonstrating the capabilities and properties of the paradigm. Figure 6.7 shows a simplified UML class diagram of the most relevant objects. The abstract base class IKalmanFilter maintains a history of abstract estimates, IBelief, which provide an interface for actual realizations, and a history of processed measurements MeasData. MeasData objects are used to represent actual measurements by a measurement vector $\mathbf{z}$, measurement covariance $\mathbf{R}$, timestamps relating to the event of perceiving information and when the measurement was received/processed in filter, the corresponding sensor's unique ID, and meta information for debugging and logging purposes. The abstract class IIsolated-KalmanFilter derives from IKalmanFilter and allows processing joint updates isolated, i.e., to couple the outputs of multiple IIsolatedKalmanFilter instances. These are maintained centralized in the IsolatedKalmanFilterHandler, which accesses instances via their unique ID. Measurements can be either provided to the individual instances or the handler.

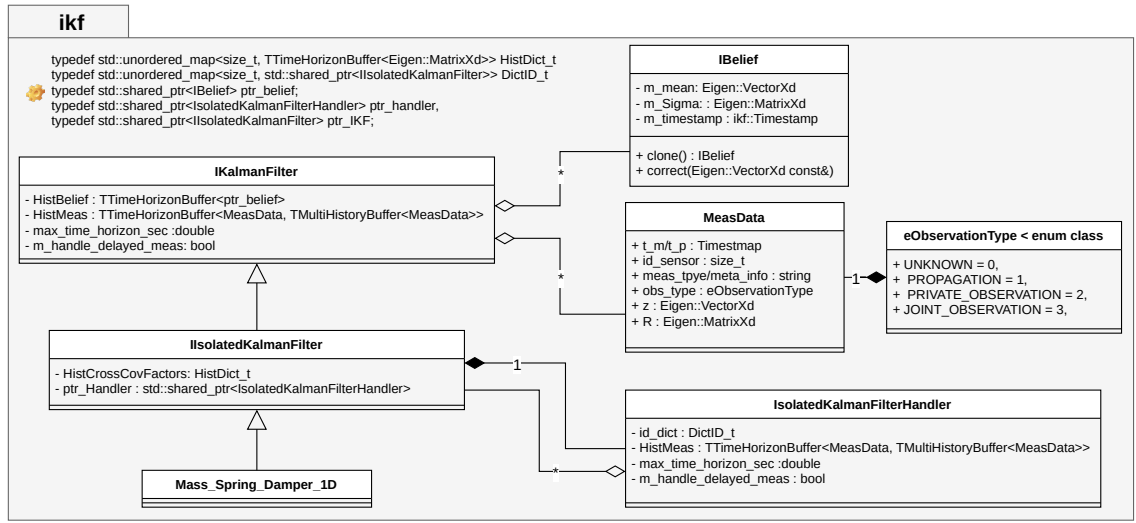

**Figure 6.7:** Shows a simplified UML class diagram of the ikf_lib.

The centralized handler is well suited to modular sensor fusion (see Chapter 4), but not for multi-agent CSE, as all filter instance, the handler, and the simulated data are held in one process. Particular focus was on handling delayed measurements, and to provide a generic and tested implementation of the paradigm.

### 6.3.6  Summary

The IKF allows exact isolated state propagation steps, since we assume decoupled dynamics and no other estimates are required. Isolated private and joint observations are exact among participants, if they are not correlated to non-participants. If participants are correlated to non-participants, the global joint covariance becomes non-positive semidefinite, meaning the gobal belief follows a non-Gaussian distribution, while the *a-priori* and *a-posteriori* stacked joint covariance of participants remains positive semidefinite. If estimators perform an update, it is assumed that non-participants are, per definition, not represented in the measurements (i.e., in the update equations) $\mathbf{H}_o = \mathbf{0}$ (see Equation (6.25)) and that non-participating estimators obtain no correction $\mathbf{K}_o = \mathbf{0}$. Therefore, correlations can only be obtained among participants of isolated observations.

The correlation between two estimators is maintained in two factorized cross-covariances $\mathcal{S}$ (see Equation (6.2)) (each estimator maintains a factor), which obtains corrections from the corresponding estimator ($\mathbf{\Phi}$ for propagation, $\mathbf{\Upsilon}$ for private update, and $\mathbf{\Lambda}$ for joint update). Once estimators perform another joint isolated observation, their cross-covariance can be restored by applying corrections to the cross-covariance factors from their previous encounter (see Figure 6.3). After the joint update, new cross-covariance factors are calculated and stored in each participant's container $\mathcal{C}_{\mathsf{S}_i}, i \in \mathbf{P}$.

In case a measurement is obtained delayed, thus out-of-sequence, the bounding two beliefs of the participant in the past needs to be found in the history of beliefs, $\mathcal{X}_{\mathsf{S}_i}, i \in \mathbf{P}$. Then one can either interpolate or perform a prediction from the older belief, to obtain a interpolated (pseudo) belief, which is corrected by the delayed measurement. Next, all state-related elements in the buffers relating to events after the delayed elements are removed from participants and post-correlated instances (see Equation (6.63)). All measurements after that event in the measurement buffers $\mathcal{Z}_{\mathsf{S}_i}, i \in \mathbf{P} \cap \mathbf{C}$ of participants and post-correlated instances are re-applied synchronously in the correct order. Finally, the delayed measurement is inserted chronologically sorted in the measurement buffer of the interim master.

## 6.4    Evaluation

In this section, both, the steady-state behavior and credibility of the proposed IKF paradigm is evaluated on a mass-spring-damper system described in the following section. The credibility is compared against a full-state Kalman filter and a naive filter. The naive filter neglects correlations between estimators by assuming independence between estimators in case of relative updates.

### 6.4.1    Mass-spring-damper model

Without loss of generality, as toy example, we estimate the position and velocity of an idealized mass-spring-damper system – a typical and well-studied problem – which is attached to the ceiling as described in Figure 6.8. The system can be described using Newton's second law of motion $\sum F = m\ddot{x}$ by

$$m\ddot{x} = \sum F = F_g - F_k - F_c = mg - kx - c\dot{x}, \tag{6.66}$$

with a spring stiffness constant $k$ in N/m, the damping constant $c$ in Ns/m, and the mass $m$ in kg A force $F_g = m * g$ is acting on the mass, with $g = 9.81\,\text{m/s}^2$ being the Earth's gravitational constant. It can be rearranged to

$$\ddot{x} = g - \frac{k}{m} - \frac{c}{m}\dot{x}. \tag{6.67}$$

The corresponding state space model with $\ddot{x} = \dot{v}$, $\dot{x} = \dot{p}$, the state $\mathbf{x} = [p; v]$, and the control input $u = g$ is

$$\begin{bmatrix} \dot{p} \\ \dot{v} \end{bmatrix} = \begin{bmatrix} 0 & 1 \\ -\frac{k}{m} & -\frac{c}{m} \end{bmatrix} \begin{bmatrix} p \\ v \end{bmatrix} + \begin{bmatrix} 0 \\ 1 \end{bmatrix} g, \tag{6.68}$$

$$\dot{\mathbf{x}} = \mathbf{F}\mathbf{x} + \mathbf{B}u, \mathbf{x}(0) = \mathbf{x}_0. \tag{6.69}$$

We assume to observe the distance between the ceiling and the mass, and the relative distance between two masses. The observed outputs are

$$z_i^k = \mathbf{H}_i \mathbf{x}^k = \begin{bmatrix} 1 & 0 \end{bmatrix} \mathbf{x}_i^k, \tag{6.70}$$

$$z_{i,j}^k = \mathbf{H}_{i,j}^k \begin{bmatrix} \mathbf{x}_i \\ \mathbf{x}_j \end{bmatrix}^k = \begin{bmatrix} -1 & 0 & 1 & 0 \end{bmatrix} \begin{bmatrix} \mathbf{x}_i \\ \mathbf{x}_j \end{bmatrix}^k. \tag{6.71}$$

The corresponding discrete-time linear time invariant system with a sample time $\Delta t$ and assuming piece wise constant control input between samples is

$$\begin{bmatrix} p \\ v \end{bmatrix}^{k+1} = \begin{bmatrix} 1 & \Delta t \\ -\Delta t\frac{k}{m} & 1 - \Delta t\frac{c}{m} \end{bmatrix}^k \begin{bmatrix} p \\ v \end{bmatrix}^k + \begin{bmatrix} 0 \\ \Delta t \end{bmatrix}^k g^k, \tag{6.72}$$

$$\mathbf{x}^{k+1} = \mathbf{\Phi}^{k+1|k}\mathbf{x}^k + \mathbf{B}_d^k u^k, \mathbf{x}^0 = \mathbf{x}_0. \tag{6.73}$$

Assuming a random white disturbance $w^k \sim \mathcal{N}\left(0, \sigma_g^2\right)$ on the control input $u^k$ and white noisy measurements $\{z_i, z_{i,j}\}$, we obtain a stochastic process for each mass-spring-

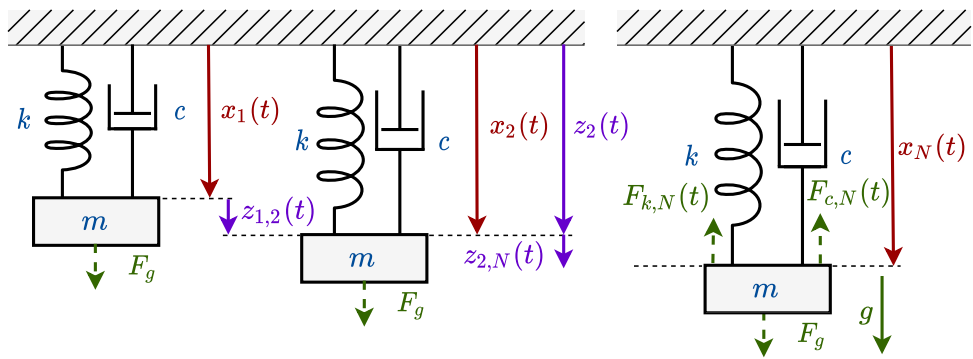

**Figure 6.8:** Shows $N$ homogeneous mass-spring-damper systems attached to the ceiling, with a spring stiffness constant $k$ in N/m, the damping constant $c$ in Ns/m, and the mass $m$ in kg. A constant force $F_g = mg$ is acting on the mass, with $g = 9.81 \, \text{m/s}^2$ being the Earth's gravitational constant. $x(t)$ is the distance from the ceiling to the mass. Arrows held in purple are distance observations from the ceiling to a mass and between vertical positions of the masses.

damper system $i$ in the form

$$\mathbf{x}_i^{k+1} = \boldsymbol{\Phi}_{ii}^{k+1|k}\mathbf{x}_i^k + \mathbf{B}_d^k(u^k + w^k), \mathbf{x}^0 = \mathbf{x}_0, \tag{6.74}$$

$$z_i = \mathbf{H}_i \mathbf{x}_i + v_i^k, \tag{6.75}$$

$$z_{i,j} = \mathbf{H}_{i,j} \begin{bmatrix} \mathbf{x}_i \\ \mathbf{x}_j \end{bmatrix}^k + v_{i,j}^k, \tag{6.76}$$

with $v_i^k \sim \mathcal{N}\left(0, \mathbf{R}_i\right), \mathbf{R}_i = \sigma_i^2, \ v_{i,j}^k \sim \mathcal{N}\left(0, \mathbf{R}_{i,j}\right), \mathbf{R}_{i,j} = \sigma_{i,j}^2$, and the estimated state $\mathbf{x}_i \sim \mathcal{N}\left(\hat{\mathbf{x}}_i, \boldsymbol{\Sigma}_{ii}\right)$. The process noise and measurement noise are not correlated.

The estimated covariance of the linear system is predicted through the state transition matrix and incorporating the control input noise $w^k$

$$\boldsymbol{\Sigma}_{ii}^{k+1} = \boldsymbol{\Phi}_{ii}^{k+1|k}\boldsymbol{\Sigma}_{ii}^k(\boldsymbol{\Phi}_{ii}^{k+1|k})^{\mathsf{T}} + \Delta t \mathbf{B}_d^k \sigma_g^2 (\mathbf{B}_d^k)^{\mathsf{T}} \tag{6.77}$$

and the estimated mean is propagated via

$$\hat{\mathbf{x}}_i^{k+1} = \boldsymbol{\Phi}_{ii}^{k+1|k}\hat{\mathbf{x}}_i^k + \mathbf{B}_d^k g_i^k. \tag{6.78}$$

### 6.4.2 Steady-state behavior

Since the IKF paradigm builds upon approximations which might lead to an inconsistent global full state, we want to assess in this section, under which conditions the IKF paradigm leads to credible estimates in comparison to a centralized Kalman filter implementation. Given a set of three dynamically decoupled nodes, we will study different observations configurations – so call observation graphs – and their expected behavior at steady-state, like Mourikis and Roumeliotis did on relative pose measurement graphs (RPMG) for ground robots in [109]. They show that an upper- and expected bound can be computed by determining the maximal process and measurement or the average process and measurement noise, respectively.

The steady-state behavior of the estimator allows us to draw conclusion to the filter's stability. Note that the stability of the filter does not need a stable the dynamic system[8], but states need to be completely controllable and observable [10].

As covariance equations of the KF are independent of the actual measurements in linear dynamic system, they can be iterated forward in time offline. Meaning the time evolution of the covariance given an initial conditions and known noise characteristics can be assessed, without obtaining actual measurements. The recursion of the covariance using a prediction and update step is known as the *Discrete-time matrix Riccati equation* (DARE)[10]

$$\boldsymbol{\Sigma}^{k+1|k} = \boldsymbol{\Phi}^k \left( \boldsymbol{\Sigma}^{k|k-1} - \boldsymbol{\Sigma}^{k|k-1}(\mathbf{H}^k)^\mathsf{T} \left( \mathbf{H}^k \boldsymbol{\Sigma}^{k|k-1}(\mathbf{H}^k)^\mathsf{T} + \mathbf{R}^k \right)^{-1} \mathbf{H}^k \boldsymbol{\Sigma}^{k|k-1} \right) (\boldsymbol{\Phi}^k)^\mathsf{T} + \mathbf{Q}^k,$$
(6.79)

which translates into our notation, with $k + 1(-) = k + 1|k$ and $k(-) = k|k-1$ by

$$\begin{aligned} \boldsymbol{\Sigma}^{k+1(-)} =& \boldsymbol{\Phi}^{k+1|k} \left( \boldsymbol{\Sigma}^{k(-)} - \boldsymbol{\Sigma}^{k(-)}(\mathbf{H}^k)^\mathsf{T} \left( \mathbf{H}^k \boldsymbol{\Sigma}^{k(-)}(\mathbf{H}^k)^\mathsf{T} + \mathbf{R}^k \right)^{-1} \mathbf{H}^k \boldsymbol{\Sigma}^{k(-)} \right) (\boldsymbol{\Phi}^{k+1|k})^\mathsf{T} \\ &+ \mathbf{Q}^{k+1|k}. \end{aligned}$$
(6.80)

The solution of the DARE for a time-invariant system converges to a finite steady-state covariance $\boldsymbol{\Sigma}^\infty$ provided that $\{\boldsymbol{\Phi}, \mathbf{H}\}$ are completely observable and that $\{\boldsymbol{\Phi}, \mathbf{C}\}$, where $\mathbf{Q} = \mathbf{C}\mathbf{C}^\mathsf{T}$ ($\mathbf{C}$ is the square root of $\mathbf{Q}$) is completely controllable [10]. $\boldsymbol{\Sigma}^\infty$ is a unique solution which is independent of the initial condition/covariance $\boldsymbol{\Sigma}^0$. The steady-state covariance is the solution of the algebraic Riccati equation

$$\boldsymbol{\Sigma}^\infty = \boldsymbol{\Phi} \left( \boldsymbol{\Sigma}^\infty - \boldsymbol{\Sigma}^\infty \mathbf{H}^\mathsf{T} (\mathbf{H}\boldsymbol{\Sigma}^\infty\mathbf{H}^\mathsf{T} + \mathbf{R})^{-1} \mathbf{H}\boldsymbol{\Sigma}^\infty \right) \boldsymbol{\Phi}^\mathsf{T} + \mathbf{Q},$$
(6.81)

which can be solved directly or iteratively, and leads to a steady-state Kalman gain $\mathbf{K}_\infty = \boldsymbol{\Sigma}^\infty \mathbf{H}^\mathsf{T} (\mathbf{H}\boldsymbol{\Sigma}^\infty\mathbf{H}^\mathsf{T} + \mathbf{R})^{-1}$.

The Ricatti equation can have a stable (steady-state) solution, even if the dynamic of the system is unstable, assuming that the system is observable and controllable. A sufficient condition to proof controllability is a full rank in the controllability matrix $\text{Rank}(\boldsymbol{\mathcal{C}}) = n$ [50] with

$$\boldsymbol{\mathcal{C}} = [\boldsymbol{B_d}^0, \boldsymbol{\Phi}^{1|0}\boldsymbol{B_d}^1, \dots, \boldsymbol{\Phi}^{k|k-1}\boldsymbol{B_d}^{k-1}],$$
(6.82)

which states that the process noise enters into each state component and prevents the uncertainty of the estimated state from converging to zero and causes the covariance to be positive definite [10]. Similarly, the observability matrix needs to have full rank $\text{Rank}(\boldsymbol{\mathcal{O}}) = n$ with

$$\boldsymbol{\mathcal{O}} = [\boldsymbol{H}^0; \boldsymbol{H}^1\boldsymbol{\Phi}^{1|0}; \dots; \boldsymbol{H}^k\boldsymbol{\Phi}^{k|k-1}],$$
(6.83)

which guarantees an information flow about each state component and prevents the uncertainty to grow unbounded. This condition yields the existence of steady-state solution $\boldsymbol{\Sigma}^\infty$ that is positive definite $\mathbb{S}_{++}$ or positive semidefinite $\mathbb{S}_+$ [10].

In the absence of measurements (open-loop, with $\mathbf{H} = \mathbf{0}$ or $\mathbf{R}^{-1} = \mathbf{0}$), a linear discrete-time dynamic system has no output that is observed, meaning it is a stochastic process, e.g., in the form

$$\mathbf{x}^{k+1} = \mathbf{A}\mathbf{x}^k + \mathbf{w}^k,$$
(6.84)

---

[8]A discrete-time dynamic system is stable and only stable, if the (discrete-time) state transition matrix $\boldsymbol{\Phi}$ is stable. This is the case, when all (complex) eigenvalues/poles are within the unit circle, i.e., the norm of eigenvalues is smaller than 1. If a single eigenvalue is greater one, the dynamic system is unstable.

with $\mathbf{x} \sim \mathcal{N}(\hat{\mathbf{x}}, \boldsymbol{\Sigma})$ and $\mathbf{w} \sim \mathcal{N}(\mathbf{0}, \mathbf{W})$ being random but uncorrelated variables. The uncertainty evolves according to the Lyapunov differential equation

$$\boldsymbol{\Sigma}^{k+1} = \mathbf{A}\boldsymbol{\Sigma}^k\mathbf{A}^\mathsf{T} + \mathbf{W}, \tag{6.85}$$

which is stable if the state transition matrix $\mathbf{A}$ is stable. If $\mathbf{A}$ is stable and $\mathbf{W}$ constant, a unique steady state covariance can be found by solving the Lyapunov equation [10]

$$\boldsymbol{\Sigma}^\infty = \mathbf{A}\boldsymbol{\Sigma}^\infty\mathbf{A}^\mathsf{T} + \mathbf{W}. \tag{6.86}$$

### Scenarios

For the steady-state analysis we studied eleven different observation graphs/configurations (denoted as scenario $\mathsf{S}_{\{1,\ldots,11\}}$) as shown in Figure 6.9 for a network of estimating nodes $\mathbf{S} := \{\mathsf{S}_i | i = 1, \ldots, N\}$. In the first scenario $\mathsf{S}_1$, the nodes are obtaining no observations, meaning that the estimated states are not observable and the uncertainty grows unbounded in case the system is not stable. In $\mathsf{S}_2$, each node obtains private observations and the rank of each subsystem's observabilty matrix is full $\text{Rank}(\boldsymbol{\mathcal{O}}_i) = n_i$. In $\mathsf{S}_3$, cyclic relative observations between nodes are obtained, thus no global information is obtained and renders the system unobservable. $\mathsf{S}_4$ extends $\mathsf{S}_3$, with a single node obtaining private observations, which renders the full state observable. In addition to $\mathsf{S}_4$, fully meshed relative observations are obtained in $\mathsf{S}_6$, while in $\mathsf{S}_5$ all nodes obtain private observations. In $\mathsf{S}_7$, no node obtains private observation, just fully meshed relative observations (a complete graph with bi-directional links). $\mathsf{S}_8$ extends $\mathsf{S}_1$ with a single joint observation between two nodes, while in $\mathsf{S}_9$ all nodes are connected.

In $\mathsf{S}_{10}$, all except the first node are obtaining private measurements, while the first node is relying on relative observation to the other nodes, leading to a star-based observation graph[9]. A unidirectional star-based observation graph rooted at the first node and without private measurements is realized in $\mathsf{S}_{11}$[10].

Please note, that Mourikis and Roumeliotis performed an in-depth performance analysis for different relative pose measurement graphs (RPMG) in [109], by investigating the complete graph ($\mathsf{S}_7$), the cyclic graph ($\mathsf{S}_3$), the disconnected graph ($\mathsf{S}_8$), and the connected graph ($\mathsf{S}_9$).

### Simulation results

We conducted a steady-state covariance analysis for the previously described scenarios using four ($N = 4$) homogeneous mass-spring-damper models (see Section 6.4.1) in MATLAB and compared the resulting covariance by using either a full-state Kalman filter $\boldsymbol{\Sigma}^\infty_{\text{kf}}$ or isolated Kalman filter instances $\boldsymbol{\Sigma}^\infty_{\text{ikf}}$. The solution for the Kalman filter's steady-state covariance was computed by processing 10000 iterations of the DARE (see Equation (6.80)) on the global full state's covariance $\mathbf{x} = [\mathbf{x}_1; \ldots; \mathbf{x}_N]$. In case of the isolated Kalman filter, the recursive filter steps were modified to support isolate private and joint updates (see Section 6.3.3).

The following parameters for mass-spring damper systems were used: a spring stiffness constant $k = 1\,\text{N/m}$, the damping constant $c = 0.1\,\text{Ns/m}$, the mass $m = 1\,\text{kg}$, the gravitational constant $g = 9.81\,\text{m/s}^2$, control input noise $\sigma_g = 0.1\,\text{m/s}^2$, distance measurement noise $\sigma_i = 0.05\,\text{m}$, relative distance noise $\sigma_{i,j} = 0.05\,\text{m}$, the sampling period $\Delta t = 0.001\,\text{s}$, and the initial uncertainty $\boldsymbol{\Sigma}^0_{ii} = \mathbf{I}$.

In Table 6.1, the trace of the full state's uncertainty, the Frobenius norm $\|\mathbf{M}\|_F = \sqrt{\sum_{i=1}^n \sum_{j=1}^m \mathbf{M}_{i,j}^2}$ of the difference between the covariance matrices $\left\|\boldsymbol{\Sigma}^\infty_{kf} - \boldsymbol{\Sigma}^\infty_{ikf}\right\|_\text{F}$ and

---

[9]This constellation will be relevant in our modular sensor fusion approach described in Chapter 4.

[10]A common configuration, when relative information to stationary landmarks is observed in, e.g., VIO.

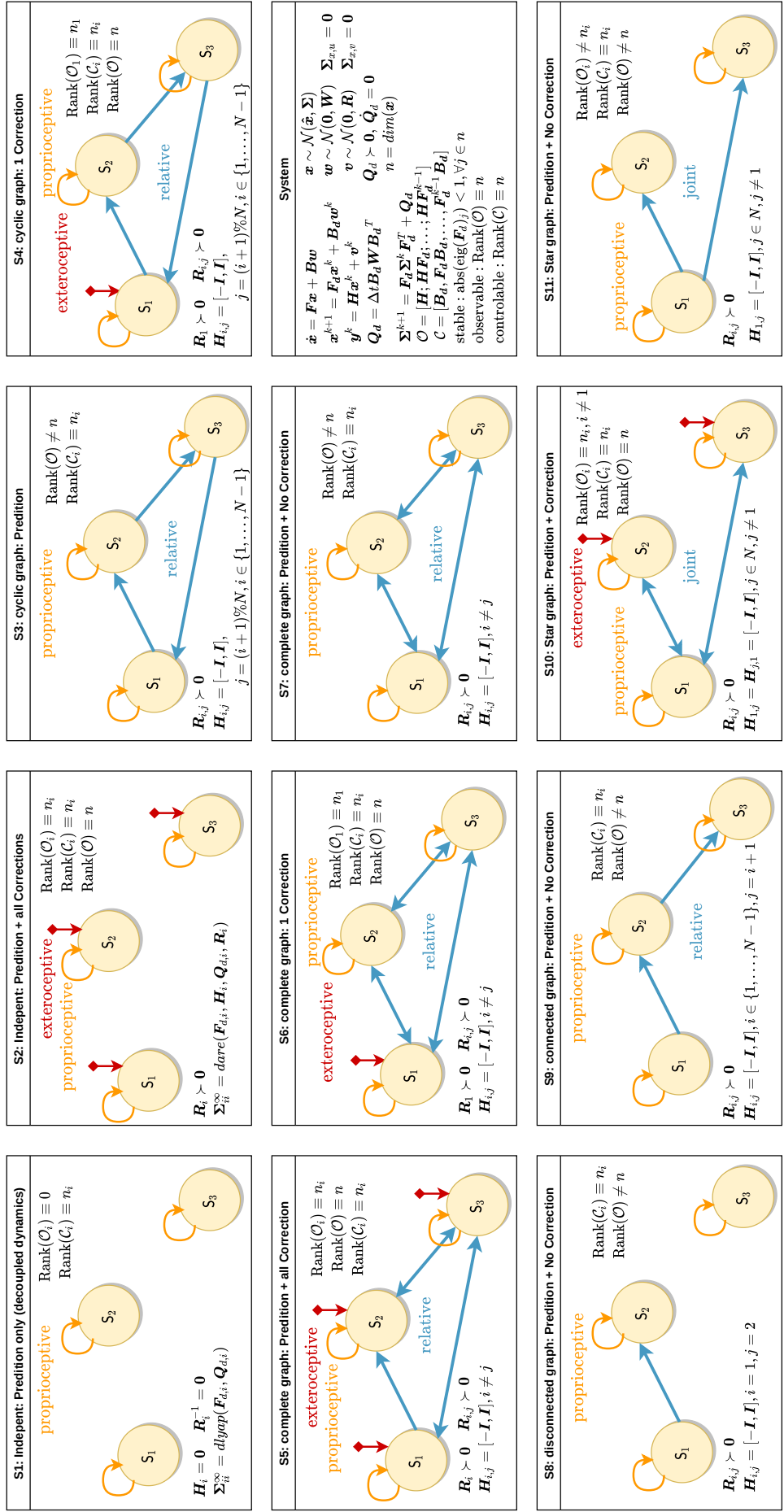

**Figure 6.9:** Shows ten different observations graphs in a network of three nodes $\{S_1, S_2, S_3\}$ that are used to study the steady-state behavior in Section 6.4.2. Arrows in orange indicate proprioceptive observations, private exteroceptive ones are held in red, and relative observations are held in blue.

| S | $N$ | $\frac{\mathbf{Q}}{\mathbf{R}_{ii}}$ | $\frac{\mathbf{R}_{ii}}{\mathbf{R}_{i,j}}$ | $\mathrm{tr}(\boldsymbol{\Sigma}^{\infty}_{kf})$ | $\mathrm{tr}(\boldsymbol{\Sigma}^{\infty}_{ikf})$ | $\mathrm{tr}(\boldsymbol{\Sigma}^{\infty}_{kf} - \boldsymbol{\Sigma}^{\infty}_{ikf})$ | $\boldsymbol{\Sigma}^{\infty}_{kf} \in \mathbb{S}_{+}$ | $\boldsymbol{\Sigma}^{\infty}_{ikf} \in \mathbb{S}_{+}$ | $\left\|\boldsymbol{\Sigma}^{\infty}_{kf} - \boldsymbol{\Sigma}^{\infty}_{ikf}\right\|_{\mathrm{F}}$ | $\left\|\boldsymbol{\mathcal{K}}^{\infty}_{kf} - \boldsymbol{\mathcal{K}}^{\infty}_{ikf}\right\|_{\mathrm{F}}$ |
|---|---|---|---|---|---|---|---|---|---|---|
| 1 | 4 | 0.04 | 1 | 7.31537946 | 7.31537946 | 0 | 1 | 1 | 0 | 0 |
| 2 | 4 | 0.04 | 1 | 0.00015293 | 0.00015285 | 8e-08 | 1 | 1 | 4e-08 | 0.00236663 |
| 3 | 4 | 0.04 | 1 | 1.82898595 | 0.91167764 | 0.91730831 | 1 | 0 | 0.84261337 | 2.78239829 |
| 4 | 4 | 0.04 | 1 | 0.00023918 | 0.00025318 | -1.4e-05 | 1 | 0 | 1.933e-05 | 1.13491998 |
| 5 | 4 | 0.04 | 1 | 0.00018888 | 0.00019349 | -4.61e-06 | 1 | 1 | 4.93e-06 | 0.41905172 |
| 6 | 4 | 0.04 | 1 | 0.00021256 | 0.00022462 | -1.206e-05 | 1 | 0 | 2.186e-05 | 1.44451797 |
| 7 | 4 | 0.04 | 1 | 1.82896187 | 0.91153385 | 0.91742803 | 1 | 0 | 0.8426316 | 2.78312948 |
| 8 | 4 | 0.04 | 1 | 3.65779502 | 3.68750993 | -0.02971491 | 1 | 0 | 0.5709065 | 1.95671019 |
| 9 | 4 | 0.04 | 1 | 1.82900879 | 1.90292098 | -0.07391219 | 1 | 0 | 0.76619192 | 3.45066432 |
| 10 | 4 | 0.04 | 1 | 0.00018997 | 0.00020251 | -1.254e-05 | 1 | 1 | 1.052e-05 | 0.75044314 |
| 11 | 4 | 0.04 | 1 | 1.82900745 | 1.87115590 | -0.04214844 | 1 | 0 | 0.7263689 | 3.33405208 |
| 4 | 4 | 0.01 | 1 | 0.00037189 | 0.00039221 | -2.032e-05 | 1 | 0 | 3.22e-05 | 1.02053597 |
| 4 | 4 | 0.1 | 1 | 0.00018349 | 0.00019385 | -1.036e-05 | 1 | 0 | 1.448e-05 | 1.19326302 |
| 4 | 4 | 1 | 1 | 9.906e-05 | 0.00010406 | -5e-06 | 1 | 0 | 7.65e-06 | 1.27264922 |
| 4 | 4 | 10 | 1 | 5.526e-05 | 5.781e-05 | -2.54e-06 | 1 | 0 | 4.26e-06 | 1.3013204 |
| 4 | 4 | 100 | 1 | 3.113e-05 | 3.254e-05 | -1.41e-06 | 1 | 0 | 2.39e-06 | 1.3075512 |
| 4 | 4 | 1000 | 1 | 1.766e-05 | 1.847e-05 | -8e-07 | 1 | 0 | 1.34e-06 | 1.304943 |
| 4 | 4 | 1 | 0.01 | 0.00027452 | 0.00028853 | -1.401e-05 | 1 | 1 | 1.039e-05 | 0.32939867 |
| 4 | 4 | 1 | 0.1 | 0.00015437 | 0.00016119 | -6.82e-06 | 1 | 1 | 5.96e-06 | 0.5338478 |
| 4 | 4 | 1 | 1 | 9.906e-05 | 0.00010406 | -5e-06 | 1 | 0 | 7.65e-06 | 1.27264922 |
| 4 | 4 | 1 | 10 | 7.107e-05 | 7.518e-05 | -4.11e-06 | 1 | 0 | 1.193e-05 | 2.26602958 |
| 4 | 4 | 1 | 100 | 5.603e-05 | 5.971e-05 | -3.68e-06 | 1 | 0 | 1.573e-05 | 2.94257421 |
| 4 | 4 | 1 | 1000 | 4.769e-05 | 5.183e-05 | -4.14e-06 | 1 | 0 | 1.839e-05 | 3.41465867 |

**Table 6.1:** Show the steady-state behavior using the mass-spring damper model in different scenarios $\mathrm{S}_{\{1,\ldots,11\}}$ with $N = 4$ instances. The experiment is described in Section 6.4.2.

the difference between the corresponding correlation matrices $\left\|\boldsymbol{\mathcal{K}}^{\infty}_{kf} - \boldsymbol{\mathcal{K}}^{\infty}_{ikf}\right\|_{\mathrm{F}}$ (see Equation (2.3)) is listed[11]. The trace of the covariance difference $\mathrm{tr}(\boldsymbol{\Sigma}^{\infty}_{kf} - \boldsymbol{\Sigma}^{\infty}_{ikf}) < 0$ allows us to determine, if the total uncertainty using the IKF paradigm is more conservative, which is always the case when global full state is rendered observable.

All, expect $\mathrm{S}_3$ and $\mathrm{S}_7$, lead to stable steady-state covariances using the IKF paradigm. In the scenarios $\mathrm{S}_3$ (cyclic graph without private observations) and $\mathrm{S}_7$ (complete graph without private observations), the IKF paradigm leads to overly optimistic, and thus, inconsistent estimates, as existing correlations between participants and non-participants are neglected. In contrast, scenario $\mathrm{S}_{11}$ leads to stable beliefs, despite performing purely relative observations without private observations, as approximated correlations are maintained in the first node.

While the full state's covariance remains positive semidefinite using the Kalman filter, it does not always by applying the IKF paradigm.

In Table 6.1, the ratio between the noise parameters was changed for scenario $\mathrm{S}_4$ and less accurate relative observations over private observation have a positive impact using the IKF paradigm as filter instances are less correlated.

In Figure 6.10, Figure 6.11, and Figure 6.12, the final uncertainties for the scenarios $\mathrm{S}_{1,\ldots,11}$ are visualized.

---

[11]Since the $\boldsymbol{\Sigma}^{\infty}_{ikf}$ is potentially not positive semidefinite, we project it to $\mathbb{S}_{+}$ by setting negative eigenvalues to zero, before the correlation matrix $\boldsymbol{\mathcal{K}}^{\infty}_{ikf}$ is computed.

### 6.4.3 Credibility analysis

In our custom MATLAB framework for MMSF (see Section 4.4), we performed 30 Monte Carlo simulation runs per fusion strategy on a set of five estimators (one for each mass-spring-damper system) leading to a full state $\mathbf{x} = \begin{bmatrix} \mathbf{x}_{\{1,\dots,5\}} \end{bmatrix}$.

We specified the following parameters for each mass-spring damper system depending on the ID $i$: a spring stiffness constant $k = 5\,\mathrm{N/m}$, the damping constant $c = 0.1\,\mathrm{Ns/m}$, the mass $m = i\,\mathrm{kg}$, the gravitational constant $g = 9.81\,\mathrm{m/s^2}$, control input noise $\sigma_g = 0.1\,\mathrm{m/s^2}$, distance measurement noise $\sigma_i = 0.1\,\mathrm{m}$, relative distance noise $\sigma_{i,j} = 0.1\,\mathrm{m}$, the sampling period $\Delta t = 0.001\,\mathrm{s}$, the simulation duration $D = 20\,\mathrm{s}$, and the initial uncertainty $\boldsymbol{\Sigma}_{ii}^0 = \mathbf{I}$. Absolute distance measurements are obtained only by the first estimator $\mathsf{S}_1$.

Table 6.2 shows the ARMSE (AR) and ANEES (AN) of the estimated states over 30 Monte Carlo simulation runs. The ANEES should be on average one for each, the position and velocity estimates. Estimates using IKF tend to be slightly less accurate compared to a centralized-equivalent estimator (C) operating in each filter step on the full-state vector. On the other hand, estimators using IKF only require a second estimator at the moment of isolated joint relative position updates.

Figure 6.13 shows the ANEES of three estimators' position and velocity estimates using either fusion strategy. The uncertainty of these estimates are predominately conservative (the ANEES of the velocity is on average below 1) for the C and IKF approach (see Table 6.2). Causes for this behavior may be discretization errors since other values are known by simulation design. As expected, the ANEES of the naive filters indicate inconsistencies, and thus, resulting in significantly higher estimation errors (see Figure 6.14). While filters using the proposed IKF approach perform almost identically as a centralized equivalent filter operating on the global full state – in particular the first estimator instance $\mathsf{S}_1$ that obtains global information.

These results are significant for practical application as it reveals how little performance gain may be obtained by considering correlations to non-participating estimators in a centralized equivalent implementation in view of its limitations regarding scalability and computational resources needed compared to IKF. On the other hand, neglecting correlations between estimators at all in the naive approach leads to inconsistencies.

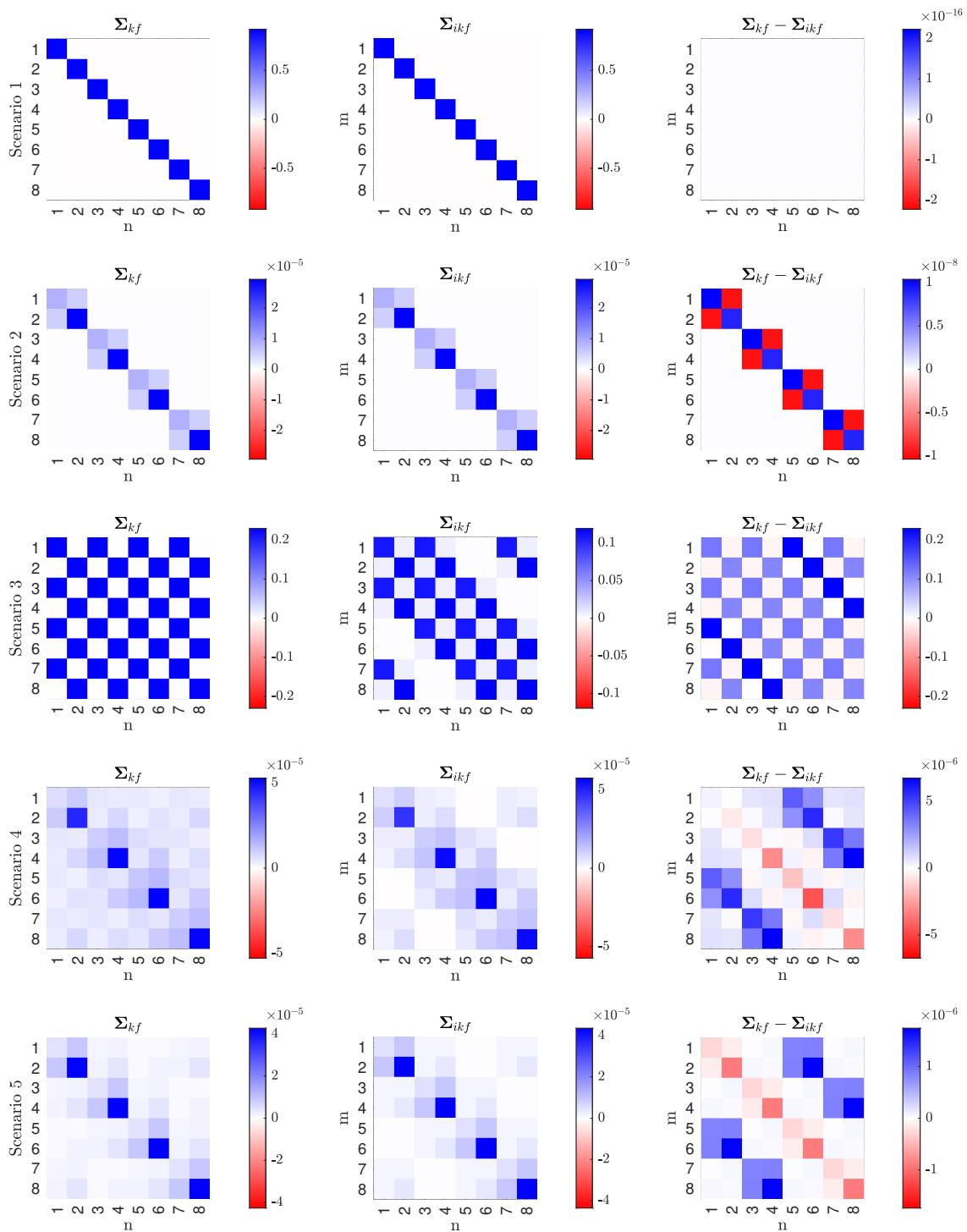

**Figure 6.10:** Observation graphs $S_{1,...,5}$: Left column shows the global covariance using a Kalman filter $\mathbf{\Sigma}_{kf}^{\infty}$, the middle column using the IKF paradigm $\mathbf{\Sigma}_{ikf}^{\infty}$, and right column shows the difference. The mass-spring damper model is described in Section 6.4.1 with noise parameters $\mathbf{Q}_n = 0.01^2$ and $\mathbf{R}_{ii} = \mathbf{R}_{ij} = 0.05^2$.

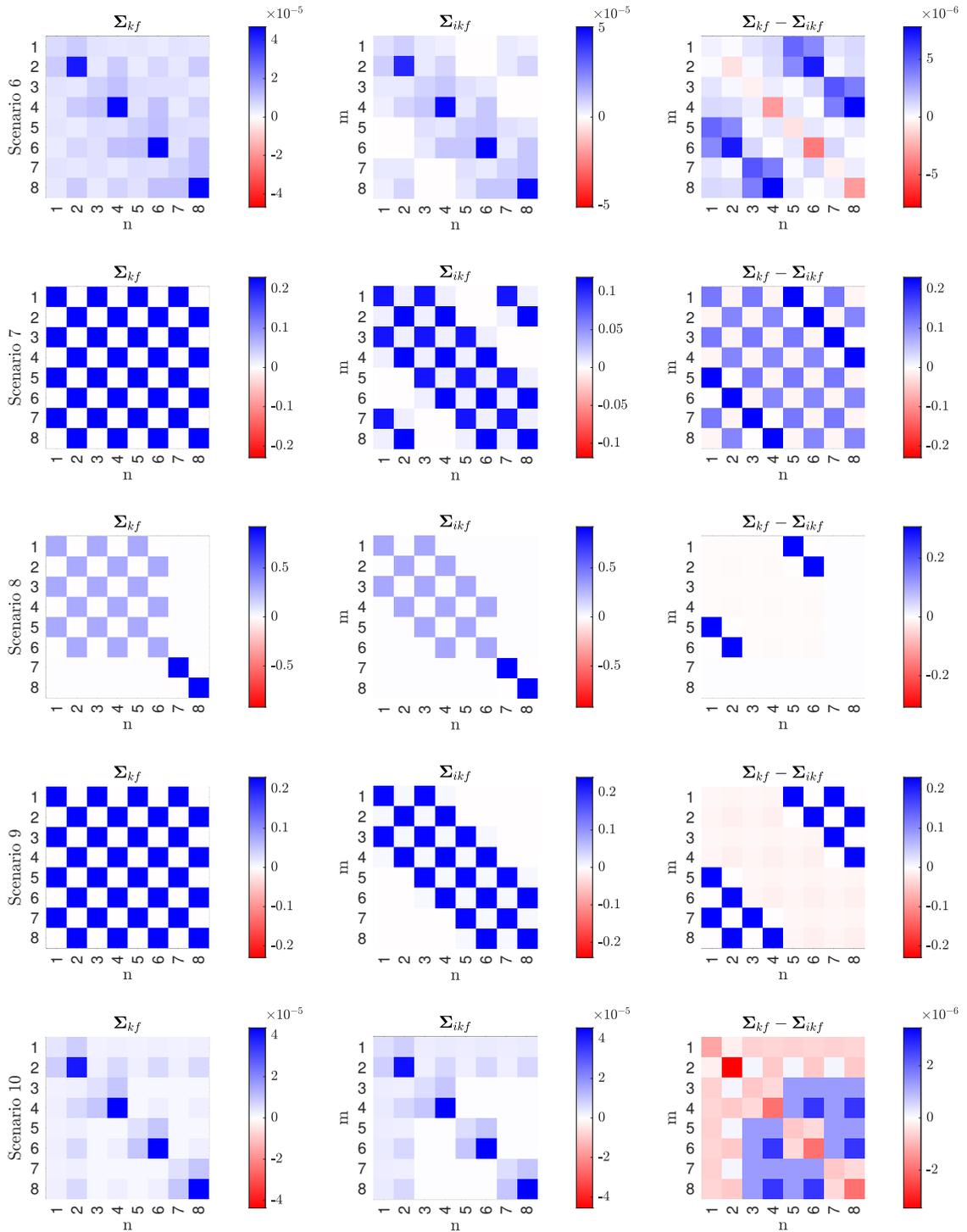

**Figure 6.11:** Observation graphs $S_{6,\dots,10}$: Left column shows the global covariance using a Kalman filter $\boldsymbol{\Sigma}_{kf}^{\infty}$, the middle column using the IKF paradigm $\boldsymbol{\Sigma}_{ikf}^{\infty}$, and right column shows the difference. The mass-spring damper model is described in Section 6.4.1 with noise parameters $\mathbf{Q}_n = 0.01^2$ and $\mathbf{R}_{ii} = \mathbf{R}_{ij} = 0.05^2$.

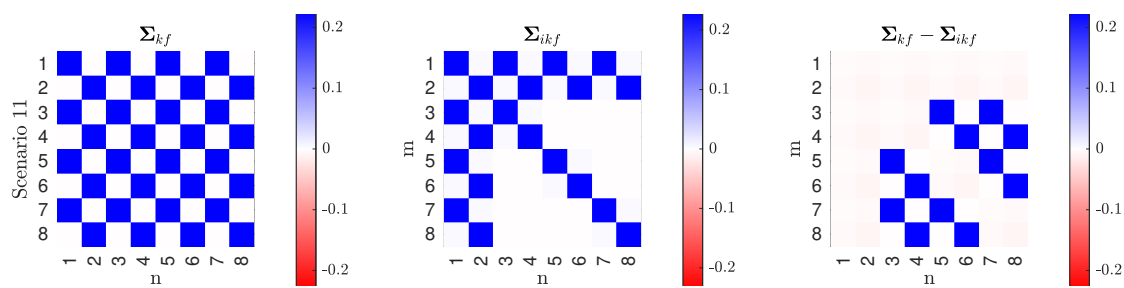

**Figure 6.12:** Observation graph $S_{11}$: Left column shows the global covariance using a Kalman filter $\boldsymbol{\Sigma}_{kf}^{\infty}$, the middle column using the IKF paradigm $\boldsymbol{\Sigma}_{ikf}^{\infty}$, and right column shows the difference. The mass-spring damper model is described in Section 6.4.1 with noise parameters $\mathbf{Q}_n = 0.01^2$ and $\mathbf{R}_{ii} = \mathbf{R}_{ij} = 0.05^2$.

| Strategy | $p_1$ [cm] | | $v_1$ [cm/s] | | $p_2$ [cm] | | $v_2$ [cm/s] | | $p_3$ [cm] | | $v_3$ [cm/s] | | $p_4$ [cm] | | $v_4$ [cm/s] | | $p_5$ [cm] | | $v_5$ [cm/s] | |
|---|---|---|---|---|---|---|---|---|---|---|---|---|---|---|---|---|---|---|---|---|
| | AR | AN | AR | AN | AR | AN | AR | AN | AR | AN | AR | AN | AR | AN | AR | AN | AR | AN | AR | AN |
| C | **0.86** | **0.65** | 5.75 | **0.23** | **1.18** | 0.67 | **7.15** | 0.27 | **1.46** | 0.63 | **7.91** | 0.29 | **1.5** | 0.66 | **7.38** | 0.32 | **1.71** | 0.71 | **7.09** | 0.32 |
| IKF | **0.86** | 0.64 | **5.55** | 0.22 | 1.31 | **0.69** | 7.36 | **0.3** | 1.85 | **0.79** | 8.93 | **0.46** | 1.85 | **0.79** | 8.59 | **0.54** | 1.85 | **0.79** | 7.67 | **0.53** |
| Naive | 2.63 | 12.41 | 8.15 | 1.74 | 4.94 | 39.22 | 12.33 | 9.45 | 7.02 | 94.37 | 14.56 | 24.16 | 7.78 | 138.07 | 14.66 | 37.65 | 6.81 | 106.53 | 12.25 | 32.02 |

**Table 6.2:** Credibility analysis: Shows the ARMSE (AR) and ANEES (AN) of the estimated states over 30 Monte Carlo simulation runs using a centralized equivalent estimator (C), the proposed IKF paradigm, and a naive approach (Naive) neglecting correlations. Best values in bold. The approaches (C) and (IKF) are slightly underconfident in the velocity estimates, while the naive one (Naive) seems to settle, after an inconsistent phase (see Figure 6.13), at higher values. The experiment is described in Section 6.4.3.

## 6.5   Conclusion

In this chapter, we have proposed a novel Kalman filter decoupling paradigm denoted as Isolated Kalman Filtering (IKF), which paves the way to address large and complex estimation problems in an efficient and distributed fashion. We have theoretically justified that the decoupling between participants and non-participants in isolated joint observations is based on an *implicit* maximum determinant completion, that leads stable decoupled estimates under certain conditions. In case of pure relative updates and therefore an unobservable system, the decoupled estimates in cyclic or complete observation graphs become inconsistent (overly optimistic), due to a fictive information flow. Star-based or connected relative observation graphs lead to stable, but pessimistic, steady-state covariances. If the system is fully observable and controllable, any relative observation graph leads to stable steady-state covariances. Based on empirical insights, we believe the following conjecture to be true:

**Conjecture 1** *The IKF paradigm leads to a finite, unique, and stable steady-state covariance $\Sigma_{ikf}^{\infty}$ if the system is completely observable and controllable. The global matrix is not necessary positive semidefinite, while stacked block matrices relating to output-coupled instances are.*

In a Monte Carlo simulation on a linear toy example, we manifest the analysis and highlight the superiority over a naive implementation, while still performing considerably well compared to an exact Kalman filter formulation.

The IKF paradigm allows to decouple the inputs and to couple the outputs of dynamic systems isolated, if the coupled systems fulfill certain criteria (see Section 6.2). Due to the isolated output coupling, corrections are not obtained by correlated but non-participating nodes, which prevents, e.g., sensor self-calibration (see Section 5.3). Hidden variables, e.g., intrinsic or extrinsic sensor states, that are observable by the system configuration in an exact filter formulation, might not be rendered observable or just partially observable using the IKF paradigm. This might reduce the overall accuracy and state convergence. As similar discussion can be found in [138], regarding tightly and loosely coupled estimators. Consequently, the observability properties can only be guaranteed among the output-coupled estimator nodes, as the participants are not aware of non-participants.

In our next chapter Chapter 7, an algorithm unifying CSE and MMSF based on IKF is presented.

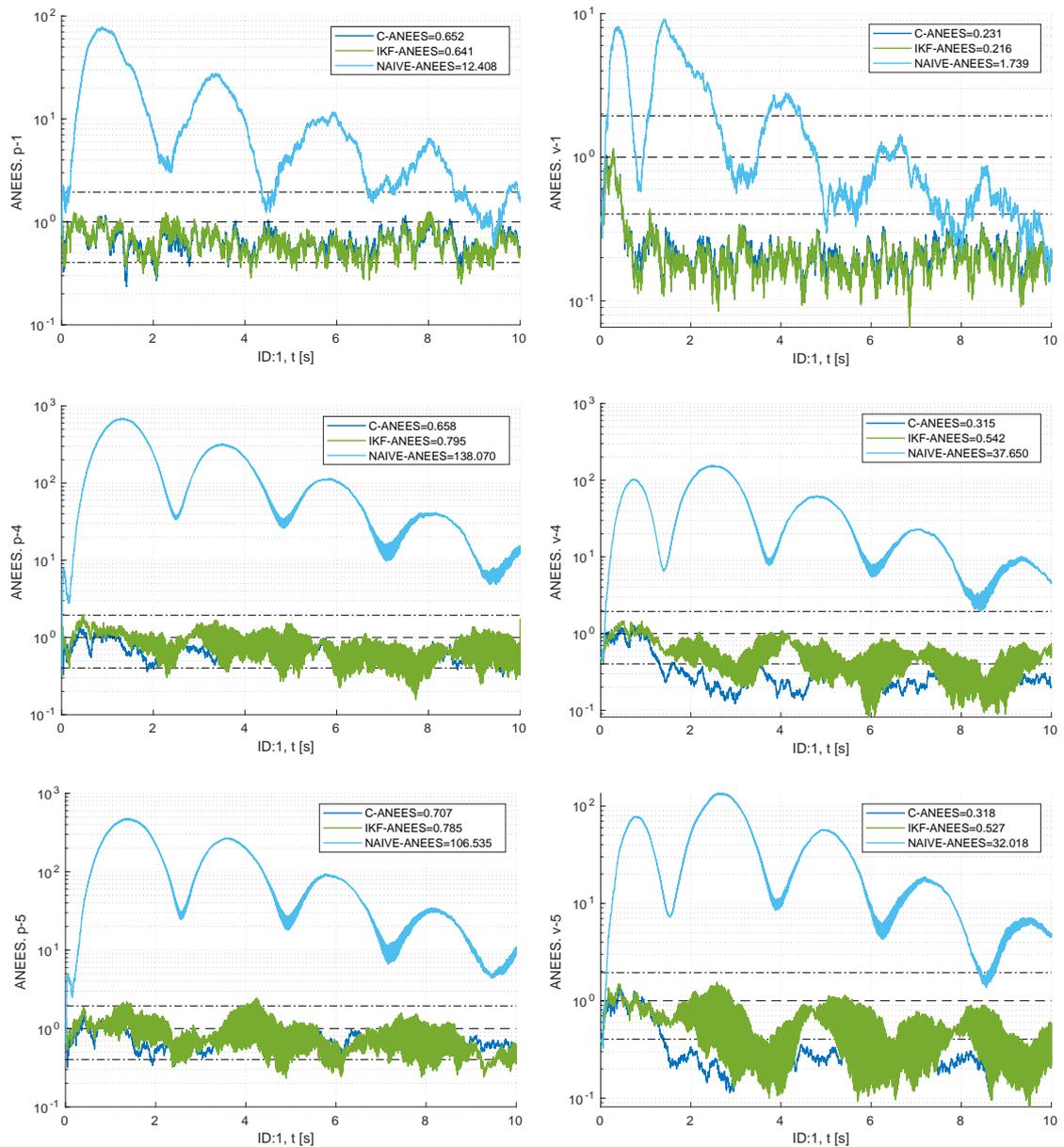

**Figure 6.13:** Credibility analysis: Shows the ANEES in a logarithmic scale, the double-sided 95 % confidence region (dotted lines), and expected ANEES value (dashed line) over 30 Monte Carlo simulation runs of the position $p_{\{1,4,5\}}$ (left column) and velocity $v_{\{1,4,5\}}$ (right column) estimate of the first $\mathsf{S}_1$, forth $\mathsf{S}_4$, and fifth estimator $\mathsf{S}_5$ using a centralized equivalent estimator (blue), the proposed Isolated Kalman filtering (IKF) paradigm (green), and the naive filter (cyan). The experiment is described in Section 6.4.3.

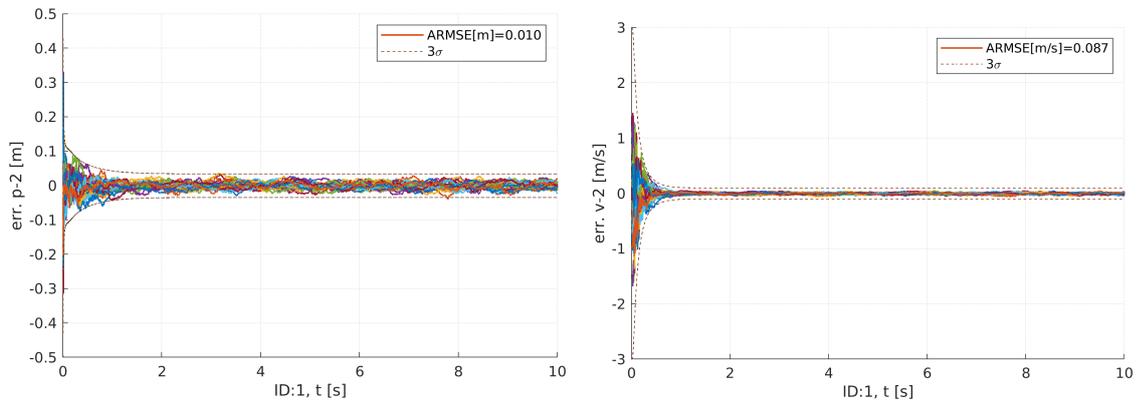

**(a)** Centralized equivalent estimator.

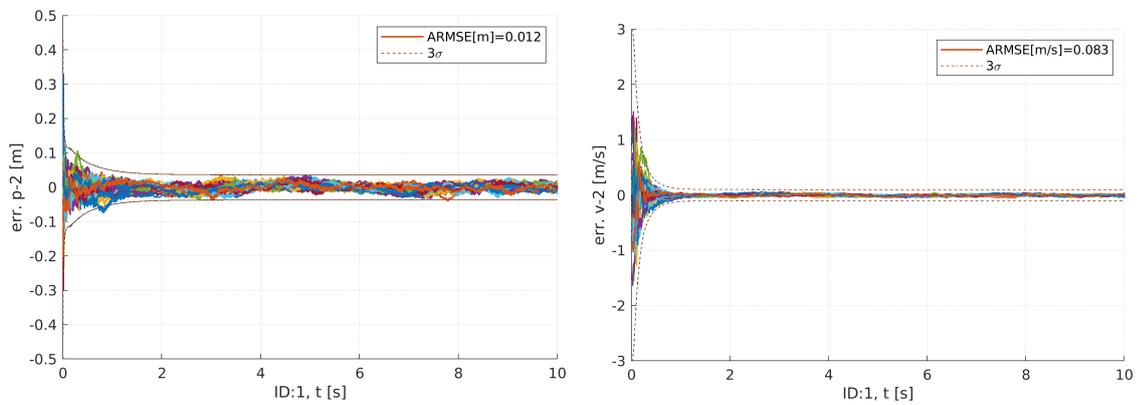

**(b)** IKF.

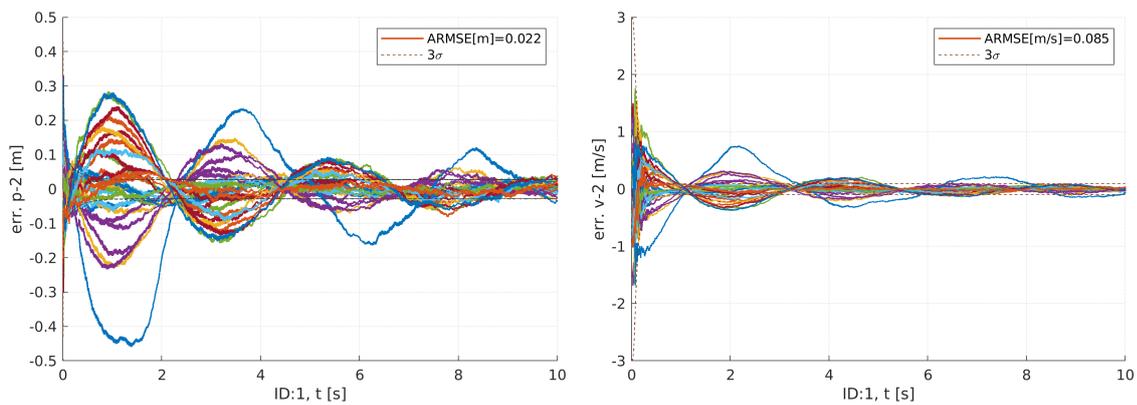

**(c)** Naive estimator.

**Figure 6.14:** Credibility analysis: Shows the estimation errors and the double-sided 99.7 % confidence region (dotted lines) over 30 Monte Carlo simulation runs of the position $\mathbf{p}_2$ (left column) and velocity $\mathbf{v}_2$ (right column) estimate of the second estimator $\mathsf{S}_2$. In the first row using a centralized equivalent estimator, in the second row the proposed Isolated Kalman filtering (IKF) paradigm, and in the last row the naive filter. The experiment is described in Section 6.4.3.

# Chapter 7

# Distributed Collaborative Modular Sensor Fusion

The objective of this chapter is to address two open issues in state-of-the-art filter-based DCSE algorithms. First, to deal with changing dynamical systems of individual nodes/agents in the swarm by, e.g., a time-varying/changing sensor constellation on an agent. For instance, if the sensor-suite is changed during mission (change of propagation sensor in case of hybrid vehicles, wheel encoder to an IMU or velocity sensor) or redundant sensors are detected as faulty and are logically removed from the estimator. Second, to deal with out-of-sequence measurements, which typically result from a (pre-) processing delay between the moment sensor data was perceived until it reaches the filter update step in the processing unit. We tackle these problems by a novel unified filter architecture for modular and collaborative sensor fusion based on the IKF paradigm (see Chapter 6) and provide a detailed pseudocode. In Monte Carlo simulations, we evaluate the filter credibility and analyzed the execution time with increasing sensor latency and increasing number of sensors.

## 7.1 Introduction

In this chapter, we investigate on a unified distributed EKF algorithm for both MMSF and CSE, which typically has been addressed in literature separately. In fact, our IKF paradigm described in Chapter 6, allows us to treat physical sensors as isolated estimator instances (e.g., mounted on an agent or distributed in the area of interest). This abstraction allows us to not only perform state estimation on agents truly modular, but also to perform inter-agent observations efficiently, requiring only the subset of directly involved estimator instances.

Challenges of modular sensor fusion are the handling of different sensor rates, delays, failure detection, self-calibration, as well as true modularity that ensures robustness against a single point of failure, e.g., due to redundancy. These requirements impose tremendous challenges regarding the increasing computation as well as the architectural complexity of the fusion framework.

At the same time, from our Definition 1, CSE aims to enable the agents' estimators to work together to achieve a common task or goal. Collaboratively estimating states or performing jointly observations can significantly improve an individual's estimates, but abrogates the statistical independence of the involved estimators, which need to be considered and properly treated in order to obtain consistent estimates.

Rendering CSE exact can (i) significantly improve the estimation performance of individual agents, (ii) provide redundancy in case of sensor failures, jamming or spoofing, and

(iii) enable agents with less accurate sensors to benefit from agents with more accurate ones [109, 122]. Therefore, it is often used in the field of CL.

Rendering CSE distributed on agents is challenging as it can degrade the performance, lead to inconsistent estimates, requires extensive bookkeeping, communication, or/and requires high computation if done naively.

To the best of our knowledge, delayed collaborative measurements in centralized or distributed CSE have not been addressed yet and impose another dimension of complexity. As collaborative measurements tend to be more time-consuming than processing sensor data on-board locally (due to the involvement of other agents each with multiple sensors), a delay with respect to other sensors and/or agents is a crucial aspect to be properly considered. Whilst handling this consistently in a centralized fusion entity is straightforward (yet with high computation demand), it imposes additional challenges and constraints in a distributed setup.

Truly modular estimators allow for adding or removing sensors during operation leading to shrinking or growing full states of individual agents, meaning that the correlation between agents change (thus the vector and matrix dimensions), in the worst case, between two consecutive inter-agent observations. To the best of our knowledge, existing centralized or distributed CSE algorithms do not explicitly focus on this circumstance.

An estimation problem consisting of various heterogeneous agents with versatile sensor suites that jointly observe each other or jointly observing object of interests, imposes a tremendous challenge for existing filter-based approaches. In our experiments, we show that the here proposed DC-MMSF algorithm outperforms other fusion strategies we ported from existing CSE formulations regarding scalability and timing, while keeping up with the accuracy and credibility of centralized-equivalent architectures. The algorithm unifies and extends our recent publications in the domain of CSE [76] and MMSF [74], and builds upon a novel and stable Kalman filter decoupling strategy denoted as IKF (see Chapter 6).

More precisely, our contributions are:

- Proposing a unified filter architecture for MMSF and CSE based on IKF that supports delayed measurements and the maintenance effort for each estimator instance is invariant to the number of correlated sensors ($\mathcal{O}(1)$).
- Providing detailed pseudocode for our proposed DC-MMSF approach.
- A nonlinear observability analysis on three different multi-agent ranging scenarios.
- An evaluation on real data for a swarm of five collaborating agents with synthetic but realistic inter-agent measurements.

The chapter is structured as follows. In Section 7.2 and in Section 7.3, the motivation and combining of CSE and MMSF is described, respectively. In Section 7.4, the algorithm is proposed. In Section 7.5, we describe the simulation framework and sensor models of an indoor navigation problem with real and augmented data, which is evaluated in terms of estimator credibility and by performing a timing analysis. Finally, we conclude the algorithm in Section 7.6.

## 7.2  Motivation

We assume a time-varying set of communicating agents $\mathbf{A} := \{A_i | \ i = 1, \ldots, M\}$ and a time-varying set of estimator instances per agent. The set of all sensor instances is defined as $\mathbf{S} := \{S_i | \ i = 1, \ldots, N\} = \{S_i | \ i = 1, \ldots, |\mathbf{A}|\}$. In modular sensor fusion, each sensor related state $\mathbf{x}_{S_i}$ needs to be regarded as part of the *global full state* $\mathbf{x}$. Since the global full state is potentially distributed among multiple agents, which can further be dividing into various sensor specific states, it is desirable to reduce the communication and computation effort whenever possible. In a centralized filter, which typically operates on the full state, the entire state and the joint covariance matrix are needed in *each* filter step. It is known that the processing time of a naive estimator increases cubically $\mathcal{O}(L^3)$ with the state vector length $L$ and additionally linearly with the update rate of non-delayed sensor measurements. On top of that, delayed sensor updates and the handling of the interim sensor observations will result in additional processing time, depending on the strategy used. With an increasing amount of sensors, the effective measurement rate and the amount of delayed measurements increases.

To overcome this burden, we propose the use our Kalman filter decoupling strategy, which we denote as Isolated Kalman Filtering (IKF), that allows isolated filter steps and to restore interdependencies between estimators/nodes when needed. Due to isolated filter steps, where only a subset of decoupled estimators are required, we are distinguishing between participating $(p)$ $\mathbf{S}_p := \{\mathsf{S}_i | \ i \in \mathbf{P}\}$ and non-participating (others $o$) $\mathbf{S}_o := \{\mathsf{S}_i | \ i \in \bar{\mathbf{P}}\}$ estimators, see Equation (6.1f) and Equation (6.1g), respectively. Further, we can differentiate between *participants'* and *non-participants'* beliefs, resulting in a stacked random variable $\mathbf{x} = \begin{bmatrix} \mathbf{x}_p; \ \mathbf{x}_o \end{bmatrix}$. $\mathbf{x}_p$ is a joint belief of participants, e.g., consisting of $\mathsf{S}_i$'s and $\mathsf{S}_j$'s belief $\mathbf{x}_p = \begin{bmatrix} \mathbf{x}_i; \ \mathbf{x}_j \end{bmatrix}$, and a joint belief of non-partipants $\mathbf{x}_o$.

## 7.3 Unifying CSE and MMSF

In this chapter, we extend our MMSF-DAH (see Chapter 4) approach to support CSE, denoted as the DC-MMSF approach. As a consequence, CSE can be performed distributed among agents, while (i) communication between agents is only required at the moment of inter-agent joint observations, (ii) one agent acts as interim master to process the *isolated state correction* (see Section 6.3.3) on the participants' stacked belief, (iii) agents can be added and removed from the swarm, while removed ones will still cause a slight maintenance overhead in the correlated ones, but will be able to join the swarm later again, (iv) each agent's local full state can vary during mission, meaning that each local sensor suite can be truly modular, and (v) delayed and multi-rate sensor updates are supported (see Section 6.3.4). The ability of (iv) and (v) are clear contributions with respect to CSE that have not been in focus of recent works of, e.g., [73, 76, 80, 95, 121].

As a consequence of inter-agent joint observations, an agent's local sensor suite will be correlated to other agents' sensor suites. A key concept is to maintain correlations between local estimator instances – the sensor specific IKF instances – on a single agent equal to correlations between agents' estimator instances. The only differences are the availability and accessibility of the required information in case of inter-agent (global) joint observations and that delayed measurements might lead to reapplying measurements on multiple agents. That said, although being different in view of the implementation, conceptually, no differences are made with respect to the local estimator instances.

Figure 7.1 depicts a block diagram of main components of the proposed DC-MMSF approach, showing a clear abstraction between agents and the estimator instances (nodes). That allows us to propose a unified estimator architecture to support both, MMSF and CSE seamlessly and is one major difference to existing filter-based fusion architectures, e.g., [18, 46, 58, 96, 150].

By using a correction buffer per estimator instance, originating from [76] and explained in more detail in Section 6.3.1, the maintenance effort for correlations between estimator instances (locally or globally) in the propagation step can be reduced to $\mathcal{O}(1)$, and thus, can be shifted to the moment when correlations are required again. This concept renders DC-MMSF (i) ideal for high prediction rates, as it is typically the case for an ESEKF based on IMU propagation, (ii) capable of performing any-sensor to any-sensor observations and private observations, (iii) capable of re-applying updates faster after delayed (out-of-order) measurements, (iv) having minimal overhead for maintaining temporally disabled estimator instances or sporadic inter-agent joint observations, and (v) requiring communication only among *participating* agents during the event of processing inter-agent joint observations[1].

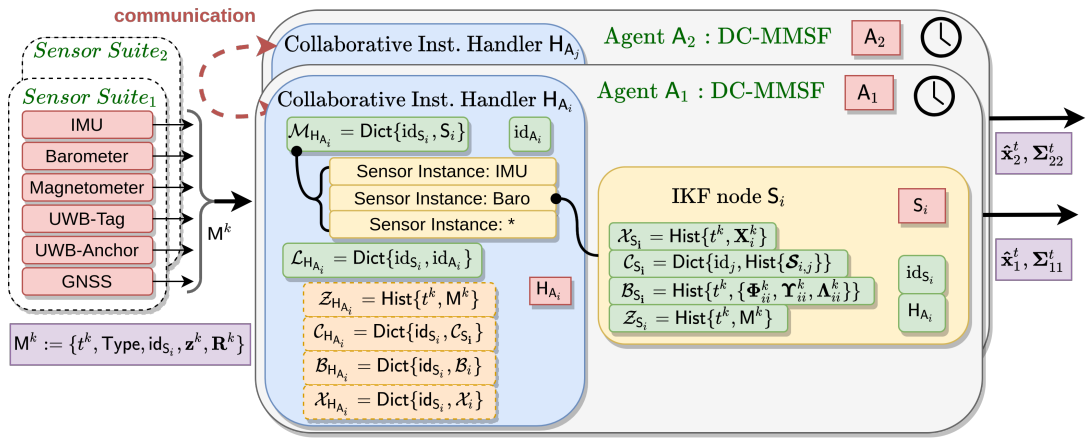

**Figure 7.1:** DC-MMSF: Shows the modular multi-sensor fusion framework of individual agents, consisting of a collaborative instance handler $\mathsf{H_A}$, held in blue, maintaining various IKF instances $\mathsf{S}$, held in yellow, for a specific sensor type. The sensor suite provides measurements to the $\mathsf{H_A}$, which again delegates them to the appropriate estimator instance $\mathsf{S}$, which performs the isolated filter steps. The $\mathsf{H_A}$ has logically access to the measurement-, belief-, factorized cross-covariance, and correction history ($\mathcal{Z}_{\mathsf{H_A}}$, $\mathcal{X}_{\mathsf{H_A}}$, $\mathcal{C}_{\mathsf{H_A}}$, and $\mathcal{B}_{\mathsf{H_A}}$, respectively), held in orange, of the entire sensor constellation. For inter-agent joint observation, information exchange between collaborative instance handlers of participants $\mathsf{H_{A}}_i$, $i \in \mathsf{P}$, is needed. Therefore, a lookup table $\mathcal{L}_{\mathsf{H_A}}$ associating sensor instances with agents is maintained. Details are provided in Section 7.4.

## 7.4   The DC-MMSF Algorithm

In this section, the architecture of a local fusion entity, the *Collaborative Instance Handler* $\mathsf{H_A}$ (see Section 7.4.1), to unify $N$ *locally* or *globally* held IKF instances $\mathsf{S}$ is proposed and depicted in Figure 7.1. This architecture allows us to do both, perform modular multi-sensor fusion on an agent and collaboratively among distributed agents, thus, it is denoted as Distributed Collaborative Modular Multi-Sensor Fusion (DC-MMSF).

As previously defined, *local* refers to estimator instances associated to (rigidly) attached sensors on an individual agent, while *global* specifies the estimator instances maintained on the other agents, meaning that they are not directly accessible and communication is required in order to exchange desired information[2]. An agent $\mathsf{A}_i$ has a unique identifier $\mathsf{id}_{\mathsf{A}_i}$ and is an element of a time-varying group of agents $\mathbf{A}$ with computational and range limited communication capabilities.

---

[1]The amount of communication links scales in the number of agents involved $n$ with $\mathcal{O}(n)$. The information to be exchanged depends on the number of involved estimator instances $m$, on the data representation, and on each estimator instances' state vector length $v_i$, which scales with $\mathcal{O}(v^2)$ with $v = \sum_{i=1}^{m} v_i$ as partitions of the joint belief of participants need to be exchanged.

[2]This implies that local estimator instances can be efficiently accessed by inter-process communication or directly by sharing the same memory within the same process.

### 7.4.1 Collaborative Instance Handler

Each agent $\mathsf{A}_i$ holds in its handler $\mathsf{H}_{\mathsf{A}_i}$ a varying set of IKF instances $\mathsf{S}$ in a dictionary $\boldsymbol{\mathcal{M}}_{\mathsf{H}_{\mathsf{A}_i}} := \mathsf{Dict}\{\mathsf{id}_{\mathsf{S}_i}, \mathsf{S}_i\}^3$. The elements are accessed via a unique sensor identifier $\mathsf{id}_{\mathsf{S}}$ that is associated to a physically real sensor of the agent's sensor suite. The sensors are of a specific and known type, e.g., an IMU, barometer, magnetometer, GNSS sensor, etc. and allows instantiating sensor specific estimator instances. The handler maintains a lookup table $\boldsymbol{\mathcal{L}}_{\mathsf{H}_{\mathsf{A}_i}} := \mathsf{Dict}\{\mathsf{id}_{\mathsf{S}_i}, \mathsf{id}_{\mathsf{A}_i}\}$ for associating unique sensor identifiers with unique identifiers for agents. This table is updated and used at the moment of global/inter-agent joint observation, where local IKF instances require data from an IKF instance maintained on other agents via (wireless network) communication. All (local) sensors measurements are delegated to the instance handler $\mathsf{H}_{\mathsf{A}_i}$, which further delegates them to the appropriate estimator instance (see Algorithm 7.1). In case of local joint observations, information exchange between local instances can be handled locally via the handler $\mathsf{H}_{\mathsf{A}_i}$, i.e. the memory can be directly accessed through pointers within a common process.

As described in Section 6.2, each estimator instance $\mathsf{S}_i$ maintains a history of recent beliefs $\boldsymbol{\mathcal{X}}_{\mathsf{S}_i}$, a dictionary $\boldsymbol{\mathcal{C}}_{\mathsf{S}_i}$ with histories of factorized cross-covariances $\mathcal{S}$ relating to other estimator instances, a history of recent corrections terms $\boldsymbol{\mathcal{B}}_{\mathsf{S}_i}$, a history or recent measurements $\boldsymbol{\mathcal{Z}}_{\mathsf{S}_i}$, a unique identifier $\mathsf{id}_{\mathsf{S}_i}$, and a reference to the local instance handler $\mathsf{H}_{\mathsf{A}_i}$, in order to request access to other local or global IKF instances to perform isolated joint observations. If the requested IKF instance is not locally available, i.e. not maintained in $\mathsf{H}_{\mathsf{A}_i}$, then the $\mathsf{H}_{\mathsf{A}_i}$ performs a lookup in the $\boldsymbol{\mathcal{L}}_{\mathsf{H}_{\mathsf{A}_i}}$. If the requested IKF instance was not found, the agent requests a list of sensor identifiers from agents in communication range. If the identifier cannot be found, the inter-agent joint observation is not performed. Otherwise, the $\mathsf{H}_{\mathsf{A}_i}$ request from the associated agent the required information as described in Section 7.4.4.

A sensor measurement is defined as a set $\mathsf{M}^k := \{t^k, \mathsf{Type}, \mathsf{id}_{\mathsf{S}_i}, \mathbf{z}^k, \mathbf{R}^k\}$, with a timestamp $t^k$ of the measurement $\mathbf{z}^k$, the measurement uncertainty $\mathbf{R}^k$, the measurement type $\mathsf{Type}$, and the sensor identifier $\mathsf{id}_{\mathsf{S}}$. Each estimator instance can, based on the measurement type $\mathsf{Type}$, issue the appropriate measurement method which leads to an isolated filter step (see Algorithm 7.4). In case of inter-agent joint observation, the measurement type needs to encode the agent association.

Each $\mathsf{H}_{\mathsf{A}_i}$ maintains a history of (chronologically sorted) recent measurements in the $\boldsymbol{\mathcal{Z}}_{\mathsf{H}_{\mathsf{A}_i}}$ buffer for compensating delayed, i.e. out-of-order measurements within the handler. If the measurement is not rejected from the NIS-based hypothesis check (Algorithm 7.6), all elements (beliefs, cross-covariance factors, correction terms) after the measurement event are deleted from all local buffers and all interim measurements are re-applied in order to properly re-compute the deleted elements. If any interim measurements are related to another agent and if it was not rejected, then **all known agents** will be notified, if in range, to re-apply their measurements after the delayed one (see Algorithm 7.3). Certainly, this is computational-wise suboptimal, as only a subset of known agents might be post-correlated, but simplifies the algorithm (ideally, only post-correlated agents are notified). Otherwise, redoing measurements among participants only, might lead to dangling cross-covariance factors as described in Section 6.3.4. Figure 7.2 underlines, that an interim global joint observations $\mathbf{H}^3_{5,3}$ with a non-participating agent needs to be reprocessed as well. Finally, the new measurement is inserted chronologically sorted into the agent's buffer $\boldsymbol{\mathcal{Z}}_{\mathsf{H}_{\mathsf{A}_i}}$.

Summarized, a handler $\mathsf{H}_{\mathsf{A}_i}$ maintains a dictionary of estimator instance $\boldsymbol{\mathcal{M}}_{\mathsf{H}_{\mathsf{A}_i}}$ and a sliding time horizon buffer of recent measurements $\boldsymbol{\mathcal{Z}}_{\mathsf{H}_{\mathsf{A}_i}}$. Via the dictionary $\boldsymbol{\mathcal{M}}_{\mathsf{H}_{\mathsf{A}_i}}$, the handler has logically access to the history of all local (i) beliefs $\boldsymbol{\mathcal{X}}_{H_{A_i}}$, (ii) correction terms $\boldsymbol{\mathcal{B}}_{H_{A_i}}$, and (iii) cross-covariance factors $\boldsymbol{\mathcal{C}}_{H_{A_i}}$, as depict in Figure 7.1.

---

[3]Careful reader will notice that in MMSF-DAH (see Section 4.3), we called the handler *Instance Handler*. The collaborative handler is a specialization of the instance handler and is capable of handling inter-agent communication requests.

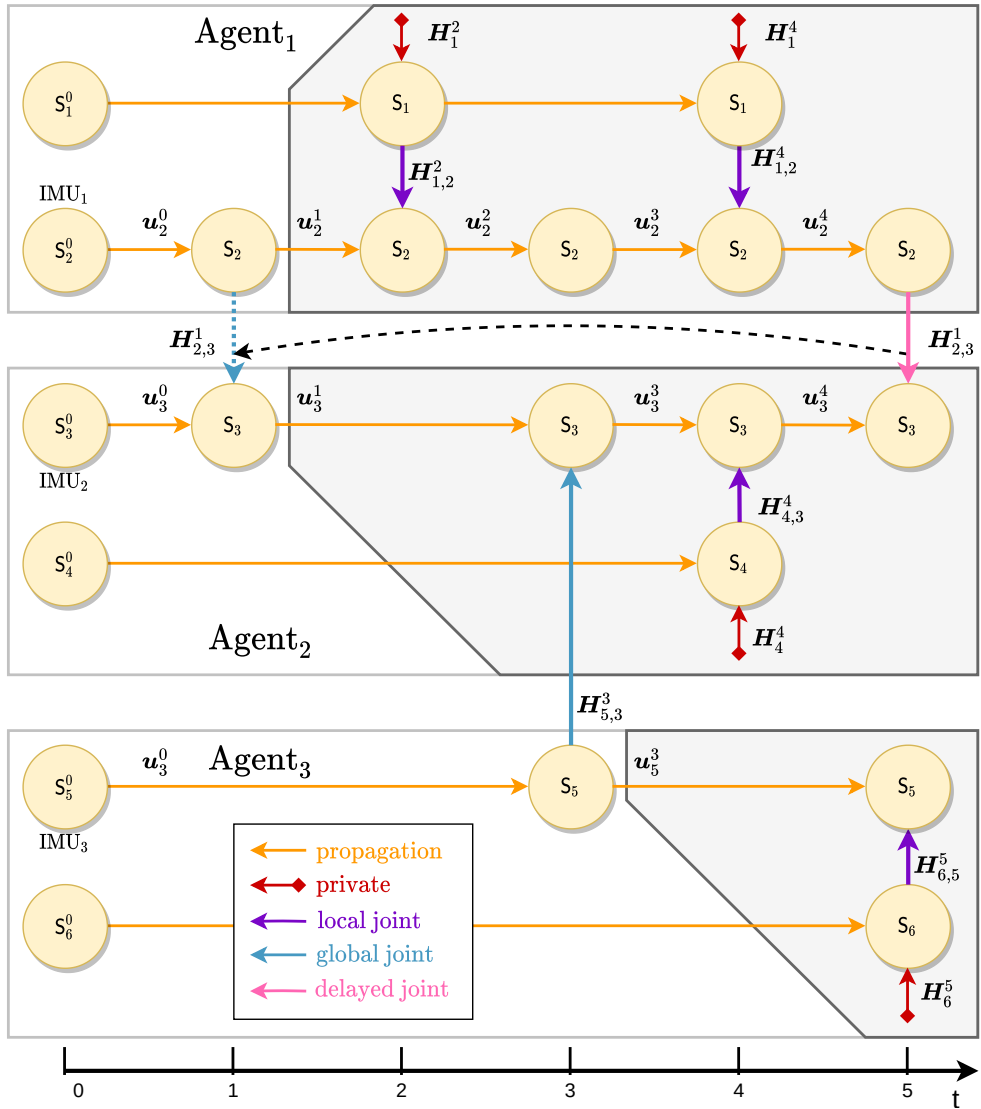

**Figure 7.2:** Shows the measurement graph of six instances $\{S_{1,\dots,6}\}$ maintained on three agents $\{A_{1,2,3}\}$. Arrows in orange are predictions, e.g., due to control inputs $\mathbf{u}$. Local private observations are held in red, local joint observations in purple, and global (inter-agent) joint observations in blue. The global joint observation at $t = 5$ held in magenta indicates a delayed measurement $\mathbf{H}_{2,3}^1$ originating from $t = 1$. In order to restore the beliefs at $t = 5$, all intermediate measurements after that event (within the gray boxes) need to be reprocessed in an appropriate order. Obviously, $A_2$ needs to trigger $A_3$ to re-apply the global joint observation $\mathbf{H}_{5,3}^3$ at $t = 3$.

### 7.4.2   Time Horizon

To reduce the memory footprint, it is desirable to have a fixed time horizon for the buffers, which leads to the challenge that required information can fall out of the time horizon of the history, as discussed in Section 6.3.1. As the time horizon of the correction buffer $\boldsymbol{\mathcal{B}}_S$ is finite, we forward-propagate factorized cross-covariances $\boldsymbol{\mathcal{S}}$, if they are about to fall outside that time horizon (see Algorithm 7.11). For example, if correlated estimator instances are not performing joint observations for a duration longer than the time horizon of $\boldsymbol{\mathcal{C}}_S$. There-fore, the horizon has to be checked periodically in propagation steps of estimator instances (see Algorithm 7.12). This ensures that the time horizon of the correction buffer $\mathcal{B}_\mathsf{S}$ is always adequately populated, while reducing the maintenance effort during the propagation and private update steps compared to [95, 121]. The length of the correction buffer's horizon has an impact on *when* and *where* that propagation happens (either propagation or joint observation step). In the limit of infinitely long time horizons, it always happens in joint observation step (when it is actually required), with limited time horizons the buffer-length check in the estimator's propagation step may trigger such a forward-propagation of factorized cross-covariances. A reasonable time horizon for all buffers on an agent would be twice the maximal sensor latency of its sensors, since it would prevent a forced forward propagation of cross-covariance factors in its estimator instances propagation step (if the agent is not correlated to other agents' estimator instances). In case of inter-agent correlations, it is likely that agents are not having a rendezvous for a longer period of time or just sporadically throughout the mission. Therefore, the agent's buffer horizons should be adjusted primarily to the local sensor configuration, and it should be expected that inter-agent cross-covariance factors $\mathcal{S}$ will be forcibly forward propagated.

### 7.4.3 Interim Master

In joint observations, the instance $\mathsf{S}_i \in \mathsf{A}_i$ that obtained the measurement acts as *interim master*. It has to (i) stack the joint belief by requesting the beliefs and cross-covariance factors $\mathcal{S}$ of all participants $\mathsf{S}_p$, (ii) to compute the measurement Jacobian $\mathbf{H}_p$, (iii) to perform the NIS hypothesis check in order to detect measurement outliers (see Algorithm 7.6), (iv) to compute the stacked state correction $\Delta\hat{\mathbf{x}}_p^{k(+)}$ and the *a-posteriori* joint covariance $\mathbf{\Sigma}_{pp}^{k(+)}$, and (v) to split and send the appropriate parts to the participants. Each individual participant then (i) corrects the *a-priori* estimate with the correction obtained, (ii) computes the correction factor $\mathbf{\Lambda}$ and adds it to the correction buffer $\mathcal{B}_\mathsf{S}$, (iii) factorizes the cross-covariances to other participants and inserts them in the dictionary $\mathcal{C}_\mathsf{S}$, and (iv) inserts the corrected belief into $\mathcal{X}_\mathsf{S}$. Details are shown in Algorithm 7.14. Without loss of generality, the inter-agent bi-joint observation (see Algorithm 7.14) can be extended to support an arbitrary number of other agents' estimator instances.

### 7.4.4 Communication

Isolated propagation (see Algorithm 7.12), private observation (see Algorithm 7.9), and local joint observations steps require no inter-agent communication, since all information is directly accessible via the local handler $\mathsf{H}_\mathsf{A}$. Isolated inter-agent joint observation require an information exchange with estimator instances maintained on different agents.

In joint observations, the interim master requires information from participating estimator instances. The local handler $\mathsf{H}_{\mathsf{A}_i}$, knows according to the sensor identifier $\mathsf{id}_{\mathsf{S}_o}$, if the requested information is (locally) maintained in $\mathcal{M}_{\mathsf{H}_{\mathsf{A}_i}}$. If the instance is not available locally, the information must be obtained from the appropriate agent, by a lookup in $\mathsf{id}_{\mathsf{A}_p} = \mathcal{L}_{\mathsf{H}_{\mathsf{A}_i}}(\mathsf{id}_{\mathsf{S}_o})$ (see Algorithm 7.5). This dictionary needs to be kept up to date, i.e. agents need to refresh it by requesting a list of estimator identifiers from agents in communication range (see Algorithm 7.2), e.g., whenever an entry was not found. If an observation was delayed, then all interim measurements have to be applied in the correct order (processed again). In case of joint inter-agent observations, all measurements from post-correlated instances need to be re-processed in order, as discussed in Section 6.3.4.

In total, the collaborative instance handler needs to support two blocking (synchronous) requests and two asynchronous messages. One blocking request is needed to obtain the list of unique estimator identifiers from other agents. In the worst case, the message would contain $|\mathbf{S}|$ elements or if a list is requested from $|\mathbf{A}|$ agents, $|\mathbf{S}|$ elements are sent to the requesting agent. The second blocking request is needed to obtain the participating estimator instance's $\mathsf{S}_o$ *a-priori* belief (mean and covariance) and all factorized cross-covariances to other participants $\boldsymbol{\mathcal{S}}_{op} := \{\mathcal{S}_{oi}, i \in \mathbf{P}\}$, with $\mathbf{P}$ being the set of participants. The message size is $m(1+m+m|\mathbf{P}|)$ elements, where $|\mathbf{P}| \leq |\mathbf{A}|$. An asynchronous (non-blocking) message is required to update the *a-posteriori* covariance and factorized cross-covariances to participating estimators, and a state correction. The size of this message is $m(1+m+m|\mathbf{P}|)$ elements (see Algorithm 7.14). The second asynchronous message is needed to inform another agent to reprocess measurements after a specific timestamp, requiring only a single element in the message, see Algorithm 7.3.

In summary, the elements described above for our DC-MMSF approach allow a delay-compensated overarching framework seamlessly merging collaborative decentralized state estimation with modular multi-sensor fusion under a single umbrella of IKF.

### 7.4.5 Algorithm

---

**Algorithm 7.1:** DC-MMSF:$\mathsf{H}_{\mathsf{A}i}$: new sensor observation

**Input** : $\{\boldsymbol{\mathcal{X}}, \boldsymbol{\mathcal{C}}, \boldsymbol{\mathcal{B}}, \boldsymbol{\mathcal{M}}, \boldsymbol{\mathcal{Z}}\}_{\mathsf{H}_{\mathsf{A}i}}$, M

1  $\{t^k, \mathsf{id}_{\mathsf{S}_i}, \mathbf{z}_{\mathsf{S}_i}^k, \mathbf{R}_{\mathsf{S}_i}^k, \mathsf{Type}\} = \mathsf{M}$

2  $\mathsf{S}_i = \boldsymbol{\mathcal{M}}_{\mathsf{H}_{\mathsf{A}i}}(\mathsf{id}_{\mathsf{S}_i})$

3  /* process measurement as sensor specific observation * /

4  /* e.g. propagation, private, joint (bi, tri, quad, ...) */

5  rejected = $\mathsf{S}_i \to$ process_sensor_measurement(M) (Alg. 7.4)

6  **if** *!rejected* **then**

7  $\quad$ redo_updates_after_t($\{\boldsymbol{\mathcal{X}}, \boldsymbol{\mathcal{C}}, \boldsymbol{\mathcal{B}}, \boldsymbol{\mathcal{M}}, \boldsymbol{\mathcal{Z}}\}_{\mathsf{H}_{\mathsf{A}i}}, t^k$) (Alg. 7.3)

8  **end**

9  $\boldsymbol{\mathcal{Z}}(t^k) = \mathsf{M}$   // store sensor measurement

---

**Algorithm 7.2:** DC-MMSF:$\mathsf{H}_{\mathsf{A}i}$: referesh_lookup_table

**Input** : $\mathsf{H}_{\mathsf{A}i}$

**Output:** $\boldsymbol{\mathcal{L}}$

1  $\boldsymbol{\mathcal{L}} = \{\}$

2  **for** $\mathsf{A}_o \in \mathbf{A}$ *in communication range* **do**

3  $\quad$ /* request a list of unique estimator identifiers from agent $\mathsf{A}_o$ */

4  $\quad$ receive from $\mathsf{A}_o$: $\{\mathsf{l}_o = \mathsf{H}_{\mathsf{A}_o} \to \boldsymbol{\mathcal{M}}_{\mathsf{H}_{\mathsf{A}_o}} \to \mathsf{keys}()\}$

5  $\quad$ $\boldsymbol{\mathcal{L}}(\mathsf{l}_o) = \mathsf{id}_{\mathsf{A}_o}$

6  **end**

7  /* update local dictionary */

8  $\mathsf{H}_{\mathsf{A}i} \to \boldsymbol{\mathcal{L}}_{\mathsf{H}_{\mathsf{A}_o}} = \boldsymbol{\mathcal{L}}$

---

---

**Algorithm 7.3:** DC-MMSF: $\mathsf{H}_{\mathsf{A}_i}$: redo_updates_after_t

---

**Input** : $\{\boldsymbol{\mathcal{X}}, \boldsymbol{\mathcal{C}}, \boldsymbol{\mathcal{B}}, \boldsymbol{\mathcal{M}}, \boldsymbol{\mathcal{Z}}\}_{\mathsf{H}_{\mathsf{A}}i}, t^k$

1  /* delete newer corrections, cross-cov and states! */
2  delete($\{\boldsymbol{\mathcal{X}}_{\mathsf{H}_{\mathsf{A}}i}, \boldsymbol{\mathcal{C}}_{\mathsf{H}_{\mathsf{A}}i}, \boldsymbol{\mathcal{B}}_{\mathsf{H}_{\mathsf{A}}i}\} > t^k$)
3  /* redo existing observation after $t^k$! */
4  **for** $\mathsf{M}_i$ **in** $\{sort(\boldsymbol{\mathcal{Z}}_{\mathsf{H}_{\mathsf{A}}i}) > t^k\}$ **do**
5      $\{t^i, \mathsf{id}_{\mathsf{S}_i}, \mathbf{z}^i_{\mathsf{S}_i}, \mathbf{R}^i_{\mathsf{S}_i}, \mathsf{Type}\} = \mathsf{M}_i$
6      $\mathsf{S}_l = \boldsymbol{\mathcal{M}}_{\mathsf{H}_{\mathsf{A}}i}(\mathsf{id}_{\mathsf{S}_l})$
7      $\{\mathsf{rejected}_i, \mathbf{P}\} = \mathsf{S}_l \rightarrow \mathsf{process\_sensor\_measurement}(\mathsf{M}_i)$ (Alg. 7.4)
8      **if** *not rejected$_i$* **then**
9          /* simplification made: redo existing observation after $t^i$ asynchronously on all known agents instead of post-correlated ones */
10         /* assuming that $\boldsymbol{\mathcal{L}}_{\mathsf{H}_{\mathsf{A}}i}$ is up to date: */
11         **for** $\mathsf{A}_o \in \boldsymbol{\mathcal{L}}_{\mathsf{H}_{\mathsf{A}}i}$ **do**
12             send an asynchronous request to $\mathsf{A}_o$: redo_updates_after_t, $t^i$
13         **end**
14     **end**
15 **end**

---

**Algorithm 7.4:** DC-MMSF: $\mathsf{S}_i$: process_sensor_measurement

---

**Input** : $\{\boldsymbol{\mathcal{X}}, \boldsymbol{\mathcal{C}}, \boldsymbol{\mathcal{B}}, \mathsf{id}\}_{\mathsf{S}_i}, \mathsf{M}$
**Output** : rejected, $\mathbf{P}$

1  $\mathbf{P} = \{\}$
2  /* Sensor estimator type specific methods ... */
3  $\{t^k, \mathsf{id}_{\mathsf{S}_i}, \mathbf{z}^k_{\mathsf{S}_i}, \mathbf{R}^k_{\mathsf{S}_i}, \mathsf{Type}\} = \mathsf{M}$
4  **if** $\mathsf{Type} \equiv$ *proprioceptive or control* **then**
5      propagate($\ldots$) (Alg. 7.12)
6  **else if** $\mathsf{Type} \equiv$ *private* **then**
7      rejected = private_observation($\ldots$) (Alg. 7.9)
8  **else if** $\mathsf{Type} \equiv$ *joint_observation* **then**
9      **if** $\mathsf{Type} \equiv$ *inter-agent bi-observation* **then**
10         $\{\mathsf{rejected}, \mathbf{P}\} =$ global_joint_observation($\ldots$) (E.g, Alg. 7.14)
11     **else**
12         $\{\mathsf{rejected}, \mathbf{P}\} =$ local_joint_observation($\ldots$)
13     **end**
14 **else**
15     rejected = True
16 **end**

---

**Algorithm 7.5:** DC-MMSF: $\mathsf{H_{A_i}}$: get_others_belief

---

**Input** : $\mathcal{M}_{\mathsf{H_{A_i}}}, \mathcal{L}_{\mathsf{H_{A_i}}}$, Type, $\mathrm{id}_{\mathsf{S_o}}, \mathbf{id}_{\mathbf{S_p}}, t^k$
**Output:** exists, $\mathbf{x}_o^k, \mathcal{S}_{op} := \mathrm{Dict}\{\mathrm{id}_\mathcal{S}, \mathcal{S}\}$

**1** exist = True
**2** **if** $\mathrm{id}_{\mathsf{S_o}} \in \mathcal{M}_{\mathsf{H_{A_i}}}$ **then**
**3**    /* obtain information from a local estimator instance */
**4**    $\mathsf{S}_o = \mathcal{M}_{\mathsf{H_{A_i}}}(\mathrm{id}_{\mathsf{S_o}})$
**5**    $\mathbf{x}_o^k = $ get_belief$(\mathsf{S}_o, t^k)$
**6**    $\mathcal{S}_{op} = $ get_cross_factors$(\mathsf{S}_o, \mathbf{id}_{\mathbf{S_p}}, t^k)$ (Alg. 7.10)
**7** **else**
**8**    /* obtain information from a remote estimator instance */
**9**    /* assuming that $\mathcal{L}_{\mathsf{H_{A_i}}}$ is not up to date: */
**10**    refresh_lookup_table() (Alg. 7.2)
**11**    $\mathrm{id}_{\mathsf{A_j}} = \mathcal{L}_{\mathsf{H_{A_i}}}(\mathrm{id}_{\mathsf{S_o}})$
**12**    request from $\mathsf{A}_j$: $\{\mathbf{x}_o^k, \mathcal{S}_{op}, \mathrm{exist}\}$ providing:$\{\mathbf{id}_{\mathbf{S_p}}, t^k\}$ (Alg. 7.5)
**13** **end**

---

**Algorithm 7.6:** DC-MMSF: check_NIS

---

**Input** : $\mathbf{r}, \mathbf{S}, p = 0.997$
**Output:** outlier

**1** $s = \mathbf{r}^\mathsf{T} \mathbf{S}^{-1} \mathbf{r}$ // Mahalanobis distance squared
**2** DoF = length$(\mathbf{r})$ (Degrees of freedom)
**3** /* Inverse of the chi-square cumulative distribution */
**4** outlier = $(s > \mathrm{chi2inv}(p, \mathrm{DoF}))$

---

**Algorithm 7.7:** DC-MMSF:$\mathsf{S}_i$: get_belief

---

**Input** : $\mathcal{X}_{\mathsf{S}_i}, t^k$
**Output:** $\mathbf{x}_i^k$

**1** **if** *not exist*$(\mathcal{X}_{\mathsf{S}_i}(t^k))$ **then**
**2**    $\{\hat{\mathbf{x}}_i^a, \mathbf{\Sigma}_{ii}^a, t^a\} = \max(\mathrm{find}(\mathcal{X}_{\mathsf{S}_i} < t^k))$
**3**    /* predict from previous state to current timestamp */
**4**    $\mathbf{x}_i^k = $ propagate$_i(\ldots, t^a, t^k)$ (Alg. 7.12)
**5** **else**
**6**    $\{\hat{\mathbf{x}}_i^k, \mathbf{\Sigma}_{ii}^k\} = \mathbf{x}_i^k = \mathcal{X}_{\mathsf{S}_i}(t^k)$
**7** **end**

---

**Algorithm 7.8:** DC-MMSF:$\mathsf{S}_i$: compute_correction

---

**Input** : $\mathcal{B}_{\mathsf{S}_i}, t^a, t^k$
**Output:** $\mathbf{M}_{ii}^{k|a}$

**1** $\mathbf{M}_{ii}^{k|a} = \mathbf{I}$
**2** $t^b = \min(\mathcal{B}_{\mathsf{S}_i} > t^a)($
**3** **for** $l \leftarrow t^b$ **to** $t^k$ **do**
**4**    $\mathbf{M}_{ii}^{k|a} = \mathcal{B}_{\mathsf{S}_i}(l)\mathbf{M}_{ii}^{k|a}$
**5** **end**

---

**Algorithm 7.9:** DC-MMSF:$\mathsf{S}_i$ : private_observation

**Input**  : $\{\{\boldsymbol{\mathcal{X}}, \boldsymbol{\mathcal{C}}, \boldsymbol{\mathcal{B}}, \mathsf{id}\}_{\mathsf{S}_i}, t^k, \mathbf{z}^k, \mathbf{R}^k\}$

**Output**: $\{\boldsymbol{\mathcal{X}}, \boldsymbol{\mathcal{C}}, \boldsymbol{\mathcal{B}}, \mathsf{id}\}_{\mathsf{S}_i}$, rejected

1   /* get existing beliefs or predict beliefs */

2   $\hat{\mathbf{x}}_i^{k(-)}, \boldsymbol{\Sigma}_{ii}^{k(-)} = \text{get\_belief}(\boldsymbol{\mathcal{X}}_{\mathsf{S}_i}, \boldsymbol{\mathcal{B}}_{\mathsf{S}_i}, t^k)$ (Alg. 7.7)

3   $\mathbf{H}^k = \left[ \frac{\partial h(\mathbf{x}_{s_i})}{\partial \mathbf{x}_{s_i}} \Big|_{\hat{\mathbf{x}}_i} \right]^{k(-)}$

4   $\mathbf{S}^k = \mathbf{H}^k \boldsymbol{\Sigma}_{ii}^{k(-)} \left(\mathbf{H}^k\right)^{\mathsf{T}} + \mathbf{R}^k$

5   $\mathbf{K}^k = \boldsymbol{\Sigma}_{ii}^{k(-)} \left(\mathbf{H}^k\right)^{\mathsf{T}} (\mathbf{S}^k)^{-1}$

6   $\mathbf{r} = \boxminus h(\hat{\mathbf{x}}_i^{k(-)})) \boxplus \mathbf{z}^k$

7   rejected = check_NIS$(\mathbf{r}, \mathbf{S}^k)$ (Alg. 7.6)

8   **if** *not rejected* **then**

9      $\hat{\mathbf{x}}_i^{k(+)} = \hat{\mathbf{x}}_{\mathsf{S}_i}^{k(-)} \boxplus \mathbf{K}^k \mathbf{r}$

10     $\boldsymbol{\Sigma}_{ii}^{k(+)} = (\mathbf{I} - \mathbf{K}^k \mathbf{H}^k)\boldsymbol{\Sigma}_{ii}^{k(-)}$

11     $\boldsymbol{\Upsilon}_{ii}^k = \boldsymbol{\Sigma}_{ii}^{k(+)} \left(\boldsymbol{\Sigma}_{ii}^{k(-)}\right)^{-1}$

12     $\boldsymbol{\mathcal{B}}_{\mathsf{S}_i}\left(t^k\right) = \boldsymbol{\Upsilon}_{ii}^k \boldsymbol{\mathcal{B}}_{\mathsf{S}_i}\left(t^k\right)$

13     $\boldsymbol{\mathcal{X}}_{\mathsf{S}_i}(t^k) = \{\hat{\mathbf{x}}_{\mathsf{S}_i}^{k(+)}, \boldsymbol{\Sigma}_{ii}^{k(+)}\}$

14  **end**

---

**Algorithm 7.10:** DC-MMSF:$\mathsf{S}_i$: get_cross_factors

**Input**  : $\boldsymbol{\mathcal{C}}_{\mathsf{S}_i}, \boldsymbol{\mathcal{B}}_{\mathsf{S}_i}, \mathbf{id}_{\mathsf{S}_o}, t^k$

**Output**: $\boldsymbol{\mathcal{S}}_{io} := \text{Dict}\{\mathsf{id}_{\mathsf{S}}, \mathcal{S}\}$

1   **for** $\{\mathsf{id}_o\}$ in $\mathbf{id}_{\mathsf{S}_o}$ **do**

2     **if** $\mathsf{id}_o \in \boldsymbol{\mathcal{C}}_i$ **then**

3       $\boldsymbol{\mathcal{S}}_{io}(\mathsf{id}_o) = \text{get\_cross\_cov\_factor}(\boldsymbol{\mathcal{C}}_{\mathsf{S}_i}, \boldsymbol{\mathcal{B}}_{\mathsf{S}_i}, \mathsf{id}_o, t^k)$ (Alg. 7.13)

4     **end**

5   **end**

---

**Algorithm 7.11:** DC-MMSF:$\mathsf{S}_i$: check_horizon

**Input**  : $\boldsymbol{\mathcal{B}}_{\mathsf{S}_i}, \boldsymbol{\mathcal{C}}_{\mathsf{S}_i}, t^k$

1   $t^o = \min(\text{find}(\boldsymbol{\mathcal{B}}_{\mathsf{S}_i} < t^k))$ (oldest correction term)

2   $t^m = t^o + (t^k - t^o)/2$ (half of the time horizon)

3   **for** $\{\mathsf{id}_{\mathsf{S}_j}\}$ in $\boldsymbol{\mathcal{C}}_{\mathsf{S}_i}$ **do**

4     **if** *find*$(\boldsymbol{\mathcal{C}}(\mathsf{id}_{\mathsf{S}_j})) \equiv t^o$ **then**

5       $\mathbf{M}_{ii}^{m|o} = \text{compute\_corr}(\boldsymbol{\mathcal{B}}_{\mathsf{S}_i}, t^o, t^m)$ (Alg. 7.8)

6       $\boldsymbol{\mathcal{S}}_{ij}^{o(-)} = \boldsymbol{\mathcal{C}}_{\mathsf{S}_i}(\mathsf{id}_{\mathsf{S}_j})(t^o)$

7       $\boldsymbol{\mathcal{C}}_{\mathsf{S}_i}(\mathsf{id}_{\mathsf{S}_j}) = \{\mathbf{M}_{ii}^{m|o} \boldsymbol{\mathcal{S}}_{ij}^{o(-)}, t^k\}$ (forward prop.)

8     **end**

9   **end**

---

**Algorithm 7.12:** DC-MMSF:$\mathsf{S}_i$: Propagation

**Input** : $\{\boldsymbol{\mathcal{B}}, \boldsymbol{\mathcal{C}}, \boldsymbol{\mathcal{X}}\}_{\mathsf{S}_i}, \mathbf{u}^k, \mathbf{N}_{ii}^k, t^{k-1}, t^k$

**Output:** $\hat{\mathbf{x}}_i^k, \boldsymbol{\Sigma}_{ii}^k, \{\boldsymbol{\mathcal{X}}, \boldsymbol{\mathcal{B}}\}_{\mathsf{S}_i}$

1   $\{\hat{\mathbf{x}}_i^{k-1}, \boldsymbol{\Sigma}_{ii}^{k-1}\} = \boldsymbol{\mathcal{X}}_{\mathsf{S}_i}(t^{k-1})$

2   $\boldsymbol{\Phi}_{ii}^{k|k-1} = \left[ \frac{\partial \phi_i(\mathbf{x}_p, \mathbf{u})}{\partial \mathbf{x}_p} \Big|_{\hat{\mathbf{x}}_i, \mathbf{u}} \right]^{k|k-1}$

3   $\mathbf{G}_{ii}^{k|k-1} = \left[ \frac{\partial \phi_i(\mathbf{x}_p, \mathbf{u})}{\partial \mathbf{u}} \Big|_{\hat{\mathbf{x}}_i, \mathbf{u}} \right]^{k|k-1}$

4   $\mathbf{Q}_{ii}^{k|k-1} = \mathbf{G}_{ii}^{k|k-1} \mathbf{N}_{ii}^k (\mathbf{G}_{ii}^{k|k-1})^{\mathsf{T}}$

5   $\hat{\mathbf{x}}_i^k = \phi_i(\hat{\mathbf{x}}_i^{k-1}, \mathbf{u}^k)$

6   $\boldsymbol{\Sigma}_{ii}^k = \boldsymbol{\Phi}_{ii}^{k|k-1} \boldsymbol{\Sigma}_{ii}^{k-1} (\boldsymbol{\Phi}_{ii}^{k|k-1})^{\mathsf{T}} + \mathbf{Q}_{ii}^{k|k-1}$

7   $\boldsymbol{\mathcal{B}}_{\mathsf{S}_i}\left(t^k\right) = \boldsymbol{\Phi}_{ii}^{k|k-1}$   // insert into sorted buffer

8   $\boldsymbol{\mathcal{X}}_{\mathsf{S}_i}(t^k) = \{\hat{\mathbf{x}}_i^k, \boldsymbol{\Sigma}_{ii}^k\}$   // insert belief sorted

9   check_horizon($\boldsymbol{\mathcal{B}}_{\mathsf{S}_i}, \boldsymbol{\mathcal{C}}_{\mathsf{S}_i}, t^k$) (Alg. 7.11)

---

**Algorithm 7.13:** DC-MMSF:$\mathsf{S}_i$: get_cross_cov_factor

**Input** : $\boldsymbol{\mathcal{C}}_{\mathsf{S}_i}, \boldsymbol{\mathcal{B}}_{\mathsf{S}_i}, \mathsf{id}_{\mathsf{S}_o}, t^k$

**Output:** $\boldsymbol{\mathcal{S}}_{io}^k$

1   **if** $\mathsf{id}_{\mathsf{S}_o} \in \boldsymbol{\mathcal{C}}_i$ **then**

2     **if** $not\ exist(\boldsymbol{\mathcal{C}}_{\mathsf{S}_i}(\mathsf{id}_{\mathsf{S}_o}, t^k))$ **then**

3       /* get the latest factor and forward propagate it */

4       $\{\boldsymbol{\mathcal{S}}_{io}^m, t^m\} = \max(\mathrm{find}(\boldsymbol{\mathcal{C}}_{\mathsf{S}_i}(\mathsf{id}_{\mathsf{S}_o}) < t^k))$

5       $\mathbf{M}^{k|m} = \text{compute\_corr}(\boldsymbol{\mathcal{B}}_{\mathsf{S}_i}, t^m, t^k)$ (Alg. 7.8)

6       $\boldsymbol{\mathcal{S}}_{io}^k = \mathbf{M}^{k|m} \boldsymbol{\mathcal{S}}_{io}^m$

7       $\boldsymbol{\mathcal{C}}_{\mathsf{S}_i}(\mathsf{id}_{\mathsf{S}_o}, t^k) = \boldsymbol{\mathcal{S}}_{io}^k$

8     **else**

9       $\boldsymbol{\mathcal{S}}_{io}^k = \boldsymbol{\mathcal{C}}_{\mathsf{S}_i}(\mathsf{id}_{\mathsf{S}_o}, t^k)$

10    **end**

11 **end**

**Algorithm 7.14:** DC-MMSF:$\mathsf{S}_{\{m,n\}}$: Inter-Agent bi-sensor observation

---

**Input** : $\{\boldsymbol{\mathcal{X}}, \boldsymbol{\mathcal{C}}, \boldsymbol{\mathcal{B}}, \mathsf{id}\}_{\{\mathsf{S}_m, \mathsf{S}_n\}}, t^k, \mathbf{z}^k, \mathbf{R}^k, \mathrm{Type}$
**Output**: $\{\boldsymbol{\mathcal{X}}, \boldsymbol{\mathcal{C}}, \boldsymbol{\mathcal{B}}, \mathsf{id}\}_{\{\mathsf{S}_m, \mathsf{S}_n\}}, \mathrm{rejected}, \mathbf{P}$

1   // Type: Agent $\mathsf{A}_i$ detects $\mathsf{A}_j$, involving $\mathsf{S}_m$ and $\mathsf{S}_n$

2   $\mathbf{P} = \{\mathsf{id}_{\mathsf{A}_j}\}$ /* set of participating agents */

3   **if** $\mathsf{S}_m \in \mathsf{A}_i$ *observes* $\mathsf{S}_n \in \mathsf{A}_j$ *at* $t^k$ **then**

4      /* Agent $\mathsf{A}_i$ is interim master*/

5      $\{\mathcal{S}_{\mathsf{S}_m \mathsf{S}_n}^m, t^m\} = \max(\mathrm{find}(\boldsymbol{\mathcal{C}}_{\mathsf{S}_m}(\mathsf{id}_{\mathsf{S}_n}) < t^k))$

6      $\mathbf{x}_{\mathsf{S}_m}^m = \mathrm{get\_belief}(\boldsymbol{\mathcal{X}}_{\mathsf{S}_m}, \boldsymbol{\mathcal{B}}_{\mathsf{S}_m}, t^m)$ (Alg. 7.7)

7      $\mathbf{M}_{\mathsf{S}_m}^{k|m} = \mathrm{compute\_correction}(\boldsymbol{\mathcal{B}}_{\mathsf{S}_m}, t^m, t^k)$ (Alg. 7.8)

8      $\mathcal{S}_{\mathsf{S}_m \mathsf{S}_n}^k = (\mathbf{M}_{\mathsf{S}_m}^{k|m} \mathcal{S}_{\mathsf{S}_m \mathsf{S}_n}^m)$

9      request from $\mathsf{A}_j$: $\{\mathbf{x}_{\mathsf{S}_n}^k, \mathcal{S}_{\mathsf{S}_n \mathsf{S}_m}^k\}$ providing:$\{\mathsf{id}_{\mathsf{S}_n}, t^k\}$ (Alg. 7.5)

10      $\boldsymbol{\Sigma}_{\mathsf{S}_m \mathsf{S}_n}^k = (\mathcal{S}_{\mathsf{S}_m \mathsf{S}_n}^k)(\mathcal{S}_{\mathsf{S}_n \mathsf{S}_m}^k)^\mathsf{T}$

11      $\boldsymbol{\Sigma}_{pp}^{k(-)} = \begin{bmatrix} \boldsymbol{\Sigma}_{\mathsf{S}_n} & \boldsymbol{\Sigma}_{\mathsf{S}_n \mathsf{S}_m} \\ \boldsymbol{\Sigma}_{\mathsf{S}_n \mathsf{S}_m}^\mathsf{T} & \boldsymbol{\Sigma}_{\mathsf{S}_m} \end{bmatrix}^k$

12      $\hat{\mathbf{x}}_p^{k(-)} = \begin{bmatrix} \hat{\mathbf{x}}_{\mathsf{S}_n} \\ \hat{\mathbf{x}}_{\mathsf{S}_m} \end{bmatrix}^k$

13      $\mathbf{H}_p = \begin{bmatrix} \frac{\partial h_{m,n}(\mathbf{x}_{\mathsf{S}_n}, \mathbf{x}_{\mathsf{S}_m})}{\partial \mathbf{x}_{\mathsf{S}_n}}\Big|_{\hat{\mathbf{x}}_p^{k(-)}} & \frac{\partial h_{m,n}(\mathbf{x}_{\mathsf{S}_n}, \mathbf{x}_{\mathsf{S}_m})}{\partial \mathbf{x}_{\mathsf{S}_m}}\Big|_{\hat{\mathbf{x}}_p^{k(-)}} \end{bmatrix}$

14      $\mathbf{S}_p = \mathbf{H}_p \boldsymbol{\Sigma}_{pp}^{k(-)} \mathbf{H}_p^\mathsf{T} + \mathbf{R}^k$

15      $\mathbf{r}_p = \boxminus h_{m,n}\left(\hat{\mathbf{x}}_p^{(-)}\right) \boxplus \mathbf{z}$

16      $\mathrm{rejected} = \mathrm{check\_NIS}(\mathbf{r}, \mathbf{S}_p)$ (Alg. 7.6)

17      **if** *not rejected* **then**

18         $\mathbf{K}_p = \boldsymbol{\Sigma}_{pp}^{k(-)} \mathbf{H}_p^\mathsf{T} (\mathbf{S}_p)^{-1}$

19         $\Delta \hat{\mathbf{x}}_p^{k(+)} = \mathbf{K}_p \mathbf{r}_p$

20         $\boldsymbol{\Sigma}_{pp}^{k(+)} = (\mathbf{I} - \mathbf{K}_p \mathbf{H}_p) \boldsymbol{\Sigma}_{pp}^{k(-)}$

21         /* Note: split $\boldsymbol{\Sigma}_{pp}^{k(+)}$ and $\Delta \hat{\mathbf{x}}_p^{k(+)}$ */

22         send to $\mathsf{A}_j$: $\{\Delta \hat{\mathbf{x}}_{\mathsf{S}_n}^{k(+)}, \boldsymbol{\Sigma}_{\mathsf{S}_n}^{k(+)}, \mathcal{S}_{\mathsf{S}_n \mathsf{S}_m}^k = \mathbf{I}, \mathrm{False}\}$

23         $\hat{\mathbf{x}}_{\mathsf{S}_m}^{k(+)} = \hat{\mathbf{x}}_{\mathsf{S}_m}^{k(-)} \boxplus \Delta \hat{\mathbf{x}}_{\mathsf{S}_m}^{k(+)}$

24         $\mathcal{S}_{\mathsf{S}_m \mathsf{S}_n}^{k(+)} = \boldsymbol{\Sigma}_{\mathsf{S}_m \mathsf{S}_n}^{k(+)}$

25         $\boldsymbol{\Lambda}_{\mathsf{S}_m}^k = \boldsymbol{\Sigma}_{\mathsf{S}_m}^{k(+)} (\boldsymbol{\Sigma}_{\mathsf{S}_m}^{k(-)})^{-1}$

26         $\boldsymbol{\mathcal{B}}_{\mathsf{S}_m}(t^k) = \boldsymbol{\Lambda}_{\mathsf{S}_m}^k \boldsymbol{\mathcal{B}}_{\mathsf{S}_m}(t^k)$

27         $\boldsymbol{\mathcal{C}}_{\mathsf{S}_m}(\mathsf{id}_{\mathsf{S}_n}) = \{\mathcal{S}_{\mathsf{S}_m \mathsf{S}_n}^{k(+)}, t^k\}$

28         $\boldsymbol{\mathcal{X}}_{\mathsf{S}_m}(t^k) = \{\hat{\mathbf{x}}_{\mathsf{S}_m}^{k(+)}, \boldsymbol{\Sigma}_{\mathsf{S}_m}^{k(+)}\}$

29      **else**

30         send to $\mathsf{A}_j$: rejected

31      **end**

32   **else**

33      /* get previous cross-covariance factors (Alg. 7.10)*/

34      $\{\mathcal{S}_{\mathsf{S}_n \mathsf{S}_m}^m, t^m\} = \max(\mathrm{find}(\boldsymbol{\mathcal{C}}_{\mathsf{S}_n}(\mathsf{id}_{\mathsf{S}_m}) < t^k))$

35      /* get existing beliefs or predict beliefs */

36      $\mathbf{x}_{\mathsf{S}_n}^m = \mathrm{get\_belief}(\boldsymbol{\mathcal{X}}_{\mathsf{S}_n}, \boldsymbol{\mathcal{B}}_{\mathsf{S}_n}, t^m)$ (Alg. 7.7)

37      $\mathbf{M}_{\mathsf{S}_n}^{k|m} = \mathrm{compute\_correction}(\boldsymbol{\mathcal{B}}_{\mathsf{S}_n}, t^m, t^k)$ (Alg. 7.8)

38      $\mathcal{S}_{\mathsf{S}_n \mathsf{S}_m}^k = (\mathbf{M}_{\mathsf{S}_n}^{k|m} \mathcal{S}_{\mathsf{S}_n \mathsf{S}_m}^m)$

39      send to $\mathsf{A}_i$: $\{\mathbf{x}_{\mathsf{S}_n}^k, \mathcal{S}_{\mathsf{S}_n \mathsf{S}_m}^k\}$

40      wait and receive from $\mathsf{A}_i$: $\{\Delta \hat{\mathbf{x}}_{\mathsf{S}_n}^{k(+)}, \boldsymbol{\Sigma}_{\mathsf{S}_n}^{k(+)}, \mathcal{S}_{\mathsf{S}_n \mathsf{S}_m}^{k(+)}, \mathrm{not\ rejected}\}$ or $\{\mathrm{rejected}\}$

41      **if** *not rejected* **then**

42         $\hat{\mathbf{x}}_{\mathsf{S}_n}^{k(+)} = \hat{\mathbf{x}}_{\mathsf{S}_n}^{k(-)} \boxplus \Delta \hat{\mathbf{x}}_{\mathsf{S}_n}^{k(+)}$

43         $\boldsymbol{\Lambda}_{\mathsf{S}_n}^k = \boldsymbol{\Sigma}_{\mathsf{S}_n}^{k(+)} (\boldsymbol{\Sigma}_{\mathsf{S}_n}^{k(-)})^{-1}$

44         $\boldsymbol{\mathcal{B}}_{\mathsf{S}_n}(t^k) = \boldsymbol{\Lambda}_{\mathsf{S}_n}^k \boldsymbol{\mathcal{B}}_{\mathsf{S}_n}(t^k)$

45         $\boldsymbol{\mathcal{C}}_{\mathsf{S}_n}(\mathsf{id}_{\mathsf{S}_m}) = \{\mathcal{S}_{\mathsf{S}_n \mathsf{S}_m}^{k(+)}, t^k\}$

46         $\boldsymbol{\mathcal{X}}_{\mathsf{S}_n}(t^k) = \{\hat{\mathbf{x}}_{\mathsf{S}_n}^{k(+)}, \boldsymbol{\Sigma}_{\mathsf{S}_n}^{k(+)}\}$

47      **end**

48   **end**

## 7.5   Evaluations

To evaluate the performance of DC-MMSF, we conducted experiments with our custom MATLAB framework, that allows to load real data from the EuRoC dataset [20] or to generate a dataset based on a synthetic trajectory. Exteroceptive measurements (private or joint observations) are generated based on the ground truth trajectory and are modified by the sensors' calibration states and noise parameters. Noisy and biased real-world IMU samples are provided by the datasets and used without modifications. This simulation environment ensures deterministic and repeatable results, and allows changing the sensor latency $t_{lat}$ of all sensors, which is important for a fair comparison between different fusion strategies and the ANEES evaluation.

Figure 7.3 depicts schematically how agents obtain their configurations and measurements. Once the sensor suite is defined or loaded from a dataset, the measurements can be processed in a multi-agent handler, which is again maintaining multiple handlers $\mathsf{H_A}$, while communication between them is handled locally.

Therefore, an optimal communication is assumed (no latency, packet loss, bandwidth limitations), as it would bias the evaluation of the processing time for different filter steps between distributed and centralized approaches. In a centralized architecture, all measurements would have to be streamed to a central fusion entity, resulting in even higher sensor latency, which further increases the processing time. The evaluation is performed single-threaded in MATLAB on an AMD Ryzen 7 3700X CPU with 32 GB DDR4 RAM.

The aim of the evaluation is twofold. In the first experiment $\mathsf{S_1}$, we compare the filter accuracy and credibility of the proposed DC-MMSF approach against a centralized equivalent estimator MDCSE-DP* and the MDCSE-DAH approach, which are described in Section 7.5.1. In the second experiment $\mathsf{S_2}$ we compare the execution time of the proposed approach against MDCSE-DAH.

Before the accuracy and credibility, and scalability are discussed in Sections 7.5.4 and 7.5.5, the sensor models and the state spaces are described in Section 7.5.2.

### 7.5.1   Architecture Overview

In the following evaluation we compare the filter accuracy and credibility of the proposed DC-MMSF approach against a centralized equivalent estimator, DCSE-DP* (see Section 3.4.4), and our DCSE-DAH (see Section 3.4.6) approach.

For that purpose, these approaches are ported and modified to support a generic sensor suite and delayed measurements. To avoid confusions, the modular variants are denoted as MDCSE-DAH and MDCSE-DP*, respectively. Note that MDCSE-DP* and MDCSE-DAH do not support adding or removing estimator instances during mission, as the correlations between agents would be violated, i.e. the state space of individual agent is not allowed to change. MDCSE-DP* is performing updates on the global full state, while the state propagation is performed isolated on each agent. Therefore, MDCSE-DP* can be considered as baseline approach performing statistically equivalent as a centralized EKF operating on the global full state.

The MDCSE-DAH approach uses per agent locally the MMSF-DP filter formulation (see Section 4.4) in the instance handler $\mathsf{H}$ which is embedded in collaborative instance handler $\mathsf{H_A}$. The $\mathsf{H_A}$ maintains additionally a history of factorized cross-covariances $\mathcal{C}_{\mathsf{H_A}}$ relating to other agents, a history of correction terms $\mathcal{B}_{\mathsf{H_A}}$, and a history of measurements $\mathcal{Z}_{\mathsf{H_A}}$, as depict in Figure 7.4, to support collaborative/inter-agent observations. Each physical sensor is associated with a filter node $\mathsf{S}$ that allows for isolated propagation steps, but private update steps are performed of the agent's local full state. Inter-agent observations require information exchange between participating agents' handlers $\mathsf{H_{A}}_i, i \in \mathsf{P}$

In the limit of a single sensor per agent, DC-MMSF behaves like MDCSE-DAH. The main difference between both is that the cross-covariance factors are maintained in the instance $\mathsf{S}$ per sensor in DC-MMSF, while in MDCSE-DAH they are maintained in the collaborative instance handler $\mathsf{H_A}$. This also means, if the $\mathsf{H_A}$ in DC-MMSF, maintains many sensor instances $\mathsf{S}$, DC-MMSF becomes more efficient, as it operates on sub-states of the agent's local full state.

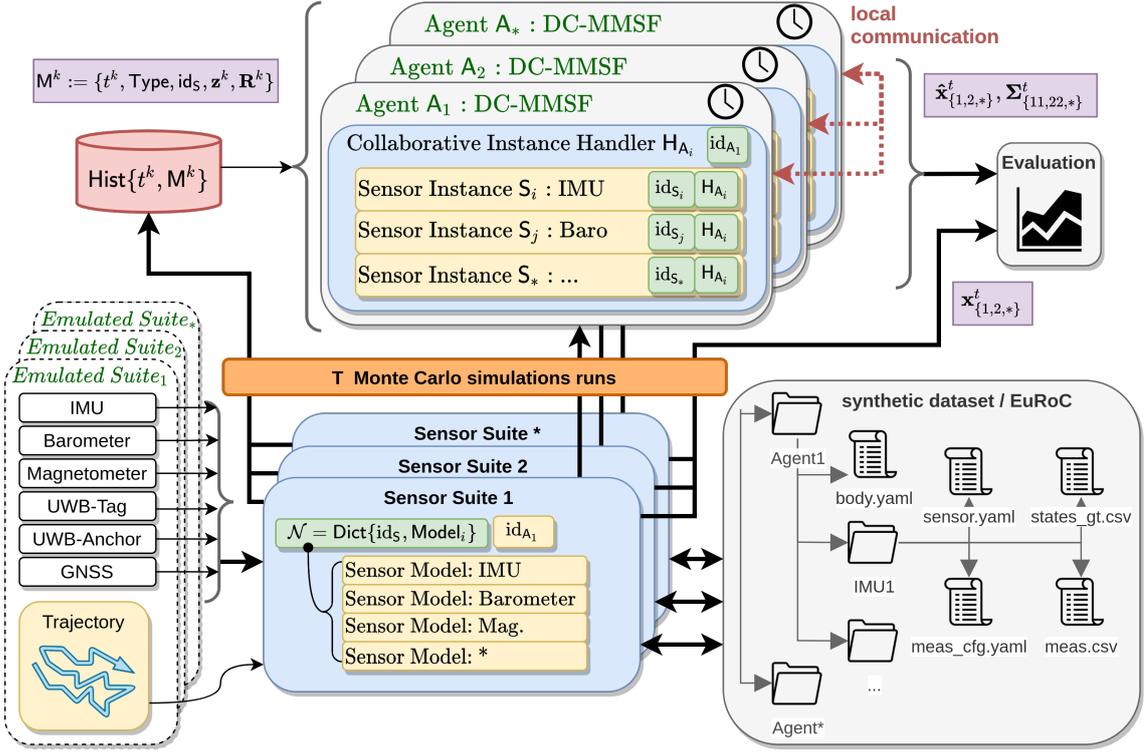

**Figure 7.3:** Shows the block diagram of the multi-agent simulation framework used in Section 7.5. Each agents' estimation framework is defined by a sensor suite consisting of multiple sensor models. It can be loaded and stored into a human-readable dataset or be generated from a set of emulated sensors. Each sensor suite allows generating a set of measurements, initial beliefs, and to obtain the true state, which is a prerequisite for the ANEES evaluation. Based on the sensor suite's sensor models, each agent's collaborative instance handler $\mathsf{H_A}$ instantiates an appropriate IKF instance $\mathsf{S}$. All measurements in $\mathsf{Hist}\{t^k, \mathsf{M}^k\}$ from the sensor suites are processed sequentially and chronologically sorted, while individual sensor delays are configurable. To guarantee reproducibility and a deterministic simulation behavior, agents perform local communication by sharing the same memory space.

### 7.5.2  Sensor Measurement Models and State Definitions

In our AINS, an indirect error estimation is performed [104], with $\mathbf{x} = \hat{\mathbf{x}} \boxplus \tilde{\mathbf{x}}$ (type-1 error). This requires observations to be expressed by their error $\tilde{\mathbf{z}} = \boxminus \hat{\mathbf{z}} \boxplus \mathbf{z}$. This measurement error needs to be linearized with respect to the error state $\tilde{\mathbf{x}}$ at the current estimate $\tilde{\mathbf{z}} = \mathbf{H}\tilde{\mathbf{x}}$ with the measurement Jacobian $\mathbf{H} = \frac{\partial h}{\partial \mathbf{x}} \frac{\mathbf{x}}{\partial \tilde{\mathbf{x}}}\big|_{\hat{\mathbf{x}}}$ for the measurement function $\mathbf{z} = h(\mathbf{x})$. Details on the error-state definition, the measurement models and Jacobians can be found in Section 2.9.

Figure 7.5 shows the spatial frame constellation of our experiments. Each agent's sensor (e.g., IMU $\mathcal{I}$, barometer $\mathcal{P}$, ranging tag $\mathcal{T}$) is rigidly attached with respect to a body reference frame $\mathcal{B}$ and requires a spatial calibration state, e.g., $^{\mathcal{B}}_{\mathcal{B}}\mathbf{p}_{\mathcal{R}}$ and $^{k}\mathbf{R}_{\mathcal{R}}$ specifying the translation and orientation of the sensor reference frame $\mathcal{R}$ with respect to the body.

Stationary ranging anchors are modeled as individual agents with communication and processing capabilities, where each estimates – as spatial calibration state – the position between the global reference frame $\{\mathcal{G}\}$ and the origin of anchor position $\{\mathcal{A}\}$ as $_{\mathcal{G}}^{\mathcal{G}}\mathbf{p}_{\mathcal{A}}$. An error in these parameters degrades the estimation performance, and an inexact offline calibration may lead to unmodeled errors and estimation inconsistencies [36]. Therefore, we treat observable calibration parameters as random variables that are estimated in the corresponding estimator instance, allowing for online self-calibration. Further, it allows modeling flexible and time-varying relative constraints (spatio-temporal extrinsics) between the sensors and the body reference, e.g., by a random walk process.

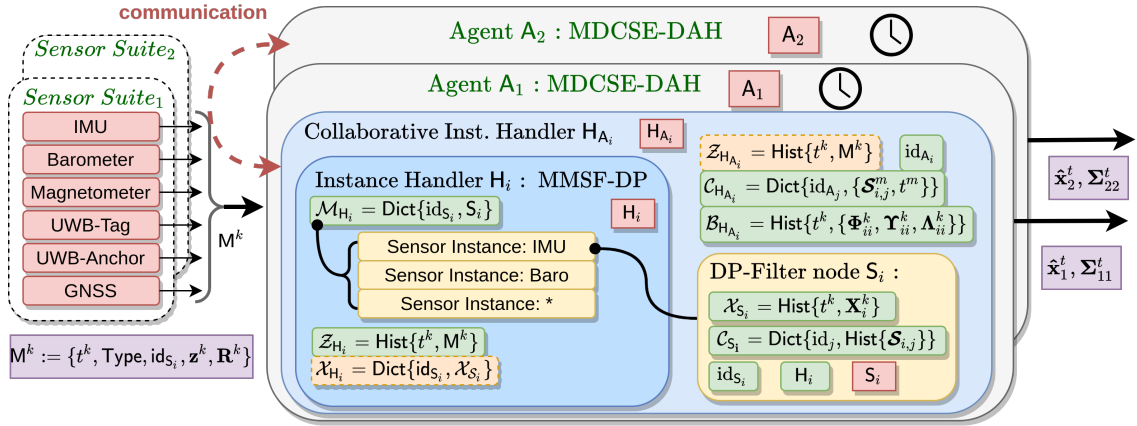

**Figure 7.4:** MDCSE-DAH: Shows the multi-sensor fusion framework of individual agents, consisting of a collaborative instance handler $H_A$ maintaining a MMSF-DP handler $H$, which again maintains various estimator instances $S$ for a specific type. The sensor suite provides measurements to the instance handler, which delegates them to the appropriate filter instance, which performs the filter steps exactly (considering locally available estimates). For inter-agent joint observation, information exchange between participating handlers $H_A$ is needed.

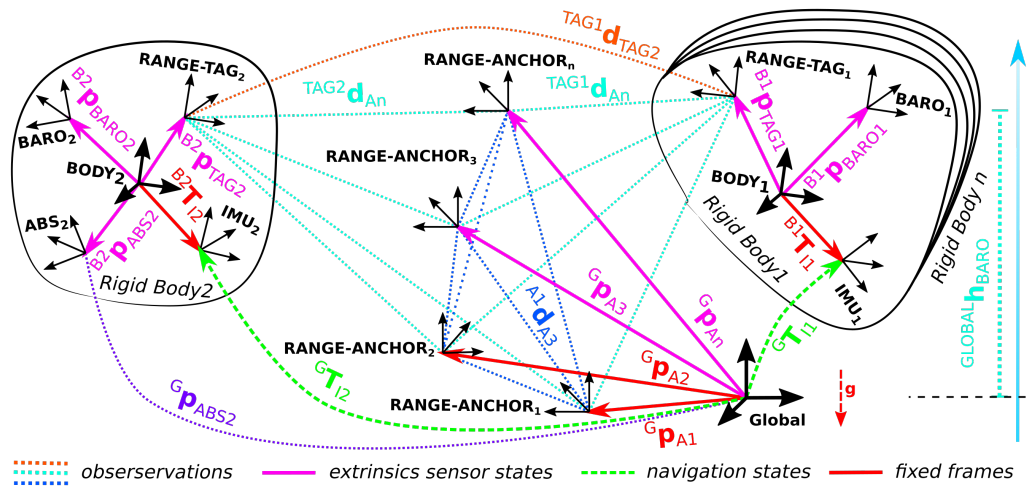

**Figure 7.5:** Spatial frame constellation of a multiagent system navigating in space using range measurements between stationary ranging anchors to recover the 6-DoF pose on each agent. Additional relative range measurements between agents' ranging tags, as well as pressure readings from a barometer are used.

The IMU is used as state propagation sensor, with the state $\mathbf{x}_{\mathcal{I}}$ defined as

$$\mathbf{x}_{\mathcal{I}} = \begin{bmatrix} _{\mathcal{G}}^{\mathcal{G}}\mathbf{p}_{\mathcal{I}}; {}^{\mathcal{G}}\mathbf{v}_{\mathcal{I}}; {}^{\mathcal{G}}\mathbf{q}_{\mathcal{I}}; {}_{\mathcal{I}}\mathbf{b}_{\omega}; {}_{\mathcal{I}}\mathbf{b}_a \end{bmatrix} \tag{7.1a}$$

with ${}^{\mathcal{G}}_{\mathcal{G}}\mathbf{p}_{\mathcal{I}}, {}^{\mathcal{G}}_{\mathcal{G}}\mathbf{v}_{\mathcal{I}}$, and ${}^{\mathcal{G}}\mathbf{q}_{\mathcal{I}}$ as the position, velocity, and orientation of the IMU $\mathcal{I}$ w.r.t. to the global navigation frame $\mathcal{G}$. ${}_{\mathcal{I}}\mathbf{b}_{\omega}$ and ${}_{\mathcal{I}}\mathbf{b}_{a}$ are the estimated gyroscope and accelerometer biases to correct the related IMU readings (see Equation (2.79)). The unobservable pose between the body frame $\{\mathcal{B}\}$ and IMU $\{\mathcal{I}\}$, ${}^{\mathcal{B}}_{\mathcal{B}}\mathbf{p}_{\mathcal{I}}$ and ${}^{\mathcal{B}}\mathbf{q}_{\mathcal{I}}$, are constants that need to be known *a-priori* . The gravity vector is assumed to be aligned with the z-axis of the navigation frame ${}_{\mathcal{G}}\mathbf{g} = [0,0,9.81]^{\mathsf{T}}$.

The barometer measures the local absolute pressure and its position ${}^{\mathcal{B}}_{\mathcal{B}}\mathbf{p}_{\mathcal{P}}$ with respect to the body reference frame leads to the sensor extrinsic state $\mathbf{x}_{\mathcal{P}}$, which is defined as

$$\mathbf{x}_{\mathcal{P}} = \begin{bmatrix} {}^{\mathcal{B}}_{\mathcal{B}}\mathbf{p}_{\mathcal{P}} \end{bmatrix} \tag{7.1b}$$

with ${}^{\mathcal{B}}_{\mathcal{B}}\mathbf{p}_{\mathcal{P}}$ being the position between the body reference frame and the sensor, which is assumed to be fixed and known *a-priori* . The measurement Jacobian and further details can be found in Section 2.9.4.

The range sensor measures the distance between two devices, e.g., a ranging device based on UWB measures the distance between two antennas. For the moving ranging tags $\{\mathcal{T}\}$ and stationary ranging anchors $\{\mathcal{A}\}$, a spatial displacement to a reference frame is estimated. The anchor state $\mathbf{x}_{\mathcal{A}}$ is

$$\mathbf{x}_{\mathcal{A}} = \begin{bmatrix} {}^{\mathcal{G}}_{\mathcal{G}}\mathbf{p}_{\mathcal{A}} \end{bmatrix}, \tag{7.1c}$$

which is not changing in time ${}^{\mathcal{G}}_{\mathcal{G}}\dot{\mathbf{p}}_{\mathcal{A}} = \mathbf{0}$. The tag state $\mathbf{x}_{\mathcal{T}}$ is

$$\mathbf{x}_{\mathcal{T}} = \begin{bmatrix} {}^{\mathcal{B}}_{\mathcal{B}}\mathbf{p}_{\mathcal{T}} \end{bmatrix} \tag{7.1d}$$

with ${}^{\mathcal{B}}_{\mathcal{B}}\mathbf{p}_{\mathcal{T}}$ being the transformation between the body reference frame $\mathcal{B}$ and the UWB tag. The spatial displacement is not changing in time ${}^{\mathcal{B}}_{\mathcal{B}}\dot{\mathbf{p}}_{\mathcal{T}} = \mathbf{0}$. The measurement Jacobian and further details can be found in Section 2.9.6.

### 7.5.3 Nonlinear Observability Analysis

Since the proposed experiment is a non-trivial estimation problem with different (interagent) joint observations and an IMU as a state propagation sensor, an nonlinear observability analysis (see Section 2.9.7) helps to understand the properties of the estimation problem at hand, but does not reflect the behavior of the linearized or actual system. Previous works, e.g., [17, 49], have shown the utility of a non-linear observability analysis of the estimation problem in predicting/interpreting the latter evaluation result. Without loss of generality, this analysis can be applied to multiagent systems, by building a full state vector from the individual agent's estimators and using the system dynamics and measurement models appropriately. By determining the rank of the observability matrix of the noise-free nonlinear system, we can analyze, if the system is local weakly observable or not. Note that unobservable dimensions can be states or a combination of them. In this section, we apply the nonlinear observability analysis to the estimation problem depicted in Figure 7.5 with different numbers of agents and anchors. Three cases are of interest: (i) a single agent with $N$ anchors, (ii) multiple agents ($K$) without anchors, and (iii) multiple agents ($K$) with $N$ anchors. In the analyses, we assume an excitation in all three axes of the IMU's accelerometer and gyroscope, and that the gravity vector is known and aligned with the z-axis of the navigation frame.

#### Single Agent with Anchors

At first, we study the sensor estimators observability proprieties in case of a single agent measuring ranges (T-A) to $K$ anchors (see Table 7.1). The full state vector has four unobservable dimensions, involving the absolute position and yaw orientation (roll and pitch are observable in a VINS [118]) of the whole system with respect to the global navigation frame $\{\mathcal{G}\}$. In other words, one can estimate the relative constellation of the agent and the anchors. Once the agent obtains absolute position measurements, e.g., provided by a GNSS sensor, the full state vector is observable. This is due to the yaw orientation being available through the positional change – a sort of pseudo measurement. Note that this information propagates via (joint) range measurements between the tag and the anchors to the anchors' estimators. On the other hand, the same effect is achievable by fixing a certain number of anchors in the global frame. Fixing in this context means, to assume that the state is known exactly and treated as known constant. Fixing one anchor makes the absolute position of the full state observable and leaves the global yaw orientation unobservable, which is rendered observable by fixing a second anchor.

| Single agent (Tag+IMU) with $K$ anchors | | | | |
|---|---|---|---|---|
| A-A | Absolute | num. fixed A | | |
| (meshed) | position | 0 | 1 | 2 |
| 0 | 0 | 4 | 1 | 0 |
| 0 | 1 | 0 | 0 | 0 |
| 1 | 0 | 4 | 1 | 0 |

**Table 7.1:** Overview of unobservable dimensions of a single agent with an IMU for propagation and an ranging tag observing $K$ anchors. Optionally, anchors can perform measurements (fully meshed) among themselves (A-A), anchors can be set to be known globally ($^{\mathcal{G}}\mathbf{p}_{\mathcal{A}_{(1,\ldots,K)}}$), and the agent can obtain absolute position measurements via e.g. an GNSS sensor. The green cell color indicates observable configurations, orange those where the global yaw orientation is unobservable, and red indicates that the global yaw orientation and absolute position is unobservable (see Section 7.5.3).

### Multiple Agents without Anchors

Second, we study the observability of multiple agents when they are beyond the anchors' sensing range (see Table 7.2). By obtaining fully meshed tag to tag (T-T) range measurements, a system of at least three agents (two agents leave one dimension ambiguity in the spheres' intersection) will encounter 9 unobservable dimensions in the full state space. Again, the estimation is lacking absolute information, spanning 9 dimension over all agents' position, velocity, and orientation in the global frame jointly observable. Remarkably, this means that the whole system knows all agents' relative position and orientation with respect to each other. Again, if one agent obtains absolute position measurements (e.g., provided by a GNSS sensor), the full state vector is rendered observable. The same holds for an estimation problem, if the range measurements are replaced by relative position measurements between the IMUs.

### Multiple Agents with Anchors

Finally, we study the case of multiple agents measuring ranges to multiple stationary anchors. Both agents and anchors optionally perform fully meshed range measurements among themselves, T-T and A-A, respectively. Further, agents potentially obtain range measurements between the anchors and the tags T-A (see Table 7.3).

If the anchors cannot communicate (exchange information) with the agents (no T-A),

| $M$ agents (Tag+IMU) | | | | |
|---|---|---|---|---|
| I-I | T-T | num. abs. position $N$ | | |
| (meshed) | (meshed) | 0 | 1 | 2 |
| 0 | 0 | $15M$ | $15(M-N)$ | $15(M-N)$ |
| 1 | 0 | 9 | 0 | 0 |
| 0 | 1 | 9 | 0 | 0 |
| 1 | 1 | 9 | 0 | 0 |

**Table 7.2:** Analysis of all unobservable dimensions spanned by $M$ agents with IMUs and ranging tags to observe each other via fully meshed inter-tag distance measurements (T-T) or via fully meshed relative IMU position measurements (I-I). Once more, agents can obtain absolute position measurements via, e.g., a GNSS sensor. A green colored cell indicates fully observable configurations, orange that the global position, velocity and orientation of all $M$ agents are jointly observable (span a 9 dimensional unobservable subspace), and red indicates that the whole state space is unobservable (see Section 7.5.3).

e.g., new anchors are out of range, but can use a bidirectional mesh with other anchors, one will find 15 unobservable dimensions in the state space. We encounter two jointly observable sub-spaces: (i) the 9 unobservable dimensions of the multiagent system that has no information from the anchors (see Section 7.5.3), and (ii) 6 additional unobservable dimension from the anchors' global pose (position and orientation). Providing at least one agent with absolute position information solves the first unobservable subspace. Similarly, fixing at least two anchors makes the second subspace observable. With no communication, both clusters are isolated, and providing absolute information to one cluster does not propagate it to the other cluster.

If T-A measurements are conducted without absolute information, by either fixing two anchors or providing absolute position measurement to one of the agents, four dimensions are unobservable as with a single agent configuration (see Section 7.5.3). Meaning that with at least three agents in the system, absolute position and yaw orientation of all agents and anchors are jointly observable. The limit of at least three agents comes from the multi-agent case without anchors, which, if violated, results in additional unobservable dimensions. The anchors in the system allow for the velocity and roll/pitch orientation to be observable. Again, fixing at least two anchors globally or supplying an agent with absolute position information, e.g., by a GNSS sensor, renders the full state observable.

### 7.5.4 Scenario $S_1$

For evaluating the estimator credibility based on an ANEES evaluation (see Section 2.6.3) and the computation time of individual estimator instances in a realistic scenario, we use all five *Machine Hall* sequences (MH_{01...05}) of the EuRoC dataset [20] in a Monte Carlo simulation with 20 runs[4], to emulate an indoor (GNSS-denied environment) inspection with five MAVs. A single simulation run takes 135 s, which is the longest flight time among the *Machine Hall* sequences between take-off and landing. Each MAV is equipped with an IMU, an emulated barometer, and an emulated ranging transceiver (tag), for both communication and pair-wise ranging between other ranging modules in communication range, using, e.g., DecaWave's[5] UWB ranging devices.

---

[4]Please note, that the IMU measurement noise, in contrast to other measurement noise, did not change between Monte Carlo simulation runs, as the real-world IMU measurements were directly used. This compromise was made to justify the applicability on MAVs.

| $M$ agents (Tag+IMU) and $K$ anchors | | | | | | |
|------|------|----------|------|------|------|------|
| A-A | T-A | num. abs. | T-T | num. fixed. anchors | | |
|     |     |          |     | 0 | 1 | 2 |
| 1 | 0 | 0 | 1 | 15 | 12 | 9 |
| 1 | 1 | 0 | 0 | 4 | 1 | 0 |
| 1 | 1 | 0 | 1 | 4 | 1 | 0 |
| 1 | 1 | 1 | 0 | 0 | 0 | 0 |

**Table 7.3:** Case study results of the unobservable dimensions of $M$ agents with IMUs and ranging tags measuring fully meshed unidirectional ranges to $K$ stationary anchors (T-A). The anchors can perform fully meshed measurements among themselves (A-A). If necessary, agents can obtain absolute position measurements via e.g. an GNSS sensor, perform fully meshed range measurements between the tags (T-T), and anchors can be set to be known globally. Green coloring indicates fully observable configurations, yellow an unobservable global yaw orientation, orange an unobservable global yaw orientation and position, and red indicates the two unobservable sub-spaces of agents and anchors (see Section 7.5.3).

Figure 7.5 depicts the spatial frame constellation. Twenty-five stationary ranging transceiver (anchors) are assumed to be deployed to cover the area of interest as shown in Figure 7.6 with a communication range of $4\,\mathrm{m}$. The communication range is on purpose short, to justify the deployment of 25 anchors and to challenge the estimation problem: (i) anchors and MAVs are meeting again, meaning that they are correlated and needs to be considered properly, (ii) some anchors have no direct link to (fixed) reference anchors, (iii) it increases the total amount of estimator instances $|\mathsf{S}|$ and handlers $|\mathsf{H_A}|$.

According to a nonlinear observability analysis assuming persistent communication and sensing capabilities, the absolute position of two anchors needs be known in order to render this nonlinear estimation problem fully observable. Since the system suffers from approximated models, linearization errors, multi-rate measurements, and limited communication range, we found 5 fixed anchors close to the take-off position of the MAVs are a good compromise regarding the convergence of the MAVs' estimates under realistic conditions. Still, the global full state is, due to limited communication and sensing range, not observable, in particular due anchor position estimates that are not overlapping with the sensing range of fixed anchors.

Since the ranging anchors are assumed to possess communication and computation capabilities, we model them as individual agents with a sensor estimator handler and one estimator instance. This results in 30 agents with in total 35 estimator instances: five agents each with an estimator instance for an IMU, barometer and ranging tag, and 25 anchors with a single estimator, while fixed sensor estimates are assumed to be constant and known, and therefore not considered in the estimation process. For this scenario, the system's global full state vector with 165 elements is

$$\mathbf{x}_{\mathsf{S}_1} = \left[ \{\mathbf{x}_{\mathcal{I}}; \mathbf{x}_{\mathcal{P}}; \mathbf{x}_{\mathcal{T}}\}_{\{1,2,3,4,5\}}; \mathbf{x}_{\mathcal{A}_{\{6,\dots,25\}}} \right]. \tag{7.2}$$

The ranging devices perform meshed and uni-directional measurements at a rate of $5\,\mathrm{Hz}$ with other devices in a range of $4\,\mathrm{m}$, with a standard deviation of $\sigma_\mathrm{r} = 0.1\,\mathrm{m}$, from $t = 0.3\,\mathrm{s}$ on, with a random message drop rate of $10\,\%$ (simulating, e.g., non-line-of-sight conditions), and a latency of $t_\mathsf{lat} = 0$.

The barometer provides pressure information at $20\,\mathrm{Hz}$, with a standard deviation of $\sigma_P = 1\,\mathrm{Pa}$ (a pressure difference of 1 Pa translates into a height difference of $\sim$8.43 cm in the standard atmosphere), a latency of $t_{\mathsf{lat}} = 0$, and a random message drop rate of 10 %.

---

[5]https://www.decawave.com/

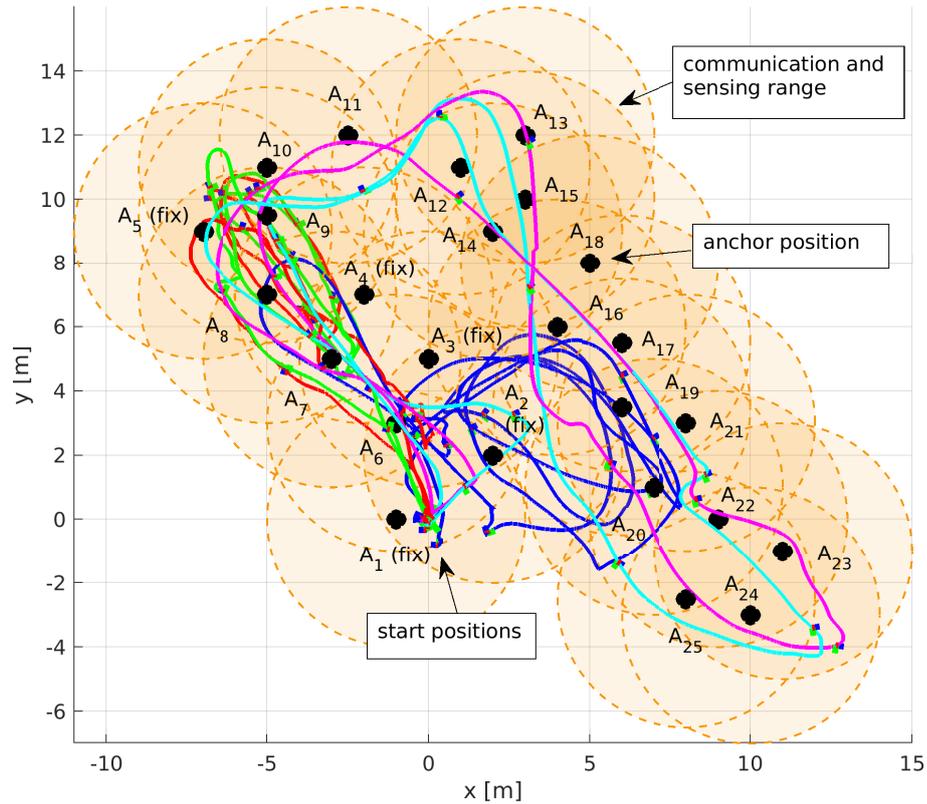

**Figure 7.6:** Scenario S$_1$: Top view on the anchor placement (bullets) and the trajectories of the *Machine Hall* sequences of the EuRoC dataset [20] in red, green, blue, cyan, and magenta. The filled orange circles show the communication range of 4 m per ranging anchor. Some anchors close to the starting position are assumed to be constant and known (*fix*), to define the navigation reference frame.

| | ${}^{\mathcal{G}}_{\mathcal{G}}\mathbf{p}_{\mathcal{A}}$ | ${}^{\mathcal{B}}_{\mathcal{B}}\mathbf{p}_{\{\mathcal{T},\mathcal{P}\}}$ | ${}^{\mathcal{G}}_{\mathcal{G}}\{\mathbf{p},\mathbf{v}\}_{\mathcal{I}}$ | ${}^{\mathcal{G}}\mathbf{q}_{\mathcal{I}}$ | ${}_{\mathcal{I}}\mathbf{b}_a$ | ${}_{\mathcal{I}}\mathbf{b}_\omega$ |
|---|---|---|---|---|---|---|
| $\boldsymbol{\sigma}^0$ | 1 cm | 1 cm | 0.1 {m, m/s} | 5 deg | 0.1 m/s$^2$ | 0.01 rad/s |

**Table 7.4:** Scenario S$_1$: Initial standard deviation of the IMU and sensor self-calibration states.

The IMU provides biased and noisy measurements at 200 Hz with a noise density of $\boldsymbol{\sigma}_a = 0.002\,\mathrm{m/s}^{1.5}$ and $\boldsymbol{\sigma}_\omega = 0.00017\,\mathrm{rad/s}^{0.5}$ for the accelerometer and gyroscope, respectively.

Note that in this constellation, relative range measurements between ranging modules always result in global joint observations, while local barometer readings result in local joint observations between the barometer and IMU. Range measurements between tags require four estimator instances, two relating to the IMUs and two relating to the ranging tags, and lead to an inter-agent quad-joint observation. Range measurements between tags and anchors result in an inter-agent triple-joint observations requiring an IMU, a tag, and an anchor.

Per simulation run, all states are initialized with a random offset around the true state based on the initial uncertainty defined in Table 7.4.

**Accuracy and Credibility**

In Tables 7.5 and 7.6, no remarkable differences in the ARMSE and ANEES between the DC-MMSF and MDCSE-DAH are noticeable, even though they perform joint observation on different sets of estimator instances. This indicates that DC-MMSF indeed performs quite close to the supposedly exact (centralized equivalent) MDCSE-DP* formulation in view of the credibility.

The ANEES of all states should be on average 3, with a lower and upper 99.7% credibility bound of $[1.63, 4.89]$ for 20 Monte Carlo runs (see Section 2.6.3). The self-calibration capabilities of the sensors' extrinsic states given the initial condition listed in Table 7.4 and noisy range measurements ($\sigma_r = 10\,\text{cm}$) can be seen in Table 7.6 (last three columns) for the beliefs $\mathbf{x}_\mathcal{P}$, $\mathbf{x}_\mathcal{T}$, and $\mathbf{x}_\mathcal{A}$. The average anchor positions ${}^\mathcal{G}_\mathcal{G}\bar{\mathbf{p}}_\mathcal{A}$ of the anchors $\mathcal{A}_{6-25}$ ($\mathsf{A}_{11-30}$) have an average error of 0.71 cm using DC-MMSF, although some anchors are not globally observable via range measurements from nearby anchors and agents, as no direct or indirect link[6] to the fixed anchors can be established. This means the estimators in these globally unobservable areas are slightly drifting, so that the tag and anchor position, were on average in this setup estimated less accurately using either strategy in contrast to the estimated anchor position of, e.g., agent $\mathsf{A}_{11}$ (anchor $\mathcal{A}_6$) Figure 7.7.

Figure 7.7 shows a subset of the global full state to indicate the performance of DC-MMSF in view of credibility. We chose to display the position of the fifth MAV's estimated position and gyroscope bias using DC-MMSF in the 20th run. We show the estimated anchor position of $\mathsf{A}_{11}$ (anchor $\mathcal{A}_6$) which near fixed anchors and of $\mathsf{A}_{28}$ (anchor $\mathcal{A}_{23}$) having no direct connections to fixed anchors. All estimates remain within the $3\sigma$, and thus, credible with a general tendency of slight underconfidence. Given the implicit maximum determinant completion in the IKF paradigm (described in Section 6.3.3), this result is expected. The estimated anchor position of $\mathsf{A}_{28}$ – due to a lacking observability – is not converging, in contrast to $\mathsf{A}_{11}$.

**Execution time**

Regarding the single threaded total execution time, DC-MMSF is $2.07\times$ faster than MDCSE-DAH as shown in Table 7.6. In the one hand MDCSE-DP*, which operates on the global full state (with 165-elements) in each filter step, leads to the smallest average trajectory error ($\sim$4% lower than our proposed DC-MMSF, while MDCSE-DAH and DC-MMSF performed similarly) as shown in Table 7.5. On the other hand, using MDCSE-DP* leads a $37.2\times$ higher computation time compared to DC-MMSF as shown in Table 7.6, assuming perfect communication between the agents and the centralized fusion entity. This renders the MDCSE-DP* approach impractical for complex swarms in real environments. Please note, that in a real application, the computation load would be distributed across multiple agents, which further reduces the total execution time for the distributed approaches, DC-MMSF and MDCSE-DAH.

In Figure 7.8, the execution time of different filter steps performed on agent $\mathsf{A}_5$ (one of the five moving MAVs) using DC-MMSF and MDCSE-DAH is shown. The average *propagation* time using DC-MMSF is $\sim$2.6$\times$ lower compared to MDCSE-DAH and highlights clearly the advantage of isolated propagation in estimator instances, over performing the propagation on the agent's full state in MDCSE-DAH. As expected, the speed-up using DC-MMSF over MDCSE-DAH for *isolated joint observation* in this particular configura-tion is small: Local joint observation are processed $\sim 1.68\times$ faster and inter-agent joint observation $\sim 1.53\times$ faster for agent $A_5$. The reason is that the stacked state vector size does not differ much in this particular experiment (therefore, similar execution times). An inter-agent (e.g., $A_i$ and $A_j$) range measurement between tags (T-T) using DC-MMSF consists of four estimator instances' states, two IMUs and two tags

$$\mathbf{x}_{\mathbf{p}} = \left[ \mathbf{x}_{\mathcal{I}_i}; \mathbf{x}_{\mathcal{T}_i}; \mathbf{x}_{\mathcal{I}_j}; \mathbf{x}_{\mathcal{T}_j} \right] \tag{7.3}$$

with 36 elements. Using MDCSE-DAH the calibration states of the barometer are included

$$\mathbf{x}_{\mathbf{p}} = \left[ \mathbf{x}_{\mathcal{I}_i}; \mathbf{x}_{\mathcal{T}_i}; \mathbf{x}_{\mathcal{P}_i}; \mathbf{x}_{\mathcal{I}_j}; \mathbf{x}_{\mathcal{T}_j}; \mathbf{x}_{\mathcal{P}_j} \right] \tag{7.4}$$

leading to 42 elements.

---

[6]An indirect link might be established if anchors or agents are still in the range of at least two fixed or fully observable anchors, thus their states are also observable. Therefore, these globally observable agent can propagate the global information to other anchors in their range, which constitutes a sort of global information relaying between agents or anchors.

| 20 MC runs | $[cm]$ | | $[cm/s]$ | | $[deg]$ | | $[cm/s^2]$ | | $[rad/s]$ | |
|---|---|---|---|---|---|---|---|---|---|---|
| Strategy | $^{\mathcal{G}}_{\mathcal{G}}\bar{\mathbf{p}}_{\mathcal{I}}$ | | $^{\mathcal{G}}_{\mathcal{G}}\bar{\mathbf{v}}_{\mathcal{G}}$ | | $^{\mathcal{G}}\bar{\mathbf{q}}_{\mathcal{I}}$ | | $_{\mathcal{I}}\bar{\mathbf{b}}_a$ | | $_{\mathcal{I}}\bar{\mathbf{b}}_\omega$ | |
| | AR | AN | AR | AN | AR | AN | AR | AN | AR | AN |
| DC-MMSF | 3.69 | 2.87 | 4.42 | 3.45 | 1.09 | **1.84** | **0.84** | 1.35 | 0.0018 | **1.42** |
| MDCSE-DAH | 3.69 | **2.92** | 4.42 | 3.45 | 1.09 | **1.84** | **0.84** | **1.36** | 0.0018 | **1.42** |
| MDCSE-DP* | **3.55** | 2.9 | **4.15** | **3.33** | **1.04** | 1.82 | 0.85 | 1.31 | **0.0016** | 1.31 |

**Table 7.5:** Scenario $S_1$: The average ARMSE and ANEES of the agents $A_{1-5}$ navigation states using different fusion strategies. Best values are in bold letters. Further details are given in Section 7.5.4.

| 20 MC runs | $[s]$ | $[cm]$ | | $[cm]$ | | $[cm]$ | |
|---|---|---|---|---|---|---|---|
| Strategy | $\bar{t}_{tot}$ | $^{\mathcal{B}}_{\mathcal{B}}\bar{\mathbf{p}}_{\mathcal{P}}$ | | $^{\mathcal{B}}_{\mathcal{B}}\bar{\mathbf{p}}_{\mathcal{T}}$ | | $^{\mathcal{G}}_{\mathcal{G}}\bar{\mathbf{p}}_{\mathcal{A}}$ | |
| | | AR | AN | AR | AN | AR | AN |
| DC-MMSF | **1575** | 0.62 | 1.94 | **0.79** | **12.3** | 0.71 | 8.34 |
| MDCSE-DAH | 3265 | 0.63 | **2.53** | **0.79** | 12.4 | 0.71 | **8.32** |
| MDCSE-DP* | 58647 | **0.58** | 2.32 | **0.79** | 13.6 | **0.62** | 8.62 |

**Table 7.6:** Scenario $S_1$: The average ARMSE and ANEES of the average over the anchor agents $A_{11-30}$ estimated states, and the average total execution time $\bar{t}_{tot}$ of all estimators using different fusion strategies. Best values are in bold letters. Further details are given in Section 7.5.4.

To demonstrate the effect of increasing the number of sensors per agent, we conduct a further experiment in Section 7.5.5 .

### 7.5.5 Scenario $S_2$

The aim of the second experiment $S_2$ is, to compare the execution time of the proposed approach against MDCSE-DAH.

In order to assess the effects of additional estimator instances and increasing sensor delays, we slightly modify the experiment defined in Section 7.5.4. No anchors are involved, meaning no tag to anchor (T-A) measurements are performed, as it would increase the total number of agents. As the agents would suffer from lacking global information, each agent obtains measurements from an absolute position sensor at a rate of 5 Hz without sensor delay. Further, the UWB range is increased to 100 m and the measurement drop

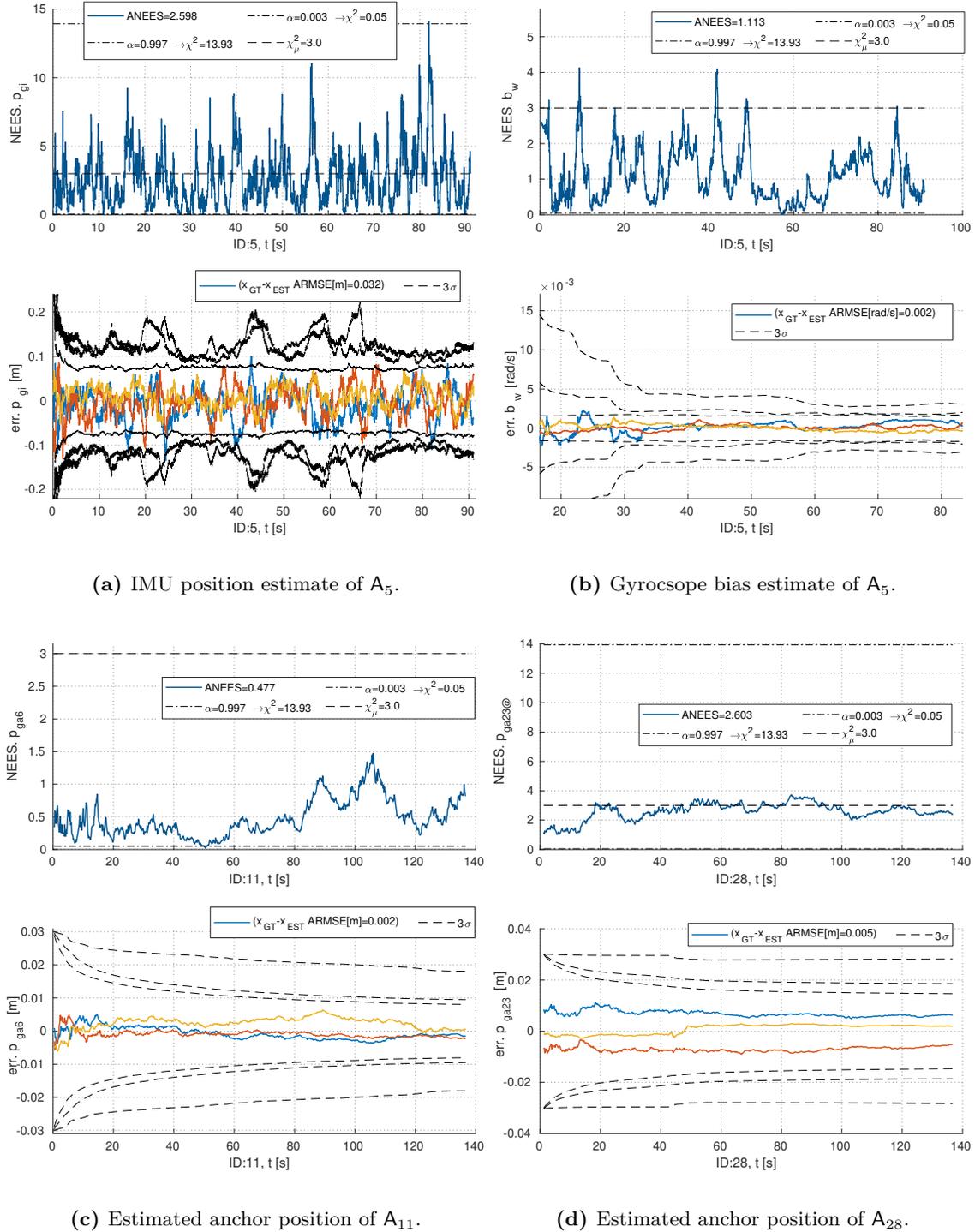

**(a)** IMU position estimate of $\mathsf{A}_5$.

**(b)** Gyroscope bias estimate of $\mathsf{A}_5$.

**(c)** Estimated anchor position of $\mathsf{A}_{11}$.

**(d)** Estimated anchor position of $\mathsf{A}_{28}$.

**Figure 7.7:** Scenario $\mathsf{S}_1$: The single run NEES (top row) and error (bottom row) of the estimated position of agent $\mathsf{A}_5$, the gyroscope bias of agent $\mathsf{A}_5$, and the anchor positions of agent $\mathsf{A}_{11}$ (sixth anchor) and $\mathsf{A}_{28}$ (23[rd] anchor) using DC-MMSF of the 20[th] simulation run. In yellow, blue, red for x, y, z axis, respectively. All estimation errors remain between their $3\sigma$ boundary (black dashed line). The experiment is described in Section 7.5.4.

rate is set to $0\,\%$, guaranteeing a persistent and fully meshed T-T measurements at $5\,\mathrm{Hz}$ and with no sensor delay modeled.

Each of the five agents has a variable number of barometers ($K = \{1, 3, 5\}$) to emulate the effect of additional sensors[7]. This increases the full state vector size per agent

$$\mathbf{x}_{\mathsf{A}} = \left[\mathbf{x}_{\mathcal{I}}; \mathbf{x}_{\mathcal{T}}; \mathbf{x}_{\mathrm{Abs}}; \mathbf{x}_{\mathcal{P}_{\{1,\ldots,K\}}}\right] \tag{7.5}$$

with the absolute position sensor' extrinsic calibration parameter $\mathbf{x}_{\mathrm{Abs}} = \left[{}^{\mathcal{B}}_{\mathcal{B}}\mathbf{p}_{\mathrm{Abs}}\right]$ and ${}^{\mathcal{B}}_{\mathcal{B}}\dot{\mathbf{p}}_{\mathrm{Abs}} = \mathbf{0}$.

For this scenario, the system's global full state vector varies with $K$ barometers per agent

$$\mathbf{x}_{\mathrm{S}_2} = \left[\mathbf{x}_{\mathcal{I}_{\{1,\ldots,5\}}}; \mathbf{x}_{\mathcal{T}_{\{1,\ldots,5\}}}; \mathbf{x}_{\mathrm{Abs}_{\{1,\ldots,5\}}}; \mathbf{x}_{\mathcal{P}_{\{1,\ldots,5\}\times K}}\right]. \tag{7.6}$$

The update rate of each barometer is lowered to $5\,\mathrm{Hz}$ with a drop rate of $0\,\%$ and the sensor latency is increased in steps ($t_{\mathsf{lat}} = \{0, 0.05, 0.1\}\,\mathrm{s}$), allowing us to show the impact on the average execution time on the individual filter steps by comparing the proposed DC-MMSF against the ported MDCSE-DAH approach.

Figure 7.9 shows clearly the advantage of performing IKF on a sensor level in our proposed DC-MMSF instead of performing it on the estimator instance handler $\mathsf{H}_{\mathsf{A}}$ level (MDCSE-DAH). First, it can be clearly seen that the average execution for MDCSE-DAH increases for the inter-agent joint observation with an increasing number of barometers (note that we added no sensor delay to T-T measurements to provoke out-of-order measurements). Second, an increased sensor delay has more impact on the execution time since more measurements – mostly IMU measurements – need to be recalculated due to the delay. In case of MDCSE-DAH, these measurements need to be applied on the agent's full state, whereas DC-MMSF performs them isolated. Theoretically, the average execution time using DC-MMSF should scale with $\mathcal{O}(1)$ being *agnostic* to the total amount of estimator instances in the system. The slight execution time increase is caused by memory access overheads, e.g., the buffer implementation of $\mathsf{Dict}$ does not ensure a constant memory access time, which is theoretically the case.

## 7.6  Conclusion

Our proposed Distributed Collaborative Modular Multi-Sensor Fusion (DC-MMSF) algorithm based on IKF leverages the seminal work in the domain of collaborative state estimation (e.g., CL) bringing them to maturity by supporting generic and changing sensor constellations, and by supporting delayed measurement. It breaks the classical full-state-per-agent paradigm up into single individual estimator instances in the swarm, blurring the lines between the strict distinction of off-board and on-board sensors. Our approach further exploits the fact that measurement sensitivity matrices in typical sensor fusion problems are sparse and that the state propagation can be performed independently. While correlations are efficiently maintained, communication is only required at the moment of interactions, and the correlations are restored at the moment they are needed again. This renders DC-MMSF as a fully scalable CSE approach with theoretically a constant complexity in communication and maintenance.

---

[7]Note that fusing multiple proprioceptive (propagation) sensors, e.g., an IMU is non-trivial as discussed in [36], since (i) relative pose constraints need to be defined in sensor updates, and (ii) the definition of the measurement Jacobian for sensor updates varies with the number of proprioceptive sensors. Supporting redundant proprioceptive sensors is important to mitigate a single point of failure, and we plan to address a generic solution to fuse multiple proprioceptive in future work.

The approach was validated on an indoor navigation scenario in a Monte-Carlo simulation based on a real dataset for MAVs with artificially augmented sensing modalities, manifesting that the proposed approach outperforms recent approaches, while achieving a similar estimation performance in terms of accuracy and credibility.

In future work, we will investigate on consistent and distributed cooperative multi-agent multi-target tracking as an extension to the proposed DC-MMSF algorithm. Furthermore, it would be interesting to integrate DC-MMSF in existing filter-based VIO algorithms.

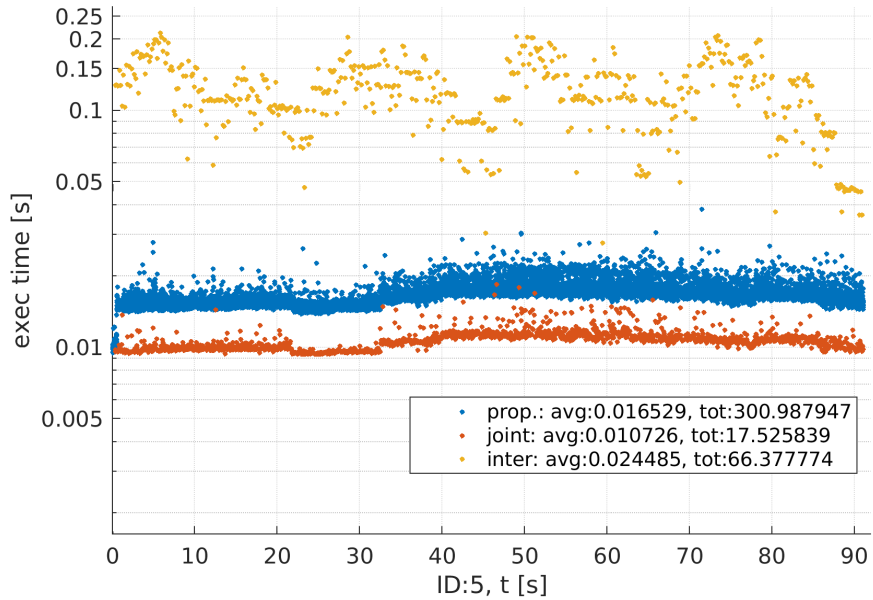

**(a)** MDCSE-DAH

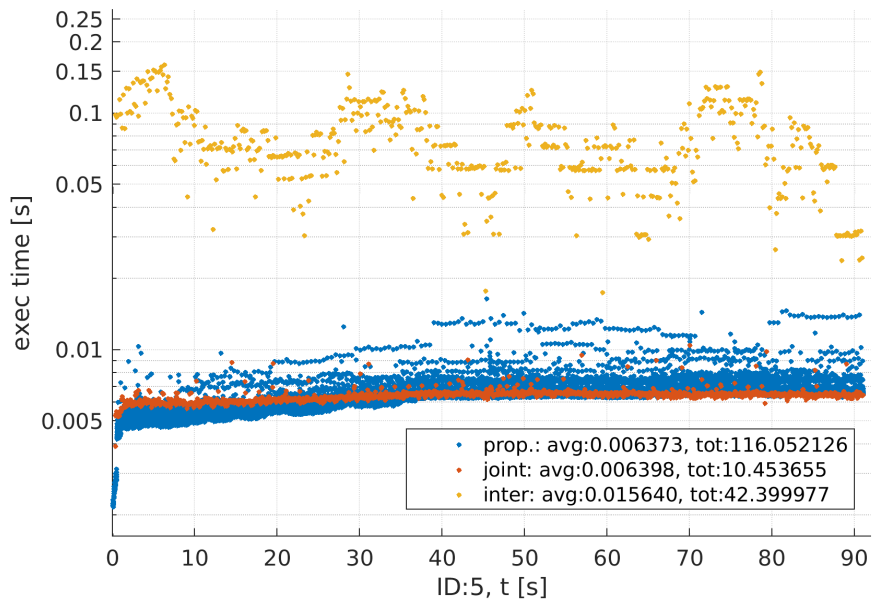

**(b)** DC-MMSF

**Figure 7.8:** Scenario $S_1$: Logarithmic duration of different filter steps on agent $A_5$ fusing different fusion strategies in the 20th Monte Carlo simulation run in the experiment described in Section 7.5.4: propagation (prop), isolated joint observation (joint), and inter-agent joint observations (inter). The corresponding entries in the legend show the average and the accumulated time of each type.

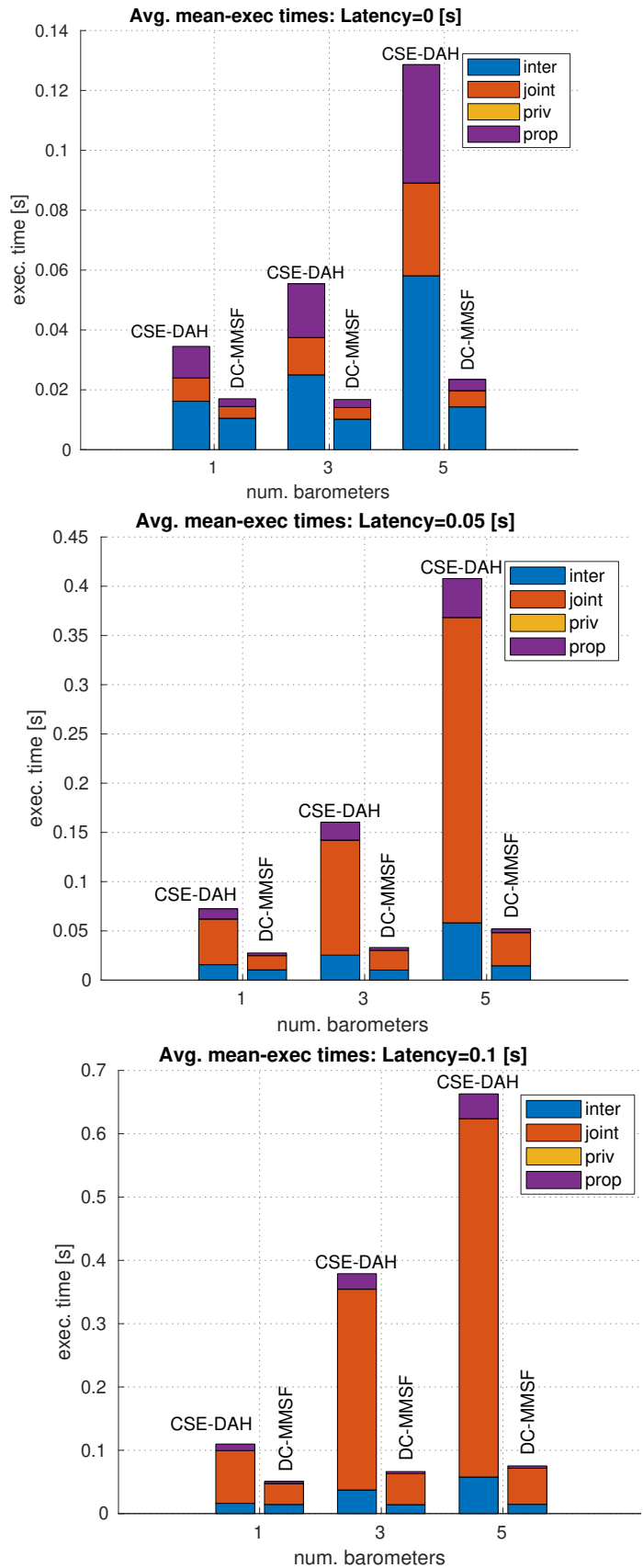

**Figure 7.9:** Scenario $S_2$: Shows the impact of both, (i) additional sensors and (ii) increasing sensor latency on the average execution time for different filter steps: propagation in purple (prop), private observations in yellow (priv), local joint observations in red (joint), and inter-agent joint observations in blue (inter). Please note the different scales in the y-axis and that no private sensor updates have been performed as described in Section 7.5.5.

# Chapter 8

# Conclusion and Outlook

In the preceding chapters, we have investigated decoupling strategies for Kalman filter formulations, in particular for addressing the key problems of scalability and handling delayed updates. In this chapter, we summarize our findings and provide an outlook on possible research directions.

## 8.1 Summary and Contributions

In Chapter 3, we aimed at generalizing filter-based Distribued CSE (DCSE). Therefore, we formulated the problem as a networked dynamic model with decoupled dynamics and sparsely coupled outputs. We carefully study the estimation problem on an example with three agents and propose a novel approximated DCSE algorithm denoted as DCSE-DAH which was published in [76]. We studied and compared different architectures and evaluated their performance with respect to scalability and accuracy in Monte Carlo simulations and on a real-world dataset for MAVs.

In Chapter 4, we generalize filter-based modular sensor fusion from a DCSE perspective, which allows us to bridge the gap between those two domains. We proposed a modular sensor framework denoted as MMSF-DAH, which lays the basis for the Isolated Kalman Filtering (IKF) paradigm and compared it against different decoupling approaches, that we have ported and adapted from DCSE. We evaluated their performance with respect to scalability and accuracy in Monte Carlo simulations and again on a real-world dataset for MAVs.

In Chapter 5, we revisited the infrastructure-based UWB ranging problem using the MMSF-DAH framework in order break the computational barrier and allows to efficiently use all range measurements in a fully meshed UWB ranging network. In extensive Monte Carlo simulations we studied the anchor self-calibration capabilities using the IKF paradigm. We extended a real-world dataset for MAVs with synthetics set of UWB ranging devices to demonstrate a real-world application.

In Chapter 6, we propose a new Kalman filter decoupling paradigm, denoted as Isolated Kalman Filtering (IKF), which is a generalization of our DCSE-DAH approach [76] and our MMSF-DAH approach [74] supporting delayed measurements. The paradigm is built upon approximations introduced by Luft *et al.* in [94], while we proof that the approximation for the cross-correlations in case of a joint measurement, was actually equivalent to a maximum determinant completion of the incomplete covariance matrix. We investigated on the filter credibility and steady-state performance in different observation graphs, and provide a reference implementation, in form of a generic C++ framework, to the public.

Finally, in Chapter 7, we unified DCSE and modular sensor fusion in the DC-MMSF framework, which again is build upon the IKF paradigm. This approach allowed used to address efficiently the issue of a time-varying sensor configuration per agents and to support asynchronous measurements. The approach was evaluated on a multi-agent indoor navigation scenario using a real-world dataset for MAVs with a synthetics set of UWB ranging devices.

By concluding this work, we are exited to propose a novel decoupling scheme for Kalman filters and hope that our findings and our view on modular estimation frameworks help the robotics community to address existing issues and to realize more generic estimation frameworks.

## 8.2 Limitation and Future Research

In order to become an integral part for commercial and research applications, exhaustive real-world experiments need to be conducted and parts of the framework need to be open-sourced. Especially navigating in closed-loop control, would reveal important aspects, such as the general applicability using commercial off-the-shelf components. A disadvantage of real-world experiments in the early stages of development is, that systematical errors in the problem formulation might be shadowed by the experimental setup.

Nevertheless, the datasets used in our evaluations are accredited in the robotics community and commonly used as benchmark for VINS. Our models and kinematics are applied in other estimation frameworks such as in MaRS [18] by Brommer *et al.* . A filter consistency and credibility analysis can only be performed in a controlled simulation environment by performing Monte Carlo simulation runs, see Section 2.6.3. Scalability analyses with parameter sweeps regarding, e.g., the number of agents, sensors, or sensor delay become quickly infeasible in real-world experiments, e.g., due to space and hardware constraints. Consequently, most of our evaluations performed can only be addressed in simulation to guarantee a fair comparison, deterministic behavior, and a reproducability of the results. The key challenge towards real-world experiments using MAVs remains in the vast engineering effort and in the complexity of designing a distributed and scalable estimation framework.

In addition to the limitations in the above discussion, several research questions can be defined. In context of modular and decoupled sensor fusion, the applicability of reference sensors for differential sensor fusion would be interesting. For instance, a local pressure sensor, magnetometers, or sun (star) sensors could be treated as additional sensor in the MMSF-DAH formulation, to compensate local disturbances or allow for relative height or heading estimation in an AINS. In that direction, an extension for target-tracking while performing localization is an integrable part of a generalized collaborative estimation framework. Targets can be handled as additional IKF nodes in the MMSF-DAH formulation, with target specific dynamic models. In the same way, stationary landmarks (persistent features) could be integrated using dedicated IKF nodes. The challenge regarding a collaborative target-tracking remains, apart form the unique (target) association, in the track-to-track fusion of the distributed estimates, while accounting for the correlations between other (decoupled) estimates efficiently. An extension for collaborative target tracking in the proposed DC-MMSF framework might be rather trivial. Agents could, e.g., exchange local beliefs of tracked targets, once they are in communication range and if the beliefs were changed meanwhile. Then they need to find a consensus about their beliefs, e.g., based on covariance intersection, and apply a correction term in their correction buffer to account for the correlation between other local beliefs. The same strategy could be applied for estimated landmark positions or other reference parameters.

Another integrable part of nowadays AINS is to fuse visual-, thermal-, radar-, or LIDAR-based information. An efficient and de facto gold standard in filter-based approaches is to formulate a MSCKF, which is augmenting the local belief by stochastic clones, that allows to formulate constraints relating to beliefs at different time instances.

Stochastic clones can be – again – handled as additional IKF nodes in the MMSF-DAH formulation and feature observations can be rendered as local joint updates incorporating a set of IKF nodes. Persistent features can be maintained natively in the MMSF-DAH formulation, as mentioned before. An investigation on the efficiency and introduced overhead of a truly-modular MSCKF in contrast to an application specific (tailored) MSCKF formulation would be interesting and if the decoupling through the IKF paradigm could compensate for the introduced overhead in the data management within the MMSF-DAH approach.

With respect to the IKF paradigm, we suggest investigating further in the theoretical analysis on the observability properties of a decoupled full-state. In particular, Conjecture 1 is lacking a theoretical proof. Our evaluation in Section 5.3 and  Section 6.4 gave some insights to that problem, but details on the convergence behavior of the decoupled state and a theoretical observability analysis on the linearized and discretized system are missing.

Regarding observability and sensor self-calibration in a modular estimator, a two-stage approach that allows a seamless switching between a centralized-equivalent and an isolated/decoupled formulation, would combine the advantages of both strategies – accuracy and efficiency. The switching could be, e.g., motion dependent, since the observability of an AINS' states is typically motion related.

Currently under investigation is an application for autonomous bridge inspection using a collaborating and communicating swarm of MAVs in partially GNSS denied areas in order to obtain a digital twin of the infrastructure and to revisit points of interests. Individual agents will rely on a VINS with a GNSS receiver and a ranging device, to sense distances to other agents and stationary ranging anchors. In the foreseen demonstration, agents will form a relative constellation that allows for a more accurate and robust localization in GNSS-denied areas (e.g., below the bridge), by incorporating relative range measurements - which allows, in some sense, for *sensor sharing* or i.e. to relay absolute position information to other agents. A similar constellation was evaluated in Section 7.5.5. The optimal relative formation can be determined by maximizing the Fisher information, as proposed by Cossette *et al.* in [28], and would allow for collision avoidance between agents and with obstacles.

# Acronyms

**AINS** Aided Inertial Navigation System
**ANEES** Average Normalized Estimation Error Squared
**ARMSE** Average Root Mean Square Error
**ATE** Absolute Trajectory Error
**AUV** Autonomous Underwater Vehicle
**BLUE** Best Linear Unbiased Estimator
**CCSE** Centralized CSE
**CCSE-DP** CCSE with Distributed Propagation
**CI** Covariance Intersection
**CL** Collaborative Localization
**CLATT** Collaborative Localization and Target Tracking
**CNN** Convolutional Neural Network
**CSE** Collaborative State Estimation
**CSLAM** Collaborative SLAM
**DACC** Distributed Approximated Cross-Covariance
**DAH** Distributed Approximated History
**DAH-CSE** DAH Collaborative State Estimation
**DC-MMSF** Distributed Collaborative Modular Multi-Sensor Fusion
**DCL** Distributed Collaborative Localization
**DCSE** Distribued CSE
**DCSE-DACC** DCSE based on Distributed Approximated Cross-Covariances
**DCSE-DAH** DCSE based on a Distributed Approximated History
**DCSE-DP** DCSE with Distributed Propagation
**DCSE-DP\*** Decentralized CSE with Distributed Propagation
**DDMV** Distributed Discorrelated Minimum Variance
**DMV** Discorrelated Minimum Variance
**DoF** Degrees of Freedom
**DP** Distributed Propagation
**EKF** Extended Kalman Filter
**ESEKF** Error-State EKF
**FC** Fusion Center
**GNSS** Global Navigation Satellite System
**GPS** Global Positioning System
**HiPR** High Precision Ranging
**ID** identifier
**IKF** Isolated Kalman Filtering

**IMU** Inertial Measurement Unit

**INS** Inertial Navigation System

**KF** Kalman Filter

**LIDAR** Laser Imaging Detection and Ranging

**LLS** Linear Least Squares

**MAP** Maximum A Posteriori

**MaRS** Modular and Robust Sensor-Fusion

**MAV** Micro Aerial Vehicle

**MEMS** Micro Electro Mechanical Systems

**MLE** Maximum Likelihood Estimation

**MMSF** Modular Multi-Sensor Fusion

**MMSF-C** Modular Multi-Sensor Fusion Centralized equivalent

**MMSF-DACC** Modular Multi-Sensor Fusion Decoupled Approximated Cross-Covariances

**MMSF-DAH** Modular Multi-Sensor Fusion Decoupled Approximated History

**MMSF-DP** Modular Multi-Sensor Fusion Decoupled Propagation

**MMSF-MaRS** Modular Multi-Sensor Fusion based on MaRS

**NEES** Mean Normalized Estimation Error Squared

**MSCKF** Multi-State Contraint Kalman Filter

**NEES** Normalized Estimation Error Squared

**NIS** Normalized Innovation Squared

**NLOS** non-line-of-sight

**NTP** Network Time Protocol

**PDOP** positional dilution of precision

**RIO** Radar-Inertial Odometry

**RMSE** Root Mean Square Error

**ROS** Robot Operating System

**RPE** Relative Pose Error

**RSS** received signal strength

**RTOF** return time of flight

**SCIF** Split-Covariance Intersection Filter

**SKF** Schmidt-Kalman Filter

**SLAM** Simultaneous Localization and Mapping

**SVD** singular value decomposition

**TDOA** time difference of arrival

**TOA** time of arrival

**UAV** Unmanned Aerial Vehicle

**UKF** Unscented Kalman Filter

**UWB** ultra-wideband

**VINS** Visual-Inertial Navigation System

**VIO** Visual-Inertial Odometry

**DDMV** Distributed Discorrelated Minimum Variance

# References

## Online sources